\documentclass{article}

     \PassOptionsToPackage{numbers, compress}{natbib}
\usepackage[preprint]{neurips_2026}

\usepackage[utf8]{inputenc} % allow utf-8 input
\usepackage[T1]{fontenc}    % use 8-bit T1 fonts
\usepackage[hidelinks]{hyperref}       % hyperlinks
\usepackage{url}            % simple URL typesetting
\usepackage{booktabs}       % professional-quality tables
\usepackage{amsfonts}       % blackboard math symbols
\usepackage{nicefrac}       % compact symbols for 1/2, etc.
\usepackage{microtype}      % microtypography
\usepackage[x11names]{xcolor}         % colors
\usepackage[export]{adjustbox} % for the grid of qualitative results

\definecolor{TabBlue}{rgb}{0.122, 0.467, 0.706}
\definecolor{TabGray}{rgb}{0.498, 0.498, 0.498}
\definecolor{TabPink}{rgb}{0.890, 0.467, 0.761}
\definecolor{TabBrown}{rgb}{0.549, 0.337, 0.294}
\definecolor{TabPurple}{rgb}{0.580, 0.404, 0.741}
\definecolor{TabGreen}{rgb}{0.173, 0.627, 0.173}
\definecolor{TabOrange}{rgb}{1.000, 0.498, 0.055}
\definecolor{TabDarkRed}{rgb}{0.550, 0.060, 0.060}
\definecolor{TabCyan}{rgb}{0.000, 0.749, 0.749}
\usepackage{graphicx}
\usepackage{caption}
\usepackage{tikz}
\usepackage{pgfplots}  
\usepackage{twemojis}
\usepackage{siunitx}  % for \SI command
\usepackage{wrapfig}
\pgfplotsset{compat=1.18}   % Match your PGFPlots version

\title{Point Cloud Sequence Encoding for Material-conditioned Graph Network Simulators}

\usepackage[acronym,nohypertypes={acronym,notation}]{glossaries}
\newcommand{\glsf}[1]{\glsreset{#1}\gls{#1}}

\newacronym{gns}{GNS}{Graph Network Simulator}
\newacronym{mgn}{MGN}{MeshGraphNet}
\newacronym{mpc}{MPC}{Model Predictive Control}
\newacronym{mbrl}{MBRL}{Model-Based Reinforcement Learning}
\newacronym{gnn}{GNN}{Graph Neural Network}
\newacronym{sofa}{SOFA}{Simulation Open Framework Architecture}
\newacronym{mpn}{MPN}{Message Passing Network}
\newacronym{mlp}{MLP}{Multilayer Perceptron}
\newacronym{cnn}{CNN}{Convolutional Neural Network}
\newacronym{mse}{MSE}{Mean Squared Error}
\newacronym{iou}{IoU}{Intersection over Union}
\newacronym{ggns}{GGNS}{Grounding Graph Network Simulator}
\newacronym{amber}{\textit{AMBER}}{Adaptive Meshing By Expert Reconstruction}
\newacronym{gcn}{GCN}{Graph Convolutional Network}
\newacronym{pde}{PDE}{Partial Differential Equation}
\newacronym{asmr}{\textit{ASMR}}{Adaptive Swarm Mesh Refinement}
\newacronym{gmm}{GMM}{Gaussian Mixture Model}
\newacronym{mp}{MP}{Movement Primitive}
\newacronym{fem}{FEM}{Finite Element Method}
\newacronym{amr}{AMR}{Adaptive Mesh Refinement}
\newacronym{amg}{AMG}{Adaptive Mesh Generation}
\newacronym{ml}{ML}{Machine Learning}
\newacronym{rl}{RL}{Reinforcement Learning}
\newacronym{il}{IL}{Imitation Learning}
\newacronym{dcd}{DCD}{Density-Aware Chamfer Distance}
\newacronym{unet}{U-Net}{U-Net}
\newacronym{cnp}{CNP}{Conditional Neural Process}
\newacronym{gno}{GNO}{Graph Neural Operator}
\newacronym{fps}{FPS}{Farthest Point Sampling}
\newacronym{knn}{KNN}{K-Nearest Neighbors}
\newacronym{sdf}{SDF}{Signed Distance Field}

\newacronym{mango}{\textit{MaNGO}}{Meta Neural Graph Operator}
\newacronym{pstnet}{\textit{PSTNet}}{Point Spatio-Temporal Network}
\newacronym{peach}{\textit{PEACH}}{Point Cloud Encoding for Accurate Context Handling}

\author{
  \textbf{Philipp Dahlinger}$^1$\thanks{Correspondence to \texttt{philipp.dahlinger@kit.edu}} %
  \And Balázs Gyenes$^1$ %
  \And Niklas Freymuth$^1$ %
  \And Luca Geminiani$^2$ %
  \And Tobias Würth$^3$ %
  \And Johannes Mitsch$^3$ %
  \And Nadja Klein$^2$ %
  \And Luise Kärger$^3$ %
  \And Gerhard Neumann$^1$
  \AND
  \normalfont $^1$Autonomous Learning Robots, $^2$Methods for Big Data, $^3$Institute of Vehicle System Technology \\
  Karlsruhe Institute of Technology, Karlsruhe
}
\usepackage{amsmath,amsfonts,bm}

\def\eqref#1{equation~\ref{#1}}
\def\1{\bm{1}}

\DeclareMathAlphabet{\mathsfit}{\encodingdefault}{\sfdefault}{m}{sl}
\SetMathAlphabet{\mathsfit}{bold}{\encodingdefault}{\sfdefault}{bx}{n}

\begin{document}

\maketitle

\begin{abstract}
Graph Network Simulators (GNSs) have emerged as powerful surrogates for complex physics-based simulation, offering inherent differentiability and orders-of-magnitude speedups over traditional solvers. 
However, GNSs typically assume access to the underlying material parameters, such as stiffness or viscosity, 
severely limiting their utility in realistic experimental settings.
While recent meta-learning approaches address the parameter dependency by inferring properties from mesh trajectories, reconstructing a mesh from an observed scene is challenging.
In this work, we introduce \glsf{peach}, a novel framework that applies in-context learning on point clouds to adapt a learned simulator to unseen physical properties during inference.
Our approach relies on a novel spatio-temporal point cloud sequence encoder, as well as two forms of auxiliary supervision to help improve simulation fidelity.
We demonstrate that \gls{peach} is capable of accurate zero-shot sim-to-real transfer on a challenging, dynamic scene.
Experiments on simulation scenes show that \gls{peach} even outperforms mesh-based baselines on prediction accuracy, while being much more practical for real-world deployment.

\end{abstract}

\section{Introduction}
\label{sec:introduction}

\glsresetall

% \textbf{What is the problem? Why graph network simulators?}
Physics simulation is central to engineering disciplines ranging from structural mechanics to robotics, yet classical solvers based on the finite element method are computationally expensive \citep{zienkiewicz2005finite,stanova2015finite}.
\glspl{gns} have emerged as fast and differentiable surrogates for mesh-based physics, offering orders-of-magnitude speedups over classical solvers~\citep{battaglia2016interaction, pfaff2020learning, brandstetter2021message, linkerhagner2023grounding, wurth2025diffusion}.
However, standard \glspl{gns} assume known process conditions, resulting in a parameter estimation problem when physical parameters change.
Classical system identification can recover such parameters, but doing so per-instance multiplies the experimental and computational cost in ways that scale poorly~\citep{aastrom1971system, lennart1999system, isakov2006inverse, wu2015galileo, murthy2020gradsim, antonova2023rethinking}.
Test-time gradient inversion through a differentiable simulator is one response, but reintroduces a per-instance optimization loop and there is no guarantee that the loss objective for inversion is well-conditioned~\citep{qzhao2022graphpde, wang2024latent}.
In-context conditioning instead amortizes identification by encoding a set of context trajectories into a latent material descriptor, enabling rapid adaptation to new material configurations at inference time~\citep{dahlinger2025mango}.
In this work,  we formulate adaptive physical simulation as a real-to-sim in-context learning problem. Using a novel spatio-temporal encoder architecture proposal combines in-context conditioning for rapid adaptation with prediction of unknown physical parameters in a latent space  to encode a more expressive and informative context as input for the simulator. 
%Predicting the unknown physical parameters in a latent space also allows the encoder to pass a more expressive and informative context to the simulator, without creating a bottleneck.
%Therefore, we formulate adaptive physical simulation as a real-to-sim in-context learning problem.

% \textbf{Why point clouds? What are the benefits and challenges?}
Operating on meshes instead of points~\cite{robocook23, hd-vpd} helps \glspl{gns} generate stable and more accurate predictions over long sequences.
However, mesh trajectories of arbitrary deforming objects are not directly observable with standard sensing; reconstructing them requires either invasive marker setups or template registration that itself depends on geometric and constitutive priors~\citep{MansourZK24,LoperMB14}.
Point clouds, in contrast, are produced directly by depth cameras and LiDAR and capture 3D geometry of deforming objects.
Because they rely on geometric structure rather than surface appearance, sim-to-real transfer is less constrained by rendering fidelity~\citep{dexpoint, 3DDA, adapt3r, pointmappolicy}.
While point cloud observations can easily be generated from meshes, point clouds from the real world do not contain correspondences between points from subsequent time steps.
This complicates the estimation of physical parameters, since the motion of individual points contains the most information about those parameters.
Unlike in mesh-based approaches, these correspondences must be inferred indirectly.

\begin{figure}[t]
    \centering
    \includegraphics[width=\textwidth]{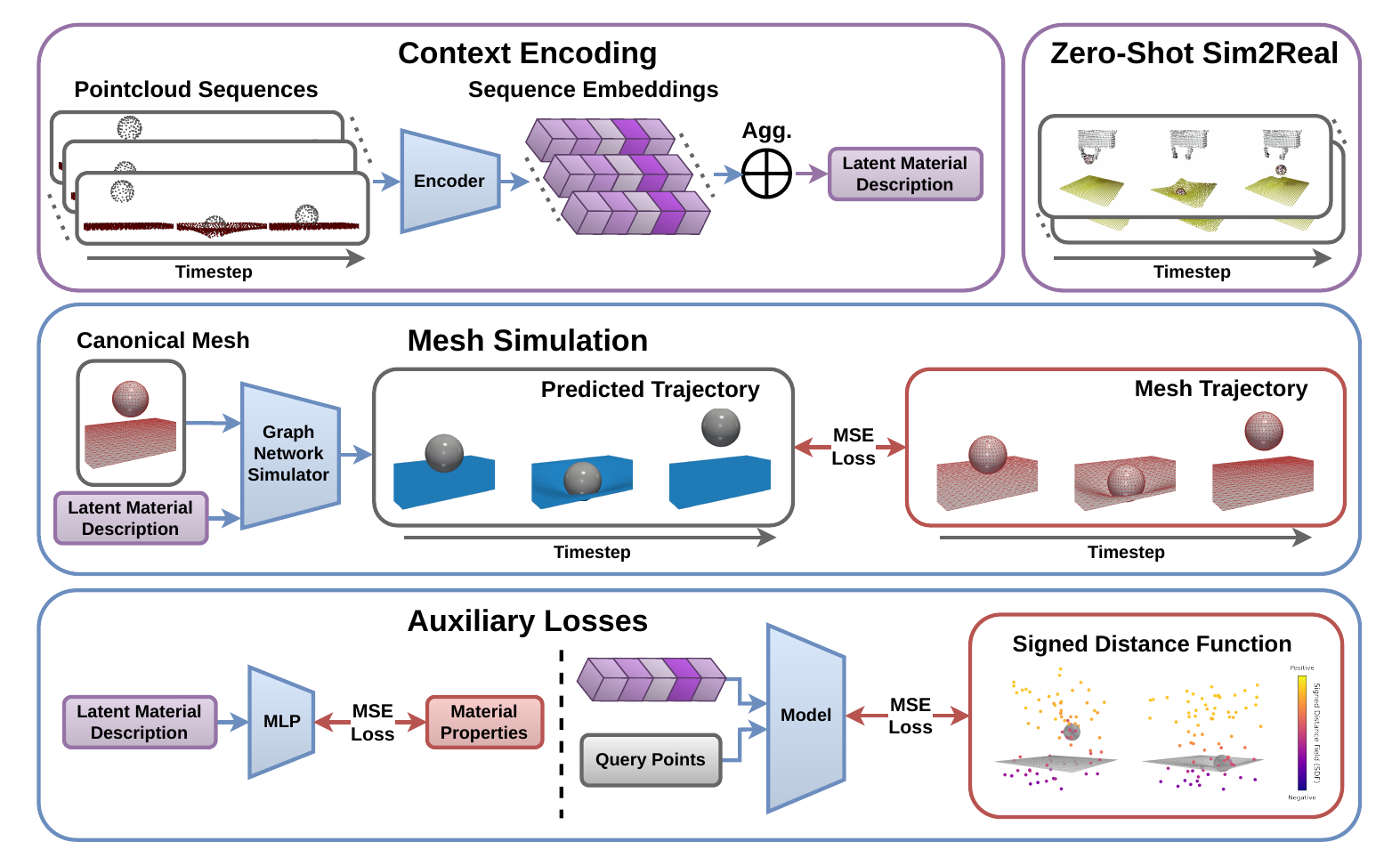}
    \caption{
    Overview of the \acrfull{peach} framework.
    Point cloud sequences from several simulated or real trajectories are encoded into a single latent material description via our novel spatio-temporal encoder architecture.
    The aggregated latent vector conditions a mesh-based simulator, which predicts complete trajectories from an initial mesh.
    % The method is trained to minimize the average simulation error, and uses an auxiliary loss on the latent material description to ensure physical consistency.
    To improve accuracy, we apply auxiliary losses based on the ground truth physical properties and the Signed Distance Field (SDF).
    % It additionally to predict full mesh dynamics, and is also used to predict material properties via an auxiliary MLP; both predictions are supervised during training. 
    During inference, only the context point cloud sequences and an initial geometry are needed to simulate behavior with novel, unseen material properties.
    }
    \vspace{-0.6cm}
    \label{fig:overview}
\end{figure}

% \textbf{Our method}
To address these challenges, we introduce \glsf{peach}\footnote{Code and videos of rollouts are provided in the supplementary material. Datasets will be released upon acceptance.}, an in-context framework that conditions a graph network simulator on a small set of observed point cloud sequences.
This framework is able to predict mesh-level dynamics for unseen materials, without any test-time parameter optimization.
\gls{peach} extends the conditional neural process framework \citep{garnelo_conditional_2018} to spatio-temporal observations: a spatio-temporal encoder maps each point cloud sequence to a latent vector, and the resulting set is aggregated into a single material descriptor that conditions the simulator.
Our novel encoder architecture extends the common point-patch paradigm~\cite{pointMAE, pointBERT, pointGPT} by treating a sequence as a single point cloud embedded in 4D space-time.
By relying on proximity in 4D space-time, the network can extract physical properties from the observation without explicit point-to-point correspondences.
% We pair this encoder with a trajectory-level simulator that predicts entire rollouts in a single forward pass, avoiding the train-test mismatch of autoregressive methods, which are trained on single-step predictions but deployed over long horizons, leading to compounding errors~\citep{egno2024, dahlinger2025mango}.
% Training the encoder–simulator stack end-to-end additional supervision: we found in preliminary experiments that the latent material descriptor is otherwise under-constrained \todo{forward-ref the ablation}.
While this architecture is already competitive, we found in preliminary experiments that additional auxiliary losses can further improve the results.
We investigate two auxiliary losses applied during training only: a regression head that grounds the latent embedding in the ground truth physical parameters, and a \gls{sdf} reconstruction loss that encourages the encoder to retain fine-grained geometric information in the earlier layers.
%prevents the encoder from collapsing context information into a representation that fits the regression target but discards scene-level signal.
For inference, the model receives an initial mesh of the canonical geometry, which is typically available from CAD or a one-time scan, and $1$ to $8$ point cloud sequences from the target process. \gls{peach} predicts the remaining mesh simulation using the implicit physical properties defined by the context.

To summarize, our contributions are as follows:
(i) We propose \gls{peach}, an end-to-end architecture that efficiently and accurately simulates a mesh-based trajectory conditioned on a handful of observed point clouds, without test-time optimization.
As a  key benefit, our novel spatio-temporal point cloud encoder is applicable to any operation on point cloud sequences due to its generality.
(ii) We demonstrate \gls{peach}'s sim-to-real capability through one-shot, rapid adaptation of a real-world scene in which a robot drops a ball onto a rubber trampoline, a setting beyond the reach of mesh-based methods.
We additionally validate on four simulation scenes featuring deformable objects and outperform baselines with access to the ground truth context mesh.
(iii) Through parameter studies, we provide insights that underscore the necessity of auxiliary losses for parameter regression and \gls{sdf}.
Furthermore, we visualize the latent space to show that it summarizes well  the underlying physical properties of the scene.

\begin{figure}[t]
    \centering
    \begin{minipage}{0.2\textwidth}
        \centering
        \includegraphics[width=\textwidth]{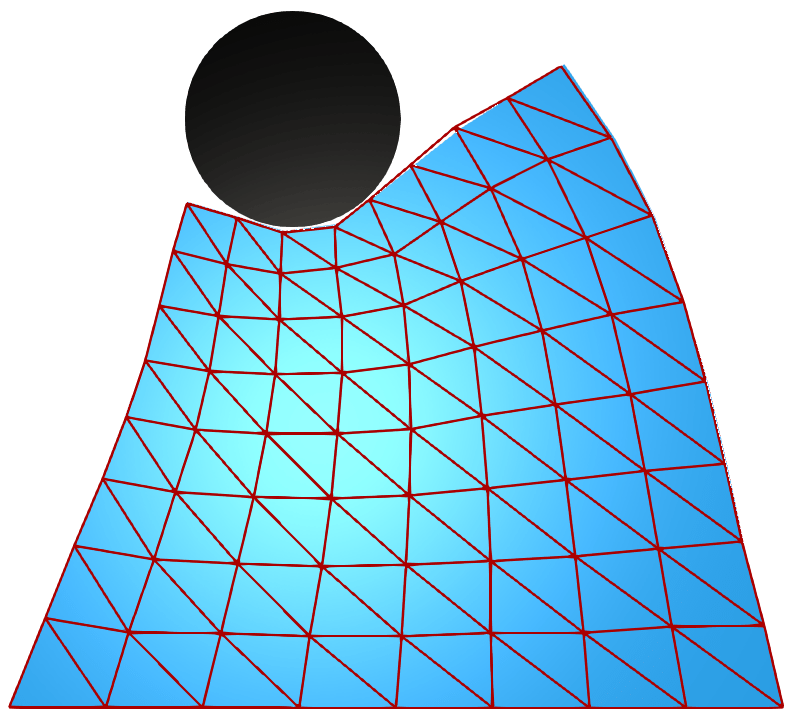}
    \end{minipage}
    \hspace{1mm}
    \begin{minipage}{0.29\textwidth}
        \centering
        \includegraphics[width=\textwidth]{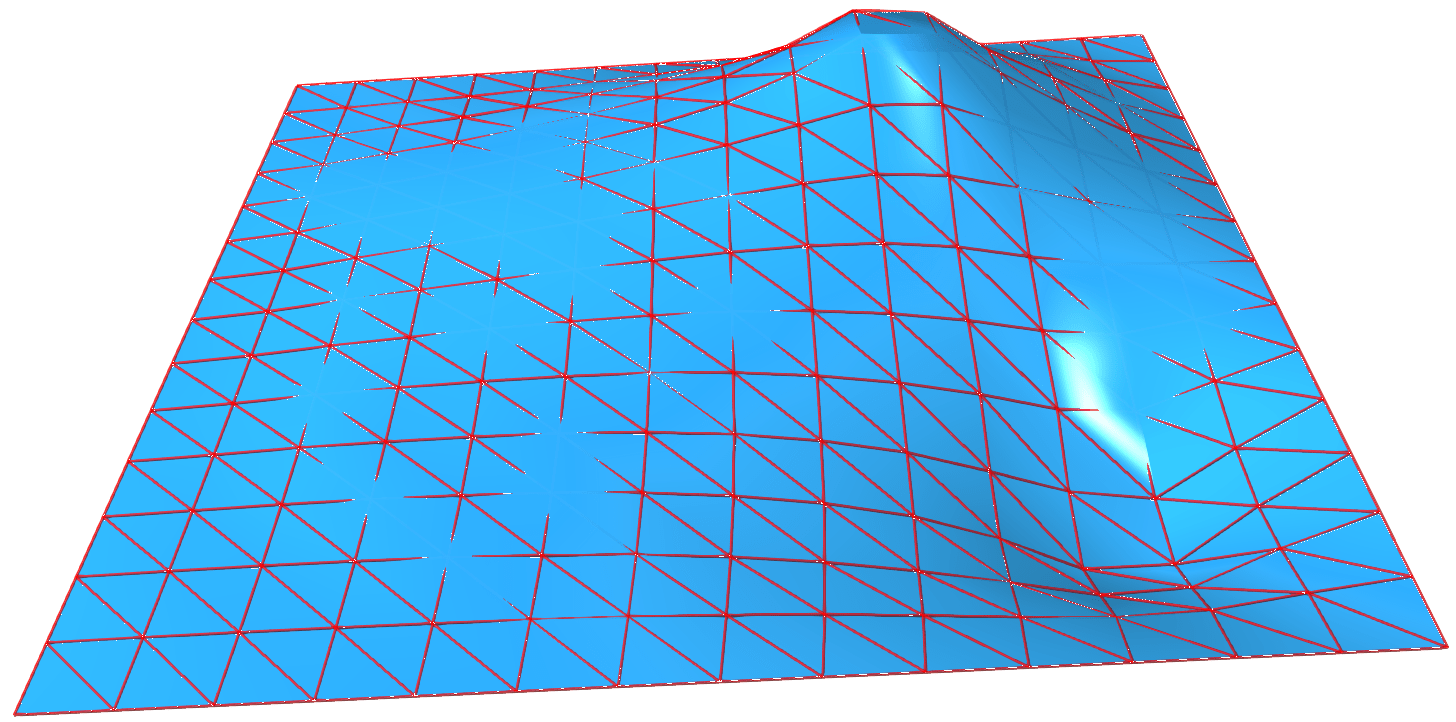}
    \end{minipage}
    \begin{minipage}{0.48\textwidth}
        \centering
        \includegraphics[width=\textwidth]{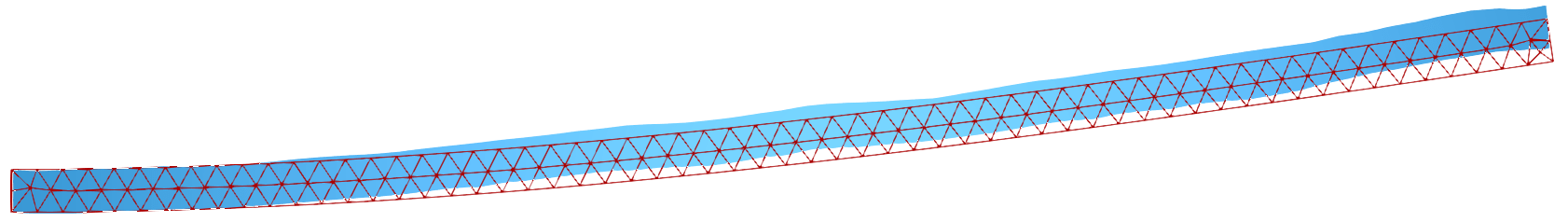}
    \end{minipage}%
\vspace{0.05mm}

    \begin{minipage}{0.2\textwidth}
        \centering
        {\small \texttt{Deforming Block}}
    \end{minipage}
    \begin{minipage}{0.29\textwidth}
        \centering
        {\small \texttt{Sheet Deformation}}
    \end{minipage}
    \begin{minipage}{0.48\textwidth}
        \centering
        {\small \texttt{Bending Beam}}
    \end{minipage}%
\vspace{-5mm}

    \begin{minipage}{\textwidth}
        \centering
        \includegraphics[width=\textwidth, trim={50 0 0 0}, clip]{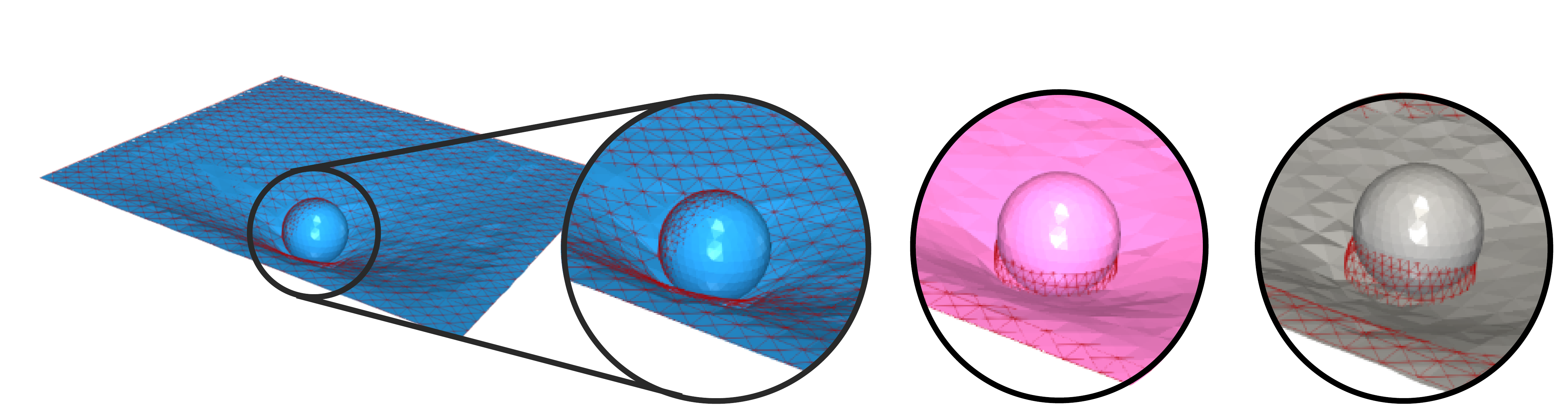}
    \end{minipage}

    % \begin{minipage}{0.325\textwidth}
    %     \centering
    %     \includegraphics[width=\textwidth]{figures/task_overview_data/trampoline_peach.png}
    % \end{minipage}
    % \begin{minipage}{0.325\textwidth}
    %     \centering
    %     \includegraphics[width=\textwidth]{figures/task_overview_data/trampoline_pstnet.png}
    % \end{minipage}
    % \begin{minipage}{0.325\textwidth}
    %     \centering
    %     \includegraphics[width=\textwidth]{figures/task_overview_data/trampoline_dummy.png}
    % \end{minipage}%
    % \vspace{0.05mm}
    
    \begin{minipage}{0.3\textwidth}
        \centering
        {\small \texttt{Trampoline }}
    \end{minipage}
    \begin{minipage}{0.2\textwidth}
        \centering
        {\small \hspace{0.6cm} \textit{PEACH}}
    \end{minipage}
    \begin{minipage}{0.23\textwidth}
        \centering
        {\small \hspace{0.8cm}\textit{PSTNet Encoder}}
    \end{minipage}
    \begin{minipage}{0.23\textwidth}
        \centering
        {\small \hspace{0.5cm} \textit{No Context}}
    \end{minipage}

\caption{
    Overview of the four simulation scenes.
    The environments cover a range of deformable object types and material models, from purely elastic to viscoelastic constitutive behaviors, with varying material properties across tasks.
    Qualitative predictions of \gls{peach} and selected baselines are shown in \textcolor{blue}{blue}, \textcolor{TabPink}{pink}, and \textcolor{gray}{gray}, respectively, with ground-truth meshes overlaid in \textcolor{red}{red}.
}
\vspace{-0.6cm}
\label{fig:task_overview}
\end{figure}

\section{Related Work}
\label{sec:related_work}

%% Long version below if you need more context!
\textbf{System Identification.}
Physical parameters such as stiffness or relaxation behavior are rarely known precisely for real-world scenes, which limits the use of \gls{pde} simulators in many engineering applications~\citep{landau2012theory, blazek2015computational, salatovic2025reliable, avril2017material, chai2024smart}.
System identification~\citep{aastrom1971system, lennart1999system} recovers such parameters from indirect observations.
In applied mechanics, related approaches include finite element model updating and the Virtual Fields Method~\citep{avril2008overview, pierron2010extension, bonnet2005inverse}.
Recent approaches pair learned models with physics engines or differentiable simulators for property estimation and control~\citep{wu2015galileo, murthy2020gradsim, antonova2023rethinking}, and train surrogate models for \gls{pde}-constrained parameter inference~\citep{qzhao2022graphpde, wang2024latent, adaptigraph24}.
% Simulators are critical for solving \glspl{pde} across a number engineering applications, yet essential physical parameters such as stiffness or relaxation behavior are rarely known precisely for a given real-world scene~\citep{landau2012theory, blazek2015computational, salatovic2025reliable, avril2017material, chai2024smart}.
% System identification \citep{aastrom1971system, lennart1999system} addresses this by recovering parameters from indirect observations, with counterparts in applied mechanics spanning updating of finite element models to the Virtual Fields Method \citep{avril2008overview, pierron2010extension, bonnet2005inverse}.
% and identifiability-aware analyses of nonlinear constitutive laws \citep{zhang2022parameter}.
% Learning-based variants couple deep models to physics engines for property estimation from video data \citep{wu2015galileo}, use differentiable simulators for gradient-based identification and control \citep{murthy2020gradsim, antonova2023rethinking}, or invert learned surrogates for \gls{pde}-constrained inverse problems and online property estimation \citep{qzhao2022graphpde, wang2024latent, adaptigraph24}.
% Learning-based approaches combine deep models with physics engines or differentiable simulators for property estimation and control \citep{wu2015galileo, murthy2020gradsim, antonova2023rethinking}, and extend to inverse problems by learning surrogate models for \gls{pde}-constrained parameter inference \citep{qzhao2022graphpde, wang2024latent, adaptigraph24}.

A different line of robotic real-to-sim work recovers deformable-object parameters via a per-instance test-time optimization.
These works include iterative FEM-comparison \citep{frank10}, to differentiable rendering and simulation \citep{diffcloud22}, to differentiable-simulator system identification on point clouds \citep{yang2025differentiable, adaptigraph24}.
Each instance requires a fresh optimization, which is typically done iteratively. 
The objective is non-convex and exhibits local minima and sloppy parameter directions~\citep{transtrum2010nonlinear, zhang2022parameter}. 
Direct-inversion alternatives sidestep the non-convexity but require dense full-field strain measurements~\citep{avril2008overview} that are unavailable in typical real-world settings.
\gls{peach} amortizes identification into a learned encoder, replacing per-instance optimization with a single forward pass. 
Although our simulator is differentiable, we avoid test-time optimization as it scales poorly across many instances and because the inverse problem is ill-conditioned~\citep{zhang2022parameter}, and instead rely on generalization from training data.

% These methods trigger a fresh inference step per instance that requires the forward model to remain well-conditioned: the iterative-optimization majority is additionally non-convex with documented local minima and sloppy parameter directions \citep{transtrum2010nonlinear, zhang2022parameter}, while direct-inversion methods avoid this only by requiring dense full-field strain measurements \citep{avril2008overview} unavailable from typical real-world sensing.
% \gls{peach} amortizes identification into a learned encoder: a single forward pass replaces the inner step, and although our forward simulator is differentiable, we avoid test-time optimization both because it scales poorly across many instances and because the inverse problem is generally ill-conditioned \citep{transtrum2010nonlinear, zhang2022parameter}, trading these failure modes for the more controllable one of imperfect generalization from training data.

\textbf{Adapting Learned Simulators to New Materials.}
% S1 — GNS as the relevant family
Graph network simulators have emerged as fast, differentiable surrogates for mesh- and particle-based physics, enabling scalable simulation of physical systems of rigid bodies \citep{battaglia2016interaction, sanchezgonzalez2020learning}, deformable solids~\citep{pfaff2020learning}, and PDE-like dynamics on unstructured meshes~\citep{brandstetter2021message}. 
Recent extensions aim to capture long-range interactions in nonlinear solid mechanics via hierarchical architectures \citep{yu2024learning, freymuth2025amber, wurth2025diffusion}.
% S2 — Adjacent paradigms
% S3 — The adaptation problem within learned simulators
These methods assume known process conditions and thus inherit the parameter-estimation problem whenever materials change.
Recent work therefore conditions learned simulators on material parameters either via parameter-efficient fine-tuning \citep{manoharan2025parameter} or via meta-learning over a context set of trajectories \citep{chakrabarty2025meta, dahlinger2025context}.
% S4 — In-context conditioning, MaNGO as the closest neighbor

In particular, MaNGO \citep{dahlinger2025mango} encodes a context set of mesh trajectories into a latent material descriptor that conditions a trajectory-level \gls{gns}, enabling adaptation to unseen materials in a single forward pass without test-time optimization. 
This is achieved via a \glsf{cnp} formulation for in-context learning \citep{garnelo_conditional_2018}.
% S5 — The gap, with goal/method distinction
However, mesh trajectories of arbitrarily deforming objects are not directly observable with real-world sensors, limiting MaNGO to synthetic data regimes.
% Since our objective is real-to-sim adaptation across varying material conditions
Instead, we condition on point cloud sequences, enabling adaptive physical simulation of real scenes with readily available 3D cameras.
Recent advances in foundation models for stereo reconstruction further support this approach by enabling the generation of cleaner and less noisy point clouds \citep{wen2025stereo}.

\begin{figure}[!t]
  \centering
  \includegraphics[width=0.192\textwidth, trim={0 2.5cm 0 3.5cm}, clip]{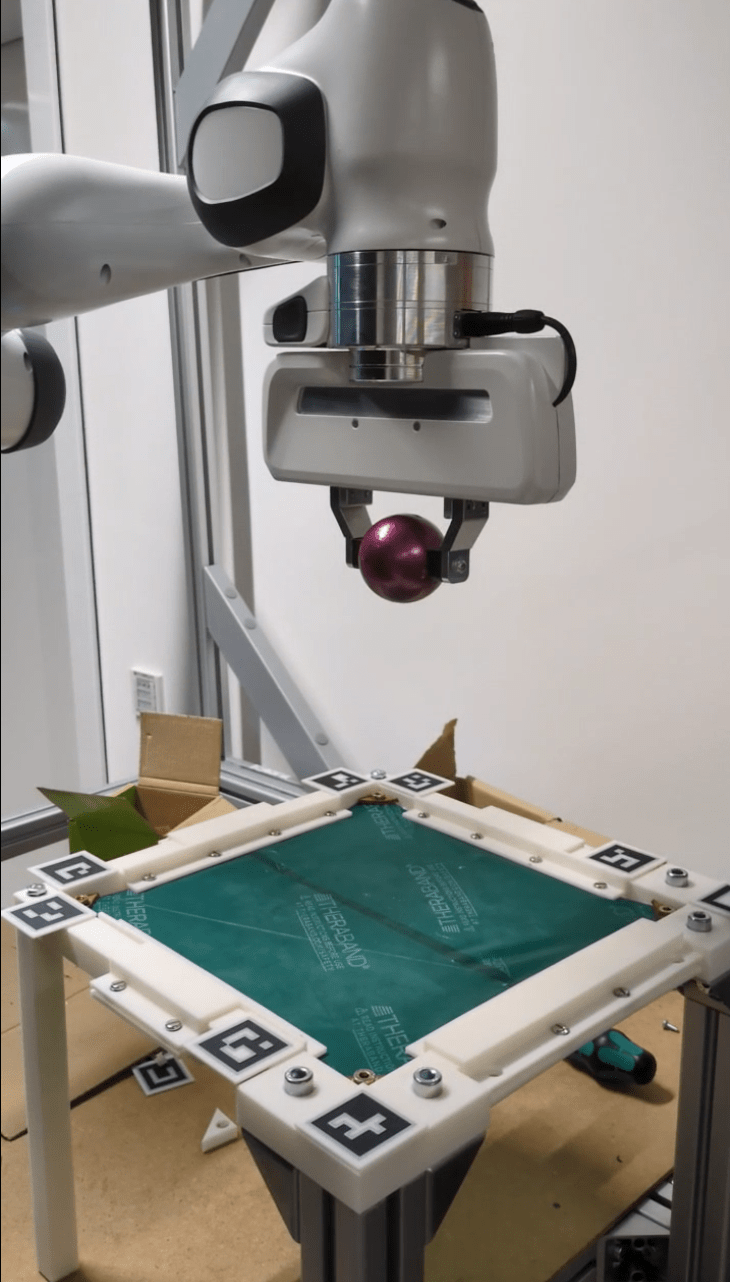}\hfill
  \includegraphics[width=0.192\textwidth, trim={0 2.5cm 0 3.5cm}, clip]{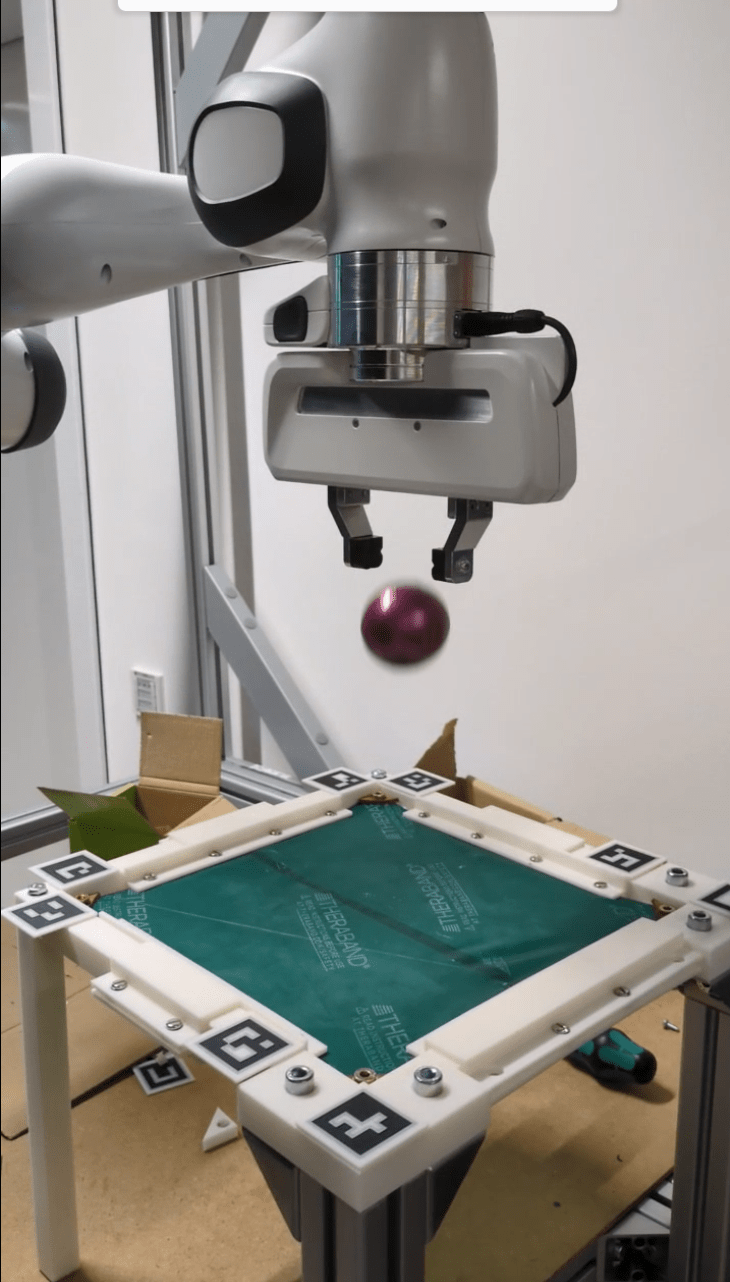}\hfill
  \includegraphics[width=0.192\textwidth, trim={0 2.5cm 0 3.5cm}, clip]{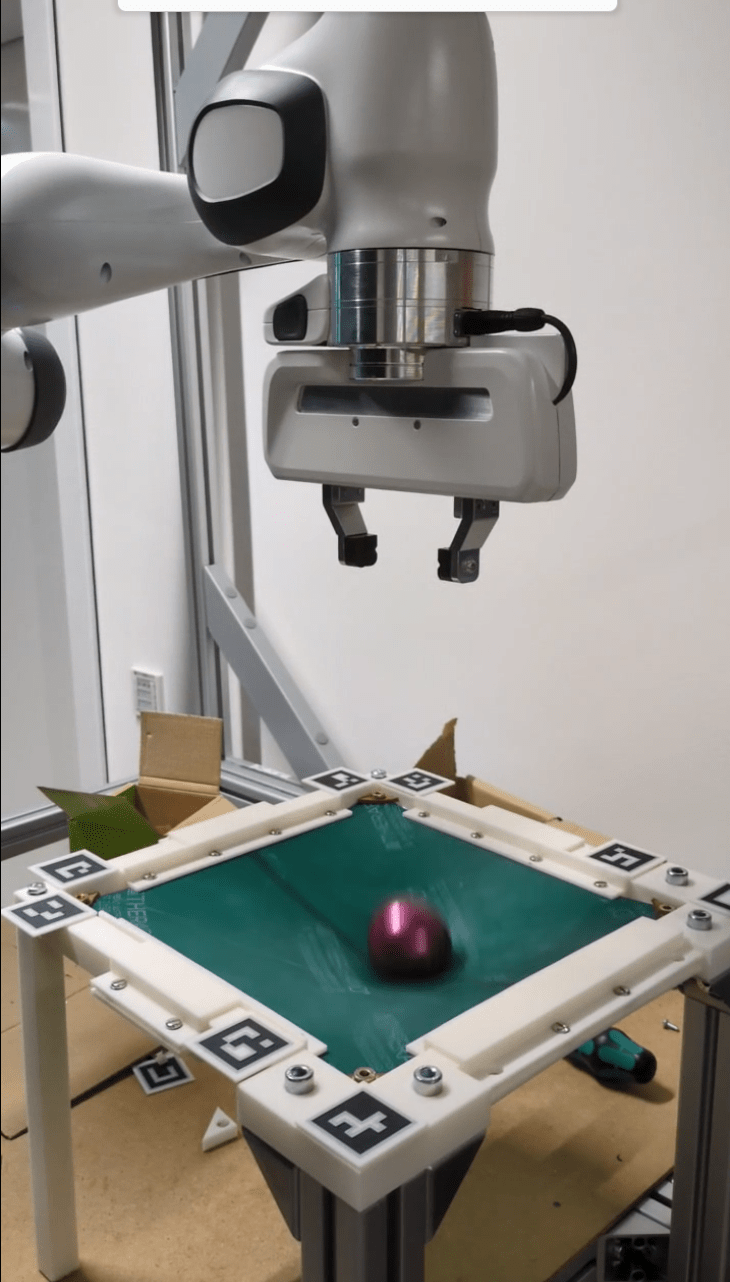}\hfill
  \includegraphics[width=0.192\textwidth, trim={0 2.5cm 0 3.5cm}, clip]{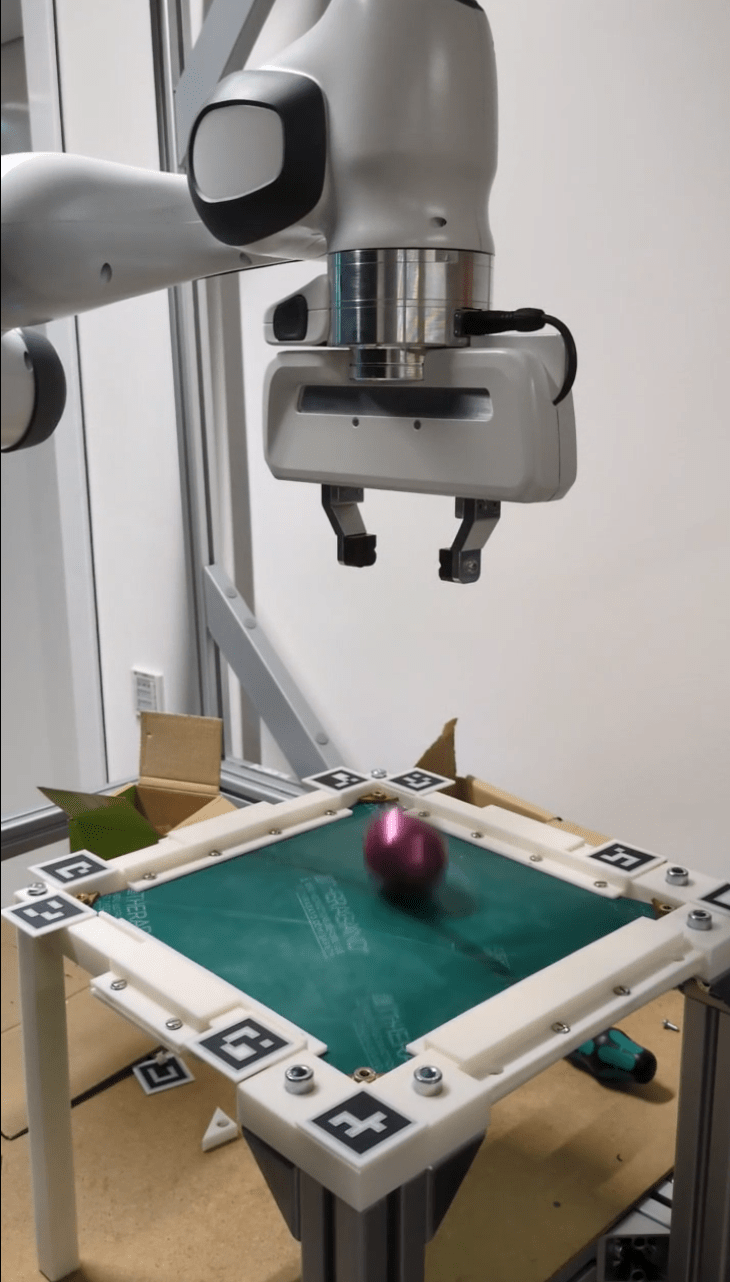}\hfill
  \includegraphics[width=0.192\textwidth, trim={0 2.5cm 0 3.5cm}, clip]{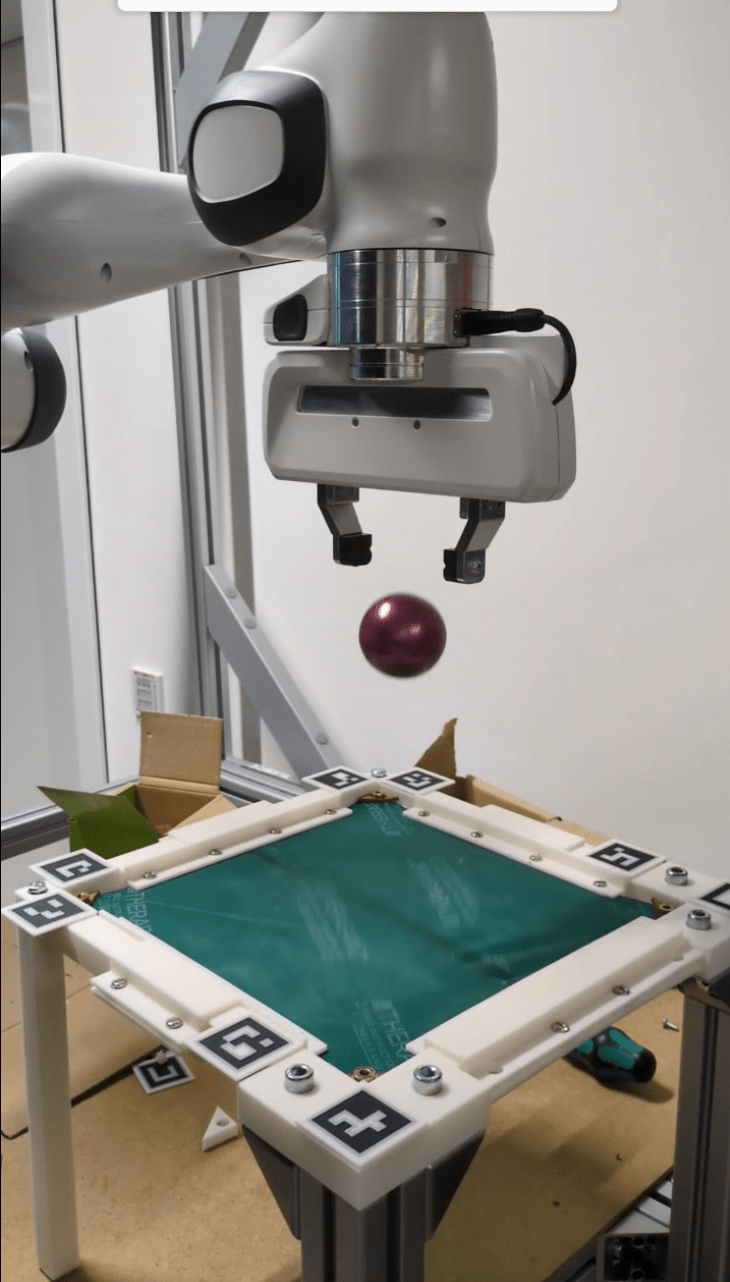}
  \caption{Real-world \texttt{Trampoline} setup. A robot arm releases a steel ball above an elastic membrane stretched across a 3D-printed frame, with ArUco markers at the corners for calibration.}
  \label{fig:realworld_setup}
  \vspace{-0.3cm}
\end{figure}

% Paragraph 3: Positioning, Engine and the Juice
\textbf{Point Cloud Sequence Observations for Learned Simulators.}
While permutation-invariant set encoders for individual point clouds \citep{qi2016pointnet, ruizhongtai2017pointnet, pointMAE, pointBERT, pointGPT} are well-established, processing point cloud sequences requires architectures that handle spatial irregularity and frame-to-frame motion without explicit point correspondences \citep{meteornet19, fan2021pstnet}.
Within learned simulation, point clouds have been used either to ground mesh-based \glspl{gns} in sensor observations~\citep{linkerhagner2023grounding}, or as the basis of particle-based dynamics models that operate directly in observation space for deformable-object manipulation \citep{li2019learning, robocook23}.
% In both cases, they are used to directly drive the dynamics rather than serve as a context representation for material inference.
In both cases, the latent physical properties are implicitly learned in the dynamics model, making adapting to new conditions less data efficient.
Instant Policy \citep{vosylius2024instantpolicyincontextimitation} instead conditions imitation learning policies on point cloud demonstrations via amortized in-context learning. The method shares PEACH's conditioning paradigm but applies to robot policies rather than material parameters and operates without a downstream physics simulator.
% PEACH targets a regime that none of the above categories addresses jointly: amortized in-context real-to-sim adaptation from point cloud sensor data, combining the sequence-encoder lineage of \citep{meteornet19, fan2021pstnet, Gyenes24} with the in-context conditioning paradigm of \citep{vosylius2024instantpolicyincontextimitation, dahlinger2025mango} to bypass both the test-time optimization of differentiable-simulator real-to-sim methods and the mesh-trajectory requirement of prior in-context simulator adaptation \citep{dahlinger2025mango}.

PEACH targets a regime not jointly addressed by prior work: amortized in-context sim-to-real adaptation from point cloud observations.
It combines sequence-based point cloud encoders \citep{meteornet19, fan2021pstnet} with in-context conditioning mechanisms \citep{vosylius2024instantpolicyincontextimitation, dahlinger2025mango}, while avoiding both the test-time optimization required by differentiable-simulator real-to-sim methods and the reliance on mesh-trajectory supervision in prior in-context simulator adaptation \citep{dahlinger2025mango}.

\section{Point Cloud Encoding for Accurate Context Handling}
\label{sec:method}

\textbf{Problem Setting.}
We consider a family of physical systems that share the same governing equations but differ in their physical properties, such as stiffness, damping, or thickness.
The goal is to learn a simulator that mimics a \gls{fem} solver and predicts a sequence of mesh states $\hat{\mathcal{G}}_{1:T}$ at discrete frames $1,\ldots,T$, without explicit knowledge of the physical properties $\rho$.
% A requirement for deployment is the availability of an initial mesh state $\mathcal{G}_0$, which serves as the starting point for simulation. 
% While this may appear restrictive, it aligns with standard practice in industry, where mesh-based representations are commonly available or can be obtained through established reconstruction pipelines.
% In this sense, our approach integrates naturally into existing simulation workflows while extending them with learned material inference from observations.
This setup is common in industrial applications, where mesh-based representations are commonly available or can be obtained through established reconstruction pipelines.
In our work, each \emph{task} corresponds to a fixed configuration of physical properties $\rho$, where the initial conditions such as mesh geometry or external forces may vary.
For each initial condition, we generate a ground truth mesh trajectory using an \gls{fem} solver.
By ray-tracing the mesh states with a virtual camera, we subsequently create an observed point cloud trajectory for each mesh trajectory.
At inference, we wish to adapt our learned simulator to the given task, conditioned only on a small context of observed point cloud trajectories.
% Instead, this information is inferred from a small context set of point cloud sequences.

\textbf{Framework Overview.}
During training, we sample a subset of $C$ point cloud trajectories from one task, where each trajectory $\mathcal{P}_i \in \mathbb{R}^{T \times N \times 3}$ consists of $T$ frames of $N$ points in 3D space.
We encode each trajectory using a point cloud sequence encoder to obtain latent codes $\{\mathbf{r}_1, \dots, \mathbf{r}_C\}$, where each $\mathbf{r}_c$ is a $d_{\mathbf r}$-dimensional vector describing the physical properties from this trajectory in latent space.
The per-sequence latent codes are further aggregated into a single task-level latent embedding $\mathbf{r}$ of the physical properties via a learned softmax aggregation, enabling the model to handle a variable number of context trajectories.
Concretely, given per-sequence latent codes $\{\mathbf{r}_1, \dots, \mathbf{r}_C\}$, the aggregated embedding is computed as
\begin{equation}
    \mathbf{r} = \sum_{c=1}^{C} \boldsymbol{\alpha}_c \odot \mathbf{r}_c, \quad \boldsymbol{\alpha}_c = \frac{\exp(\beta \mathbf{r}_c)}{\sum_{c'=1}^{C} \exp(\beta \mathbf{r}_{c'})},
\end{equation}
where $\beta$ is a learnable inverse-temperature parameter and the softmax is applied element-wise across the context dimension.
Together with an initial mesh state $\mathcal{G}_0$ sampled from the same task but independent of the context trajectories, $\mathbf{r}$ is passed to a differentiable mesh-based simulator to predict the remaining trajectory $\hat{\mathcal{G}}_{1:T}$, supervised against the ground truth $\mathcal{G}_{1:T}$.
The simulator is realized as a \gls{gns}.
Gradients are propagated end-to-end through both the simulator and the point cloud encoder, ensuring that the learned physical representation is optimized for predictive accuracy.

At inference time, the material parameters $\rho$ are unknown and only the mesh state $\mathcal{G}_0$ of the initial configuration is available.
The model must therefore rely entirely on the learned context encoding to estimate the material properties from a given sequence of context point clouds and simulate the resulting deformation.
% Crucially, real-world point cloud observations can be used directly as context during inference, bridging the gap between simulation and physical deployment.
As we show in our experiments, the model is capable of generalizing to unseen physical properties in simulation and even to real data.
Note that no real-world data is used for training, as neither ground truth mesh trajectories nor the true physical properties $\rho$ are available. We next describe the graph network simulator used as the dynamics backbone of our framework.

\textbf{Graph Network Simulators.}
\glspl{gns} generally follow one of two paradigms.
Step-based simulators~\cite{pfaff2020learning,wurth2025diffusion} iteratively predict the next mesh state from the current one, which introduces a train-to-test gap due to error accumulation during rollout.
Trajectory-based simulators~\cite{dahlinger2025mango, dahlinger2025context, egno2024} instead predict the entire remaining trajectory in a single forward pass, at the cost of higher memory requirements.
%This perspective integrates naturally into our in-context learning setup, as the inferred latent material parameters condition the entire predicted trajectory.
\gls{peach} utilizes the trajectory-based \gls{mango} simulator~\cite{dahlinger2025mango} there conditioning on the entire predicted trajectory, which we briefly review here for completeness.

Rather than maintaining separate graphs per timestep, \gls{mango} represents the trajectory as a single static graph $\mathcal{G} = (\mathcal{V}, \mathcal{E})$ with temporally indexed node and edge feature tensors $\mathbf{m}_v \in \mathbb{R}^{T \times |\mathcal{V}| \times d_\text{node}}$ and $\mathbf{m}_e \in \mathbb{R}^{T \times |\mathcal{E}| \times d_\text{edge}}$. The initial node features and trajectory-level conditioning variables, such as latent physical parameters, are broadcast across all time steps and augmented with a learned time embedding per timestep. Edge features are constructed from relative nodal positions. Both are then linearly projected into a latent dimension to obtain the initial feature tensors $\mathbf{m}_v^0$ and $\mathbf{m}_e^0$.
Spatial reasoning within each time step is performed via a message passing \gls{gnn} layer, which updates node and edge features based on local neighborhood interactions:
\begin{equation*}
    \mathbf m_e^{k+1} = f_\mathcal{E}^k(\mathbf m_e ^k, \mathbf m_v ^k, \mathbf m_w ^k)\text{, with } e=(v, w), \qquad \tilde {\mathbf m}_v^{k+1} = f_\mathcal{V}^k(\mathbf m_v ^k, \bigoplus_{e \in \mathcal{E}_v} \mathbf m_e ^{k+1}).
\end{equation*}
Here, $\mathcal{E}_v \subseteq \mathcal{E}$ denotes the edges incident to node $v$, and $\bigoplus$ is a permutation-invariant aggregation. The functions $f_\mathcal{E}^k$ and $f_\mathcal{V}^k$ are learned \glspl{mlp}.

The resulting node features $\tilde{\mathbf{m}}_v^{k+1}$ are then processed independently over time by a 1D residual \gls{cnn} to obtain ${\mathbf{m}}_v^{k+1}$.
After $K$ rounds of spatial message passing interleaved with temporal \gls{cnn} processing, the model is trained by summing per-timestep \gls{mse} losses against ground truth mesh states over the full rollout. To condition the simulator on observed data, we require an encoder that maps point cloud trajectories into a compact representation of the underlying material properties. 

% \begin{figure}[t]
%     \centering
%     \makebox[\textwidth][c]{
%     \input{figures/quantitative_results/legend}
%     }
%     \scalebox{0.39}{
%         \input{figures/quantitative_results/db/ml_loss/bar_db_bar}
%     }
%     \scalebox{0.39}{
%         \input{figures/quantitative_results/sd/ml_loss/bar_sd_bar}
%     }
%     \scalebox{0.39}{
%         \input{figures/quantitative_results/bbv/ml_loss/bar_bbv_bar}
%     }
%     \scalebox{0.39}{
%         \input{figures/quantitative_results/trampoline/ml_loss/bar_trampoline_v4_sim_bar}
%     }
%     % \input{figures/quantitative_results/db/ml_loss/bar_db_bar}
%     % \includegraphics[width=0.24\textwidth]{figures/quantitative_results/sd/ml_loss/bar_sd_bar.pdf}
%     % \includegraphics[width=0.24\textwidth]{figures/quantitative_results/bbv/ml_loss/bar_bbv_bar.pdf}
%     % \includegraphics[width=0.24\textwidth]{figures/quantitative_results/trampoline/ml_loss/bar_trampoline_v4_sim_bar.pdf}
%     \caption{
%     Full simulation rollout and material property \gls{mse} across different encoder architectures as a function of the number of context sequences for (\textbf{left}) \texttt{Deforming Block} and (\textbf{right}) \texttt{Sheet Deformation}.
%     \gls{peach} consistently performs similar to the mesh-based \gls{mango}, improving over other point cloud encoders and achieving near-Oracle performance on \texttt{Deforming Block}.
%     }
%     \label{fig:quantitative_comparison}
% \end{figure}

\begin{figure}[t]
    \centering
    % Legend at the top
    \makebox[\textwidth][c]{
        % \begin{tikzpicture}
%     \tikzstyle{every node}=[font=\scriptsize]
%     \input{figures/quantitative_results/tikzcolors}
%     \begin{axis}[%
%     hide axis,
%     xmin=10,
%     xmax=50,
%     ymin=0,
%     ymax=0.1,
%     legend style={
%         draw=white!15!black,
%         legend cell align=left,
%         legend columns=3,
%         legend style={
%             draw=none,
%             column sep=1ex,
%             line width=1pt,
%         }
%     },
%     ]

%     \addlegendimage{
%         line legend,
%         tabblue,
%         thick,
%         mark=square*,
%         mark options={line width=1.5pt, rotate=45}
%     }
%     \addlegendentry{\textbf{PEACH Encoder} (Ours)}
%     \addlegendimage{line legend, tabgreen, thick, mark=triangle*, mark options={line width=1.5pt, rotate=0}}
%     \addlegendentry{GNN Encoder}
%     \addlegendimage{
%         line legend,
%         taborange,
%         thick,
%         mark=square*,
%         mark options={line width=1.5pt, rotate=0}
%     }
%     \addlegendentry{PSTNet Encoder}
%     % New Line
%     % \addlegendimage{empty legend}
%     % \addlegendentry{}

%     \addlegendimage{line legend, tabolive, thick, mark=x, mark size=3pt, mark options={line width=1pt}}
%     \addlegendentry{Oracle Encoder}
%     \addlegendimage{line legend, tabcyan, thick, mark=+, mark size=2.5pt, mark options={line width=2pt}}
%     \addlegendentry{MaNGO Mesh Encoder}
%     \addlegendimage{line legend, tabgray, thick, mark=*}
%     \addlegendentry{No Context Encoder}    
%     \end{axis}
% \end{tikzpicture}

\begin{tikzpicture}
    \tikzstyle{every node}=[font=\scriptsize]
    \input{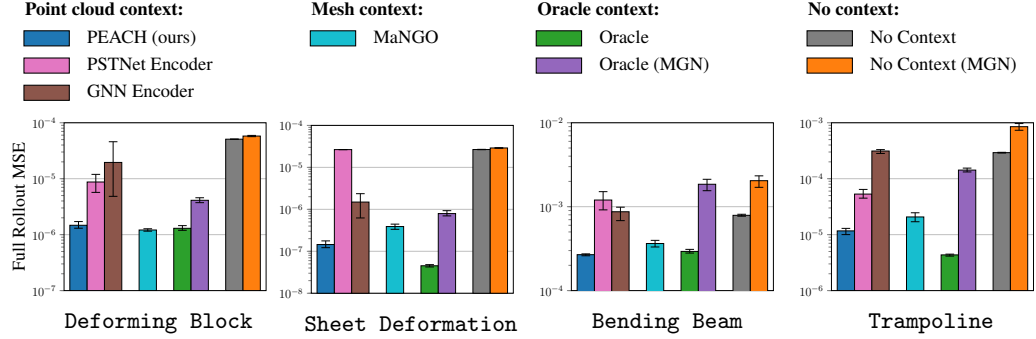}
    \begin{axis}[%
        hide axis,
        xmin=10,
        xmax=50,
        ymin=0,
        ymax=0.1,
        legend style={
            draw=none,
            legend cell align=left,
            legend columns=7,
            column sep=1ex,
        }
    ]
        \addlegendimage{area legend, fill=white, draw=white, line width=0pt}
    \addlegendentry{\hspace{-8.2mm}\textbf{Point cloud context:}}
    \addlegendimage{area legend, fill=white, draw=white, line width=0pt}
        \addlegendentry{\hspace{-1ex}}
    \addlegendimage{area legend, fill=white, draw=white, line width=0pt}
    \addlegendentry{\hspace{-8.2mm}\textbf{Mesh context:}}
        \addlegendimage{area legend, fill=white, draw=white, line width=0pt}
        \addlegendentry{\hspace{-1ex}}
    \addlegendimage{area legend, fill=white, draw=white, line width=0pt}
    \addlegendentry{\hspace{-8.2mm}\textbf{Oracle context:}}
        \addlegendimage{area legend, fill=white, draw=white, line width=0pt}
        \addlegendentry{\hspace{-1ex}}
    \addlegendimage{area legend, fill=white, draw=white, line width=0pt}
    \addlegendentry{\hspace{-8.2mm}\textbf{No context:}}
    % Col 1: PEACH
    \addlegendimage{area legend, fill=tabblue}
    \addlegendentry{PEACH (ours)}
    % Col 3: spacer
    \addlegendimage{area legend, fill=white, draw=white, line width=0pt}
    \addlegendentry{\hspace{-1ex}}
    % Col 4: MaNGO
    \addlegendimage{area legend, fill=tabcyan}
    \addlegendentry{MaNGO}
    % Col 5: spacer
    \addlegendimage{area legend, fill=white, draw=white, line width=0pt}
    \addlegendentry{\hspace{-1ex}}
    % Col 6: Oracle + Oracle MGN
    \addlegendimage{area legend, fill=tabgreen}
    \addlegendentry{Oracle}
    % Col 7: spacer
    \addlegendimage{area legend, fill=white, draw=white, line width=0pt}
    \addlegendentry{\hspace{-1ex}}
    % Col 8: No Context + No Context MGN
    \addlegendimage{area legend, fill=tabgray}
    \addlegendentry{No Context}
    % Row 2
        \addlegendimage{area legend, fill=tabpink}
    \addlegendentry{PSTNet Encoder}
    \addlegendimage{area legend, fill=white, draw=white, line width=0pt}
    \addlegendentry{\hspace{-1ex}}
    \addlegendimage{area legend, fill=white, draw=white, line width=0pt}
    \addlegendentry{\hspace{-1ex}}
    \addlegendimage{area legend, fill=white, draw=white, line width=0pt}
    \addlegendentry{\hspace{-1ex}}
    \addlegendimage{area legend, fill=tabpurple}
    \addlegendentry{Oracle (MGN)}
    \addlegendimage{area legend, fill=white, draw=white, line width=0pt}
    \addlegendentry{\hspace{-1ex}}
    \addlegendimage{area legend, fill=taborange}
    \addlegendentry{No Context (MGN)}

        % Col 2: PSTNet
\addlegendimage{area legend, fill=tabbrown}
    \addlegendentry{GNN Encoder}
    \end{axis}
\end{tikzpicture}
    }
    % Row of plots with labels underneath
    \begin{minipage}{0.24\textwidth}
        \centering
        \scalebox{0.39}{\begin{tikzpicture}
\definecolor{darkgray}{RGB}{169,169,169}
\definecolor{darkorange25512714}{RGB}{255,127,14}
\definecolor{darkslategray38}{RGB}{38,38,38}
\definecolor{darkturquoise23190207}{RGB}{23,190,207}
\definecolor{gray127}{RGB}{127,127,127}
\definecolor{lightgray204}{RGB}{204,204,204}
\definecolor{mediumpurple148103189}{RGB}{148,103,189}
\definecolor{orchid227119194}{RGB}{227,119,194}
\definecolor{sienna1408675}{RGB}{140,86,75}
\definecolor{steelblue31119180}{RGB}{31,119,180}
\definecolor{steelblue76114176}{RGB}{76,114,176}
\definecolor{tabgreen}{RGB}{44,160,44}

\begin{semilogyaxis}[
  xmin=-0.5, xmax=6.5,
  xtick=\empty,
  ymin=1e-7, ymax=1e-4,
  ymajorgrids,
  tick align=outside,
  ylabel={Full Rollout MSE},
  ylabel style={font=\LARGE},
  ytick pos=left,
  tick label style={font=\large},
]

% Group 1: PEACH (0), pstnet (0.6), gnn_encoder (1.2)
\draw[fill=steelblue31119180, draw=black, line width=0.5pt] (axis cs:-0.3, 1e-7) rectangle (axis cs:0.3,  1.4707e-6);
\draw[fill=orchid227119194,   draw=black, line width=0.5pt] (axis cs:0.3,  1e-7) rectangle (axis cs:0.9,  8.7298e-6);
\draw[fill=sienna1408675,     draw=black, line width=0.5pt] (axis cs:0.9,  1e-7) rectangle (axis cs:1.5,  1.9537e-5);

% Group 2: MaNGO (2.4)
\draw[fill=darkturquoise23190207, draw=black, line width=0.5pt] (axis cs:2.1, 1e-7) rectangle (axis cs:2.7, 1.2112e-6);

% Group 3: Oracle (3.6), Oracle MGN (4.2)
\draw[fill=tabgreen,              draw=black, line width=0.5pt] (axis cs:3.3, 1e-7) rectangle (axis cs:3.9, 1.2985e-6);
\draw[fill=mediumpurple148103189, draw=black, line width=0.5pt] (axis cs:3.9, 1e-7) rectangle (axis cs:4.5, 4.1116e-6);

% Group 4: No Context (5.4), No Context MGN (6.0)
\draw[fill=gray127,            draw=black, line width=0.5pt] (axis cs:5.1, 1e-7) rectangle (axis cs:5.7, 5.0834e-5);
\draw[fill=darkorange25512714, draw=black, line width=0.5pt] (axis cs:5.7, 1e-7) rectangle (axis cs:6.3, 5.7853e-5);

% Error bars
% PEACH (center 0)
\draw[black, semithick] (axis cs:0,    1.3030e-6) -- (axis cs:0,    1.7132e-6);
\draw[black, semithick] (axis cs:-0.15,1.3030e-6) -- (axis cs:0.15, 1.3030e-6);
\draw[black, semithick] (axis cs:-0.15,1.7132e-6) -- (axis cs:0.15, 1.7132e-6);
% pstnet (center 0.6)
\draw[black, semithick] (axis cs:0.6,  5.6935e-6) -- (axis cs:0.6,  1.1953e-5);
\draw[black, semithick] (axis cs:0.45, 5.6935e-6) -- (axis cs:0.75, 5.6935e-6);
\draw[black, semithick] (axis cs:0.45, 1.1953e-5) -- (axis cs:0.75, 1.1953e-5);
% gnn_encoder (center 1.2)
\draw[black, semithick] (axis cs:1.2,  4.8388e-6) -- (axis cs:1.2,  4.5789e-5);
\draw[black, semithick] (axis cs:1.05, 4.8388e-6) -- (axis cs:1.35, 4.8388e-6);
\draw[black, semithick] (axis cs:1.05, 4.5789e-5) -- (axis cs:1.35, 4.5789e-5);
% MaNGO (center 2.4)
\draw[black, semithick] (axis cs:2.4,  1.1531e-6) -- (axis cs:2.4,  1.2742e-6);
\draw[black, semithick] (axis cs:2.25, 1.1531e-6) -- (axis cs:2.55, 1.1531e-6);
\draw[black, semithick] (axis cs:2.25, 1.2742e-6) -- (axis cs:2.55, 1.2742e-6);
% oracle (center 3.6)
\draw[black, semithick] (axis cs:3.6,  1.1952e-6) -- (axis cs:3.6,  1.4497e-6);
\draw[black, semithick] (axis cs:3.45, 1.1952e-6) -- (axis cs:3.75, 1.1952e-6);
\draw[black, semithick] (axis cs:3.45, 1.4497e-6) -- (axis cs:3.75, 1.4497e-6);
% mgn_oracle (center 4.2)
\draw[black, semithick] (axis cs:4.2,  3.7312e-6) -- (axis cs:4.2,  4.5653e-6);
\draw[black, semithick] (axis cs:4.05, 3.7312e-6) -- (axis cs:4.35, 3.7312e-6);
\draw[black, semithick] (axis cs:4.05, 4.5653e-6) -- (axis cs:4.35, 4.5653e-6);
% No Context (center 5.4)
\draw[black, semithick] (axis cs:5.4,  5.0709e-5) -- (axis cs:5.4,  5.0960e-5);
\draw[black, semithick] (axis cs:5.25, 5.0709e-5) -- (axis cs:5.55, 5.0709e-5);
\draw[black, semithick] (axis cs:5.25, 5.0960e-5) -- (axis cs:5.55, 5.0960e-5);
% No Context MGN (center 6.0)
\draw[black, semithick] (axis cs:6.0,  5.6436e-5) -- (axis cs:6.0,  5.9399e-5);
\draw[black, semithick] (axis cs:5.85, 5.6436e-5) -- (axis cs:6.15, 5.6436e-5);
\draw[black, semithick] (axis cs:5.85, 5.9399e-5) -- (axis cs:6.15, 5.9399e-5);

\end{semilogyaxis}
\end{tikzpicture}}\\
        {\small \hspace{6mm}\texttt{Deforming Block}}
    \end{minipage}
    \begin{minipage}{0.24\textwidth}
        \centering
        \scalebox{0.39}{\begin{tikzpicture}
\definecolor{tabblue}{RGB}{31,119,180}
\definecolor{tabgreen}{RGB}{44,160,44}
\definecolor{tabcyan}{RGB}{23,190,207}
\definecolor{tabgray}{RGB}{127,127,127}
\definecolor{taborange}{RGB}{255,127,14}
\definecolor{tabpurple}{RGB}{148,103,189}
\definecolor{tabpink}{RGB}{227,119,194}
\definecolor{tabbrown}{RGB}{140,86,75}

\begin{semilogyaxis}[
  xmin=-0.5, xmax=6.5,
  xtick=\empty,
  ymin=1e-8, ymax=1e-4,
  ymajorgrids,
  ytick pos=left,
  tick align=outside,
  tick label style={font=\large},
]

% Group 1: PEACH/tabblue (0), pstnet/tabpink (0.6), gnn_encoder/tabbrown (1.2)
\draw[fill=tabblue,  draw=black, line width=0.5pt] (axis cs:-0.3, 1e-8) rectangle (axis cs:0.3, 1.45e-7);
\draw[fill=tabpink,  draw=black, line width=0.5pt] (axis cs:0.3,  1e-8) rectangle (axis cs:0.9, 2.65e-5);
\draw[fill=tabbrown, draw=black, line width=0.5pt] (axis cs:0.9,  1e-8) rectangle (axis cs:1.5, 1.49e-6);

% Group 2: MaNGO/tabcyan (2.4)
\draw[fill=tabcyan,   draw=black, line width=0.5pt] (axis cs:2.1, 1e-8) rectangle (axis cs:2.7, 3.89e-7);

% Group 3: Oracle/tabgreen (3.6), Oracle MGN/tabpurple (4.2)
\draw[fill=tabgreen,  draw=black, line width=0.5pt] (axis cs:3.3, 1e-8) rectangle (axis cs:3.9, 4.52e-8);
\draw[fill=tabpurple, draw=black, line width=0.5pt] (axis cs:3.9, 1e-8) rectangle (axis cs:4.5, 7.92e-7);

% Group 4: No Context/tabgray (5.4), No Context MGN/taborange (6.0)
\draw[fill=tabgray,   draw=black, line width=0.5pt] (axis cs:5.1, 1e-8) rectangle (axis cs:5.7, 2.66e-5);
\draw[fill=taborange, draw=black, line width=0.5pt] (axis cs:5.7, 1e-8) rectangle (axis cs:6.3, 2.90e-5);

% Error bars
% tabblue/PEACH (center 0)
\draw[black, semithick] (axis cs:0,    1.22e-7) -- (axis cs:0,    1.77e-7);
\draw[black, semithick] (axis cs:-0.15,1.22e-7) -- (axis cs:0.15, 1.22e-7);
\draw[black, semithick] (axis cs:-0.15,1.77e-7) -- (axis cs:0.15, 1.77e-7);
% tabpink/pstnet (center 0.6)
\draw[black, semithick] (axis cs:0.6,  2.64e-5) -- (axis cs:0.6,  2.66e-5);
\draw[black, semithick] (axis cs:0.45, 2.64e-5) -- (axis cs:0.75, 2.64e-5);
\draw[black, semithick] (axis cs:0.45, 2.66e-5) -- (axis cs:0.75, 2.66e-5);
% tabbrown/gnn_encoder (center 1.2)
\draw[black, semithick] (axis cs:1.2,  6.23e-7) -- (axis cs:1.2,  2.36e-6);
\draw[black, semithick] (axis cs:1.05, 6.23e-7) -- (axis cs:1.35, 6.23e-7);
\draw[black, semithick] (axis cs:1.05, 2.36e-6) -- (axis cs:1.35, 2.36e-6);
% tabcyan/MaNGO (center 2.4)
\draw[black, semithick] (axis cs:2.4,  3.34e-7) -- (axis cs:2.4,  4.44e-7);
\draw[black, semithick] (axis cs:2.25, 3.34e-7) -- (axis cs:2.55, 3.34e-7);
\draw[black, semithick] (axis cs:2.25, 4.44e-7) -- (axis cs:2.55, 4.44e-7);
% tabgreen/oracle (center 3.6)
\draw[black, semithick] (axis cs:3.6,  4.23e-8) -- (axis cs:3.6,  4.81e-8);
\draw[black, semithick] (axis cs:3.45, 4.23e-8) -- (axis cs:3.75, 4.23e-8);
\draw[black, semithick] (axis cs:3.45, 4.81e-8) -- (axis cs:3.75, 4.81e-8);
% tabpurple/mgn_oracle (center 4.2)
\draw[black, semithick] (axis cs:4.2,  6.98e-7) -- (axis cs:4.2,  9.29e-7);
\draw[black, semithick] (axis cs:4.05, 6.98e-7) -- (axis cs:4.35, 6.98e-7);
\draw[black, semithick] (axis cs:4.05, 9.29e-7) -- (axis cs:4.35, 9.29e-7);
% tabgray/No Context (center 5.4)
\draw[black, semithick] (axis cs:5.4,  2.64e-5) -- (axis cs:5.4,  2.69e-5);
\draw[black, semithick] (axis cs:5.25, 2.64e-5) -- (axis cs:5.55, 2.64e-5);
\draw[black, semithick] (axis cs:5.25, 2.69e-5) -- (axis cs:5.55, 2.69e-5);
% taborange/No Context MGN (center 6.0)
\draw[black, semithick] (axis cs:6.0,  2.85e-5) -- (axis cs:6.0,  2.94e-5);
\draw[black, semithick] (axis cs:5.85, 2.85e-5) -- (axis cs:6.15, 2.85e-5);
\draw[black, semithick] (axis cs:5.85, 2.94e-5) -- (axis cs:6.15, 2.94e-5);

\end{semilogyaxis}
\end{tikzpicture}}\\
        {\small \hspace{4mm}\texttt{Sheet Deformation}}
    \end{minipage}
    \begin{minipage}{0.24\textwidth}
        \centering
        \scalebox{0.39}{\begin{tikzpicture}
\definecolor{tabblue}{RGB}{31,119,180}
\definecolor{tabgreen}{RGB}{44,160,44}
\definecolor{tabcyan}{RGB}{23,190,207}
\definecolor{tabgray}{RGB}{127,127,127}
\definecolor{taborange}{RGB}{255,127,14}
\definecolor{tabpurple}{RGB}{148,103,189}
\definecolor{tabpink}{RGB}{227,119,194}
\definecolor{tabbrown}{RGB}{140,86,75}

\begin{semilogyaxis}[
  xmin=-0.5, xmax=6.5,
  xtick=\empty,
  ymin=1e-4, ymax=1e-2,
  ymajorgrids,
  tick align=outside,
  ytick pos=left,
  tick label style={font=\large},
]

% Group 1: PEACH/tabblue (0), pstnet/tabpink (0.6), gnn_encoder/tabbrown (1.2)
\draw[fill=tabblue,  draw=black, line width=0.5pt] (axis cs:-0.3, 1e-5) rectangle (axis cs:0.3, 2.68e-4);
\draw[fill=tabpink,  draw=black, line width=0.5pt] (axis cs:0.3,  1e-5) rectangle (axis cs:0.9, 1.20e-3);
\draw[fill=tabbrown, draw=black, line width=0.5pt] (axis cs:0.9,  1e-5) rectangle (axis cs:1.5, 8.70e-4);

% Group 2: MaNGO/tabcyan (2.4)
\draw[fill=tabcyan,   draw=black, line width=0.5pt] (axis cs:2.1, 1e-5) rectangle (axis cs:2.7, 3.64e-4);

% Group 3: Oracle/tabgreen (3.6), Oracle MGN/tabpurple (4.2)
\draw[fill=tabgreen,  draw=black, line width=0.5pt] (axis cs:3.3, 1e-5) rectangle (axis cs:3.9, 2.93e-4);
\draw[fill=tabpurple, draw=black, line width=0.5pt] (axis cs:3.9, 1e-5) rectangle (axis cs:4.5, 1.85e-3);

% Group 4: No Context/tabgray (5.4), No Context MGN/taborange (6.0)
\draw[fill=tabgray,   draw=black, line width=0.5pt] (axis cs:5.1, 1e-5) rectangle (axis cs:5.7, 7.89e-4);
\draw[fill=taborange, draw=black, line width=0.5pt] (axis cs:5.7, 1e-5) rectangle (axis cs:6.3, 2.04e-3);

% Error bars
% tabblue/PEACH (center 0)
\draw[black, semithick] (axis cs:0,    2.60e-4) -- (axis cs:0,    2.74e-4);
\draw[black, semithick] (axis cs:-0.15,2.60e-4) -- (axis cs:0.15, 2.60e-4);
\draw[black, semithick] (axis cs:-0.15,2.74e-4) -- (axis cs:0.15, 2.74e-4);
% tabpink/pstnet (center 0.6)
\draw[black, semithick] (axis cs:0.6,  9.16e-4) -- (axis cs:0.6,  1.51e-3);
\draw[black, semithick] (axis cs:0.45, 9.16e-4) -- (axis cs:0.75, 9.16e-4);
\draw[black, semithick] (axis cs:0.45, 1.51e-3) -- (axis cs:0.75, 1.51e-3);
% tabbrown/gnn_encoder (center 1.2)
\draw[black, semithick] (axis cs:1.2,  6.82e-4) -- (axis cs:1.2,  9.86e-4);
\draw[black, semithick] (axis cs:1.05, 6.82e-4) -- (axis cs:1.35, 6.82e-4);
\draw[black, semithick] (axis cs:1.05, 9.86e-4) -- (axis cs:1.35, 9.86e-4);
% tabcyan/MaNGO (center 2.4)
\draw[black, semithick] (axis cs:2.4,  3.31e-4) -- (axis cs:2.4,  3.97e-4);
\draw[black, semithick] (axis cs:2.25, 3.31e-4) -- (axis cs:2.55, 3.31e-4);
\draw[black, semithick] (axis cs:2.25, 3.97e-4) -- (axis cs:2.55, 3.97e-4);
% tabgreen/oracle (center 3.6)
\draw[black, semithick] (axis cs:3.6,  2.80e-4) -- (axis cs:3.6,  3.10e-4);
\draw[black, semithick] (axis cs:3.45, 2.80e-4) -- (axis cs:3.75, 2.80e-4);
\draw[black, semithick] (axis cs:3.45, 3.10e-4) -- (axis cs:3.75, 3.10e-4);
% tabpurple/mgn_oracle (center 4.2)
\draw[black, semithick] (axis cs:4.2,  1.55e-3) -- (axis cs:4.2,  2.12e-3);
\draw[black, semithick] (axis cs:4.05, 1.55e-3) -- (axis cs:4.35, 1.55e-3);
\draw[black, semithick] (axis cs:4.05, 2.12e-3) -- (axis cs:4.35, 2.12e-3);
% tabgray/No Context (center 5.4)
\draw[black, semithick] (axis cs:5.4,  7.70e-4) -- (axis cs:5.4,  8.12e-4);
\draw[black, semithick] (axis cs:5.25, 7.70e-4) -- (axis cs:5.55, 7.70e-4);
\draw[black, semithick] (axis cs:5.25, 8.12e-4) -- (axis cs:5.55, 8.12e-4);
% taborange/No Context MGN (center 6.0)
\draw[black, semithick] (axis cs:6.0,  1.70e-3) -- (axis cs:6.0,  2.33e-3);
\draw[black, semithick] (axis cs:5.85, 1.70e-3) -- (axis cs:6.15, 1.70e-3);
\draw[black, semithick] (axis cs:5.85, 2.33e-3) -- (axis cs:6.15, 2.33e-3);

\end{semilogyaxis}
\end{tikzpicture}}\\
        {\small \hspace{3mm}\texttt{Bending Beam}}
    \end{minipage}
    \begin{minipage}{0.24\textwidth}
        \centering
        \scalebox{0.39}{\begin{tikzpicture}
\definecolor{darkgray}{RGB}{169,169,169}
\definecolor{darkorange25512714}{RGB}{255,127,14}
\definecolor{darkslategray38}{RGB}{38,38,38}
\definecolor{darkturquoise23190207}{RGB}{23,190,207}
\definecolor{gray127}{RGB}{127,127,127}
\definecolor{lightgray204}{RGB}{204,204,204}
\definecolor{mediumpurple148103189}{RGB}{148,103,189}
\definecolor{orchid227119194}{RGB}{227,119,194}
\definecolor{sienna1408675}{RGB}{140,86,75}
\definecolor{steelblue31119180}{RGB}{31,119,180}
\definecolor{tabgreen}{RGB}{44,160,44}

\begin{semilogyaxis}[
  xmin=-0.5, xmax=6.5,
  xtick=\empty,
  ymin=1e-6, ymax=1e-3,
  ymajorgrids,
  ytick pos=left,
  tick align=outside,
  tick label style={font=\large},
]

% Group 1: PEACH/steelblue (0), pstnet/orchid (0.6), gnn_encoder/sienna (1.2)
\draw[fill=steelblue31119180, draw=black, line width=0.5pt] (axis cs:-0.3, 1e-6) rectangle (axis cs:0.3, 1.1675e-5);
\draw[fill=orchid227119194,   draw=black, line width=0.5pt] (axis cs:0.3,  1e-6) rectangle (axis cs:0.9, 5.3091e-5);
\draw[fill=sienna1408675,     draw=black, line width=0.5pt] (axis cs:0.9,  1e-6) rectangle (axis cs:1.5, 3.1270e-4);

% Group 2: MaNGO/darkturquoise (2.4)
\draw[fill=darkturquoise23190207, draw=black, line width=0.5pt] (axis cs:2.1, 1e-6) rectangle (axis cs:2.7, 2.0718e-5);

% Group 3: Oracle/tabgreen (3.6), Oracle MGN/mediumpurple (4.2)
\draw[fill=tabgreen,              draw=black, line width=0.5pt] (axis cs:3.3, 1e-6) rectangle (axis cs:3.9, 4.3457e-6);
\draw[fill=mediumpurple148103189, draw=black, line width=0.5pt] (axis cs:3.9, 1e-6) rectangle (axis cs:4.5, 1.4253e-4);

% Group 4: No Context/gray (5.4), No Context MGN/darkorange (6.0)
\draw[fill=gray127,            draw=black, line width=0.5pt] (axis cs:5.1, 1e-6) rectangle (axis cs:5.7, 2.9218e-4);
\draw[fill=darkorange25512714, draw=black, line width=0.5pt] (axis cs:5.7, 1e-6) rectangle (axis cs:6.3, 8.5081e-4);

% Error bars
% steelblue/PEACH (center 0)
\draw[black, semithick] (axis cs:0,    1.0020e-5) -- (axis cs:0,    1.2944e-5);
\draw[black, semithick] (axis cs:-0.15,1.0020e-5) -- (axis cs:0.15, 1.0020e-5);
\draw[black, semithick] (axis cs:-0.15,1.2944e-5) -- (axis cs:0.15, 1.2944e-5);
% orchid/pstnet (center 0.6)
\draw[black, semithick] (axis cs:0.6,  4.4918e-5) -- (axis cs:0.6,  6.4341e-5);
\draw[black, semithick] (axis cs:0.45, 4.4918e-5) -- (axis cs:0.75, 4.4918e-5);
\draw[black, semithick] (axis cs:0.45, 6.4341e-5) -- (axis cs:0.75, 6.4341e-5);
% sienna/gnn_encoder (center 1.2)
\draw[black, semithick] (axis cs:1.2,  2.8242e-4) -- (axis cs:1.2,  3.3119e-4);
\draw[black, semithick] (axis cs:1.05, 2.8242e-4) -- (axis cs:1.35, 2.8242e-4);
\draw[black, semithick] (axis cs:1.05, 3.3119e-4) -- (axis cs:1.35, 3.3119e-4);
% darkturquoise/MaNGO (center 2.4)
\draw[black, semithick] (axis cs:2.4,  1.7016e-5) -- (axis cs:2.4,  2.4613e-5);
\draw[black, semithick] (axis cs:2.25, 1.7016e-5) -- (axis cs:2.55, 1.7016e-5);
\draw[black, semithick] (axis cs:2.25, 2.4613e-5) -- (axis cs:2.55, 2.4613e-5);
% tabgreen/oracle (center 3.6)
\draw[black, semithick] (axis cs:3.6,  4.1340e-6) -- (axis cs:3.6,  4.4844e-6);
\draw[black, semithick] (axis cs:3.45, 4.1340e-6) -- (axis cs:3.75, 4.1340e-6);
\draw[black, semithick] (axis cs:3.45, 4.4844e-6) -- (axis cs:3.75, 4.4844e-6);
% mediumpurple/mgn_oracle (center 4.2)
\draw[black, semithick] (axis cs:4.2,  1.3208e-4) -- (axis cs:4.2,  1.5467e-4);
\draw[black, semithick] (axis cs:4.05, 1.3208e-4) -- (axis cs:4.35, 1.3208e-4);
\draw[black, semithick] (axis cs:4.05, 1.5467e-4) -- (axis cs:4.35, 1.5467e-4);
% gray/No Context (center 5.4)
\draw[black, semithick] (axis cs:5.4,  2.8797e-4) -- (axis cs:5.4,  2.9636e-4);
\draw[black, semithick] (axis cs:5.25, 2.8797e-4) -- (axis cs:5.55, 2.8797e-4);
\draw[black, semithick] (axis cs:5.25, 2.9636e-4) -- (axis cs:5.55, 2.9636e-4);
% darkorange/No Context MGN (center 6.0)
\draw[black, semithick] (axis cs:6.0,  7.3378e-4) -- (axis cs:6.0,  9.6784e-4);
\draw[black, semithick] (axis cs:5.85, 7.3378e-4) -- (axis cs:6.15, 7.3378e-4);
\draw[black, semithick] (axis cs:5.85, 9.6784e-4) -- (axis cs:6.15, 9.6784e-4);

\end{semilogyaxis}
\end{tikzpicture}}\\
        {\small \hspace{4mm}\texttt{Trampoline}}
    \end{minipage}

    \caption{
        Simulation accuracy (MSE) on simulation datasets using 8 context trajectories.
        \gls{peach} consistently outperforms other methods relying on point cloud observations for simulator adaptation.
        \gls{peach} performs comparably to \glsunset{mango}\gls{mango}, which uses privileged information in the form of meshes, and sometimes achieves performance competitive with the Oracle models.
    }
    \vspace{-0.4cm}
    \label{fig:main_results}
\end{figure}

\textbf{Spatio-Temporal Point Cloud Encoder.}
The core challenge of the point cloud encoder is to compress a large number of spatially dense but informationally sparse points into a compact and expressive latent representation.
Furthermore, there are numerous ways to treat a set of points across both space and time coordinates. 
While typical point-patch style encoders~\cite{pointMAE, pointBERT, pointGPT} are effective for tokenizing individual point clouds, material properties reveal themselves most clearly in the dynamics of the scene rather than in any single time step.
Therefore, we adapt this paradigm to point cloud sequences and develop a novel encoder architecture by considering points in 4D space-time.

Space-time coordinates are computed by normalizing the sequence duration and applying a scaling factor of $\tau$.
Setting $\tau {\rightarrow} 0$ has the effect of ignoring the time axis, while $\tau {\rightarrow} \infty$ recovers the case where each time step is encoded independently of all others.
We sample $M$ patch centers using \gls{fps}~\cite{ruizhongtai2017pointnet} and extract patches using \gls{knn}, both over space-time coordinates.
We compute the features of each point using a Fourier feature projection~\cite{tancik2020fourier, gyenes2026fourier} of its space-time coordinates \mbox{$\gamma:\mathbb{R}\to\mathbb{R}^{2L}$} given by
\begin{equation*}
    \gamma_k(x) = \biggl[ \sin \Bigl( \frac{2 \pi x}{\lambda_k } \Bigr), \, \cos \Bigl( \frac{2 \pi x}{\lambda_k} \Bigr) \biggr]^\textbf{T},
    \quad
    \lambda_k = \lambda_\text{max} \left( \frac{\lambda_\text{min}}{\lambda_\text{max}} \right)^{\tfrac{k-1}{L-1}},
    \quad
    k = 1, \dots, L
    \text{,}
    \label{eq:ff_scalar}
\end{equation*}
where $L$ is the number of frequency bands.
The Fourier-encoded point features are concatenated with an additional object type feature, describing the object to which each point belongs.
We use the PointMAE tokenizer~\cite{pointMAE} to get a patch embedding $\mathbf{s} \in \mathbb{R}^{d}$ from the point features of each patch, which is similar to PointNet \cite{ruizhongtai2017pointnet} and uses max-pooling to aggregate over the patch.
The point features of the patch center are passed through another \gls{mlp} to yield a patch positional embedding $\mathbf{c} \in \mathbb{R}^{d}$, which is then added to the patch embeddings to produce  the patch tokens $\mathbf{t} = \mathbf{s} + \mathbf{c}$.

The $M$ tokens $\{\mathbf{t}_1, \dots, \mathbf{t}_M\}$ are passed into a transformer encoder, which refines them via self-attention.
A final attention pooling layer compresses them into a single output token $\mathbf{r}_i$ for the point cloud sequence $\mathcal {P}_i$.
In contrast to simple mean or max pooling, the attention pooling layer uses a learned query token to attend over the transformer outputs, allowing the model to selectively focus on the most informative spatio-temporal regions.
We found that a lightweight architecture with a single transformer layer suffices and outperforms deeper variants, suggesting that the tokenizer already captures most of the relevant structure and only light global reasoning is needed on top.

\textbf{Auxiliary Losses.}
To encourage the latent representation to capture physically meaningful material properties and stabilize training, we introduce two auxiliary training objectives.
As in \gls{mango}, we supervise the trajectory-level latent material description $\mathbf{r}$ using ground-truth physical properties through a lightweight decoder $\hat{\rho} = \mathrm{MLP}(\mathbf{r})$.
The predicted $\hat{\rho}$ can be interpreted as an explicit system identification of the current task.

We additionally supervise the patch tokenizer using an \gls{sdf} objective inspired by~\citet{vosylius2024instantpolicyincontextimitation}. 
An \gls{sdf} \mbox{$f: \mathbb{R}^3 \to \mathbb{R}$} assigns to each point $\mathbf p$ its signed distance to the nearest mesh surface, with positive values outside and negative values inside. For 2D sheet-like geometries we treat the underside as interior. 
The distances are scaled and passed through a $\tanh$ nonlinearity to obtain normalized targets $s = \tanh(\alpha\, f(\mathbf p))$, which act similarly to truncation~\cite{curless1996volumetric} and smoothly approach an occupancy function~\cite{mescheder2019occupancy} as $\alpha{\rightarrow}\infty$. 
For each trajectory and time step, we sample query points %such that two thirds lie near object surfaces and the rest are drawn uniformly from the scene's bounding box, 
as visualized in the bottom right of Figure~\ref{fig:overview}. 
For each query point, we find its $K_{\text{SDF}}$ nearest patch tokens in 4D space-time and Fourier-encode their relative positions. 
These encodings are concatenated with the token embeddings and processed by an attention-based MLP that predicts a scalar weight per neighbor, biased by the normalized distance to emphasize spatio-temporal locality. 
A weighted aggregation of the features is decoded by a small MLP into a predicted signed distance, which is supervised against the ground-truth $s$ using a smooth $\ell_1$ loss.

\textbf{Data Augmentation.}
% Our framework can be directly applied without further training at inference time to real-world point cloud observations, enabling simulation of physical processes without access to ground-truth material parameters or mesh trajectories. Given a sequence of observed point clouds, the model infers a latent material representation and predicts the corresponding system dynamics.
%
% A requirement for deployment is the availability of an initial mesh state $\mathcal{G}_0$, which serves as the starting point for simulation. While this may appear restrictive, it aligns with standard practice in industry, where mesh-based representations are commonly available or can be obtained through established reconstruction pipelines. In this sense, our approach integrates naturally into existing simulation workflows while extending them with learned material inference from observations.
%
To bridge the gap between simulated training data and real-world sensor inputs, we improve the robustness of the point cloud encoder via data augmentation. 
In particular, we perturb simulated point cloud sequences during training to mimic typical artifacts observed in real-world data. 
These augmentations include multi-scale Gaussian noise to model sensor uncertainty and structured region dropout~\cite{cutout} to simulate occlusions or missing observations. 
To account for sensor failures and reconstruction errors, we additionally inject sporadic point artifacts by sampling a small number of points uniformly from the scene's bounding box.
These augmentations enable generalization to noisy and incomplete point cloud observations, facilitating reliable material inference and simulation in real-world scenarios.

\section{Experiments}
\label{sec:experiments}

% \begin{figure}[t]
%   \centering
%     % \makebox[\textwidth][c]{
%     %     \input{figures/quantitative_results/legend_realworld}
%     % }
%     \makebox[\textwidth][r]{%
%   \makebox[0.77\textwidth][c]{%
%     \input{figures/quantitative_results/legend_realworld}%
%   }%
% }
%   \begin{minipage}{0.22\textwidth}
%     \includegraphics[width=\textwidth, trim={0 30mm 0 100mm}, clip]{figures/qualitative_realworld_data/03.png}      
%   \end{minipage}\hfill
%   \begin{minipage}{0.44\textwidth}
%     \includegraphics[width=\textwidth]{figures/qualitative_realworld_data/real_world_qualitative2.pdf}    
%   \end{minipage}
%   \begin{minipage}{0.33\textwidth}
%         \centering
%         \scalebox{0.73}{\input{figures/quantitative_realworld_data/loss_over_time_per_method}}\\
%         % {\footnotesize\texttt{Trampoline (Real-world)}}
%     \end{minipage}
%     \hfill
% \caption{Real-world trampoline experiment. \textbf{Left:} Physical setup with a ball interacting with a latex sheet. \textbf{Center:} Qualitative comparison at $t=0.2\,\text{s}$ (maximal sheet elongation) between a model without context encoding and PEACH. The red point cloud shows the ground truth observations; note that the very bottom of the sheet is occluded by the ball and thus absent from the point cloud. \textbf{Right:} Point-to-mesh distance aggregated over all real-world trials, plotted over time, with ball-sheet contact occurring between $t=0.12\,\text{s}$ and $t=0.28\,\text{s}$.}
%   \label{fig:qualitative_realworld}
% \end{figure}

\begin{figure}[t]
  \centering

\makebox[\textwidth][l]{%
\scalebox{1.01}{
  \hspace*{0.015\textwidth}%
  \begin{minipage}[t]{0.25\textwidth}
    \vspace{2pt}
    \includegraphics[
      width=\textwidth,
      trim={0 30mm 0 100mm},
      clip
    ]{figures/qualitative_realworld_data/03.png}      
  \end{minipage}\hspace{0.030\textwidth}%
  \begin{minipage}[t]{0.68\textwidth}
    \vspace{0pt}

    % Legend next to the left image, not above the whole figure

    % \begin{tikzpicture}
%     \tikzstyle{every node}=[font=\scriptsize]
%     \input{figures/quantitative_results/tikzcolors}
%     \begin{axis}[%
%     hide axis,
%     xmin=10,
%     xmax=50,
%     ymin=0,
%     ymax=0.1,
%     legend style={
%         draw=white!15!black,
%         legend cell align=left,
%         legend columns=3,
%         legend style={
%             draw=none,
%             column sep=1ex,
%             line width=1pt,
%         }
%     },
%     ]

%     \addlegendimage{
%         line legend,
%         tabblue,
%         thick,
%         mark=square*,
%         mark options={line width=1.5pt, rotate=45}
%     }
%     \addlegendentry{\textbf{PEACH Encoder} (Ours)}
%     \addlegendimage{line legend, tabgreen, thick, mark=triangle*, mark options={line width=1.5pt, rotate=0}}
%     \addlegendentry{GNN Encoder}
%     \addlegendimage{
%         line legend,
%         taborange,
%         thick,
%         mark=square*,
%         mark options={line width=1.5pt, rotate=0}
%     }
%     \addlegendentry{PSTNet Encoder}
%     % New Line
%     % \addlegendimage{empty legend}
%     % \addlegendentry{}

%     \addlegendimage{line legend, tabolive, thick, mark=x, mark size=3pt, mark options={line width=1pt}}
%     \addlegendentry{Oracle Encoder}
%     \addlegendimage{line legend, tabcyan, thick, mark=+, mark size=2.5pt, mark options={line width=2pt}}
%     \addlegendentry{MaNGO Mesh Encoder}
%     \addlegendimage{line legend, tabgray, thick, mark=*}
%     \addlegendentry{No Context Encoder}    
%     \end{axis}
% \end{tikzpicture}

\begin{tikzpicture}
    \tikzstyle{every node}=[font=\scriptsize]
    \input{figures/quantitative_results/tikzcolors}

    \begin{axis}[%
        hide axis,
        xmin=10,
        xmax=50,
        ymin=0,
        ymax=0.1,
        legend style={
            draw=none,
            legend cell align=left,
            legend columns=3,
            /tikz/every odd column/.append style={column sep=1ex},
            /tikz/every even column/.append style={column sep=1.6ex},
        }
    ]

    % --- Bar-style legend entries ---
    \addlegendimage{area legend, fill=tabblue}
    \addlegendentry{PEACH (ours)}

    \addlegendimage{area legend, fill=tabpurple}
    \addlegendentry{PEACH (no data augm.)}

    \addlegendimage{area legend, fill=taborange}
    \addlegendentry{No Context (MGN)}

    \addlegendimage{area legend, fill=tabgray}
    \addlegendentry{No Context}
    
    \addlegendimage{area legend, fill=tabgreen}
    \addlegendentry{``Oracle``}
    \end{axis}
\end{tikzpicture}\\[-7mm]

    \hspace{0.000\textwidth}
    \begin{minipage}[t]{0.57\textwidth}
      \vspace{0pt}
      \includegraphics[
        width=\textwidth
      ]{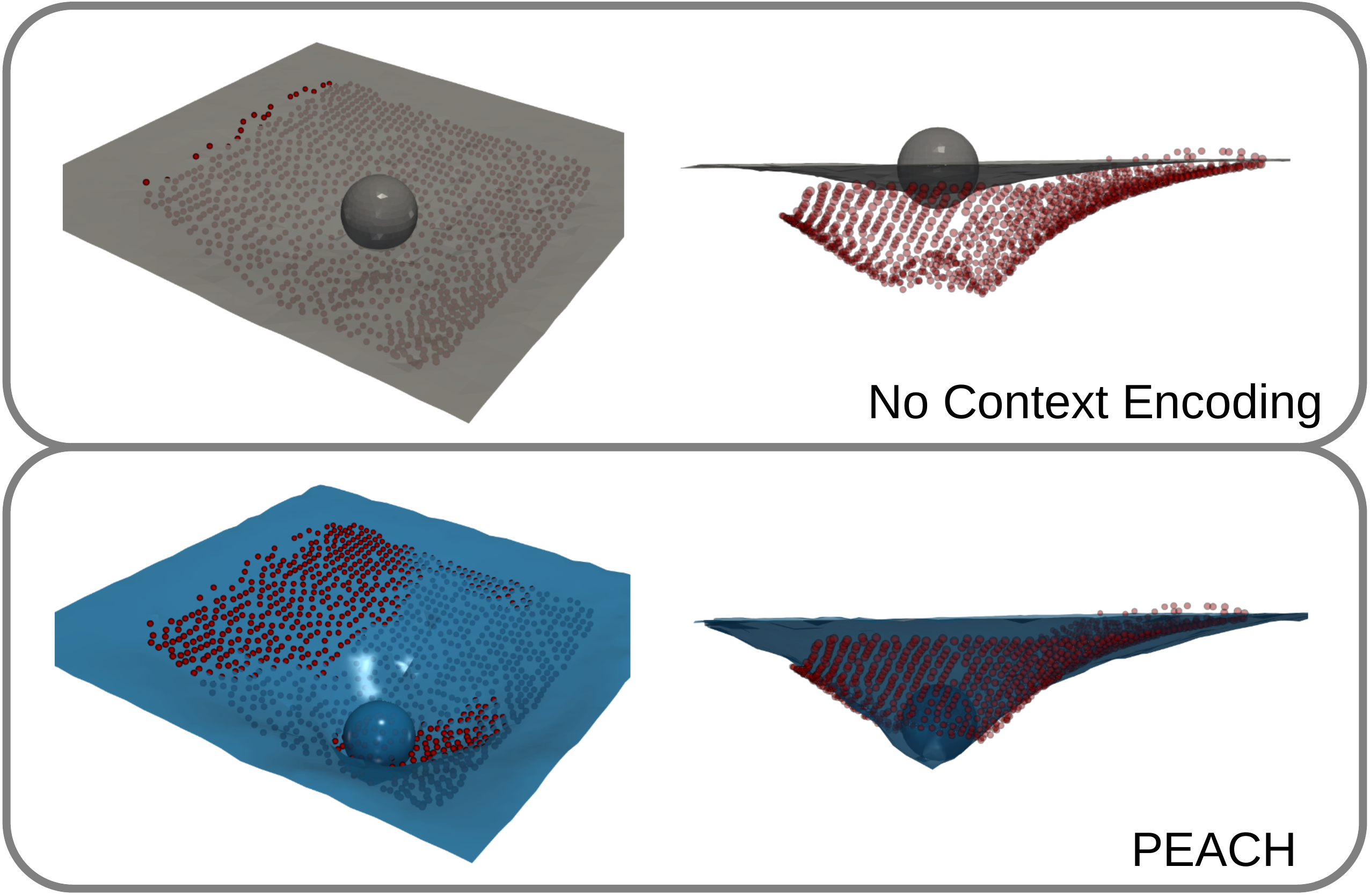}
    \end{minipage}\hspace{0.025\textwidth}%
      \hspace{-0pt}
    \begin{minipage}[t]{0.30\textwidth}
      \vspace{-10pt}
      \centering
      \scalebox{0.60}{%
        % This file was created with matplot2tikz v0.5.1.
\begin{tikzpicture}

\definecolor{darkgray}{RGB}{169,169,169}
\definecolor{darkorange25512714}{RGB}{255,127,14}
\definecolor{darkslategray38}{RGB}{38,38,38}
\definecolor{forestgreen4416044}{RGB}{44,160,44}
\definecolor{gray127}{RGB}{127,127,127}
\definecolor{lightgray204}{RGB}{204,204,204}
\definecolor{steelblue31119180}{RGB}{31,119,180}
\definecolor{mediumpurple148103189}{RGB}{148,103,189}

\begin{axis}[
    line width=0.8pt,
    axis line style={black},
    height=6.9cm,
    width=5.5cm,
    legend cell align={left},
    legend style={
      fill opacity=0.8,
      draw opacity=1,
      text opacity=1,
      at={(0.03,0.97)},
      anchor=north west,
      draw=lightgray204
    },
    % legend to name=distlegend,
    % legend columns=4,
    % legend cell align={left},
    % legend style={
    %   draw=lightgray204,
    %   fill=white,
    %   fill opacity=0.9,
    %   text opacity=1,
    %   font=\scriptsize,
    %   /tikz/every even column/.append style={column sep=5pt}
    % },
    tick align=outside,
    xlabel={Time [s]},
    xlabel style={font=\large, color=black},
    ylabel={Point-to-mesh distance [$\times10^{-3}$]},
    scaled y ticks=false,
    yticklabel={
      \pgfmathparse{\tick*1000}
      \pgfmathprintnumber[fixed,precision=1]{\pgfmathresult}
    },
    ylabel style={font=\large, color=black, xshift=-15pt,yshift=-1pt},
    tick label style={font=\small, color=black},
    xmajorgrids,
    xmajorticks=true,
    xtick pos=bottom,
    xmin=0, xmax=10,
    xtick={0,1,2,3,4,5,6,7,8,9,10},
    xticklabels={0, , 0.08, , 0.16, , 0.24, , 0.32, , 0.40},
    xtick style={color=black, line width=0.8pt},
    x grid style={lightgray!70},
    ymajorgrids,
    ymin=-0.00013, ymax=0.00169170054904424,
    ytick pos=left,
    ytick style={color=black, line width=0.8pt},
    y grid style={lightgray!70},
    mark size=1.5pt,
]
\path [draw=gray127, fill=gray127, opacity=0.3]
(axis cs:0,8.11894773903532e-06)
--(axis cs:0,5.12359331139578e-06)
--(axis cs:1,1.24465407043317e-05)
--(axis cs:2,3.70390871940742e-05)
--(axis cs:3,0.000170088983293226)
--(axis cs:4,0.000538726298991605)
--(axis cs:5,0.00111711371160709)
--(axis cs:6,0.0011539952494968)
--(axis cs:7,0.000493944165682478)
--(axis cs:8,0.000363235438599077)
--(axis cs:9,0.000430066142989745)
--(axis cs:10,0.000541865832110489)
--(axis cs:10,0.000579374625937817)
--(axis cs:10,0.000579374625937817)
--(axis cs:9,0.000462110629396193)
--(axis cs:8,0.000397154613974635)
--(axis cs:7,0.000502686903510039)
--(axis cs:6,0.00117717037701368)
--(axis cs:5,0.00120230794382223)
--(axis cs:4,0.000562686173348084)
--(axis cs:3,0.000175566404236633)
--(axis cs:2,3.94724731199858e-05)
--(axis cs:1,1.48924968993924e-05)
--(axis cs:0,8.11894773903532e-06)
--cycle;

\path [draw=darkorange25512714, fill=darkorange25512714, opacity=0.3]
(axis cs:0,5.40829057285919e-06)
--(axis cs:0,5.12708819030649e-06)
--(axis cs:1,1.26935416273e-05)
--(axis cs:2,3.87968943675787e-05)
--(axis cs:3,0.000167262485056199)
--(axis cs:4,0.00048637886969118)
--(axis cs:5,0.00101407638474484)
--(axis cs:6,0.00132488404211472)
--(axis cs:7,0.000618741954658617)
--(axis cs:8,0.00059531467425586)
--(axis cs:9,0.000698201175291615)
--(axis cs:10,0.000747446074487471)
--(axis cs:10,0.00161138177505563)
--(axis cs:10,0.00161138177505563)
--(axis cs:9,0.0013830343025802)
--(axis cs:8,0.00107759078622621)
--(axis cs:7,0.0010224056098923)
--(axis cs:6,0.00148174712423497)
--(axis cs:5,0.00157530098460938)
--(axis cs:4,0.000571288045344374)
--(axis cs:3,0.000169920463772542)
--(axis cs:2,4.35155251142305e-05)
--(axis cs:1,1.37517620123617e-05)
--(axis cs:0,5.40829057285919e-06)
--cycle;

\path [draw=forestgreen4416044, fill=forestgreen4416044, opacity=0.3]
(axis cs:0,5.02886968320126e-06)
--(axis cs:0,5.00629528346508e-06)
--(axis cs:1,1.22442673364276e-05)
--(axis cs:2,3.64980695599115e-05)
--(axis cs:3,0.000170152035236697)
--(axis cs:4,0.000532057652440017)
--(axis cs:5,0.000854251318701245)
--(axis cs:6,0.000737646454490459)
--(axis cs:7,0.00045989315263796)
--(axis cs:8,0.000471600066430256)
--(axis cs:9,0.000512676674770773)
--(axis cs:10,0.000583336630952544)
--(axis cs:10,0.00059101728056703)
--(axis cs:10,0.00059101728056703)
--(axis cs:9,0.000519406470630202)
--(axis cs:8,0.000481395700517169)
--(axis cs:7,0.00048684069643059)
--(axis cs:6,0.000778906577484122)
--(axis cs:5,0.000879407519557844)
--(axis cs:4,0.000544052049326638)
--(axis cs:3,0.000173569310650237)
--(axis cs:2,3.67153696726064e-05)
--(axis cs:1,1.23036071607885e-05)
--(axis cs:0,5.02886968320126e-06)
--cycle;

\path [draw=mediumpurple148103189, fill=mediumpurple148103189, opacity=0.3]
(axis cs:0,5.71805394145031e-06)
--(axis cs:0,5.01869094364338e-06)
--(axis cs:1,1.22791177000181e-05)
--(axis cs:2,3.65818932192497e-05)
--(axis cs:3,0.000167055152474461)
--(axis cs:4,0.000440917325113333)
--(axis cs:5,0.000484037372771127)
--(axis cs:6,0.00066896209913466)
--(axis cs:7,0.000405467094214259)
--(axis cs:8,0.000302540388506713)
--(axis cs:9,0.000360184482306067)
--(axis cs:10,0.000494188704960834)
--(axis cs:10,0.000520897211245028)
--(axis cs:10,0.000520897211245028)
--(axis cs:9,0.000368405658828124)
--(axis cs:8,0.000314538892644123)
--(axis cs:7,0.000422770715022125)
--(axis cs:6,0.000736019156793191)
--(axis cs:5,0.000525294678845967)
--(axis cs:4,0.000479866894897896)
--(axis cs:3,0.000170394556655538)
--(axis cs:2,3.7254243580378e-05)
--(axis cs:1,1.29903018734012e-05)
--(axis cs:0,5.71805394145031e-06)
--cycle;

\path [draw=steelblue31119180, fill=steelblue31119180, opacity=0.3]
(axis cs:0,5.32954746347514e-06)
--(axis cs:0,5.02559393680713e-06)
--(axis cs:1,1.23061595667195e-05)
--(axis cs:2,3.64593142171543e-05)
--(axis cs:3,0.000167017648384444)
--(axis cs:4,0.000454926606626032)
--(axis cs:5,0.000454583113850049)
--(axis cs:6,0.000523227875182783)
--(axis cs:7,0.000330028291073177)
--(axis cs:8,0.00029419582178889)
--(axis cs:9,0.000349462154376852)
--(axis cs:10,0.000491633596502652)
--(axis cs:10,0.00054824294982609)
--(axis cs:10,0.00054824294982609)
--(axis cs:9,0.000401096724481249)
--(axis cs:8,0.000327221116303917)
--(axis cs:7,0.000380109390453072)
--(axis cs:6,0.000647119674776377)
--(axis cs:5,0.000497473443410854)
--(axis cs:4,0.000475108261580317)
--(axis cs:3,0.000171479885157169)
--(axis cs:2,3.71906361107221e-05)
--(axis cs:1,1.26398031545705e-05)
--(axis cs:0,5.32954746347514e-06)
--cycle;

\addplot [very thick, gray127, mark=none, forget plot]
table {%
0 6.16846661046111e-06
1 1.33566141312258e-05
2 3.79323264291998e-05
3 0.000172875483426651
4 0.000550706236169845
5 0.00116767553210593
6 0.00116614555628075
7 0.000498315534596259
8 0.000380195026286856
9 0.000446161982363265
10 0.00056057571105157
};
\addplot [very thick, darkorange25512714, mark=none, forget plot]
table {%
0 5.25982570195538e-06
1 1.32226518198308e-05
2 4.09810629054164e-05
3 0.000168443404438676
4 0.000525215963987193
5 0.00126655838633724
6 0.00138342683052542
7 0.000767640316489633
8 0.000780909799859728
9 0.00096147732127065
10 0.00109070352106755
};
\addplot [very thick, forestgreen4416044, mark=none, forget plot]
table {%
0 5.01956598668585e-06
1 1.2269513994454e-05
2 3.66266437509921e-05
3 0.000171860672943467
4 0.000536364111319472
5 0.000864933194968671
6 0.000756467673586485
7 0.000473366924534275
8 0.000477138728074351
9 0.000516041572700487
10 0.000587167208759638
};
\addplot [very thick, mediumpurple148103189, mark=none, forget plot]
table {%
0 5.35674467698755e-06
1 1.26106139231297e-05
2 3.68227403703258e-05
3 0.000169110989764931
4 0.000464556738356805
5 0.000504666025808547
6 0.000708642764411707
7 0.000413382586743865
8 0.000308661070916969
9 0.000364295070567096
10 0.000507542958102931
};
\addplot [very thick, steelblue31119180, mark=none, forget plot]
table {%
0 5.16803246910058e-06
1 1.24482370745227e-05
2 3.68488107579878e-05
3 0.000169248766770806
4 0.000465017434103174
5 0.0004779231458906
6 0.000585161942171908
7 0.000353386584361033
8 0.000307302393166537
9 0.000375522828585417
10 0.000519938273164371
};

\draw [red, thick] (axis cs:3, 0.00003) -- (axis cs:7, 0.00003);
\draw [red, thick] (axis cs:3, -0.0000) -- (axis cs:3,  0.00006);
\draw [red, thick] (axis cs:7, -0.0000) -- (axis cs:7,  0.00006);
\node [red, font=\small] at (axis cs:5, -0.000055) {Contact Phase};
\end{axis}

\end{tikzpicture}%
      }
    \end{minipage}
  \end{minipage}%
}
}

    \caption{
        Real-world \texttt{Trampoline} scene.
        \textbf{Left:} Physical setup with a ball bouncing on a latex sheet.
        \textbf{Center:} Qualitative comparison at $t{=}0.2\,\text{s}$ (maximal sheet stretching) between \gls{peach} and a context-free simulator.
        The \textcolor{red}{red} point cloud shows the ground truth observations. The  bottom of the sheet is occluded by the ball and thus absent from the point cloud.
        \textbf{Right:} Mean point-to-mesh distance over all real-world trials, plotted over time, with ball-sheet contact occurring in $[0.12,0.28]\, \text{s}$.
    }
    \vspace{-0.4cm}
    \label{fig:qualitative_realworld}
\end{figure}

\textbf{Scenes.}
% We evaluate \gls{peach} on its ability to provide latent material descriptions from raw point cloud sequences to facilitate downstream simulation of a canonical mesh.
% We consider four simulated environments, namely \texttt{Deforming Block}, \texttt{Sheet Deformation}~\citep{dahlinger2025mango}\footnote{Respectively called \texttt{Deformable Plate} (\textit{easy}) and \texttt{Planar Bending} in \citet{dahlinger2025mango}.}, \texttt{Bending Beam} and \texttt{Trampoline}, as well as one real-world task \texttt{Trampoline Real World}.
% We evaluate \gls{peach} on inferring latent material representations from raw point cloud sequences for downstream simulation on a canonical mesh.
% Our benchmark comprises four simulated and one real-world environment.
% We adopt two existing datasets, \texttt{Deforming Block} and \texttt{Sheet Deformation}~\citep{dahlinger2025mango}\footnote{Called \texttt{Deformable Plate} (\textit{easy}) and \texttt{Planar Bending}, respectively, in \citet{dahlinger2025mango}.}, and introduce two novel simulated environments, \texttt{Bending Beam} and \texttt{Trampoline (Sim)}, as well as one real-world environment, \texttt{Trampoline (Real World)}.
We evaluate the sim-to-real capability of \gls{peach} on a novel scene called \texttt{Trampoline} that involves dynamic motion, complex contact interactions, and deformable objects.
Additionally, we introduce the novel simulation-only dataset \texttt{Bending Beam} inspired by \citet{wurth2025diffusion}, and adopt two existing datasets, \texttt{Deforming Block} and \texttt{Sheet Deformation}~\citep{dahlinger2025mango}\footnote{Called \texttt{Deformable Plate} (\textit{easy}) and \texttt{Planar Bending}, respectively, in \citet{dahlinger2025mango}.}.
These scenes span 2D and 3D domains, and are inspired by simulation-assisted process optimization in industrial contexts.
They feature varying physical parameters such as Young's modulus (all except \texttt{Deforming Block}), Poisson's ratio (all except \texttt{Sheet Deformation}), membrane thickness, shear relaxation ratio (\texttt{Trampoline}), and relaxation time (\texttt{Bending Beam} and \texttt{Trampoline}).
Rollout length varies by scene from $25$ (\texttt{Trampoline}) to $100$ (\texttt{Bending Beam}).
Although the \texttt{Trampoline} scene has a shorter horizon, it has the most complex geometry, with roughly $10$ times more triangles than other scenes.
A visual overview of the scenes is provided in Figure~\ref{fig:task_overview}, with further details in Appendix~\ref{app:dataset}.
%trampoline: only 25 steps (orig. 50, but memory)
% but lots of triangles, 10x more then the others
% bb: 100 timesteps , so long in time,
% but: small geometry, does not work well but likely due to the simulator: need hierarchical GNS
% We had data of the other tasks (mango)
% showcase simple singular material properties (you can underestand/guess the mat properties by looking at the resutls)

In simulation, we obtain per-step point clouds by ray casting from a static camera and applying \gls{fps}~\citep{ruizhongtai2017pointnet} to retain $512$ points; due to the finer geometry \texttt{Trampoline} uses $1024$ points.
% To reduce the point cloud sequence length, we apply a temporal stride of $2$, using every other simulation step.
To reduce the point cloud sequence length, we drop every $2$nd time step. 
In \texttt{Trampoline} and \texttt{Deforming Block}, collider points are explicitly marked.
% Unlike mesh trajectories, these point clouds lack temporal correspondences, making the setting substantially more challenging.
% In addition to simulated evaluation, we assess \gls{peach} on real-world point cloud observations in \texttt{Trampoline (Real World)}.
% This environment consists of real trampoline recordings without ground-truth material parameters or mesh trajectories, where a robot drops balls of varying density onto natural latex sheets of different thicknesses.
% We record the point clouds using a ZED Mini Stereo Camera (Stereolabs). 
% Since ground-truth meshes are unavailable, we evaluate rollouts using an asymmetric point-to-face Chamfer distance, computed as the squared distance from each observed point to the closest point on the predicted mesh surface.
% We report this metric over time to analyze when prediction errors occur.
For the real world \texttt{Trampoline} dataset, we use a robot to drop balls of varying density onto latex sheets of different thicknesses and record depth images with a ZED Mini Stereo Camera (Stereolabs).
Neither meshes nor ground truth physical parameters are known for these recordings.
For the ``\textit{Oracle}'' baseline, measurable material properties are determined experimentally  while the remaining parameters are chosen based on informed estimates from literature values.
Further information on the real world data is in Appendix~\ref{app:real}.

% For all four environments, we generate point clouds for each simulation step via raycasting from a static, front-facing camera followed by \gls{fps}~\citep{eldar1997farthest} to retain $512$ points per time step.
% For \texttt{Deforming Block}, points belonging to the collider are explicitly marked.
% The lack of temporal point correspondences in these sequences significantly increases task difficulty compared to mesh-based inputs. More information about the environments is given in Appendix \ref{app:dataset}.

\textbf{Baselines.}
We compare \gls{peach} against methods using $4$ different context types: point cloud contexts, mesh contexts, oracle, and no context.
Methods using point cloud contexts differ only in the encoder architecture used to encode point cloud sequences.
\textit{PSTNet Encoder}~\citep{fan2021pstnet} captures local geometric evolution via space-time tubes anchored at \gls{fps}-sampled reference points, applying decoupled spatial and temporal convolutions hierarchically across the point cloud sequence.
\textit{GNN Encoder} adapts the graph network paradigm from mesh-based simulation to point cloud sequences.
It tokenizes each timestep's point cloud into local patch embeddings, constructs a spatio-temporal graph connecting tokens within and across adjacent frames, and processes it via message passing with a final readout into a trajectory-level latent code.

\gls{mango}~\citep{dahlinger2025mango} instead computes the latent embedding using privileged mesh contexts.
Since such contexts are not readily available in real-world settings, \gls{mango} is not directly comparable to our method, but demonstrates the impact of replacing explicit mesh observations with point clouds in simulated tasks.
\textit{Oracle} methods receive the ground truth physical properties at every time step as additional node features.
\textit{No Context} methods predict trajectories solely from the initial state without additional task information.
All baselines that predict a latent embedding also receive auxiliary supervision via ground truth physical properties.

Lastly, we study the impact of the simulator component by replacing the MaNGO simulator with the \gls{mgn} simulator~\cite{pfaff2020learning}, a mesh-based autoregressive simulator trained for one-step prediction.
We instantiate the \gls{mgn} baseline in \textit{Oracle (\gls{mgn})} and \textit{No Context (\gls{mgn})} variants.
While it is possible to integrate \gls{mgn} into the \gls{peach} framework, early experiments indicated that it was outperformed by the MaNGO simulator.
Appendix~\ref{app:methods} provides further details on baselines.

\begin{figure}[t]
    \makebox[\textwidth][c]{
        \begin{tikzpicture}
    \tikzstyle{every node}=[font=\scriptsize]
    \input{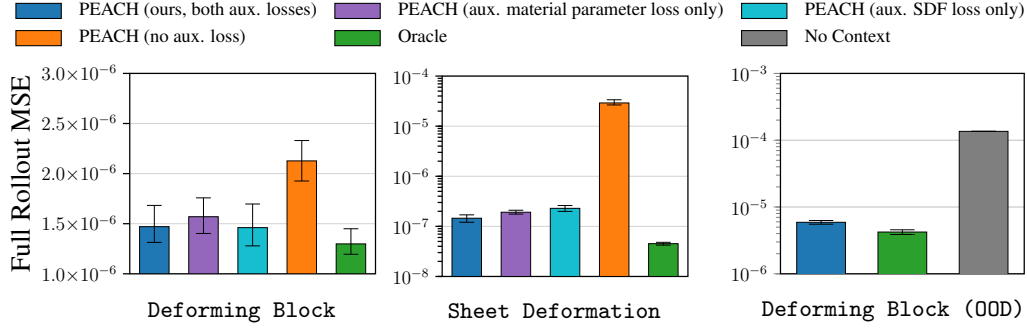}

    \begin{axis}[%
        hide axis,
        xmin=10,
        xmax=50,
        ymin=0,
        ymax=0.1,
        legend style={
            draw=none,
            legend cell align=left,
            legend columns=3,
            column sep=1ex,
        }
    ]

    % --- Bar-style legend entries ---
    \addlegendimage{area legend, fill=tabblue}
    \addlegendentry{PEACH (ours, both aux. losses)}

    \addlegendimage{area legend, fill=tabpurple}
    \addlegendentry{PEACH (aux. material parameter loss only)}

    \addlegendimage{area legend, fill=tabcyan}
    \addlegendentry{PEACH (aux. SDF loss only)}

    \addlegendimage{area legend, fill=taborange}
    \addlegendentry{PEACH (no aux. loss)}

    \addlegendimage{area legend, fill=tabgreen}
    \addlegendentry{Oracle}

    \addlegendimage{area legend, fill=tabgray}
    \addlegendentry{No Context}

    \end{axis}
\end{tikzpicture}
    }
  \centering
     \begin{minipage}{0.32\textwidth}
        \scalebox{0.6}{\begin{tikzpicture}
\definecolor{darkgray}{RGB}{169,169,169}
\definecolor{lightgray204}{RGB}{204,204,204}
\definecolor{tabblue}{RGB}{31,119,180}
\definecolor{taborange}{RGB}{255,127,14}
\definecolor{tabpurple}{RGB}{148,103,189}
\definecolor{tabred}{RGB}{214,39,40}
\definecolor{tabcyan}{RGB}{23,190,207}
\definecolor{tabgreen}{RGB}{44,160,44}

\begin{axis}[
    line width=0.8pt,
    axis line style={black},
    height=6cm,
    width=7cm,
    scaled y ticks=false,
    xmin=-0.5, xmax=4.5,
    xtick=\empty,
    ymin=1.0e-6, ymax=3.0e-6,
    ymajorgrids,
    y grid style={lightgray!70},
    tick align=outside,
    ylabel={Full Rollout MSE},
    ylabel style={font=\LARGE, color=black},
    tick label style={font=\large, color=black},
    ytick style={color=black, line width=0.8pt},
    ytick={1.0e-6, 1.5e-6, 2.0e-6, 2.5e-6, 3.0e-6},
    yticklabels={
        \(1.0{\times}10^{-6}\),
        \(1.5{\times}10^{-6}\),
        \(2.0{\times}10^{-6}\),
        \(2.5{\times}10^{-6}\),
        \(3.0{\times}10^{-6}\)
    },
    ytick pos=left,
]

% 0: pointpatch (tab:blue)
\draw[fill=tabblue, draw=black, line width=0.5pt]
    (axis cs:-0.3, 1e-6) rectangle (axis cs:0.3, 1.4707e-6);
% 1: peach_only_ml_and_matprop (tab:purple)
\draw[fill=tabpurple, draw=black, line width=0.5pt]
    (axis cs:0.7, 1e-6) rectangle (axis cs:1.3, 1.5698e-6);
% 2: peach_only_ml_and_sdf (tab:red)
\draw[fill=tabcyan, draw=black, line width=0.5pt]
    (axis cs:1.7, 1e-6) rectangle (axis cs:2.3, 1.4608e-6);
% 3: peach_only_ml (tab:orange)
\draw[fill=taborange, draw=black, line width=0.5pt]
    (axis cs:2.7, 1e-6) rectangle (axis cs:3.3, 2.1269e-6);
% 4: oracle (tab:green)
\draw[fill=tabgreen, draw=black, line width=0.5pt]
    (axis cs:3.7, 1e-6) rectangle (axis cs:4.3, 1.2985e-6);

% Error bars
% 0: pointpatch
\draw[black, semithick] (axis cs:0, 1.3136e-6) -- (axis cs:0, 1.6826e-6);
\draw[black, semithick] (axis cs:-0.15, 1.3136e-6) -- (axis cs:0.15, 1.3136e-6);
\draw[black, semithick] (axis cs:-0.15, 1.6826e-6) -- (axis cs:0.15, 1.6826e-6);
% 1: peach_only_ml_and_matprop
\draw[black, semithick] (axis cs:1, 1.4025e-6) -- (axis cs:1, 1.7580e-6);
\draw[black, semithick] (axis cs:0.85, 1.4025e-6) -- (axis cs:1.15, 1.4025e-6);
\draw[black, semithick] (axis cs:0.85, 1.7580e-6) -- (axis cs:1.15, 1.7580e-6);
% 2: peach_only_ml_and_sdf
\draw[black, semithick] (axis cs:2, 1.2787e-6) -- (axis cs:2, 1.6967e-6);
\draw[black, semithick] (axis cs:1.85, 1.2787e-6) -- (axis cs:2.15, 1.2787e-6);
\draw[black, semithick] (axis cs:1.85, 1.6967e-6) -- (axis cs:2.15, 1.6967e-6);
% 3: peach_only_ml
\draw[black, semithick] (axis cs:3, 1.9261e-6) -- (axis cs:3, 2.3300e-6);
\draw[black, semithick] (axis cs:2.85, 1.9261e-6) -- (axis cs:3.15, 1.9261e-6);
\draw[black, semithick] (axis cs:2.85, 2.3300e-6) -- (axis cs:3.15, 2.3300e-6);
% 4: oracle
\draw[black, semithick] (axis cs:4, 1.1949e-6) -- (axis cs:4, 1.4497e-6);
\draw[black, semithick] (axis cs:3.85, 1.1949e-6) -- (axis cs:4.15, 1.1949e-6);
\draw[black, semithick] (axis cs:3.85, 1.4497e-6) -- (axis cs:4.15, 1.4497e-6);

\end{axis}
\end{tikzpicture}}\\
        {\small \hspace*{18mm} \texttt{Deforming Block}}
     \end{minipage}
     \hspace{4mm}
     \begin{minipage}{0.32\textwidth}
        \scalebox{0.6}{\begin{tikzpicture}
\definecolor{lightgray204}{RGB}{204,204,204}
\definecolor{tabblue}{RGB}{31,119,180}
\definecolor{taborange}{RGB}{255,127,14}
\definecolor{tabpurple}{RGB}{148,103,189}
\definecolor{tabred}{RGB}{214,39,40}
\definecolor{tabcyan}{RGB}{23,190,207}
\definecolor{tabgreen}{RGB}{44,160,44}

\begin{semilogyaxis}[
    line width=0.8pt,
    axis line style={black},
    height=6cm,
    width=7cm,
    xmin=-0.5, xmax=4.5,
    xtick=\empty,
    ymin=1e-8, ymax=1e-4,
    ymajorgrids,
    y grid style={lightgray!70},
    tick align=outside,
    ylabel style={font=\LARGE, color=black},
    tick label style={font=\large, color=black},
    ytick style={color=black, line width=0.8pt},
    ytick={1e-8, 1e-7, 1e-6, 1e-5, 1e-4},
    yticklabels={
        \(10^{-8}\),
        \(10^{-7}\),
        \(10^{-6}\),
        \(10^{-5}\),
        \(10^{-4}\)
    },
    ytick pos=left,
]

% Bars (bottom anchored at ymin=1e-8)
% 0: pointpatch (tab:blue)
\draw[fill=tabblue, draw=black, line width=0.5pt]
    (axis cs:-0.3, 1e-8) rectangle (axis cs:0.3, 1.4541e-7);
% 1: peach_only_ml_and_matprop (tab:purple)
\draw[fill=tabpurple, draw=black, line width=0.5pt]
    (axis cs:0.7, 1e-8) rectangle (axis cs:1.3, 1.9139e-7);
% 2: peach_only_ml_and_sdf (tab:red)
\draw[fill=tabcyan, draw=black, line width=0.5pt]
    (axis cs:1.7, 1e-8) rectangle (axis cs:2.3, 2.2891e-7);
% 3: peach_only_ml (tab:orange)
\draw[fill=taborange, draw=black, line width=0.5pt]
    (axis cs:2.7, 1e-8) rectangle (axis cs:3.3, 2.9077e-5);
% 4: oracle (tab:green)
\draw[fill=tabgreen, draw=black, line width=0.5pt]
    (axis cs:3.7, 1e-8) rectangle (axis cs:4.3, 4.5173e-8);

% Error bars
% 0: pointpatch
\draw[black, semithick] (axis cs:0, 1.2069e-7) -- (axis cs:0, 1.6982e-7);
\draw[black, semithick] (axis cs:-0.15, 1.2069e-7) -- (axis cs:0.15, 1.2069e-7);
\draw[black, semithick] (axis cs:-0.15, 1.6982e-7) -- (axis cs:0.15, 1.6982e-7);
% 1: peach_only_ml_and_matprop
\draw[black, semithick] (axis cs:1, 1.7561e-7) -- (axis cs:1, 2.0982e-7);
\draw[black, semithick] (axis cs:0.85, 1.7561e-7) -- (axis cs:1.15, 1.7561e-7);
\draw[black, semithick] (axis cs:0.85, 2.0982e-7) -- (axis cs:1.15, 2.0982e-7);
% 2: peach_only_ml_and_sdf
\draw[black, semithick] (axis cs:2, 1.9907e-7) -- (axis cs:2, 2.6067e-7);
\draw[black, semithick] (axis cs:1.85, 1.9907e-7) -- (axis cs:2.15, 1.9907e-7);
\draw[black, semithick] (axis cs:1.85, 2.6067e-7) -- (axis cs:2.15, 2.6067e-7);
% 3: peach_only_ml
\draw[black, semithick] (axis cs:3, 2.6516e-5) -- (axis cs:3, 3.3661e-5);
\draw[black, semithick] (axis cs:2.85, 2.6516e-5) -- (axis cs:3.15, 2.6516e-5);
\draw[black, semithick] (axis cs:2.85, 3.3661e-5) -- (axis cs:3.15, 3.3661e-5);
% 4: oracle
\draw[black, semithick] (axis cs:4, 4.2259e-8) -- (axis cs:4, 4.8071e-8);
\draw[black, semithick] (axis cs:3.85, 4.2259e-8) -- (axis cs:4.15, 4.2259e-8);
\draw[black, semithick] (axis cs:3.85, 4.8071e-8) -- (axis cs:4.15, 4.8071e-8);

\end{semilogyaxis}
\end{tikzpicture}}\\
        {\small \hspace*{8mm}\texttt{Sheet Deformation}}
     \end{minipage}
     \hspace{-4mm}
    \begin{minipage}{0.32\textwidth}
        \centering
        \scalebox{0.6}{\begin{tikzpicture}
\definecolor{darkgray}{RGB}{169,169,169}
\definecolor{darkorange25512714}{RGB}{255,127,14}
\definecolor{darkslategray38}{RGB}{38,38,38}
\definecolor{darkturquoise23190207}{RGB}{23,190,207}
\definecolor{gray127}{RGB}{127,127,127}
\definecolor{lightgray204}{RGB}{204,204,204}
\definecolor{mediumpurple148103189}{RGB}{148,103,189}
\definecolor{orchid227119194}{RGB}{227,119,194}
\definecolor{sienna1408675}{RGB}{140,86,75}
\definecolor{steelblue31119180}{RGB}{31,119,180}
\definecolor{steelblue76114176}{RGB}{76,114,176}
\definecolor{tabgreen}{RGB}{44,160,44}

\begin{semilogyaxis}[
  line width=0.8pt,
  axis line style={black},
  xmin=-0.5, xmax=2.5,
  xtick=\empty,
  ymin=1e-6, ymax=1e-3,
  ymajorgrids,
  height=6cm,
  ytick pos=left,
  % width=7cm,
  % x grid style={lightgray204},
  % y grid style={lightgray204},
  % axis line style={darkgray},
  tick align=outside,
  ylabel style={font=\LARGE},
  tick label style={font=\large},
]

% Bars (bottom anchored at ymin=1e-7)
% 0: pointpatch (tab:blue)
\draw[fill=steelblue31119180, draw=black, line width=0.5pt] (axis cs:-0.3, 1e-7) rectangle (axis cs:0.3, 0.00000590559102420229);
% 1: oracle (tab:green)
\draw[fill=tabgreen, draw=black, line width=0.5pt] (axis cs:0.7, 1e-7) rectangle (axis cs:1.3, 0.00000421955195406554);
% 3: dummy (tab:gray)
\draw[fill=gray127, draw=black, line width=0.5pt] (axis cs:1.7, 1e-7) rectangle (axis cs:2.3, 0.000135723821585998);

% Error bars
% 0: pointpatch
\draw[black, semithick] (axis cs:0, 0.00000553021868654469
) -- (axis cs:0, 0.0000062809633618599);
\draw[black, semithick] (axis cs:-0.15, 0.00000553021868654469) -- (axis cs:0.15, 0.00000553021868654469);
\draw[black, semithick] (axis cs:-0.15, 0.0000062809633618599) -- (axis cs:0.15, 0.0000062809633618599);
% 1: oracle
\draw[black, semithick] (axis cs:1, 0.00000387946136015671) -- (axis cs:1, 0.00000455964254797436);
\draw[black, semithick] (axis cs:0.85, 0.00000387946136015671) -- (axis cs:1.15, 0.00000387946136015671);
\draw[black, semithick] (axis cs:0.85, 0.00000455964254797436) -- (axis cs:1.15, 0.00000455964254797436);
% 3: dummy
\draw[black, semithick] (axis cs:2, 0.000135091322590597) -- (axis cs:2, 0.000136356320581399);
\draw[black, semithick] (axis cs:1.85, 0.000135091322590597) -- (axis cs:2.15, 0.000135091322590597);
\draw[black, semithick] (axis cs:1.85, 0.000136356320581399) -- (axis cs:2.15, 0.000136356320581399);

\end{semilogyaxis}
\end{tikzpicture}}\\
        {\small \hspace*{4mm}\texttt{Deforming Block (OOD)}}
        
    \end{minipage}
  \caption{  \textbf{Left, Center:} Ablation study of auxiliary losses in PEACH on the
  \texttt{Deforming Block} and \texttt{Sheet Deformation} environments, respectively.
  \textbf{Right:} Performance of PEACH on out-of-distribution test tasks for
  \texttt{Deforming Block}.}
  \vspace{-0.4cm}
  \label{fig:generalizations_ablations}
\end{figure}

%PCA latent space visualization of all the Deforming Block dataset. Each dot is an encoded task, the color is its ground truth Poisson's ratio. Latent space visualizations of all further tasks are given in Appendix \todo{ref}.

\textbf{Training.}
We train each method end-to-end. % and evaluate simulation fidelity and material prediction accuracy using \gls{mse} against the full trajectory and ground-truth material vector, respectively.
All in-context methods are trained using batches of $1$ to $8$ randomly sampled context trajectories to improve adaptation capability.
We repeat each experiment across $4$ random seeds using the AdamW optimizer~\citep{adamw} with a learning rate of $5.0{\times}10^{-5}$.
All methods were trained on either an A100 or H100 GPU for $2$ days, with the exception of \texttt{Bending Beam}, which converged after $1$ day.
Appendix~\ref{app:hp} lists hyperparameters and Appendix~\ref{sec:runtime} provides runtime and parameter counts.

% We train each method end-to-end, evaluating simulation fidelity and material prediction accuracy using \gls{mse} against the full trajectory and ground truth vector, respectively.
% For the trajectory evaluation, we also include an upper bound \textit{Oracle Encoder}, which receives ground-truth material properties.
% We similarly include a \textit{No Context Encoder} with an empty context set to establish a lower bound representing regression toward mean material behavior.
% All models are trained using batches of $1$ to $8$ randomly sampled trajectories to improve adaptation capability.
% We repeat each experiment across $3$ random seeds using the AdamW optimizer~\citep{adamw} with a learning rate of $5.0{\times}10^{-5}$.
% Appendix~\ref{app:methods} provides an overview of used methods and lists detailed hyperparameters.

\section{Results}
\label{sec:results}

\textbf{Simulation.}
We evaluate the simulation fidelity using the \gls{mse} averaged across all steps in a complete rollout.
Evaluation uses an in-distribution test split; both the physical properties as well as the initial conditions used for testing are unseen during training.
Figure \ref{fig:main_results} shows that on simulation data, \gls{peach} consistently outperforms all point cloud context methods by a substantial margin.
Furthermore, \gls{peach} outperforms the mesh-based \gls{mango} on three out of four environments, despite \gls{mango} operating on privileged mesh context.
We attribute \gls{mango}'s underperformance to the relatively shallow \gls{cnn} and Deep Sets architecture of its encoder, which limits its representational capacity.
% Increasing model size would likely worsen the overfitting to training material properties we already observe.
On \texttt{Bending Beam}, \gls{peach} surpasses even the \textit{Oracle} baseline.
We attribute this to the expressiveness of the latent embedding of physical properties, which can jointly encode geometric properties such as beam thickness alongside scalar material parameters, providing the simulator with a richer conditioning signal than the explicit oracle features alone.
We note that replacing the simulator with the step-based \gls{mgn} generally reduces accuracy, even with oracle information. 
Figure \ref{fig:task_overview} shows qualitative predictions of \gls{peach} across all four scenes, demonstrating accurate simulation of diverse physical behaviors over long time horizons.
Additional qualitative results for all simulated tasks are provided in Appendix~\ref{app:vis}.

\textbf{Real World.}
Figure~\ref{fig:realworld_setup} shows an example trajectory of our real-world \texttt{Trampoline} setup. 
A robot arm releases a ball of unknown size and weight above an elastic membrane of varying thickness.
Figure~\ref{fig:qualitative_realworld} demonstrates that \gls{peach} is able to adapt to contexts from this real world setup without ever having seen real-world point clouds during training.
Qualitatively, \gls{peach} adapts to the interaction dynamics determined by the ball's size and weight and the rubber sheet's thickness, accurately capturing the sheet deformation at peak elongation.
In contrast, the \textit{No Context Encoding} baseline regresses to the mean, predicting only minimal deformation regardless of the physical parameters.

Quantitatively, \gls{peach} outperforms all baselines, including the \textit{Oracle} baseline where material parameters were measured directly from the physical setup.
All methods perform well prior to ball-sheet contact, having learned to accurately predict free-fall dynamics, but baselines deteriorate significantly during the contact phase. 
Toward the end of the trajectory, prediction error increases across all methods, which we attribute to the inherently chaotic post-bounce behavior of the ball, whose direction after impact is difficult to predict deterministically.
Combined, these results indicate our method's ability to accurately simulate real-world scenes from only point cloud observations.

\textbf{Parameter Study.}
We test the extrapolation capability of \gls{peach} using an additional, out-of-distribution set of \texttt{Deforming Block} tasks in simulation. As illustrated in Figure~\ref{fig:ood_split} of the Appendix, test tasks are drawn from the tails of the Poisson ratio distribution, ensuring that the evaluated material properties lie outside the training range.
% , where the Poisson's ratio values for test tasks are sampled outside the training distribution.
% Figure \ref{fig:generalizations_ablations} (right) further shows the out-of-distribution evaluation on \texttt{Deforming Block (OOD)}, where test tasks have Poisson's ratio values outside the training regime.
Figure \ref{fig:generalizations_ablations} (right) shows that \gls{peach} remains close to the oracle performance even under this distribution shift, demonstrating that the learned latent material representation generalizes beyond the training distribution.
A general analysis of performance as a function of context size is provided in Figure~\ref{fig:context_mse} of the Appendix~\ref{app:quant}.

\begin{wrapfigure}{r}{0.4\textwidth}
    \vspace{-0.4cm}
    \centering
    \includegraphics[width=\linewidth]{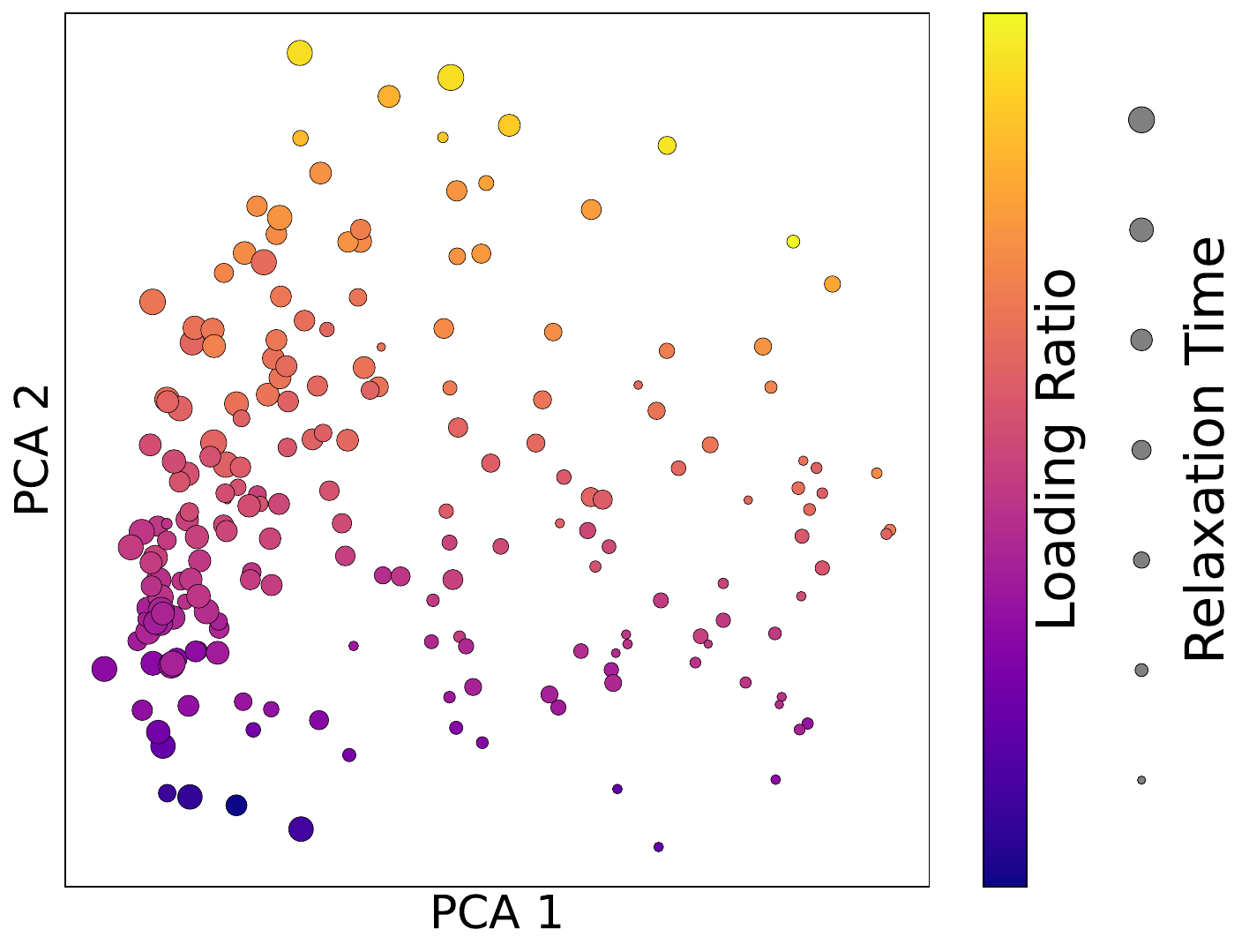}
\caption{
    Visualization of the first and second PCA component of the latent space on the \texttt{Trampoline} dataset.
    Each point represents one simulation test task.
    Color corresponds to ratio of mass to product of Young's modulus and sheet thickness, i.e., loading ratio, which governs the degree of sheet deformation.
    Size corresponds to relaxation time.
    }
    \vspace{-0.4cm}
    \label{fig:latent_space_vis_main}
\end{wrapfigure}

\textbf{Ablations.}
Figure~\ref{fig:generalizations_ablations} (left, center) presents an ablation of the auxiliary losses on \texttt{Deforming Block} and \texttt{Sheet Deformation}.
Removing both auxiliary losses leads to a substantial performance drop, confirming that auxiliary supervision is critical for learning a useful latent material representation.
% While \gls{peach} benefits from combining both losses, each one already delivers a substantial performance boost in isolation.
% Notably, training with only the SDF auxiliary loss remains competitive.
While \gls{peach} benefits from combining both losses, training with only the SDF auxiliary loss remains competitive.
Since \gls{sdf} labels can be computed purely from mesh geometry, this is a practical option for settings where ground truth physical properties are unavailable.
Figure~\ref{fig:qualitative_realworld} (right) contains an ablation of our method without data augmentation.
Simulation accuracy is somewhat reduced in the post-bounce phase before falling again, indicating that data augmentation provides a small but meaningful improvement.

\textbf{Latent Space Analysis.}
Figure~\ref{fig:latent_space_vis_main} visualizes the PCA projection of the latent material codes encoded by \gls{peach} on the \texttt{Trampoline} dataset.
Despite no explicit structure being imposed on the latent space, the projection reveals a clear organization: the first principal component separates tasks by relaxation time, while the second separates them by loading ratio.
This emergent disentanglement of physically meaningful material properties suggests that \gls{peach} learns structured representations purely from point cloud observations. Latent space visualizations for all environments are provided in Figure~\ref{fig:latents_appendix} of the Appendix.

\section{Conclusion}
\label{sec:conclusion}

We present \glsf{peach}, an in-context framework that conditions a graph network simulator on a handful of point cloud sequences to predict mesh-based dynamics for unseen materials.
% Relying only on point cloud contexts further allows \gls{peach} to simulate real-world scenes without requiring either explicit parameter estimation or any test-time optimization.
By conditioning on point clouds instead of e.g., meshes, context data can be obtained easily from real world scenes.
This allows \gls{peach} to simulate real-world scenes without explicit parameter estimation or test-time optimization, and we demonstrate zero-shot sim-to-real transfer on a challenging trampoline scene.
% As such, it facilitates efficient test-time real-to-sim transfer, in turn enabling downstream tasks such as planning and control on deformable objects directly from raw sensor data.
Across four challenging simulated datasets, \gls{peach} performs on par with or better than mesh-based learned simulators, even when these have access to the ground truth parameters.
% Our method combines a spatio-temporal encoder that treats context sequences as point clouds in 4D space-time with a trajectory-based learned simulator conditioned on the resulting encoded material parameters.
Crucial to our method are our novel spatio-temporal encoder that treats context sequences as point clouds in 4D space-time, and auxiliary supervision in the form of direct parameter regression and prediction of \gls{sdf} values.

\textbf{Limitations and future work.} 
\gls{peach} assumes access to an initial scene geometry and requires context trajectories that share physical properties with the target process. 
Generating an initial mesh from observation would relax the geometry assumption, while accumulating context online during, e.g., robot execution would handle material drift across a process.
Additionally, \gls{peach} currently predicts the full trajectory in a single forward pass, which may impose memory constraints for long sequences and high-resolution meshes. 
Predicting sparse keyframes and training a separate network to interpolate between them is a promising way to decouple memory cost from horizon length.
Further extensions could include other sensor modalities such as temperature readings and touch sensors for more accurate parameter estimation.
We discuss broader impact in Appendix~\ref{app:broader_impact}.

% We introduced \glsf{peach}, an in-context learning framework that adapts graph network simulators to unseen material properties using only sequences of point cloud observations.
% By decoupling material inference from mesh-based dynamics prediction, \gls{peach} extends prior mesh-dependent approaches to more realistic observation settings while retaining the efficiency and accuracy of trajectory-level graph network simulators.
% Our experiments on simulated deformation environments demonstrate that \gls{peach} achieves performance on par with mesh-based context methods, despite relying on significantly sparser input data.

% While this work takes an important step toward real-world adaptive simulation, all evaluations are conducted using simulated point clouds. Thus, real-world sensing effects such as noise, occlusions, and domain shift are not yet considered.
% Furthermore, inference still relies on a canonical simulation mesh, and we do not explicitly study geometric mismatch between simulated and observed systems.

\bibliography{peach, balazs}
\bibliographystyle{unsrtnat}

\newpage

\appendix

\section{Broader Impact}
\label{app:broader_impact}
\gls{peach} is a method for simulating deformable objects from point cloud observations, aimed at making physics-based simulation more accessible when material properties are unknown or hard to measure.
Efficient real-to-sim transfer of this kind could reduce reliance on costly physical experiments in engineering and scientific workflows and support digital twin construction for monitoring and design.
It can additionally lower the barrier to high-fidelity simulation in domains such as biomechanics, materials science, and soft-body engineering, where calibrating constitutive models is traditionally laborious.

At the same time, any improvement in simulation fidelity carries dual-use potential, for instance in the design of weapons or other harmful artifacts. We view \gls{peach} as a generic enabling technology in this regard, with no direct path to harm beyond what is already implied by progress in learned simulation more broadly. 
Like any data-driven model, \gls{peach} may also inherit biases from its training distribution. In particular, materials, geometries, and process conditions that are underrepresented during training are likely to be predicted less accurately, which could lead to silent failures if the method is used outside its validated regime. 
We therefore encourage practitioners to validate \gls{peach} on their target domain before relying on its predictions in decision-critical settings.
\section{Methods and Hyperparameters} \label{app:methods}

\subsection{Point Cloud Encoder} \label{app:pc_encoder}
In this section, we provide additional details on the point cloud encoders used in our comparisons.
\paragraph{PSTNet.}
We closely follow the method proposed in \cite{fan2021pstnet} and use the official implementation and architecture from \url{https://github.com/hehefan/Point-Spatio-Temporal-Convolution}. The spatial kernel radius is adapted to $0.1$ to match our task normalization, while all other architectural components and hyperparameters are kept unchanged.

\paragraph{GNN Encoder.}
At each timestep, we perform \gls{fps} to select $52$ center points per point cloud in the sequence. For each center point, we connect its $16$ nearest neighbors and apply a PointNet layer~\citep{qi2016pointnet} to each resulting patch, yielding an embedding of dimension $32$. We concatenate a Fourier encoding of the corresponding timestep to this embedding.

The resulting spatio-temporal tokens are then assembled into a graph: within each timestep, every node is connected to its $4$ nearest neighbors. In addition, each node is connected to its $2$ nearest neighbors in the subsequent timestep and to its single nearest neighbor in the following timestep. This construction yields a spatio-temporal graph encoding the complete point cloud sequence.

The graph is processed using a message-passing graph neural network~\citep{sanchezgonzalez2020learning} without global features, consisting of $5$ layers with a latent dimension of $128$. Finally, an attention-based readout module is applied to obtain a single output token.

% \paragraph{\textit{PEACH} Transformer Encoder.}
% For our method, we employ a spatio-temporal furthest point sampling to select the center points. All timesteps are first normalized to the unit interval and then scaled by a factor of 16, producing a temporal scaling comparable to the spatial normalization. For each center point, we extract its 16 nearest neighbors and apply a PointNet layer to the resulting patch. Temporal and spatial Fourier encodings are then added to these embeddings to incorporate both spatial and temporal information.

% We implement an Attention Pooling module that combines a Transformer encoder with a cross-attention-based pooling mechanism. Input tokens are first processed through a Transformer encoder using multi-head self-attention, allowing them to exchange information and build contextual representations. The resulting token embeddings are then passed to an Attention Pooling layer, where a single learnable query token attends to the encoded sequence via cross-attention, summarizing the variable-length input into a fixed set of output tokens. 

\subsection{Mesh Encoder and Simulator}
\paragraph{Mango Mesh Encoder.}
We closely follow the architecture of \cite{dahlinger2025mango} and apply a one-dimensional convolutional neural network to each node along the temporal dimension. The mesh structure provides consistent point correspondences over time, enabling this temporal processing. Node features are then aggregated using a Deep Set architecture \citep{cai2023deepset} to produce a permutation-invariant representation of the mesh. We use the official implementation, available at \url{https://github.com/ALRhub/mango}.

\paragraph{Mango Simulator.}
We closely follow the implementation from \cite{dahlinger2025mango}, using the official code available at \url{https://github.com/ALRhub/mango}.

\subsection{Parameters and Runtime} \label{sec:runtime}
\gls{peach} has a total of $3.96$M parameters, of which $3.01$M belong to the MaNGO simulator and $0.96$M to the point cloud encoder and auxiliary loss heads.

We report inference timings on an NVIDIA RTX~5090 for the \texttt{Trampoline} environment.
The encoding time scales with context size, ranging from $84$\,ms for a single context trajectory to $219$\,ms for eight.

The simulation rollout takes approximately $352$\,ms regardless of context size, as it operates solely on the aggregated latent code. In comparison, the FEM simulator used to generate the training data required 4 minutes per rollout on an AMD EPYC 74F3 24-Core Processor.
Compared to classical system identification methods that require iterative test-time optimization, \gls{peach} performs material inference in a single forward pass, making it orders of magnitude faster in practice.

\subsection{Hyperparameters} \label{app:hp}
We provide a list of the used hyperparameters in Table \ref{tab:appx_training_setup}.

\begin{table}[t]
\centering
\caption{List of the used hyperparameters}
\label{tab:appx_training_setup}
\begin{tabular}{lc}
\toprule
    Parameter & Value  \\ 
\midrule
Num Points in Pointcloud & 512 \\
Num Points in Pointcloud (Trampoline) & 1024 \\
FPS downsampling ratio (Tokenizer) & 0.1 \\
Patch size (Tokenizer) & 16 \\
Time scaling (Tokenizer) & 16 \\
Num. Fourier frequencies & 8 \\
Transformer Attention heads & 4 \\
Latent material representation dimension  & $128$ \\
Simulator Node feature dimension & $128$ \\
Simulator Message passing blocks & $15$ \\
Simulator Aggregation function & Mean \\
Simulator Activation function & Leaky ReLU \\
Optimizer & AdamW~\citep{adamw} \\
Learning rate & $5.0 \times 10^{-5}$ \\
Weight decay & $1.0 \times 10^{-4}$ \\
Min Context Size (Training) & $1$ \\
Max Context Size (Training) & $8$ \\
Latent representation aggregation & Learned Softmax \\
Auxiliary Loss Scale (Phys. Parameters) & $0.02$ \\
Auxiliary Loss Scale (SDF) & $0.5$ \\
Noise Scale MGN (Def. Block) & $0.0007$ \\
Noise Scale MGN (Bend. Beam) & $0.0007$ \\
Noise Scale MGN (Trampoline) & $0.0005$ \\
Noise Scale MGN (Sheet Def.) & $0.0005$ \\
\bottomrule
\end{tabular}
\end{table}

\section{Simulation Datasets} \label{app:dataset}
This section provides detailed information about the datasets used in our experiments. 
The key characteristics of each dataset are summarized in Table~\ref{tab:dataset_description}. \texttt{Deforming Block} and \texttt{Sheet Deformation} consist of $16$ simulation trials per task with various initial conditions, \texttt{Trampoline} and \texttt{Bending Beam} consist of $10$ simulation trials per task.
\begin{table}[h]
\centering
\caption{Dataset descriptions}
\label{tab:dataset_description}
\begin{tabular}{lccc}
\toprule
Name & Train/Val/Test Splits & Number of Steps & Number of Nodes \\
\midrule
Deforming Block       & 600/100/100   & 52  & 81 \\
Sheet Deformation & 460/50/50 & 50  &  225 \\
Bending Beam & 380/60/60 & 100 & ~180 \\
Trampoline  & 600/100/100 & 25  &  ~1700 \\
\bottomrule
\end{tabular}
\end{table}

\subsection{Trampoline.}
%
% \texttt{Trampoline (Sim)} consists of a \texttt{2D} deformable membrane clamped by a rigid frame and deformed by a spherical impactor of varying size and density falling under gravity.
% The membranes follow a nearly incompressible viscoelastic material model, with Young's modulus $E$ sampled uniformly from $[1,5]$ MPa, Poisson's ratio fixed to $\nu=0.499$, membrane thickness $t$ uniformly from $[0.15,0.4]$ mm, shear relaxation ratio $g$ uniformly from $[0.01,0.3]$, and relaxation time $\tau$ log-uniformly from $[10^{-2},10^{1}]$ s.
% The dataset includes $600/100/100$ train/val/test tasks with $10$ trajectories of $25$ steps each. Each scene contains roughly $1700$ nodes and $3300$ triangular elements.
% 
This dataset models a thin square deformable sheet clamped by a rigid frame and deformed by a spherical impactor of varying size and density falling under gravity. Rigid body motion prediction of the ball is enforced by predicting per-node velocities and averaging them across all ball nodes
The sheet has initial stress-free edge length $l_{\textrm{init}}=245$~mm.
The sheet is discretized by $9604$ triangular membrane elements and $4901$ nodes, with membrane thickness $t$ varied as described below. 
To emulate the rigid frame in the physical setup, the sheet is pre-stretched to an edge length of $l=260$~mm by applying Dirichlet boundary conditions along its outer boundary.
\par
Data generation follows a two-stage sampling procedure. 
In the first stage, a simulation configuration is defined by sampling the sheet material parameters together with the properties of a rigid spherical impacter. 
The sheet material is modeled as a nearly incompressible viscoelastic solid with rubber-like behavior. 
The instantaneous Young's modulus is sampled uniformly from $E \in [1,5]$~MPa, the Poisson's ratio is fixed to $\nu=0.499$, and the membrane thickness is sampled from $t \in [0.15,0.4]$~mm. 
Rate dependence is modeled using a single-term Prony series with shear relaxation ratio $g \in [0.01,0.3]$ and relaxation time $\tau \in [10^{-2},10^{1}]$~s.
Volumetric relaxation is neglected ($k=0$). 
The impacter diameter is chosen from $15$ to $55$~mm in steps of $5$~mm, with each diameter represented by a pre-meshed geometry of approximately $1000$ elements. 
Its mass is sampled from $[0.009,0.269]$~kg, subject to density bounds between $268$~kg/m$^3$ (3D-printed PLA with $10\%$ infill) and $7850$~kg/m$^3$ (steel).
Gravitational loading is applied as an equivalent concentrated force at the impacter's center of mass. 
\par
In the second stage, $10$ trajectories are generated for each fixed stage-one configuration by varying only the initial conditions of the spherical impacter. 
Specifically, the in-plane drop position is sampled from a Gaussian distribution centered at the sheet origin and truncated to $[-75,75]$~mm in each in-plane direction, while the drop height is sampled within $[120,280]$~mm.
Contact between the impacter and the sheet is modeled using hard normal contact to prevent interpenetration. 
Tangential contact is governed by a Coulomb friction law with coefficient $\mu=0.7$. 
All simulations are performed in \textsc{Abaqus/Explicit} using explicit time integration.

\subsection{Bending Beam.}
%
% \texttt{Bending Beam} consists of $2$D beams clamped on the left side that bend and deform over time due to external forces applied on the right side. 
% The beams follow an isotropic Kelvin--Voigt viscoelastic material model, with Young's modulus $E$ sampled log-uniformly from $[0.5, 5.0]$, Poisson's ratio $\nu$ uniformly from $[0.0, 0.45]$, and viscosity $\tau$ log-uniformly from $[0.05, 0.5]$.
% The dataset includes $380/60/60$ train/val/test tasks with $10$ trajectories of $100$ steps each.
%
This scene consists of $2$D beams clamped on the left side that bend and deform over time due to time-varying external forces applied on the right side.
The beam is modeled as a rectangular domain embedded in two dimensional space.
The left boundary of the beam is clamped with zero displacement, while external forces are applied to the right boundary.
The load consists of one horizontal and one vertical force component.
The horizontal force magnitude is sampled uniformly in $[-1.0{\cdot}10^{-7}, 1.0{\cdot}10^{-7}]$, while the vertical force magnitude is sampled uniformly in $[-3.0{\cdot}10^{-5}, 3.0{\cdot}10^{-5}]$.
The forces are varied smoothly over time using cubic spline interpolation, starting from zero force, reaching the sampled target magnitude, and returning to zero before the end of the simulation.
The beam height is sampled normally with mean $0.3$ and standard deviation $0.01$, and the beam length is sampled normally with mean $10.0$ and standard deviation $1.0$.
The maximum characteristic mesh length, used for mesh generation, is sampled normally with mean $0.2$ and standard deviation $0.02$.
The material behavior is modeled using an isotropic Kelvin--Voigt viscoelastic material model, combining an elastic stress response with a viscous stress contribution.
Young's modulus $E$ is sampled log-uniformly in $[0.5, 5.0]$, Poisson's ratio $\nu$ is sampled uniformly in $[0.0, 0.45]$, and the viscosity parameter $\tau$ is sampled log-uniformly in $[0.05, 0.5]$.
Geometric non-linearities due to large deformations are considered.
The beam is discretized using linear triangular elements.
The simulations are implemented in \texttt{scikit-fem} and solved using Newton iterations with a residual tolerance of $10^{-8}$.
We create a total of $500$ simulations, each consisting of $40{,}000$ time steps. For the \gls{gns}, we subsample to $100$ time steps per trajectory.

\subsection{Deforming Block.}
%
% \texttt{Deforming Block} consists of $2$D trapezoids deformed by a circular collider.
% The material varies by Poisson's ratio, which is sampled uniformly from $[-0.9, 0.49]$.
% The dataset contains $600/100/100$ train/val/test tasks, each with $16$ trajectories of $52$ simulation steps.
% \texttt{Deforming Block} consists of $2$D trapezoids with Poisson's ratios uniformly sampled from $[-0.9, 0.49]$ that are deformed by a circular collider. 
% The dataset for this environment contains $600/100/100$ train/val/test tasks, each containing $16$ trajectories of $52$ simulation steps.
%
This scene consists of $2$D trapezoids deformed by a circular collider. 
This environment was originally implemented by \citet{linkerhagner2023grounding} and uses \gls{sofa} \citep{faure2012sofa}. 
This dataset considers the deformation of a two dimensional rectangular block in two dimensional space. 
There are different trapezoidal initial geometries of the block.
The nodes located at the bottom edge of the block are clamped with zero displacement.
The load is applied by means of a contact between a rigid circular collider and the top surface of the block.
Across simulation trials, the collider radius is sampled uniformly between $10\,\%$ and $40\,\%$ of the block size and moves vertically downward at a constant velocity.
The contact is modeled using a hard contact definition, which is rigid in the normal direction and frictionless in the tangential direction. 
The material behavior of the block is modeled using isotropic linear elasticity with a fixed Young’s modulus~$E$, while the Poisson’s ratio $\nu$ is sampled uniformly per task from~$[-0.9,\,0.49]$.
The block is discretized using $81$ nodes and $128$ triangular elements.

\begin{figure}[t]
    \centering
    \includegraphics[width=\textwidth]{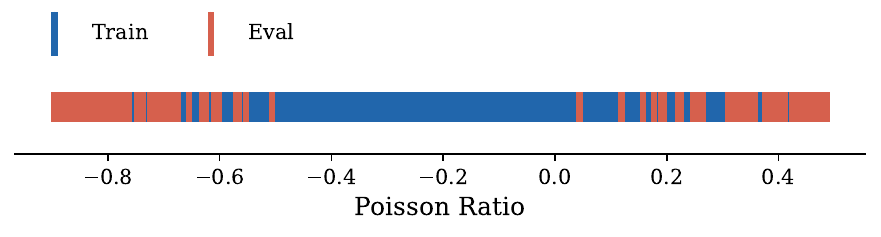}
    \caption{Out-of-distribution evaluation split for the \texttt{Deforming Block OOD} environment. Each tick marks one task, colored by split assignment. Eval tasks are drawn from the tails of the Poisson ratio distribution, ensuring that evaluated material properties lie outside the training range.}
    \label{fig:ood_split}
\end{figure}

The out-of-distribution evaluation split for this environment is visualized in Figure~\ref{fig:ood_split}, where eval tasks are drawn from the tails of the Poisson ratio range $[-0.9,\,0.49]$.

\subsection{Sheet Deformation.}
%
% \texttt{Sheet Deformation} consists of a $2$D sheet subjected to out-of-plane forces, resulting in $3$D deformations.
% The material is parameterized by Young's modulus, which is sampled log-uniformly from $[10, 500]$.
% The dataset contains $460/50/50$ train/val/test tasks, each with $16$ trajectories of $50$ simulation steps.
% In \texttt{Sheet Deformation}, a $2$D sheet is subjected to external forces perpendicular to the sheet, resulting in $3$D deformations. 
% The sheet's Young's Modulus is sampled log-uniformly from $[10, 500]$.
% Here, the dataset includes $460/50/50$ train/val/test tasks with $16$ trajectories of $50$ steps each.
%
This dataset considers the deformation of a thin square $2$D deformable sheet subjected to out-of-plane forces, resulting in $3$D deformations.
The sheet has initial stress-free edge length $l=140\;\text{mm}$. 
The thickness of the sheet is intrinsically defined as $t = 2.0 \;\text{mm}$. 
Nodes located at the outer edge of the sheet are clamped with zero displacement. The load is applied by two discrete external forces acting perpendicular to the thickness direction of the sheet. 
The force magnitude is fixed, while the normal force direction (upward or downward) and point of application are sampled uniformly.
To enable a continuous range of force application locations, the forces are applied uniformly to all nodes within a small radius  around the sampled application location. 
The forces are ramped up linearly from zero to their target magnitude over the simulated time. 
The material behavior of the sheet is modeled using isotropic linear elasticity with Young's modulus $E$ sampled log-uniformly in $[10, 500] \; \text{MPa}$ and Poisson's ratio $\nu{=}0.3$. 
The sheet is discretized using $225$ nodes and $392$ triangular elements. 
Geometric non-linearities due to large deformations are considered.
The simulation for the described environment is implemented in the finite element software \texttt{Abaqus/Standard} and solved using implicit time integration.

\section{Real World Experiments} \label{app:real}

We create a real-world version of \textbf{Trampoline} dataset for use in our sim to real experiments.
To mimic the simulated scene, solid balls of various materials and densities (see Table \ref{tab:real_spheres}) are dropped into rubber sheets of various thicknesses (see Table \ref{tab:real_sheets}).
The balls are dropped using a Franka Emika Panda robot, where the in-plane drop position is sampled from a volume of \SI{4}{\cm} x \SI{4}{\cm} x \SI{5}{\cm}.
The scene is recorded using a StereoLabs Zed Mini stereo camera at a resolution of $720$x$1280$ frame rate of \SI{30}{fps}.

\begin{table}[ht]
\centering
\begin{tabular}{c c l}
\toprule
\textbf{Diameter} (\si{\mm}) & \textbf{Mass} (\si{\g}) & \textbf{Material} \\
\midrule
40 & 269 & Steel Solid \\
40 & 69  & Steel Hollow \\
50 & 110 & Steel Hollow \\
60 & 162 & Steel Hollow \\
40 & 21  & PLA 50\% Infill \\
\bottomrule
\end{tabular}
\vspace{0.1cm}
\caption{Sphere specifications used in the real-world experiments.}
\label{tab:real_spheres}
\end{table}

\begin{table}[ht]
\centering
\begin{tabular}{l c c}
\toprule
\textbf{Color} & \textbf{Thickness} (\si{\mm}) & \textbf{Average Density} (\si{\kg\per\m\cubed}) \\
\midrule
Black  & 0.381  & 992.7 \\
Green  & 0.254  & 992.7 \\
Yellow & 0.1524 & 992.7 \\
\bottomrule
\end{tabular}
\vspace{0.1cm}
\caption{Sheet specifications used in the real-world experiments.}
\label{tab:real_sheets}
\end{table}

Depth estimates for each recorded frame are recomputed using FoundationStereo~\cite{FoundationStereo} (smaller architecture based on Vit-small).
Images are resized to $480$x$853$ and are converted into point clouds by unprojecting the depth using the cameras' intrinsic parameters.
The camera extrinsics are calibrated by mounting Aruco markers on the frame around the rubber sheet and optimizing for minimal reprojection error (OpenCV's \texttt{solvePnPRansac}).
Point clouds are then transformed into a canonical world frame such that the bottom left corner of the sheet is at the origin and the sheet is on the xy plane.

We discard time steps before the ball begins to fall and limit trajectories to 20 time steps (approx. \SI{0.667}{\s}).
Points in the background of the scene are cropped out, and voxel downsampling is applied with a voxel edge length of \SI{5}{\mm}.
To remove spurious artifacts, any points that have fewer than $20$ neighbours within a radius of \SI{15}{\mm} are filtered out.
Finally, \gls{fps}~\cite{ruizhongtai2017pointnet} is applied to reduce the point clouds to 2048 points.

Sphere points are identified via color segmentation, isolating the distinctly colored ball from the scene.
The color labels are refined through an iterative algorithm that updates each point's label based on its own color and the colors of its neighbors, smoothing spurious misclassifications at object boundaries.
Given the known sphere diameter, a spherical mask is fitted to the identified sphere points by optimizing the center position, which is then used to remove robot arm points that fall outside the mask.
Finally, \gls{fps} is applied to reduce the point clouds to 1024 points.

\section{Additional Quantitative Results.} \label{app:quant}
\begin{figure}[htbp]
    \centering
    \makebox[\textwidth][c]{
        % \begin{tikzpicture}
%     \tikzstyle{every node}=[font=\scriptsize]
%     \input{figures/quantitative_results/tikzcolors}
%     \begin{axis}[%
%     hide axis,
%     xmin=10,
%     xmax=50,
%     ymin=0,
%     ymax=0.1,
%     legend style={
%         draw=white!15!black,
%         legend cell align=left,
%         legend columns=3,
%         legend style={
%             draw=none,
%             column sep=1ex,
%             line width=1pt,
%         }
%     },
%     ]

%     \addlegendimage{
%         line legend,
%         tabblue,
%         thick,
%         mark=square*,
%         mark options={line width=1.5pt, rotate=45}
%     }
%     \addlegendentry{\textbf{PEACH Encoder} (Ours)}
%     \addlegendimage{line legend, tabgreen, thick, mark=triangle*, mark options={line width=1.5pt, rotate=0}}
%     \addlegendentry{GNN Encoder}
%     \addlegendimage{
%         line legend,
%         taborange,
%         thick,
%         mark=square*,
%         mark options={line width=1.5pt, rotate=0}
%     }
%     \addlegendentry{PSTNet Encoder}
%     % New Line
%     % \addlegendimage{empty legend}
%     % \addlegendentry{}

%     \addlegendimage{line legend, tabolive, thick, mark=x, mark size=3pt, mark options={line width=1pt}}
%     \addlegendentry{Oracle Encoder}
%     \addlegendimage{line legend, tabcyan, thick, mark=+, mark size=2.5pt, mark options={line width=2pt}}
%     \addlegendentry{MaNGO Mesh Encoder}
%     \addlegendimage{line legend, tabgray, thick, mark=*}
%     \addlegendentry{No Context Encoder}    
%     \end{axis}
% \end{tikzpicture}

\begin{tikzpicture}
    \tikzstyle{every node}=[font=\scriptsize]
    \input{figures/quantitative_results/tikzcolors}
    \begin{axis}[%
        hide axis,
        xmin=10,
        xmax=50,
        ymin=0,
        ymax=0.1,
        legend style={
            draw=none,
            legend cell align=left,
            legend columns=7,
            column sep=1ex,
        }
    ]
        \addlegendimage{area legend, fill=white, draw=white, line width=0pt}
    \addlegendentry{\hspace{-8.2mm}\textbf{Point cloud context:}}
    \addlegendimage{area legend, fill=white, draw=white, line width=0pt}
        \addlegendentry{\hspace{-1ex}}
    \addlegendimage{area legend, fill=white, draw=white, line width=0pt}
    \addlegendentry{\hspace{-8.2mm}\textbf{Mesh context:}}
        \addlegendimage{area legend, fill=white, draw=white, line width=0pt}
        \addlegendentry{\hspace{-1ex}}
    \addlegendimage{area legend, fill=white, draw=white, line width=0pt}
    \addlegendentry{\hspace{-8.2mm}\textbf{Oracle context:}}
        \addlegendimage{area legend, fill=white, draw=white, line width=0pt}
        \addlegendentry{\hspace{-1ex}}
    \addlegendimage{area legend, fill=white, draw=white, line width=0pt}
    \addlegendentry{\hspace{-8.2mm}\textbf{No context:}}
    % Col 1: PEACH
    \addlegendimage{area legend, fill=tabblue}
    \addlegendentry{PEACH (ours)}
    % Col 3: spacer
    \addlegendimage{area legend, fill=white, draw=white, line width=0pt}
    \addlegendentry{\hspace{-1ex}}
    % Col 4: MaNGO
    \addlegendimage{area legend, fill=tabcyan}
    \addlegendentry{MaNGO}
    % Col 5: spacer
    \addlegendimage{area legend, fill=white, draw=white, line width=0pt}
    \addlegendentry{\hspace{-1ex}}
    % Col 6: Oracle + Oracle MGN
    \addlegendimage{area legend, fill=tabgreen}
    \addlegendentry{Oracle}
    % Col 7: spacer
    \addlegendimage{area legend, fill=white, draw=white, line width=0pt}
    \addlegendentry{\hspace{-1ex}}
    % Col 8: No Context + No Context MGN
    \addlegendimage{area legend, fill=tabgray}
    \addlegendentry{No Context}
    % Row 2
        \addlegendimage{area legend, fill=tabpink}
    \addlegendentry{PSTNet Encoder}
    \addlegendimage{area legend, fill=white, draw=white, line width=0pt}
    \addlegendentry{\hspace{-1ex}}
    \addlegendimage{area legend, fill=white, draw=white, line width=0pt}
    \addlegendentry{\hspace{-1ex}}
    \addlegendimage{area legend, fill=white, draw=white, line width=0pt}
    \addlegendentry{\hspace{-1ex}}
    \addlegendimage{area legend, fill=tabpurple}
    \addlegendentry{Oracle (MGN)}
    \addlegendimage{area legend, fill=white, draw=white, line width=0pt}
    \addlegendentry{\hspace{-1ex}}
    \addlegendimage{area legend, fill=taborange}
    \addlegendentry{No Context (MGN)}

        % Col 2: PSTNet
\addlegendimage{area legend, fill=tabbrown}
    \addlegendentry{GNN Encoder}
    \end{axis}
\end{tikzpicture}
    }
    \includegraphics[width=0.245\textwidth]{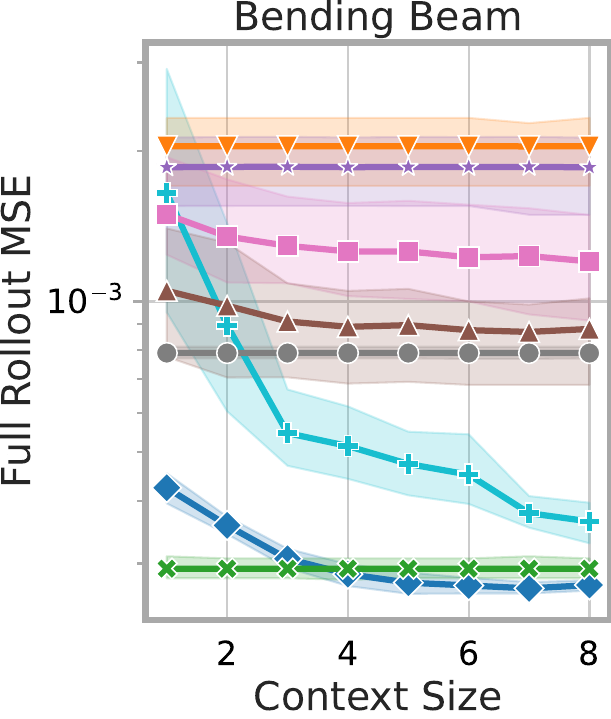}\hfill
    \includegraphics[width=0.245\textwidth]{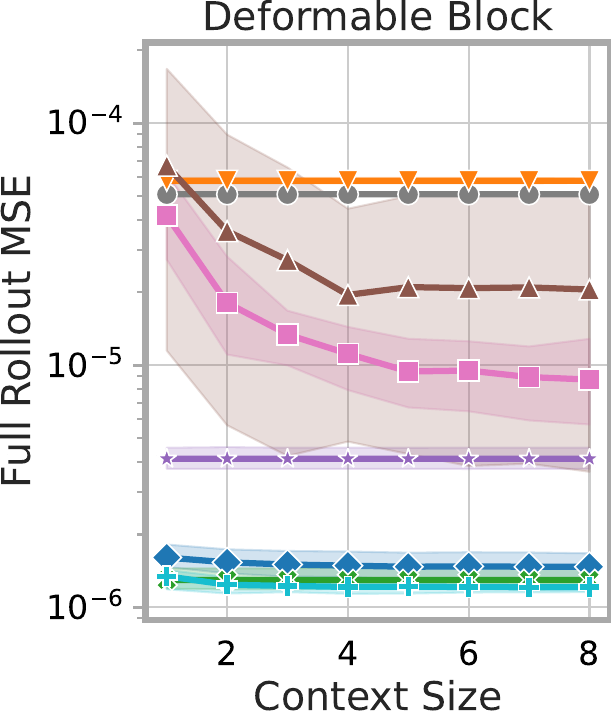}\hfill
    \includegraphics[width=0.245\textwidth]{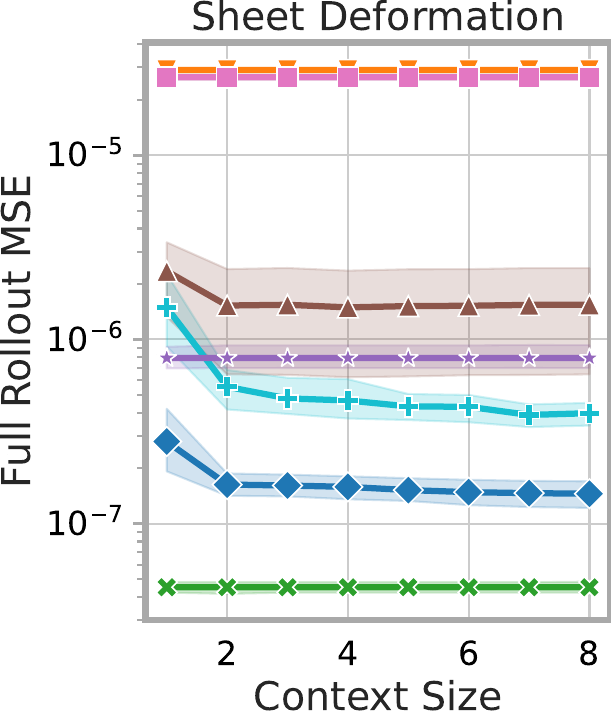}\hfill
    \includegraphics[width=0.245\textwidth]{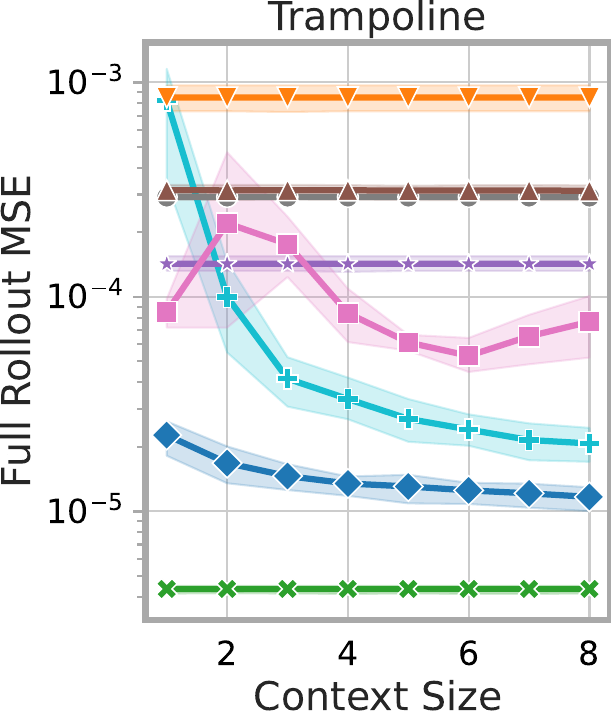}
    \caption{Full simulation rollout MSE across different baselines using 8 context trajectories.
    PEACH consistently performs similarly to the mesh-based MaNGO, improving over other
    point cloud-based baselines and achieving performance competitive with the Oracle models.}
    \label{fig:context_mse}
\end{figure}

We provide full rollout MSE as a function of context size for all four environments in Figure~\ref{fig:context_mse}. \gls{peach} consistently improves with more context trajectories and achieves performance competitive with the Oracle baselines, while outperforming all point cloud-based methods across environments.

\section{Visualizations} \label{app:vis}

We provide additional qualitative results comparing all evaluated methods on both datasets.

\subsection{Latent Space Visualizations}

\begin{figure}[htbp]
    \centering

    \begin{minipage}{0.24\textwidth}
        \centering
        \includegraphics[width=\textwidth]{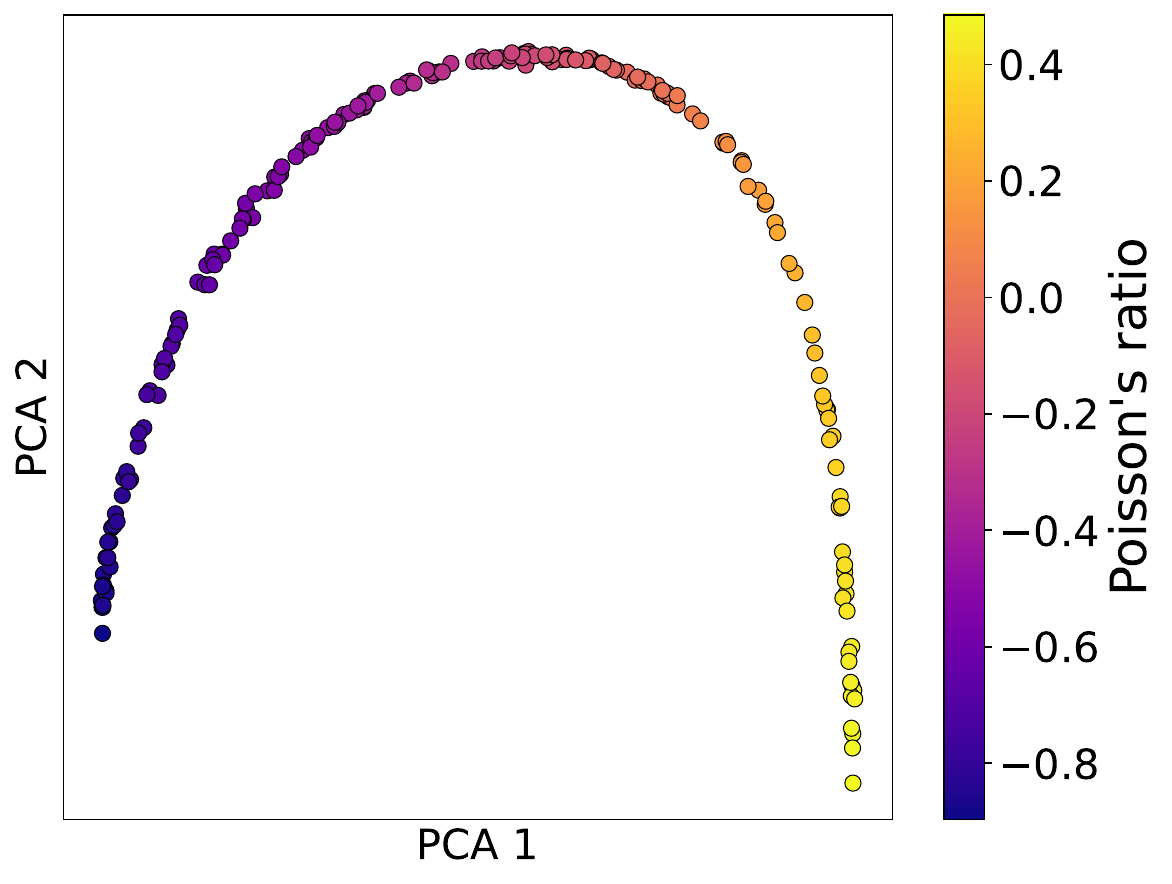}\\
        {\small \hspace{-1mm}\texttt{Deforming Block}}
    \end{minipage}
    \begin{minipage}{0.24\textwidth}
        \centering
        \includegraphics[width=\textwidth]{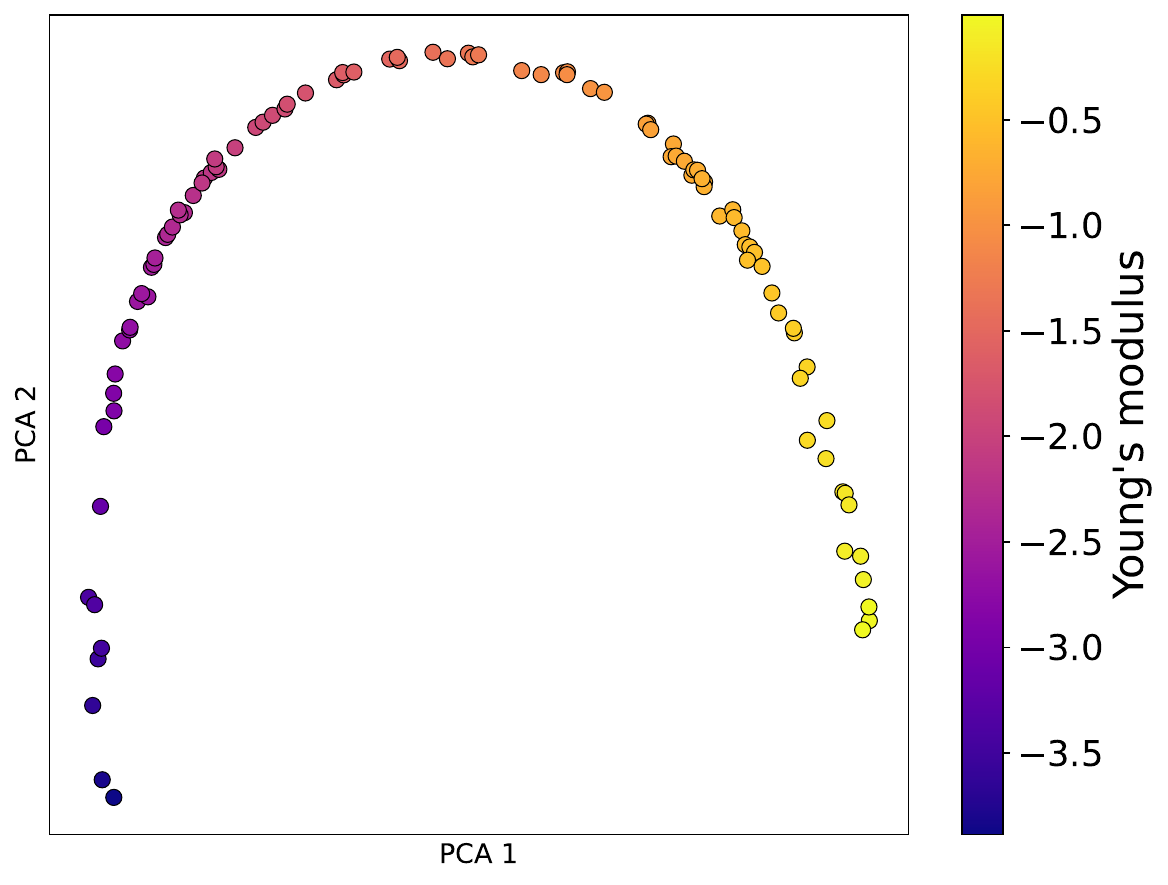}\\
        {\small \hspace{-2mm}\texttt{Sheet Deformation}}
    \end{minipage}
    \begin{minipage}{0.24\textwidth}
        \centering
        \includegraphics[width=\textwidth]{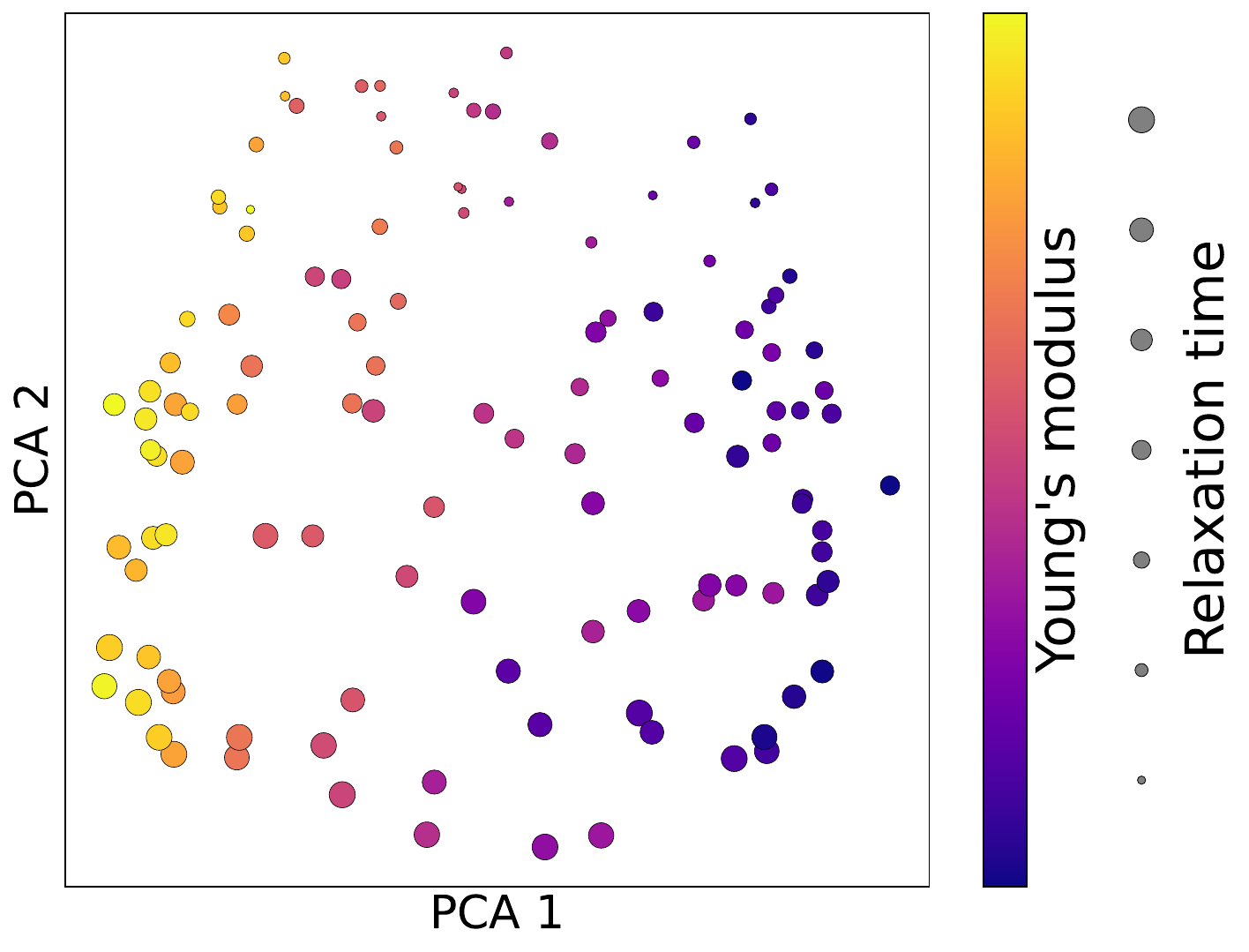}\\
        {\small \hspace{-1mm}\texttt{Bending Beam}}
    \end{minipage}
    \begin{minipage}{0.24\textwidth}
        \centering
        \includegraphics[width=\textwidth]{figures/latent_space_vis/pca_peach_trampoline_bivariate.pdf}\\
        {\small \hspace{-2mm}\texttt{Trampoline}}
    \end{minipage}

    \caption{
        PCA visualization of the latent space on the Deforming block, Sheet deformation, Bending beam and Trampoline environment (left to right).
        Each point represents an encoded task, with different coloring and scaling references based on the task's material properties.
    }
    \label{fig:latents_appendix}
\end{figure}

Figure~\ref{fig:latents_appendix} shows a PCA projection of the task-level latent codes $\mathbf{r}$ produced by the \gls{peach} encoder across all four environments. Each point corresponds to a single test task, colored by its material properties. 
The smooth, structured organization of the latent space, particularly the near-perfect one-dimensional manifold recovered for the single-parameter environments \texttt{Sheet Deformation} and \texttt{Deforming Block}, demonstrates that the encoder successfully disentangles material properties from trajectory-level variation, without any explicit supervision on the latent space geometry.

\subsection{Qualitative Trajectory Predictions}

\begin{figure*}[p]
    \centering
    {\Large \textbf{Deforming Block}}\\[1em]
    \vfill

    % helper macros (adjust trim once here)
    \newcommand{\rowlabel}[1]{%
        \begin{minipage}[c]{0.05\textwidth}\centering\rotatebox{90}{\scriptsize\textbf{\shortstack{#1}}}\end{minipage}%
    }
    \newcommand{\img}[1]{%
        \includegraphics[width=0.158\textwidth, trim=9cm 2cm 9cm 2cm, clip, valign=m]{#1}%
    }
    \newcommand{\pointcloud}[1]{%
        \includegraphics[width=\textwidth, trim=10cm 4cm 10cm 4cm, clip, valign=m]{#1}%
    }

    % Row 1: PEACH
    \rowlabel{PEACH}%
    \img{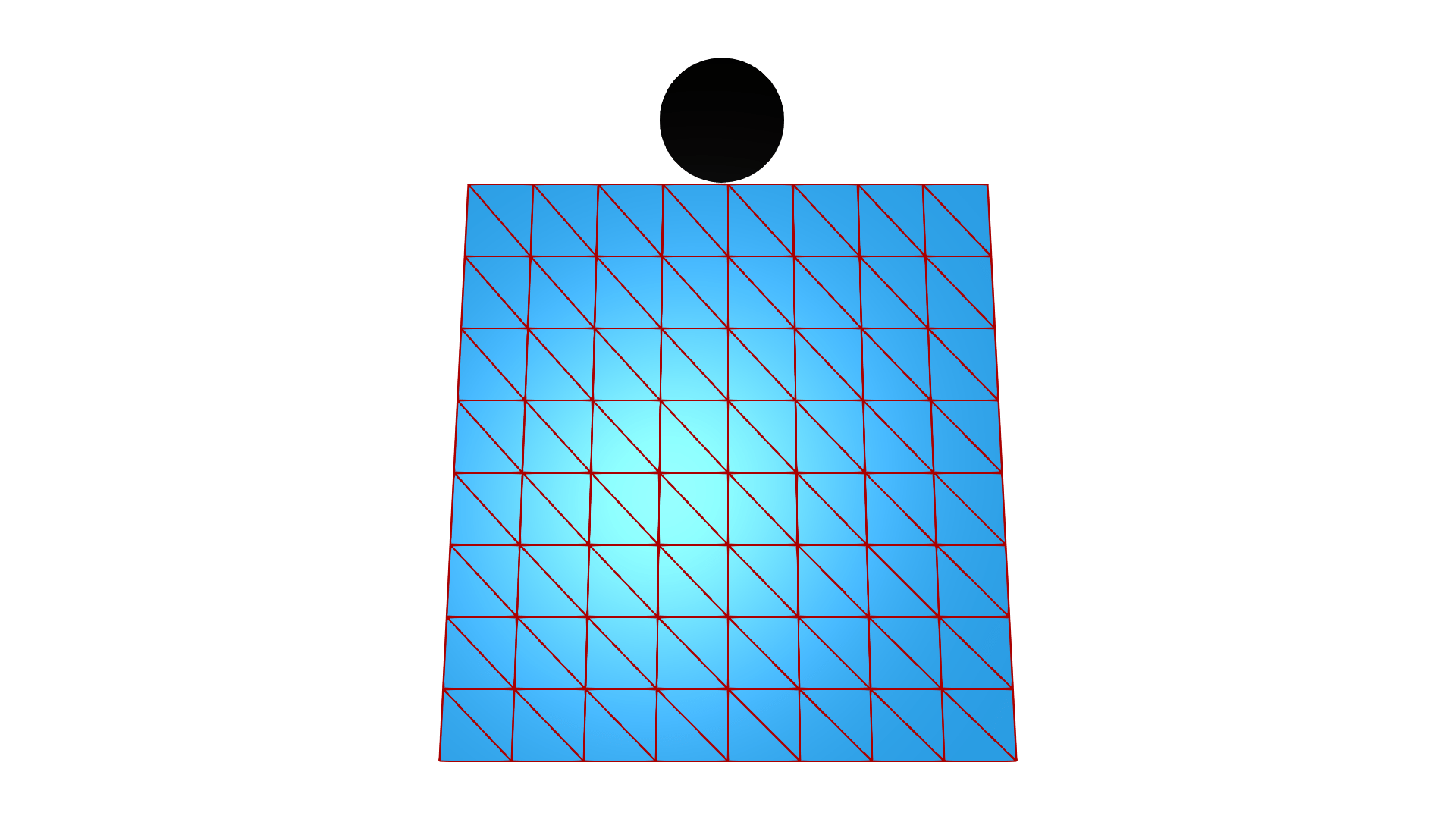}\hspace{-1pt}%
    \img{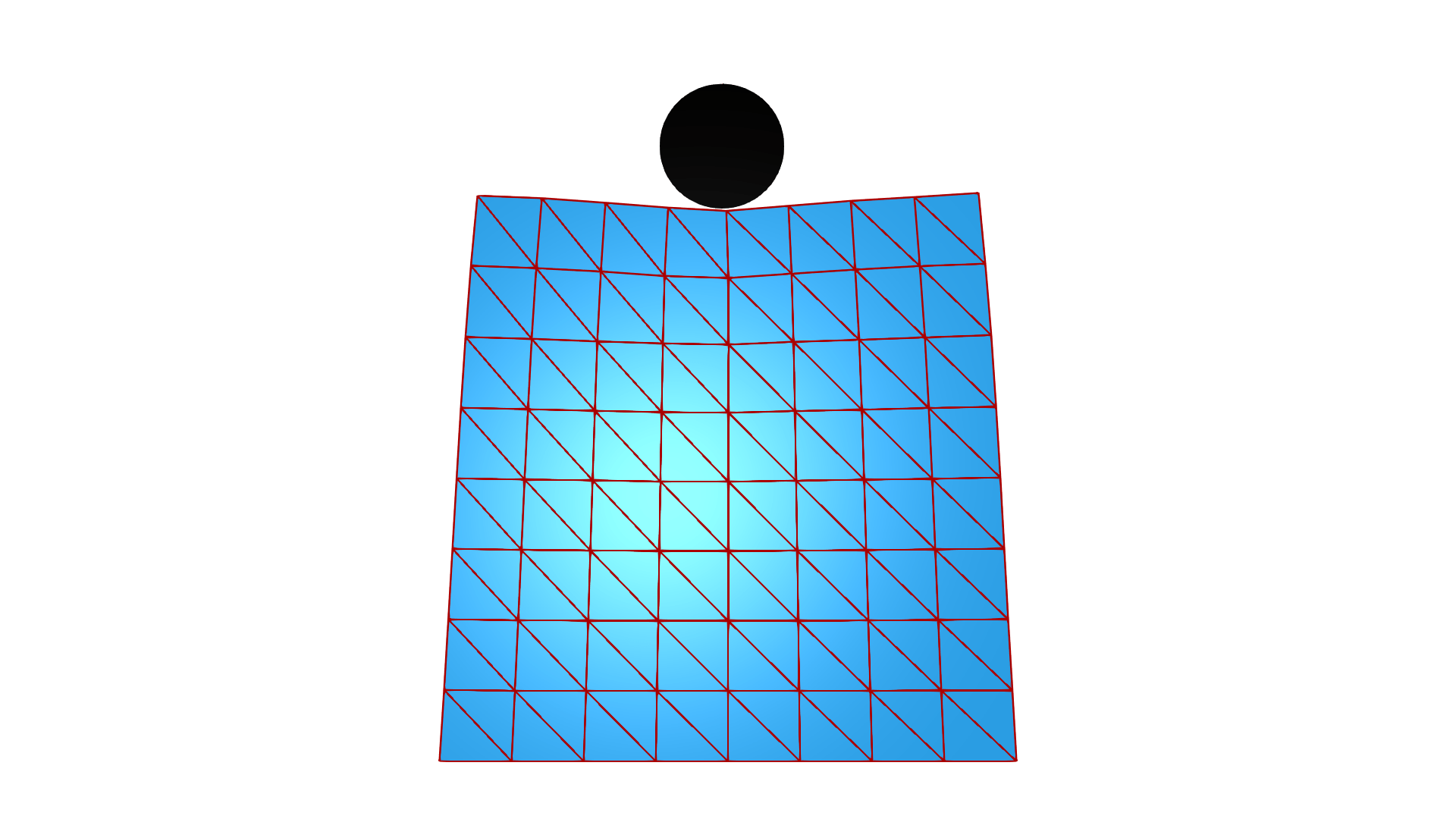}\hspace{-1pt}%
    \img{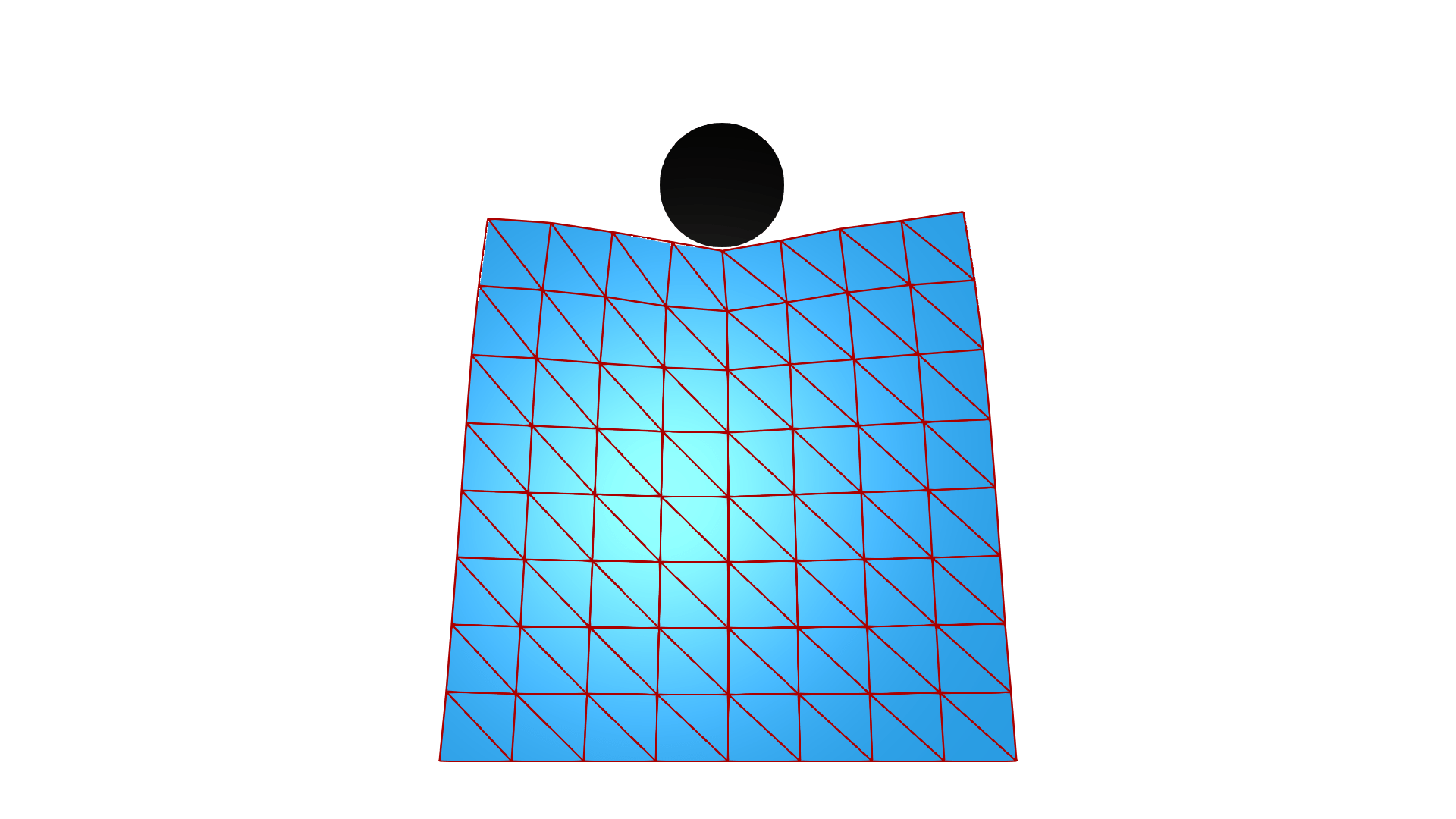}\hspace{-1pt}%
    \img{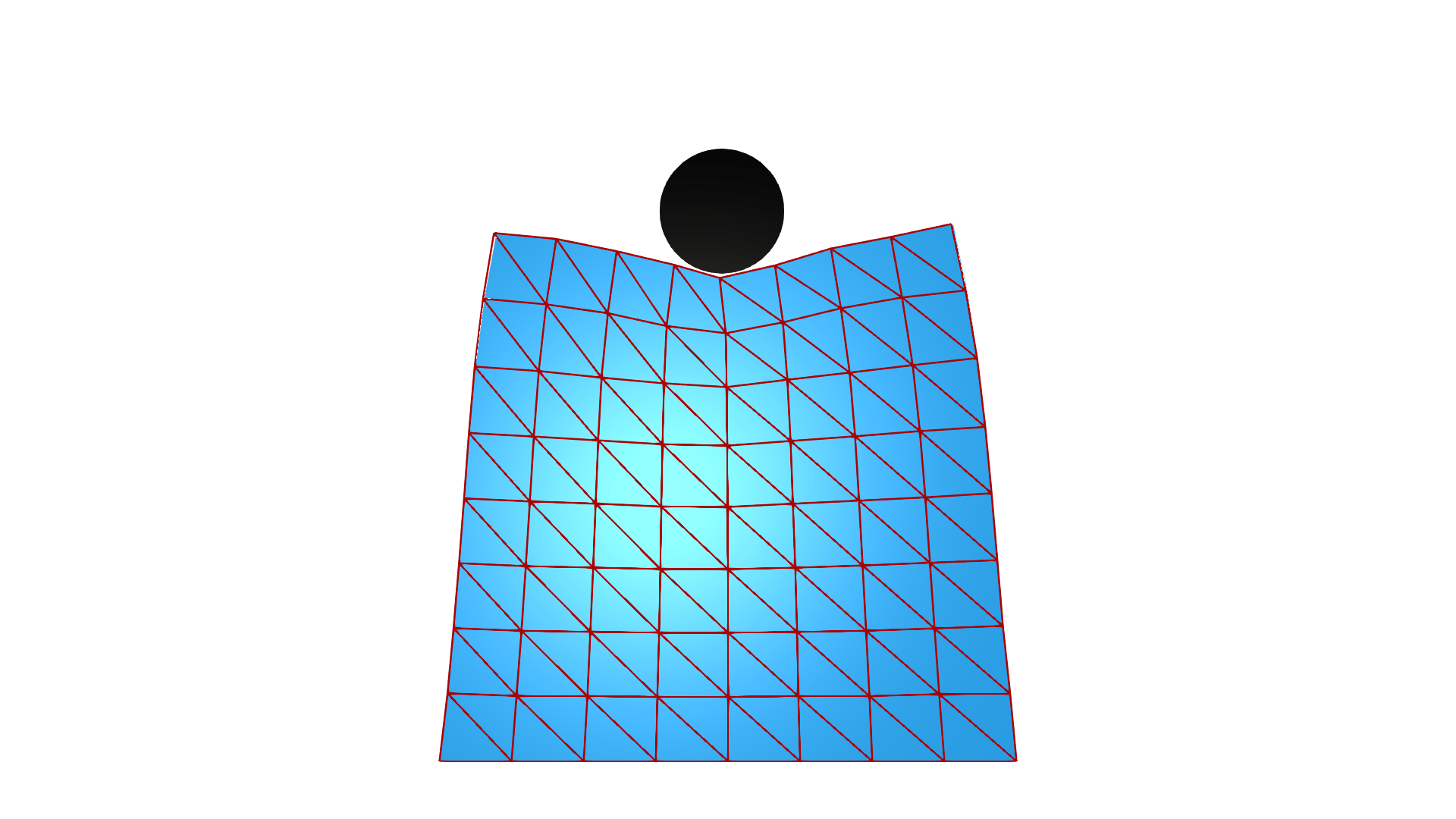}\hspace{-1pt}%
    \img{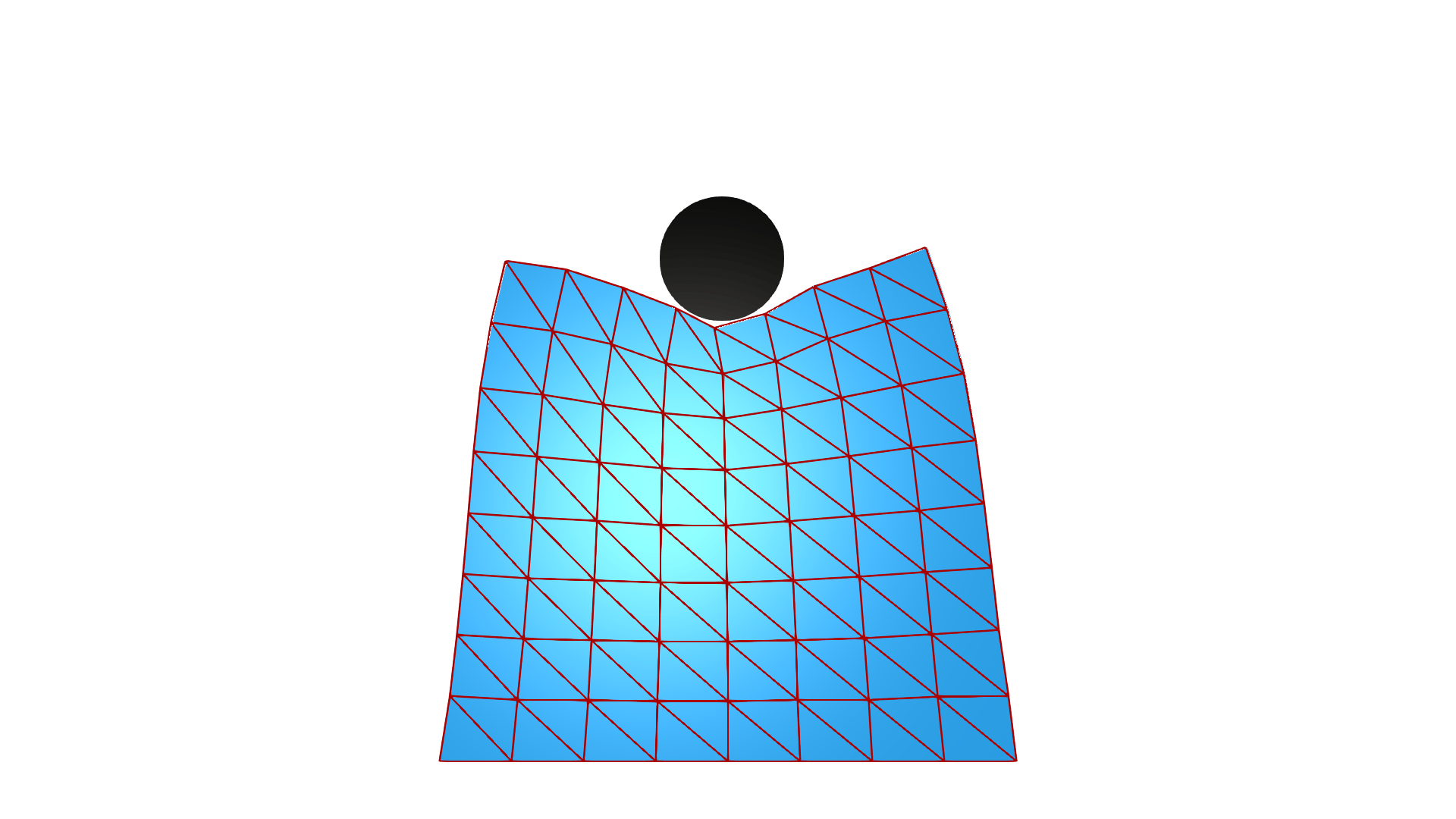}\hspace{-1pt}%
    \img{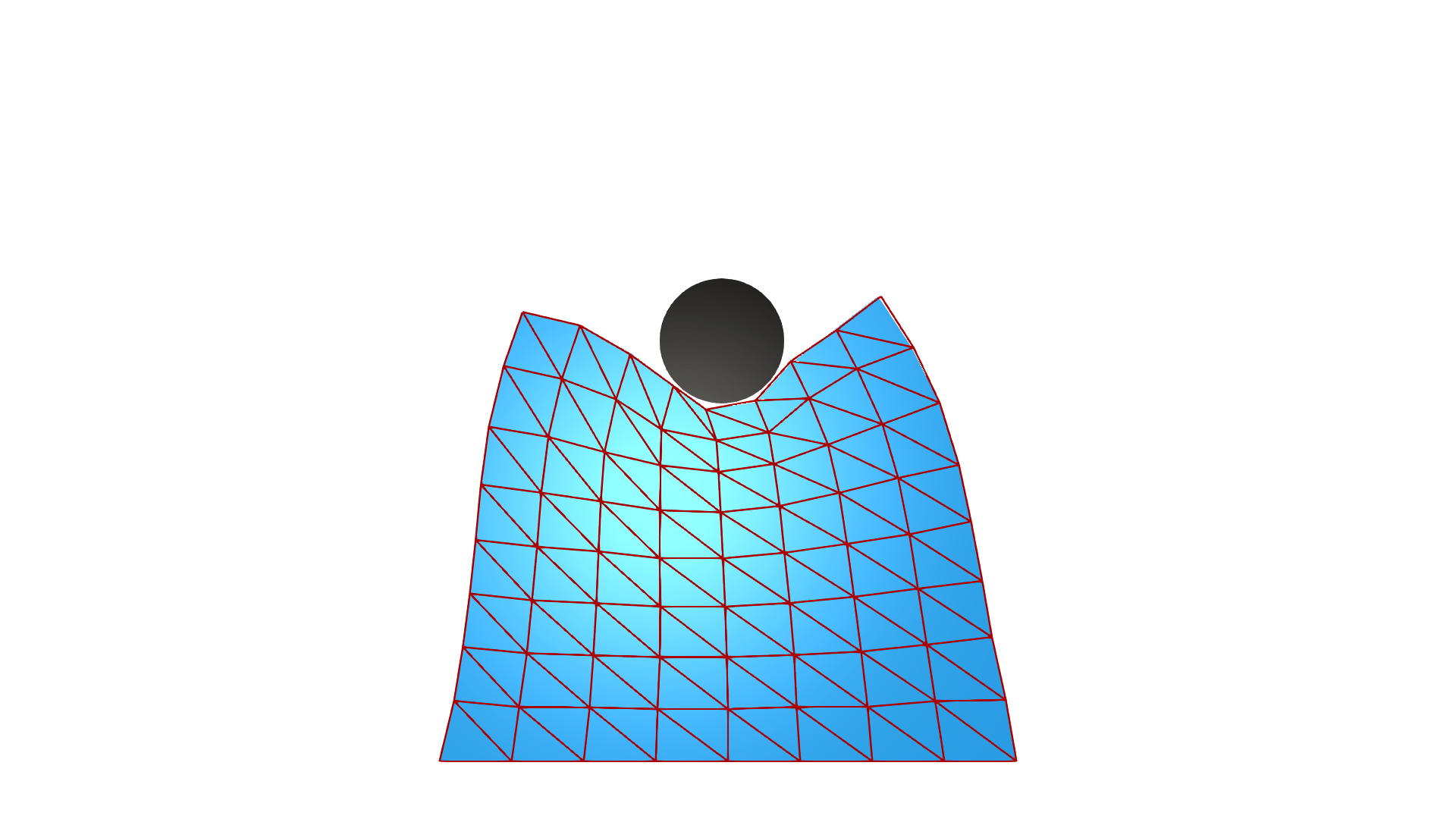}\\[0.4em]

    % Row 2: No Context
    \rowlabel{No Context}%
    \img{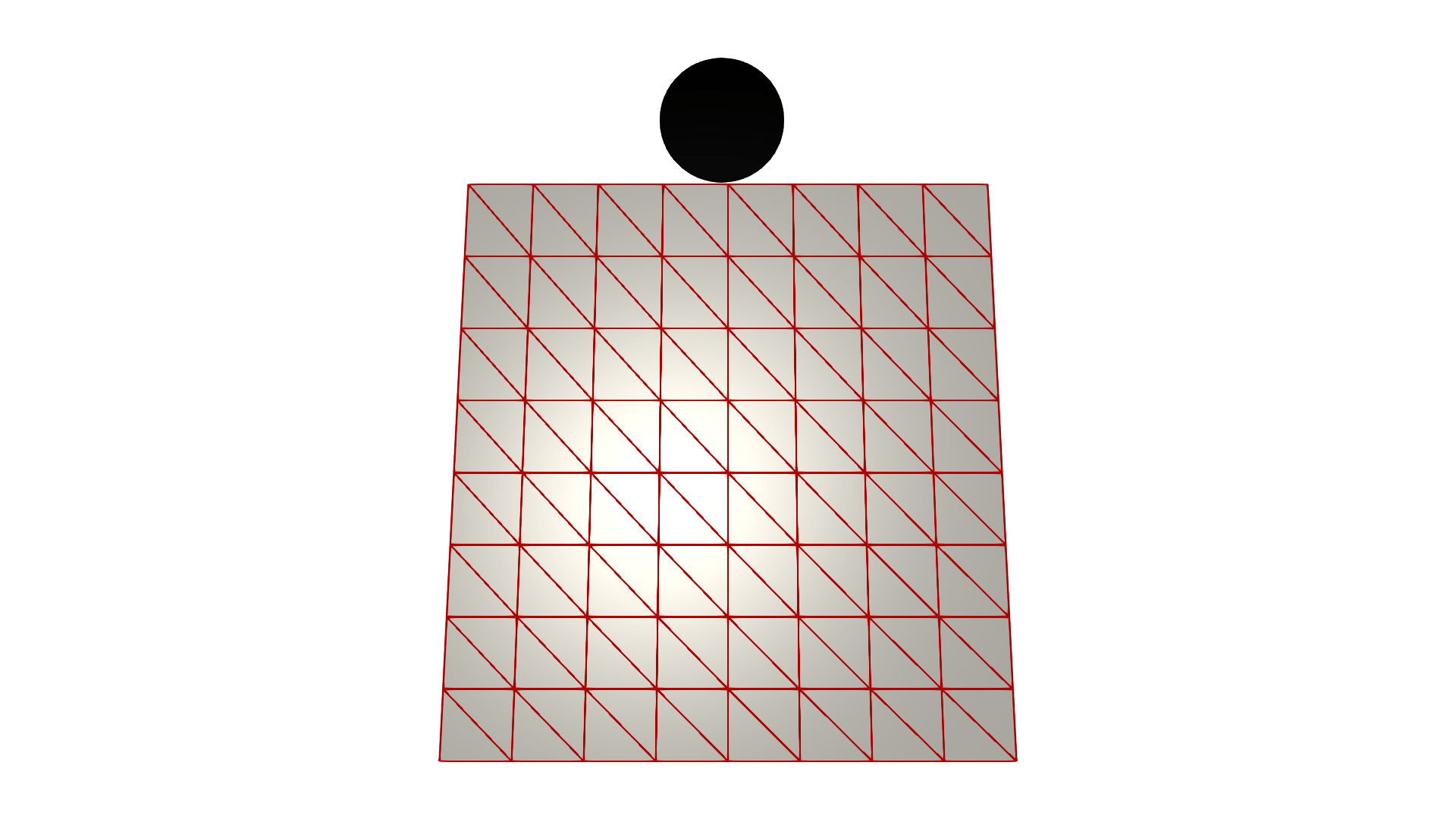}\hspace{-1pt}%
    \img{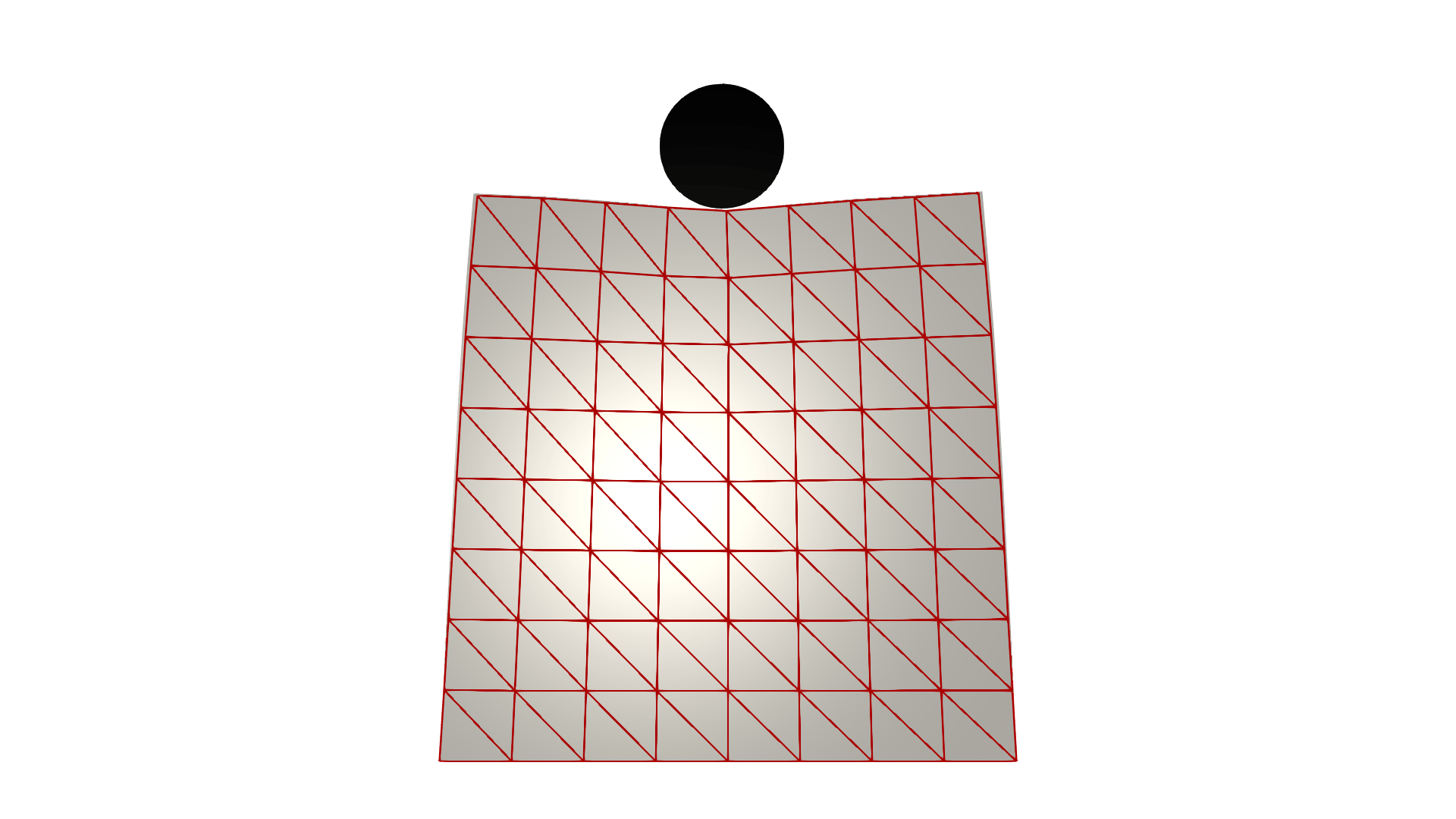}\hspace{-1pt}%
    \img{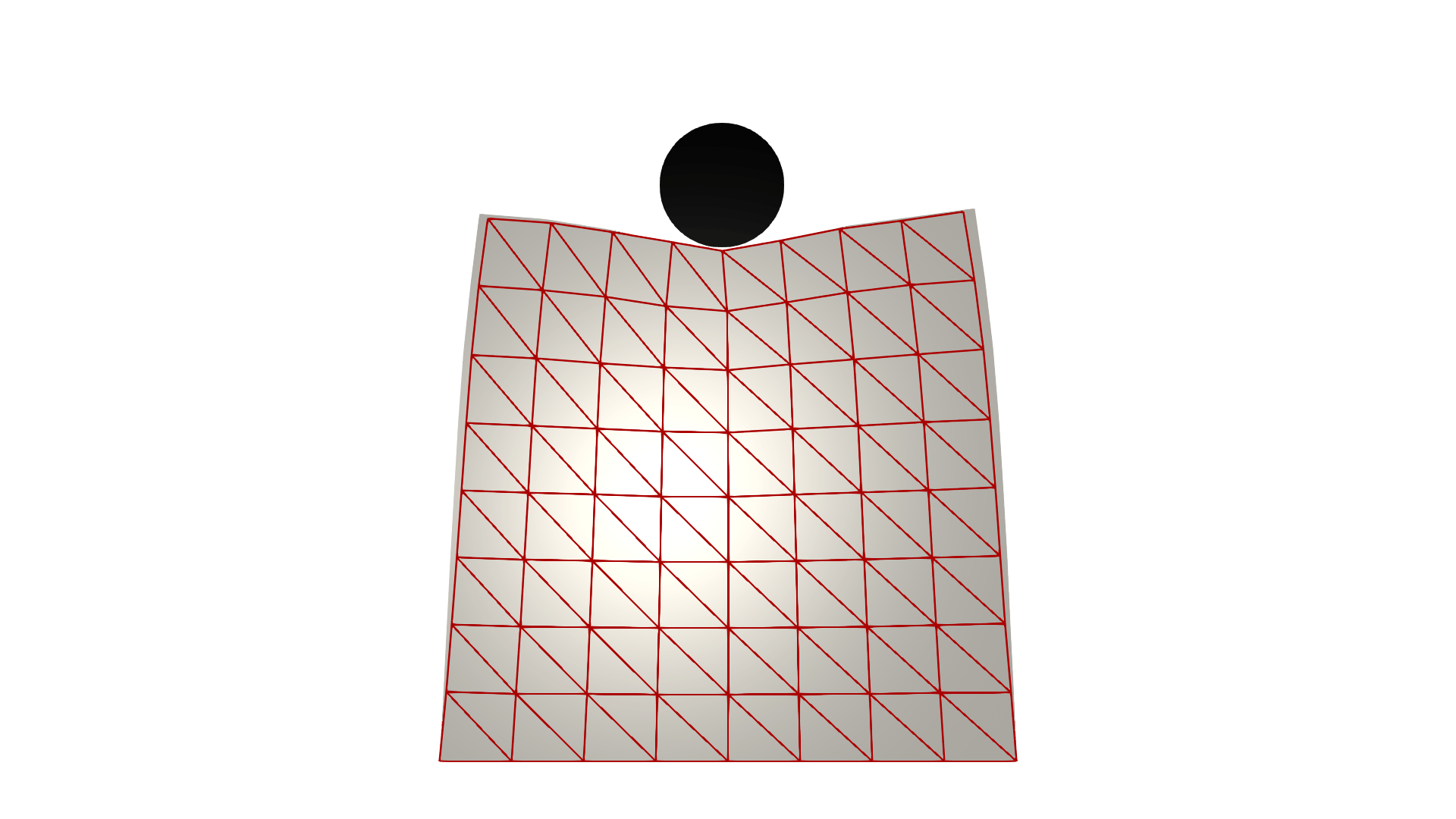}\hspace{-1pt}%
    \img{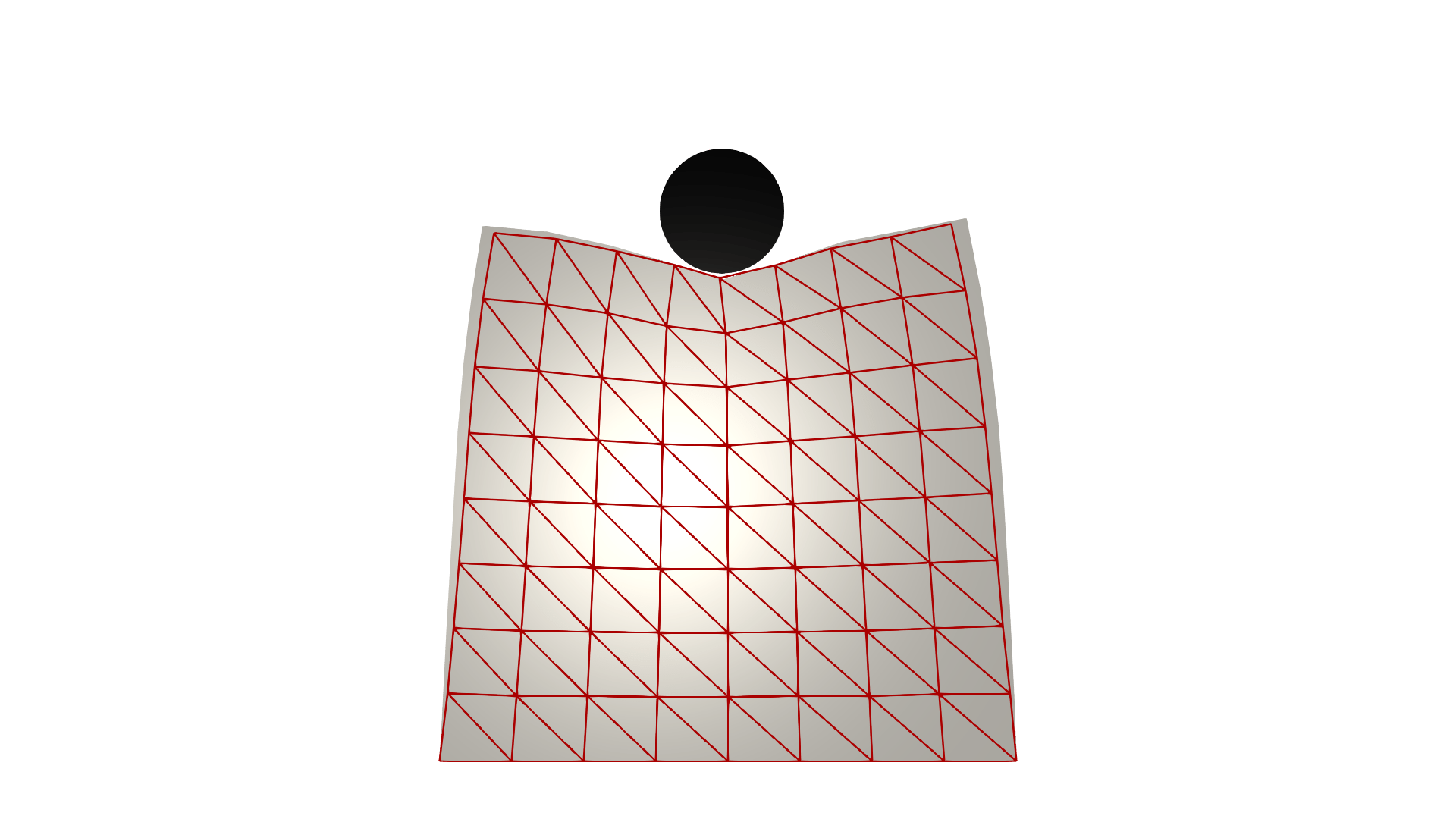}\hspace{-1pt}%
    \img{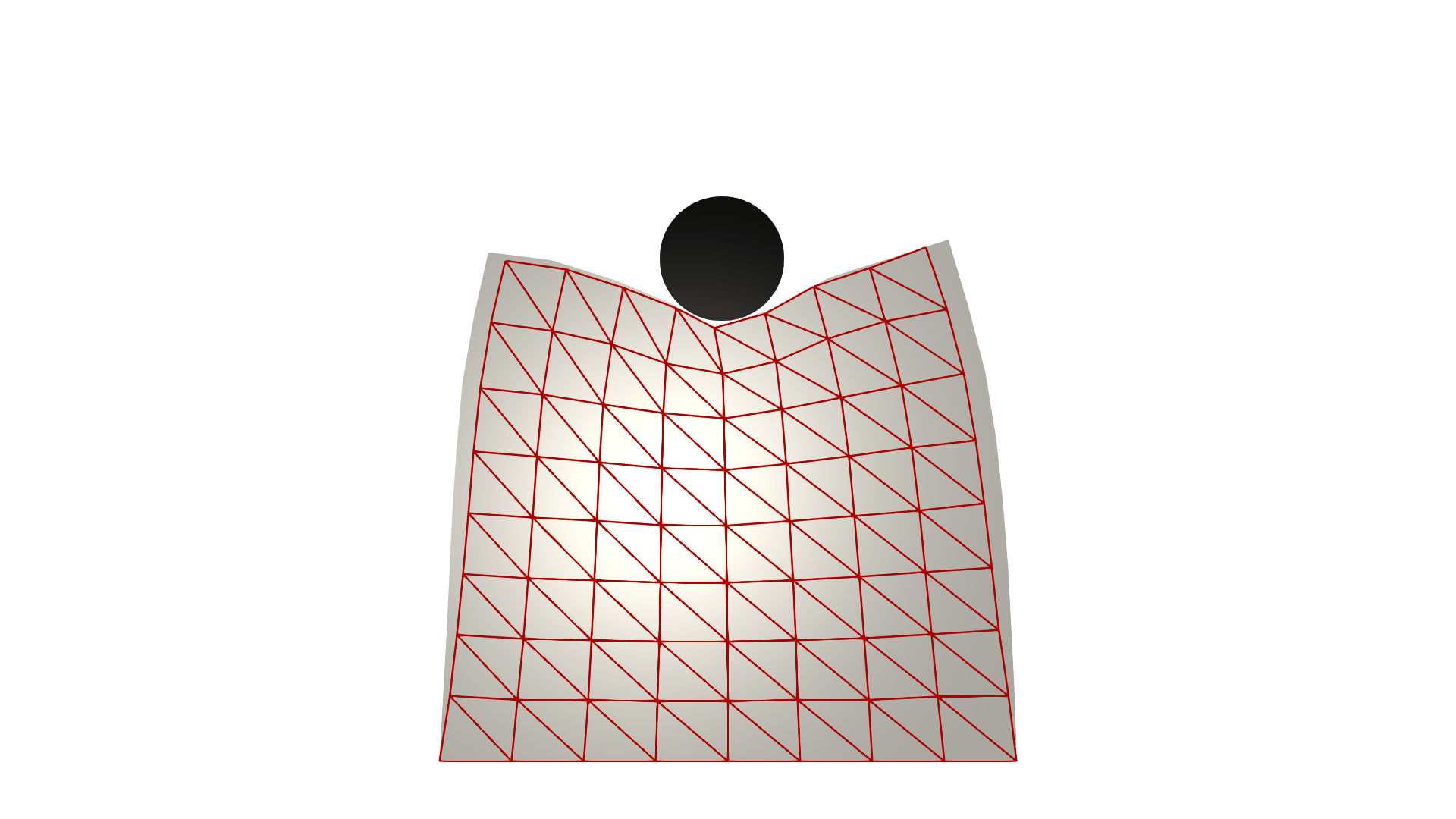}\hspace{-1pt}%
    \img{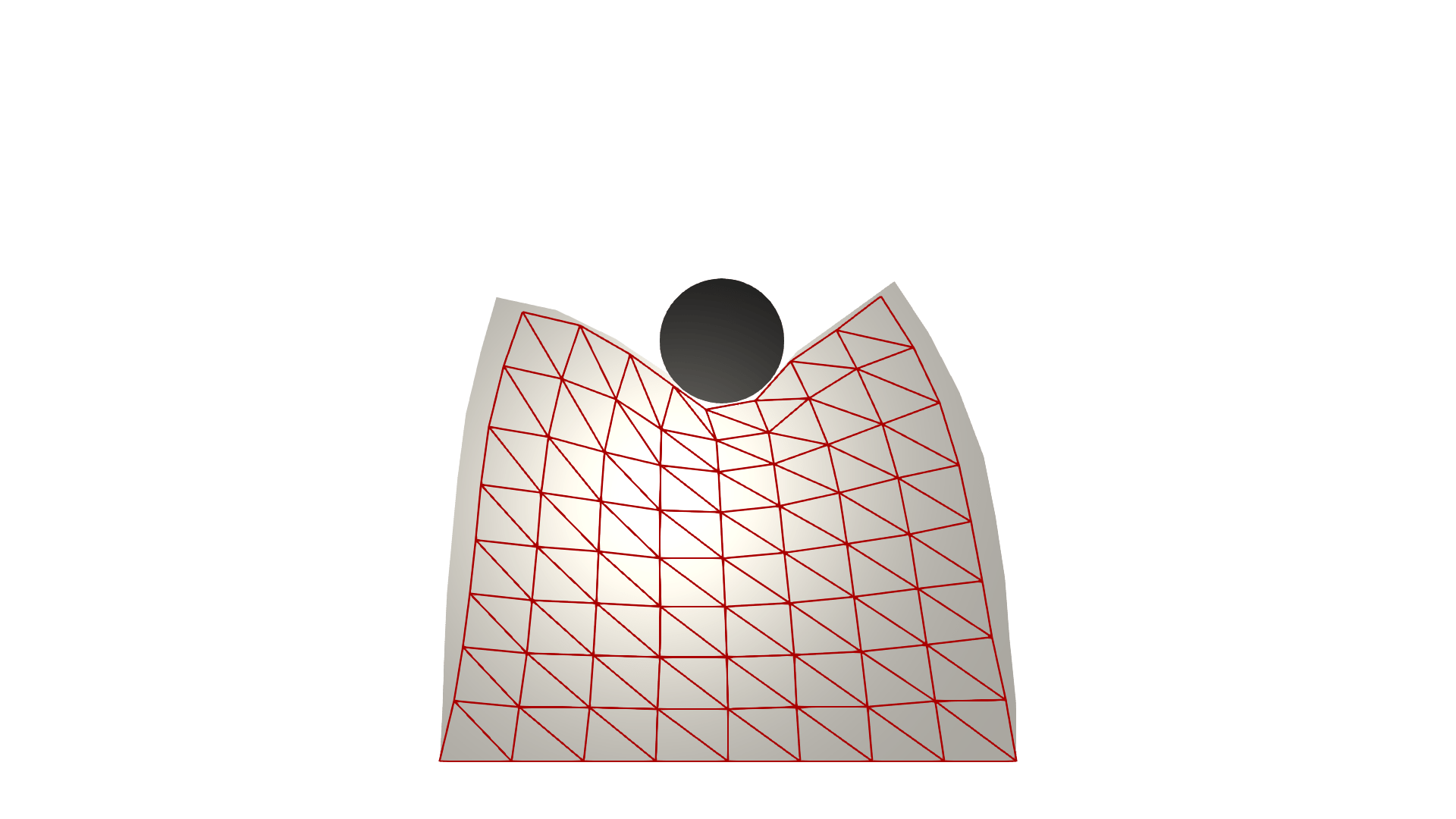}\\[0.4em]

    % Row 3: MANGO Decoder
    \rowlabel{MANGO}%
    \img{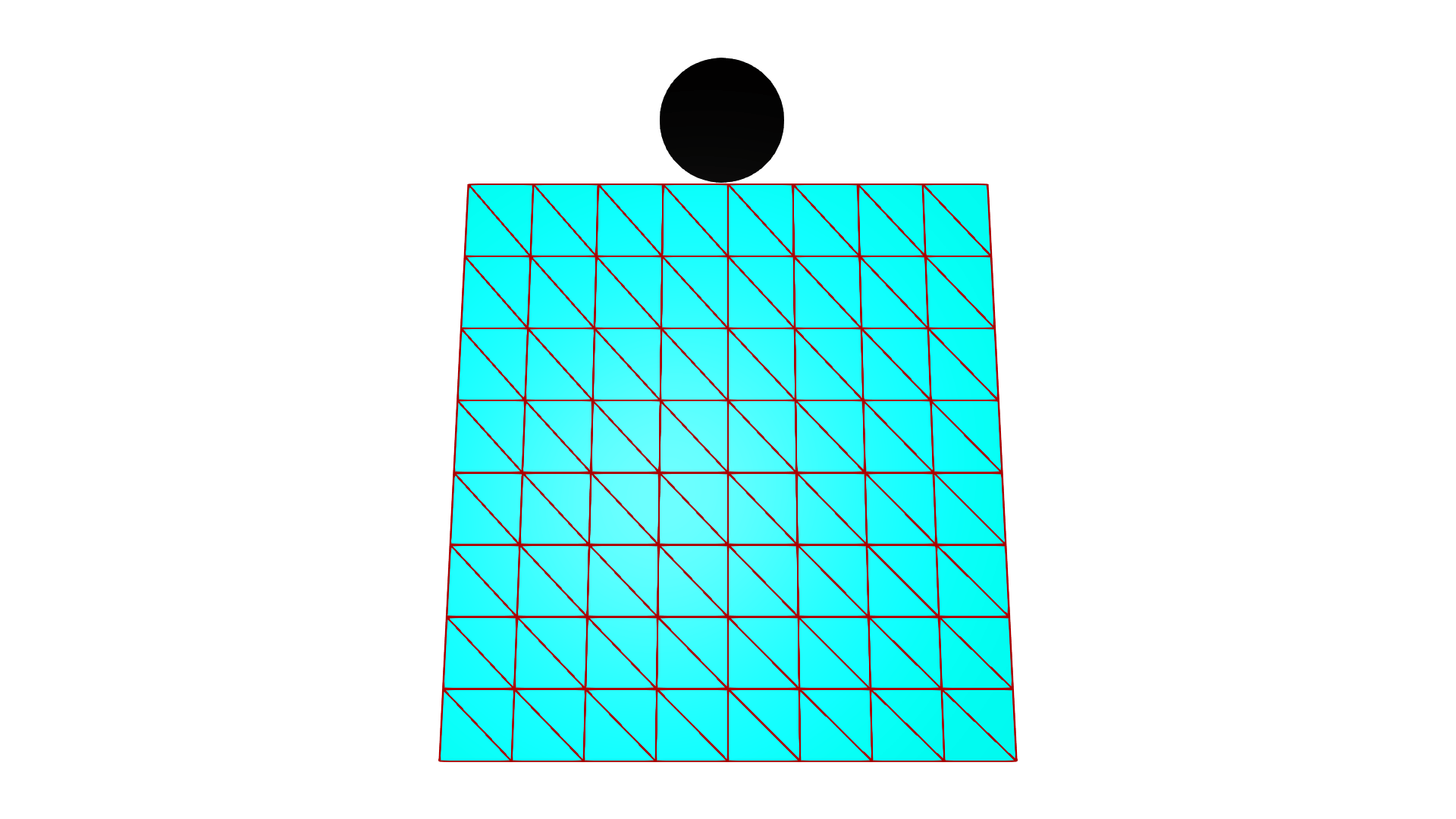}\hspace{-1pt}%
    \img{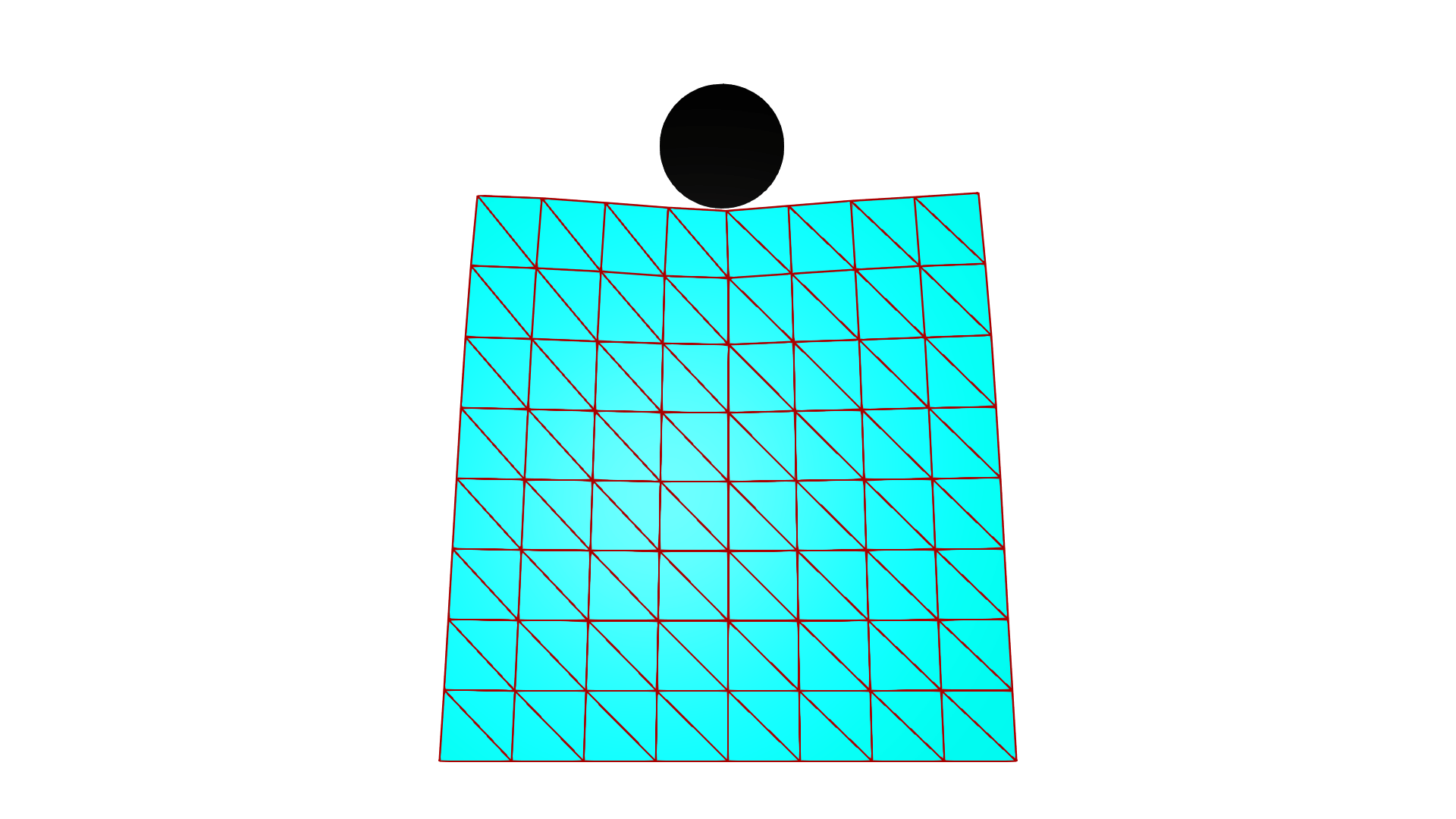}\hspace{-1pt}%
    \img{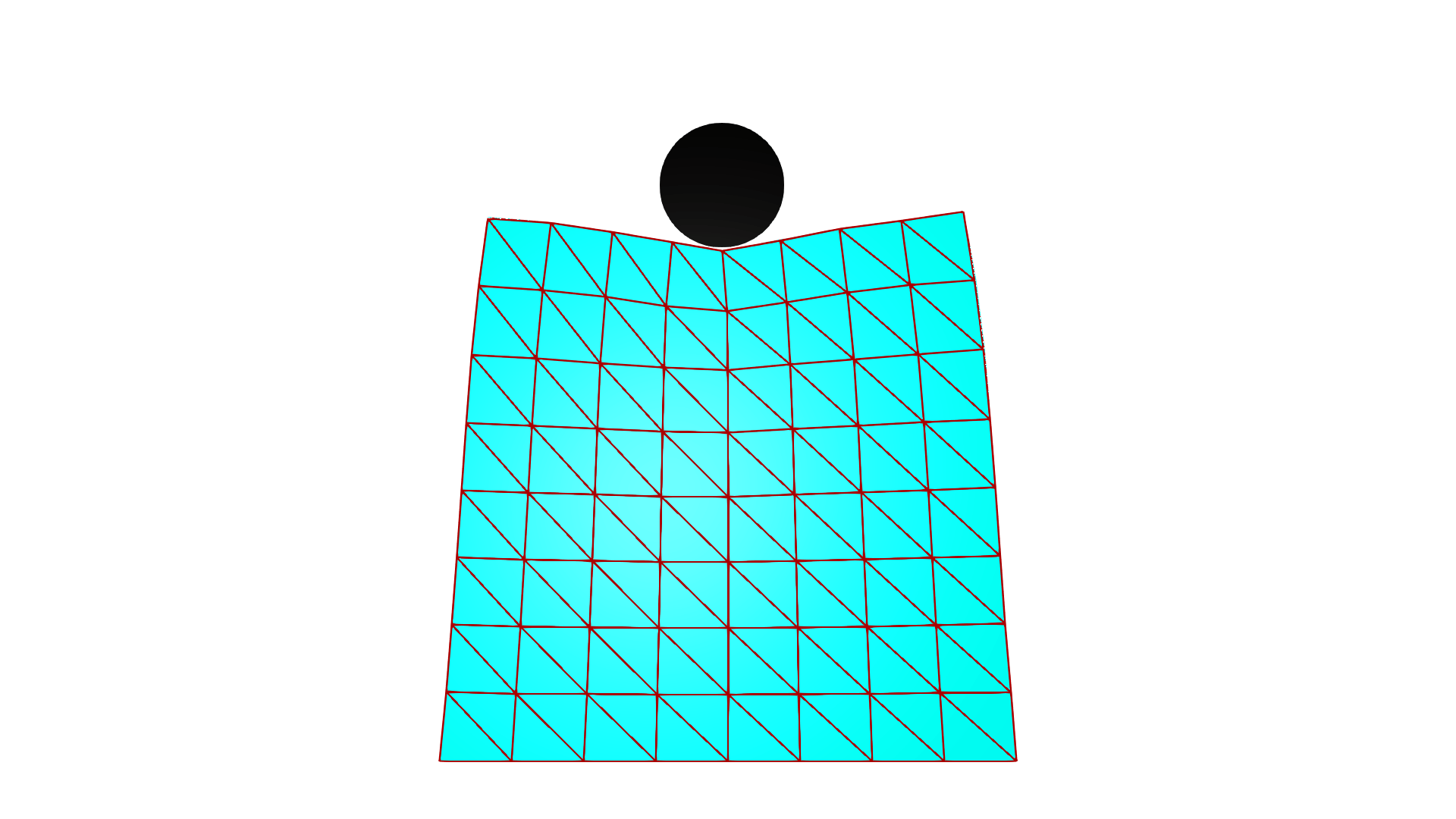}\hspace{-1pt}%
    \img{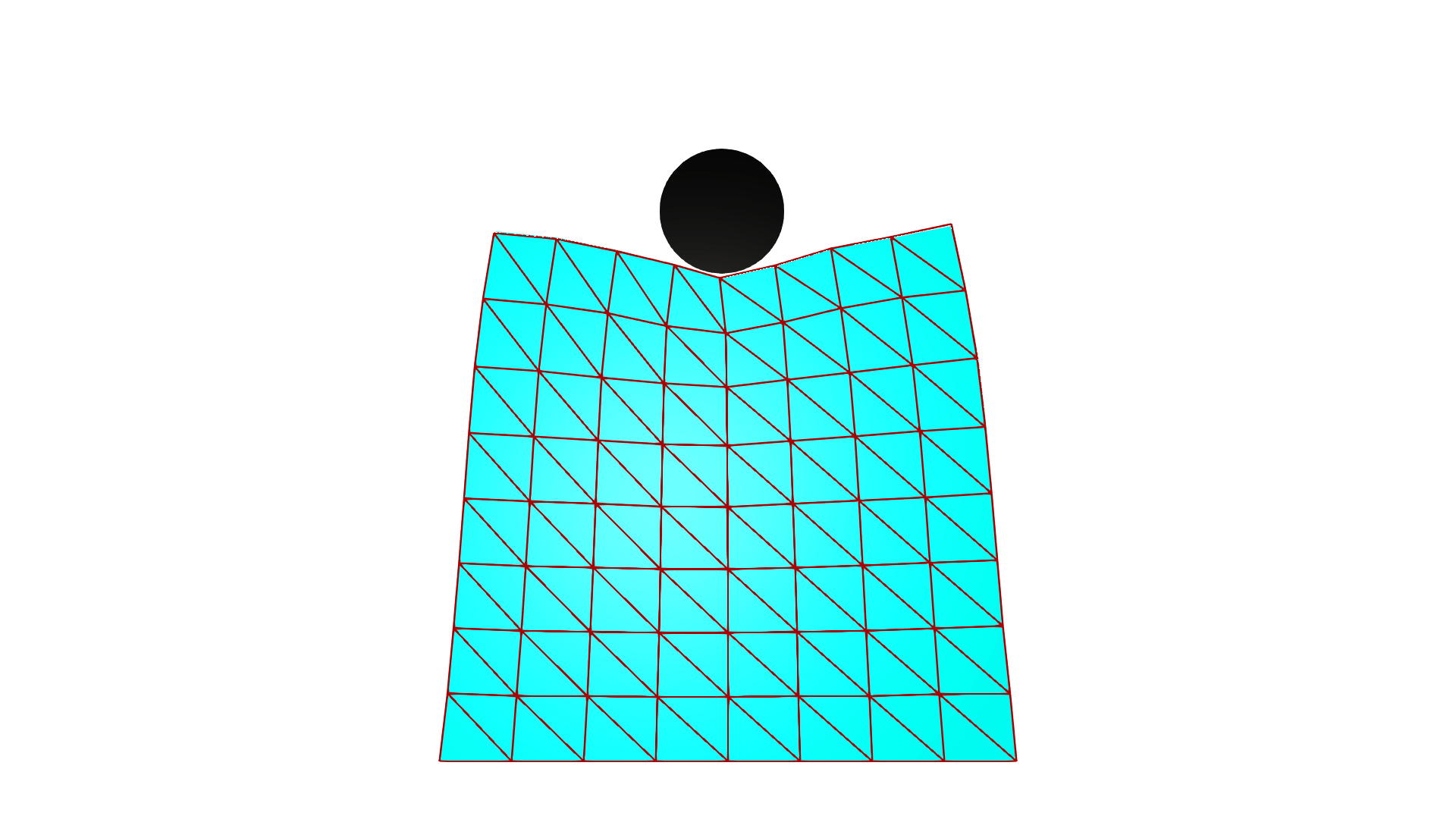}\hspace{-1pt}%
    \img{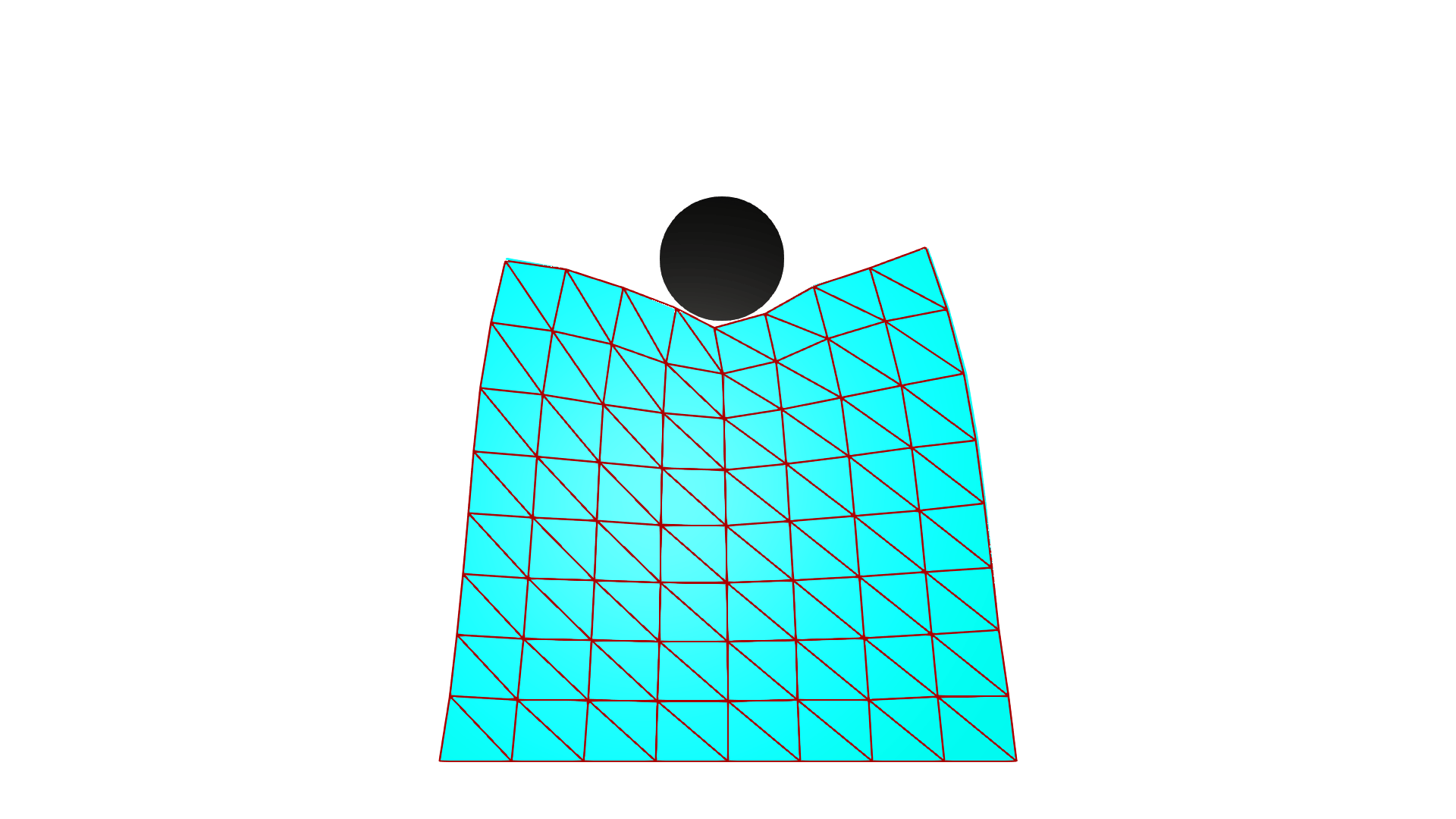}\hspace{-1pt}%
    \img{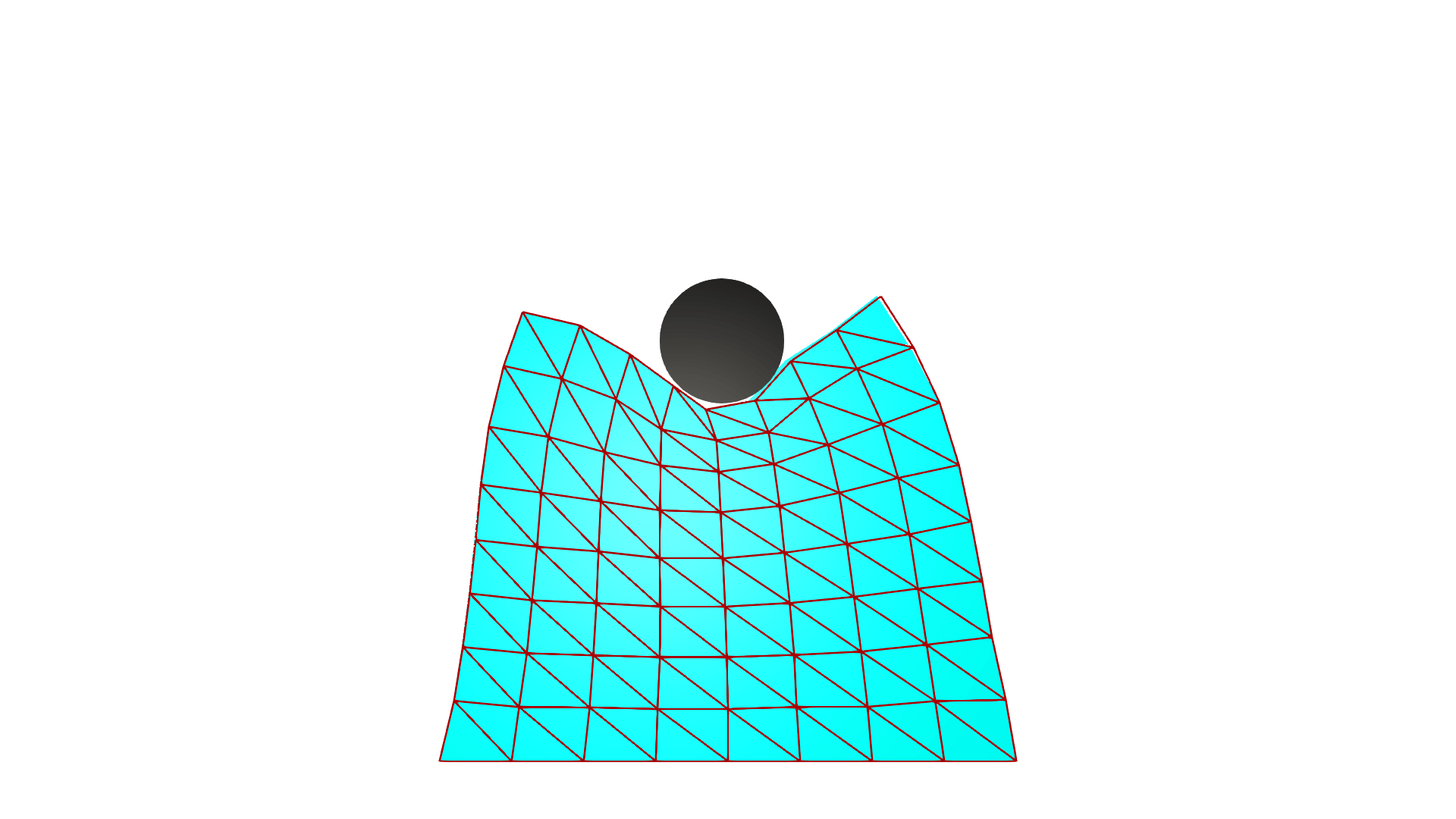}\\[0.4em]

    % Row 4: MANGO Oracle
    \rowlabel{Oracle}%
    \img{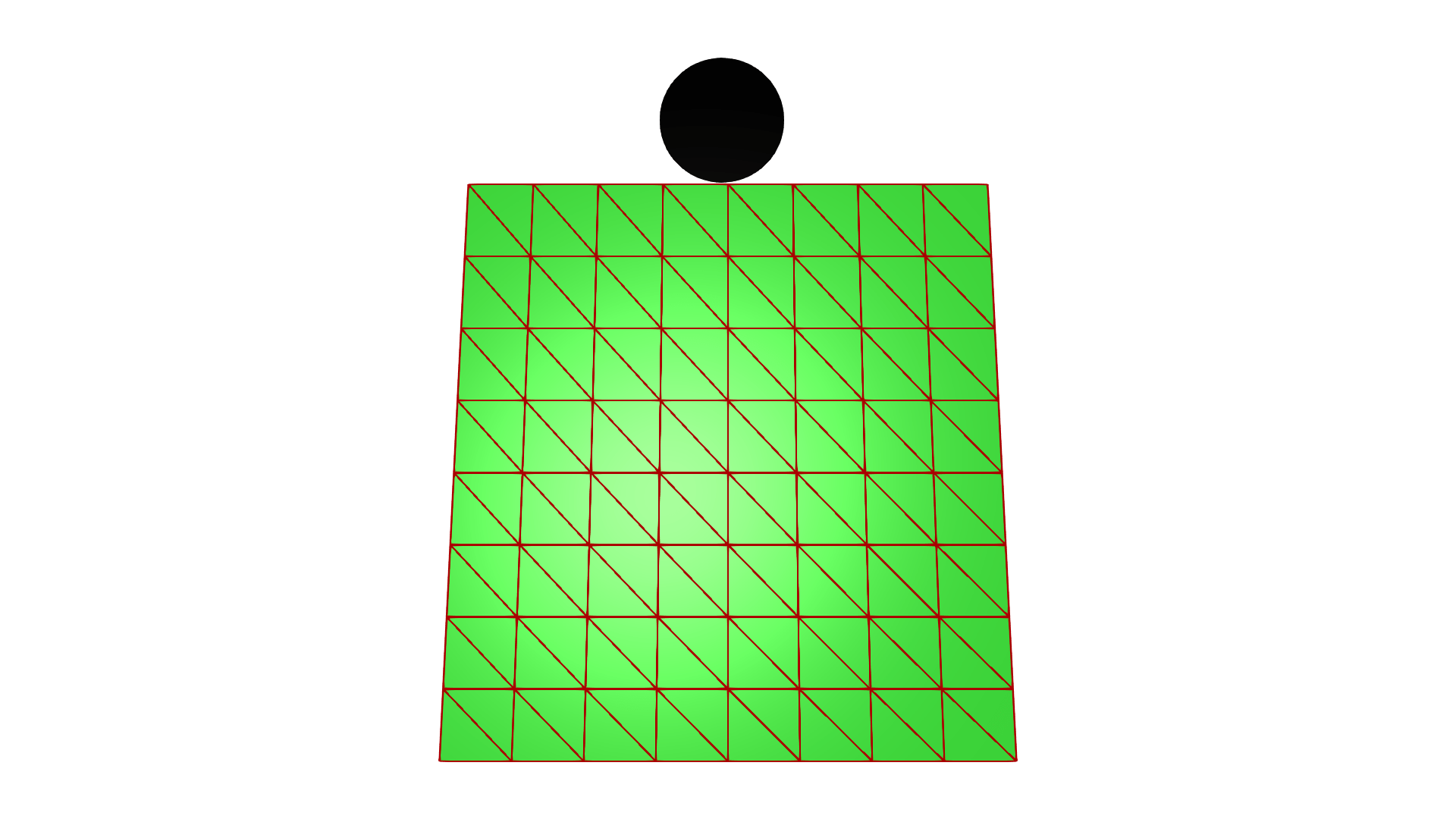}\hspace{-1pt}%
    \img{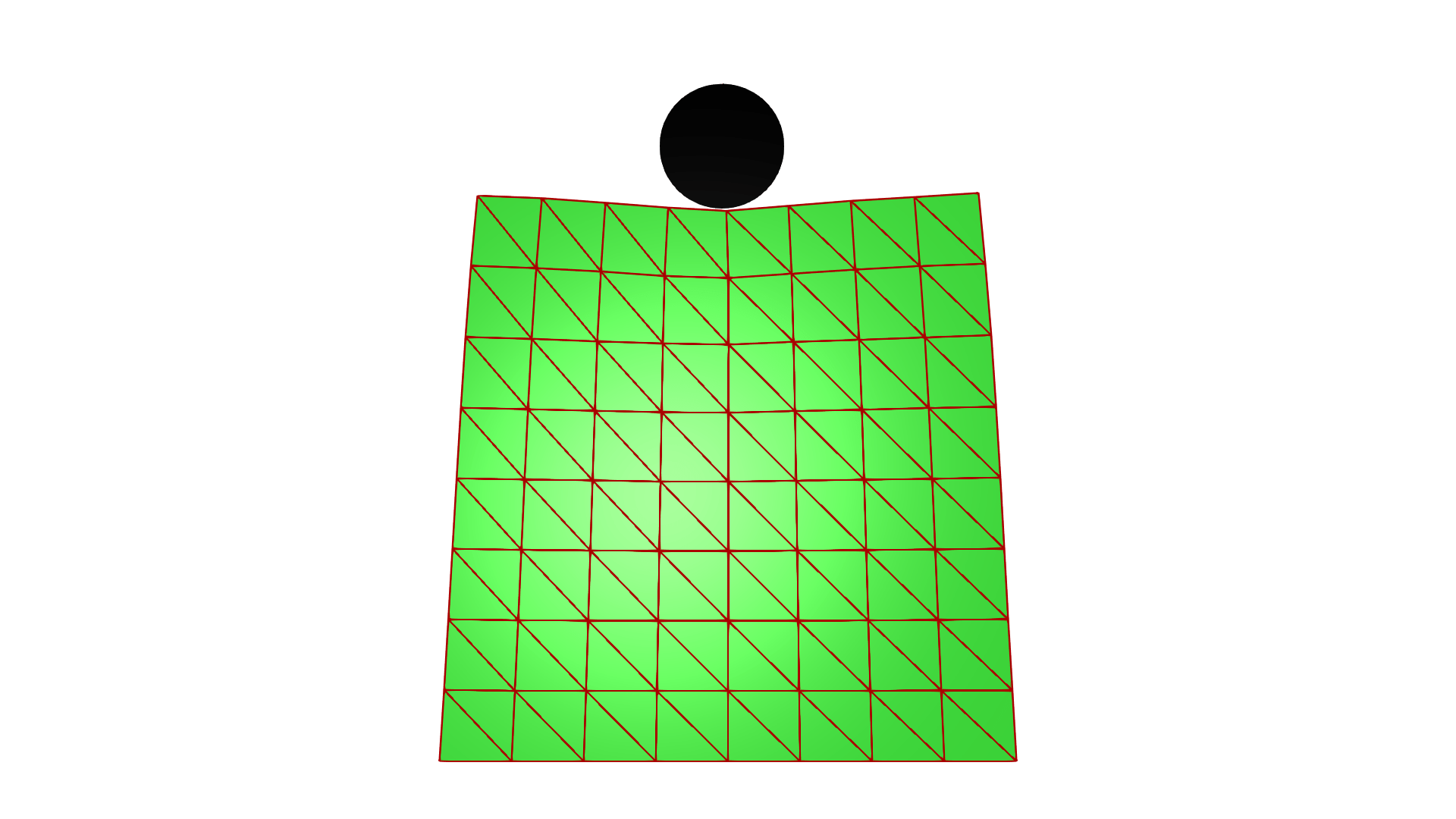}\hspace{-1pt}%
    \img{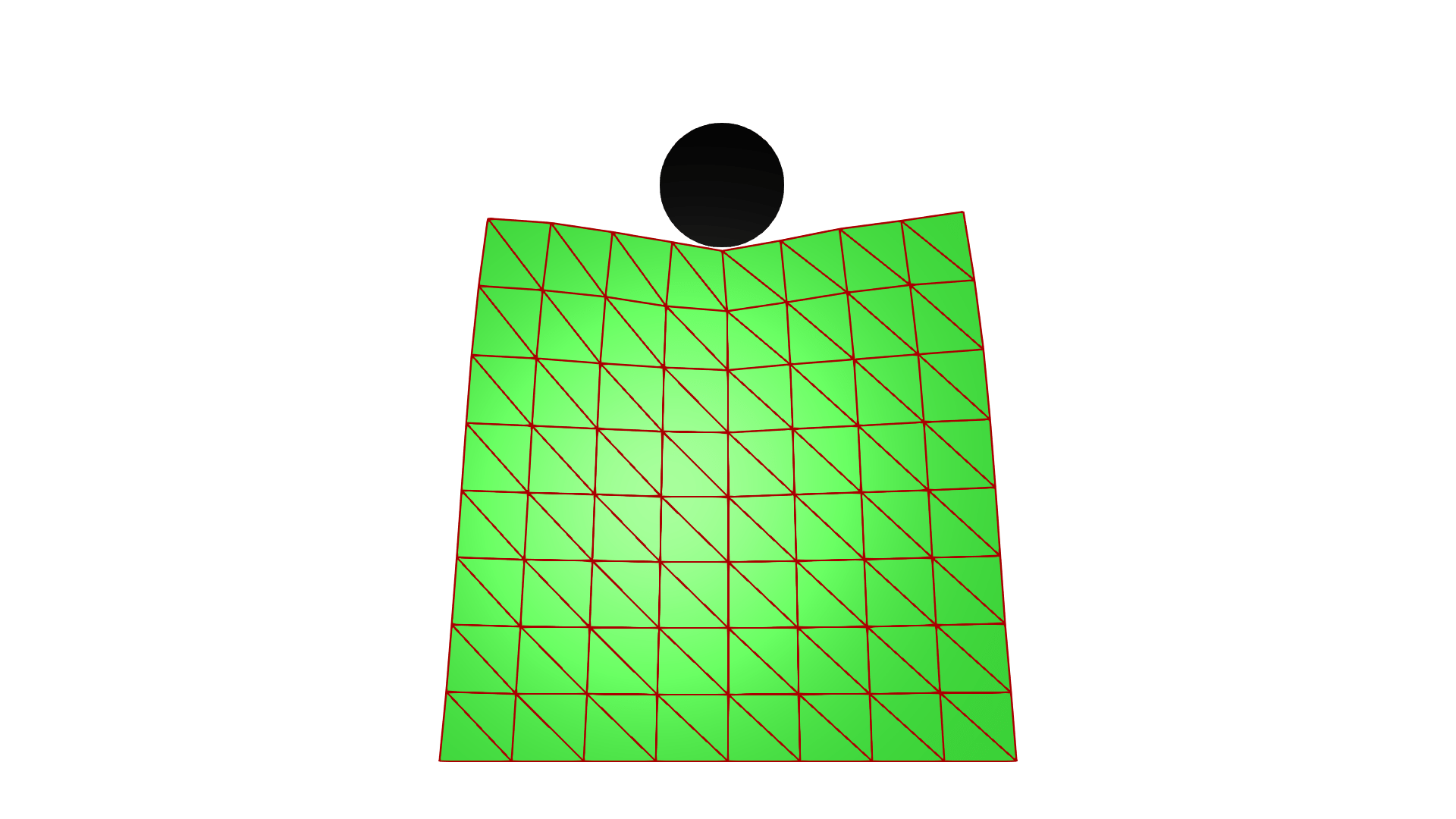}\hspace{-1pt}%
    \img{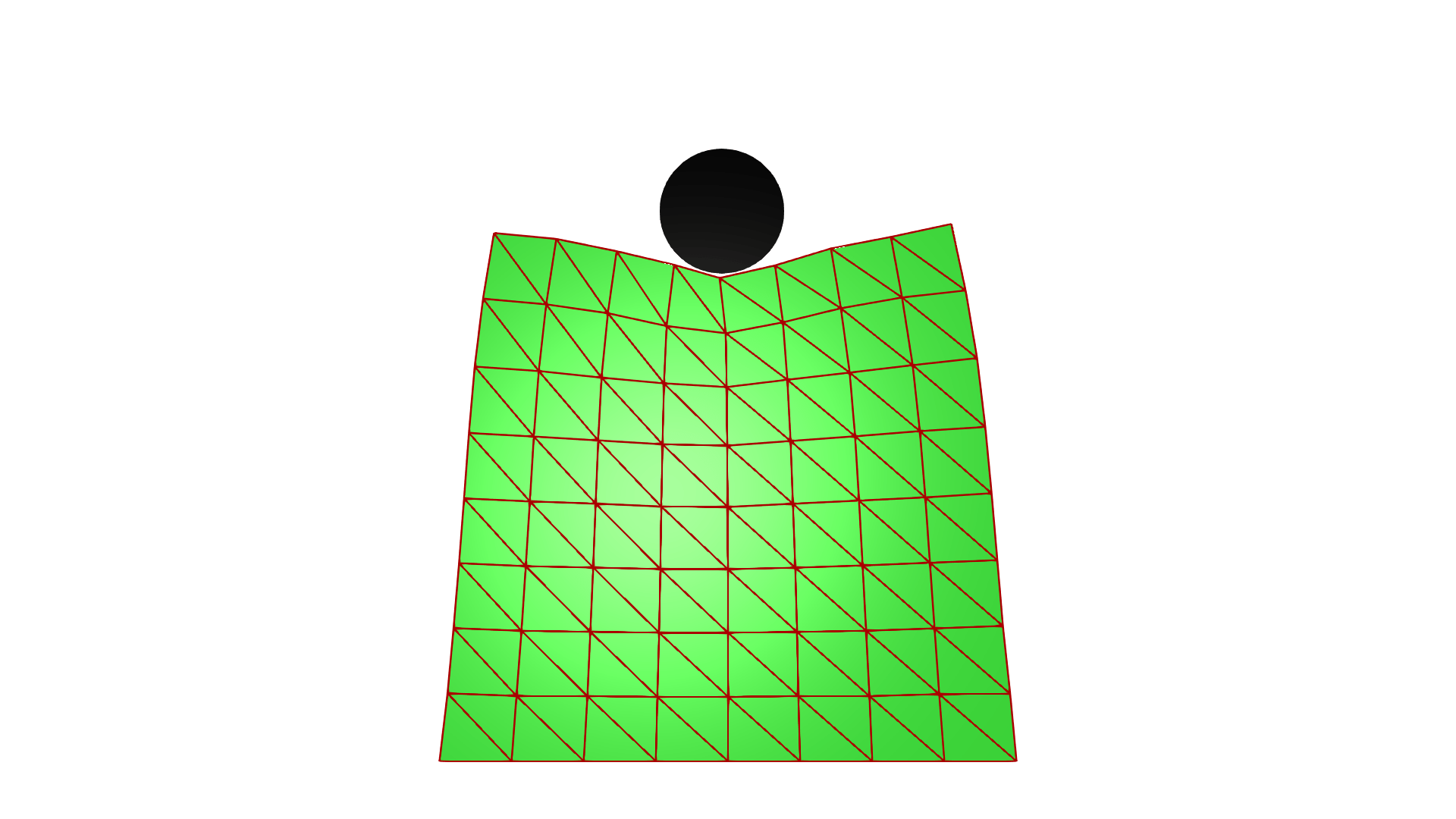}\hspace{-1pt}%
    \img{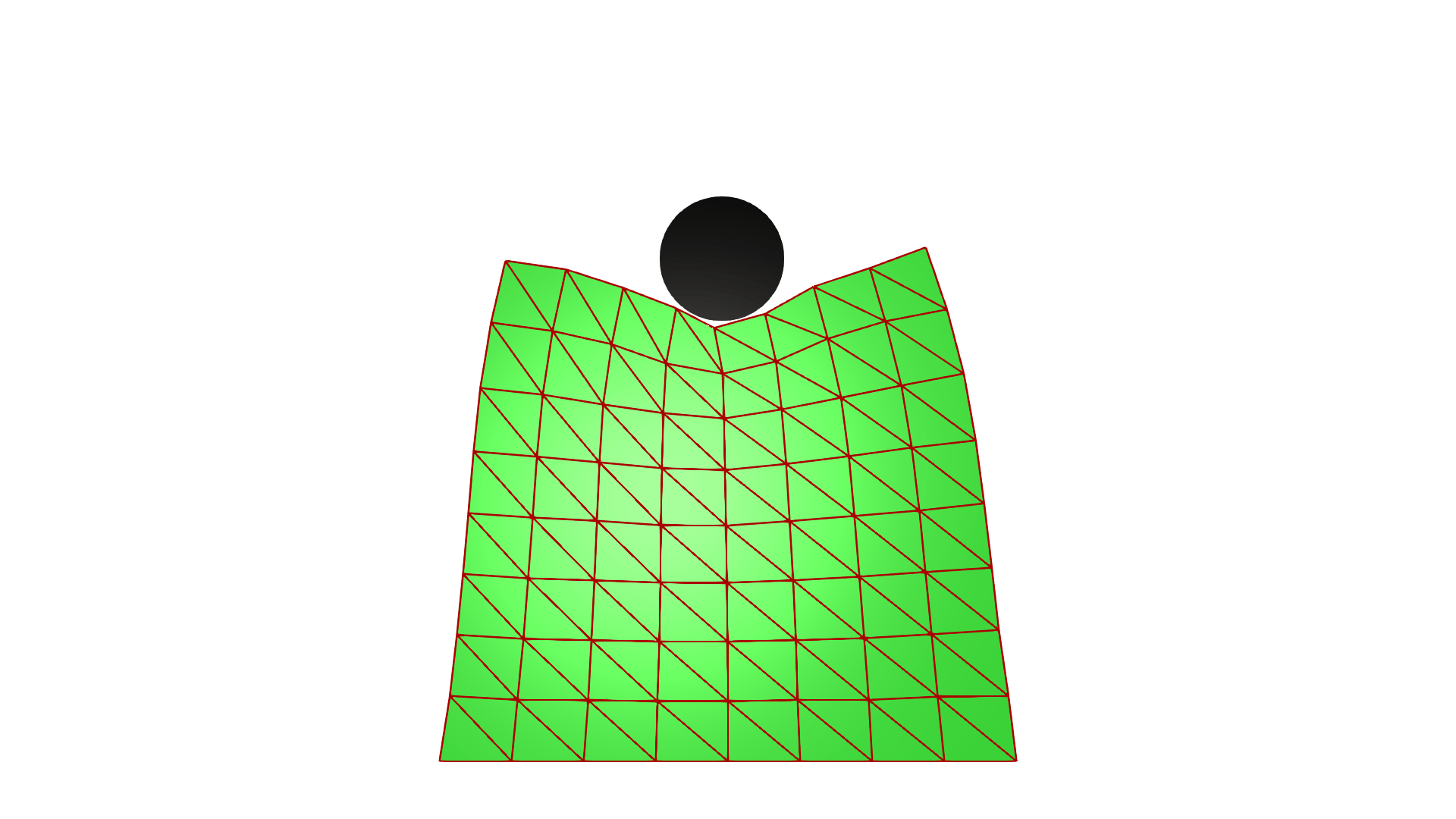}\hspace{-1pt}%
    \img{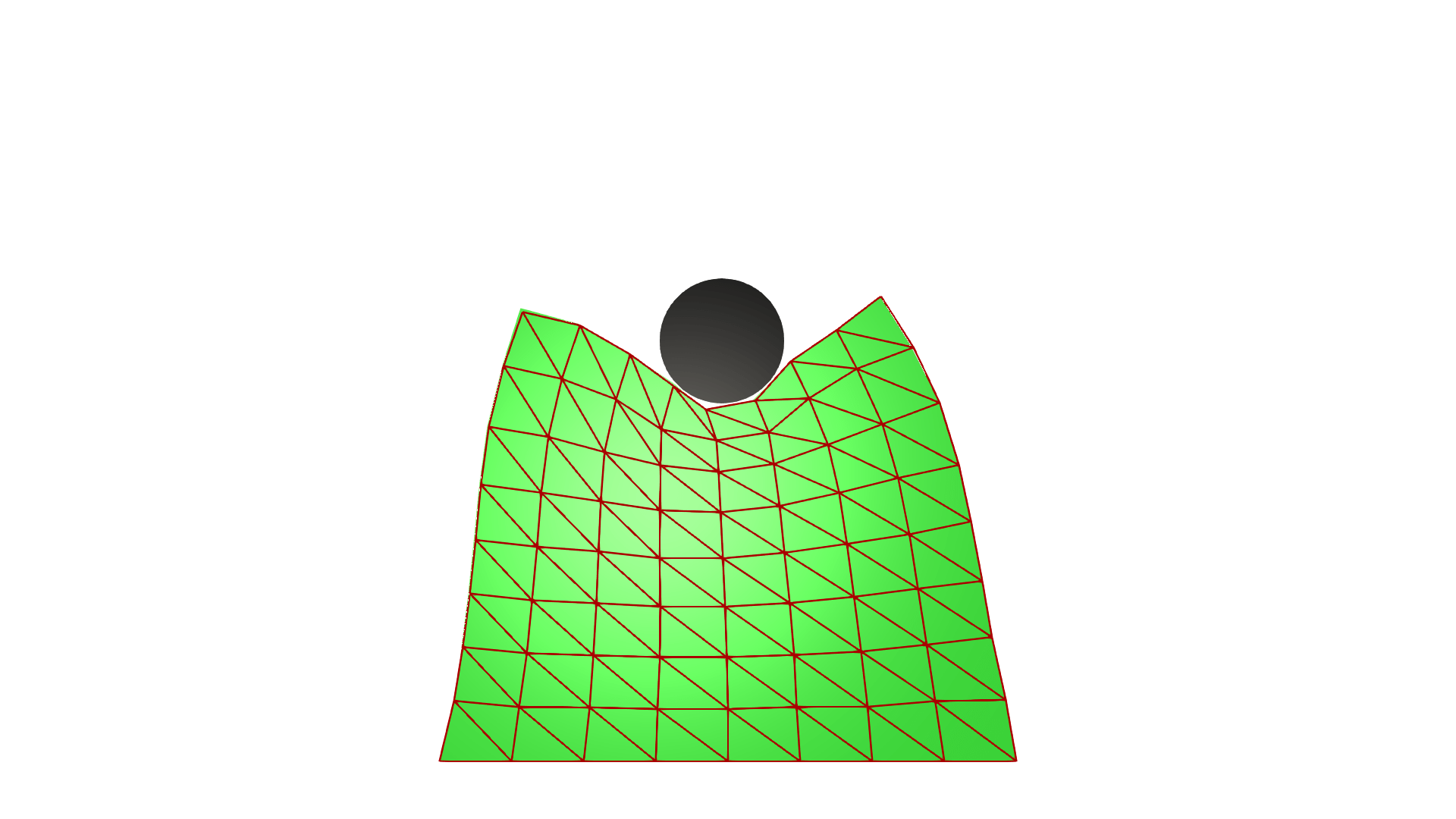}\\[0.4em]

    % Row 5: MGN
    \rowlabel{No Context \\ (MGN)}%
    \img{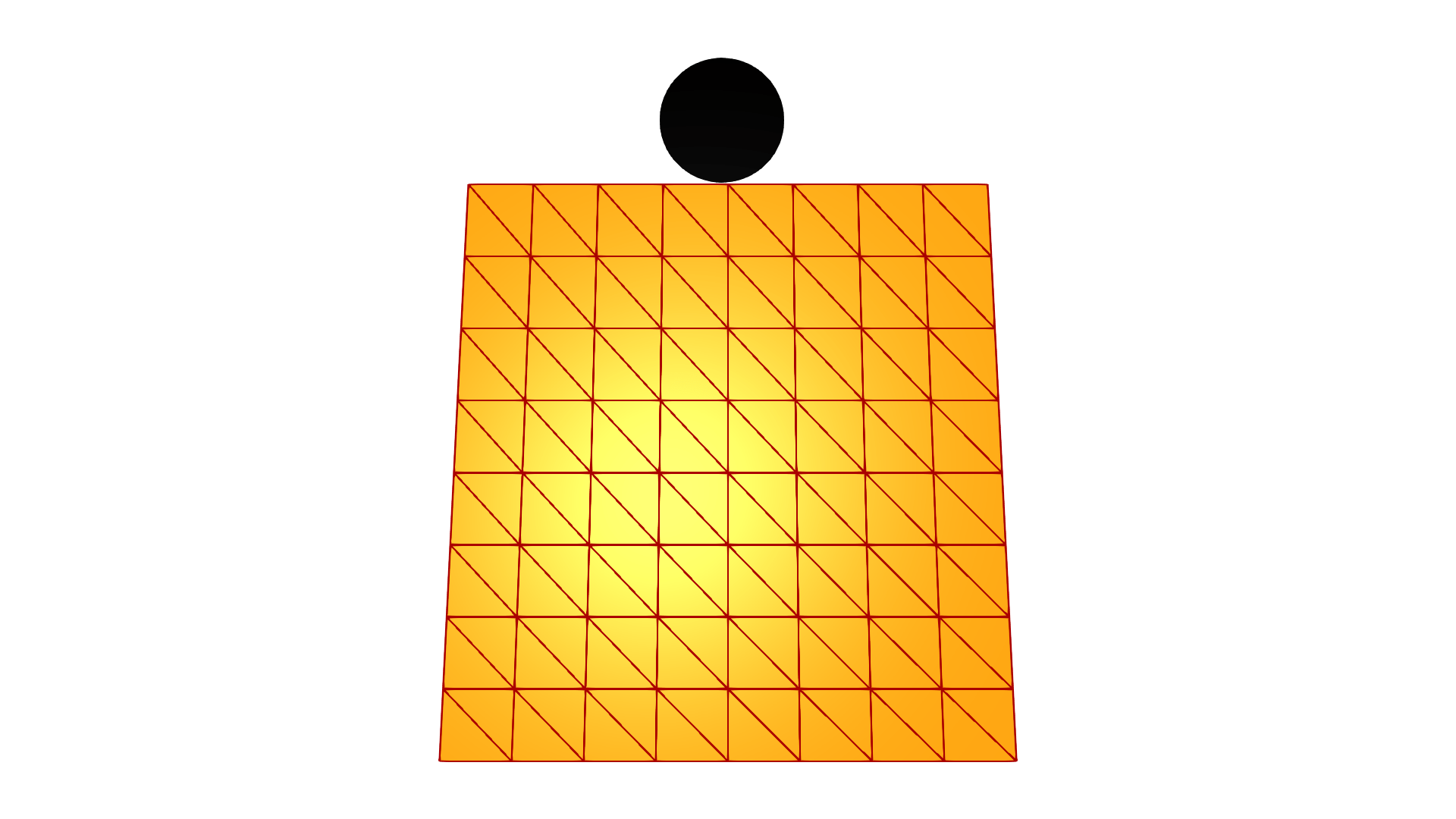}\hspace{-1pt}%
    \img{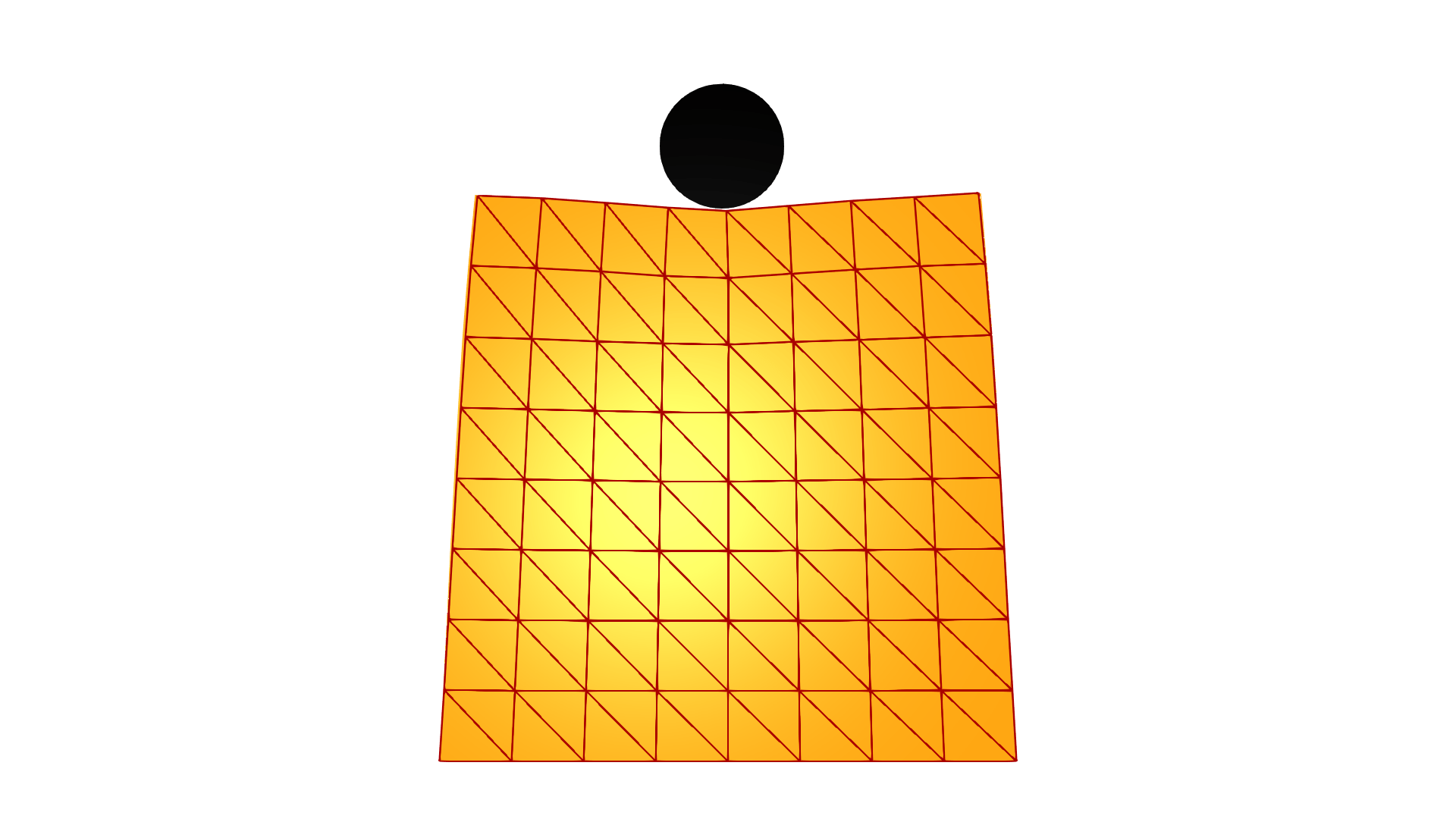}\hspace{-1pt}%
    \img{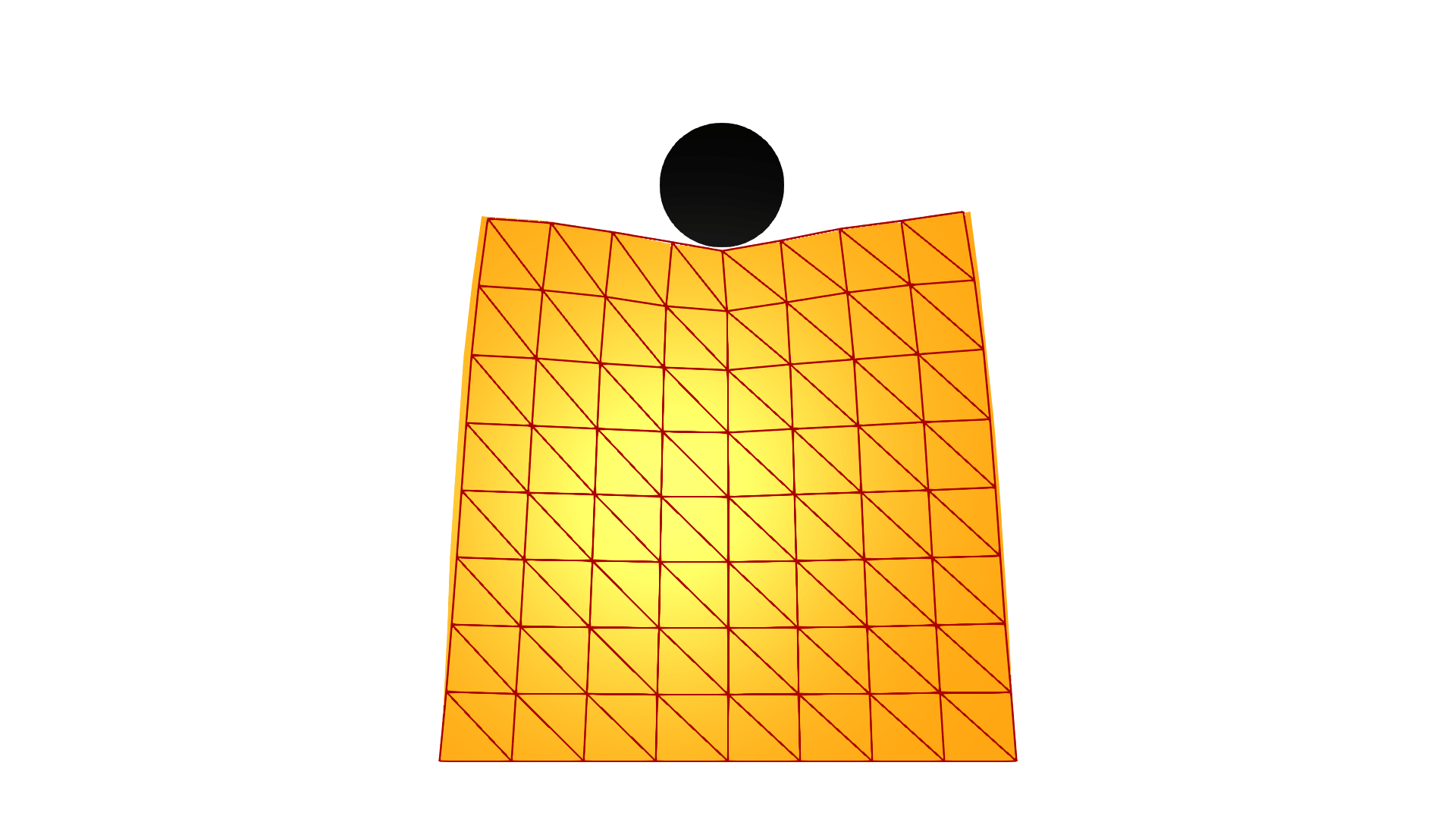}\hspace{-1pt}%
    \img{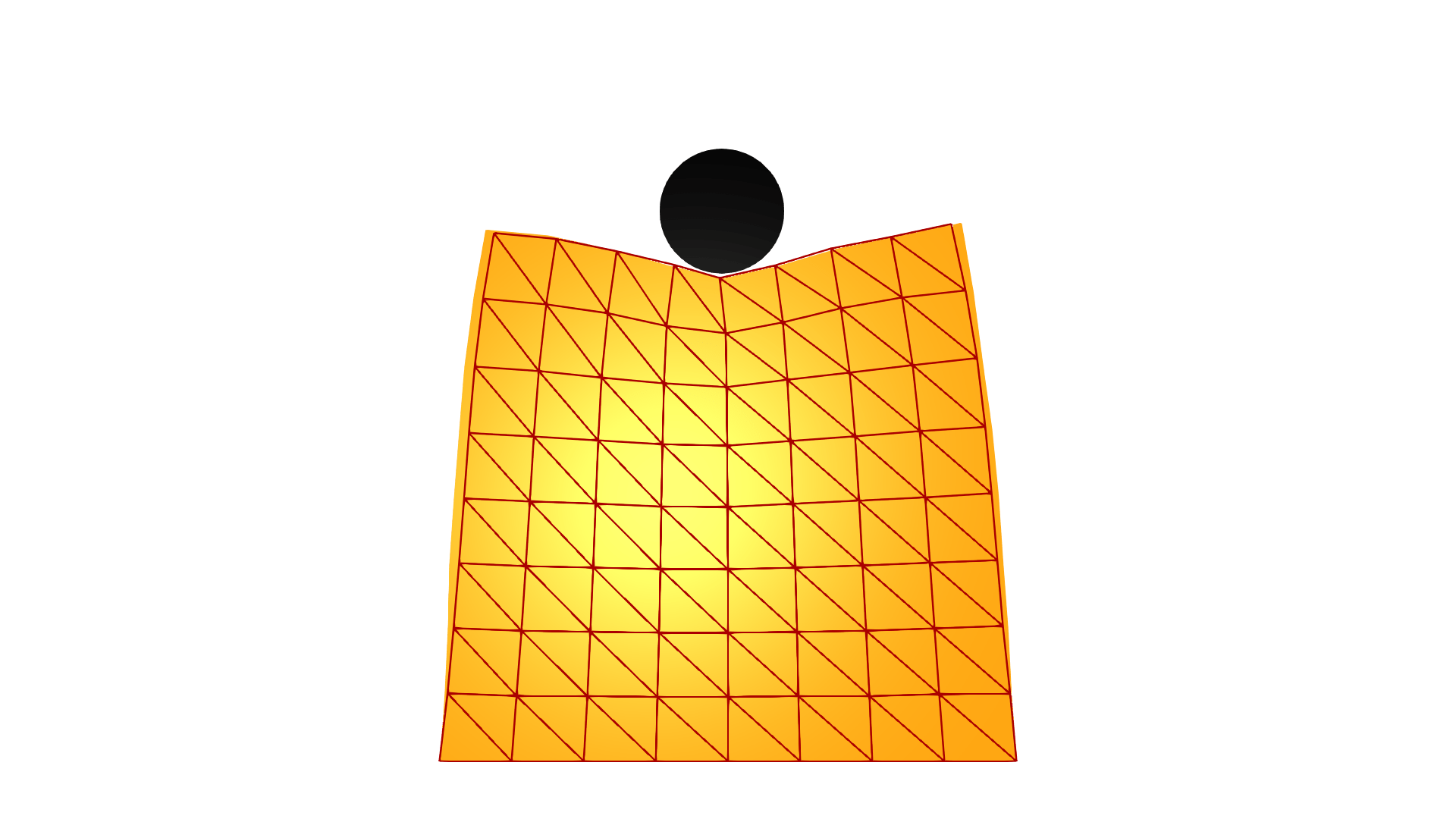}\hspace{-1pt}%
    \img{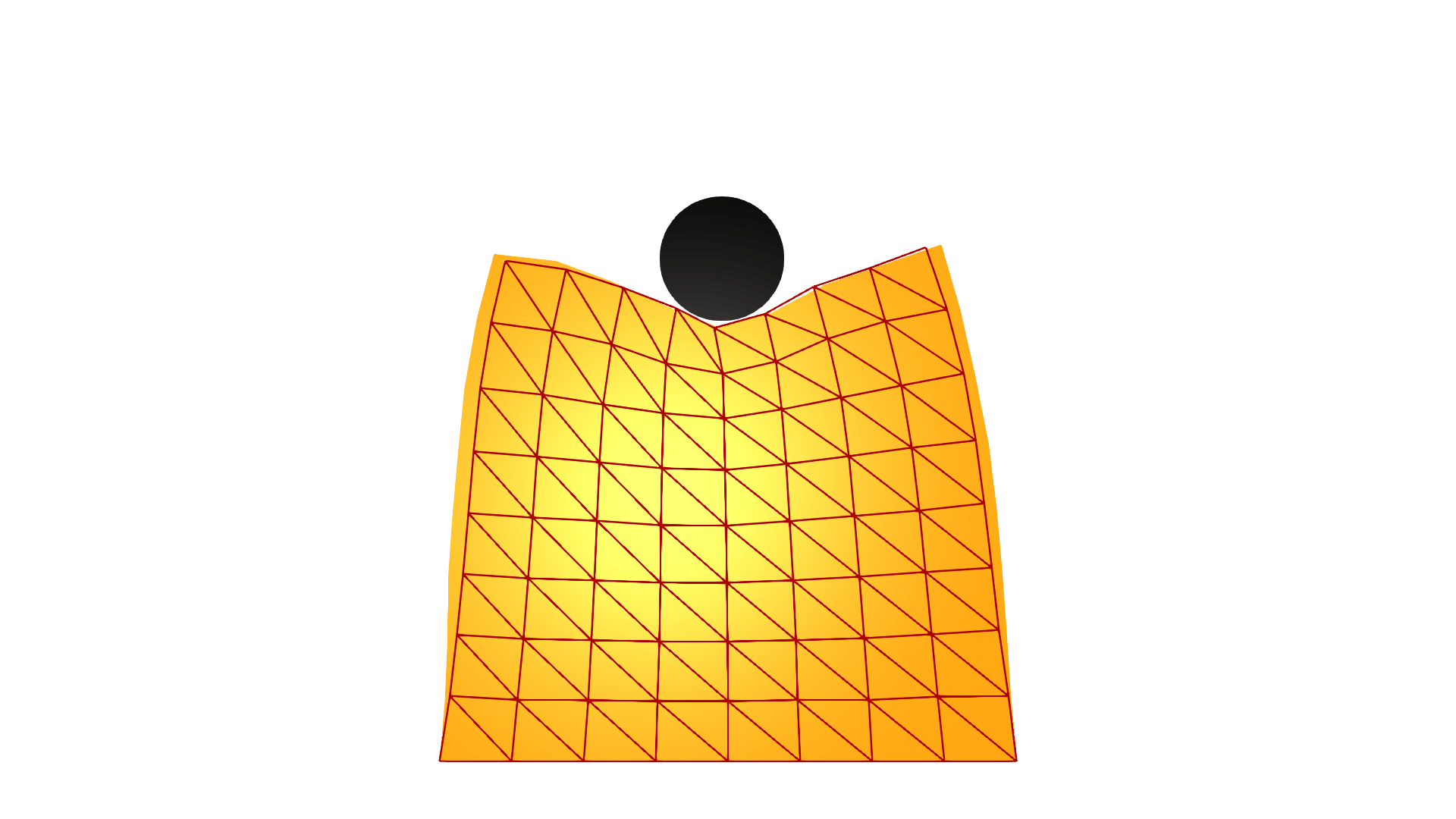}\hspace{-1pt}%
    \img{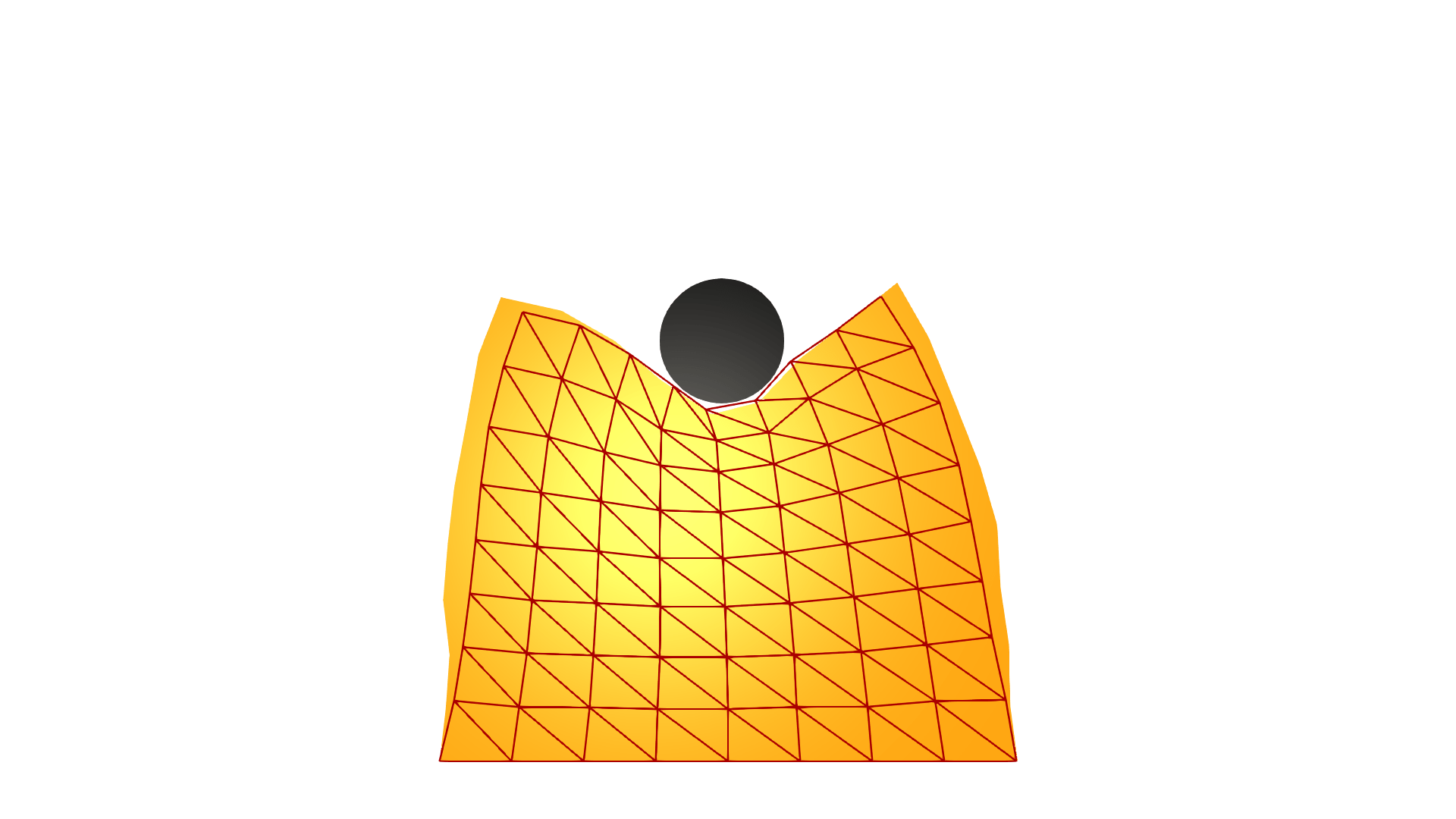}\\[0.4em]

    % Row 6: MGN Oracle
    \rowlabel{Oracle \\ (MGN)}%
    \img{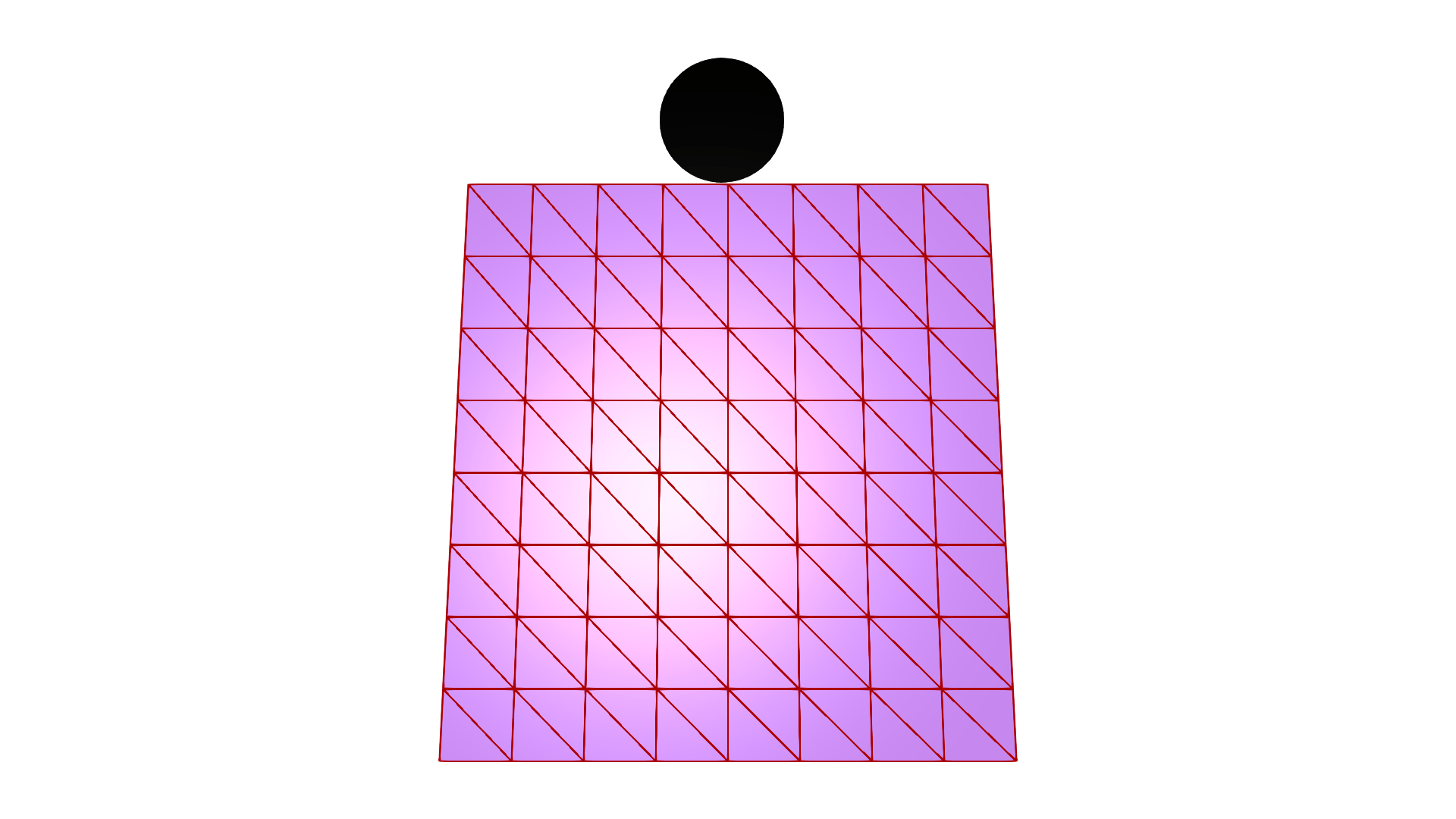}\hspace{-1pt}%
    \img{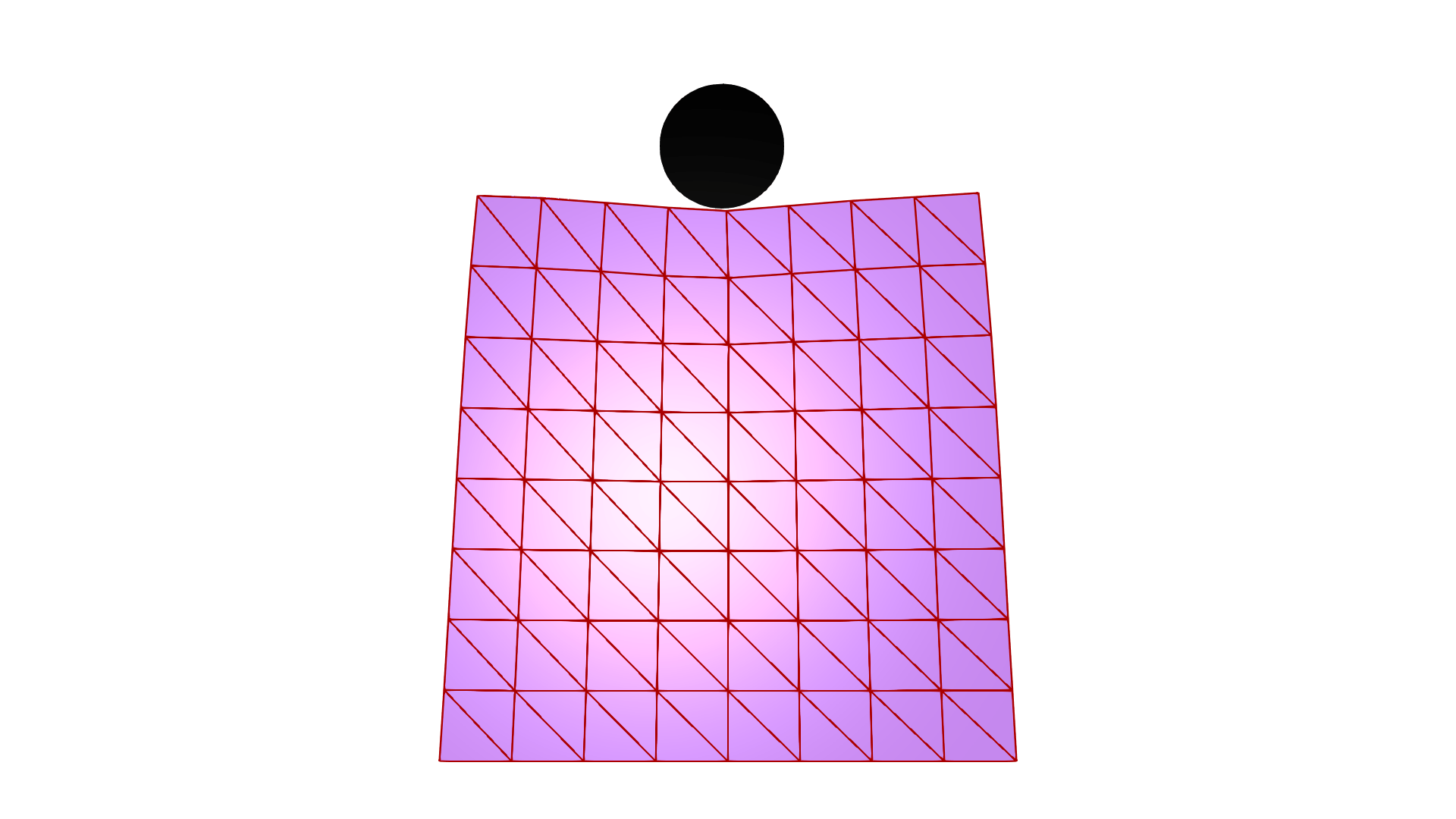}\hspace{-1pt}%
    \img{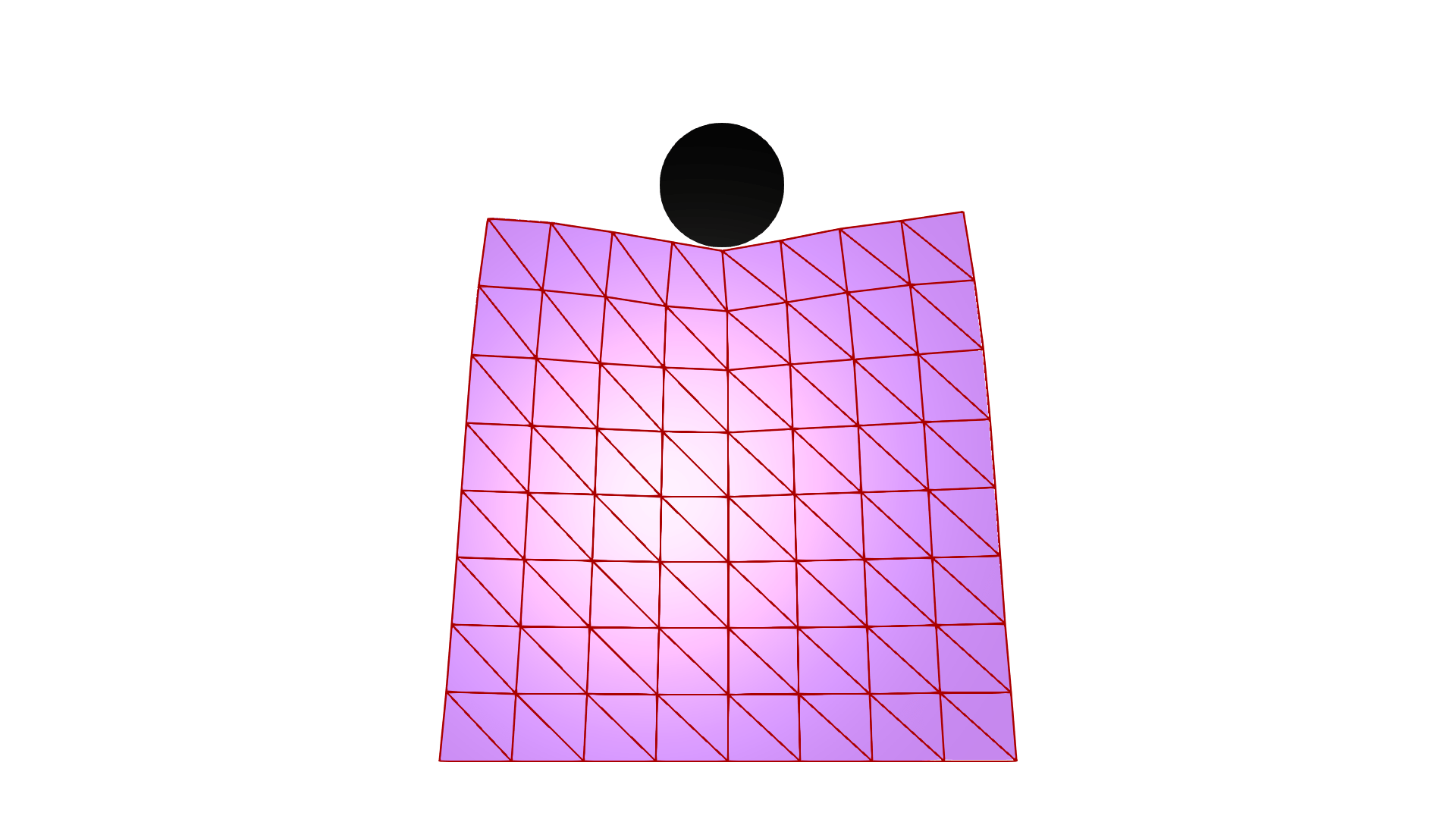}\hspace{-1pt}%
    \img{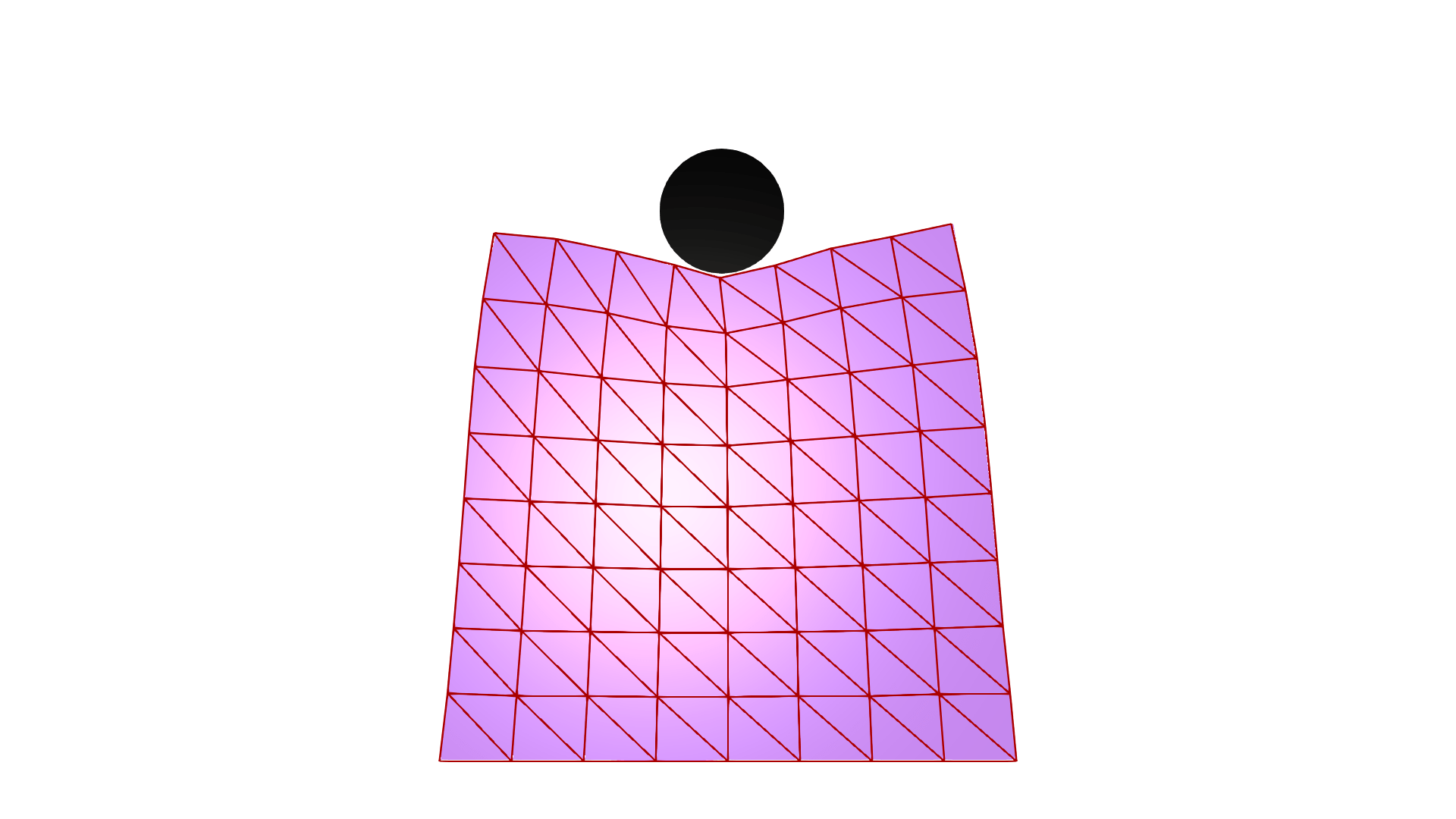}\hspace{-1pt}%
    \img{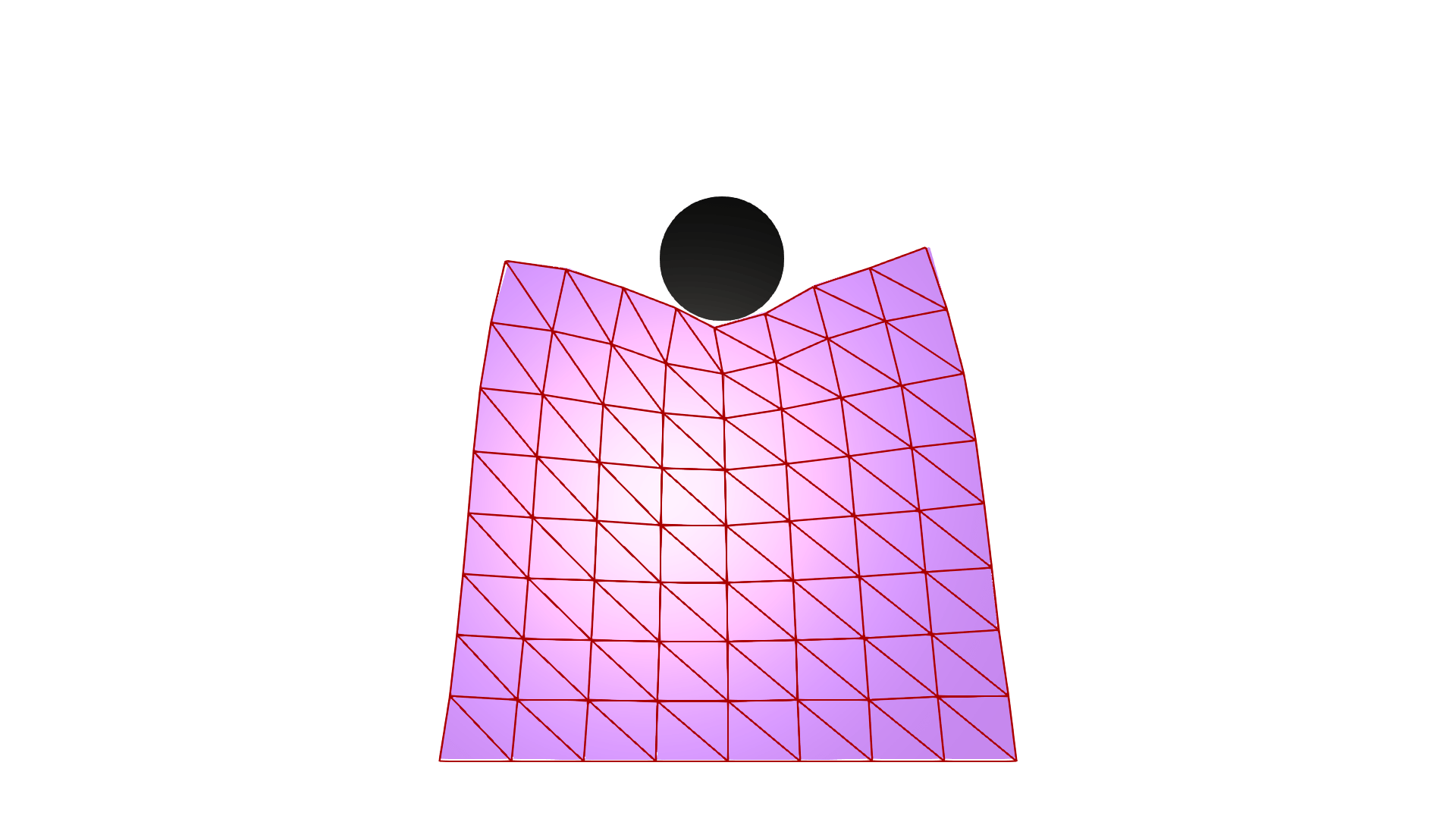}\hspace{-1pt}%
    \img{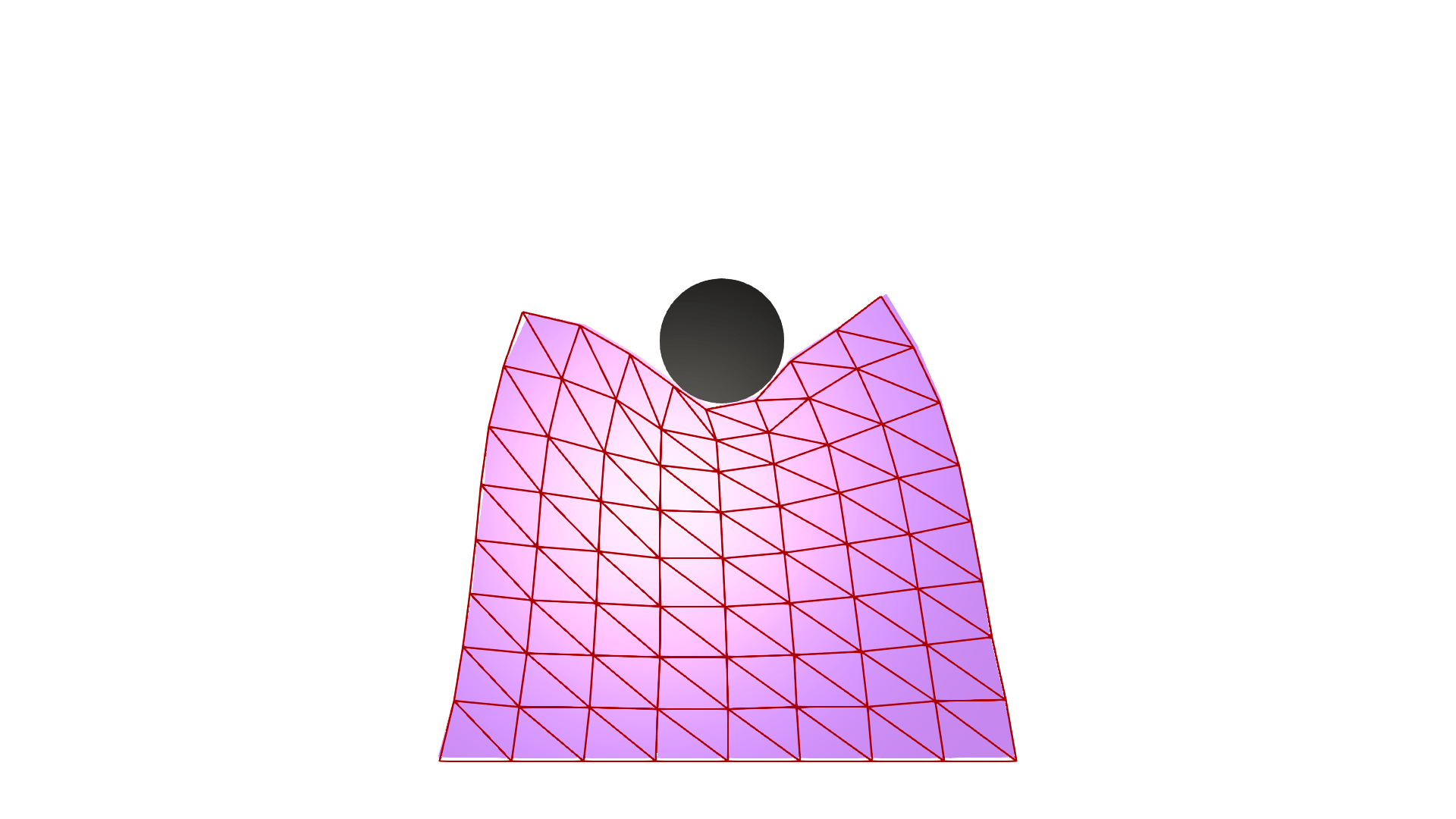}\\[0.4em]

    % Row 7: GNN Encoder
    \rowlabel{GNN \\ Encoder}%
    \img{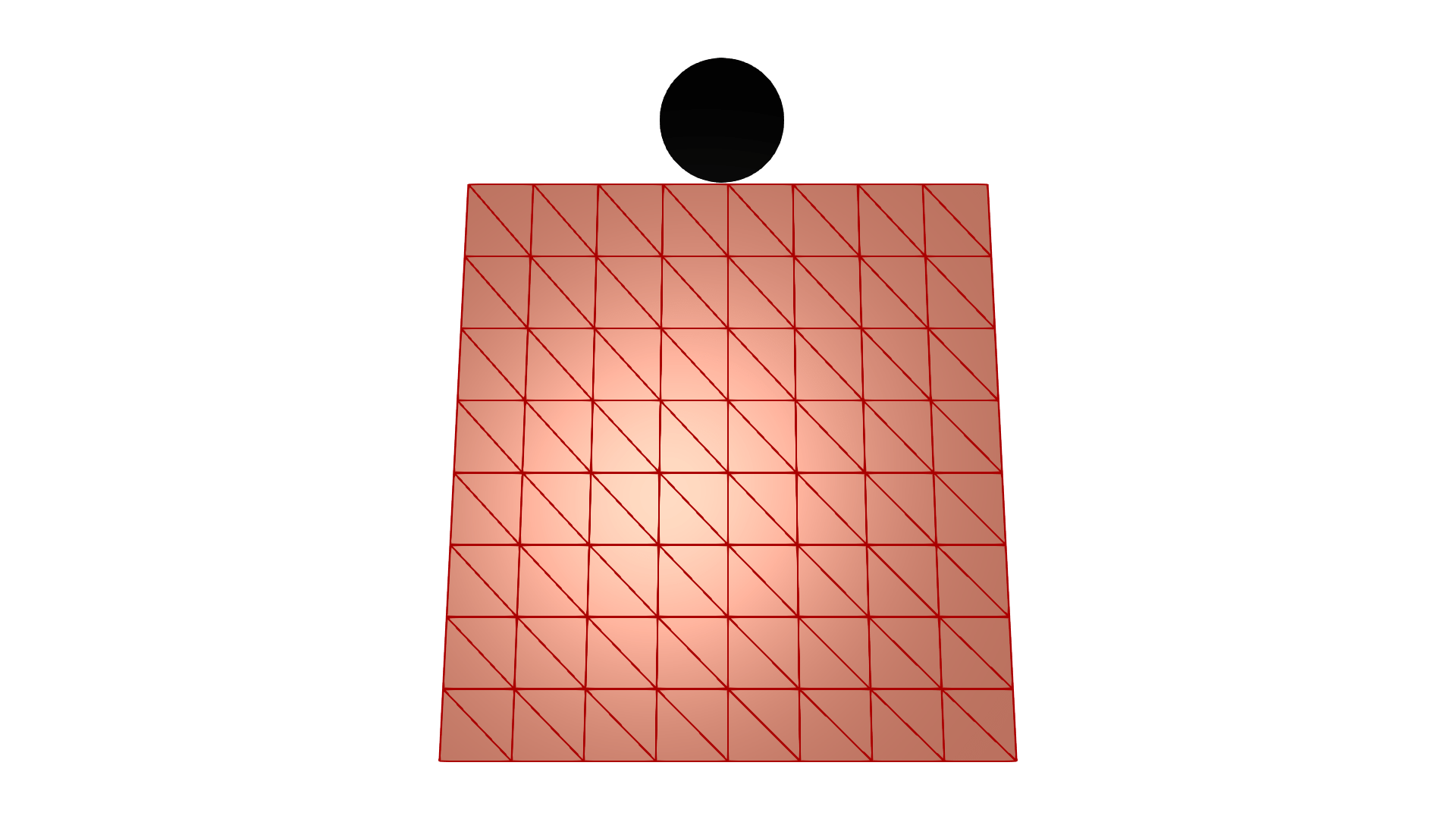}\hspace{-1pt}%
    \img{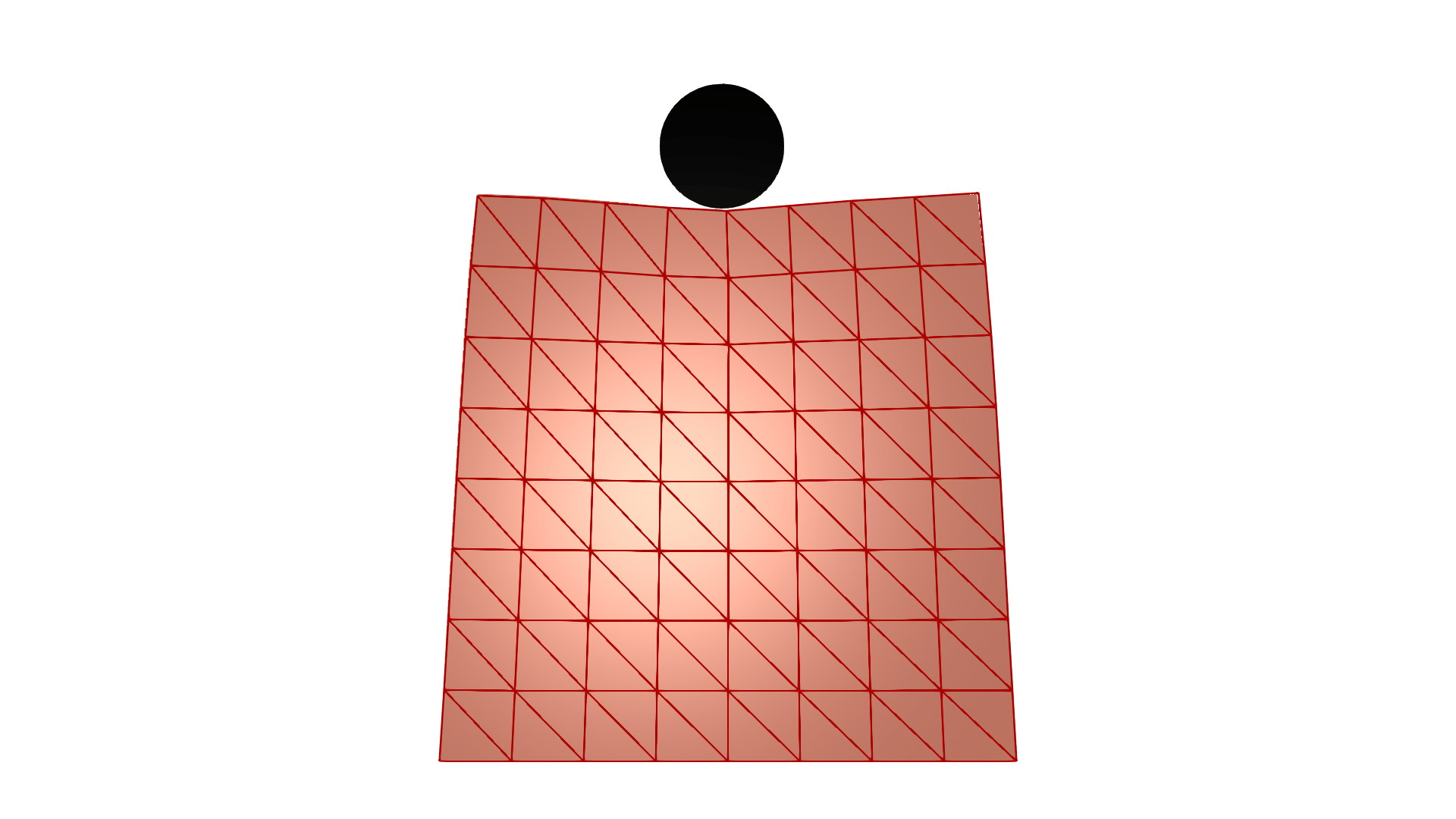}\hspace{-1pt}%
    \img{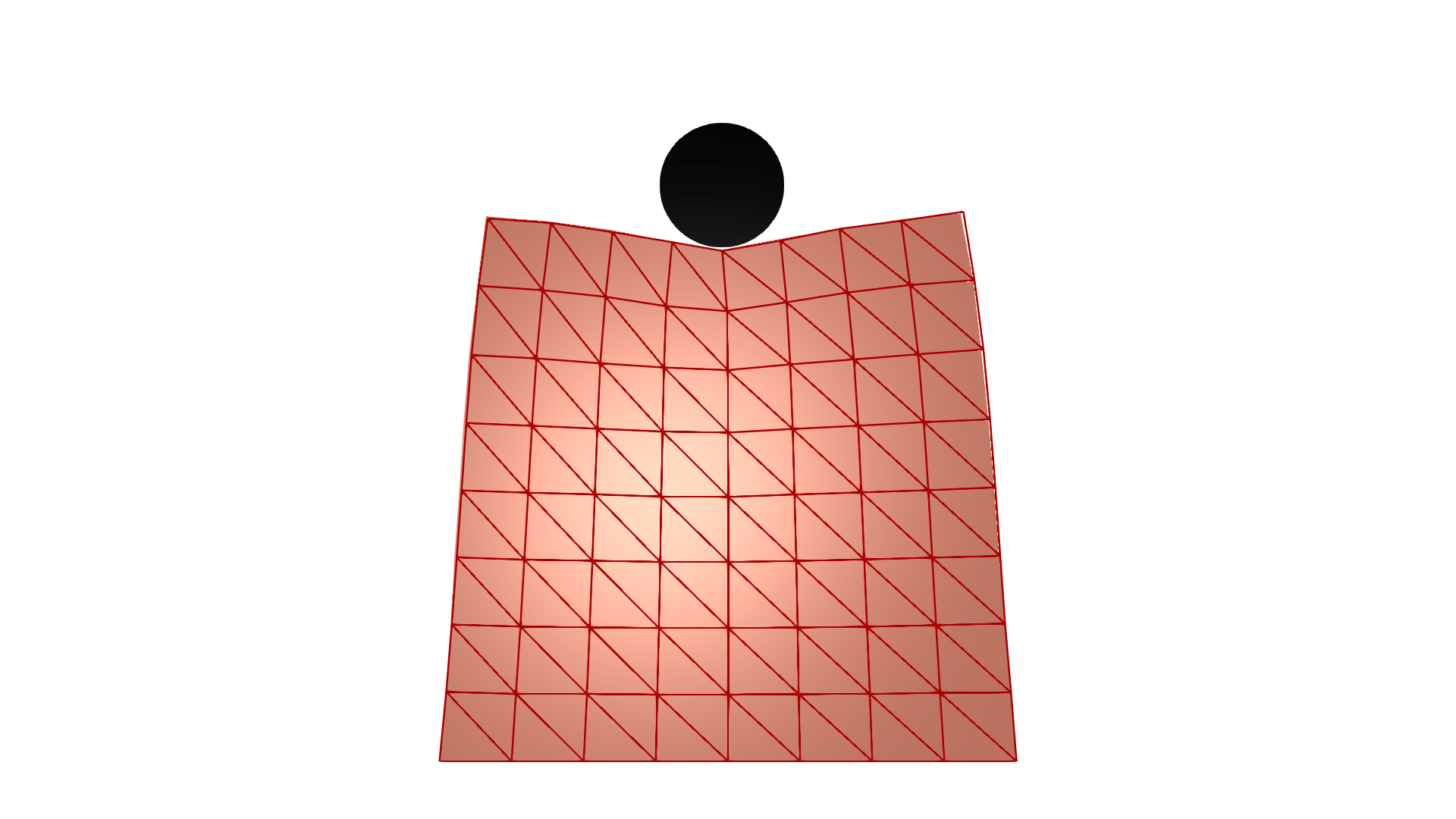}\hspace{-1pt}%
    \img{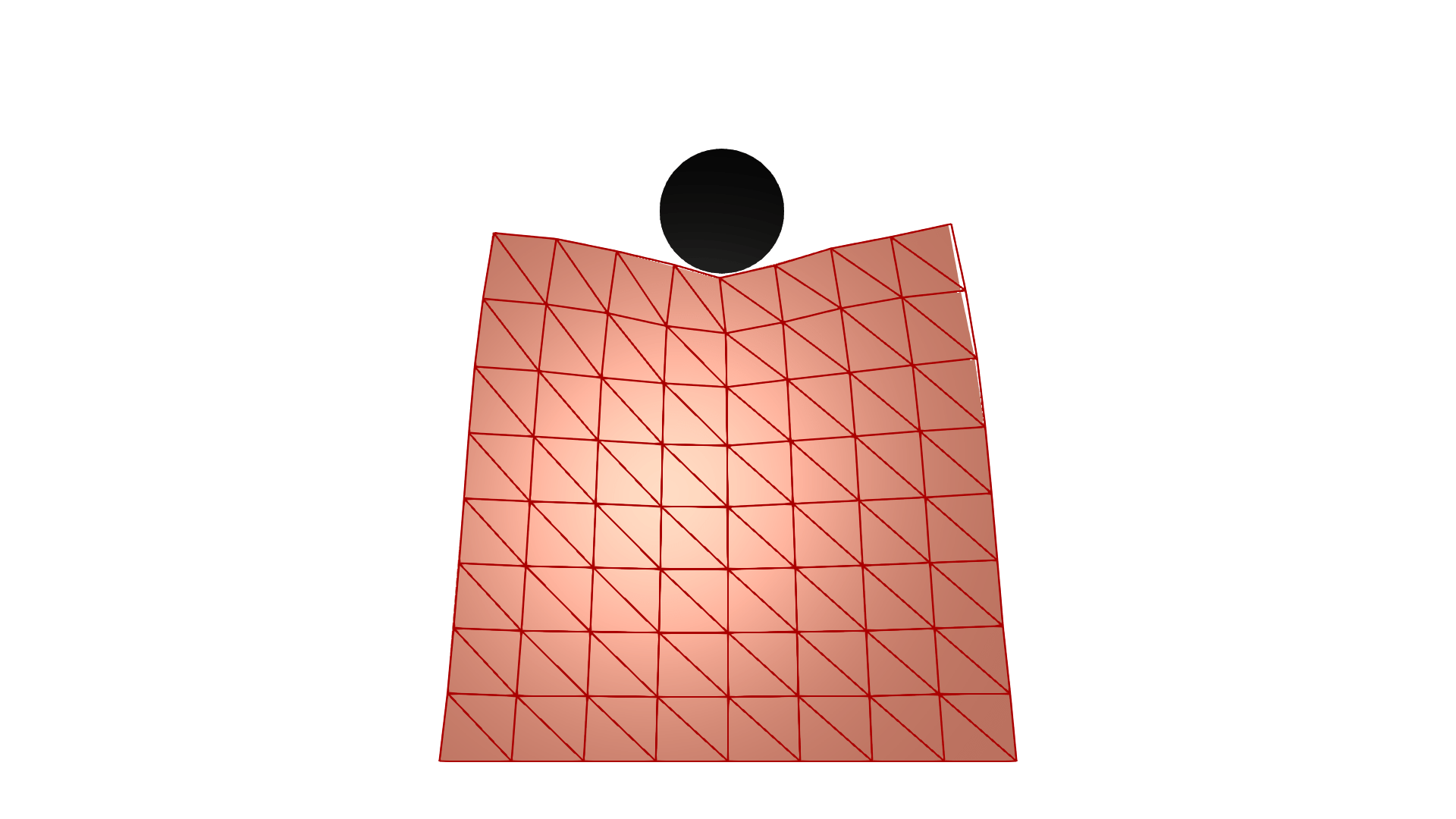}\hspace{-1pt}%
    \img{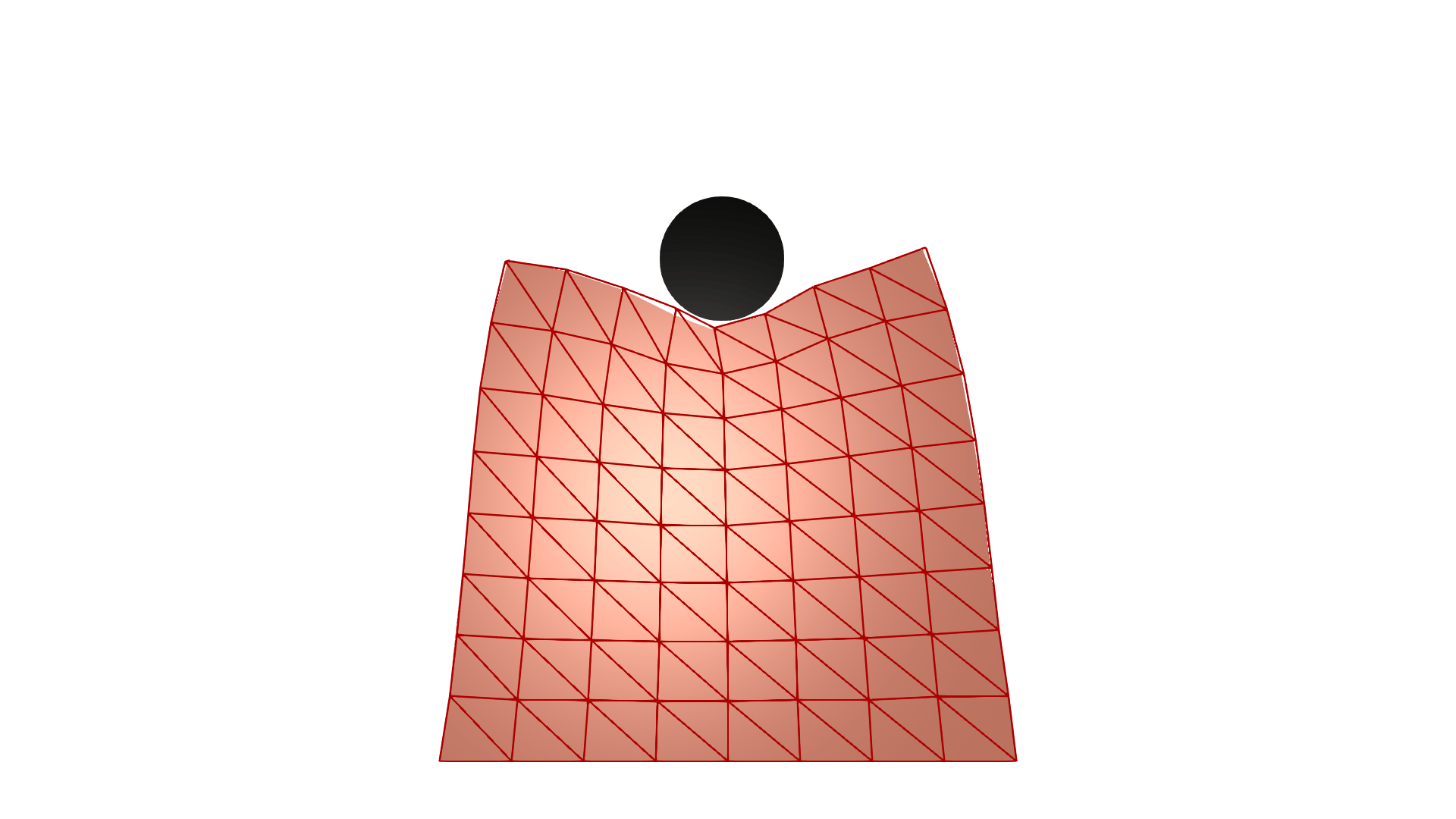}\hspace{-1pt}%
    \img{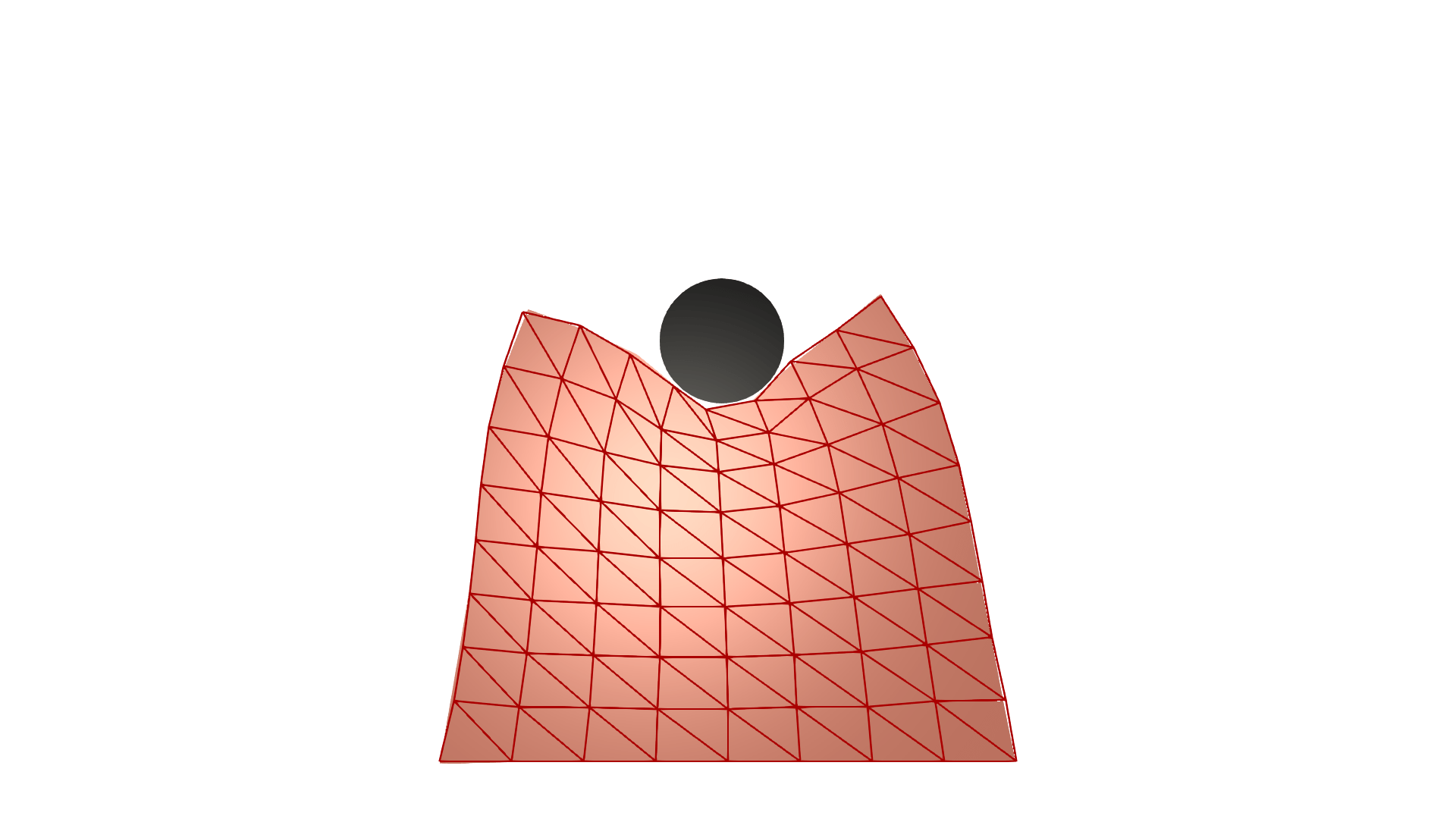}\\[0.4em]

    % Row 8: PSTNet Encoder
    \rowlabel{PSTNet \\ Encoder}%
    \img{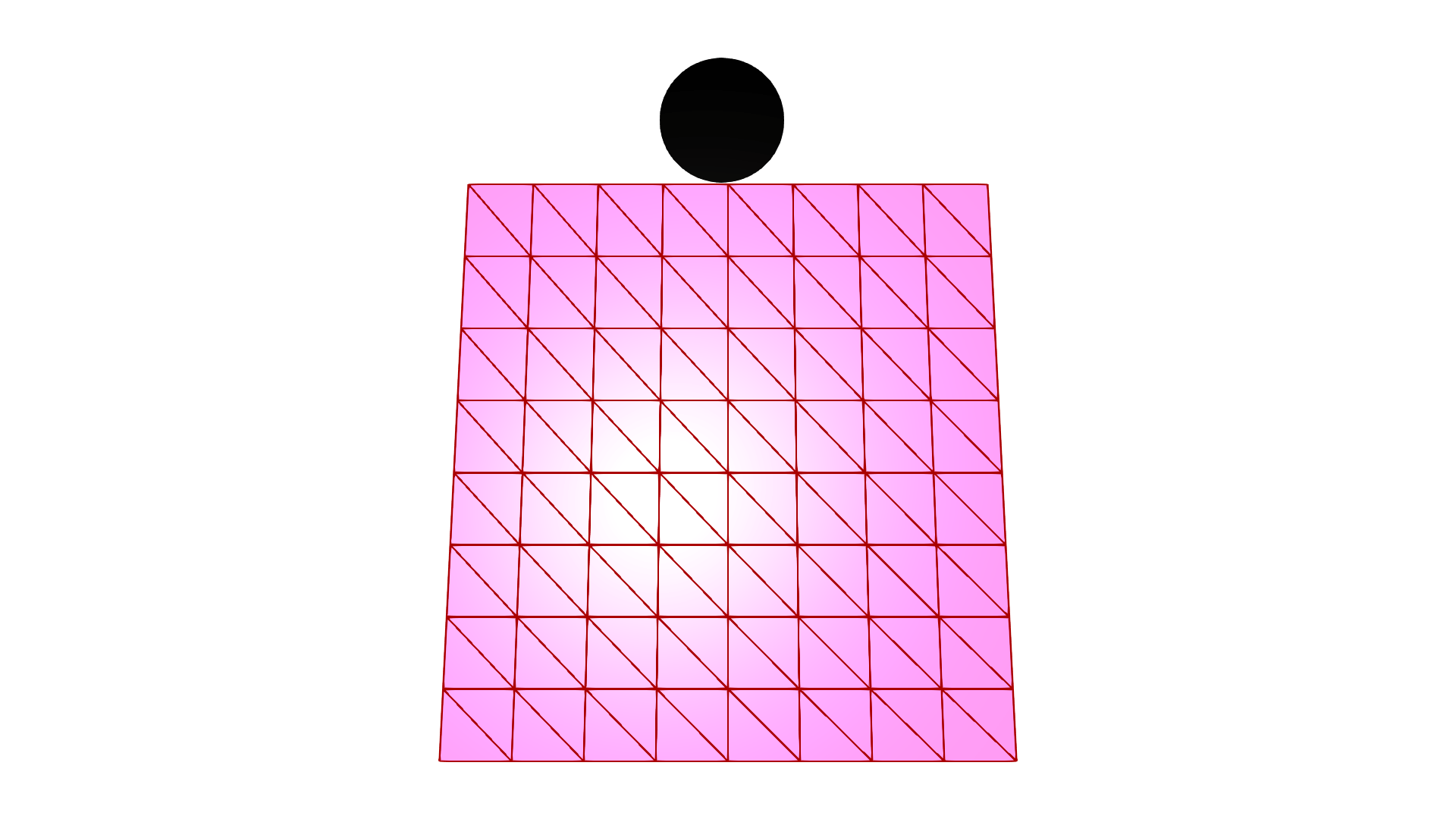}\hspace{-1pt}%
    \img{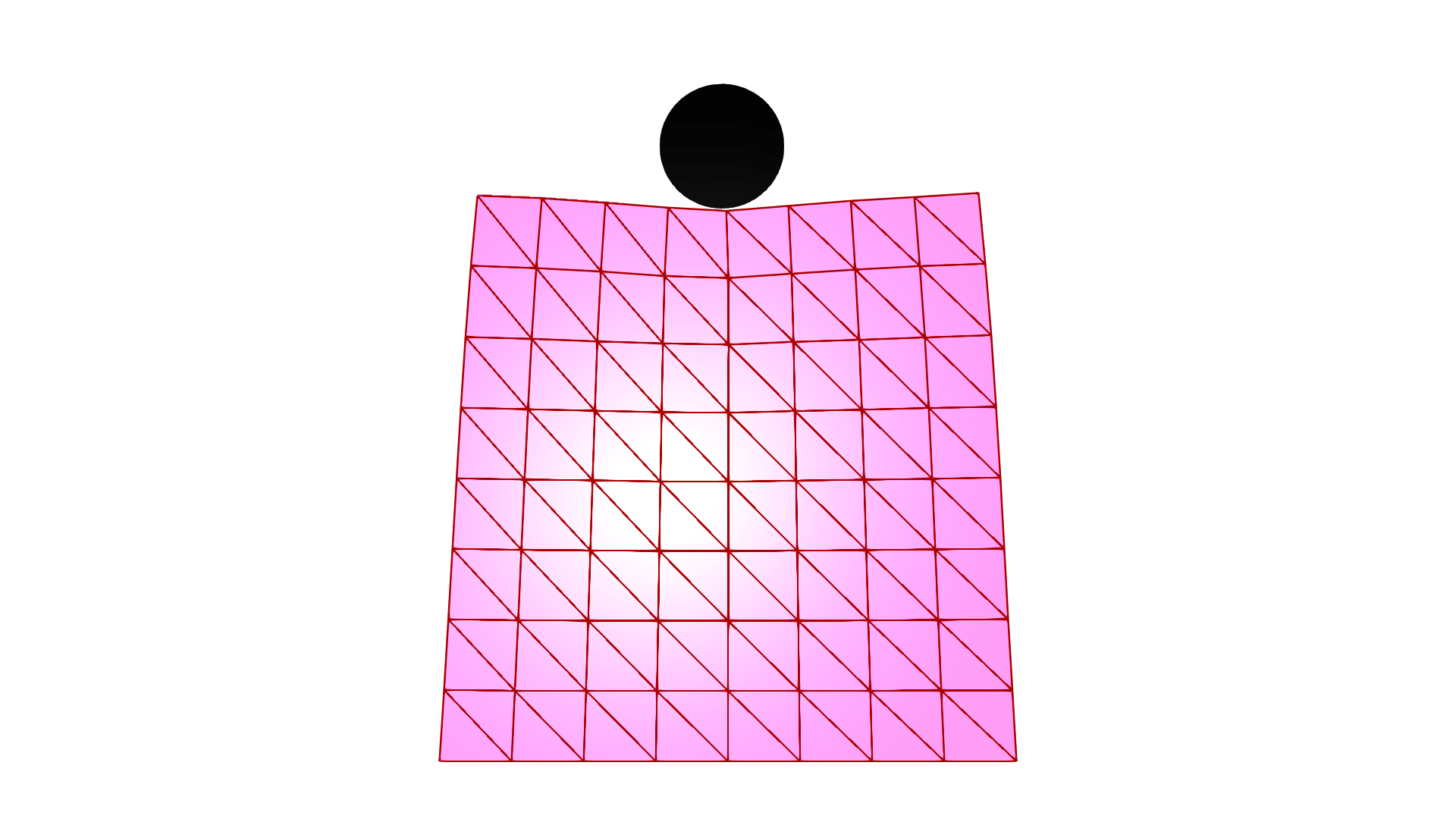}\hspace{-1pt}%
    \img{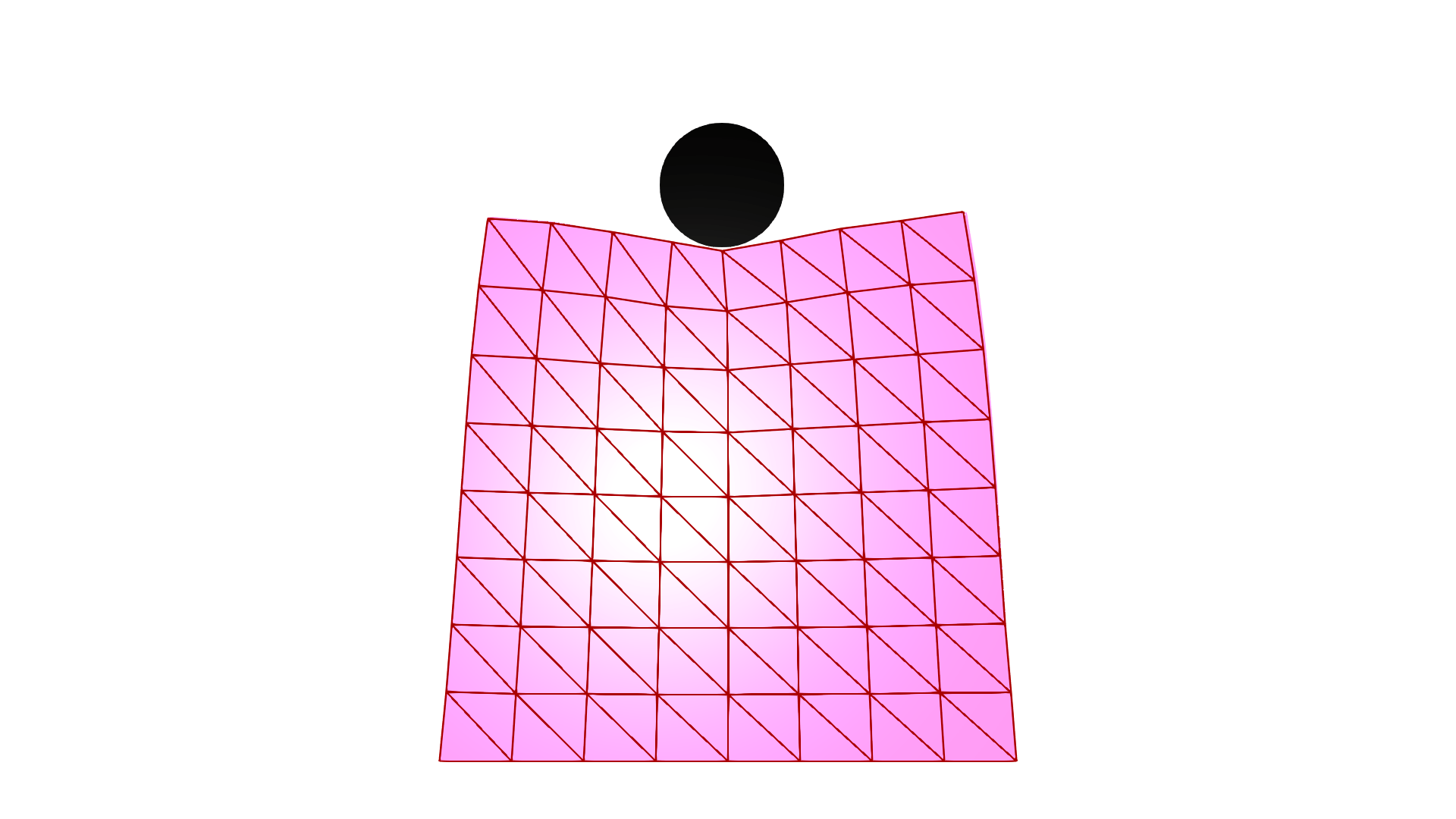}\hspace{-1pt}%
    \img{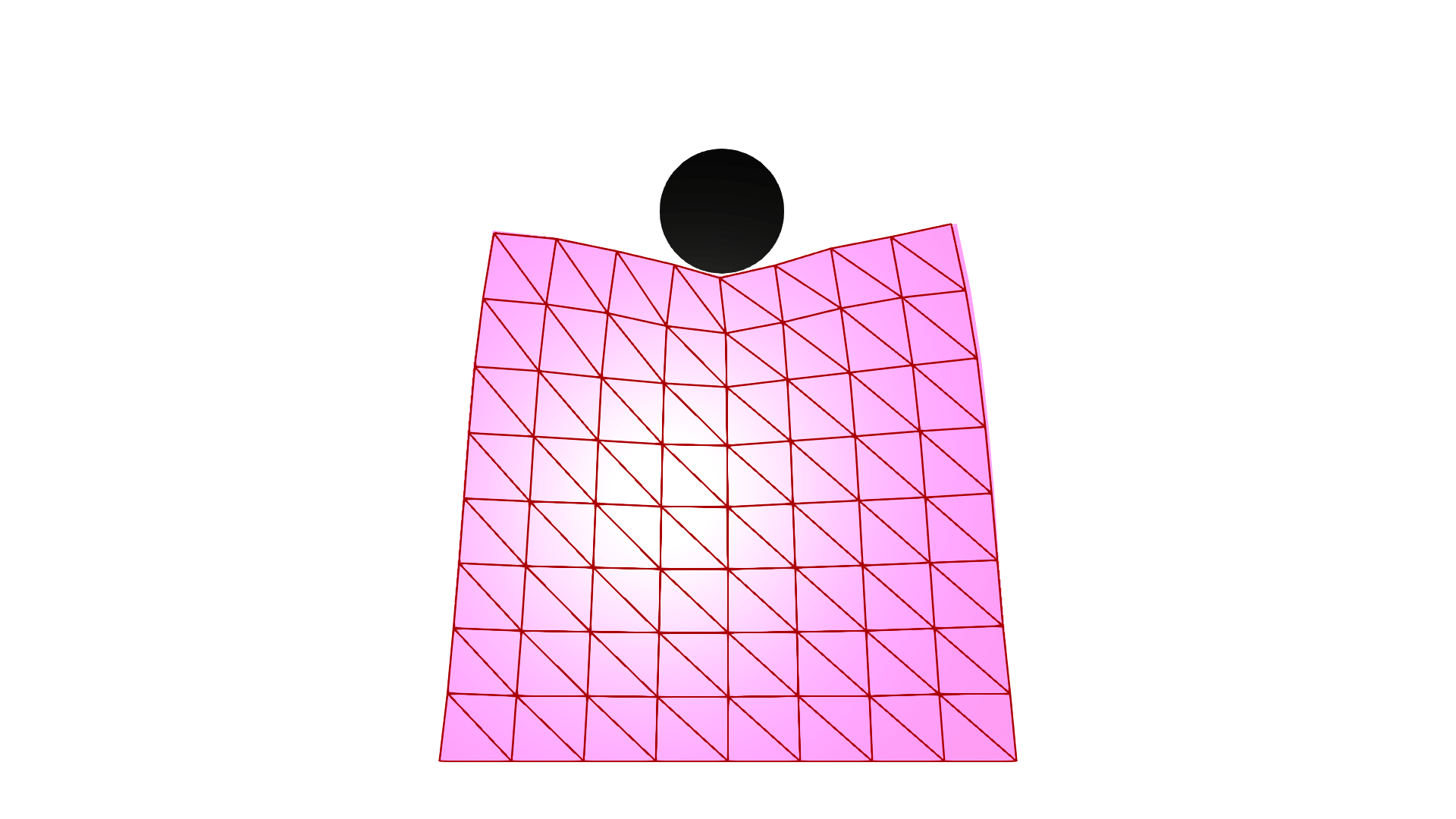}\hspace{-1pt}%
    \img{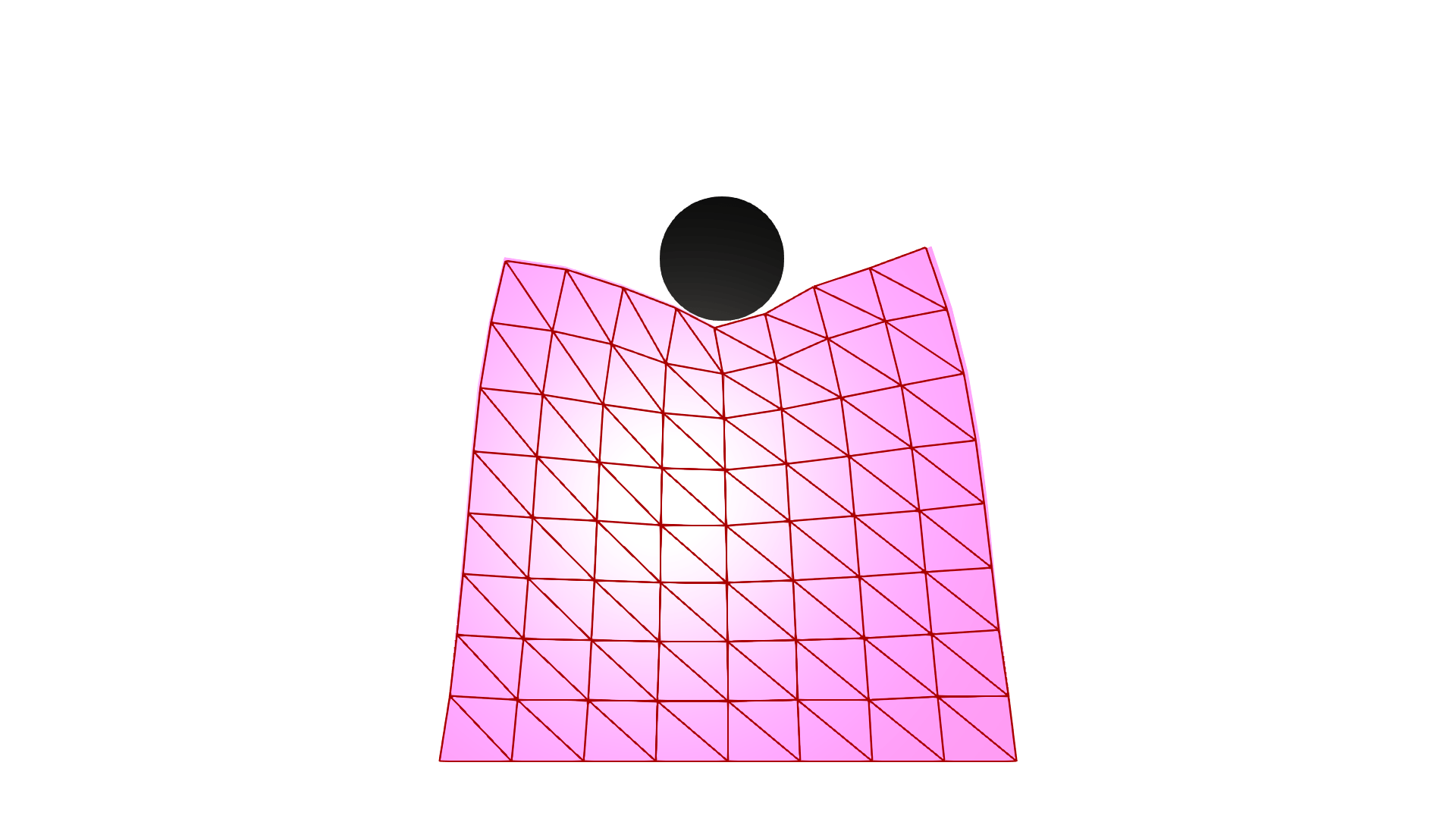}\hspace{-1pt}%
    \img{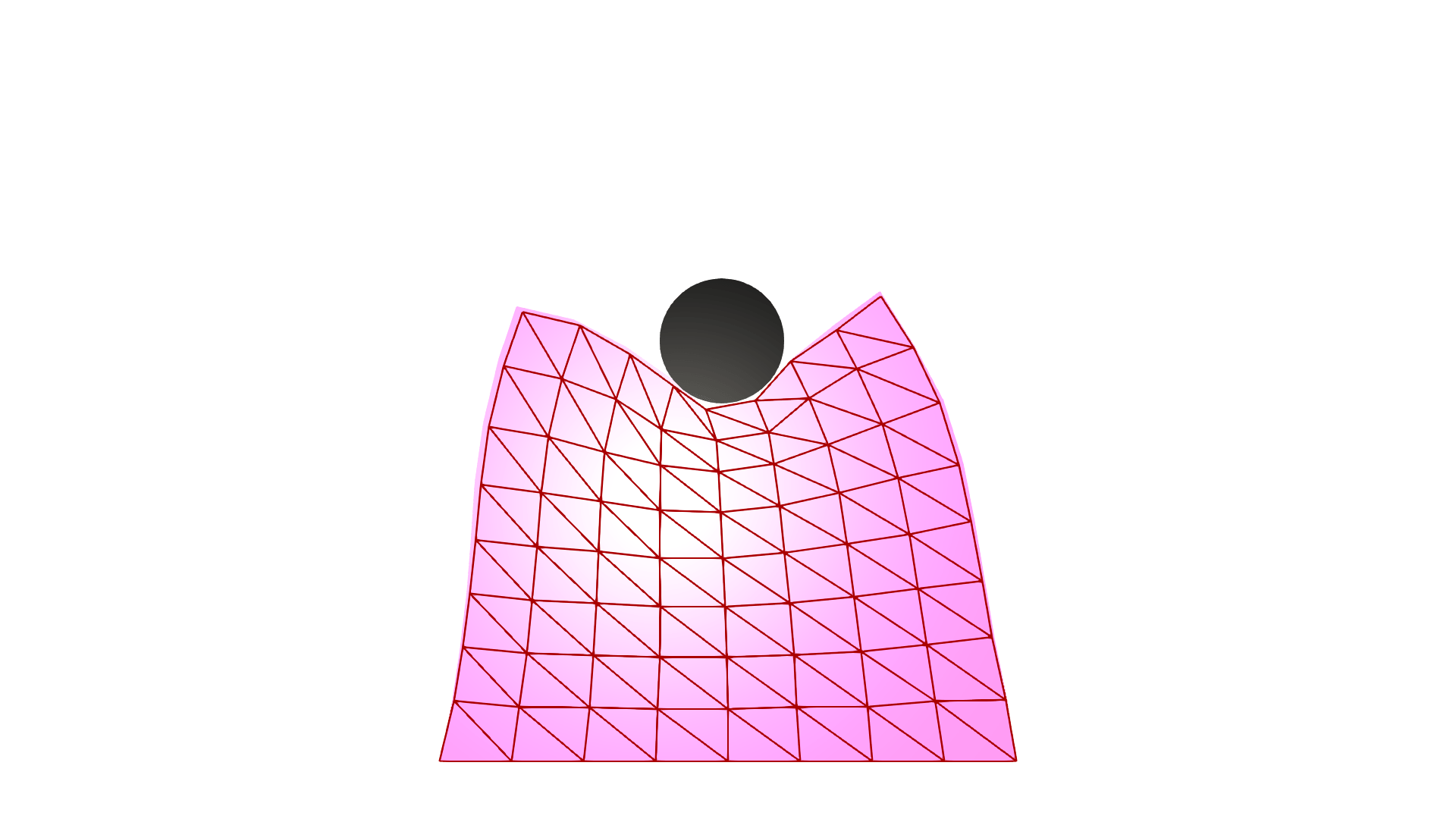}\\[0.4em]

    % Row 9: Pointcloud (special crop + labels)
    \rowlabel{Pointcloud}%
    \begin{minipage}[c]{0.158\textwidth}\centering
        \pointcloud{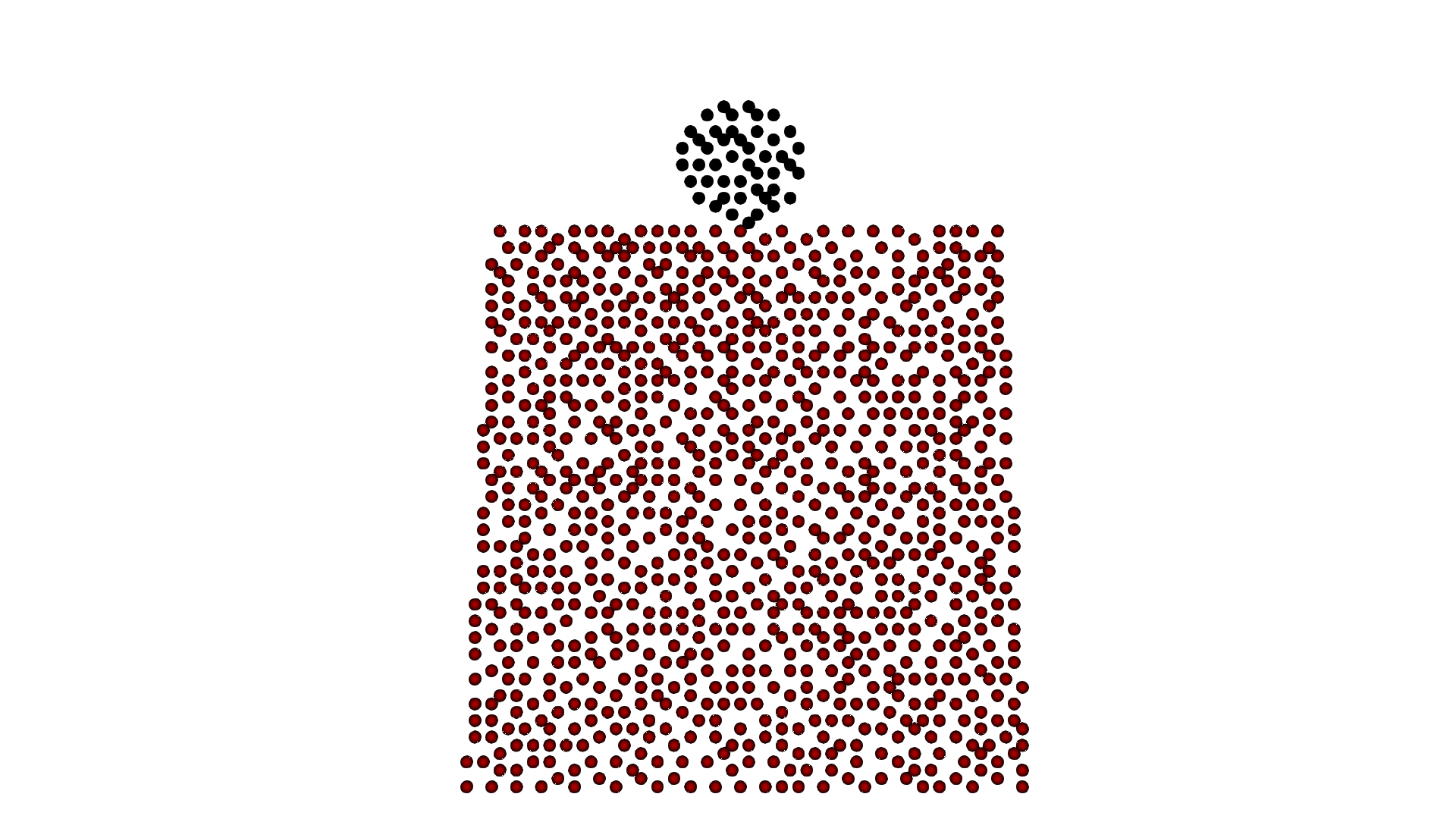}\\[2pt]
        \footnotesize$t{=}0$
    \end{minipage}%
    \begin{minipage}[c]{0.158\textwidth}\centering
        \pointcloud{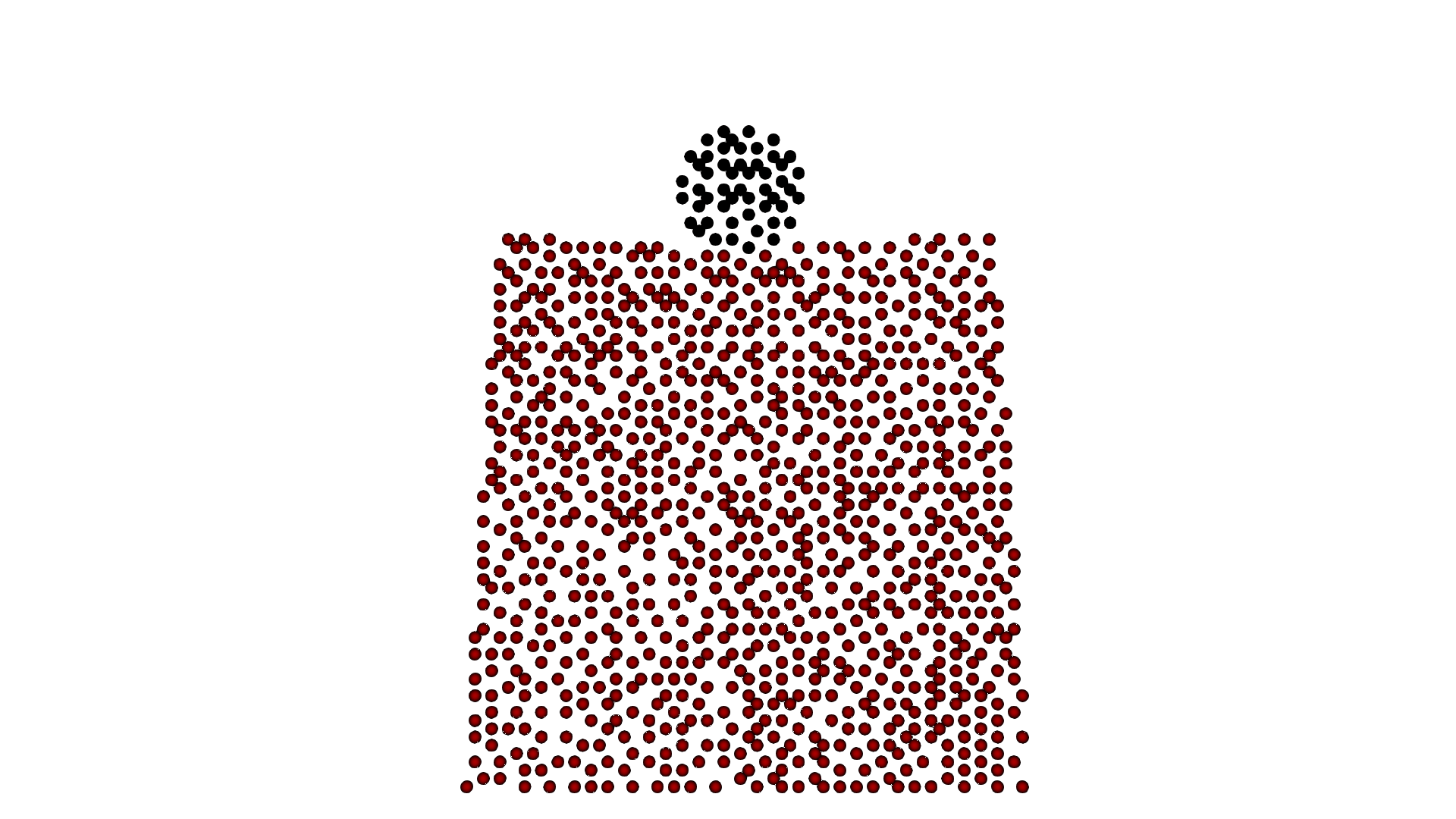}\\[2pt]
        \footnotesize$t{=}6$
    \end{minipage}%
    \begin{minipage}[c]{0.158\textwidth}\centering
        \pointcloud{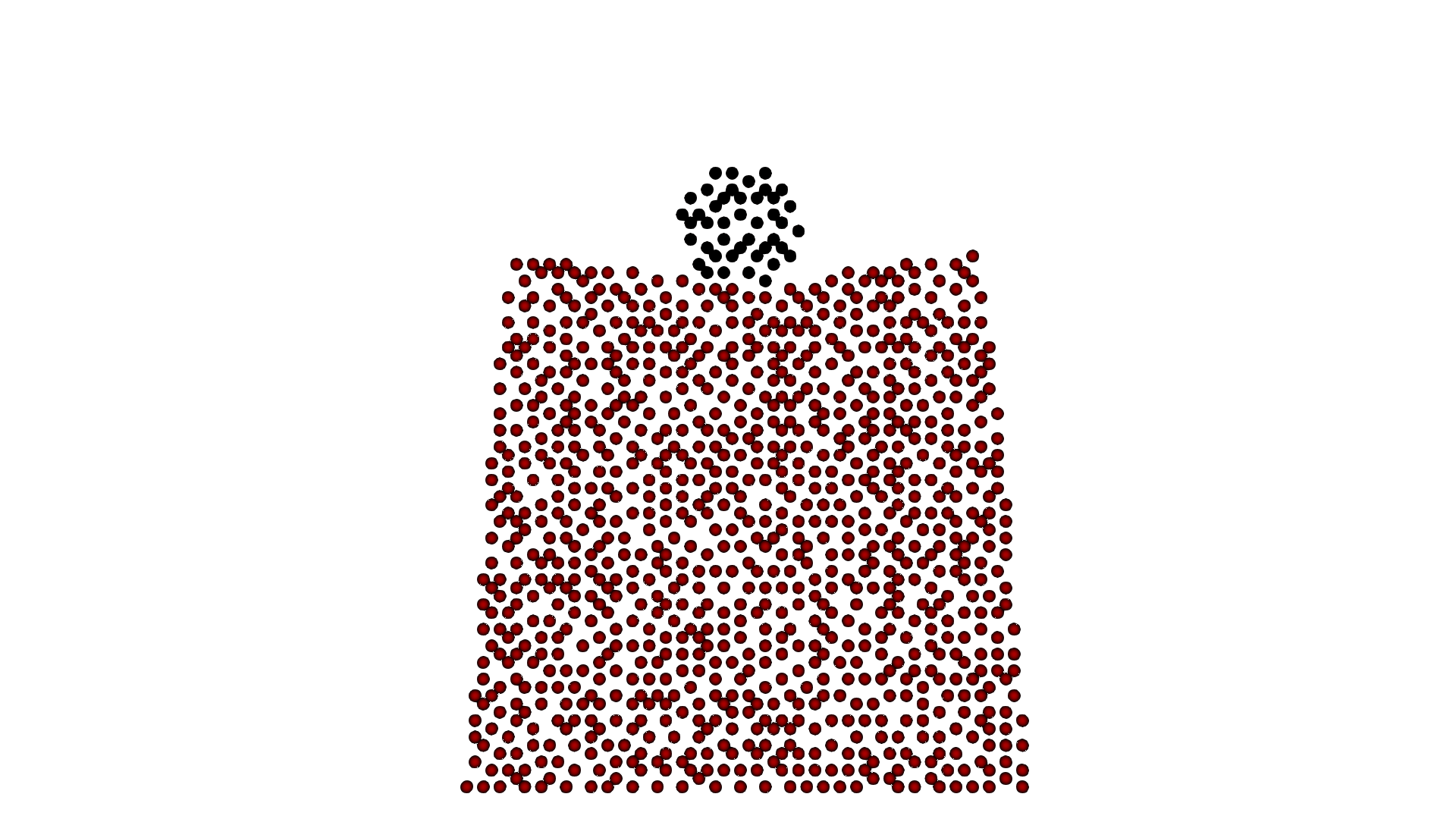}\\[2pt]
        \footnotesize$t{=}15$
    \end{minipage}%
    \begin{minipage}[c]{0.158\textwidth}\centering
        \pointcloud{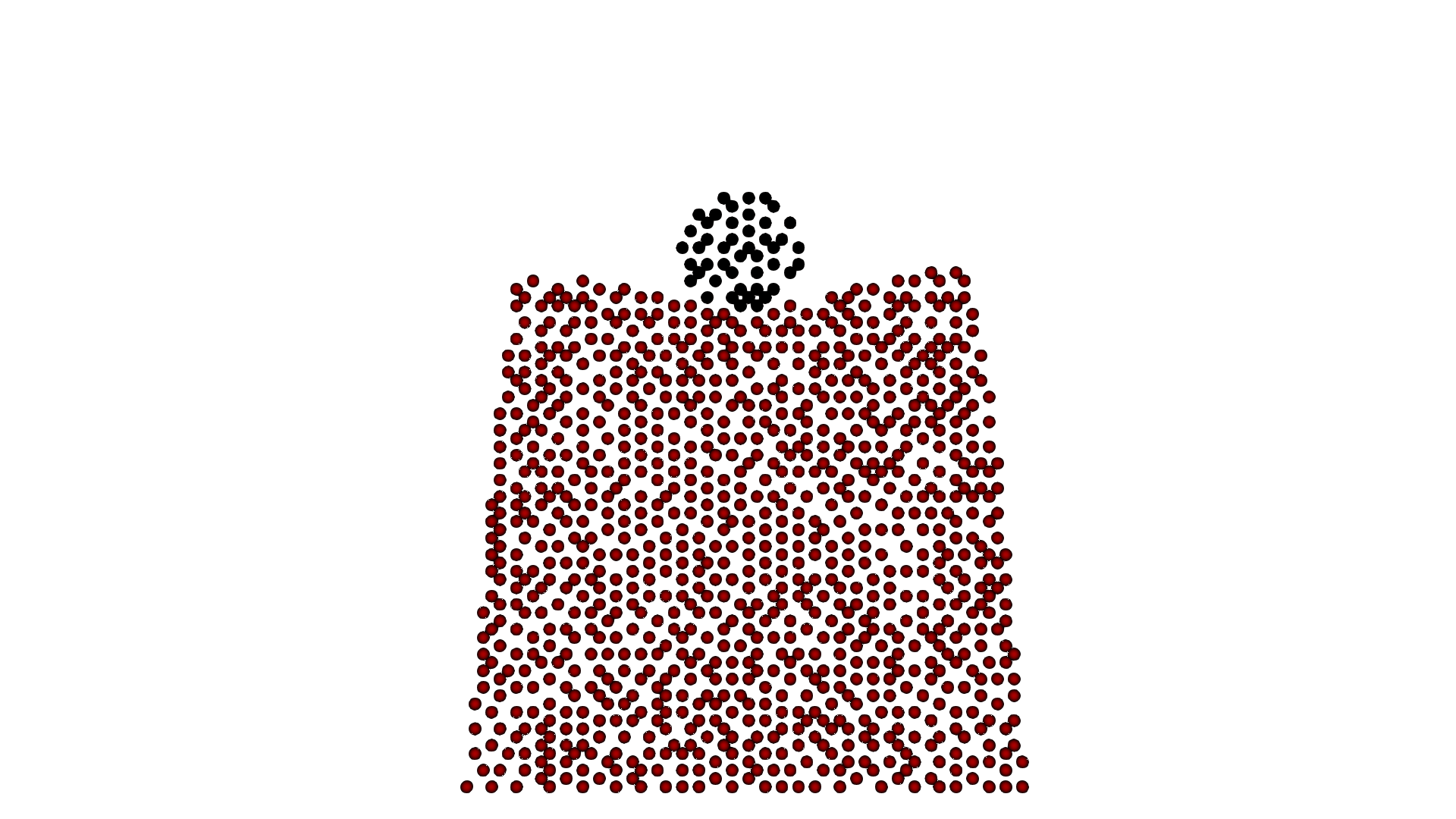}\\[2pt]
        \footnotesize$t{=}21$
    \end{minipage}%
    \begin{minipage}[c]{0.158\textwidth}\centering
        \pointcloud{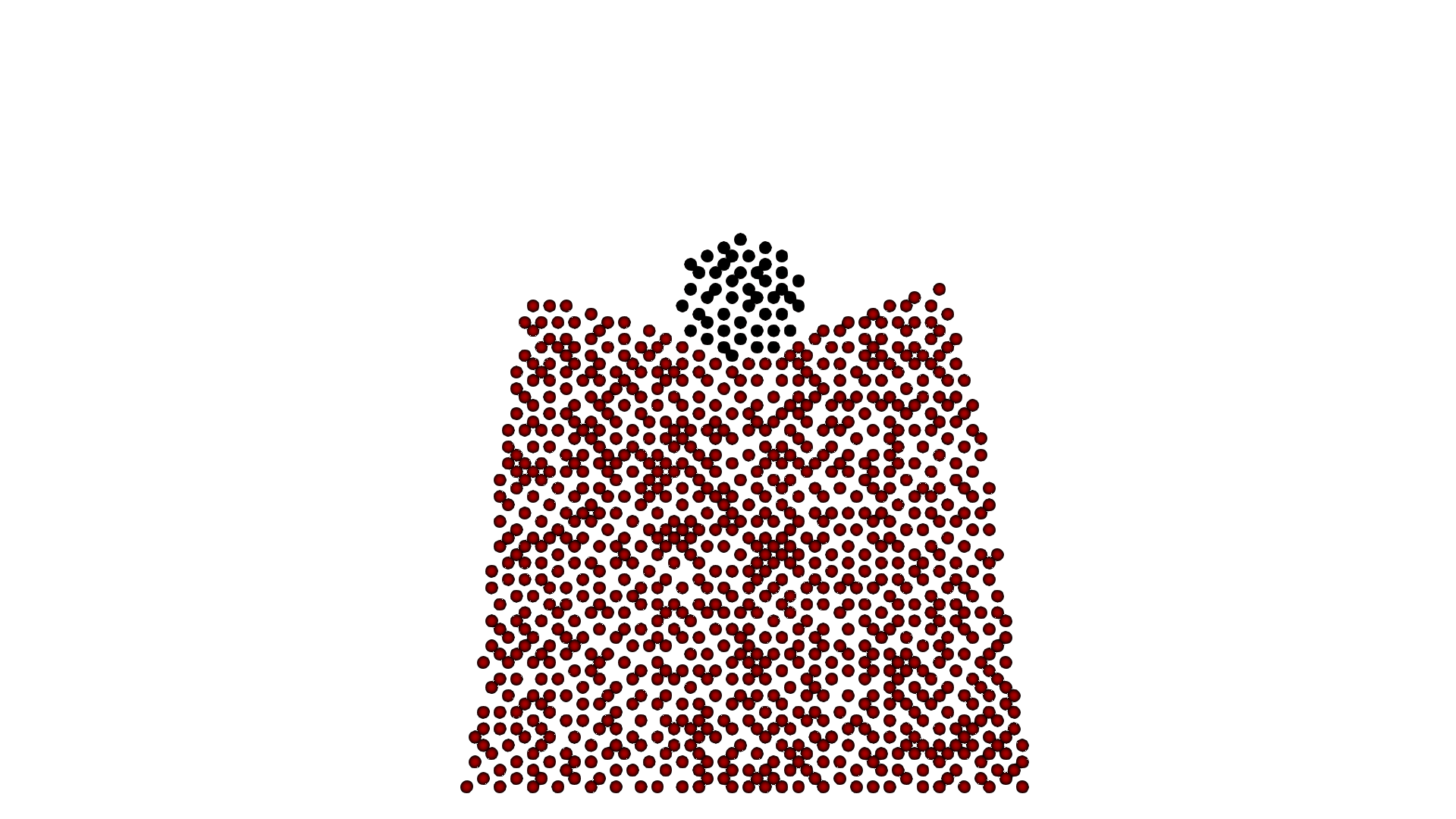}\\[2pt]
        \footnotesize$t{=}32$
    \end{minipage}%
    \begin{minipage}[c]{0.158\textwidth}\centering
        \pointcloud{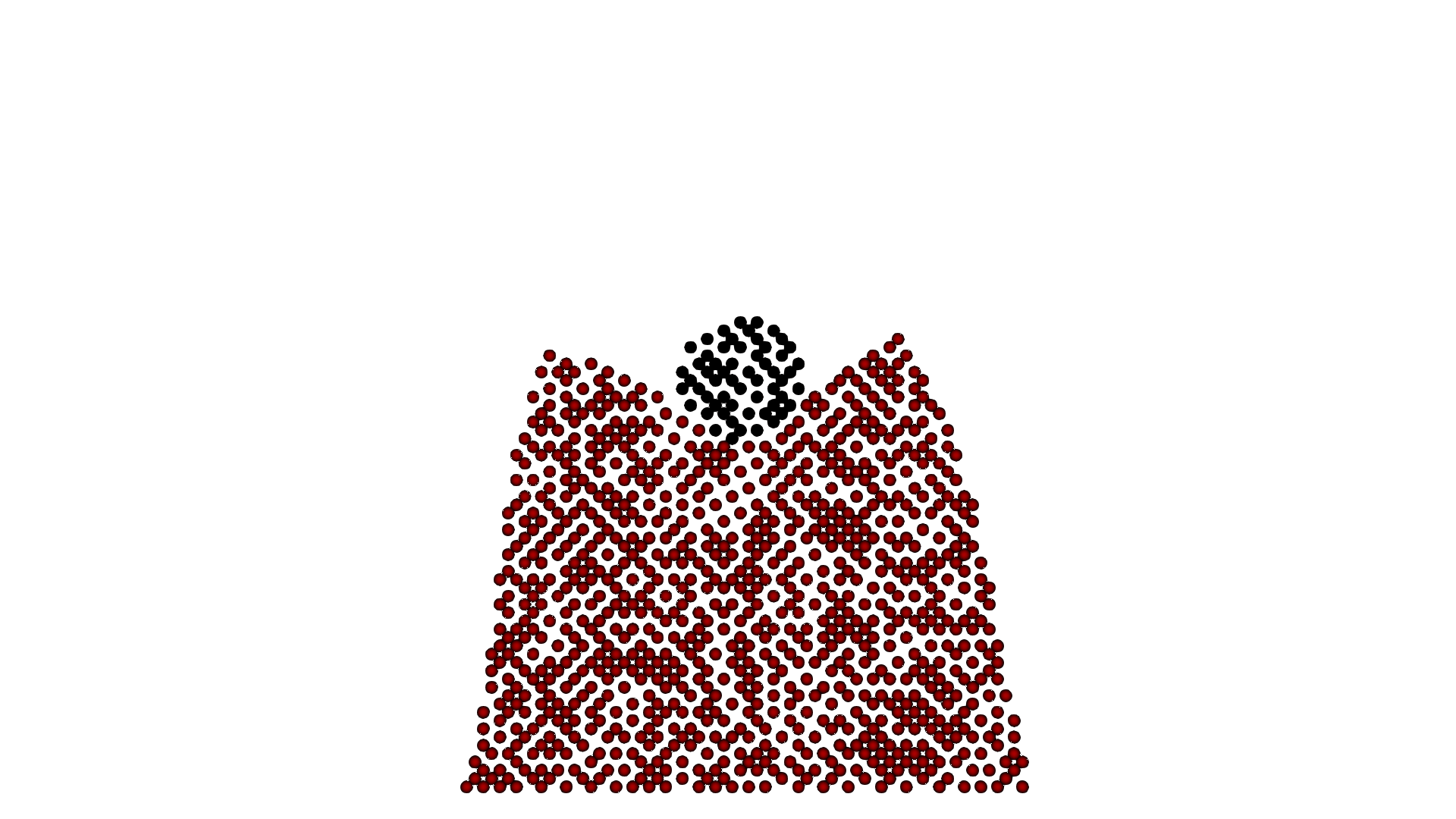}\\[2pt]
        \footnotesize$t{=}51$
    \end{minipage}

    \par\vfill
    \caption{
Predicted simulation of a \texttt{Deforming Block} test task by
\textcolor{TabBlue}{PEACH} (blue),
\textcolor{TabGray}{No Context} (gray),
\textcolor{TabCyan}{MANGO} (cyan),
\textcolor{TabGreen}{Oracle} (green),
\textcolor{TabOrange}{No Context (MGN)} (orange),
\textcolor{TabPurple}{Oracle (MGN)} (purple),
\textcolor{TabBrown}{GNN Encoder} (brown), and
\textcolor{TabPink}{PSTNet Encoder} (pink).
All visualizations show the colored \textbf{predicted mesh}, a \textbf{\textcolor{darkgray}{collider}}, and a \textbf{\textcolor{red}{wireframe}} (red) of the ground-truth simulation. The last row shows an exemplary point cloud sequence from the context set.
    }
    \label{fig:qualitative_trajectories_db}
\end{figure*}

\begin{figure*}[p]
    \centering
    {\Large \textbf{Sheet Deformation}}\\[1em]
    \vfill

    % helper macros (adjust trim once here)
    \newcommand{\rowlabel}[1]{%
        \begin{minipage}[c]{0.05\textwidth}\centering\rotatebox{90}{\scriptsize\textbf{\shortstack{#1}}}\end{minipage}%
    }
    \newcommand{\img}[1]{%
        \includegraphics[width=0.158\textwidth, trim=4cm 1cm 4cm 1cm, clip, valign=m]{#1}%
    }
    \newcommand{\pointcloud}[1]{%
        \includegraphics[width=\textwidth, trim=4cm 1cm 4cm 1cm, clip, valign=m]{#1}%
    }

    % Row 1: PEACH
    \rowlabel{PEACH}%
    \img{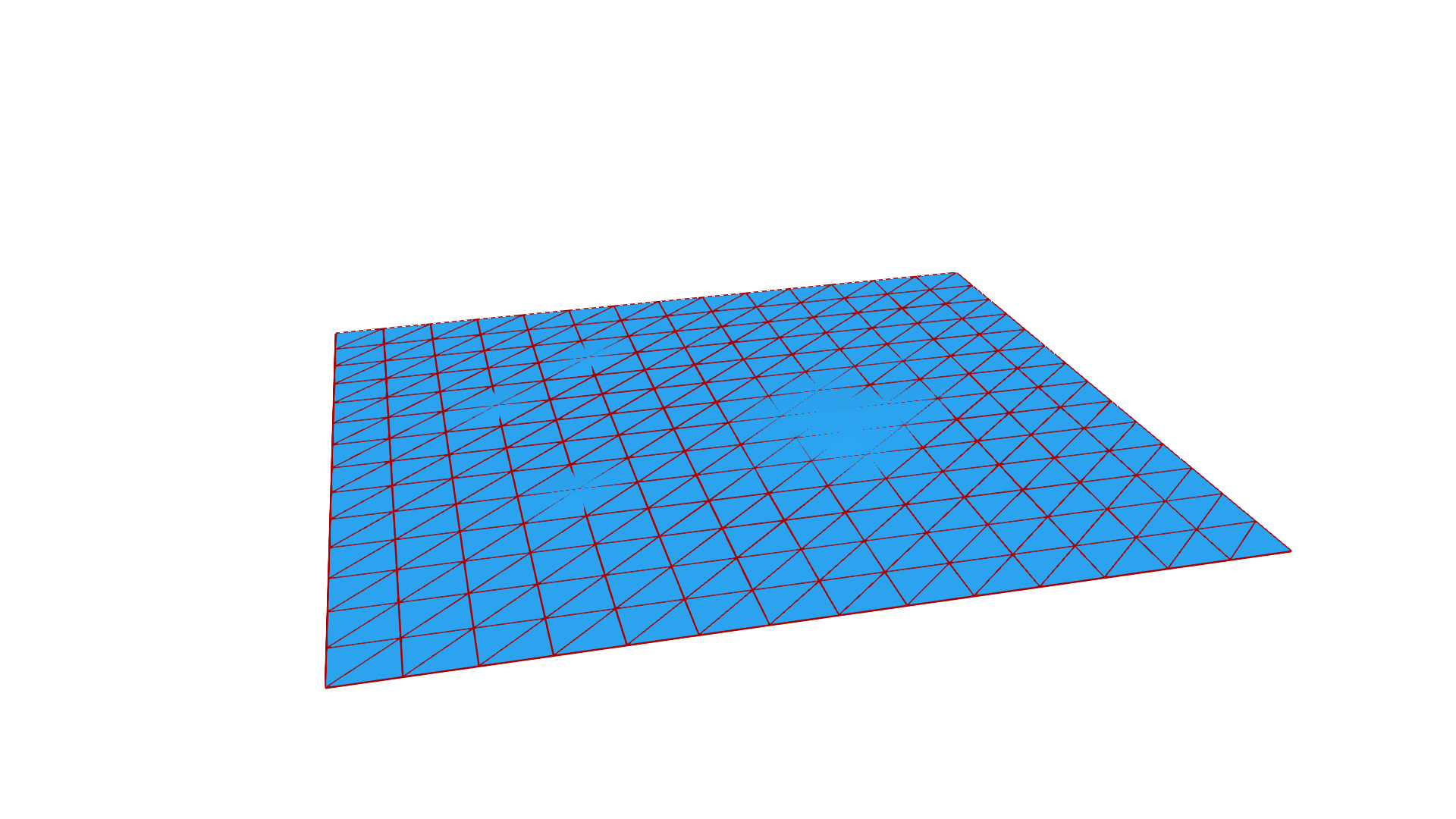}%
    \img{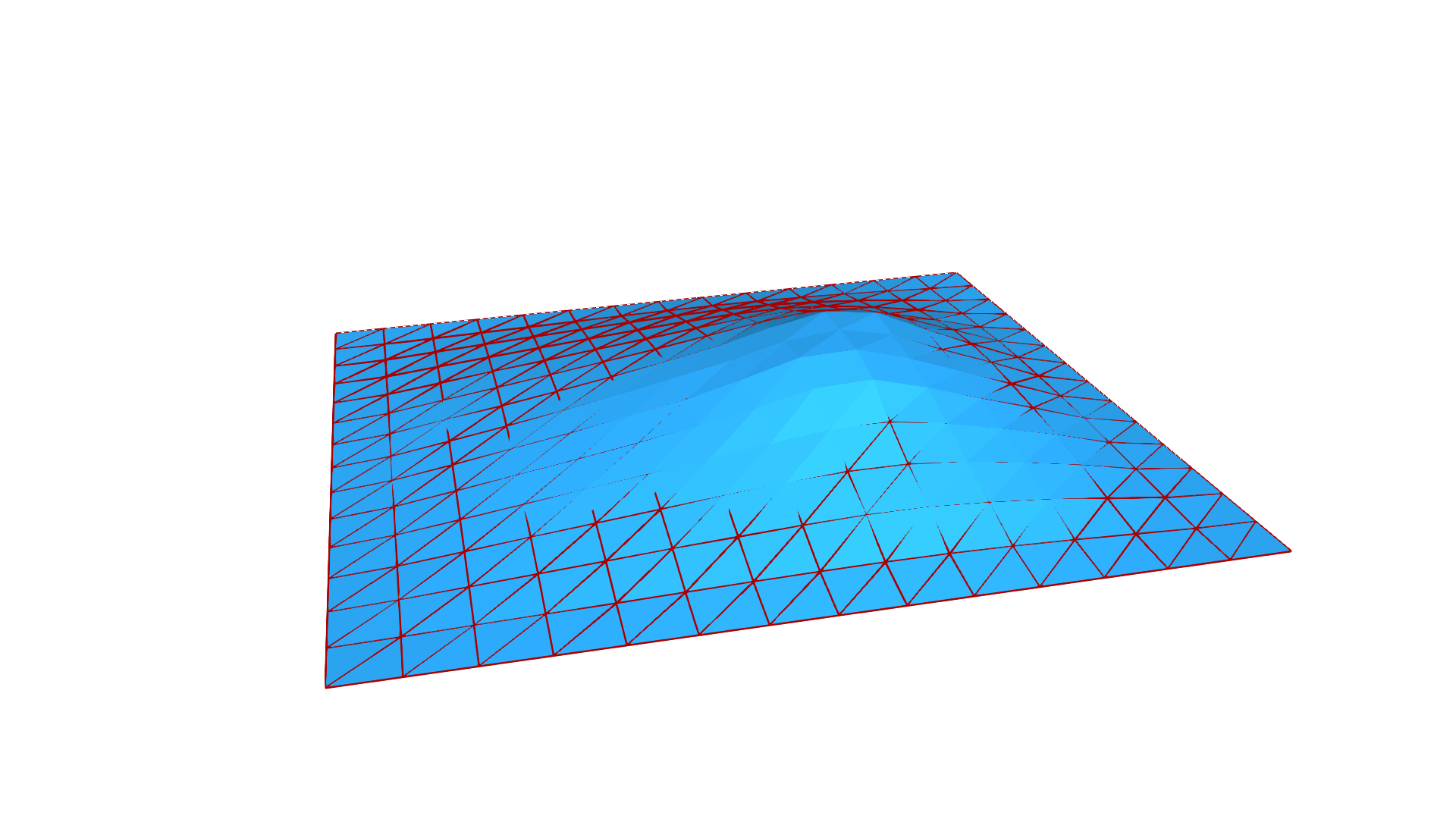}%
    \img{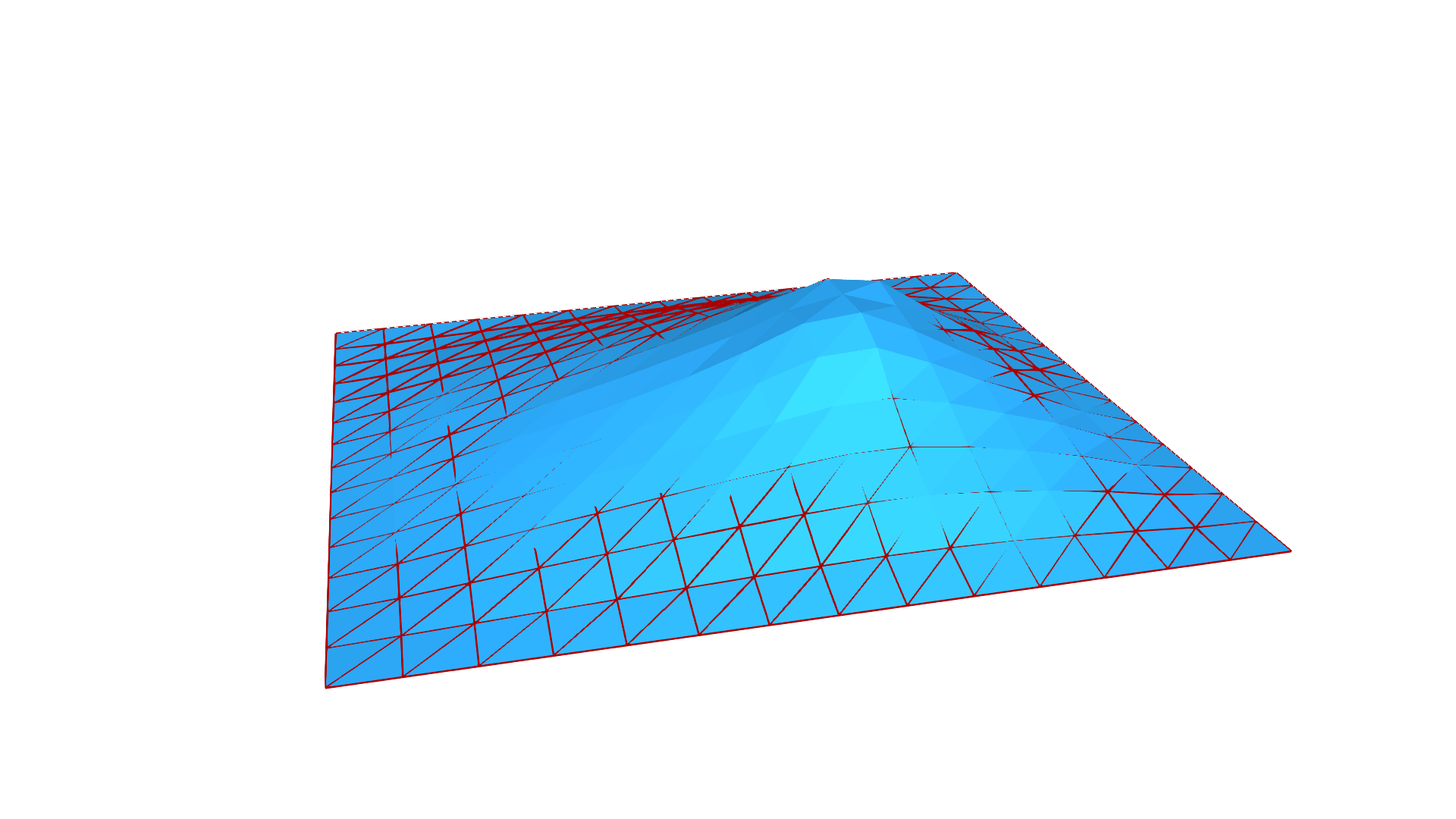}%
    \img{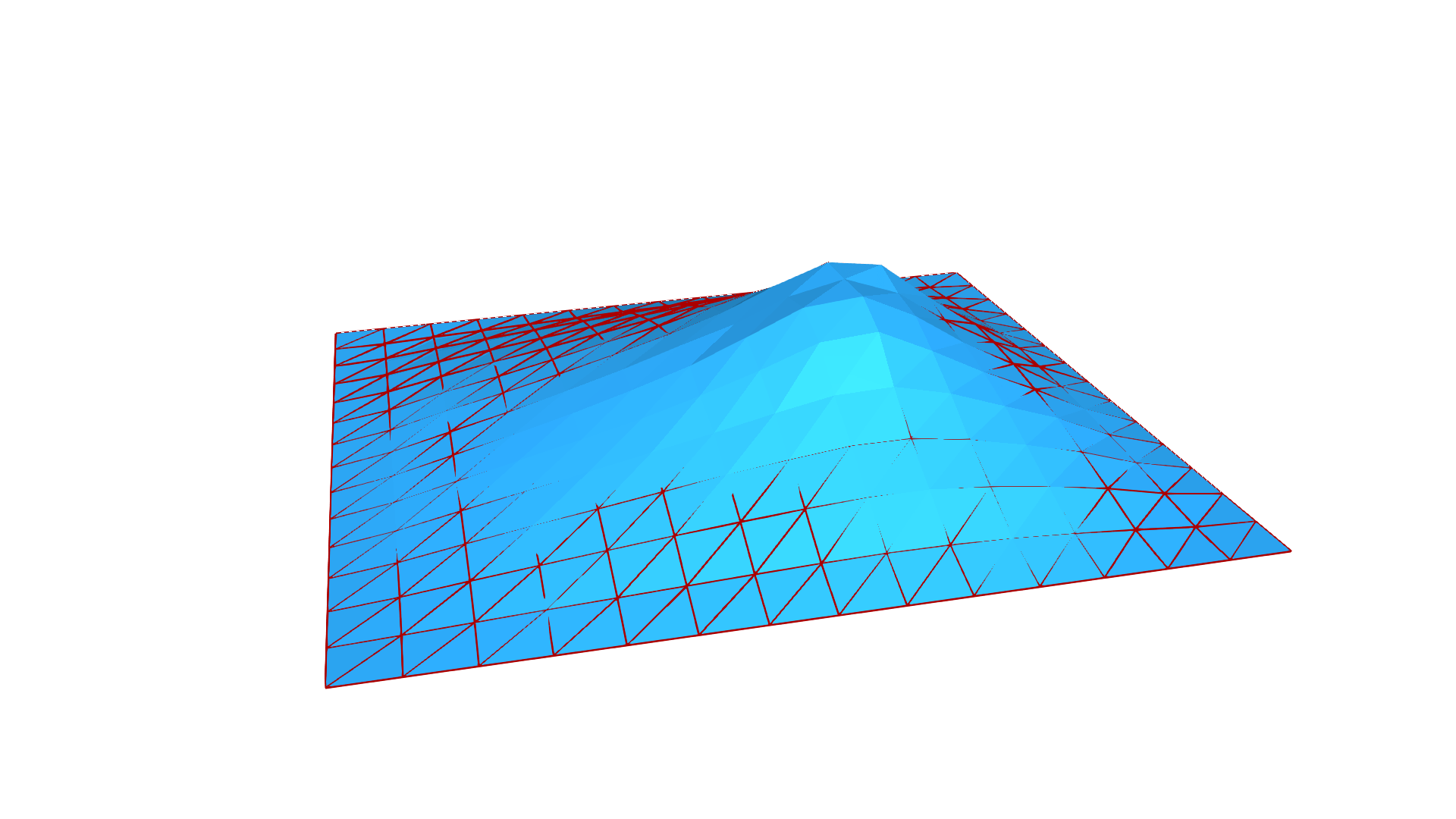}%
    \img{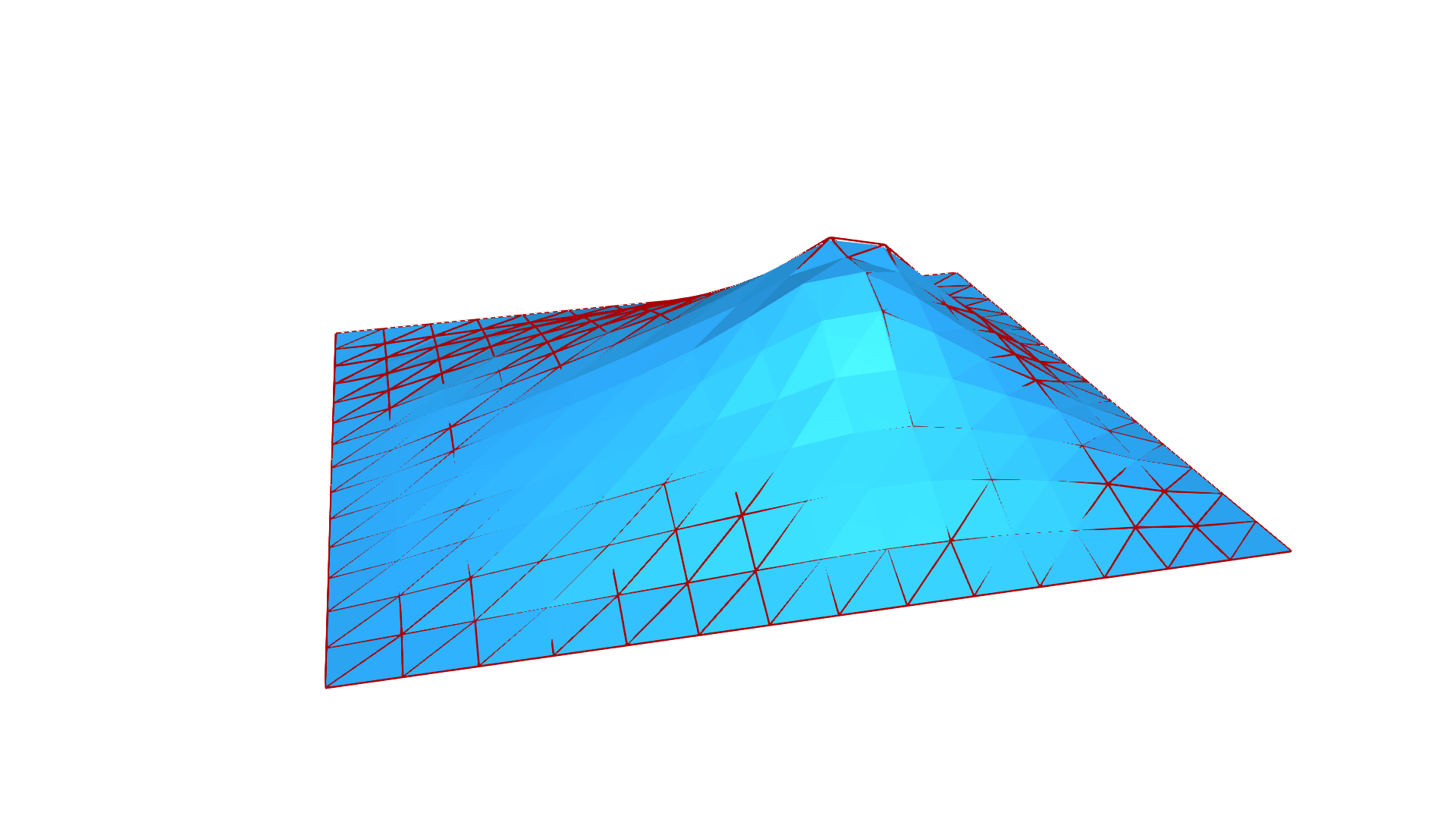}%
    \img{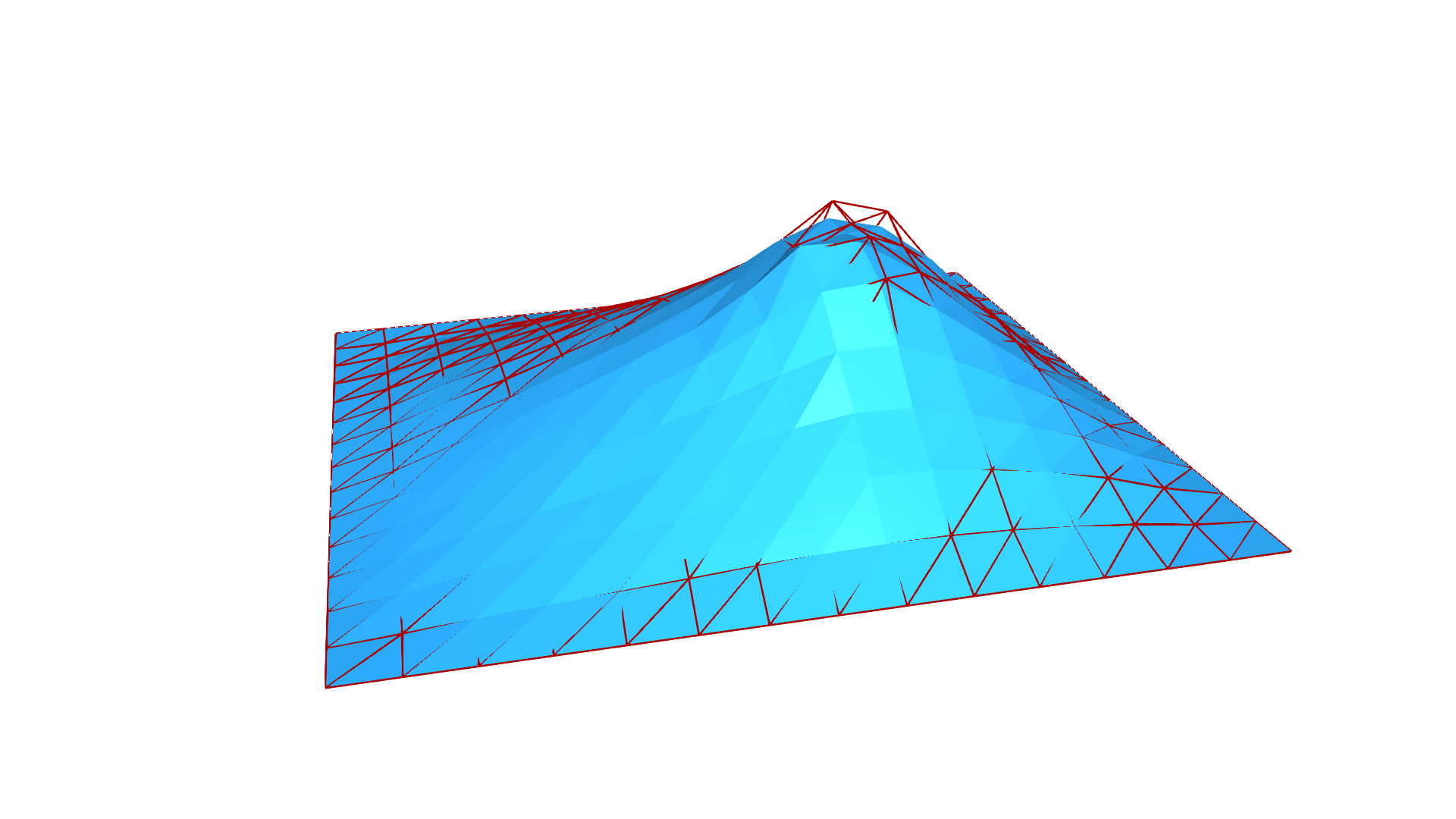}\\[0.4em]

    % Row 2: No Context
    \rowlabel{No Context}%
    \img{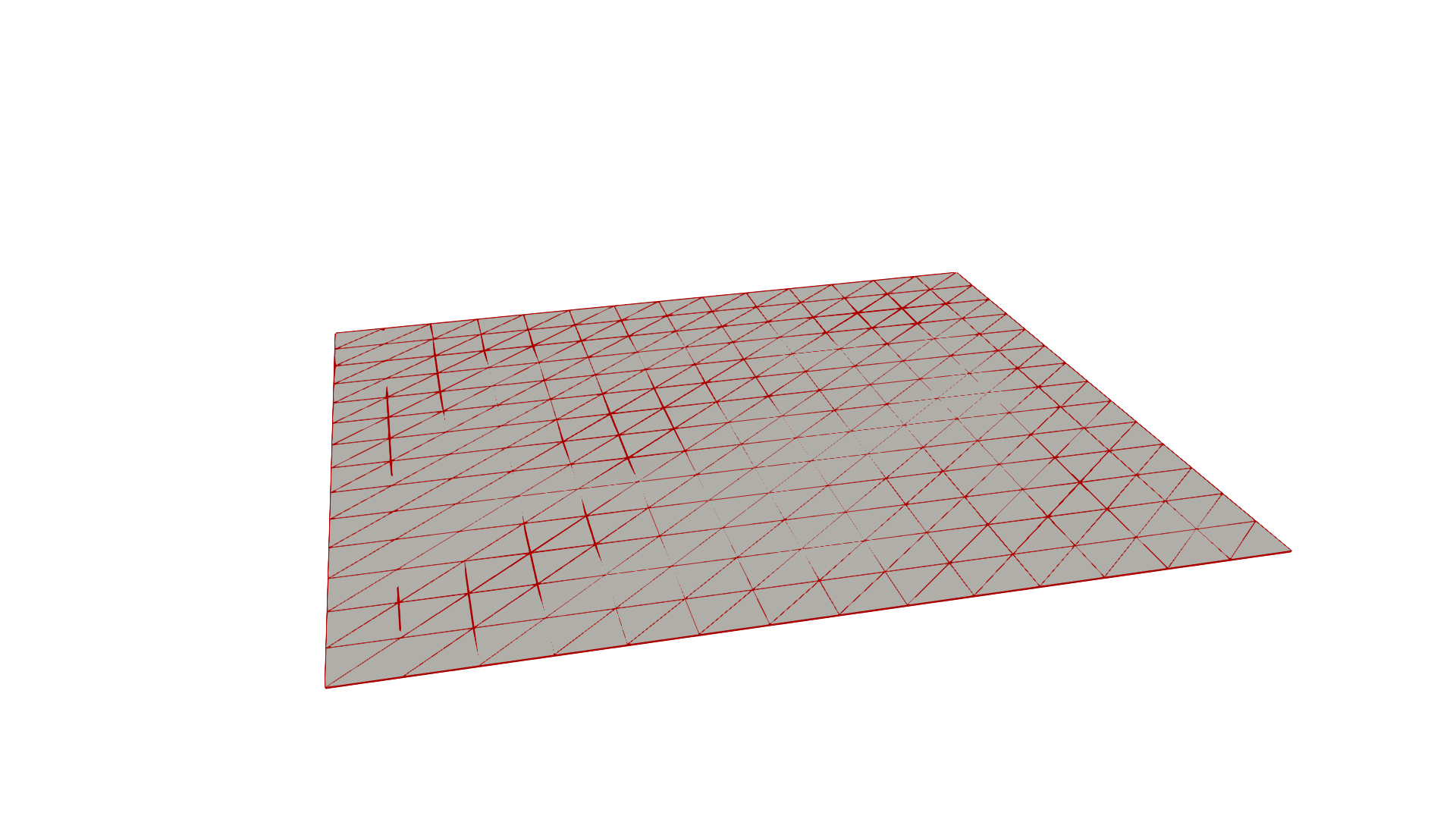}%
    \img{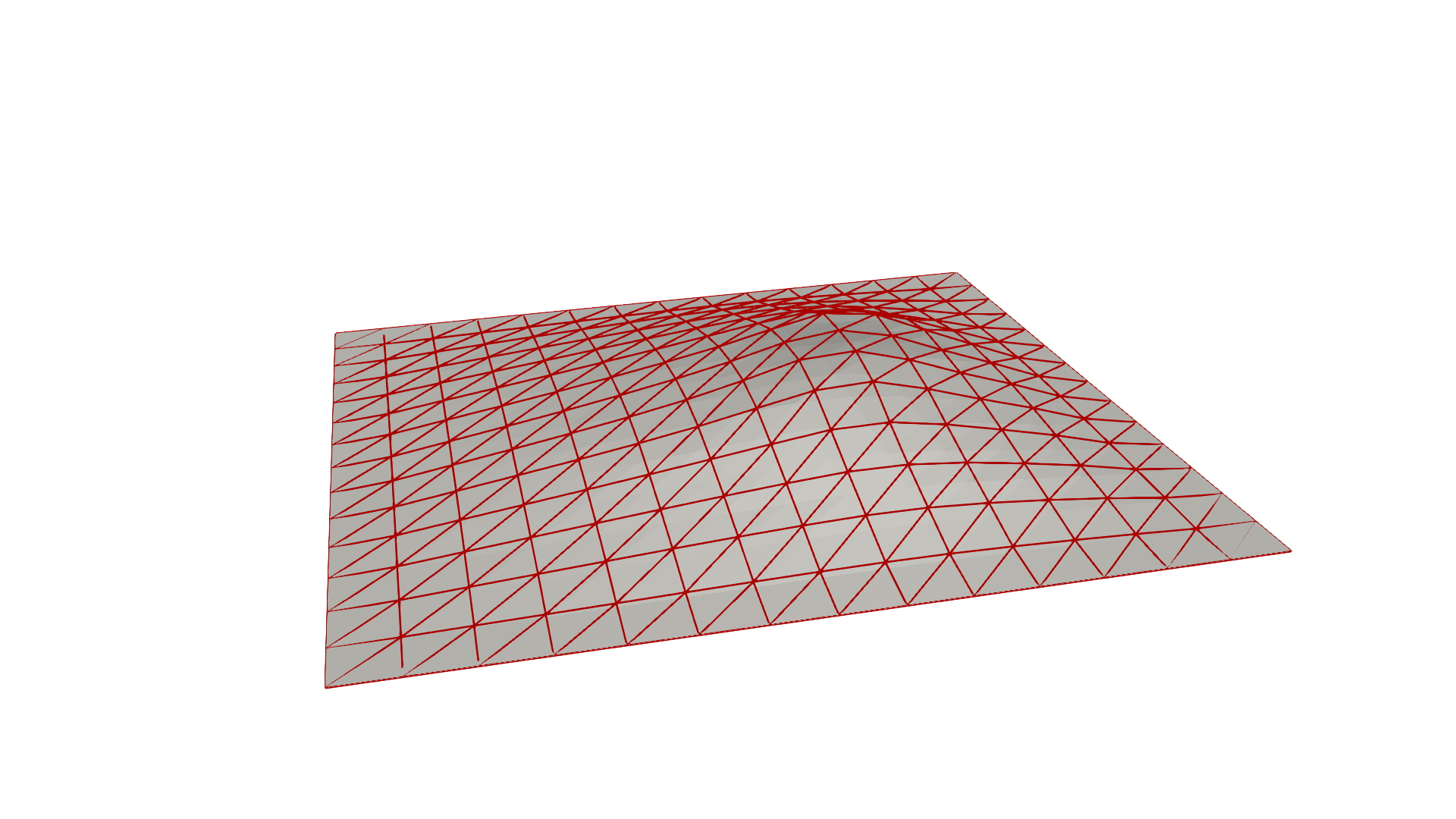}%
    \img{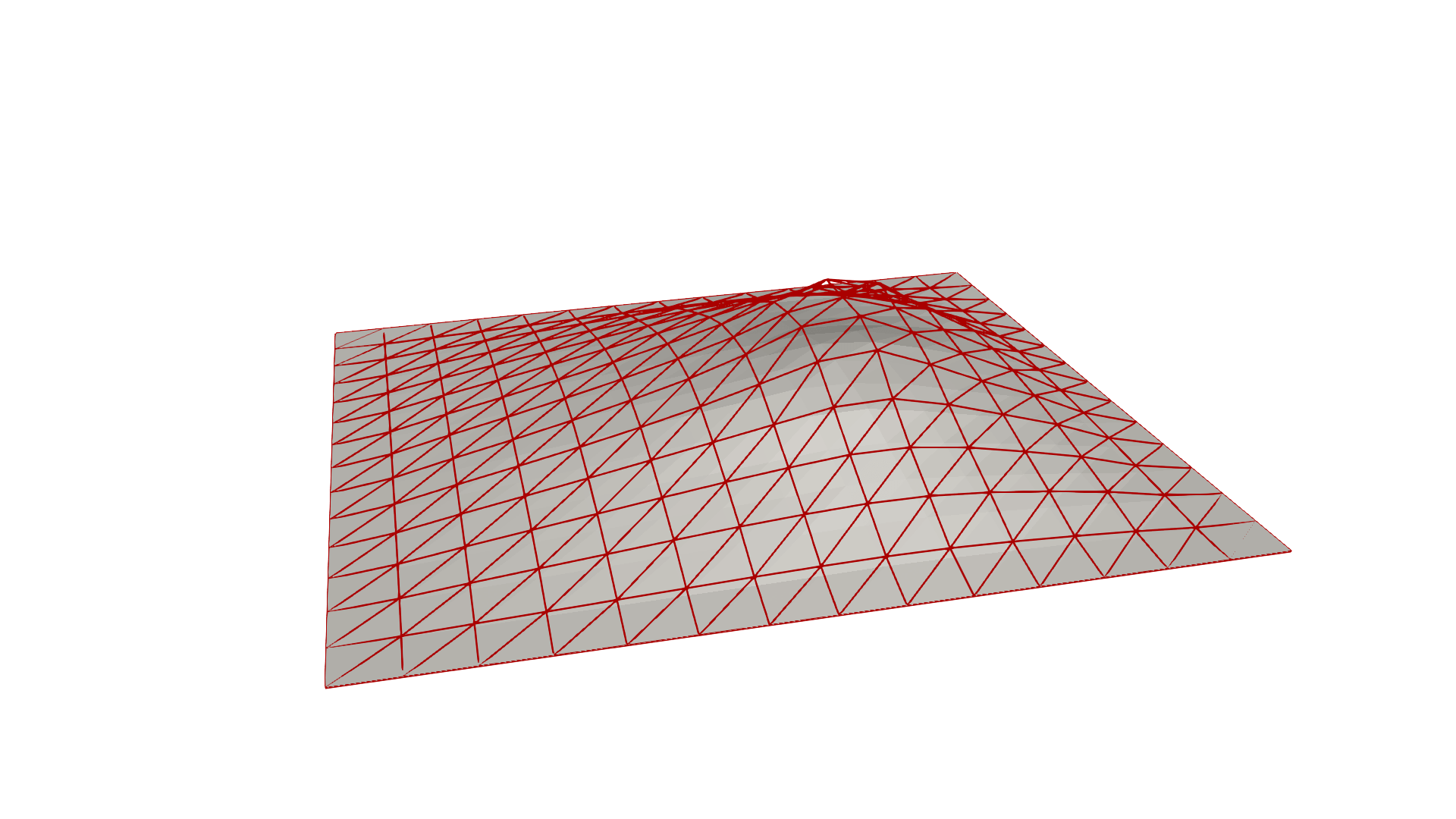}%
    \img{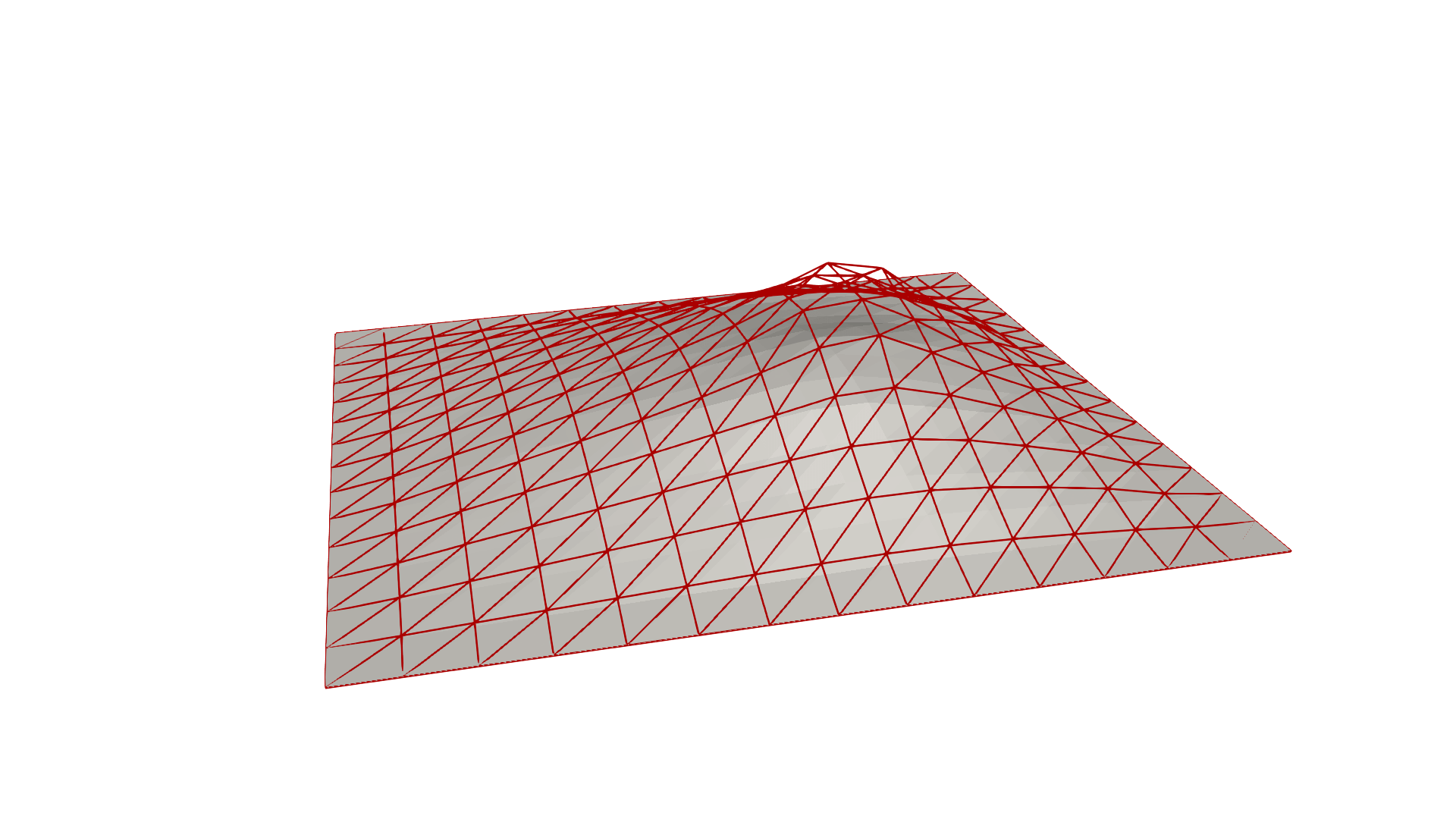}%
    \img{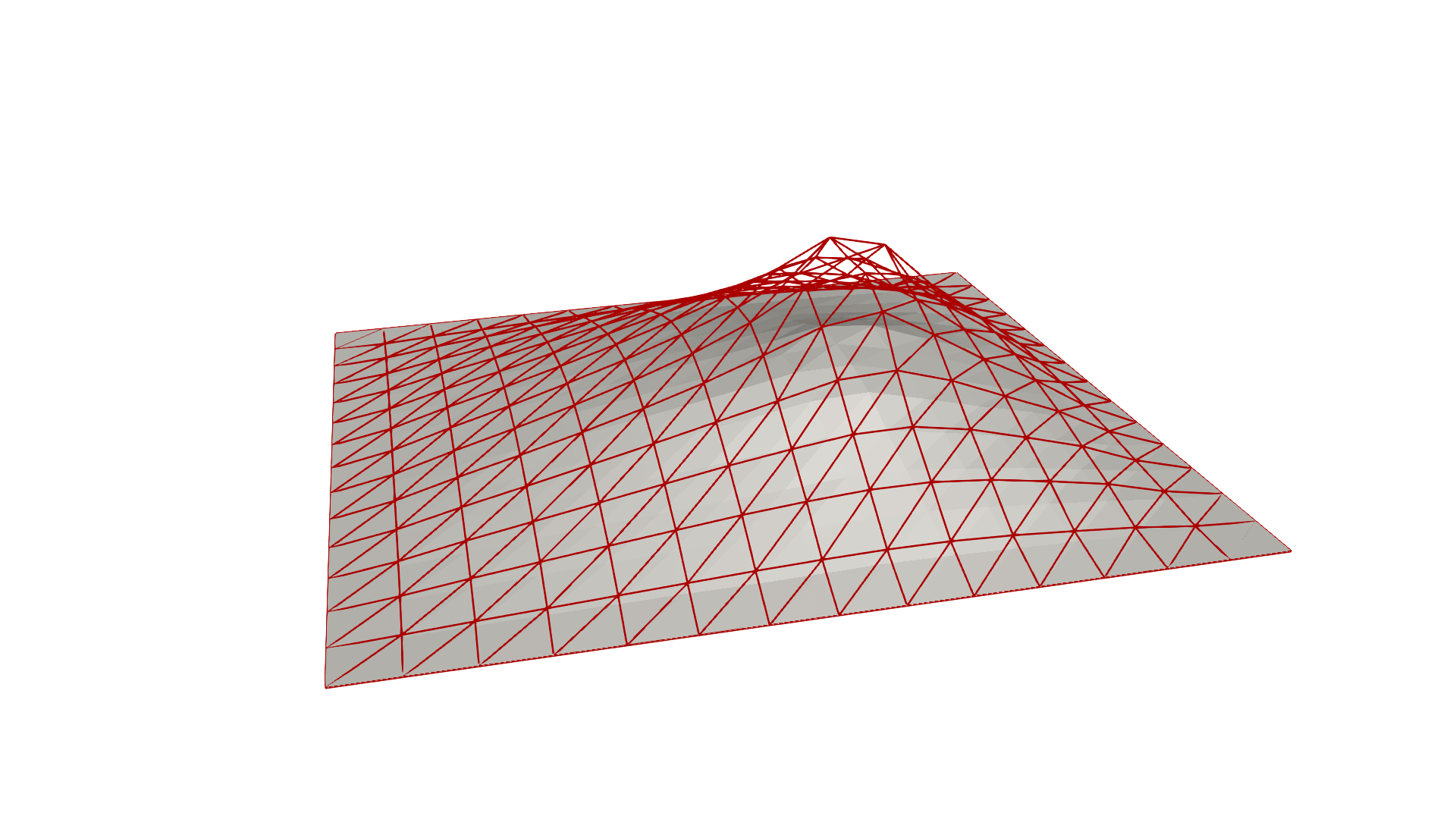}%
    \img{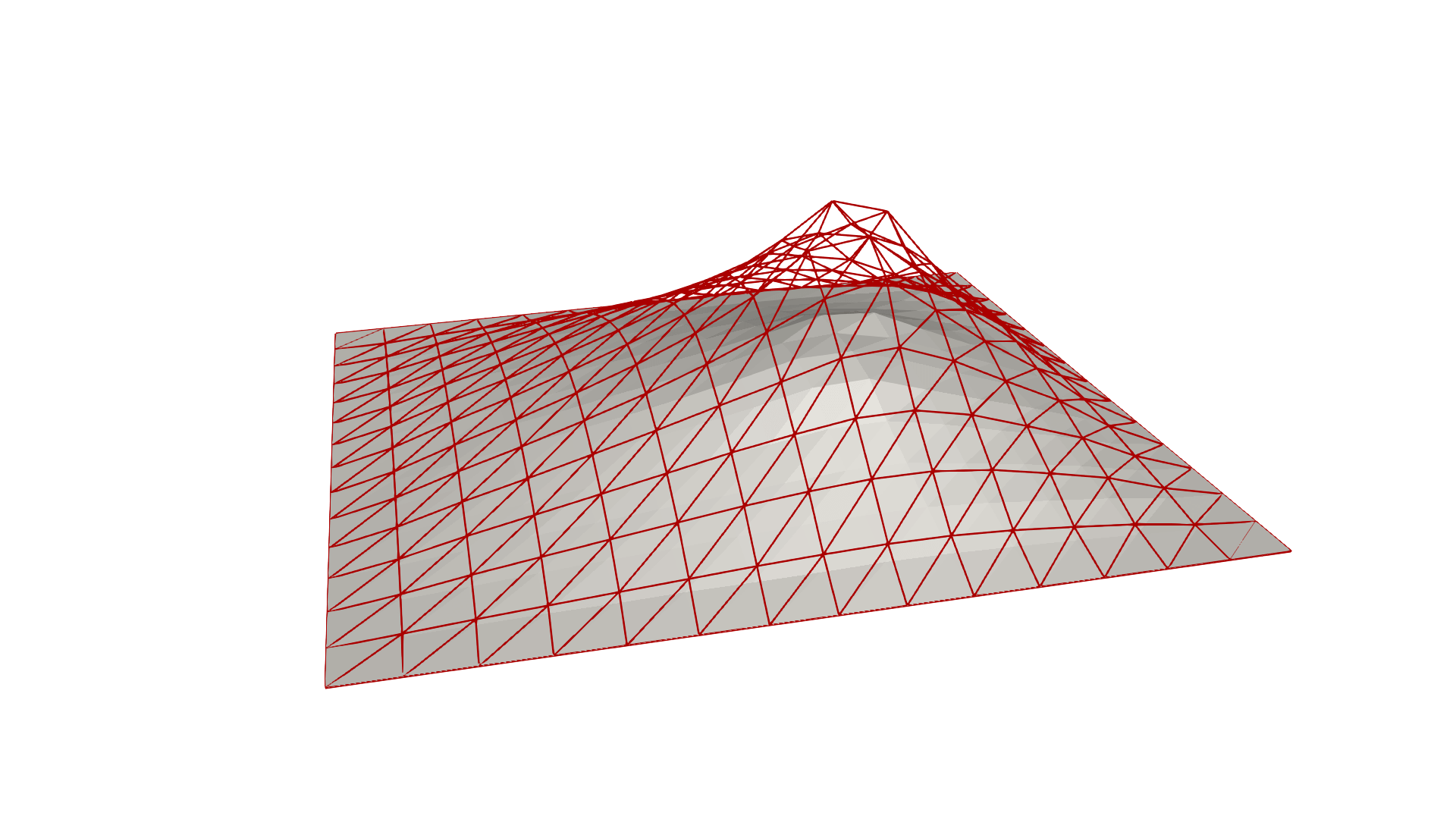}\\[0.4em]

    % Row 3: MANGO Decoder
    \rowlabel{MANGO}%
    \img{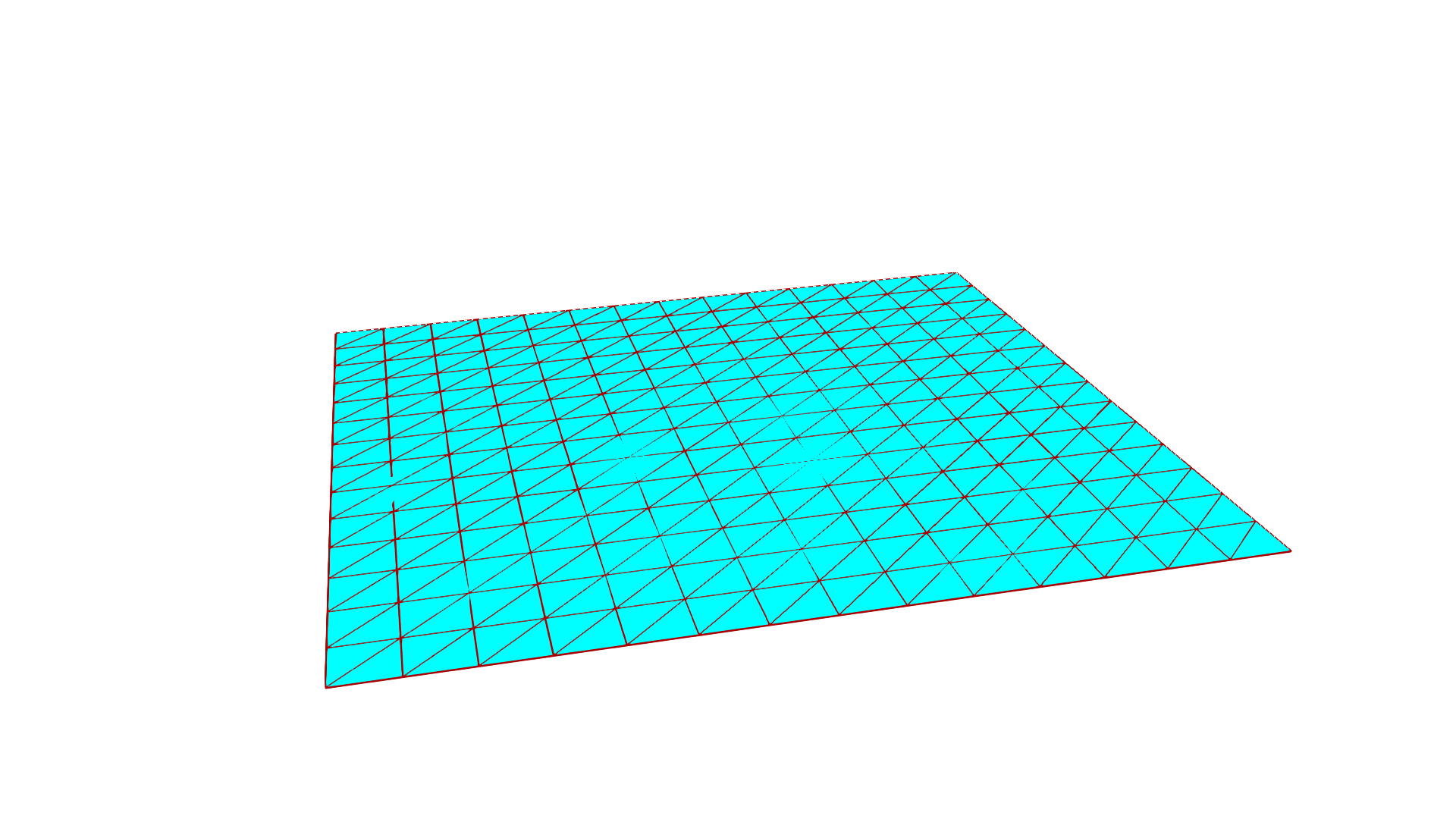}%
    \img{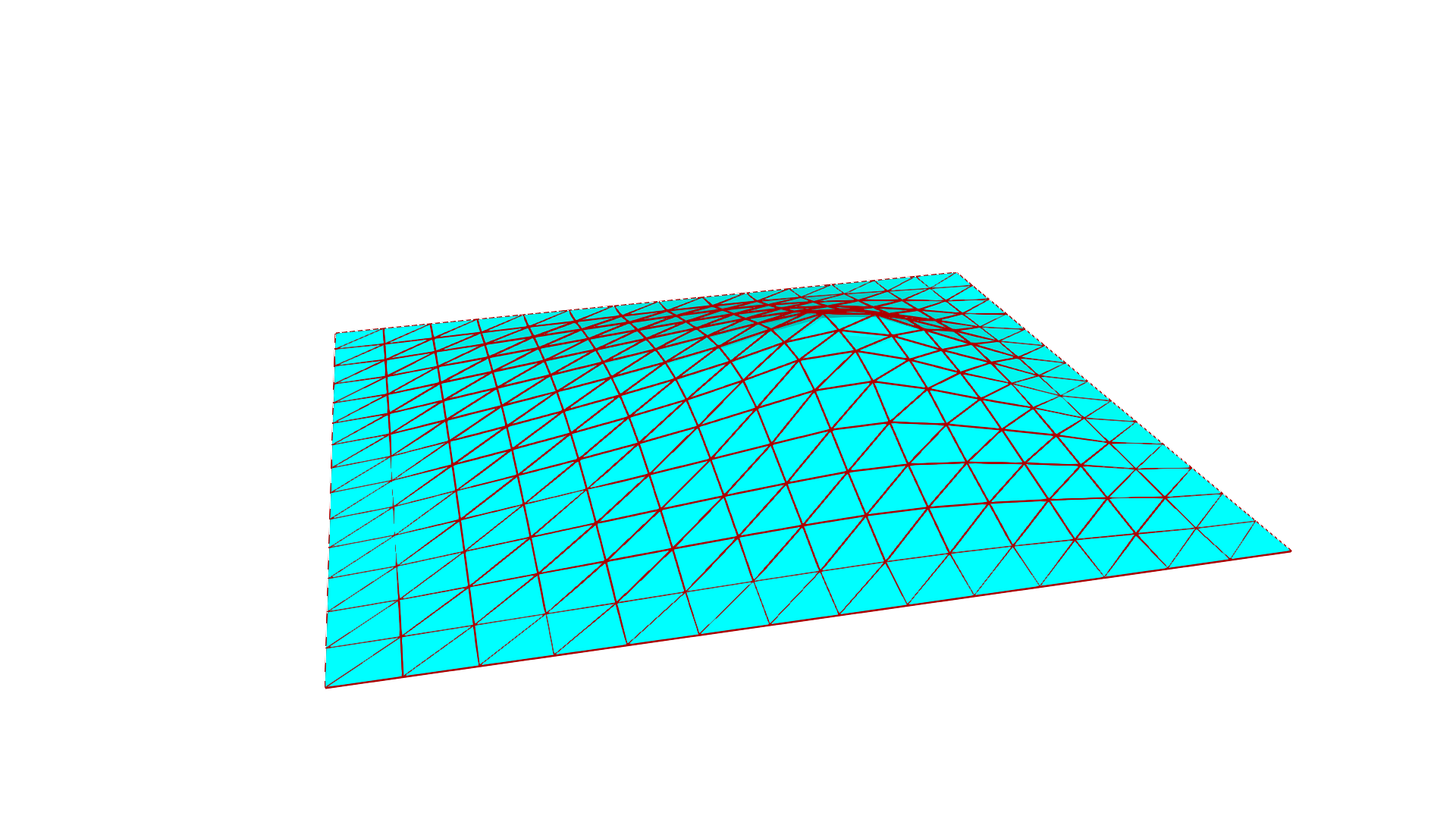}%
    \img{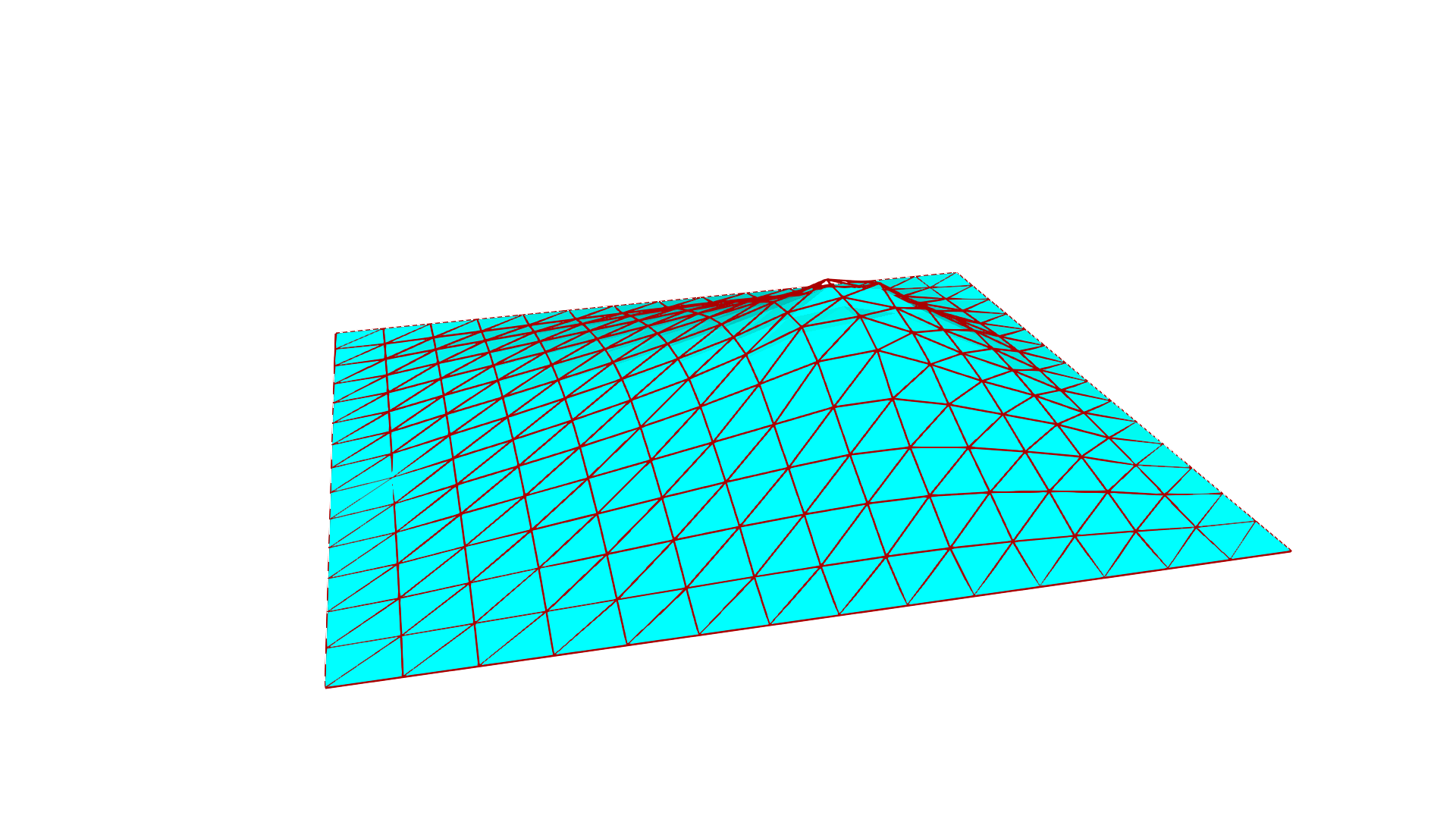}%
    \img{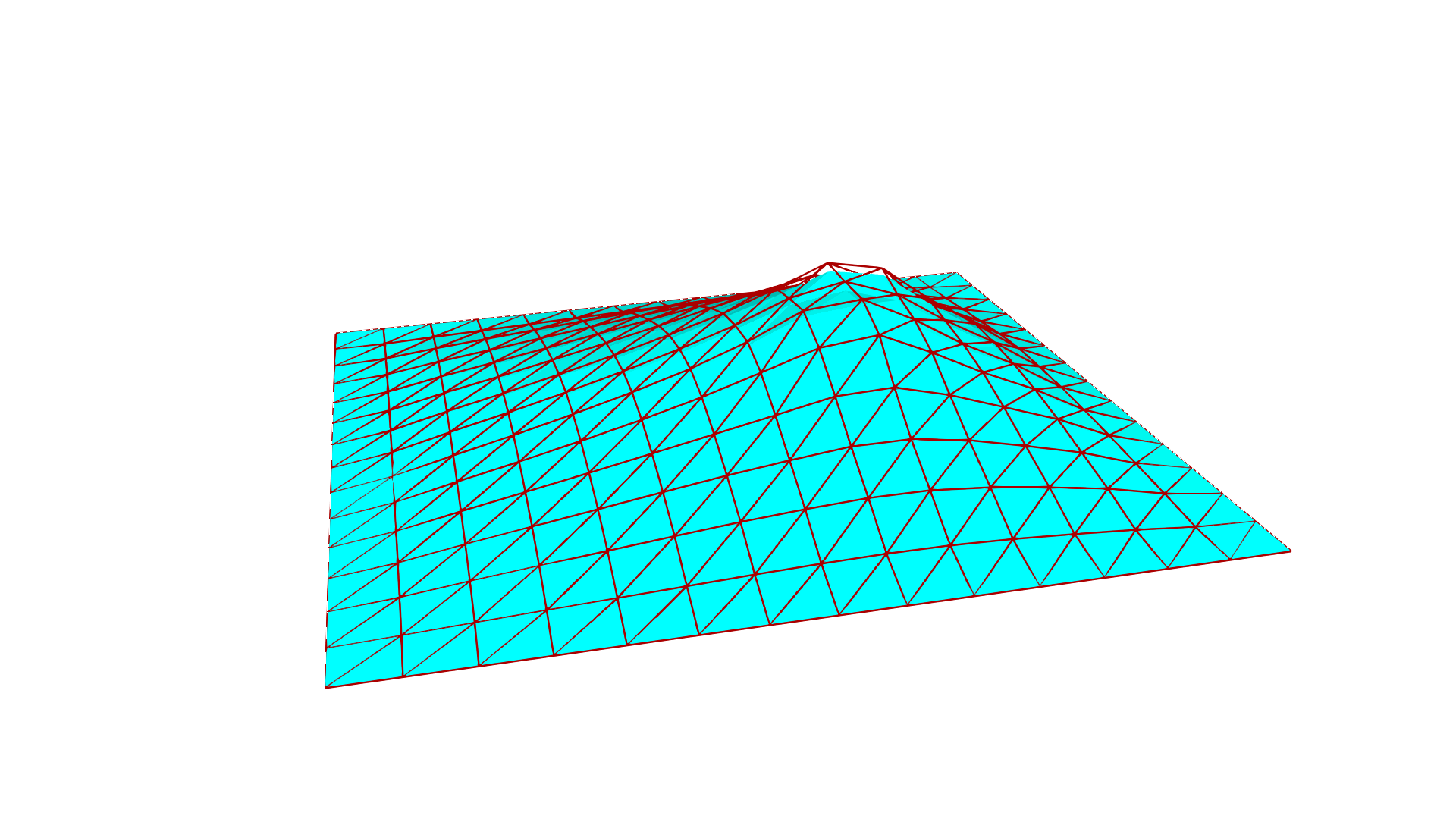}%
    \img{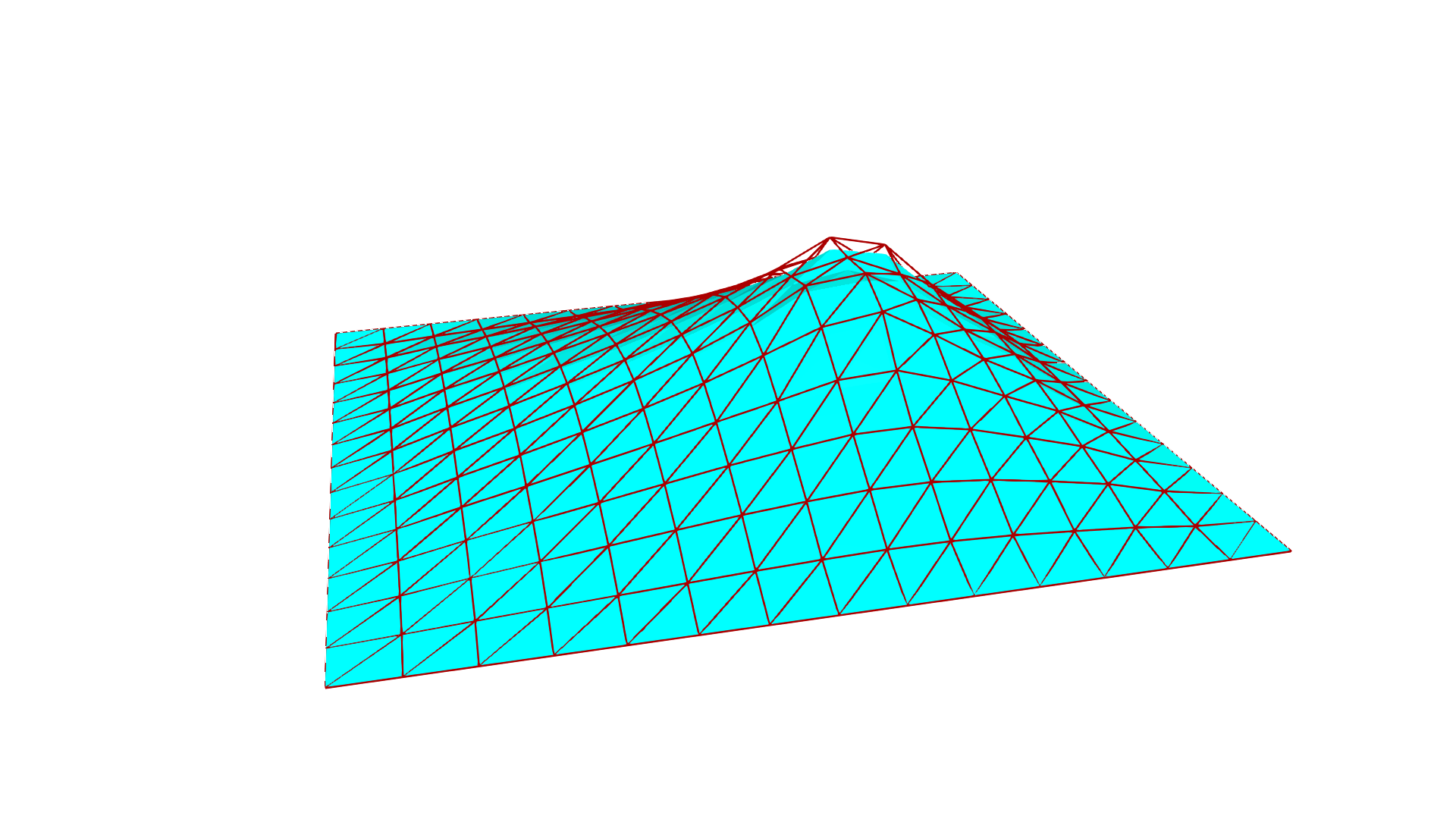}%
    \img{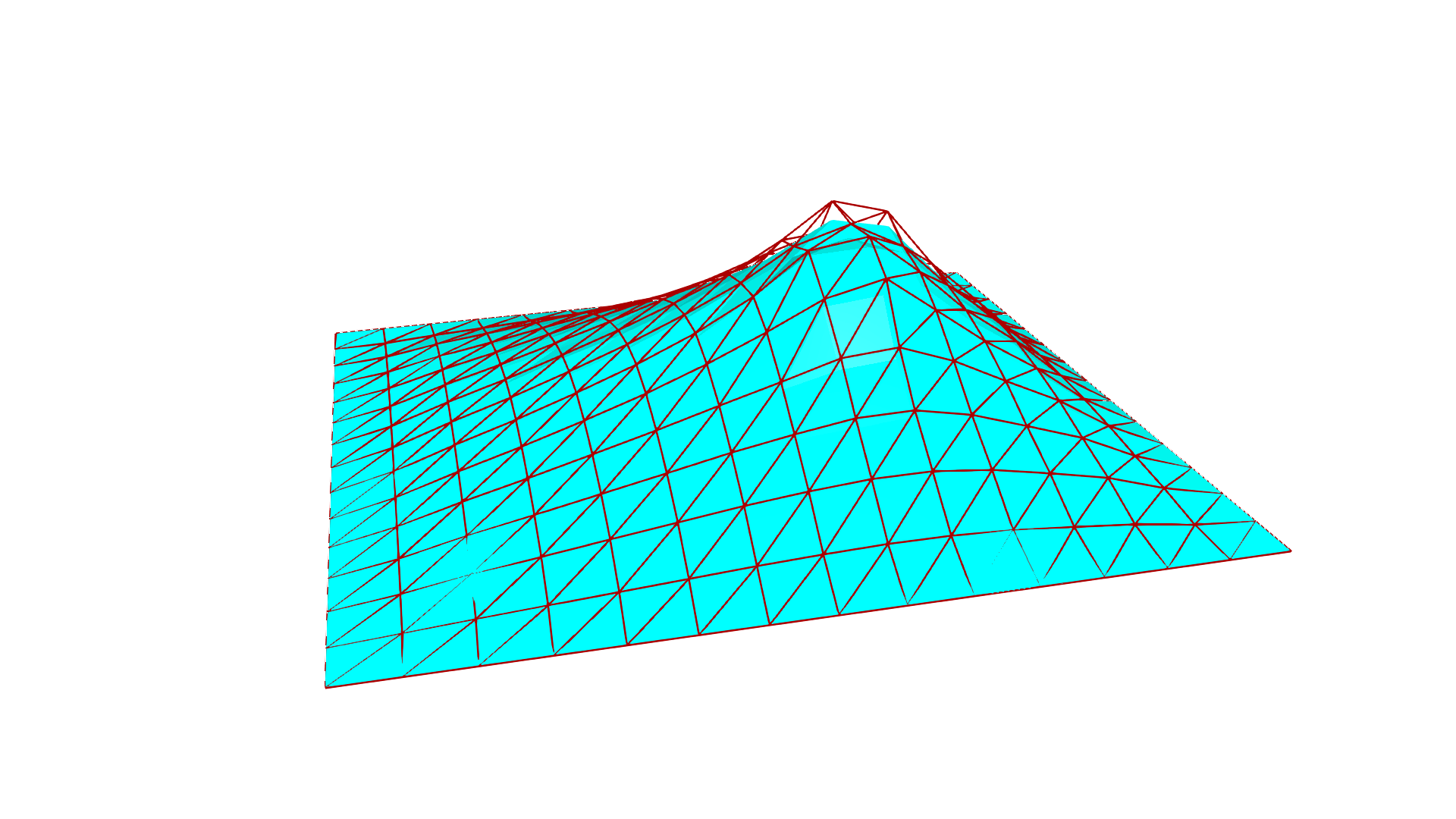}\\[0.4em]

    % Row 4: MANGO Oracle
    \rowlabel{Oracle}%
    \img{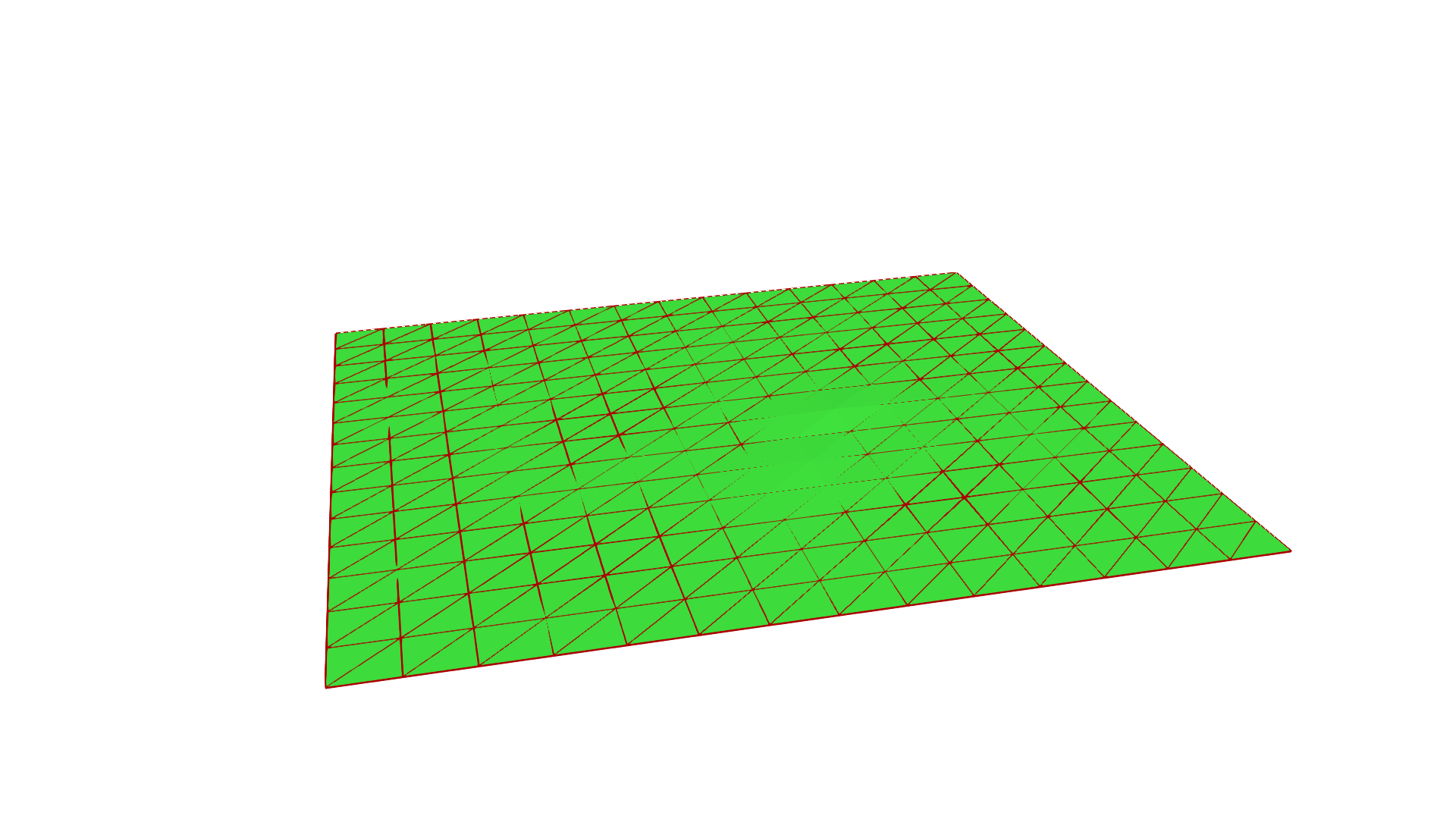}%
    \img{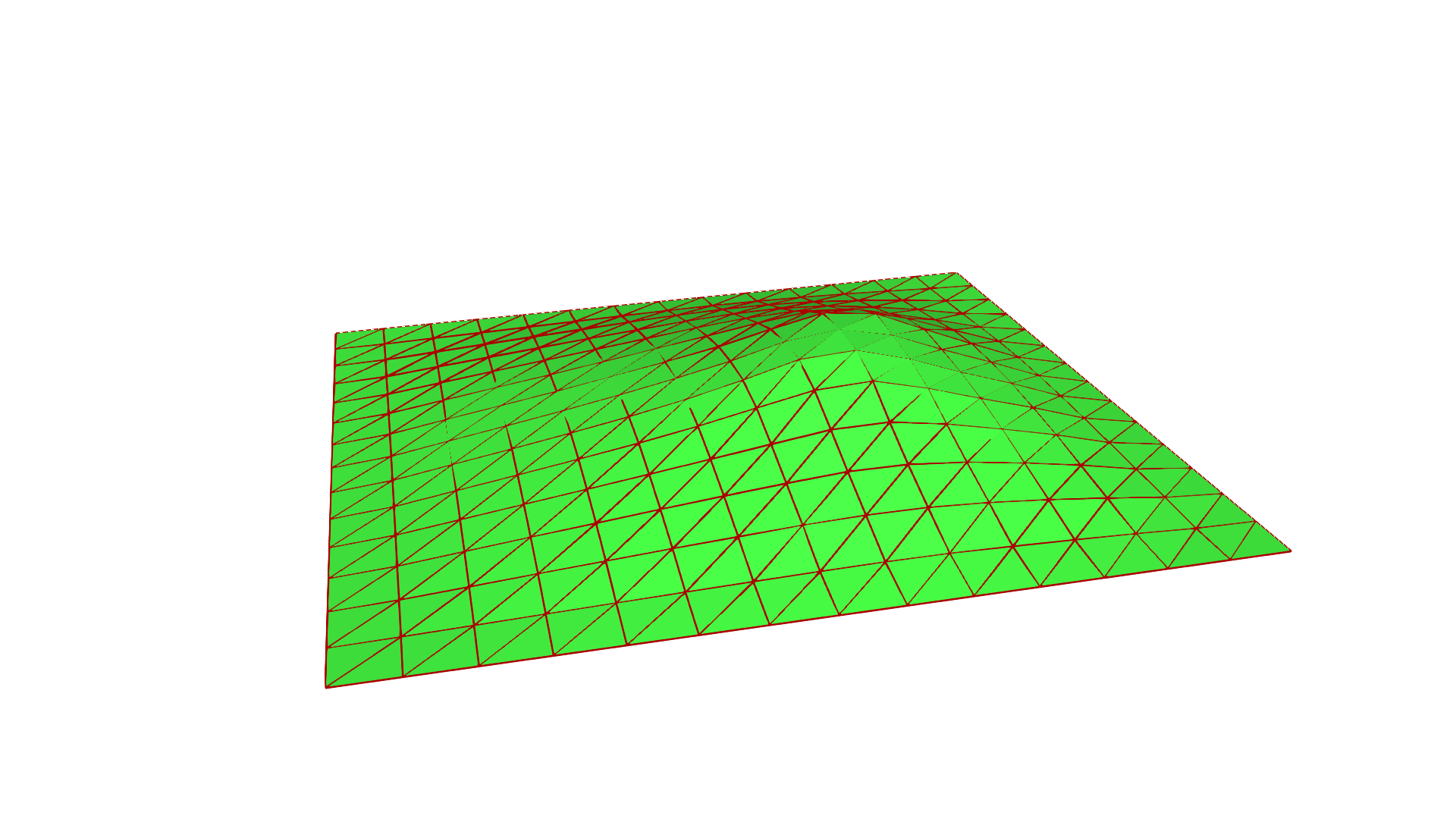}%
    \img{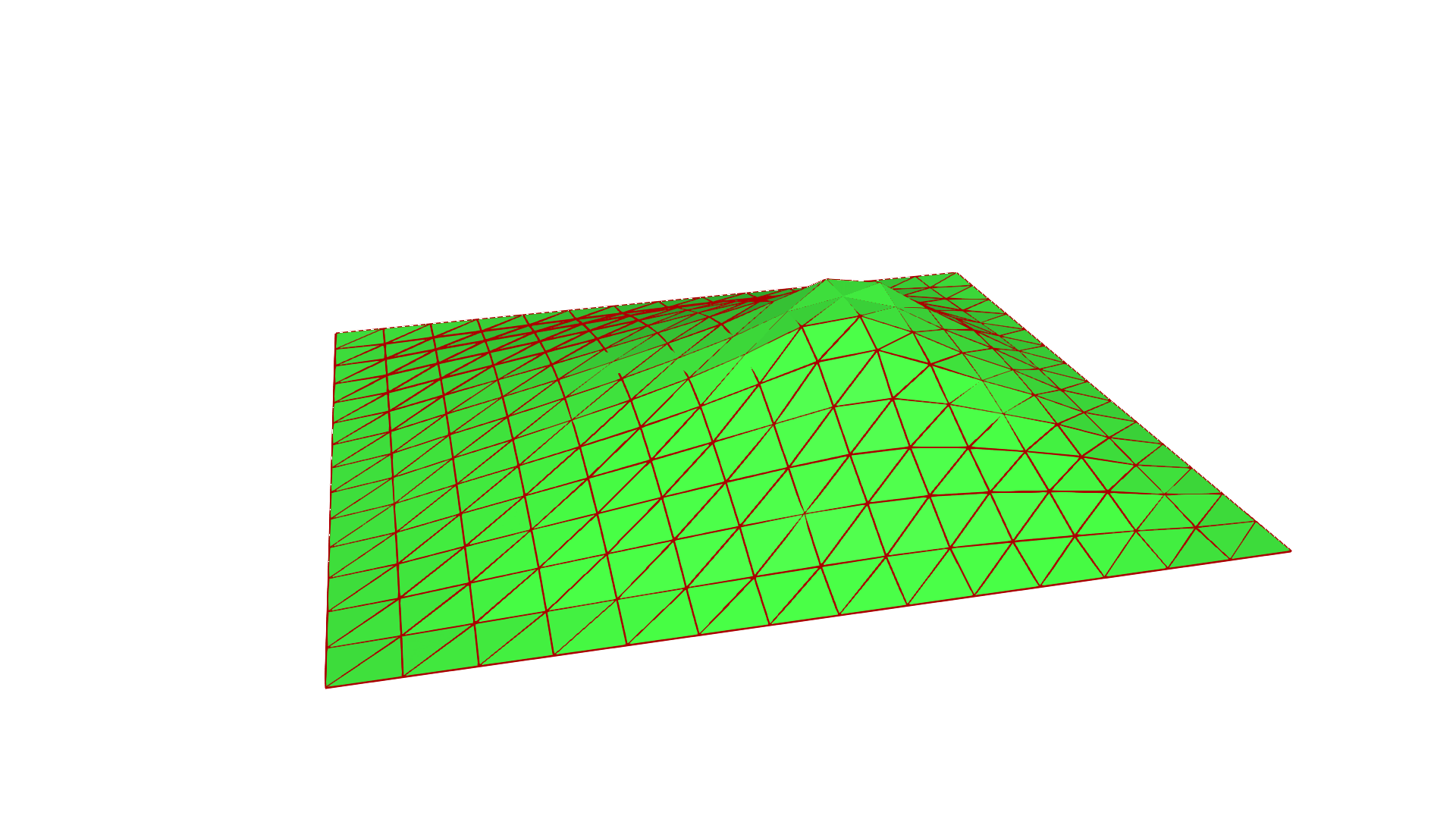}%
    \img{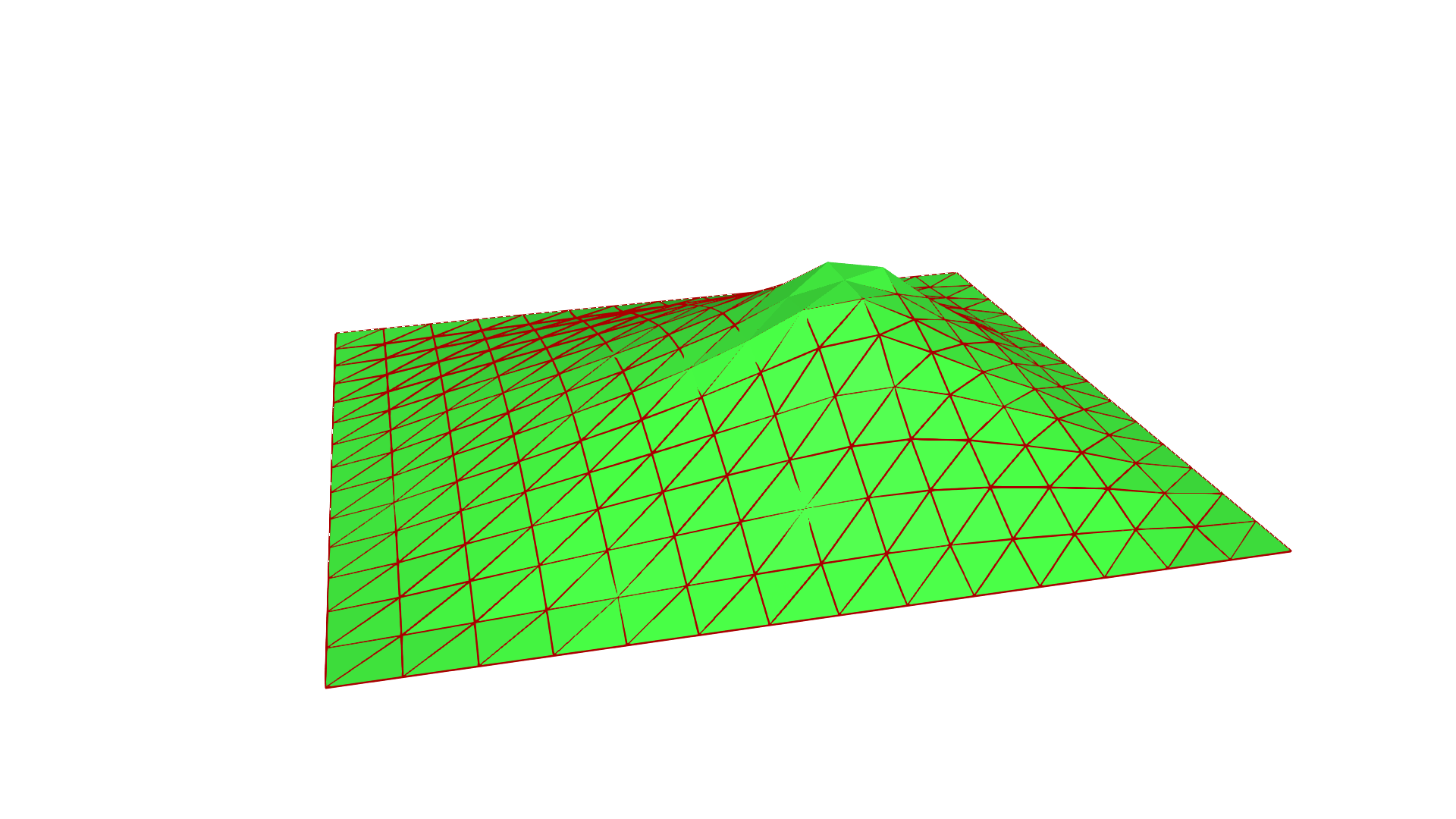}%
    \img{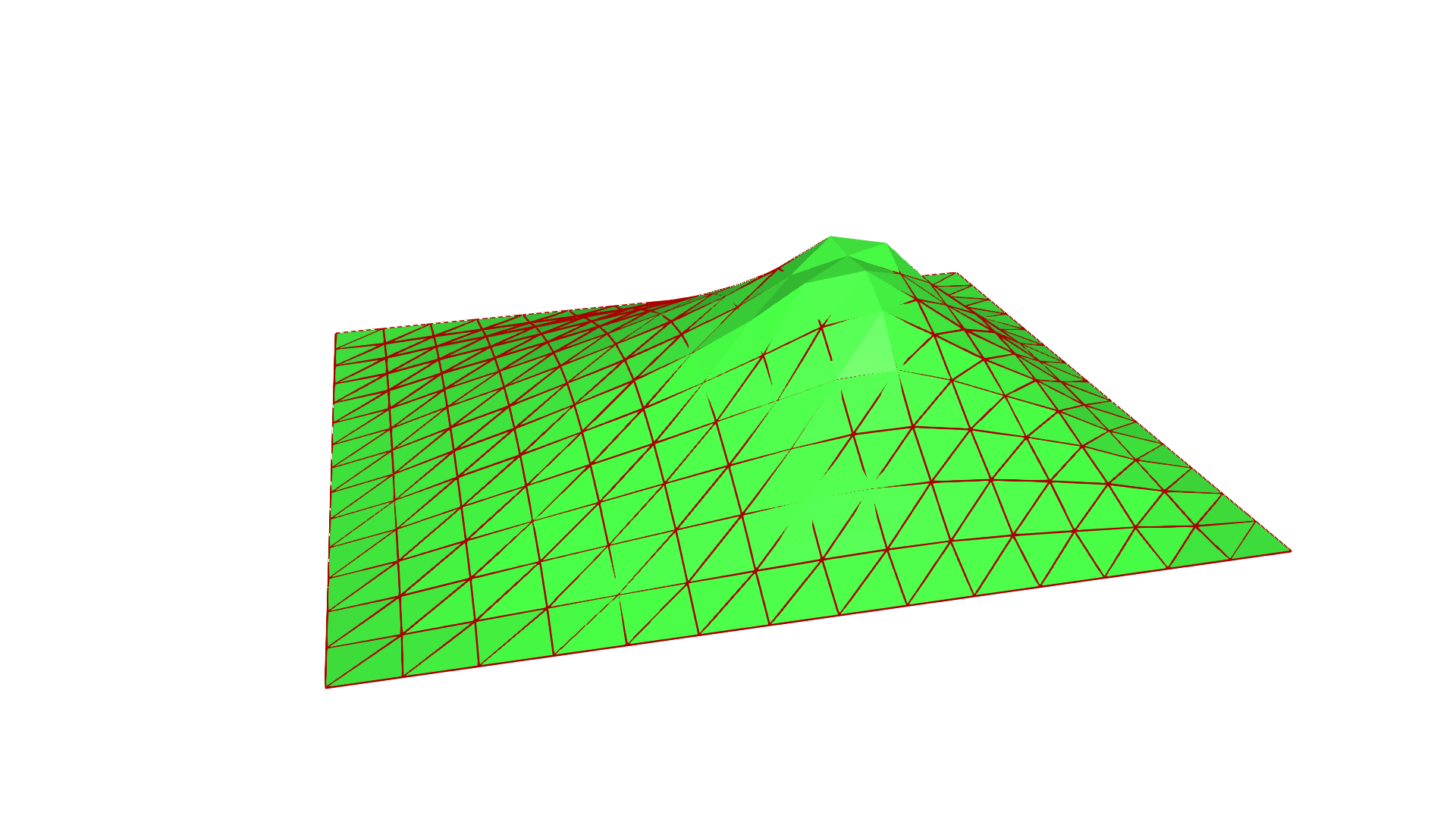}%
    \img{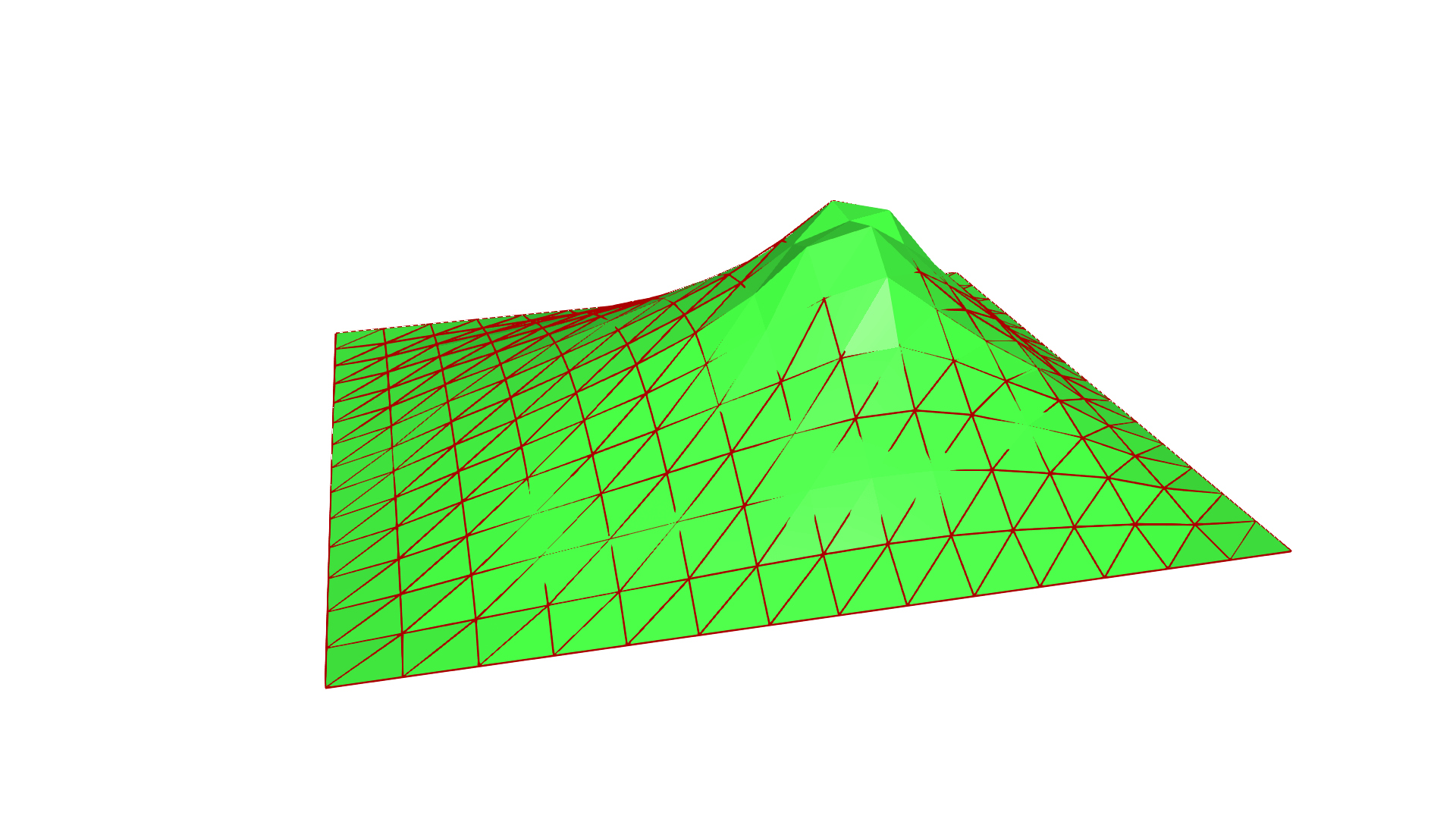}\\[0.4em]

    % Row 5: MGN
    \rowlabel{No Context \\ (MGN)}%
    \img{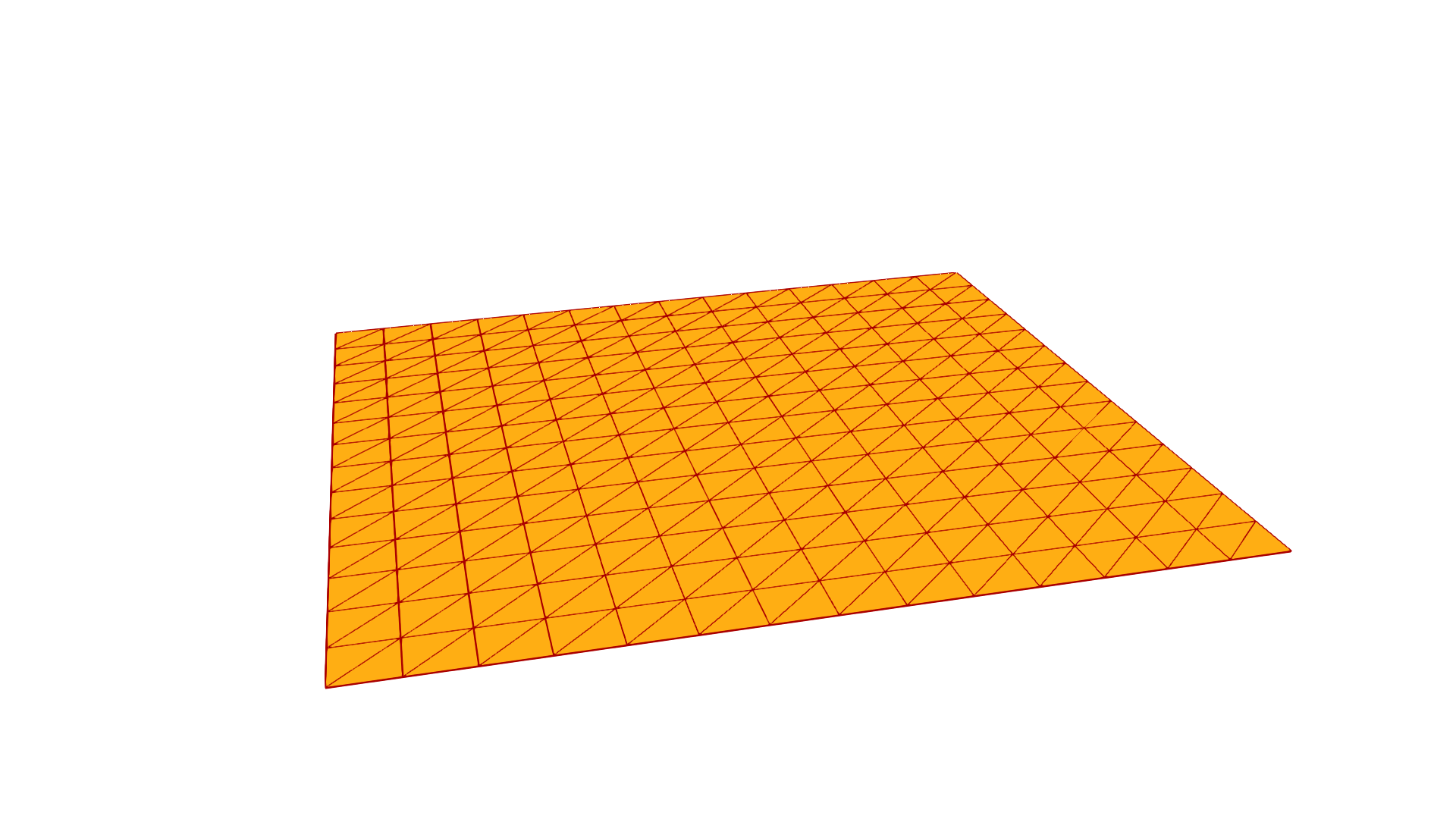}%
    \img{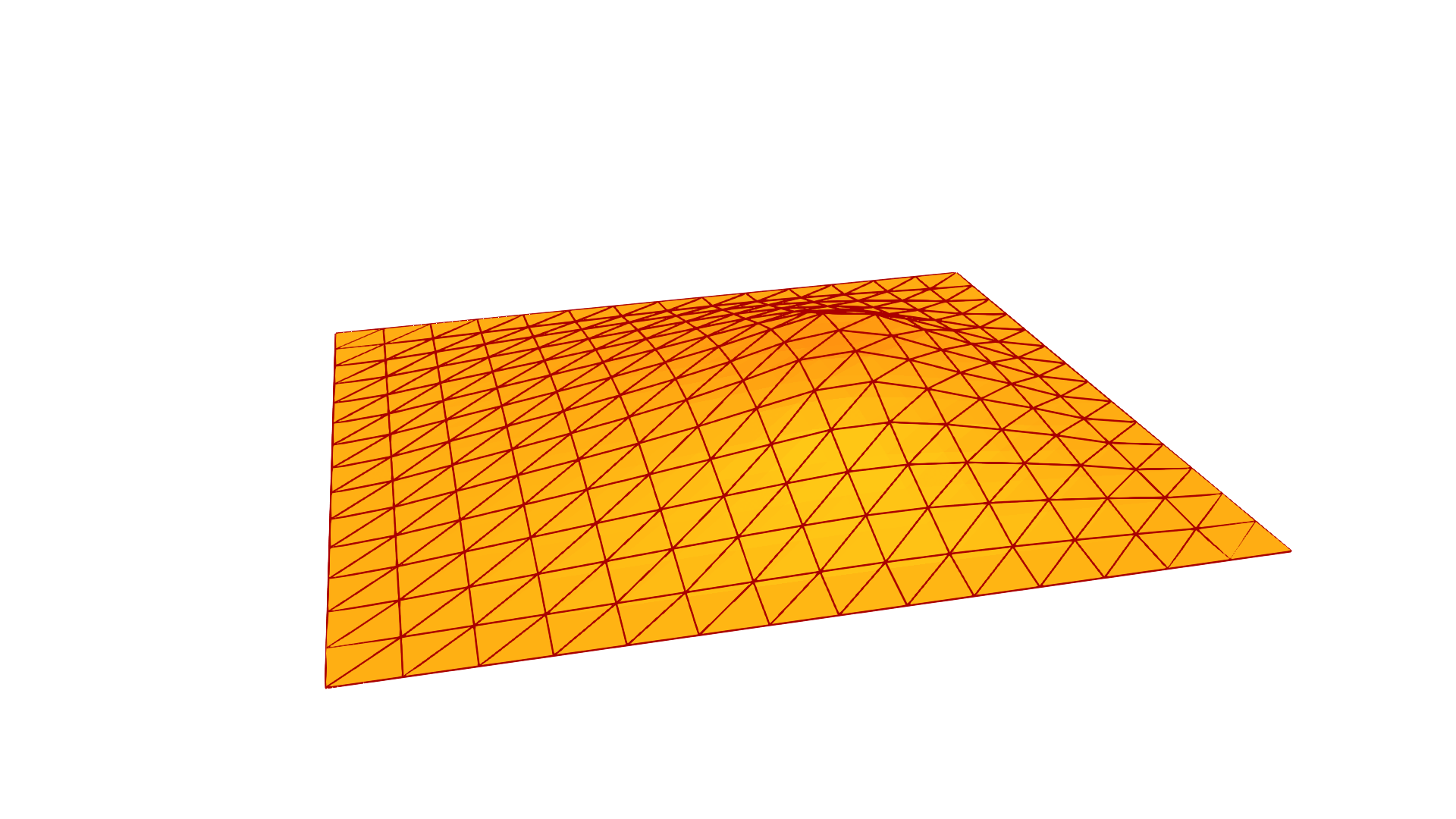}%
    \img{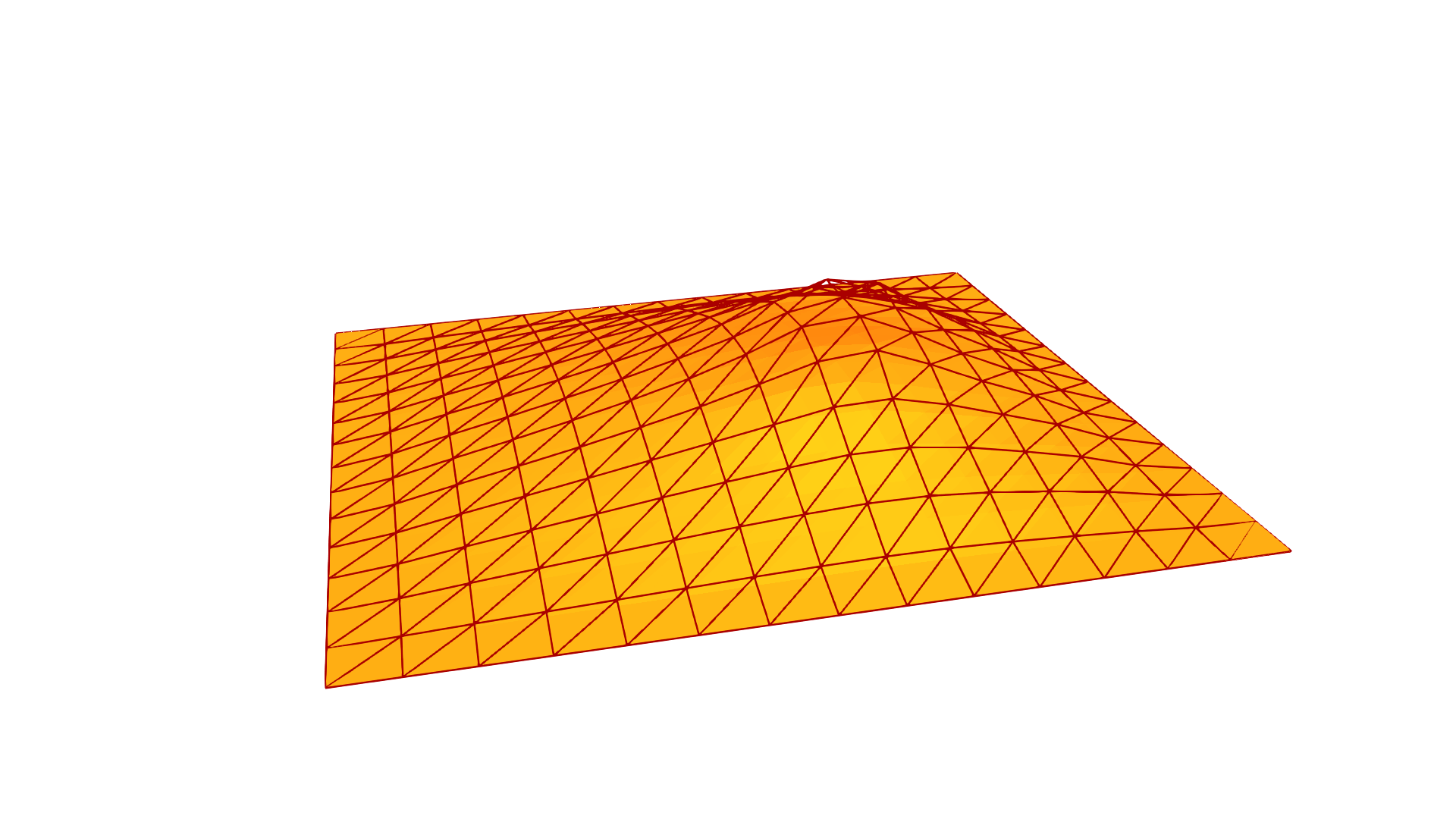}%
    \img{figures/qualitative_trajectories_sd/MGN/frame_0021.png}%
    \img{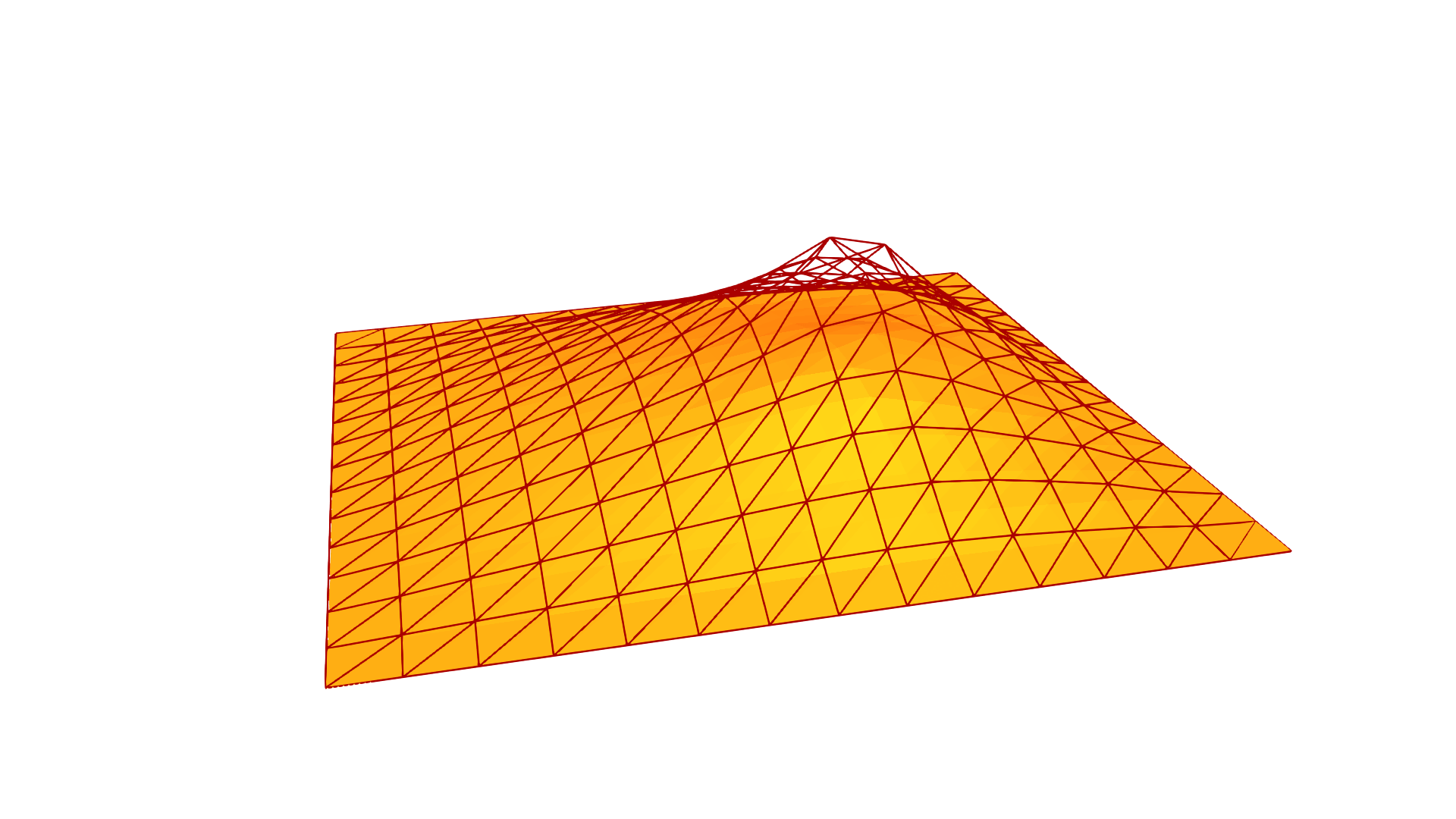}%
    \img{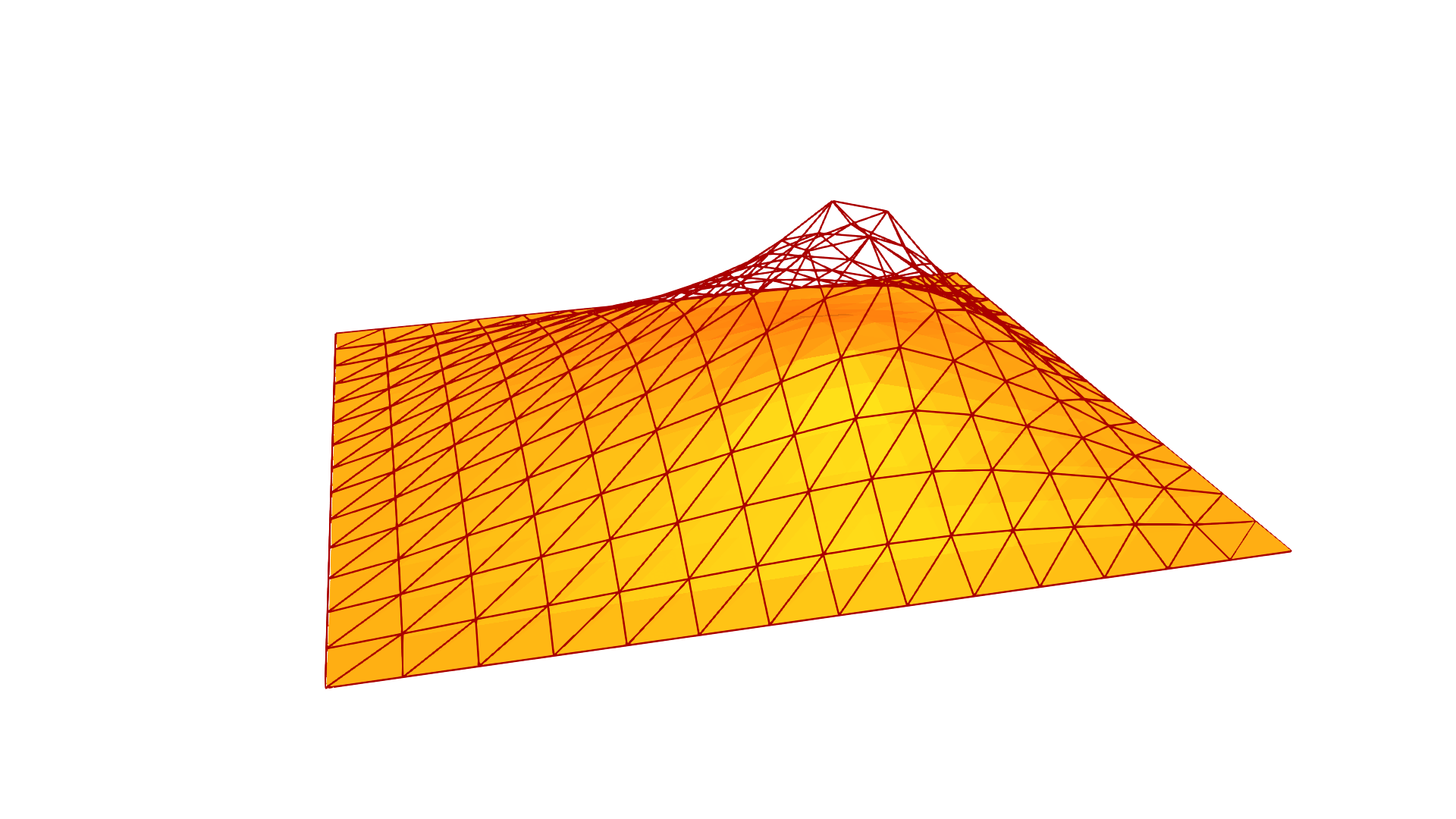}\\[0.4em]

    % Row 6: MGN Oracle
    \rowlabel{Oracle \\ (MGN)}%
    \img{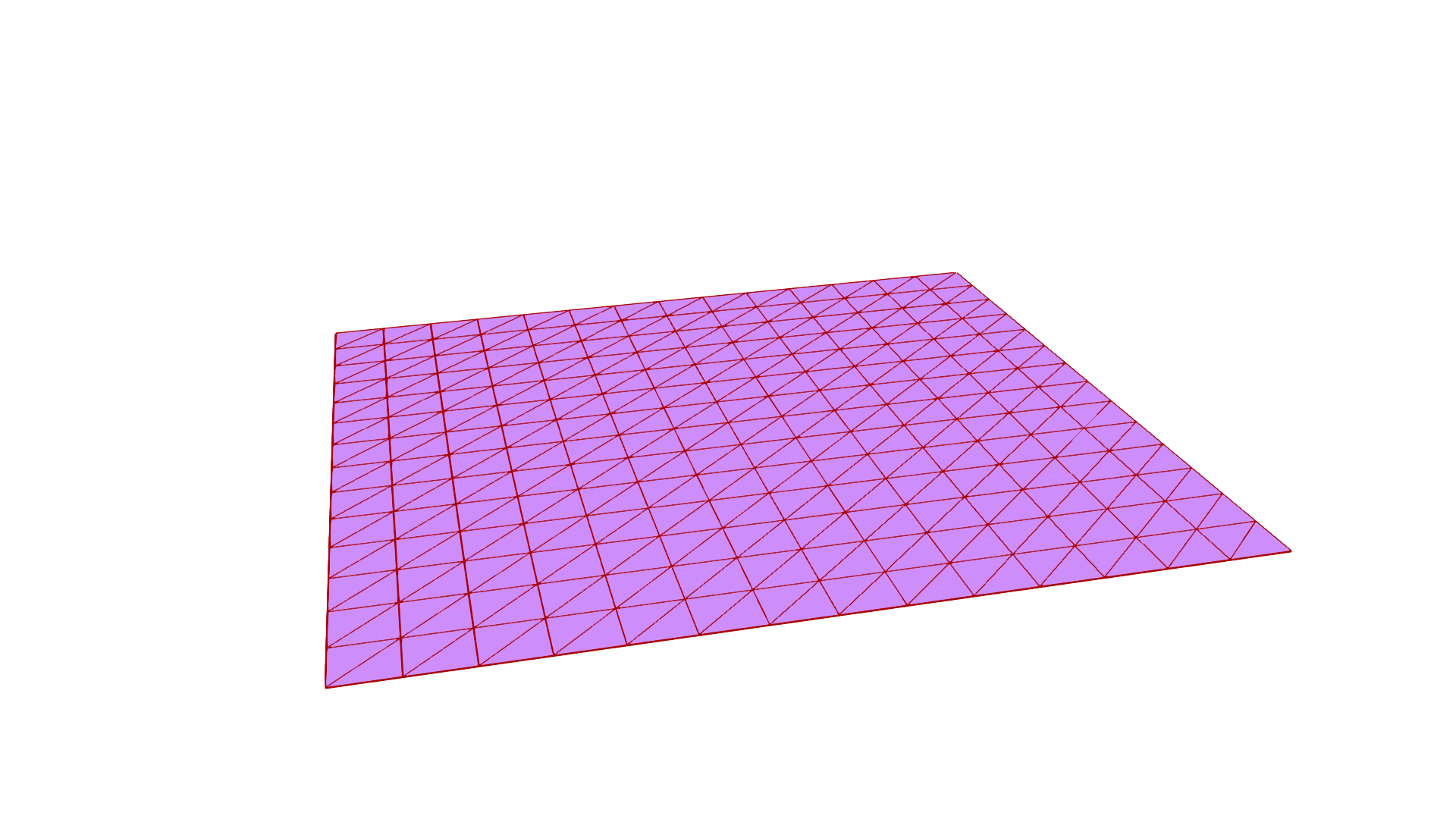}%
    \img{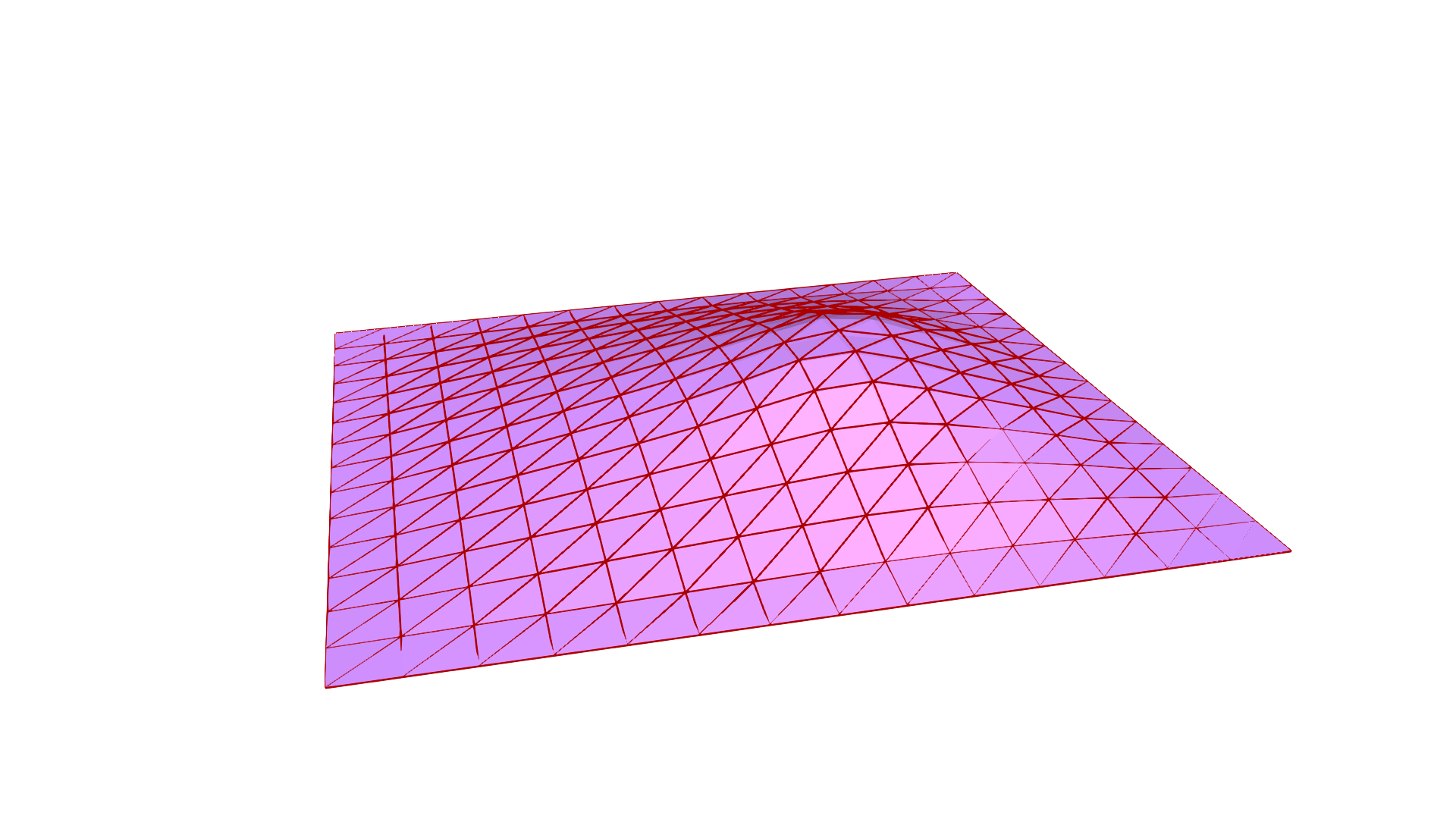}%
    \img{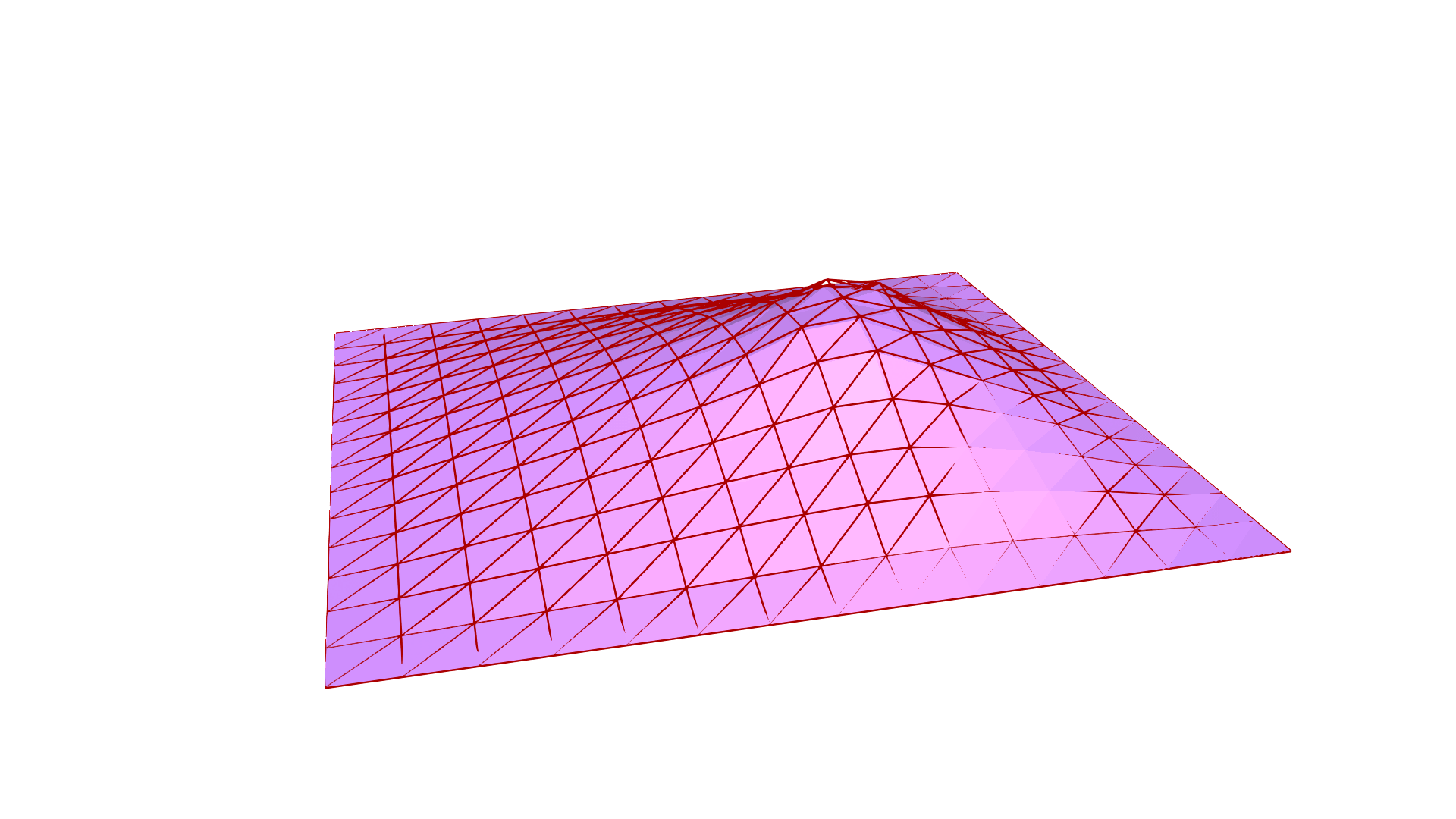}%
    \img{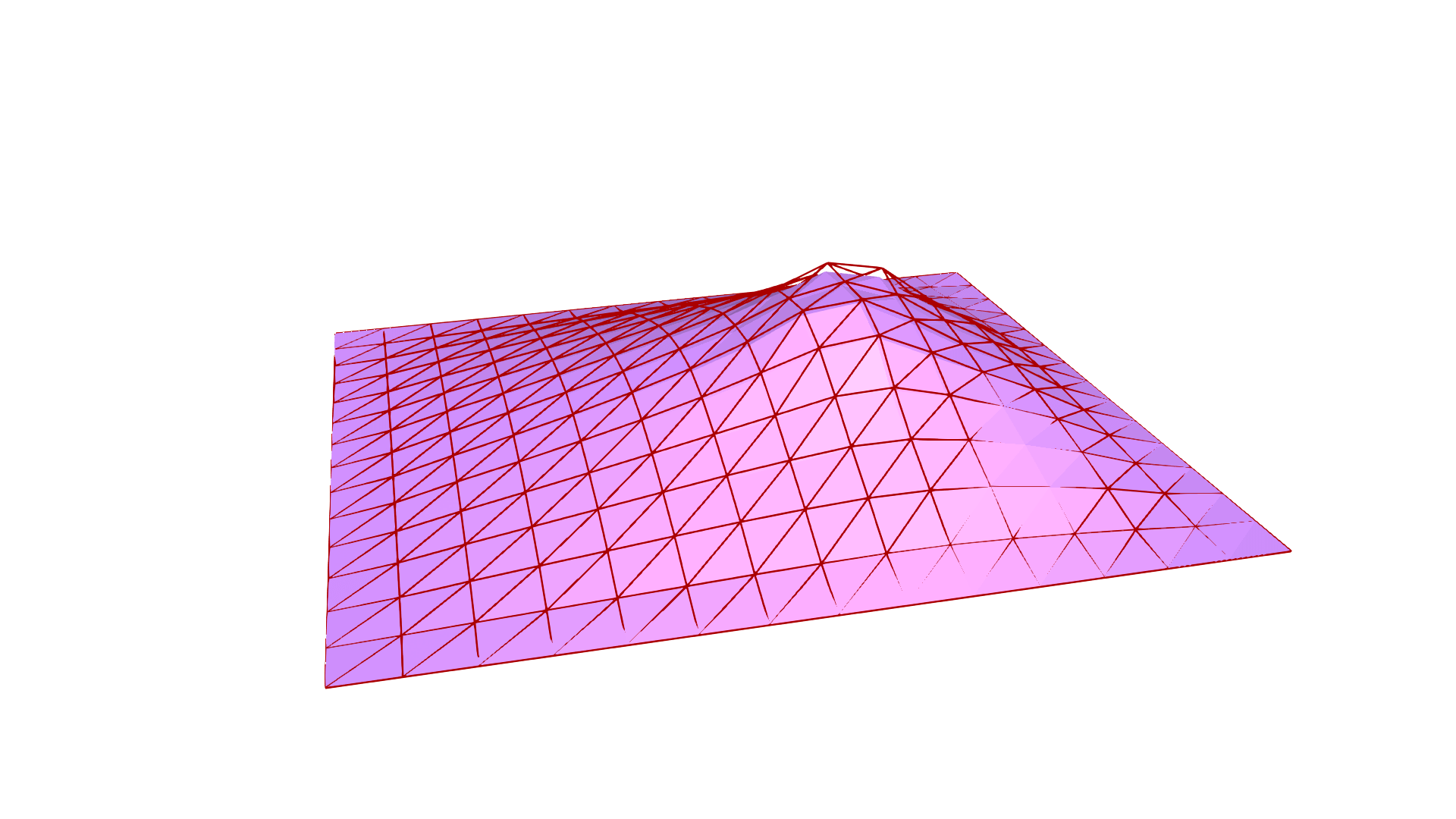}%
    \img{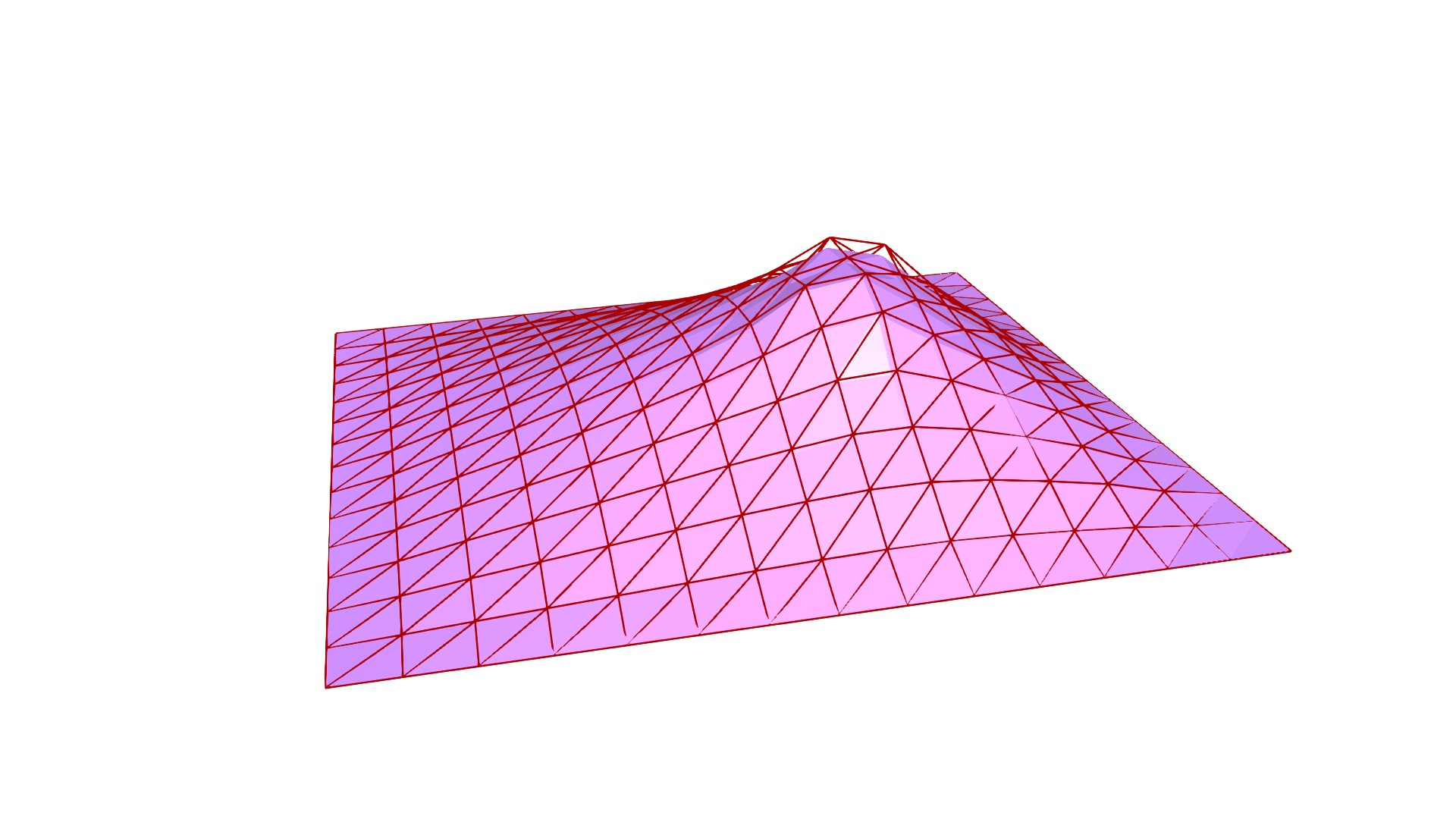}%
    \img{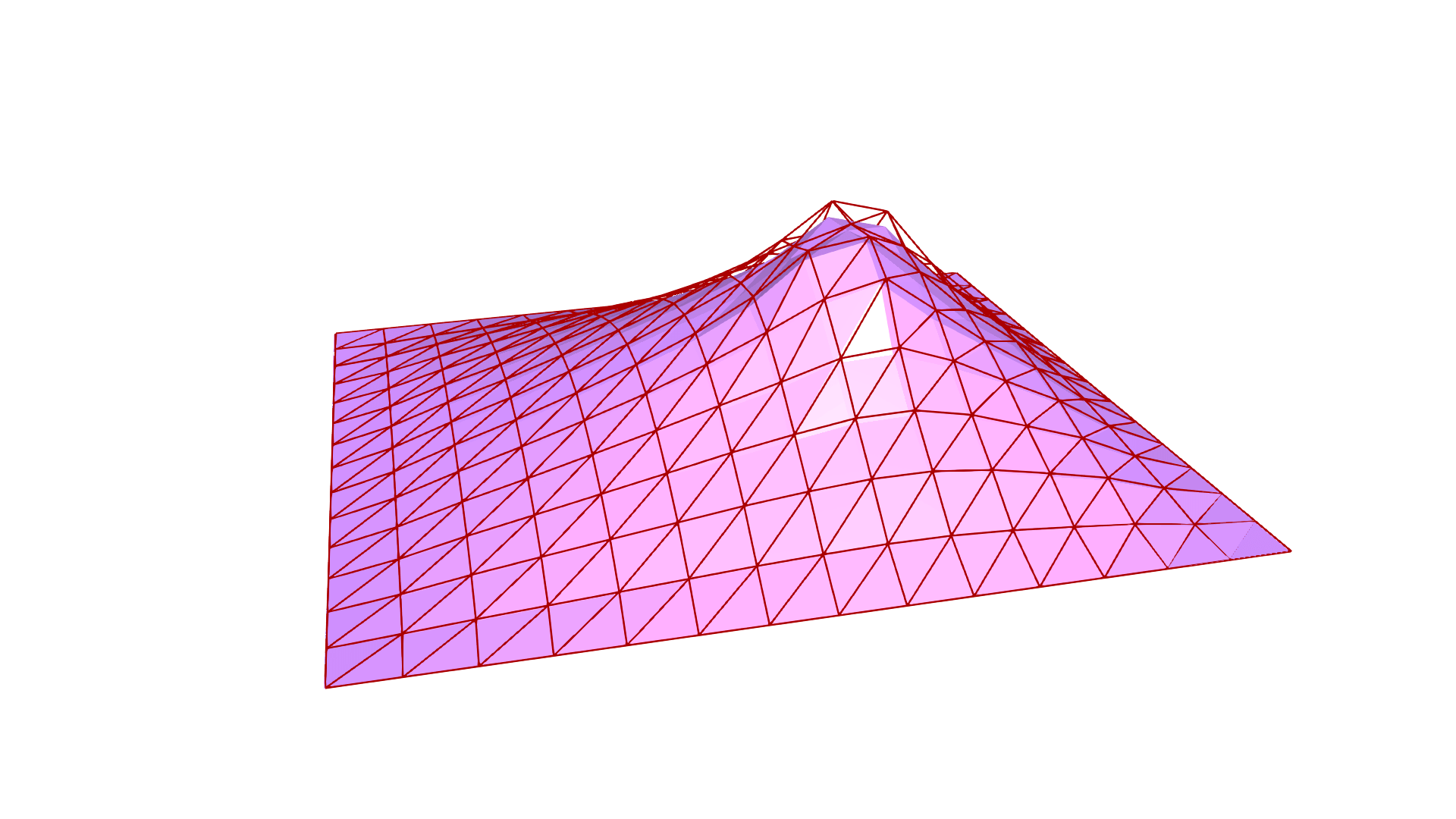}\\[0.4em]

    % Row 7: GNN Encoder
    \rowlabel{GNN \\ Encoder}%
    \img{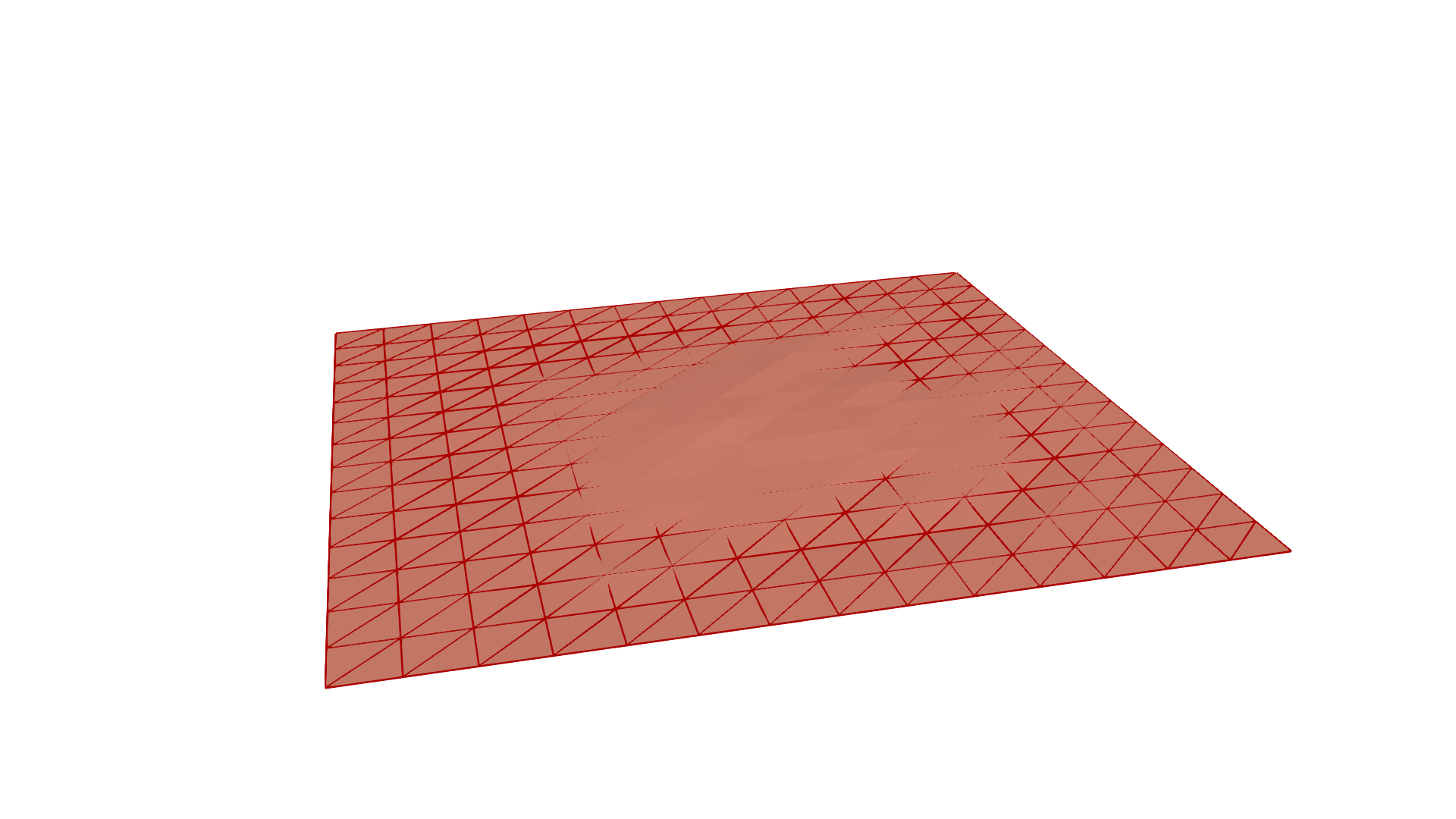}%
    \img{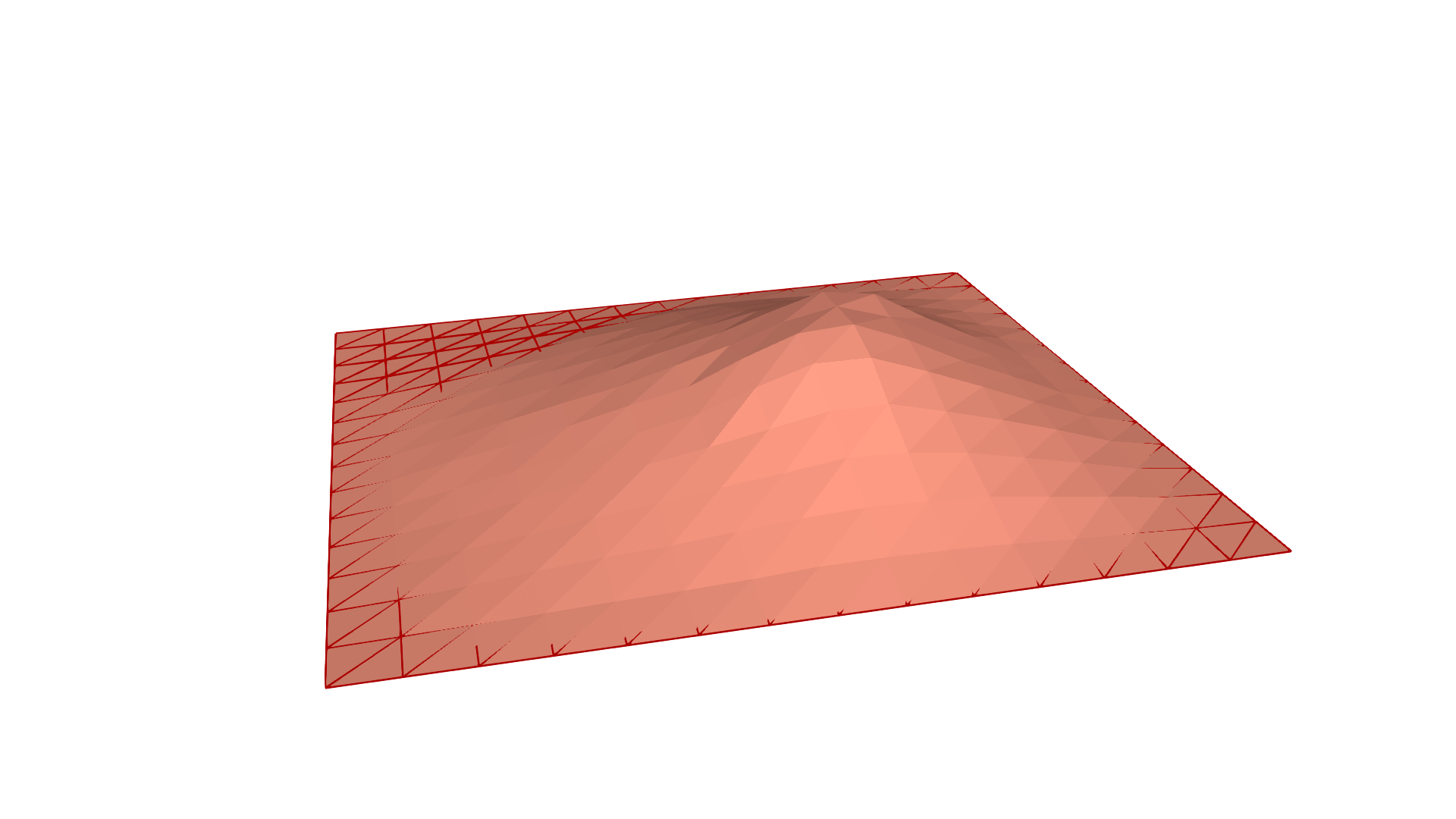}%
    \img{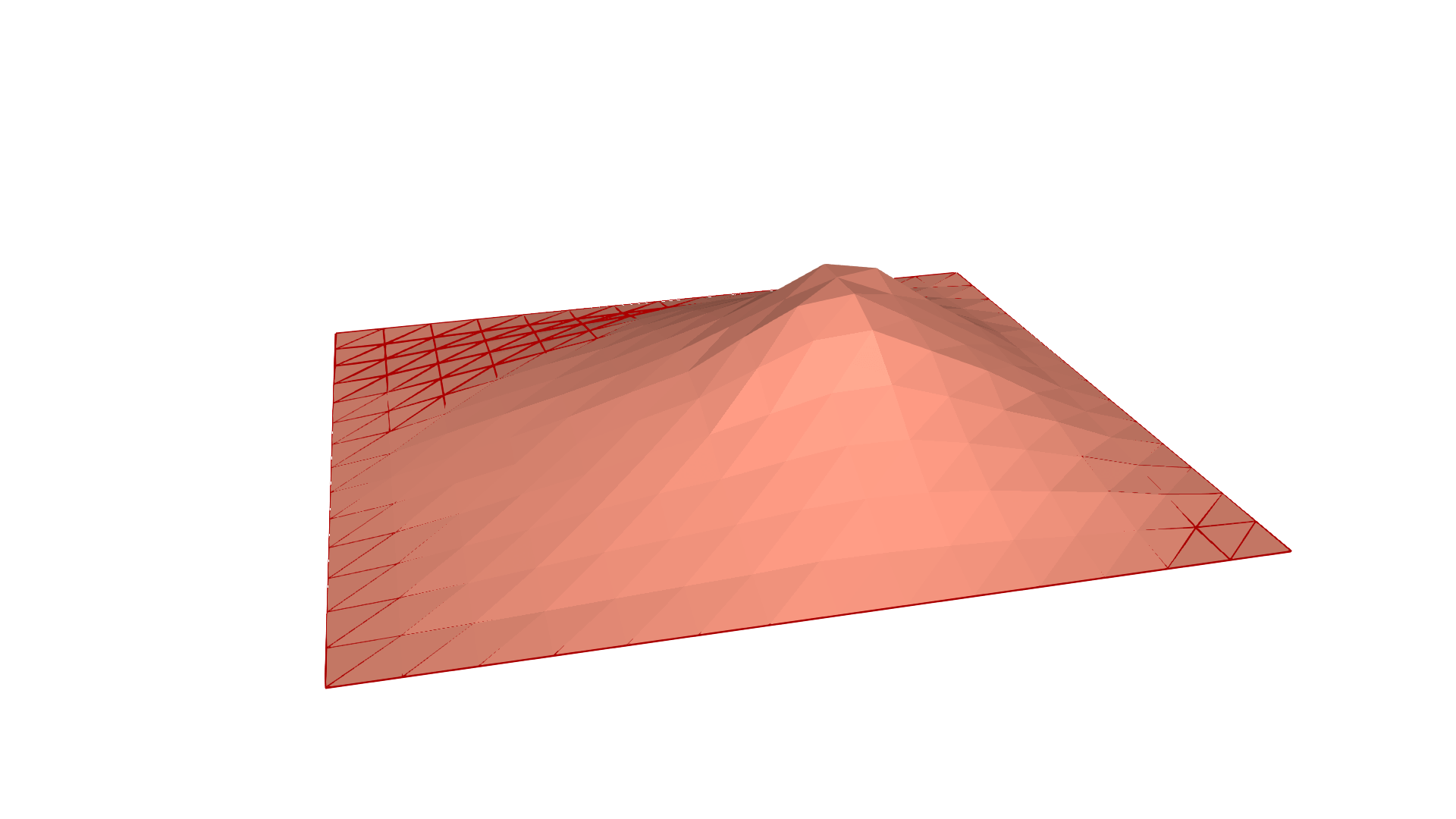}%
    \img{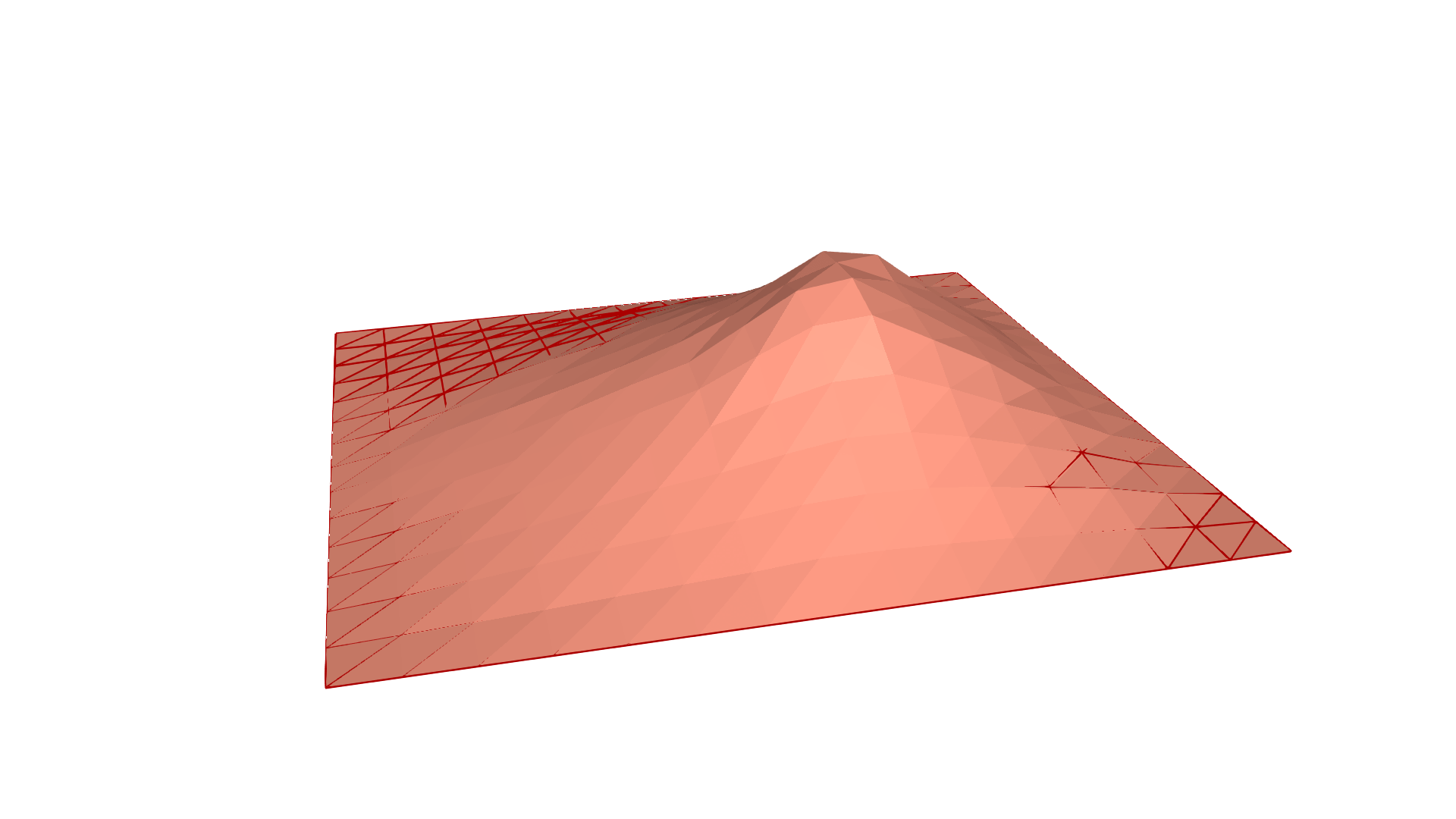}%
    \img{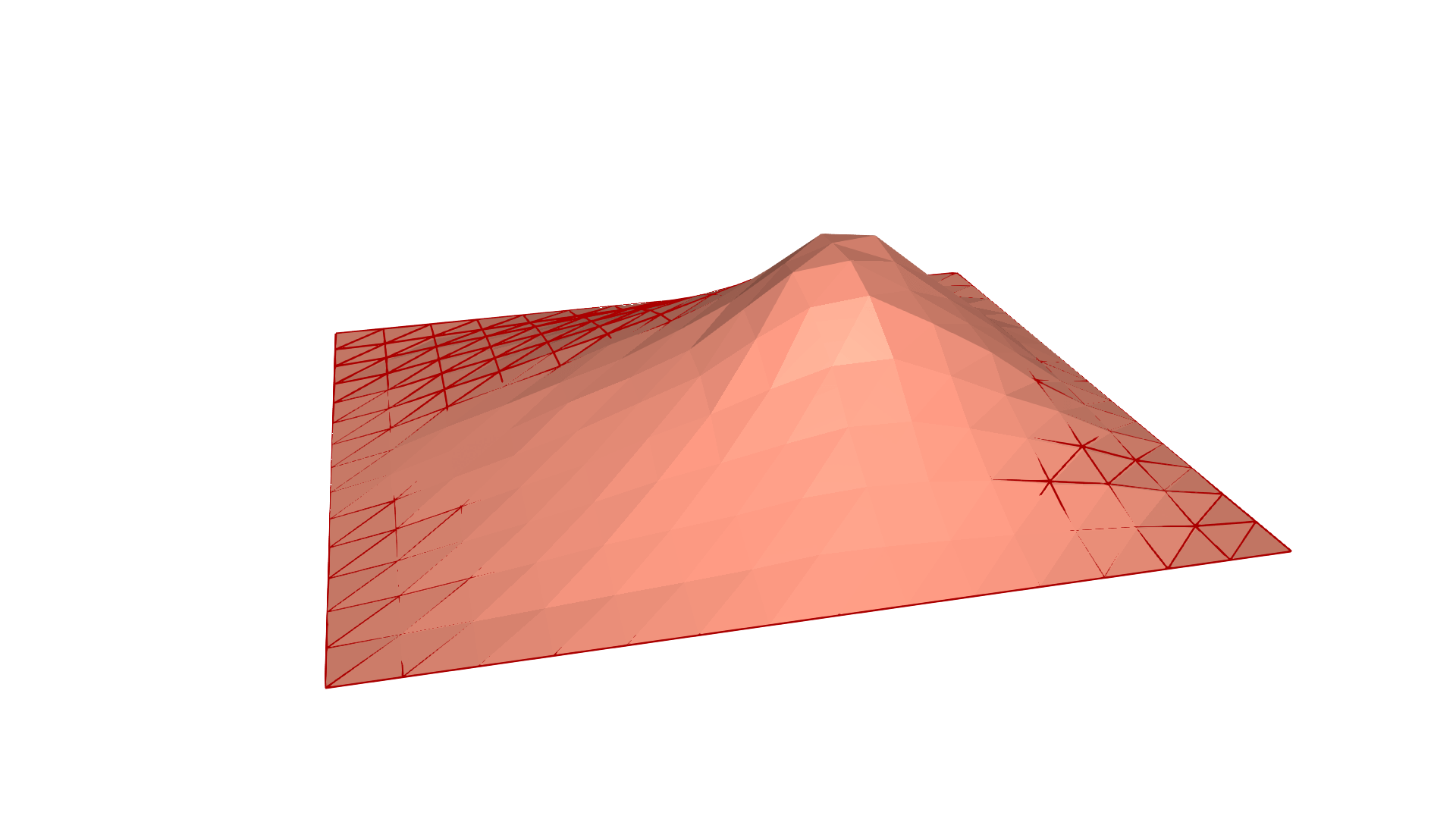}%
    \img{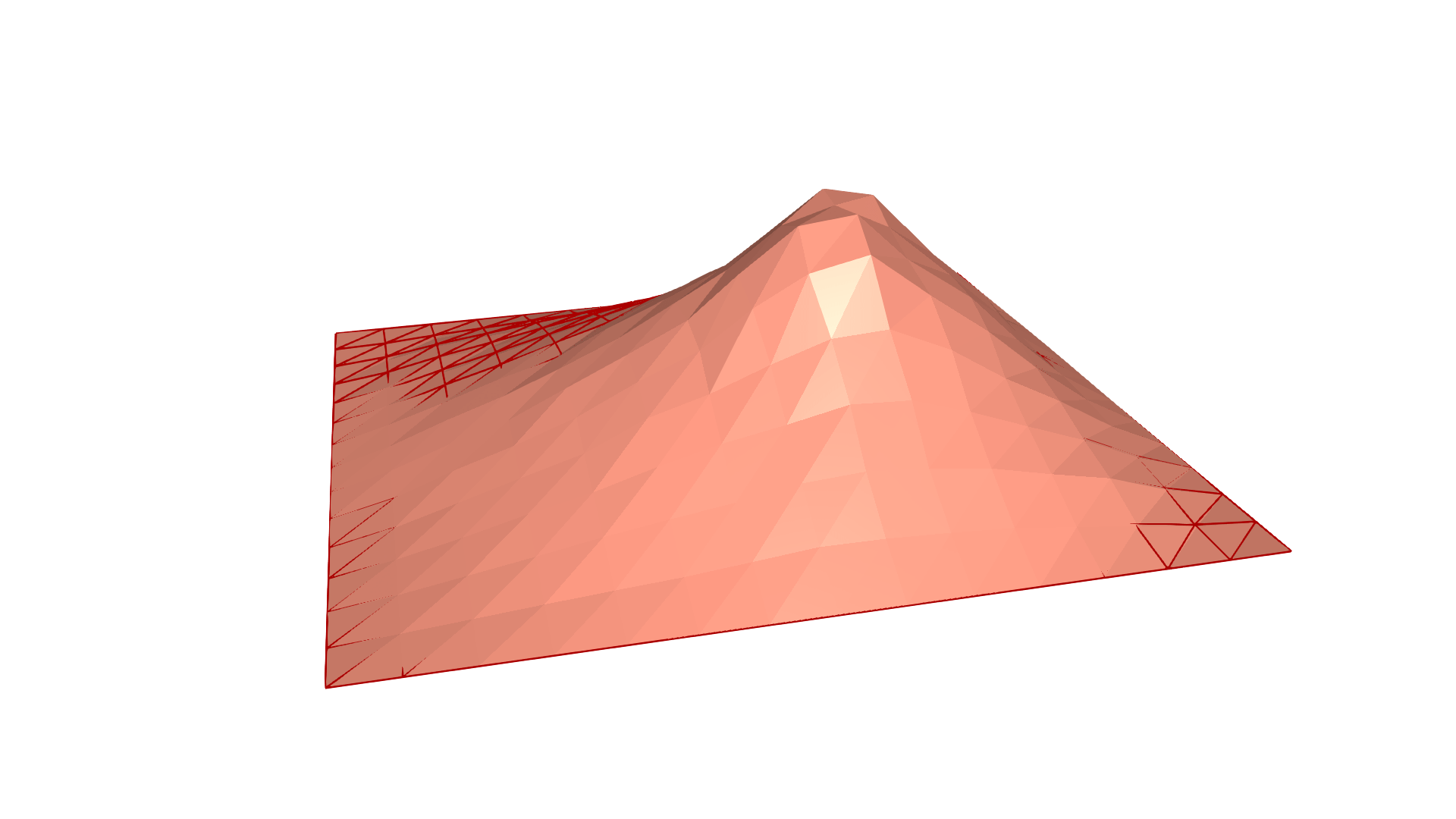}\\[0.4em]

    % Row 8: PSTNet Encoder
    \rowlabel{PSTNet \\ Encoder}%
    \img{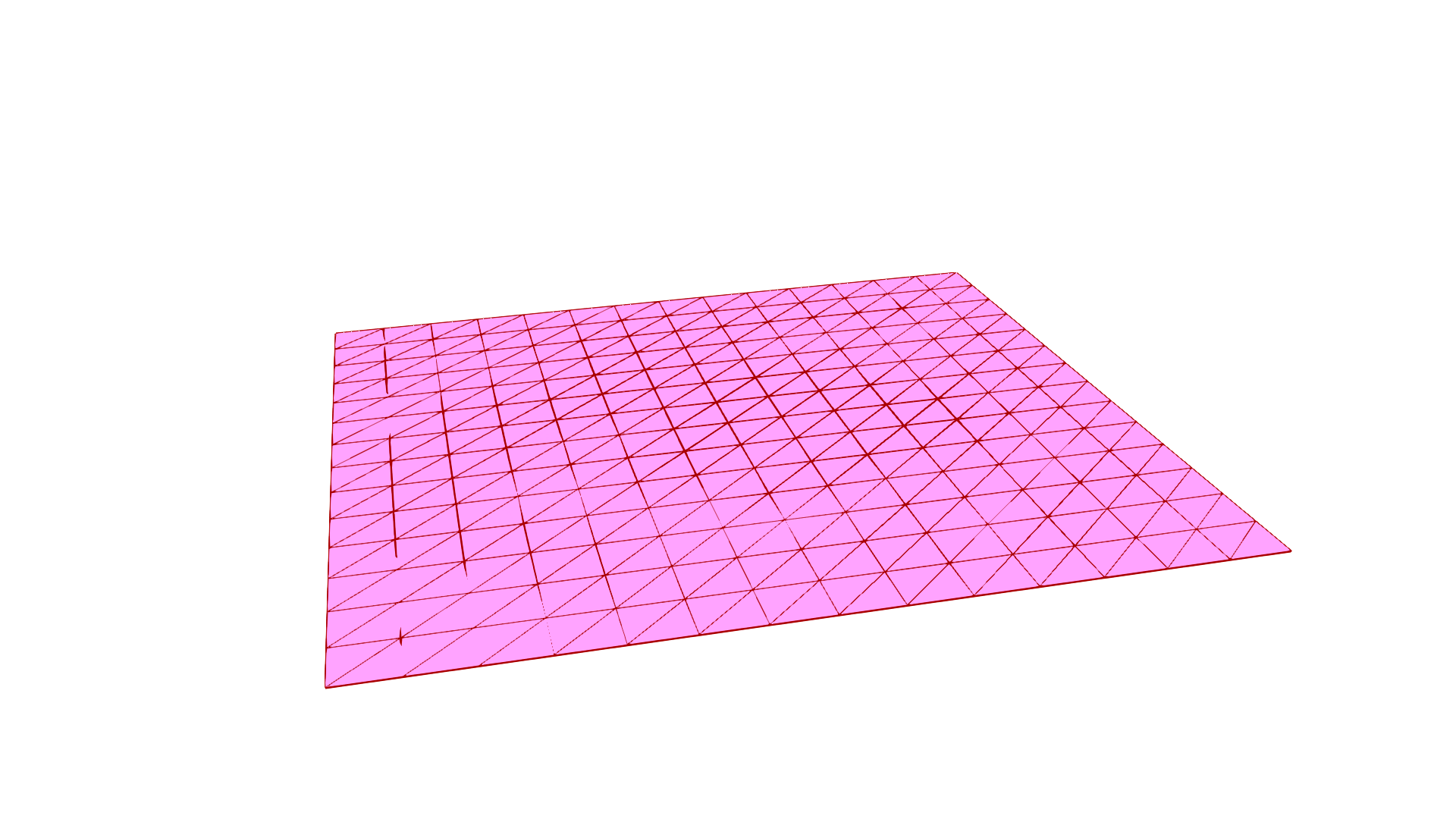}%
    \img{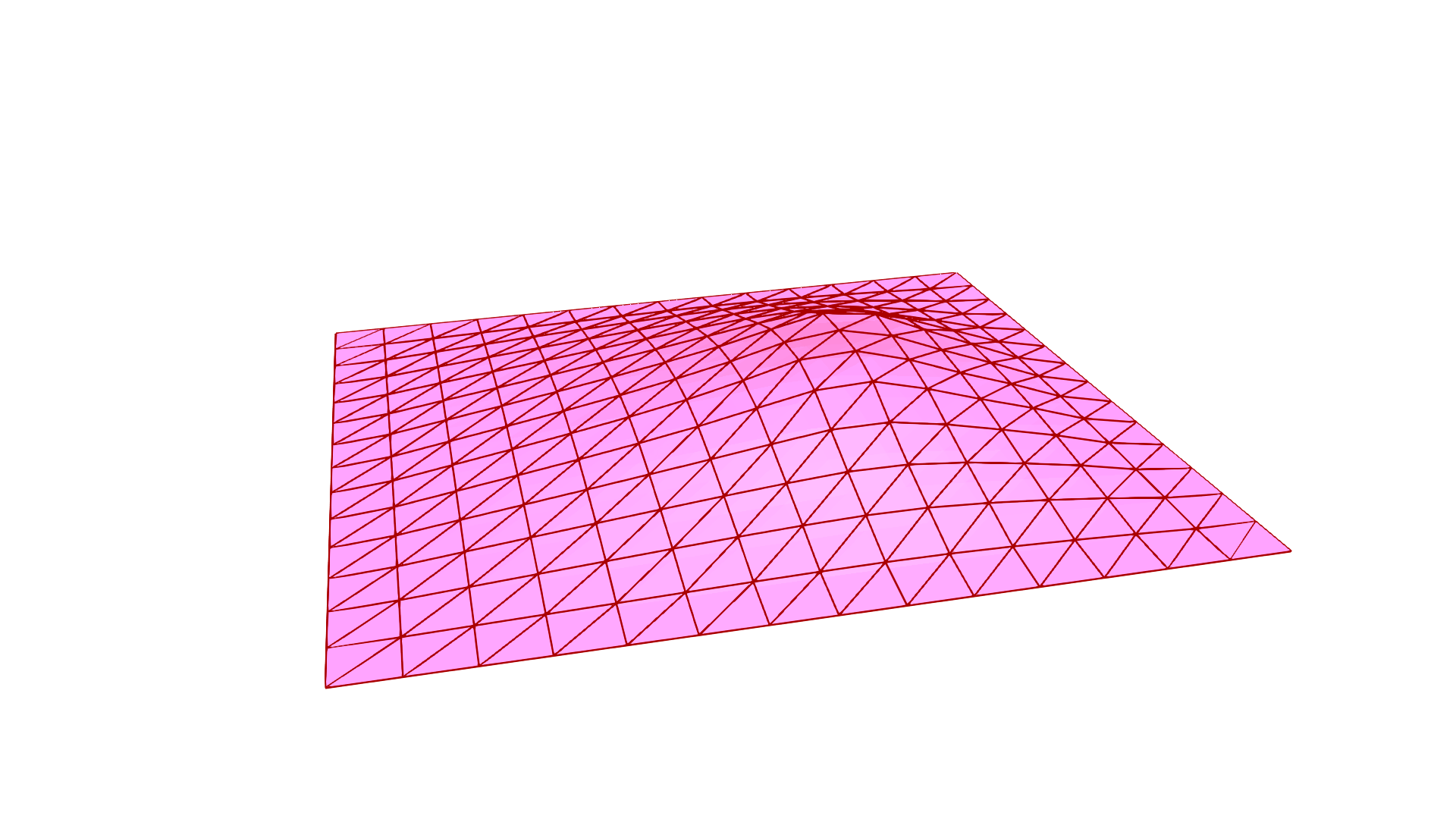}%
    \img{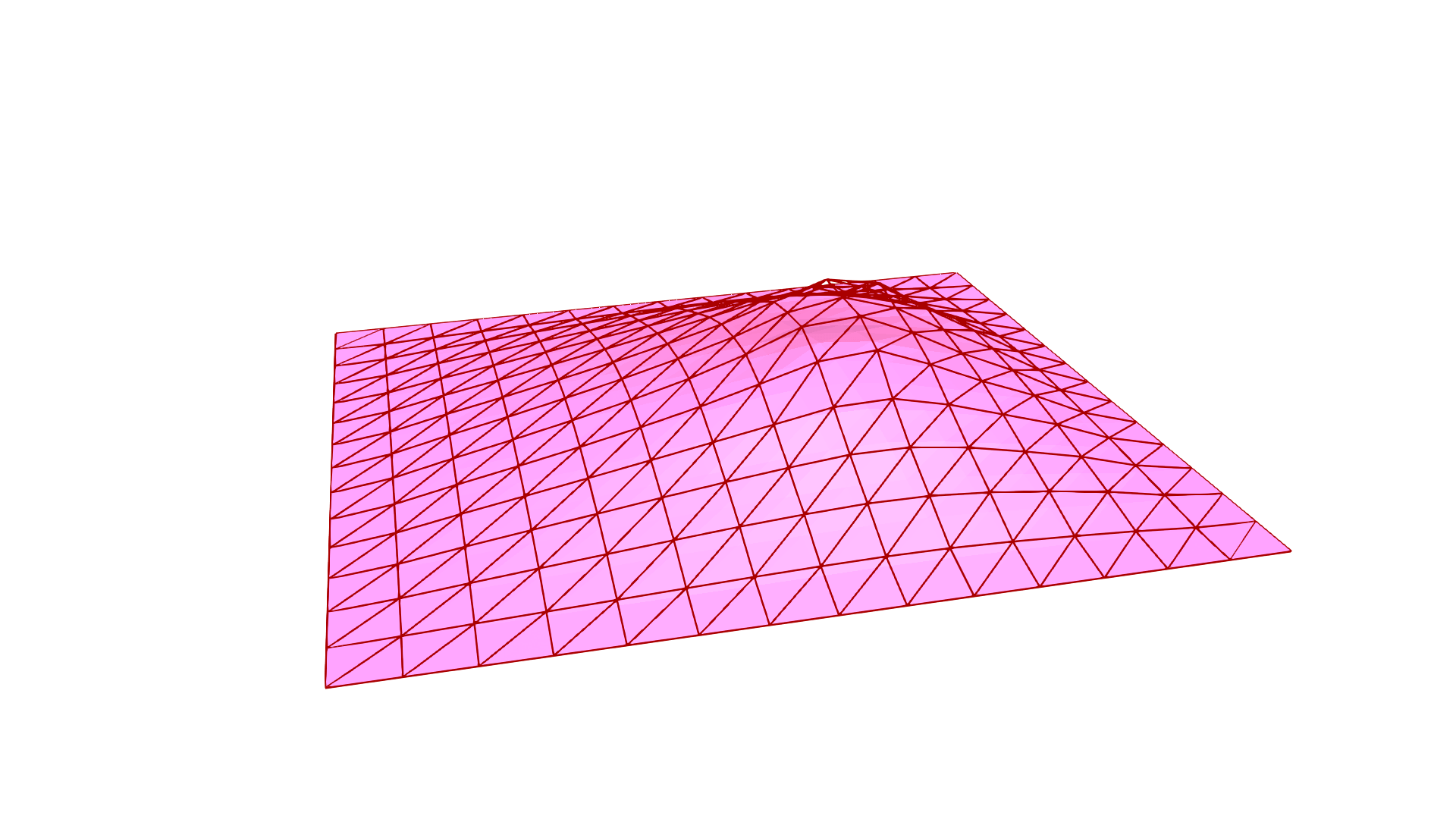}%
    \img{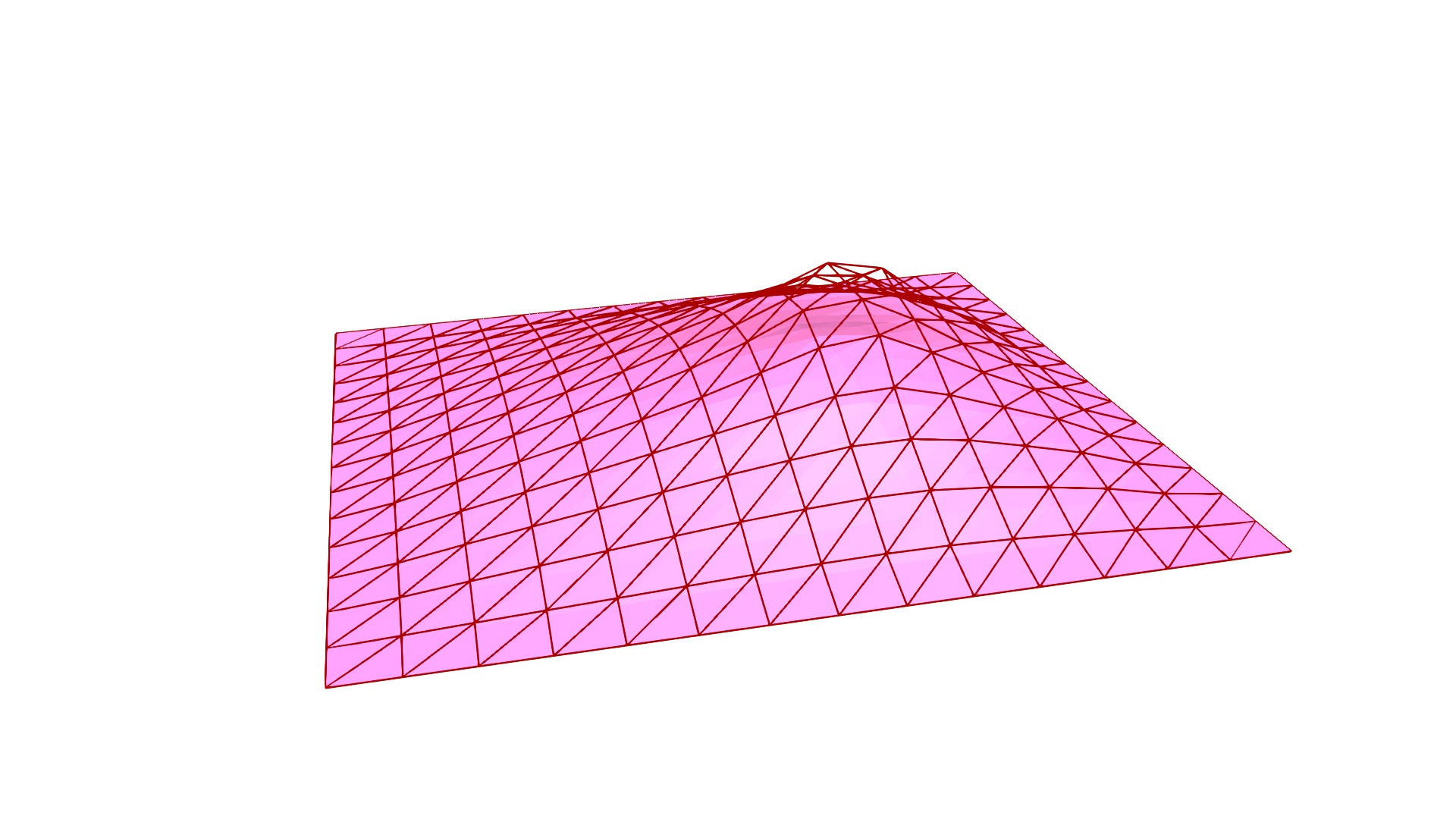}%
    \img{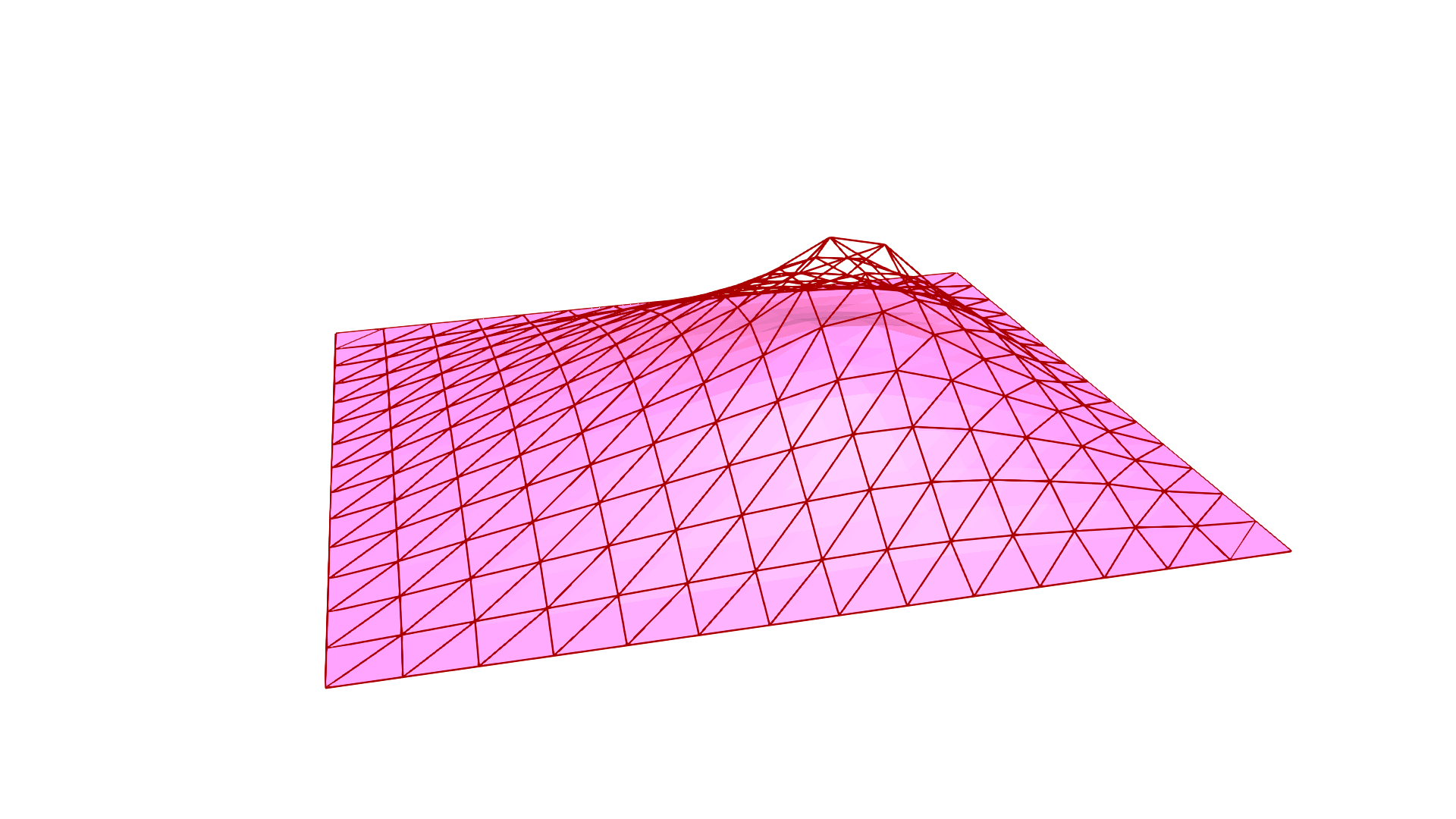}%
    \img{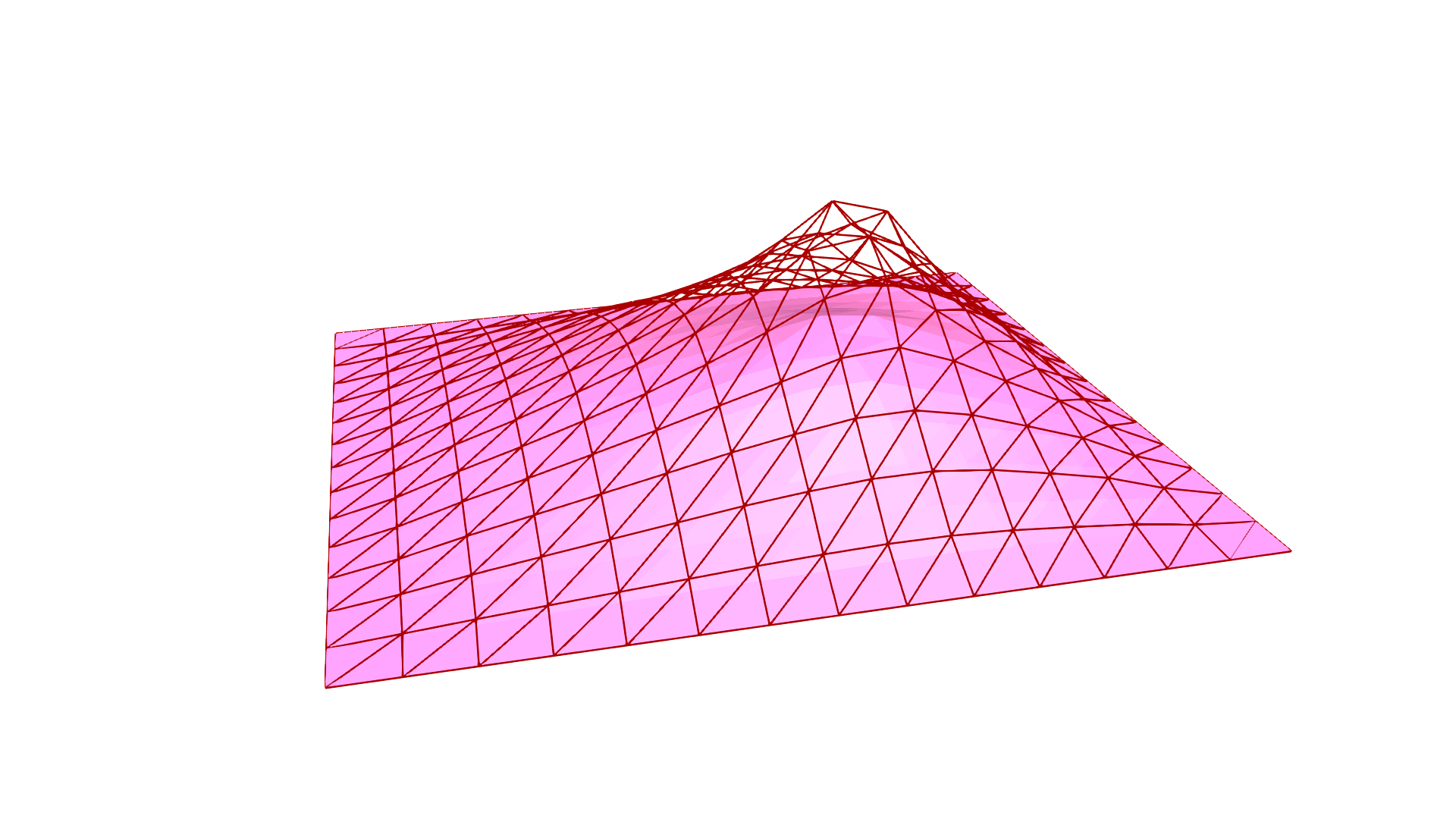}\\[0.4em]

    % Row 9: Pointcloud (same crop + labels)
    \rowlabel{Pointcloud}%
    \begin{minipage}[c]{0.158\textwidth}\centering
        \pointcloud{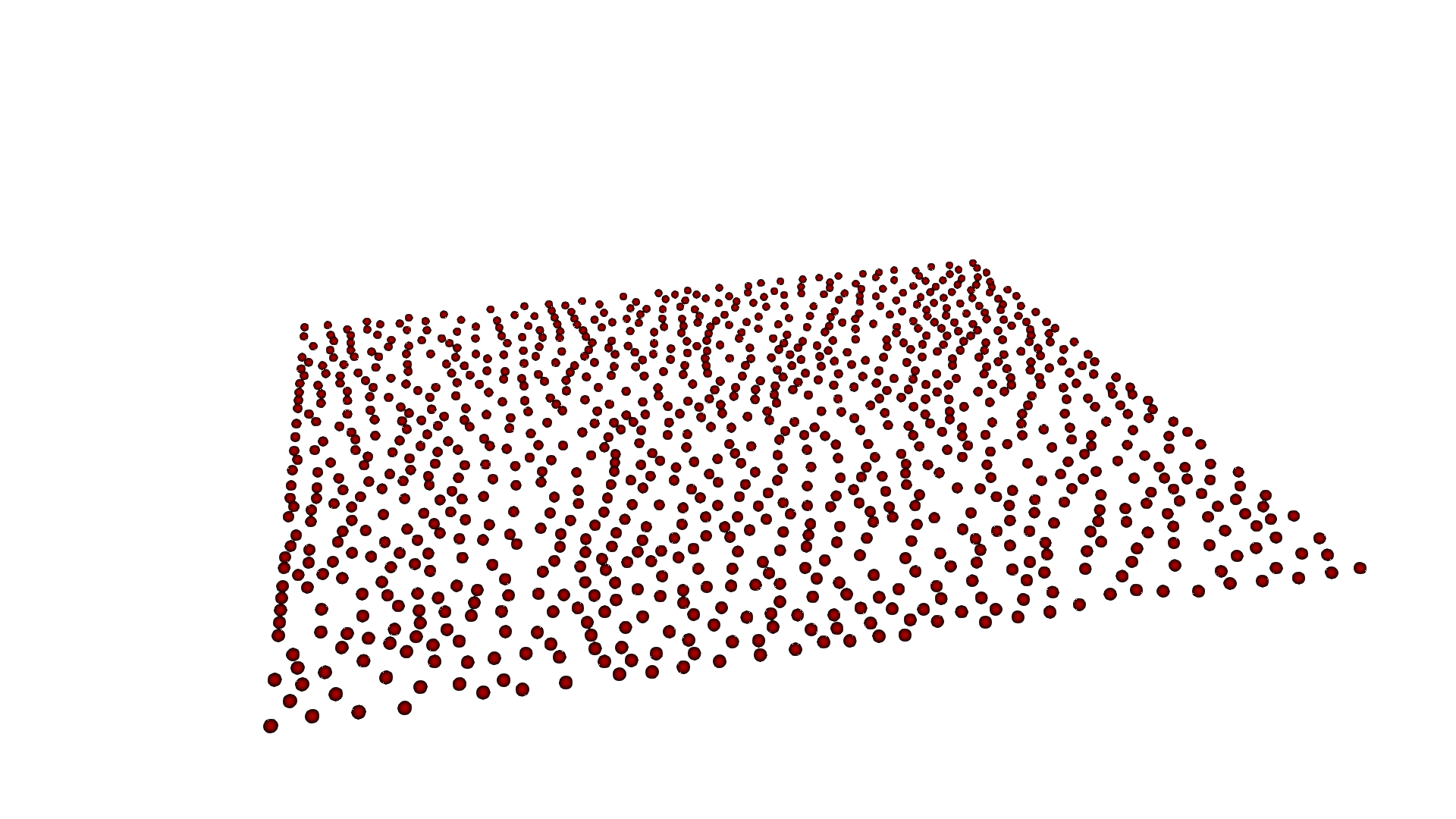}\\[2pt]
        \footnotesize$t{=}0$
    \end{minipage}%
    \begin{minipage}[c]{0.158\textwidth}\centering
        \pointcloud{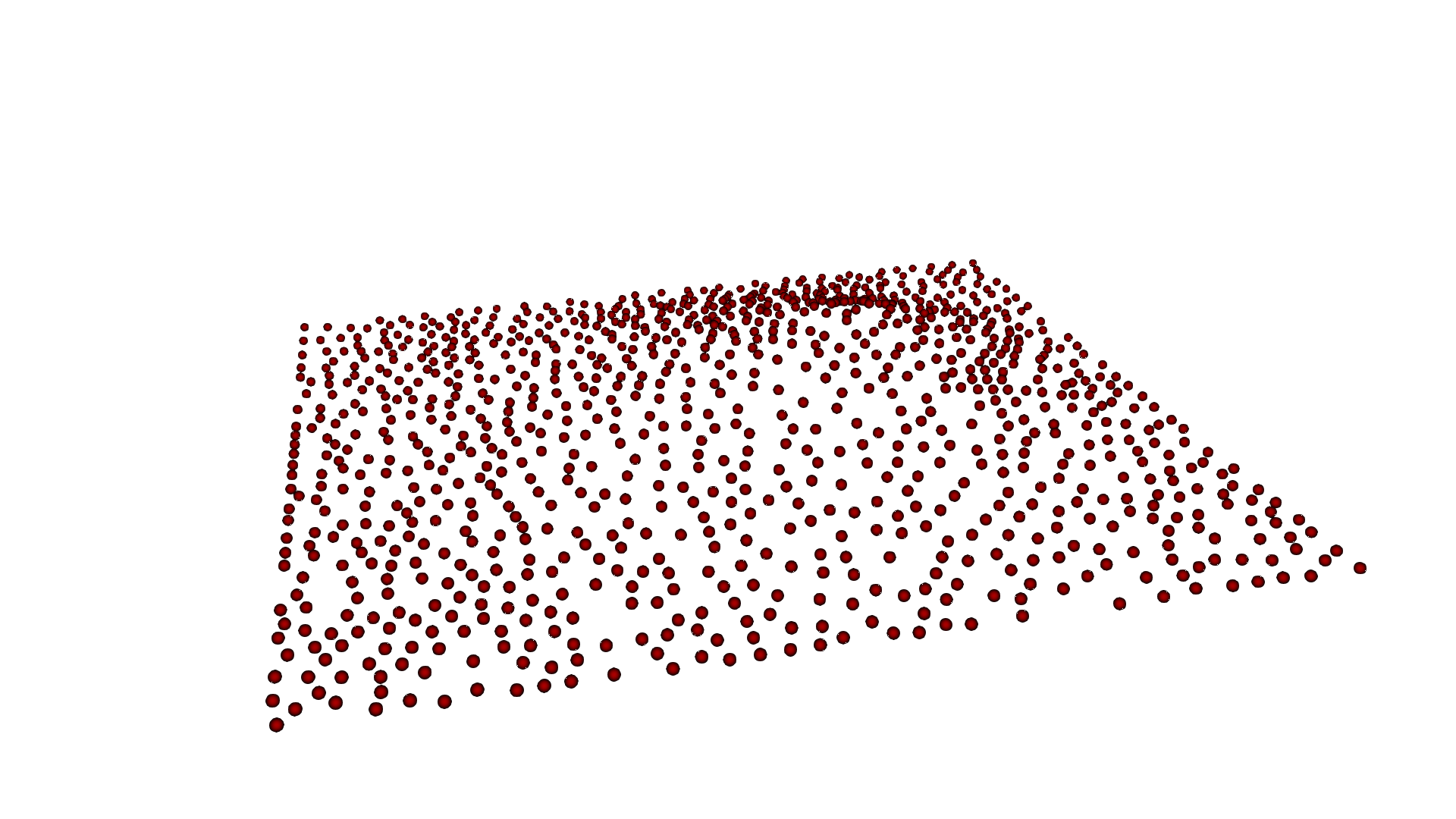}\\[2pt]
        \footnotesize$t{=}6$
    \end{minipage}%
    \begin{minipage}[c]{0.158\textwidth}\centering
        \pointcloud{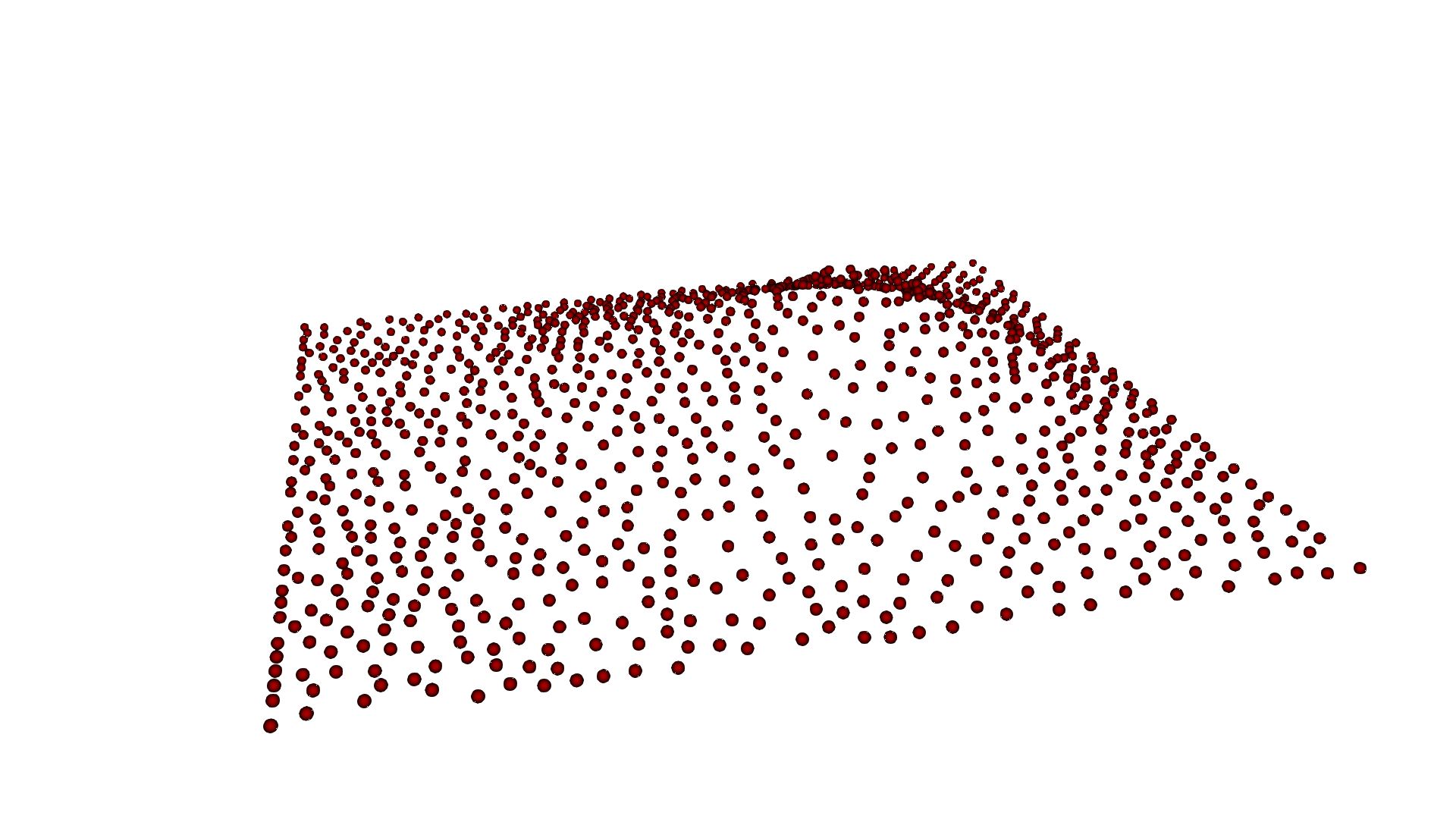}\\[2pt]
        \footnotesize$t{=}15$
    \end{minipage}%
    \begin{minipage}[c]{0.158\textwidth}\centering
        \pointcloud{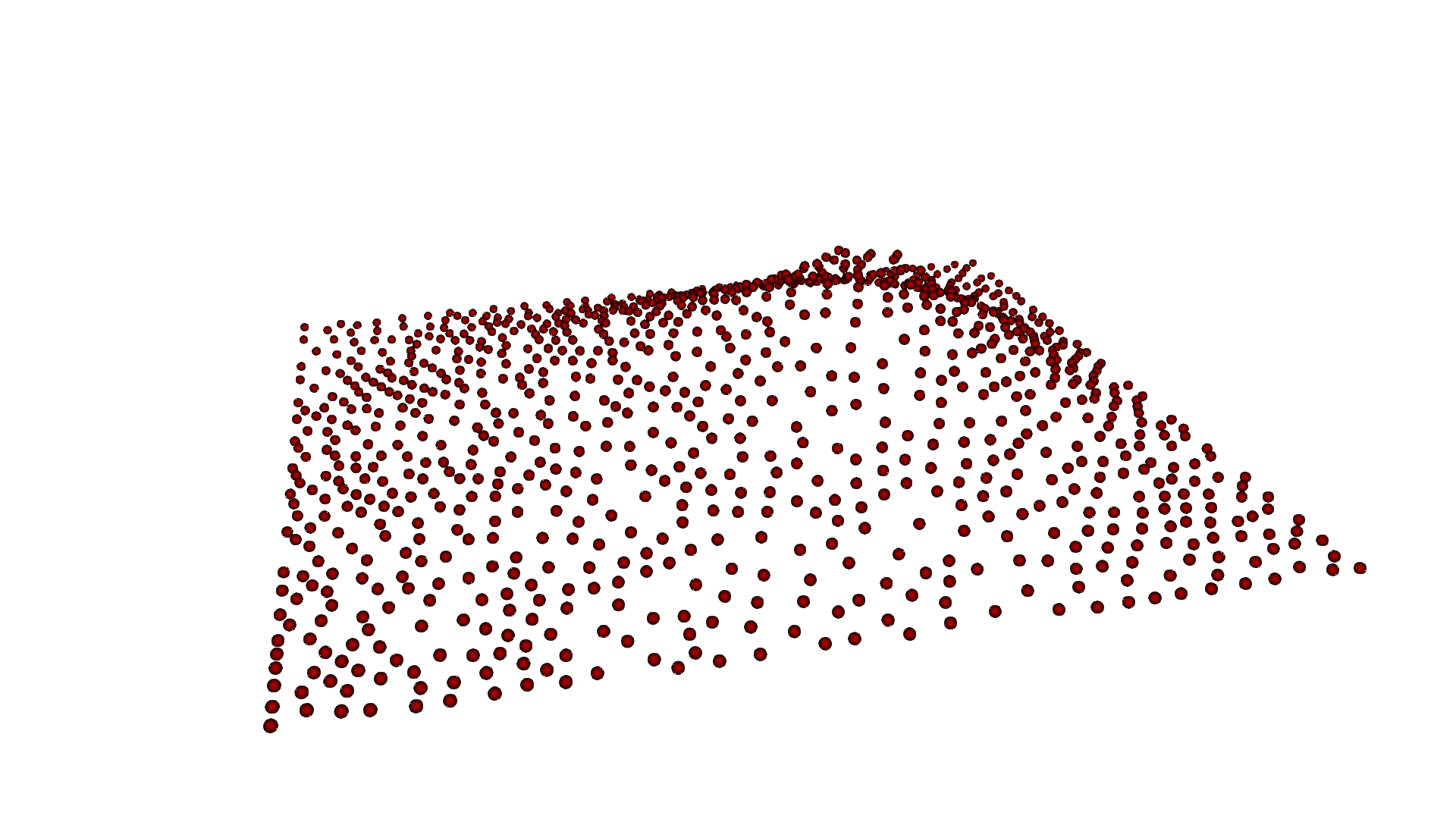}\\[2pt]
        \footnotesize$t{=}21$
    \end{minipage}%
    \begin{minipage}[c]{0.158\textwidth}\centering
        \pointcloud{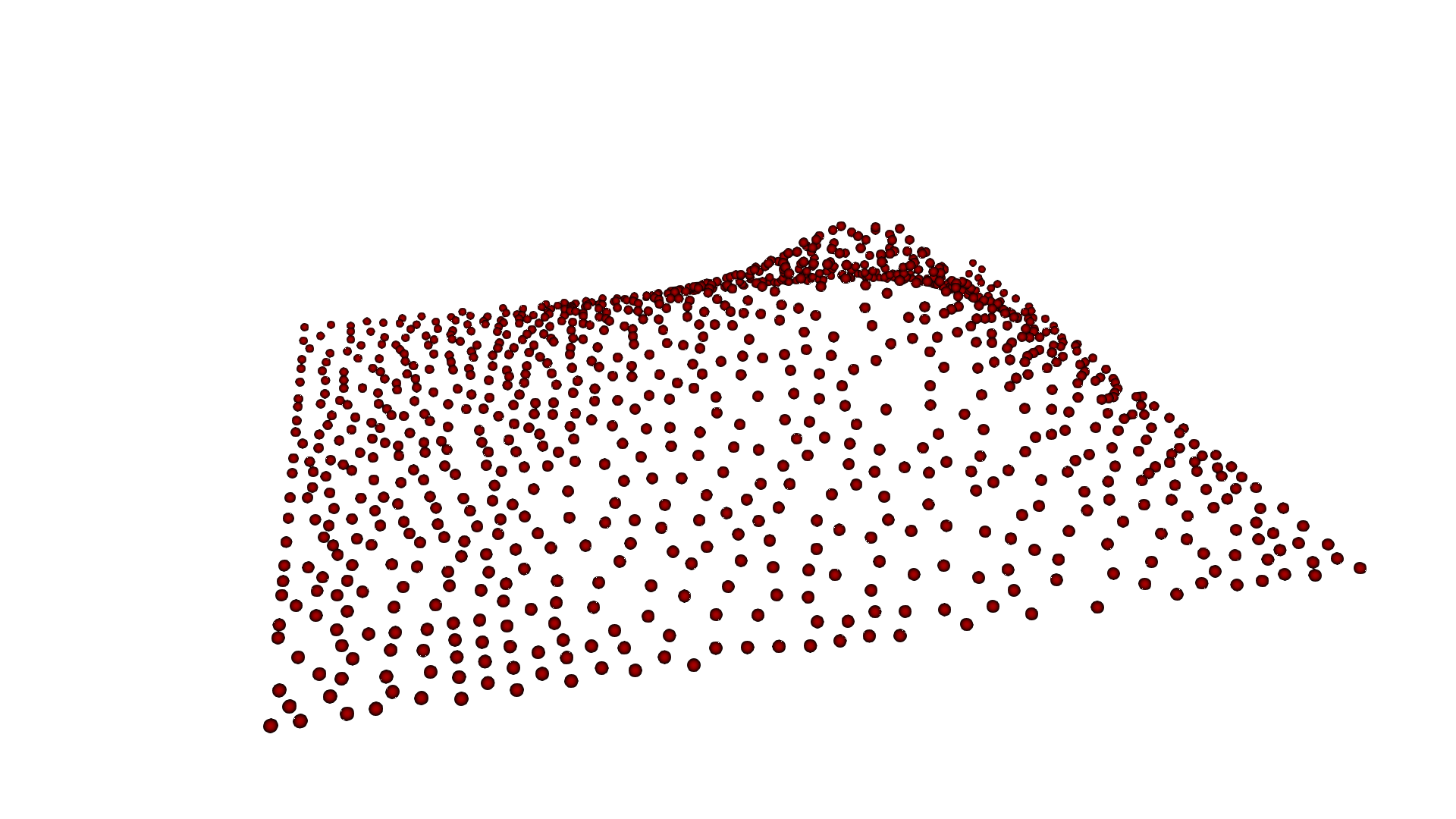}\\[2pt]
        \footnotesize$t{=}32$
    \end{minipage}%
    \begin{minipage}[c]{0.158\textwidth}\centering
        \pointcloud{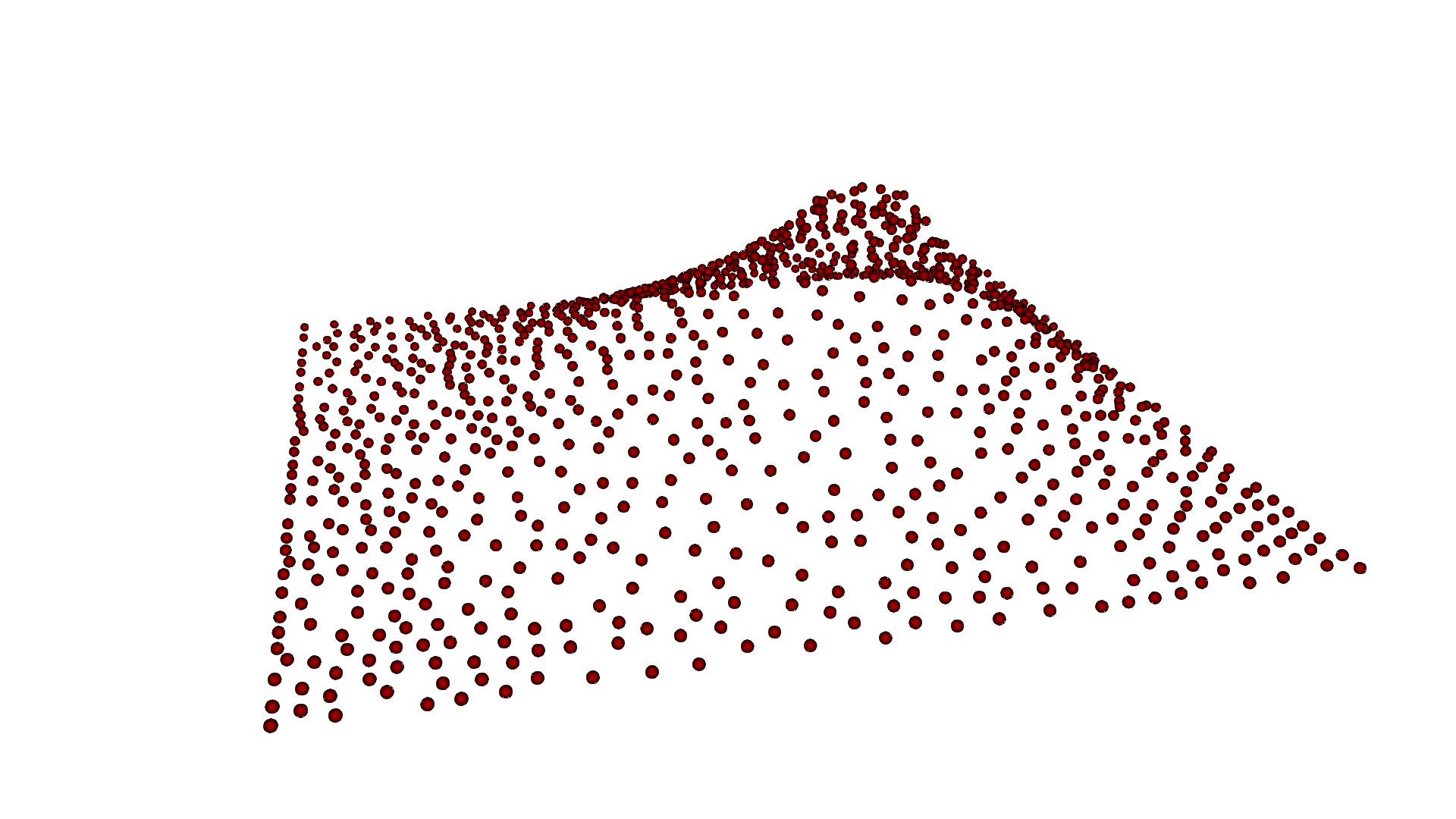}\\[2pt]
        \footnotesize$t{=}50$
    \end{minipage}

    \par\vfill
    \caption{
Predicted simulation of a \texttt{Sheet Deformation} test task by
\textcolor{TabBlue}{PEACH} (blue),
\textcolor{TabGray}{No Context} (gray),
\textcolor{TabCyan}{MANGO} (cyan),
\textcolor{TabGreen}{Oracle} (green),
\textcolor{TabOrange}{No Context (MGN)} (orange),
\textcolor{TabPurple}{Oracle (MGN)} (purple),
\textcolor{TabBrown}{GNN Encoder} (brown), and
\textcolor{TabPink}{PSTNet Encoder} (pink).
All visualizations show the colored \textbf{predicted mesh} and a \textbf{\textcolor{red}{wireframe}} (red) of the ground-truth simulation. The last row shows an exemplary point cloud sequence from the context set.
    }
    \label{fig:qualitative_trajectories_sd}
\end{figure*}

\begin{figure*}[p]
    \centering
    {\Large \textbf{Trampoline}}\\[0.1em]

    % global cropped image macro
    \newcommand{\rowlabel}[1]{%
        \begin{minipage}[c]{0.05\textwidth}\centering\rotatebox{90}{\scriptsize\textbf{\shortstack{#1}}}\end{minipage}%
    }
    \newcommand{\img}[1]{%
        \includegraphics[width=0.155\textwidth, trim=11cm 0.01cm 11cm 0.01cm, clip, valign=m]{#1}%
    }
    \newcommand{\pointcloud}[1]{%
        \includegraphics[width=\textwidth, trim=11cm 1.5cm 11cm 1.5cm, clip, valign=m]{#1}%
    }

    % Row 1: PEACH
    \rowlabel{PEACH}%
    \img{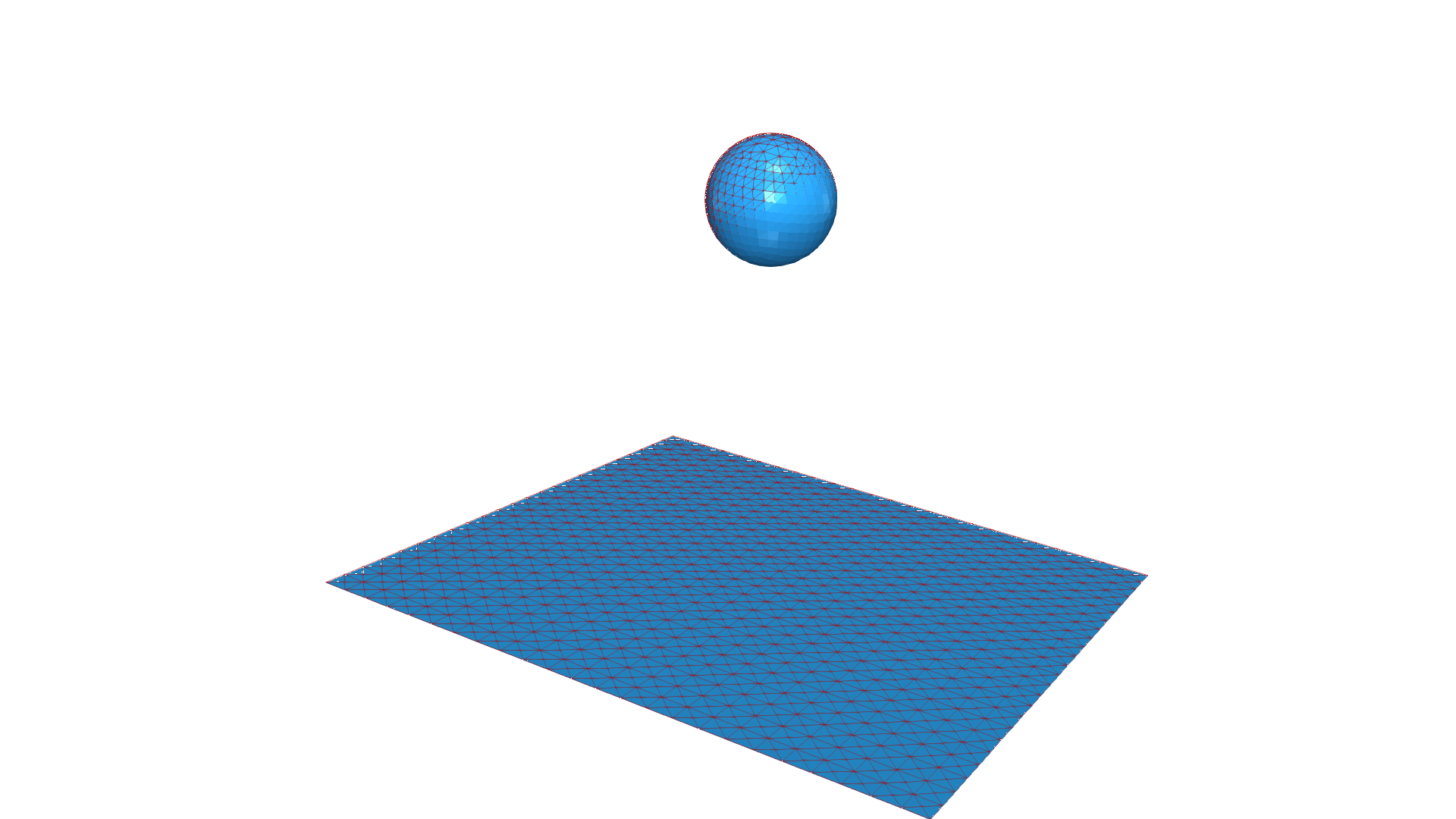}%
    \img{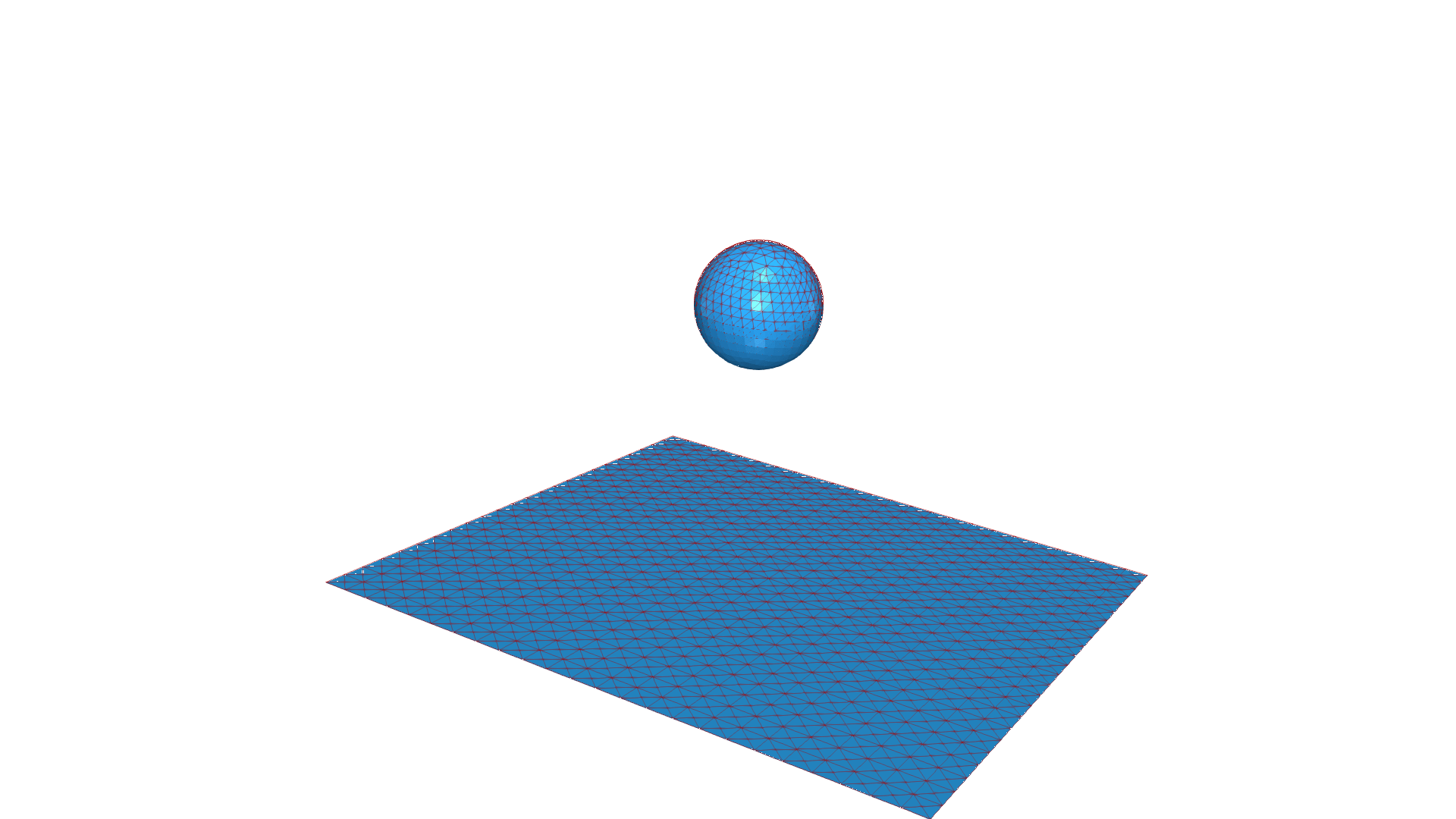}%
    \img{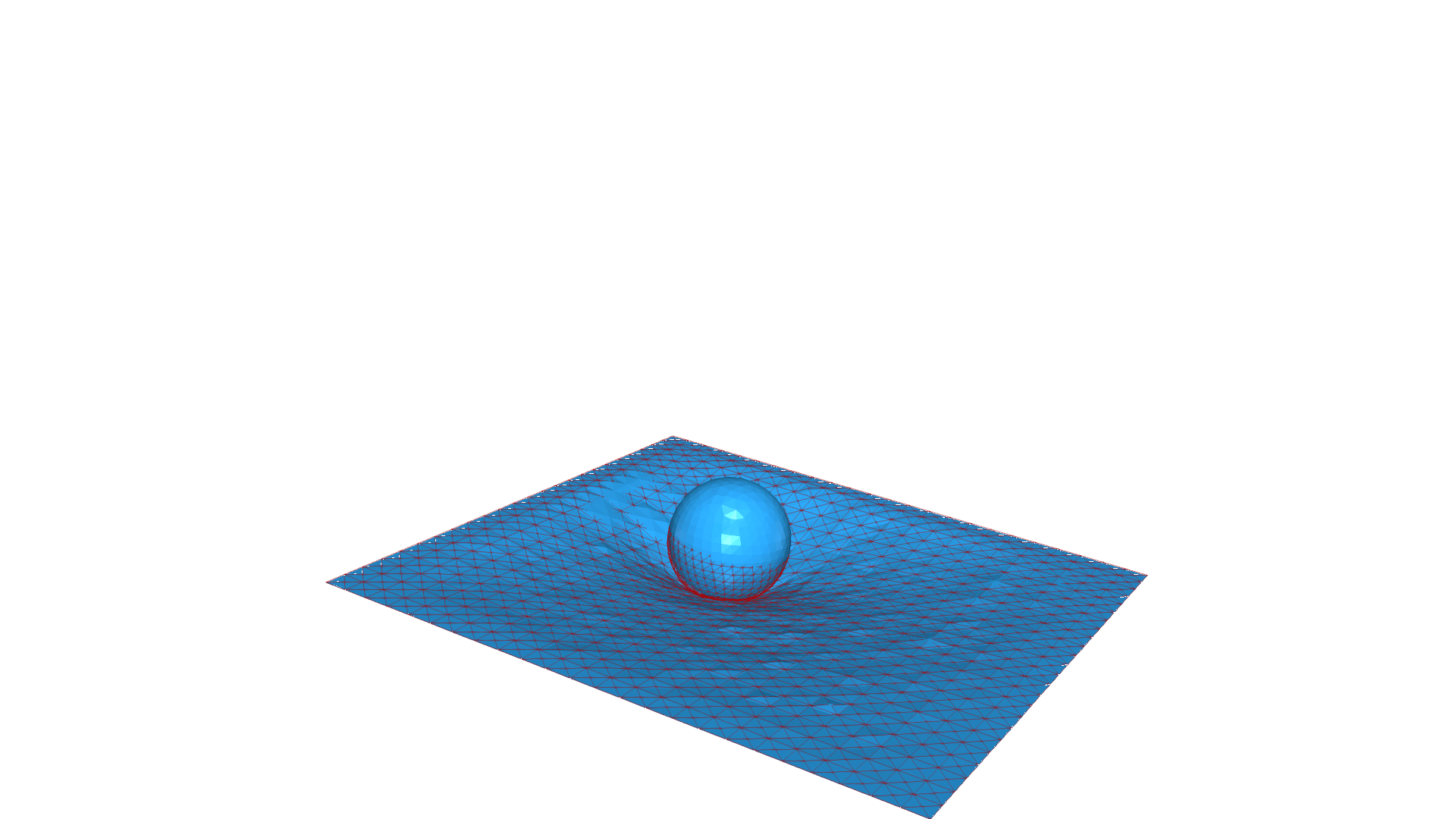}%
    \img{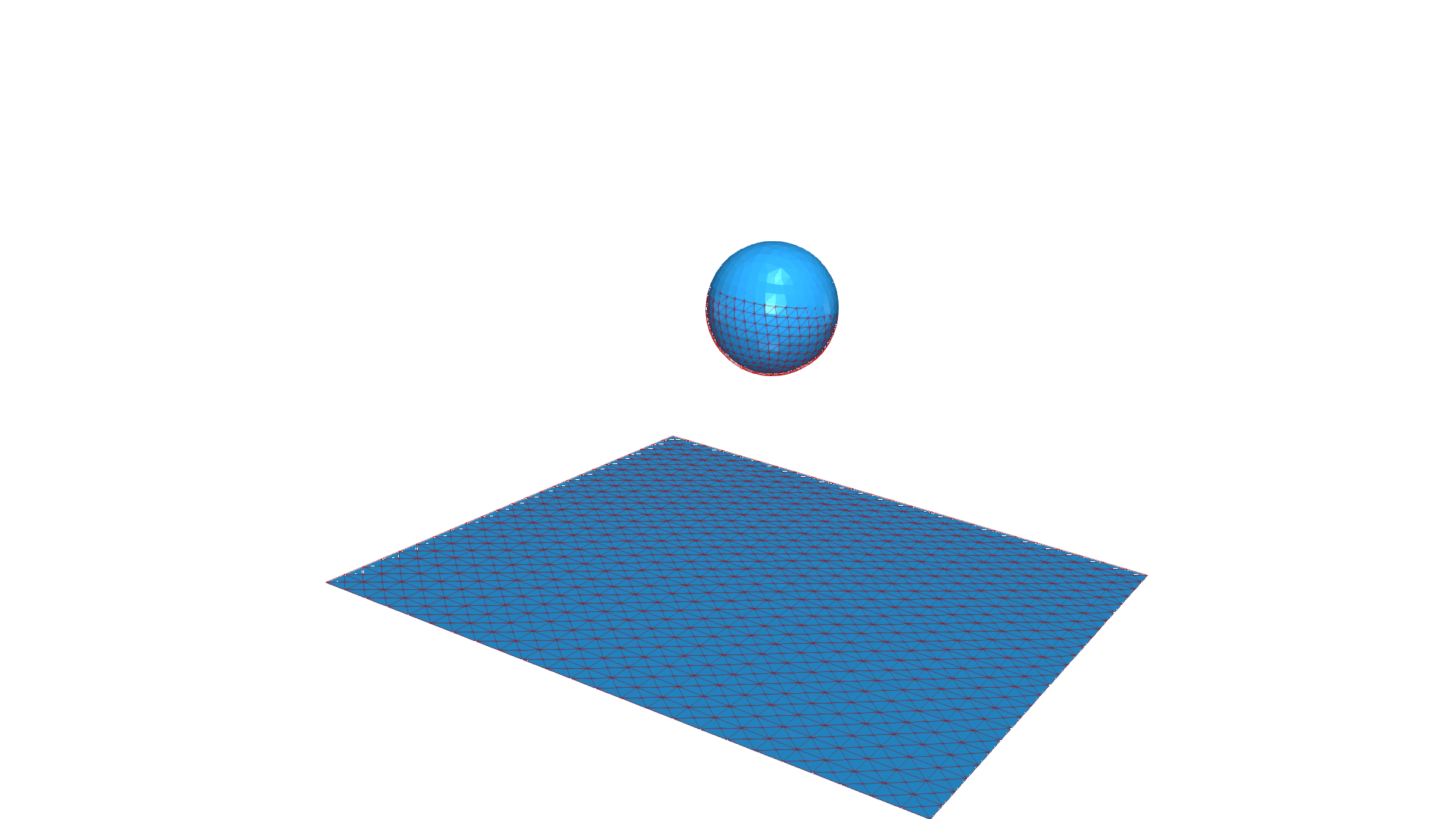}%
    \img{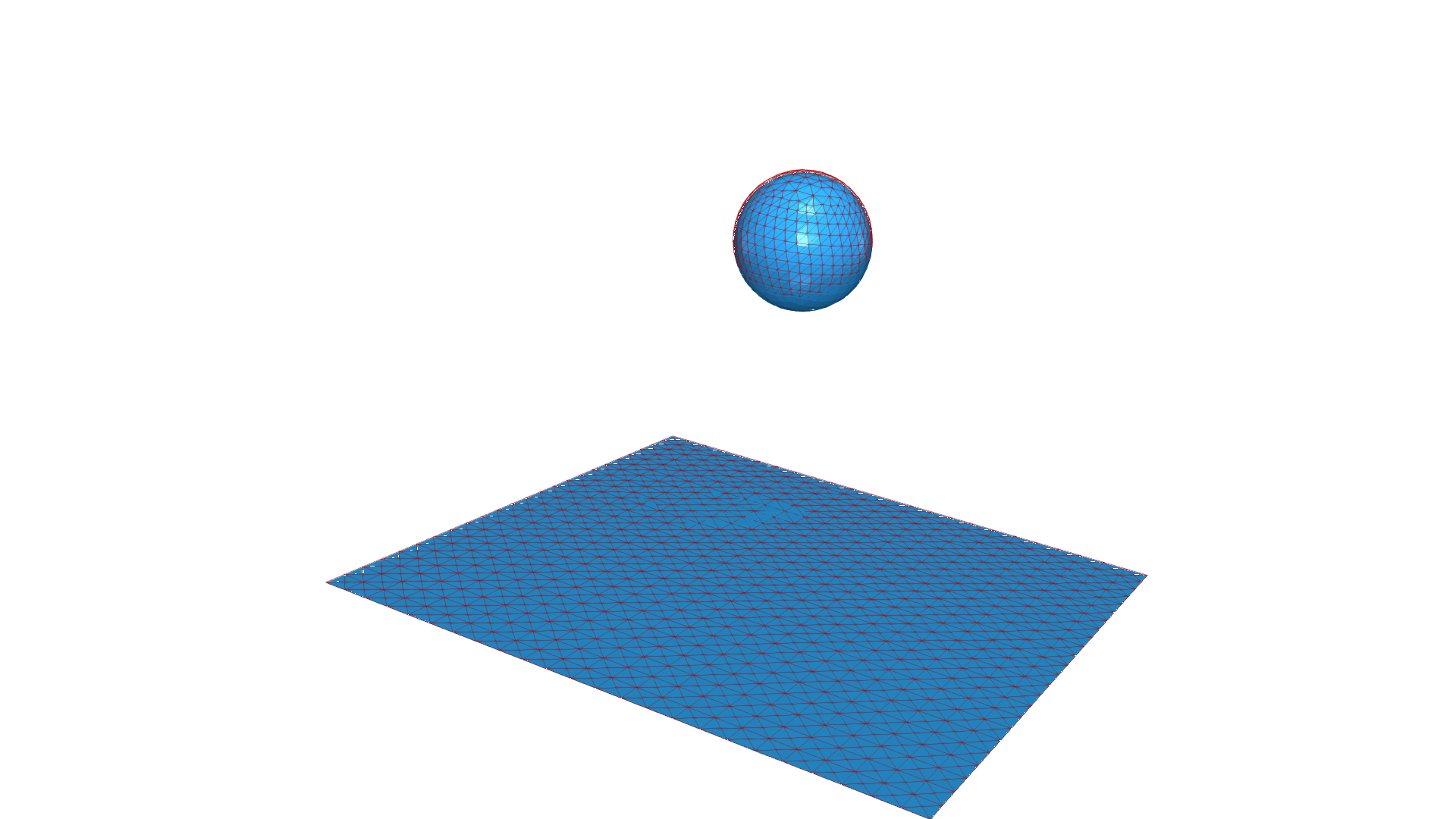}%
    \img{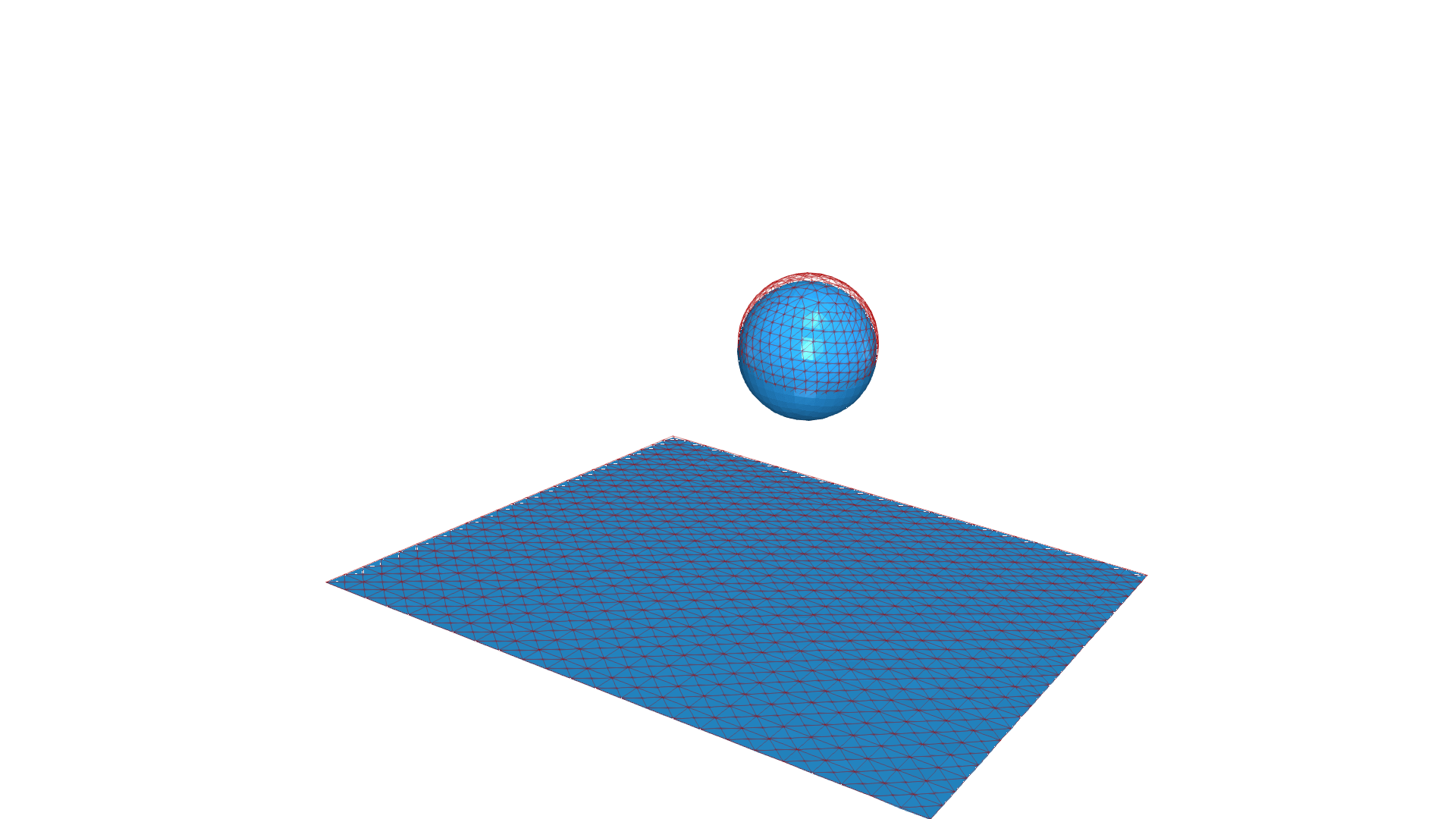}\\[0.15em]

    % Row 2: No Context
    \rowlabel{No Context}%
    \img{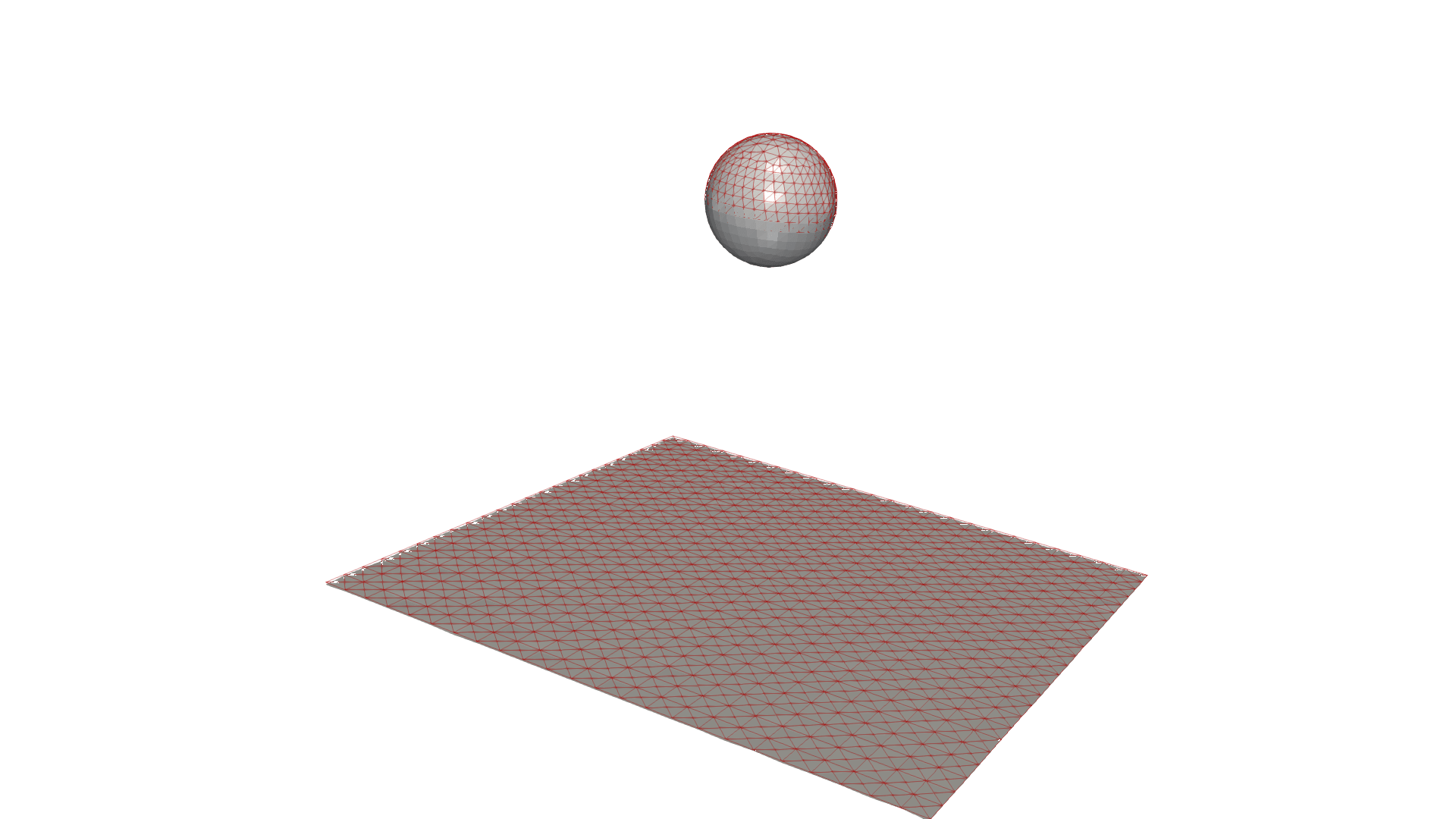}%
    \img{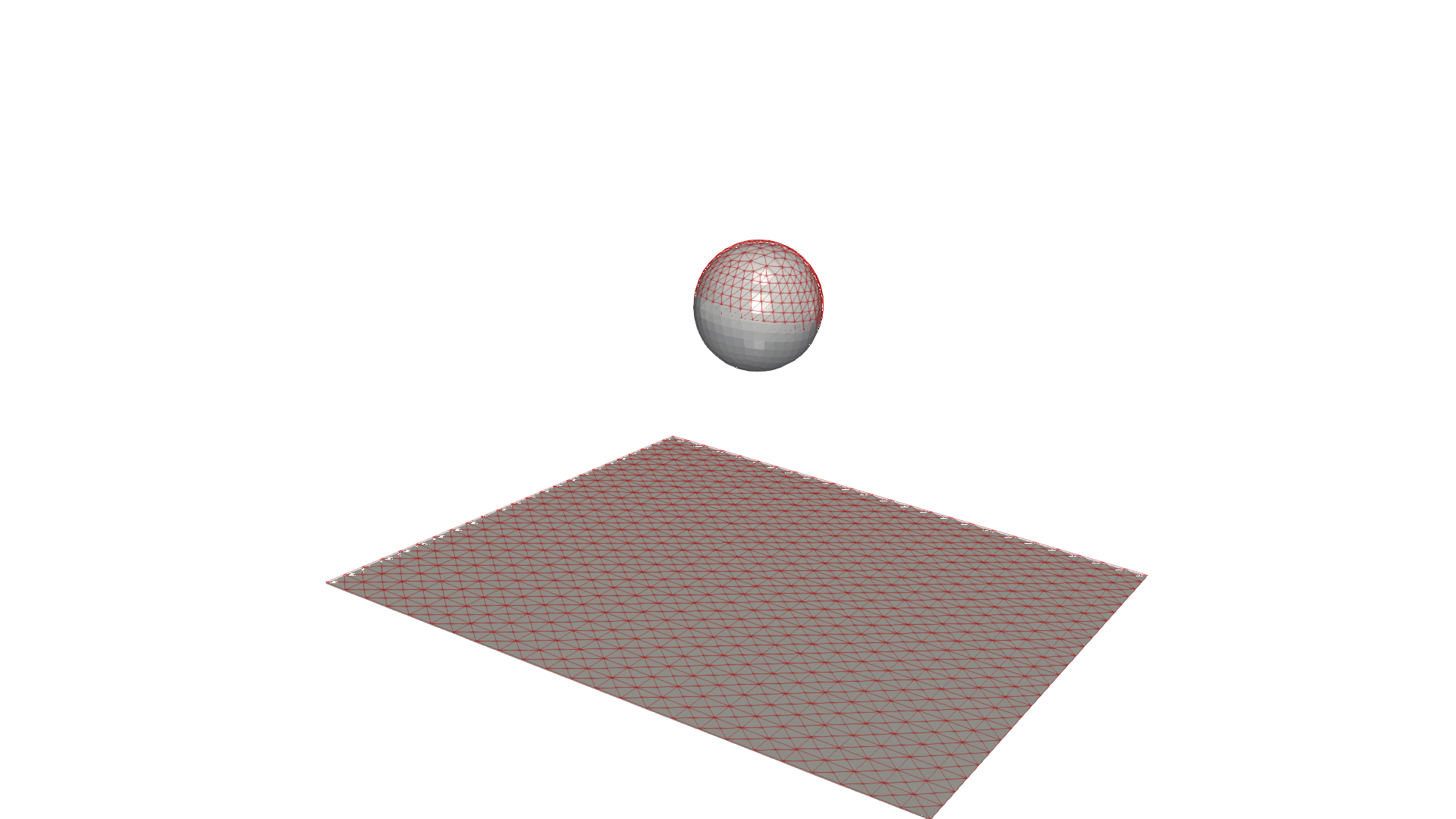}%
    \img{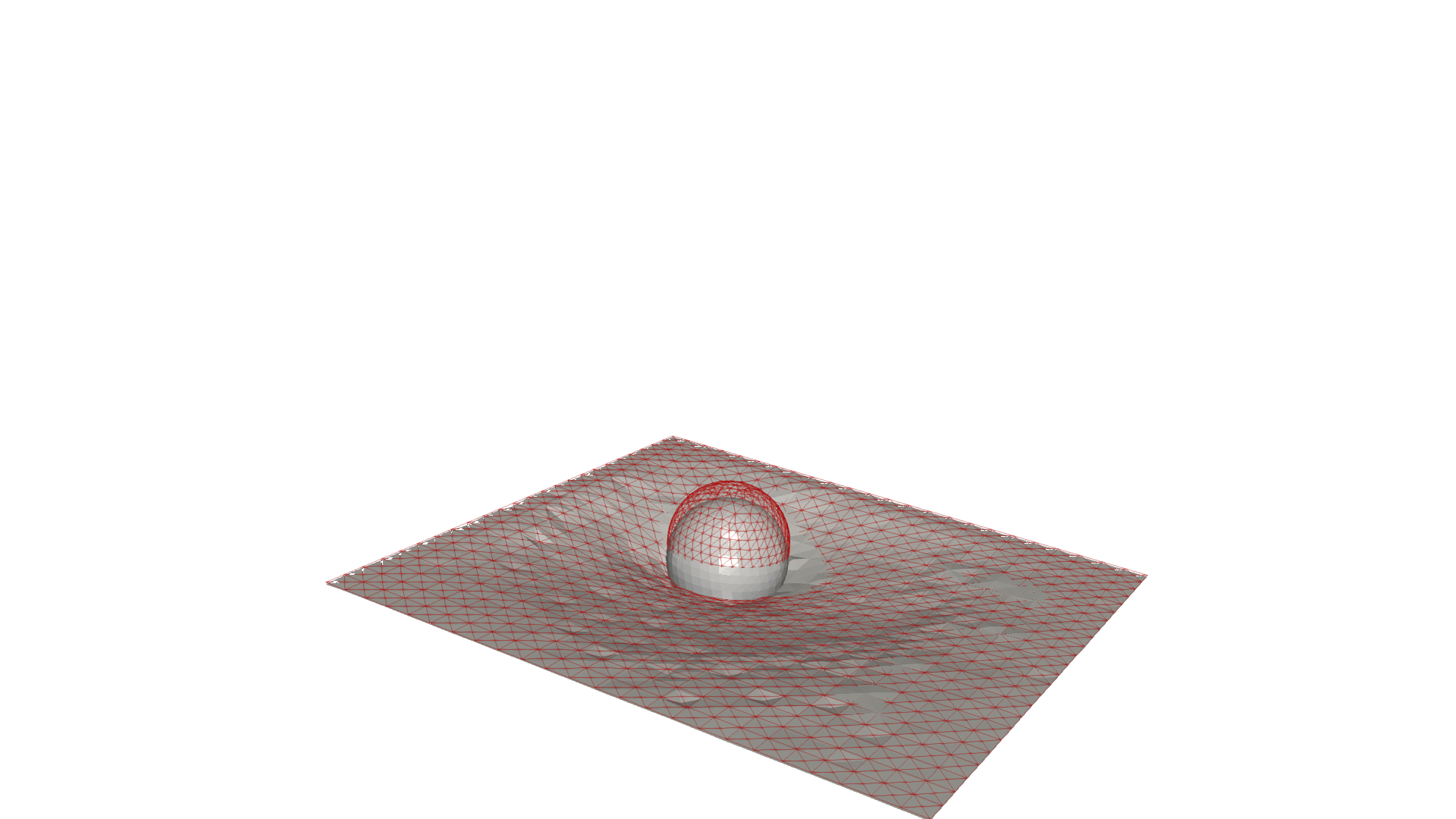}%
    \img{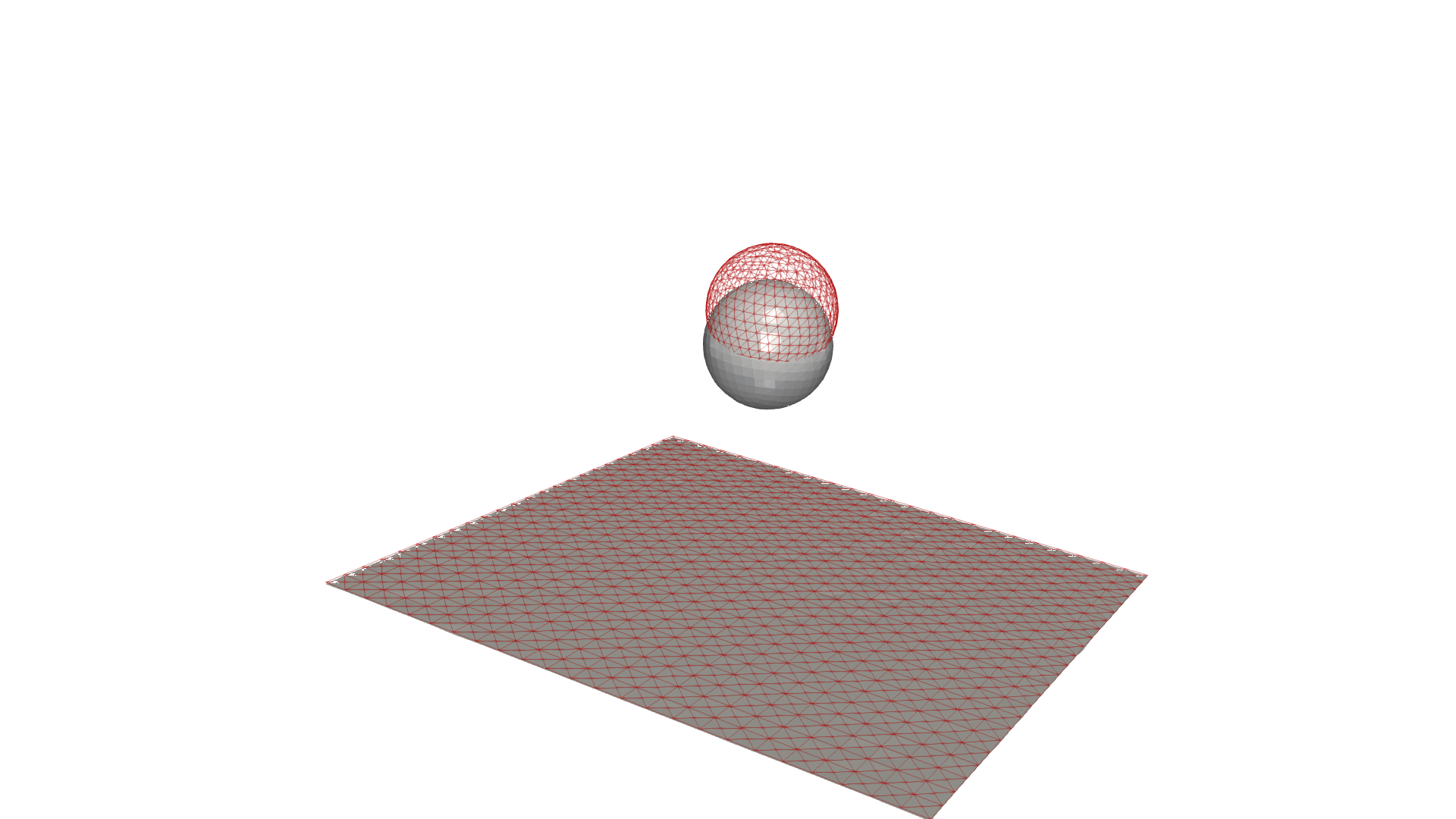}%
    \img{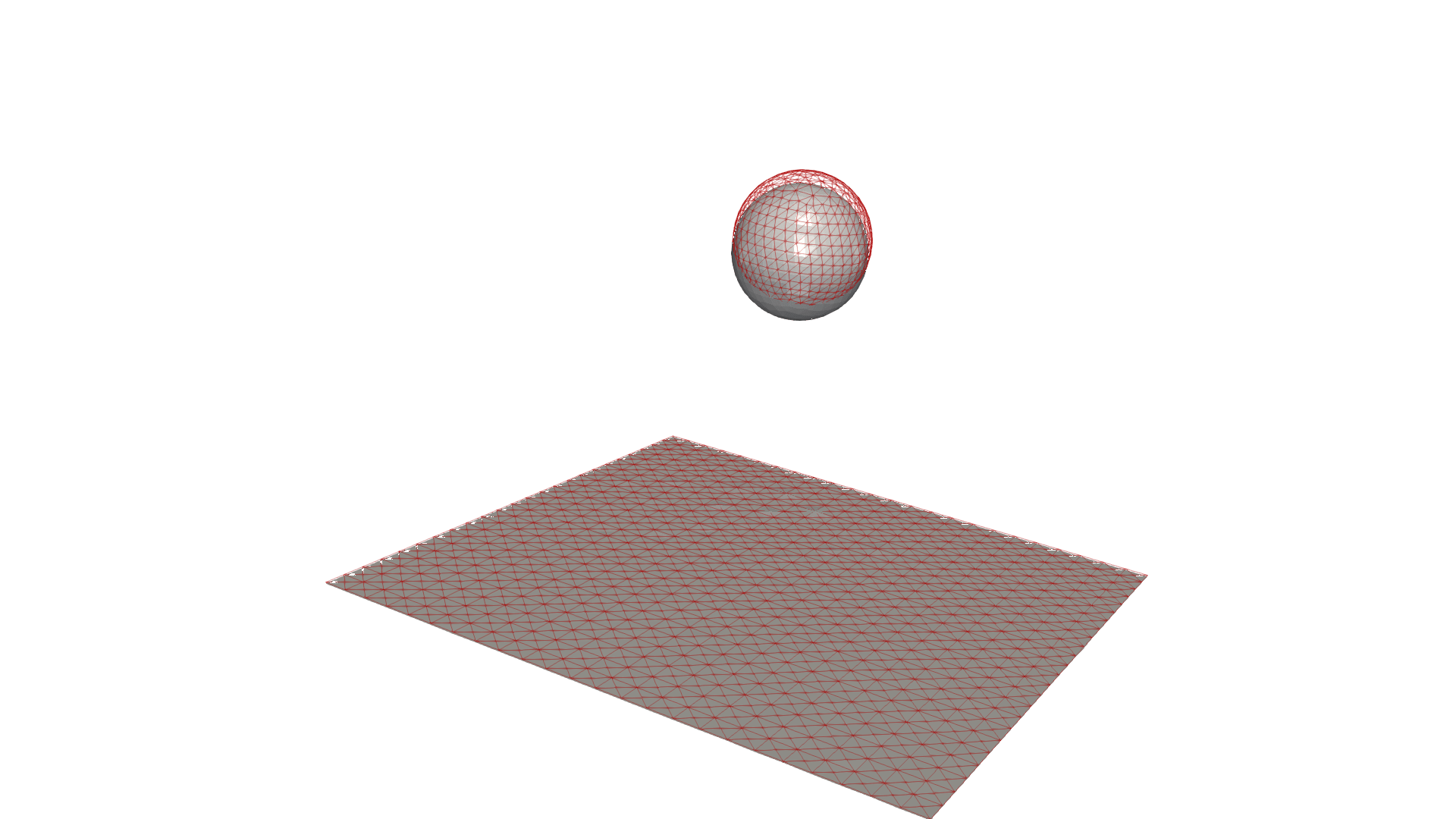}%
    \img{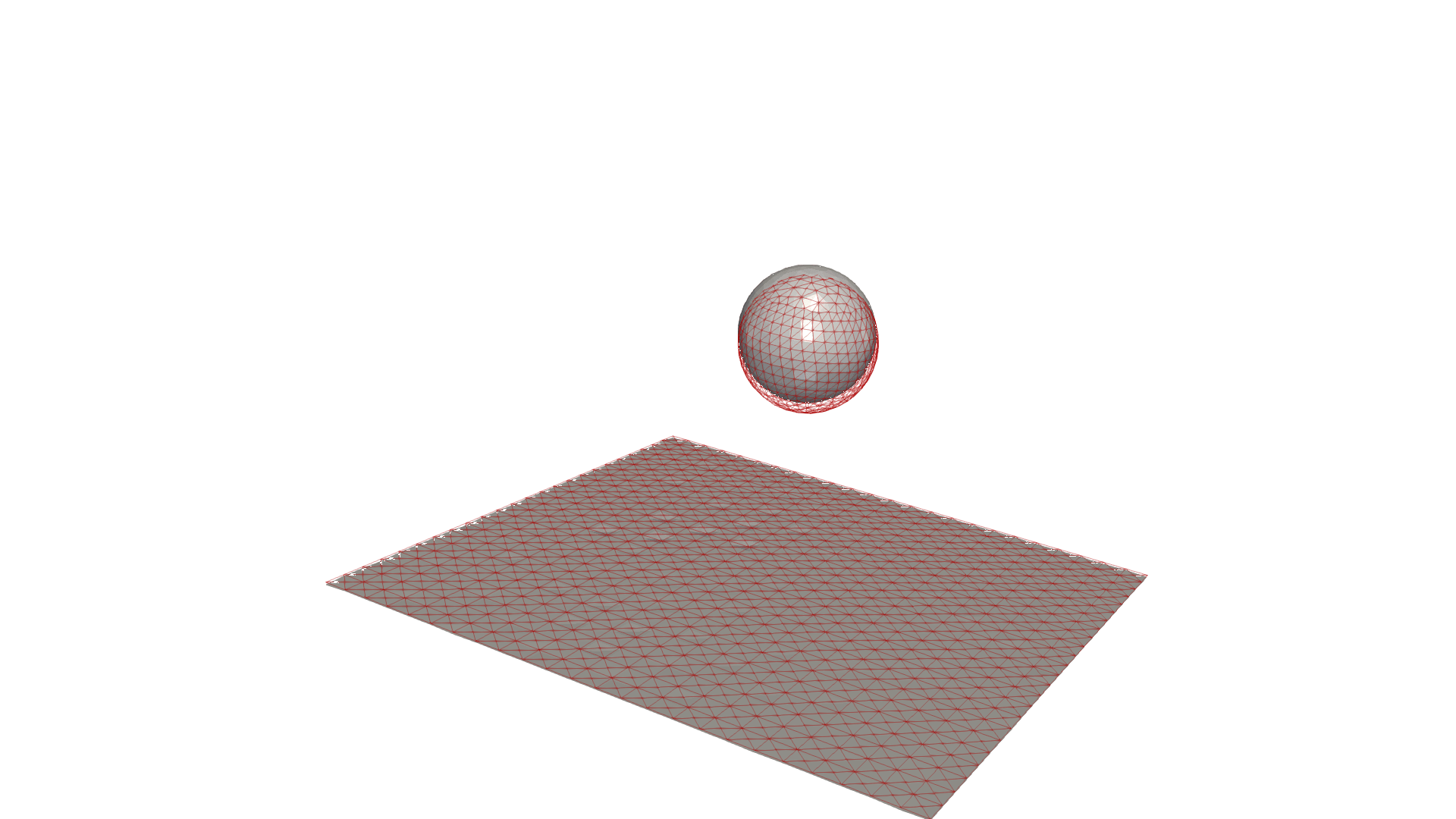}\\[0.15em]

    % Row 3: MANGO Dec
    \rowlabel{MANGO}%
    \img{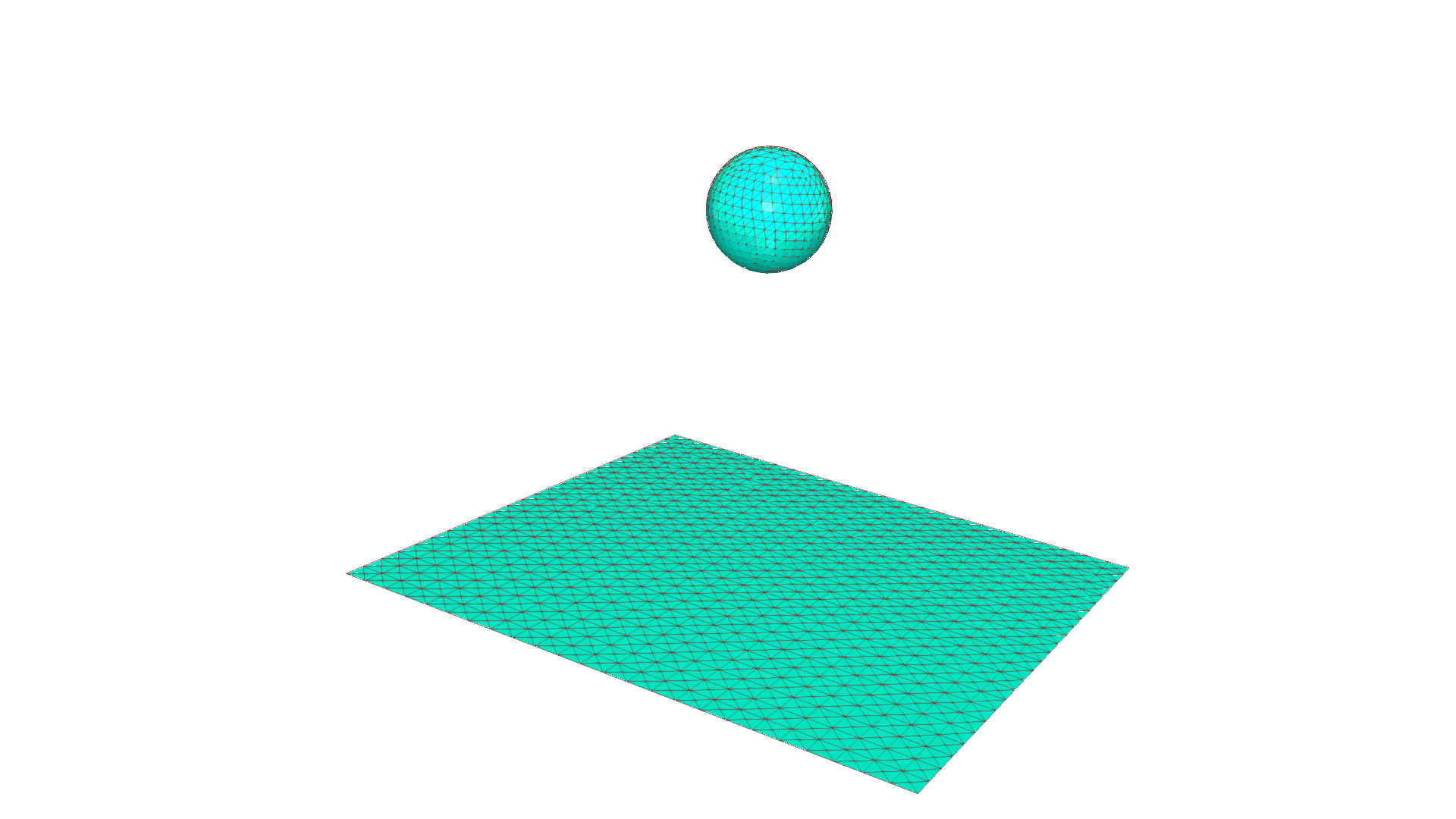}%
    \img{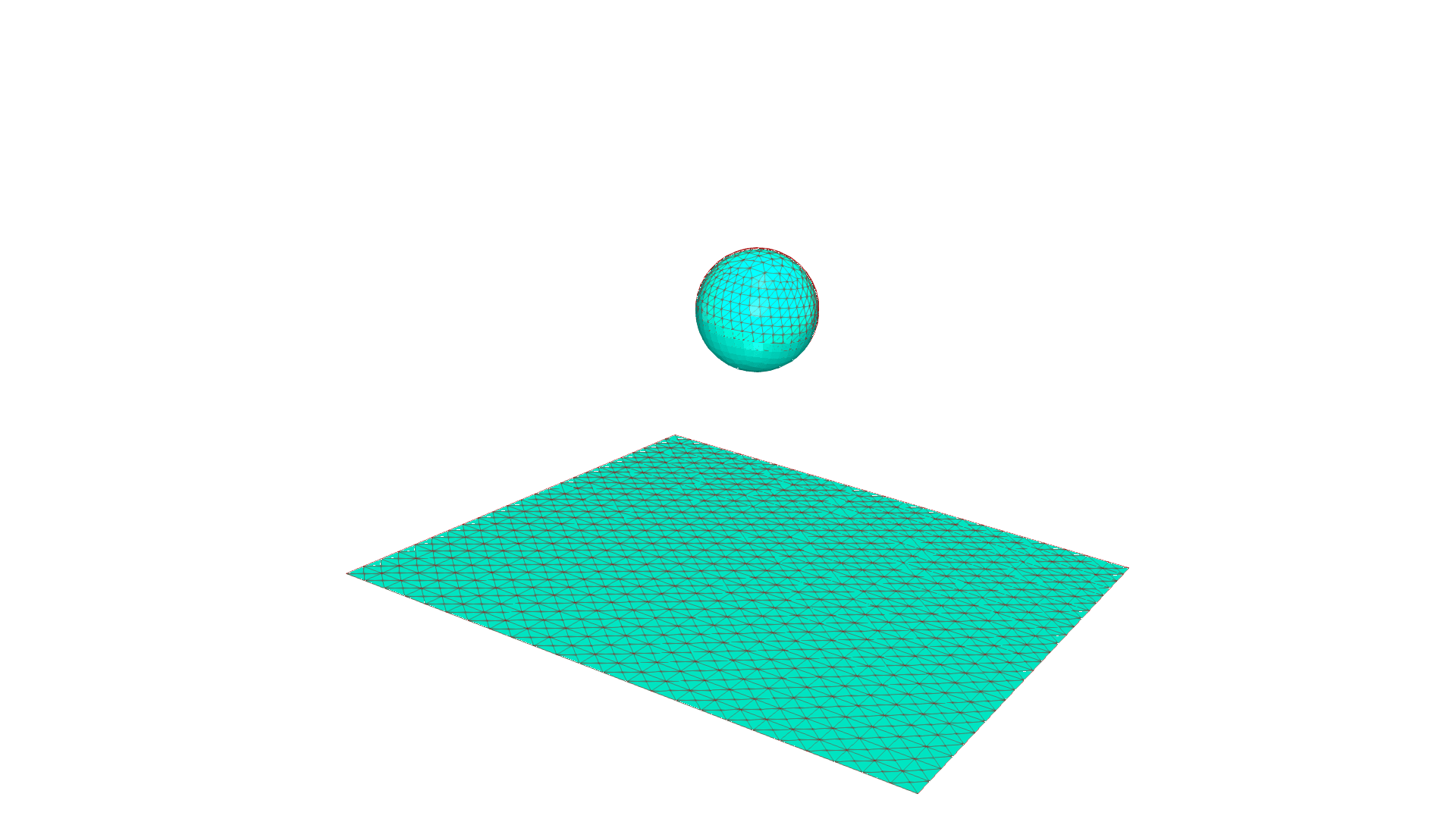}%
    \img{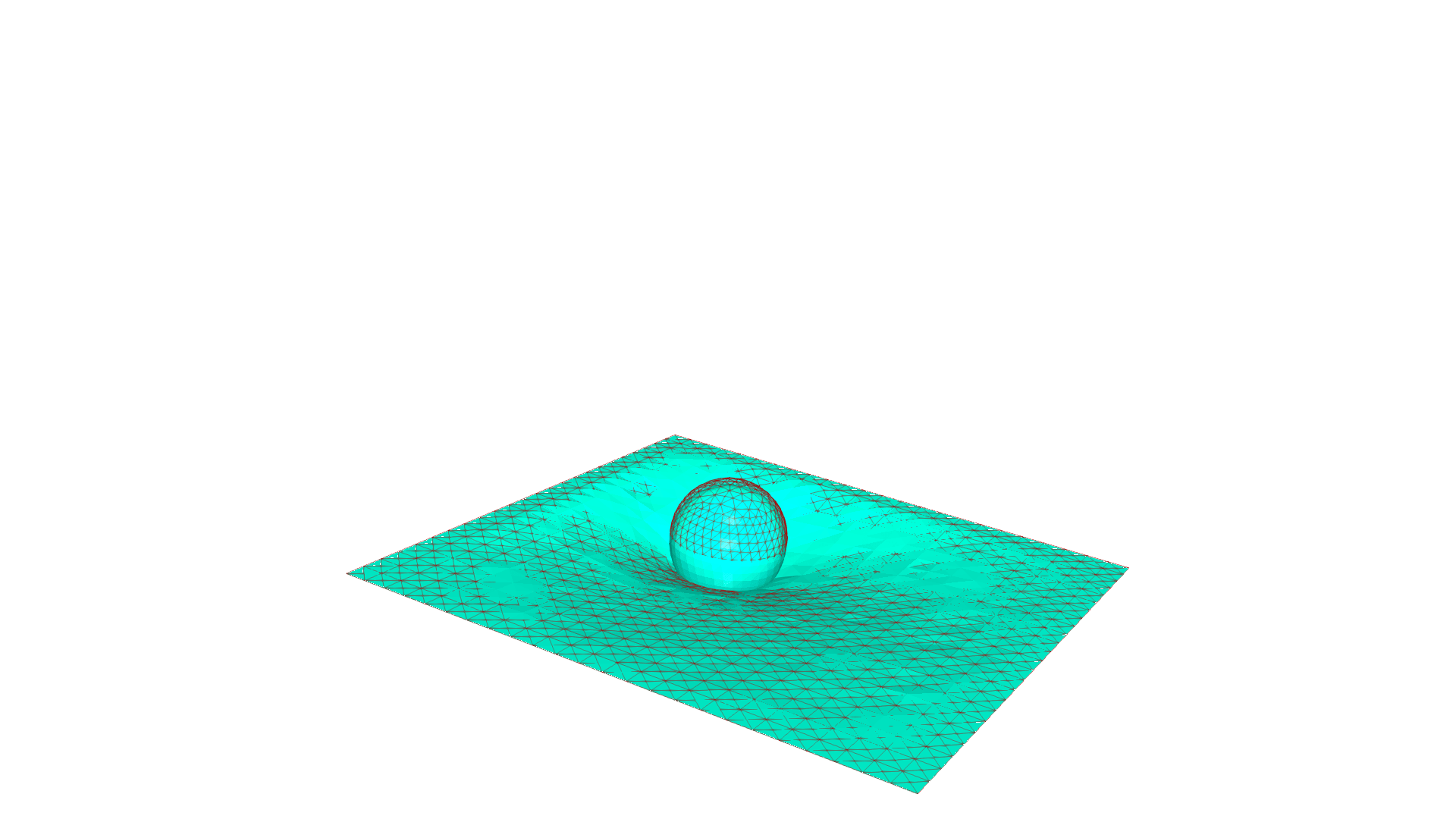}%
    \img{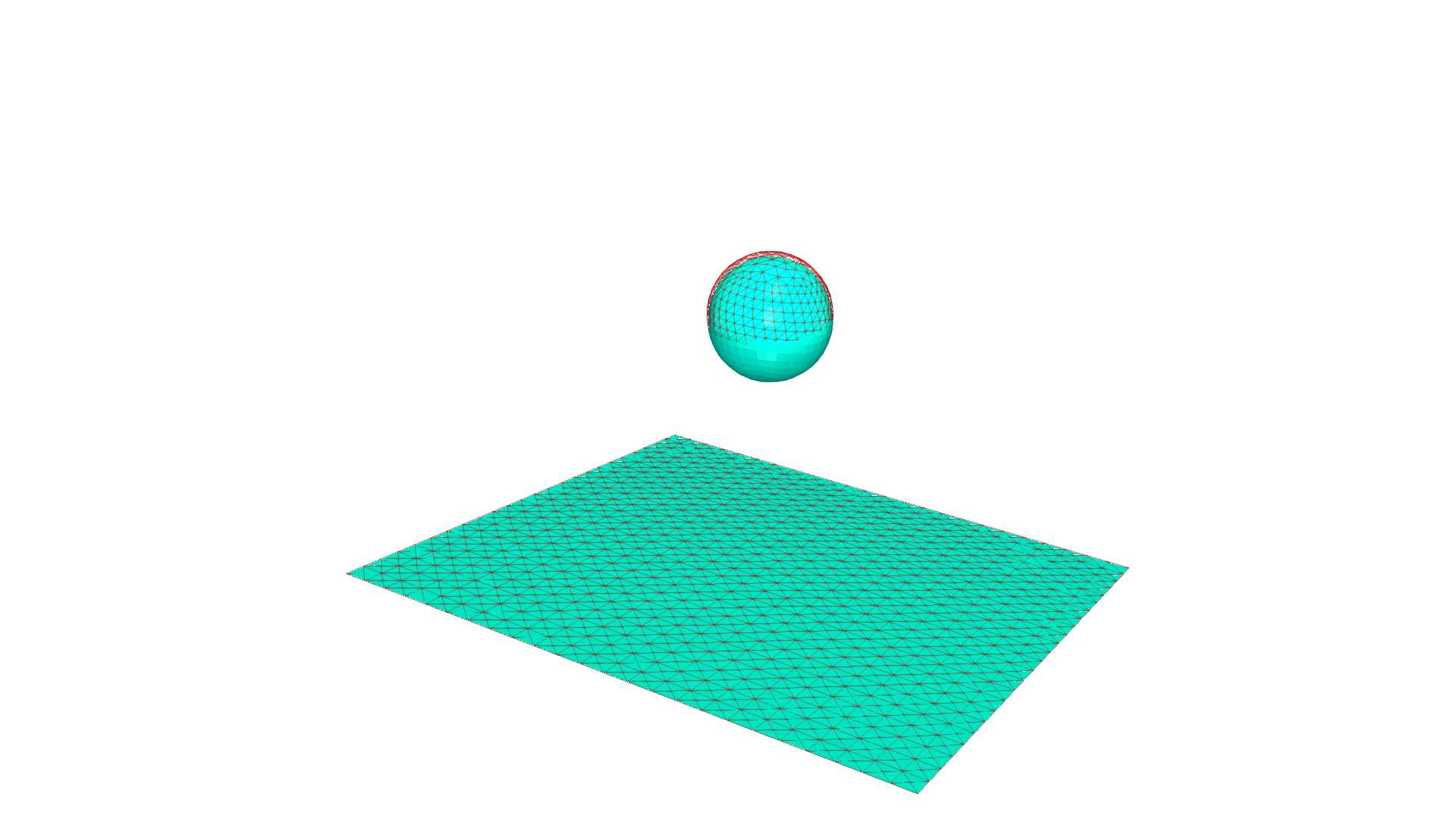}%
    \img{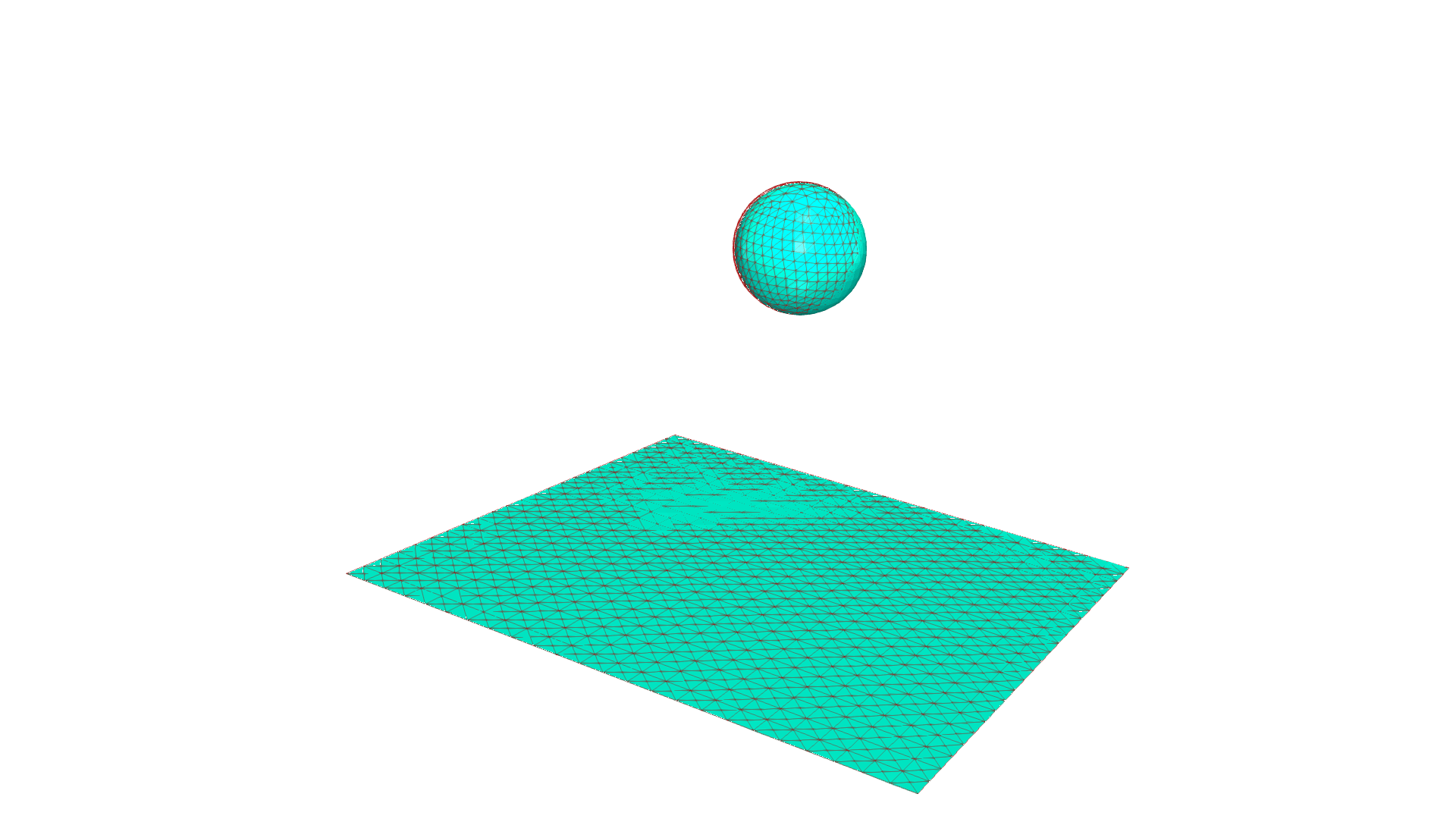}%
    \img{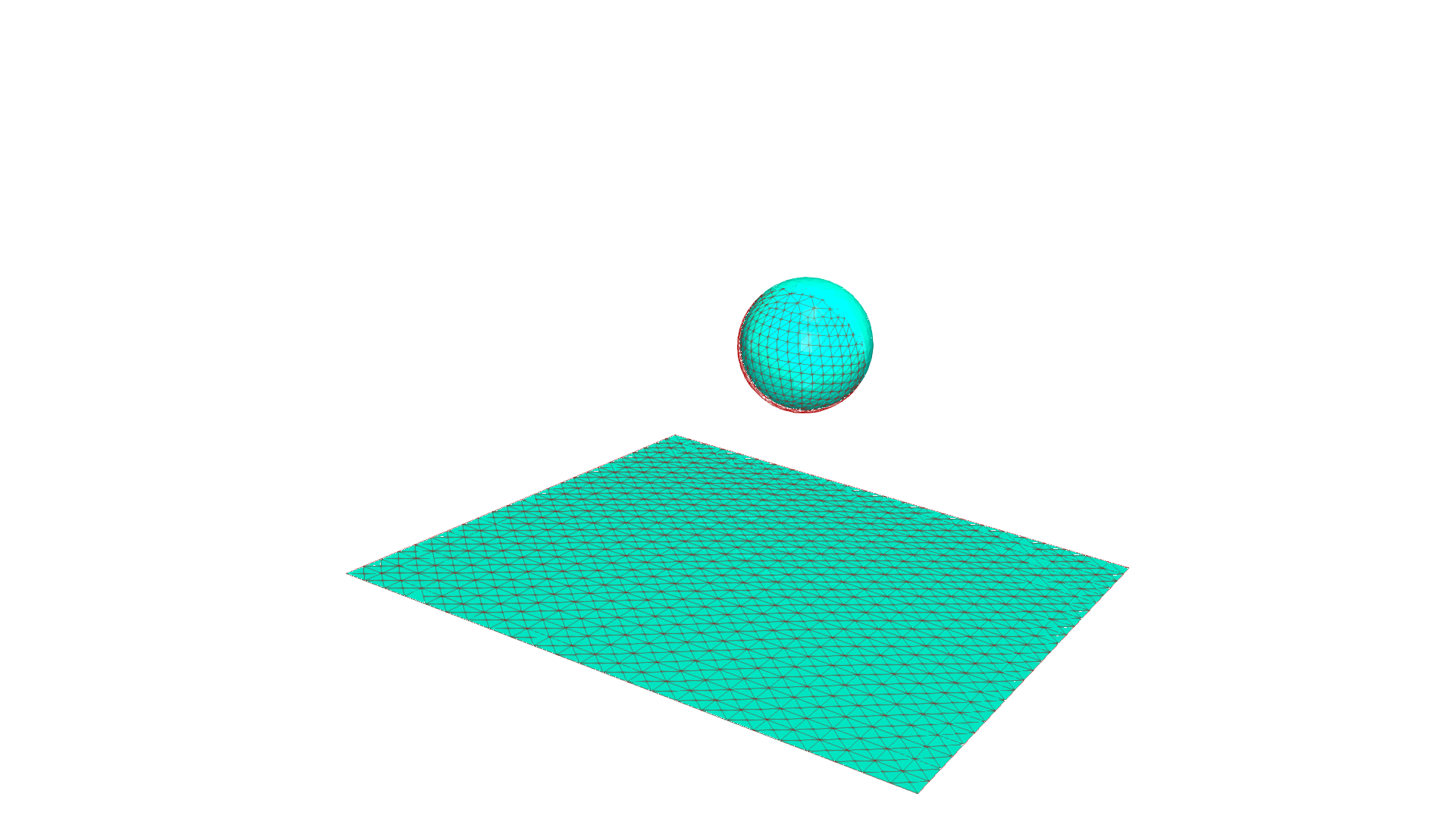}\\[0.15em]

    % Row 4: MANGO Oracle
    \rowlabel{Oracle}%
    \img{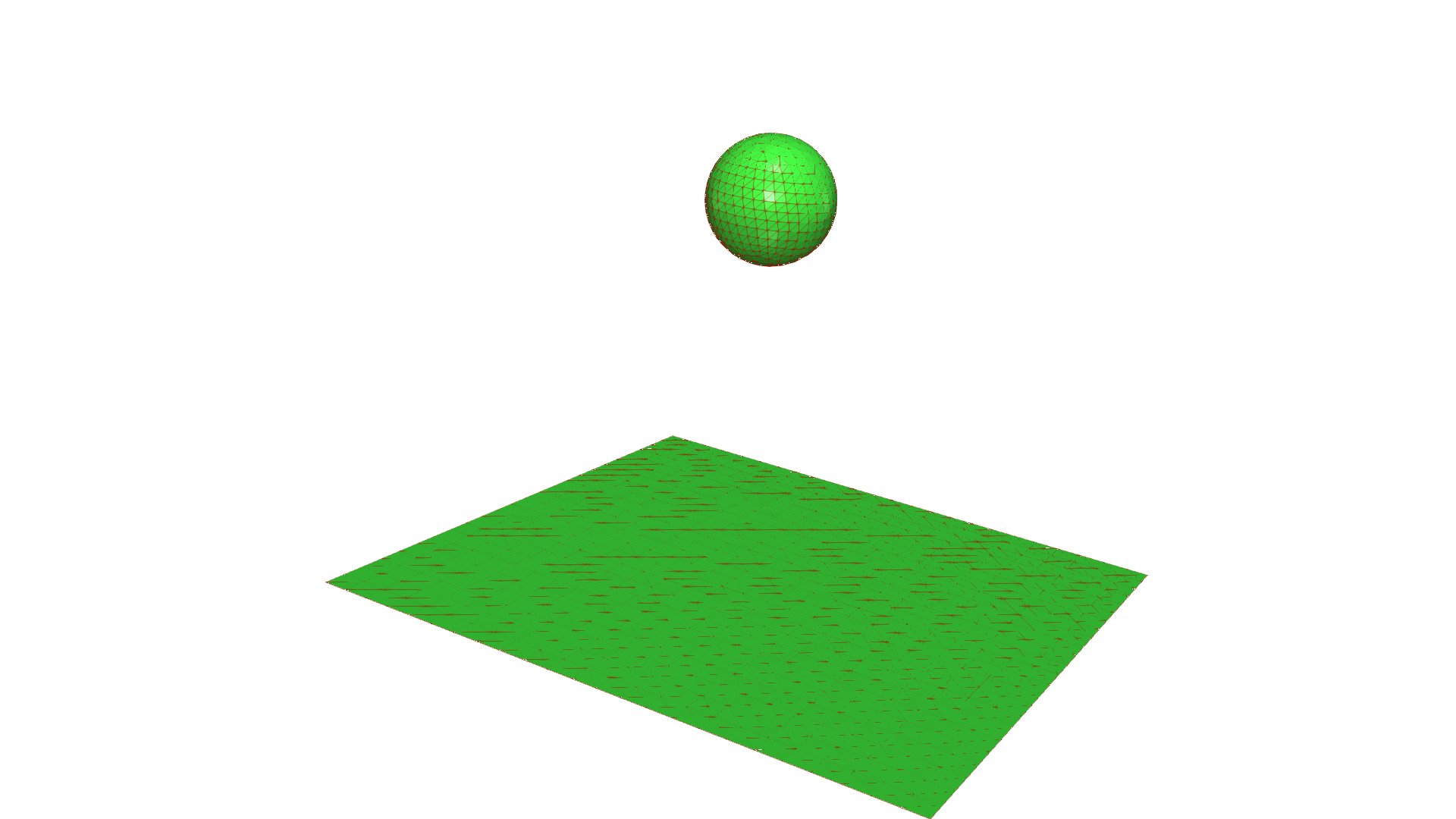}%
    \img{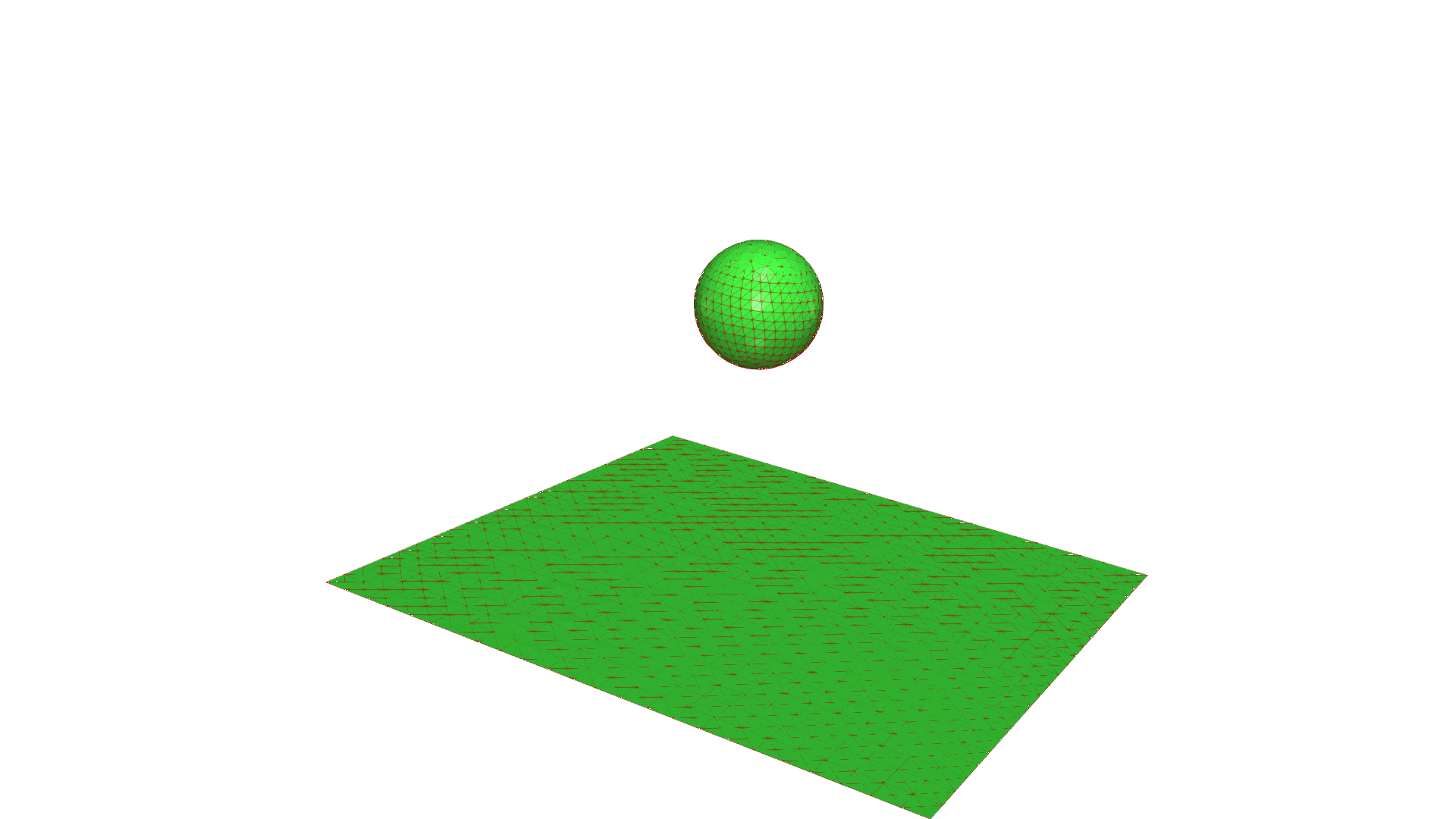}%
    \img{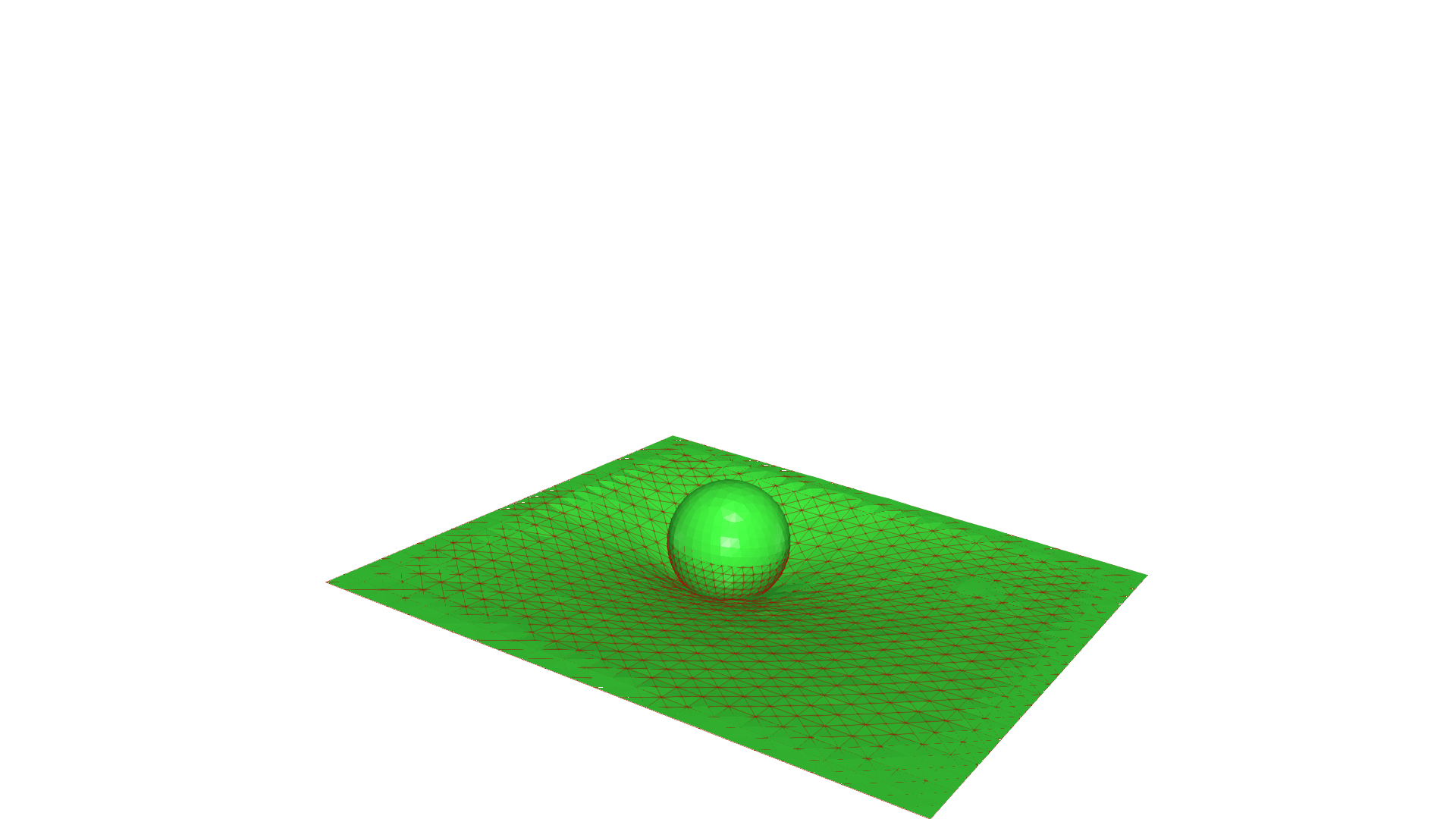}%
    \img{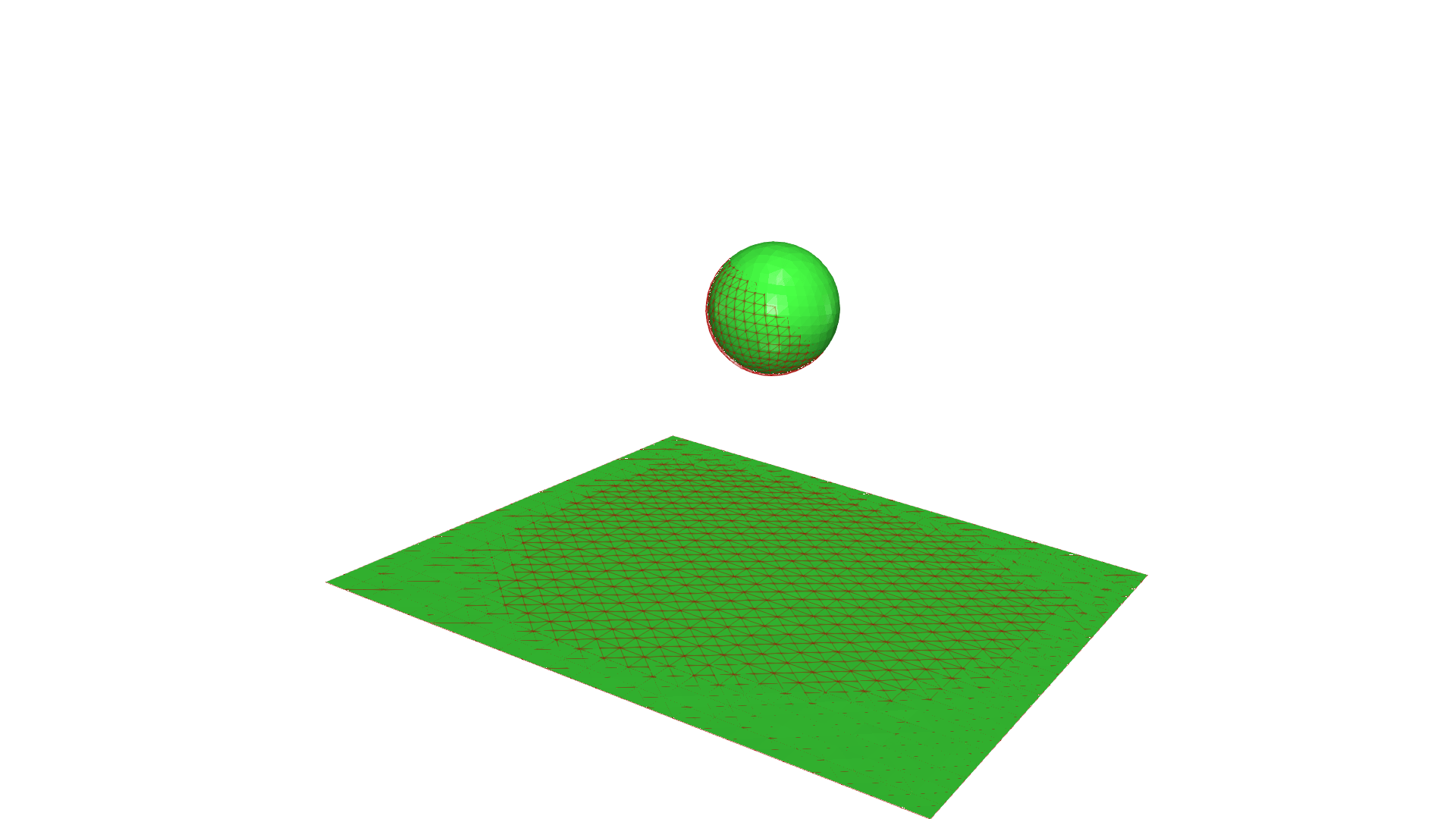}%
    \img{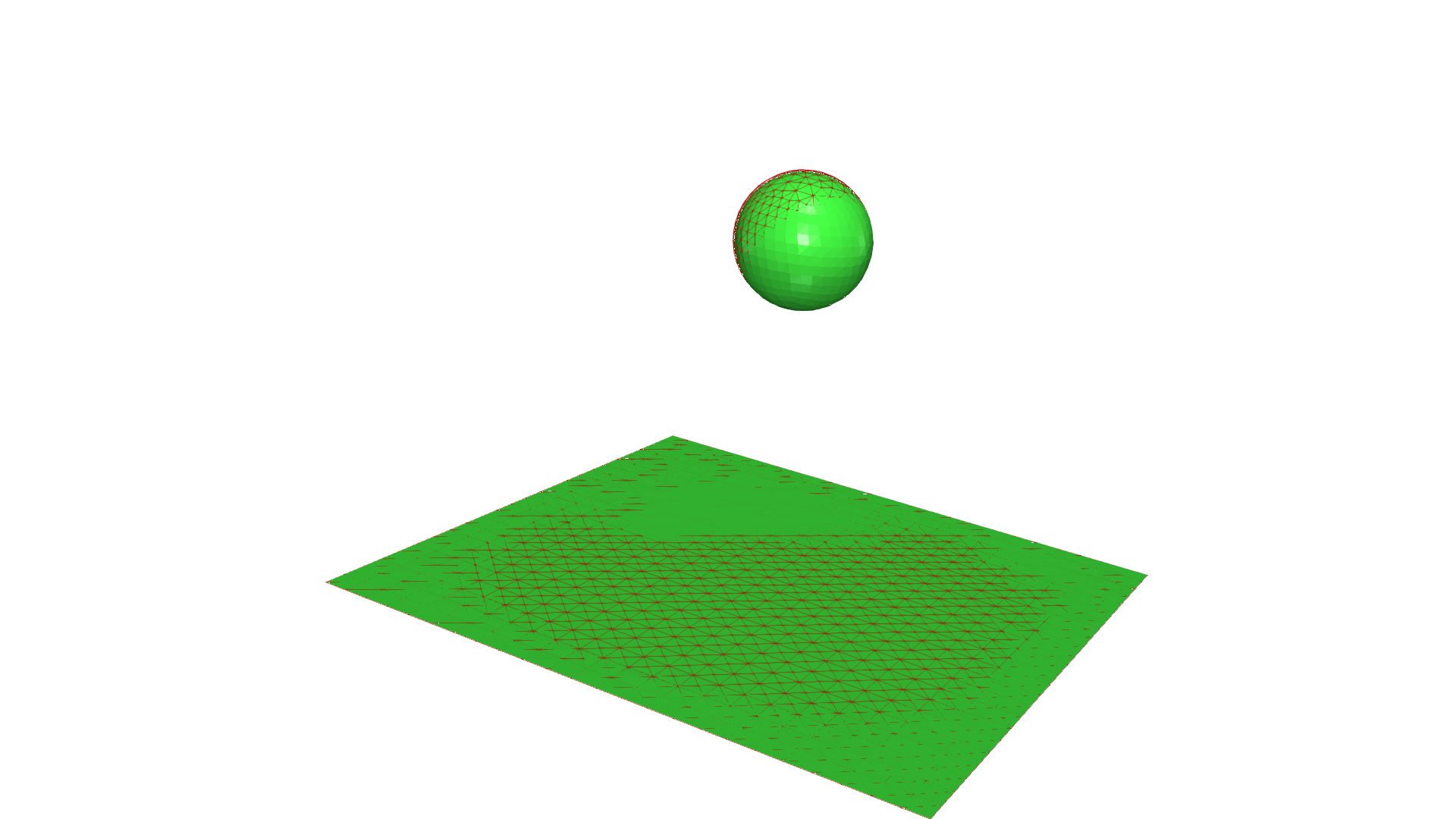}%
    \img{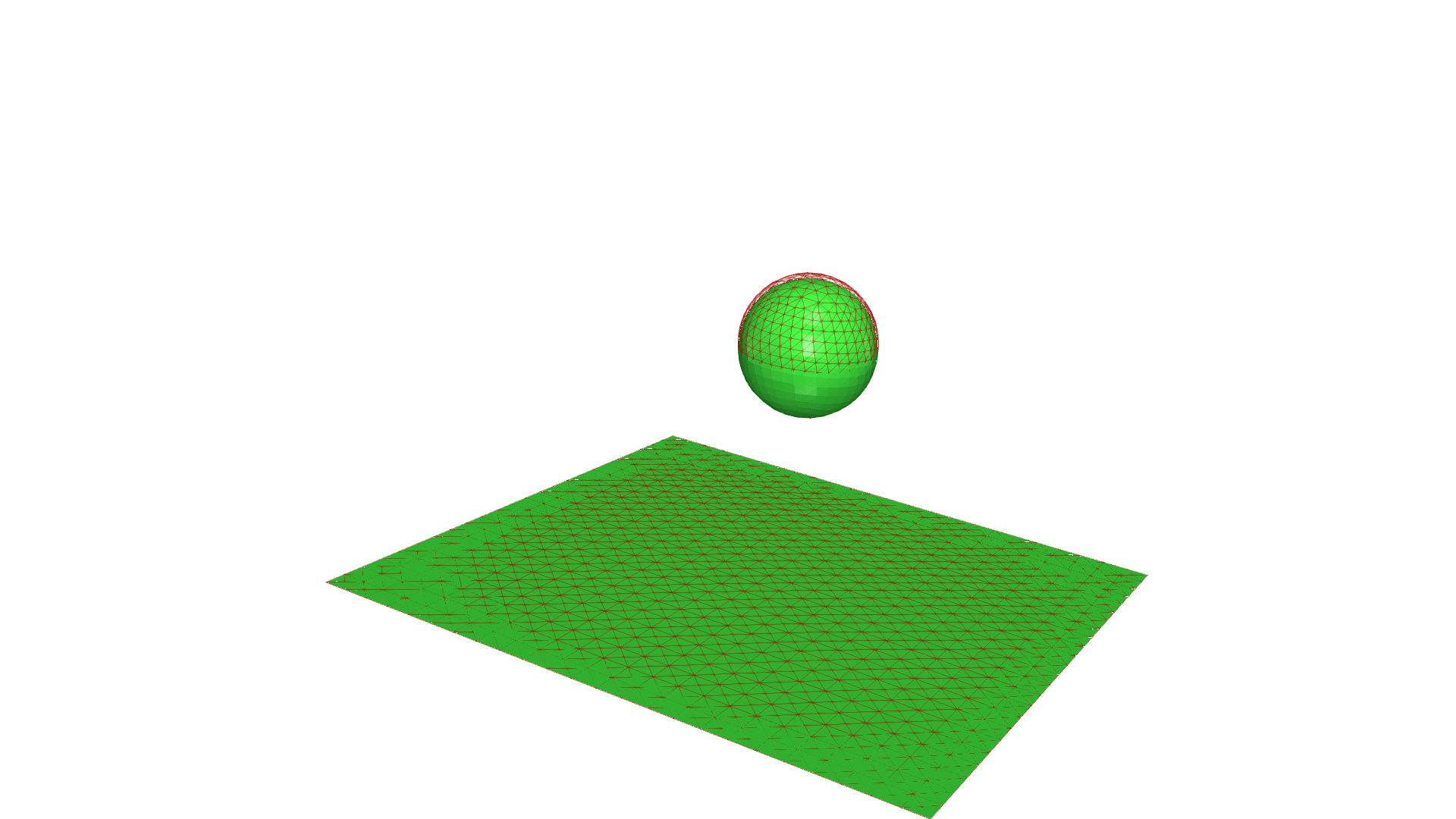}\\[0.15em]
    \par\vspace{0.15em}

    % Row 5: MGN
    \rowlabel{No Context \\ (MGN)}%
    \img{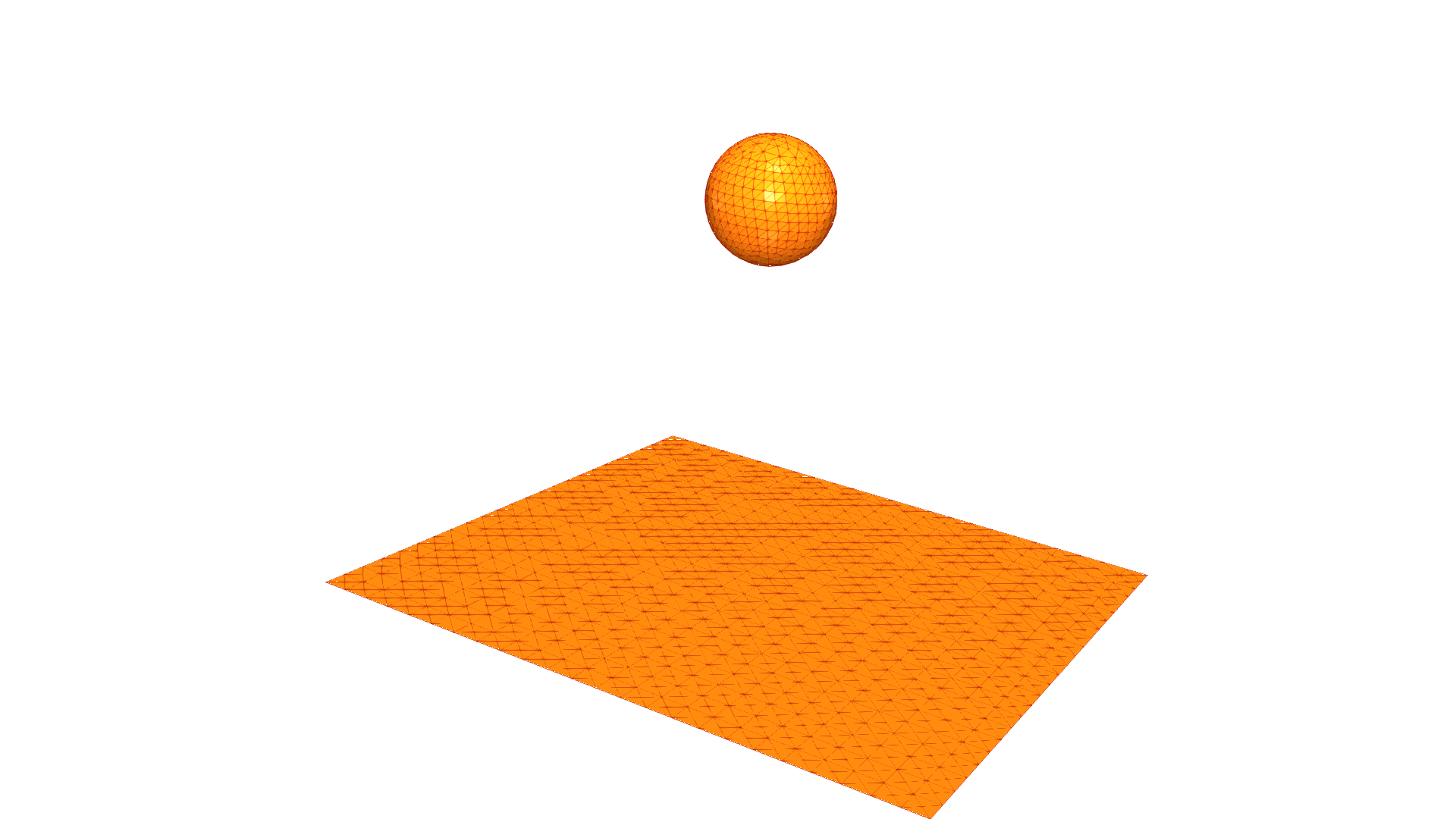}%
    \img{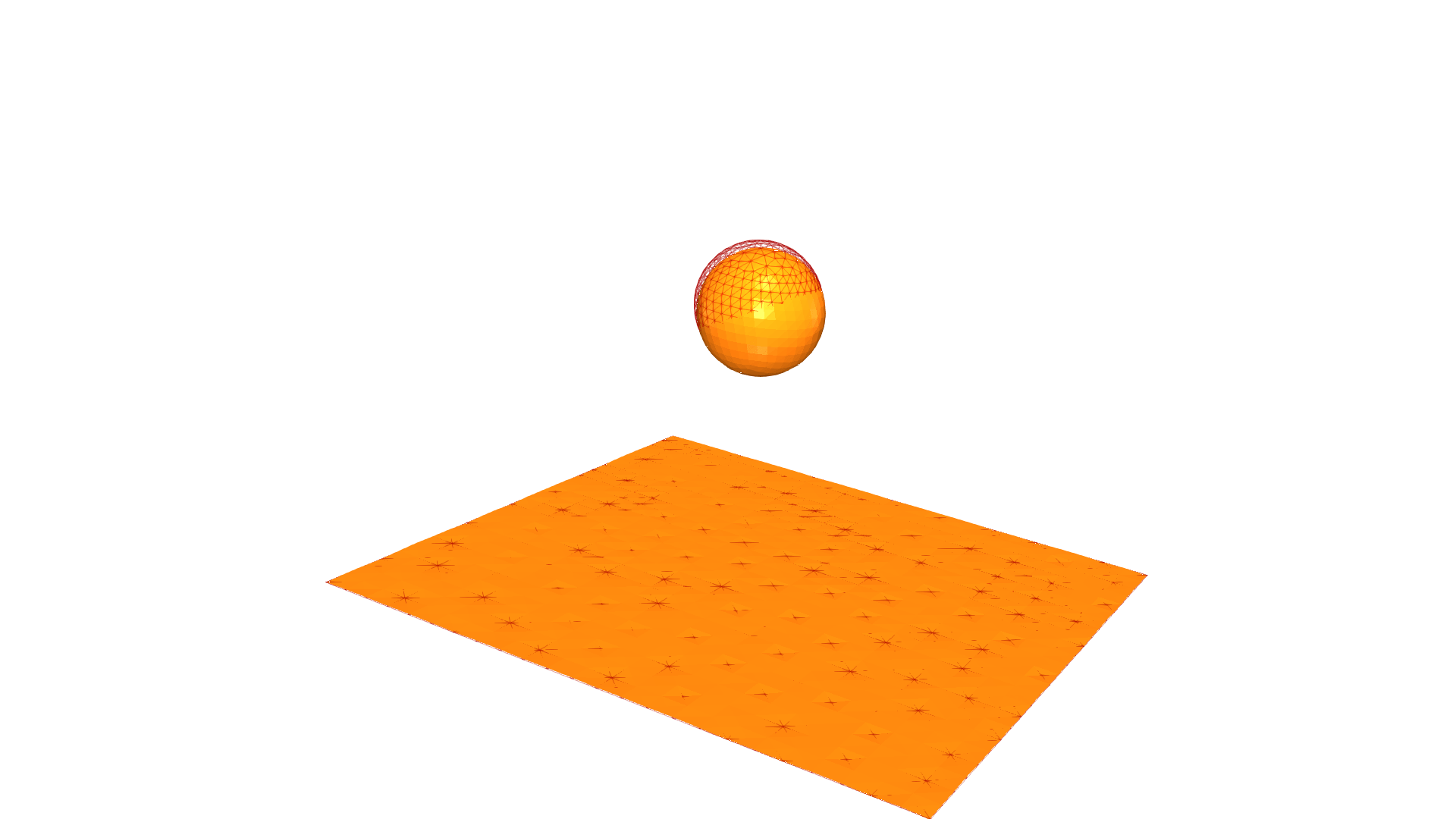}%
    \img{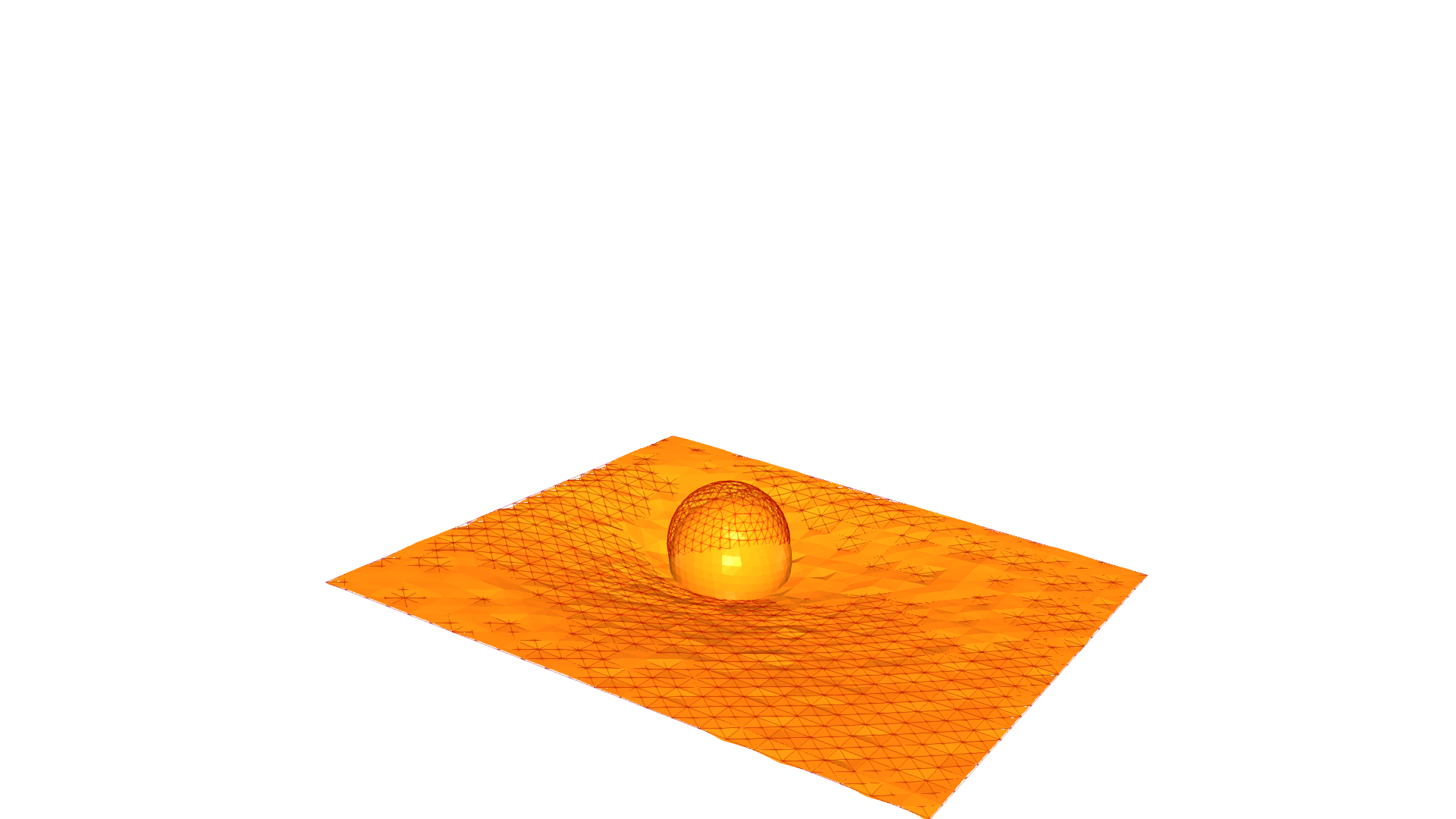}%
    \img{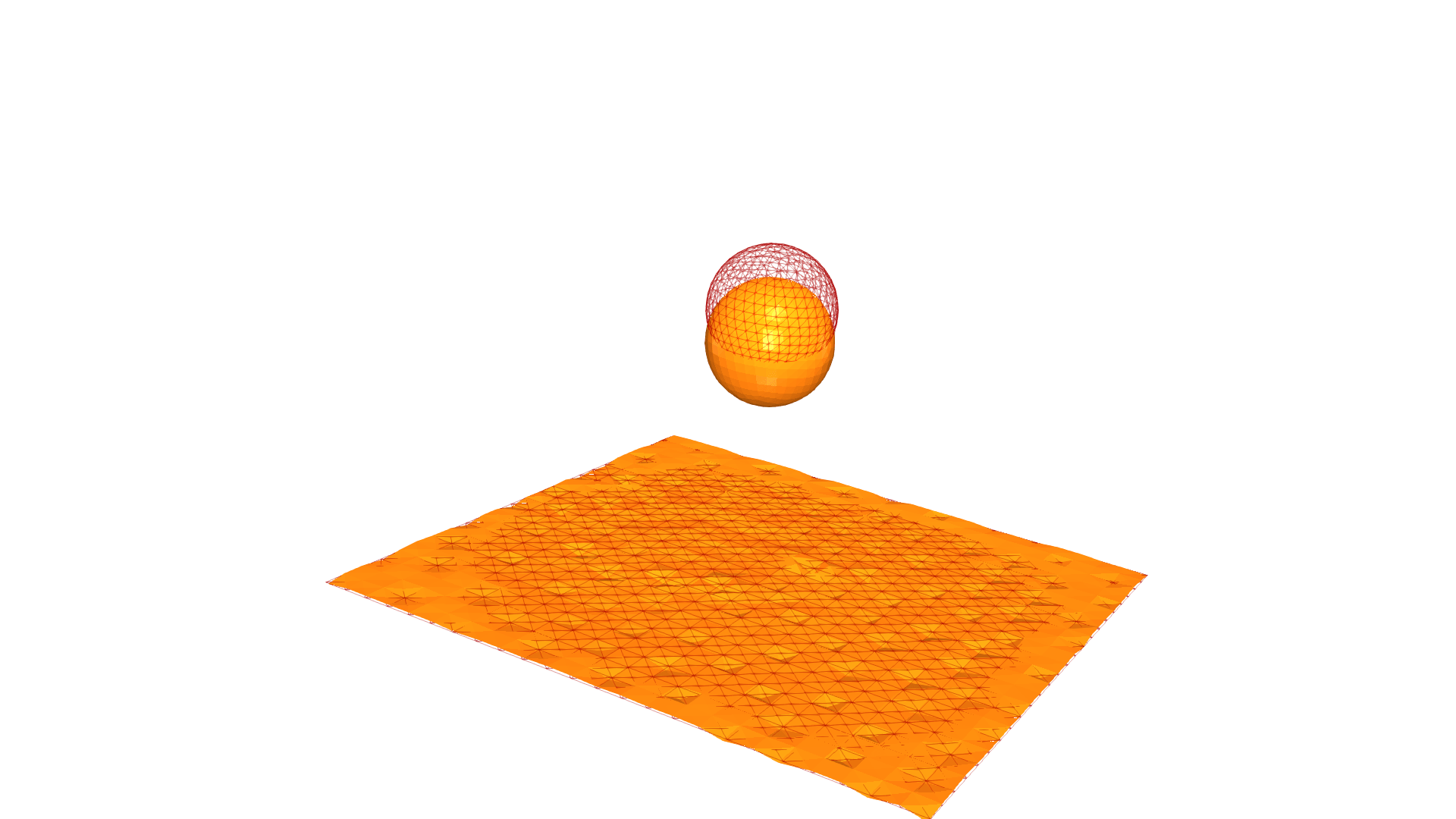}%
    \img{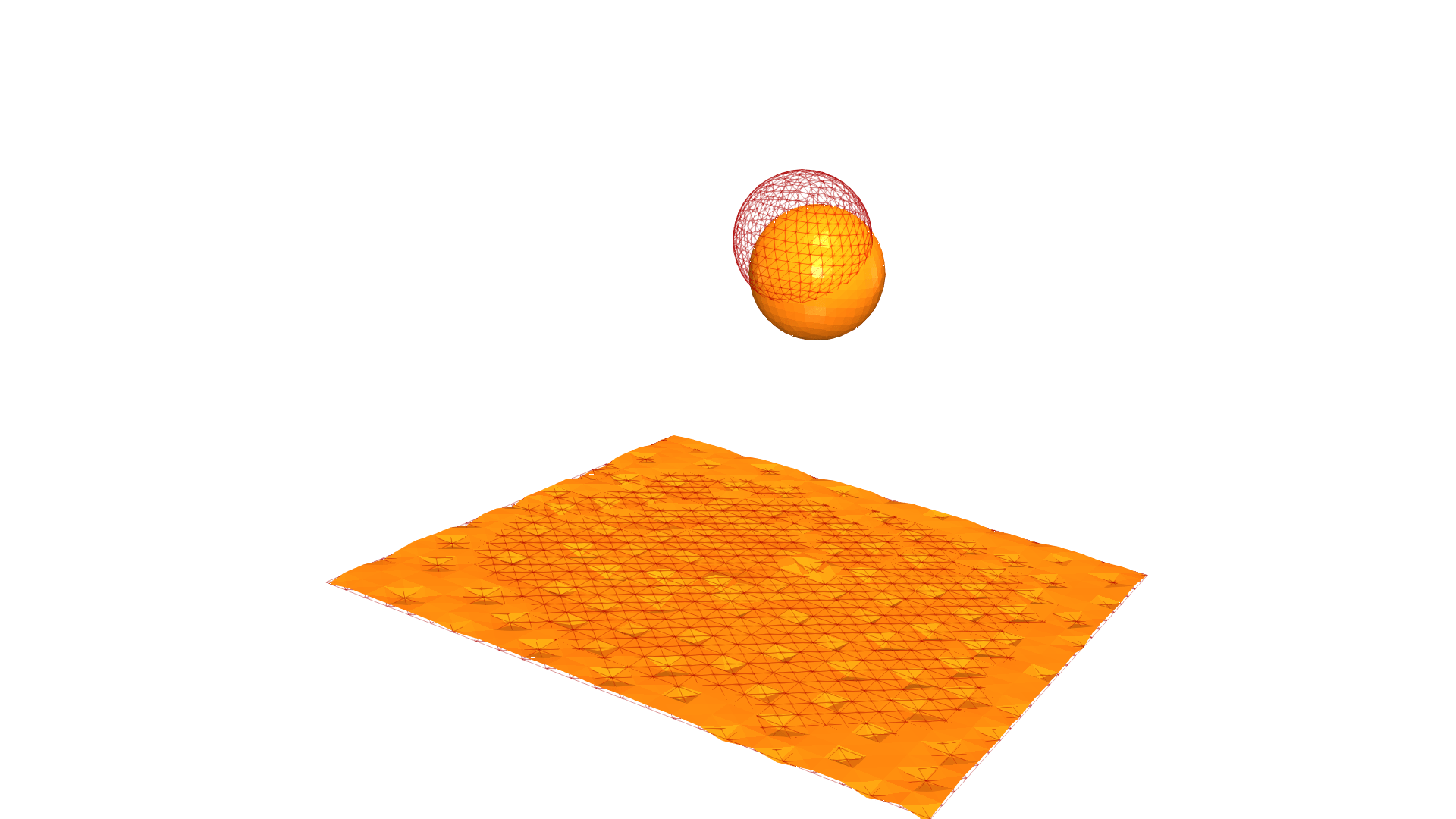}%
    \img{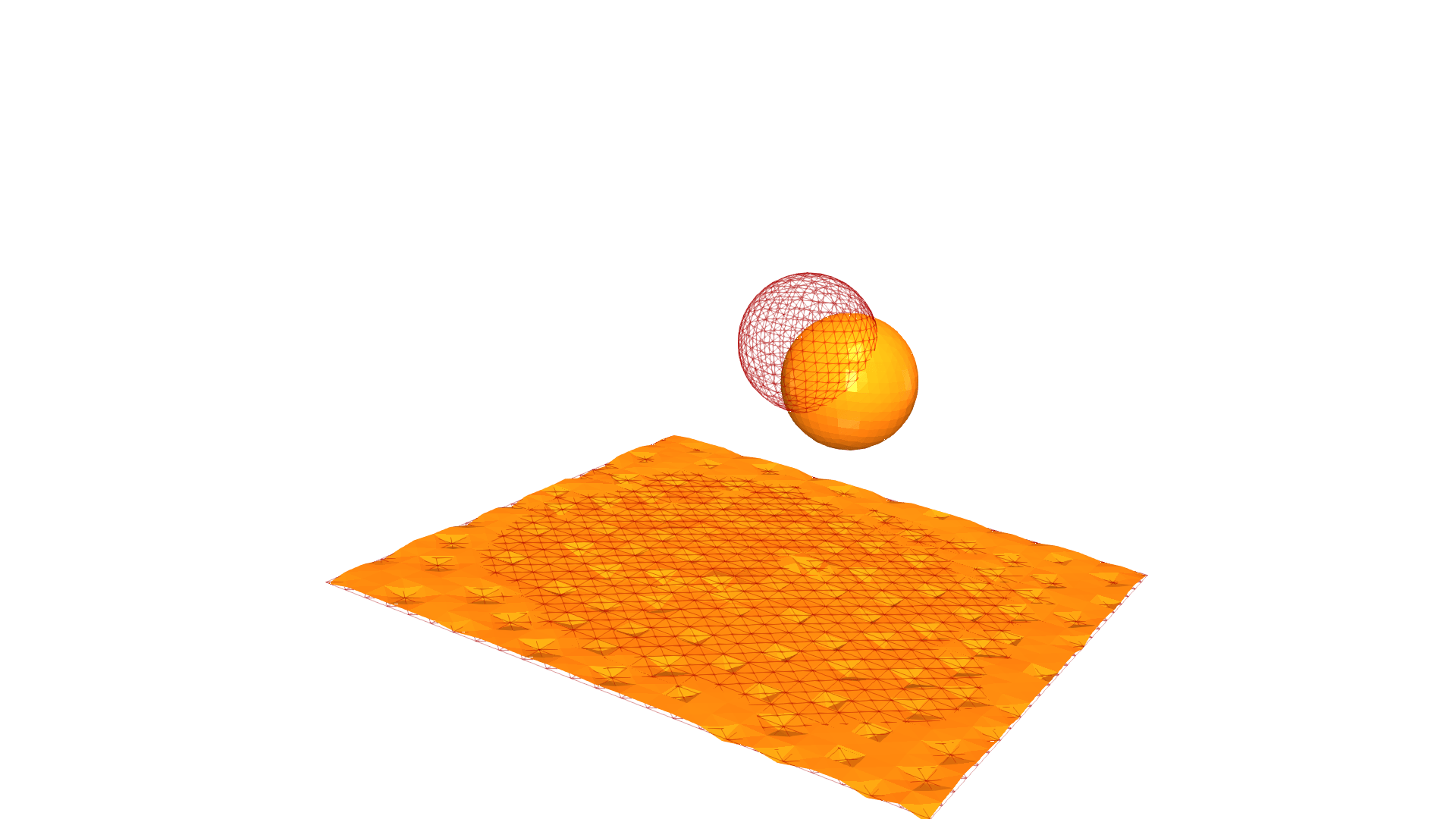}\\[0.15em]

    % Row 6: MGN Oracle
    \rowlabel{Oracle \\ (MGN)}%
    \img{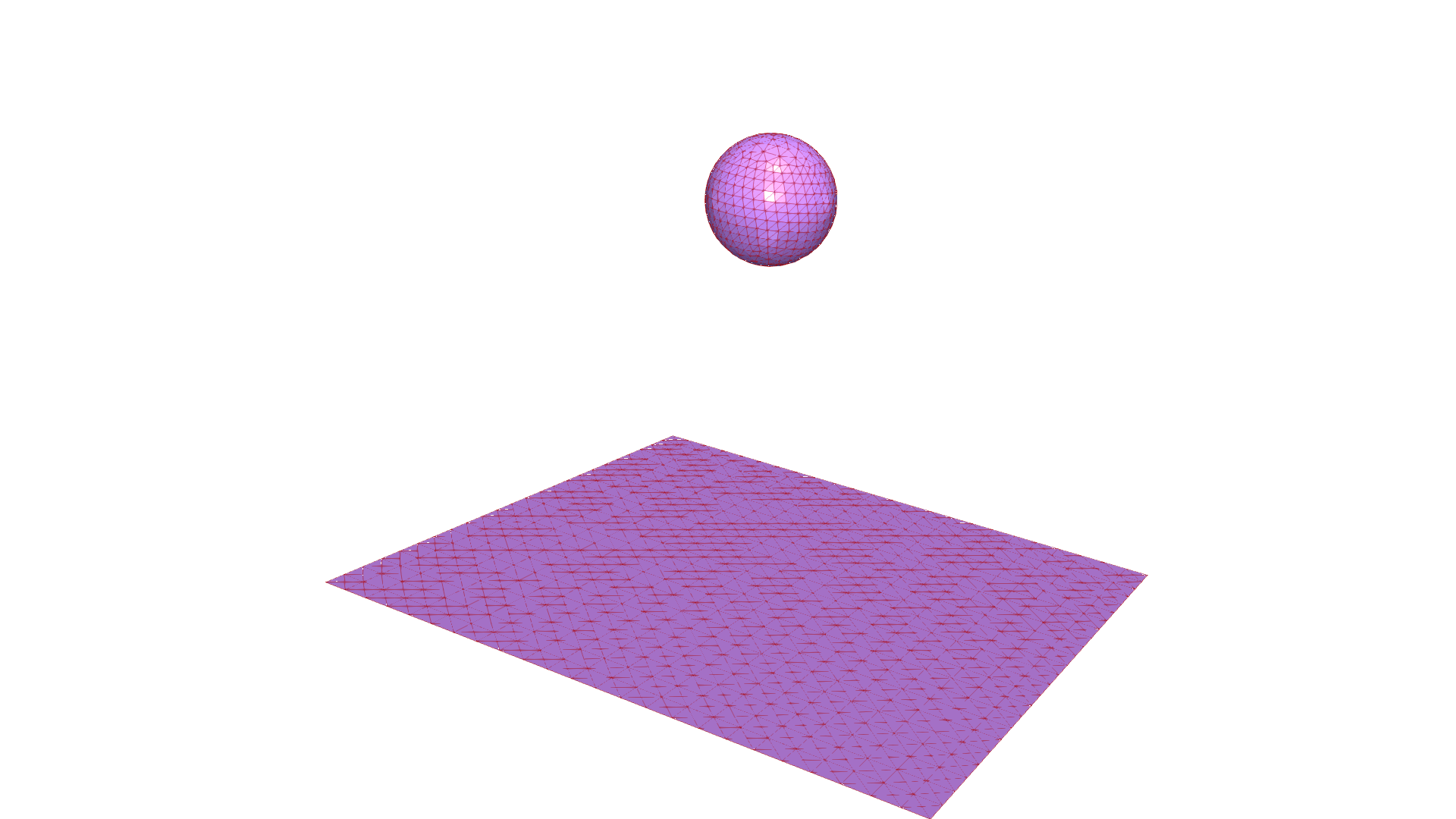}%
    \img{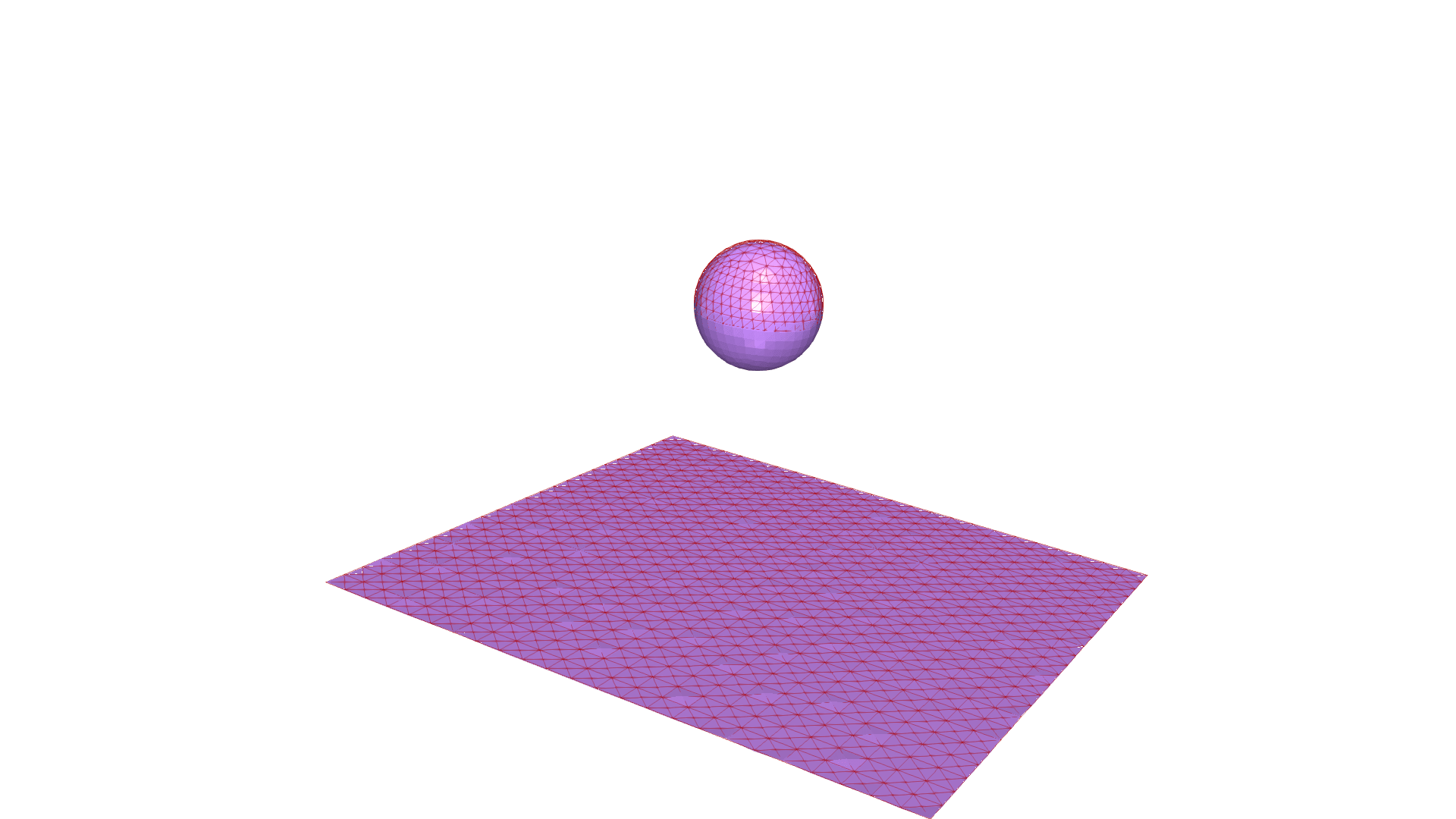}%
    \img{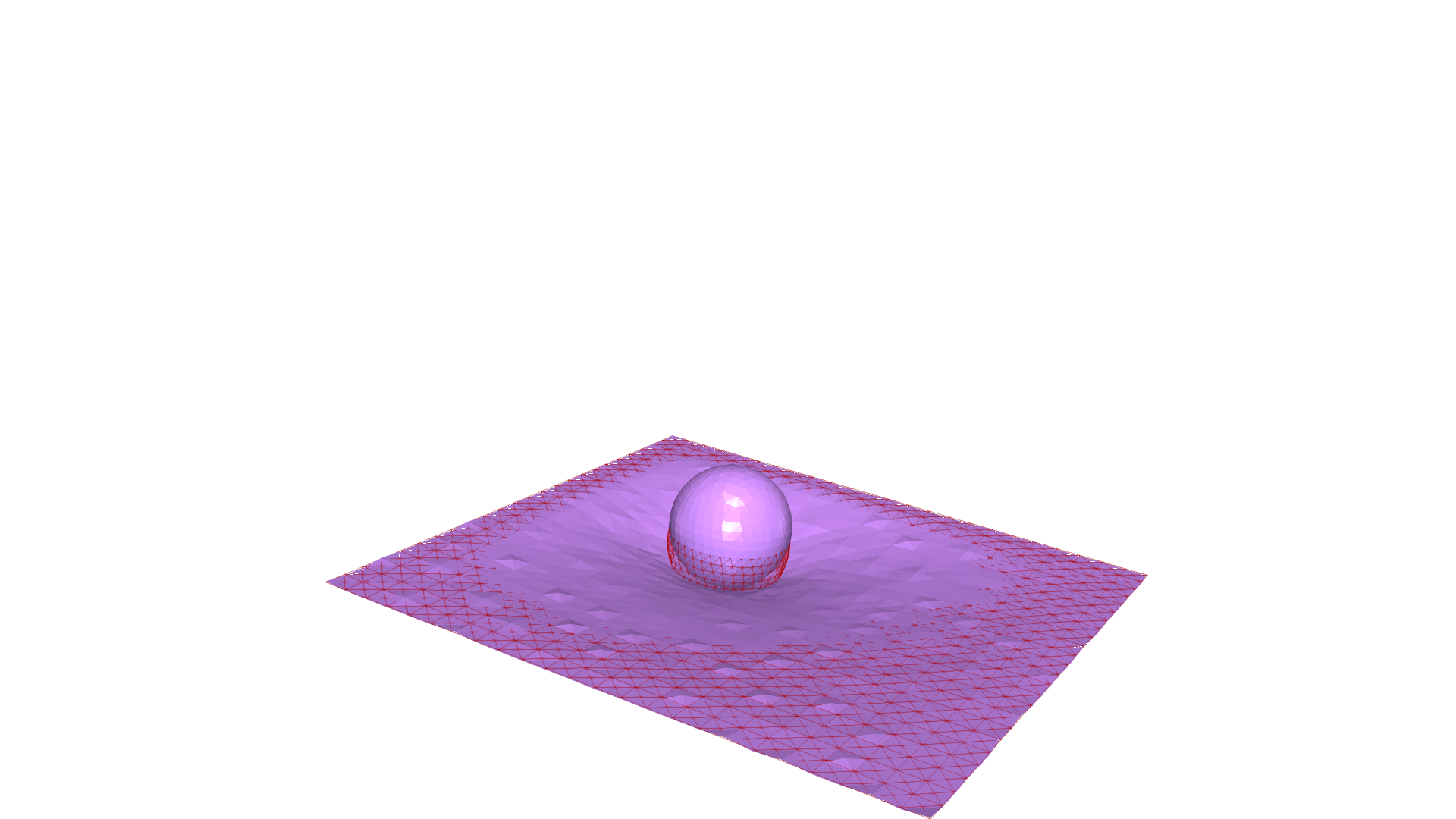}%
    \img{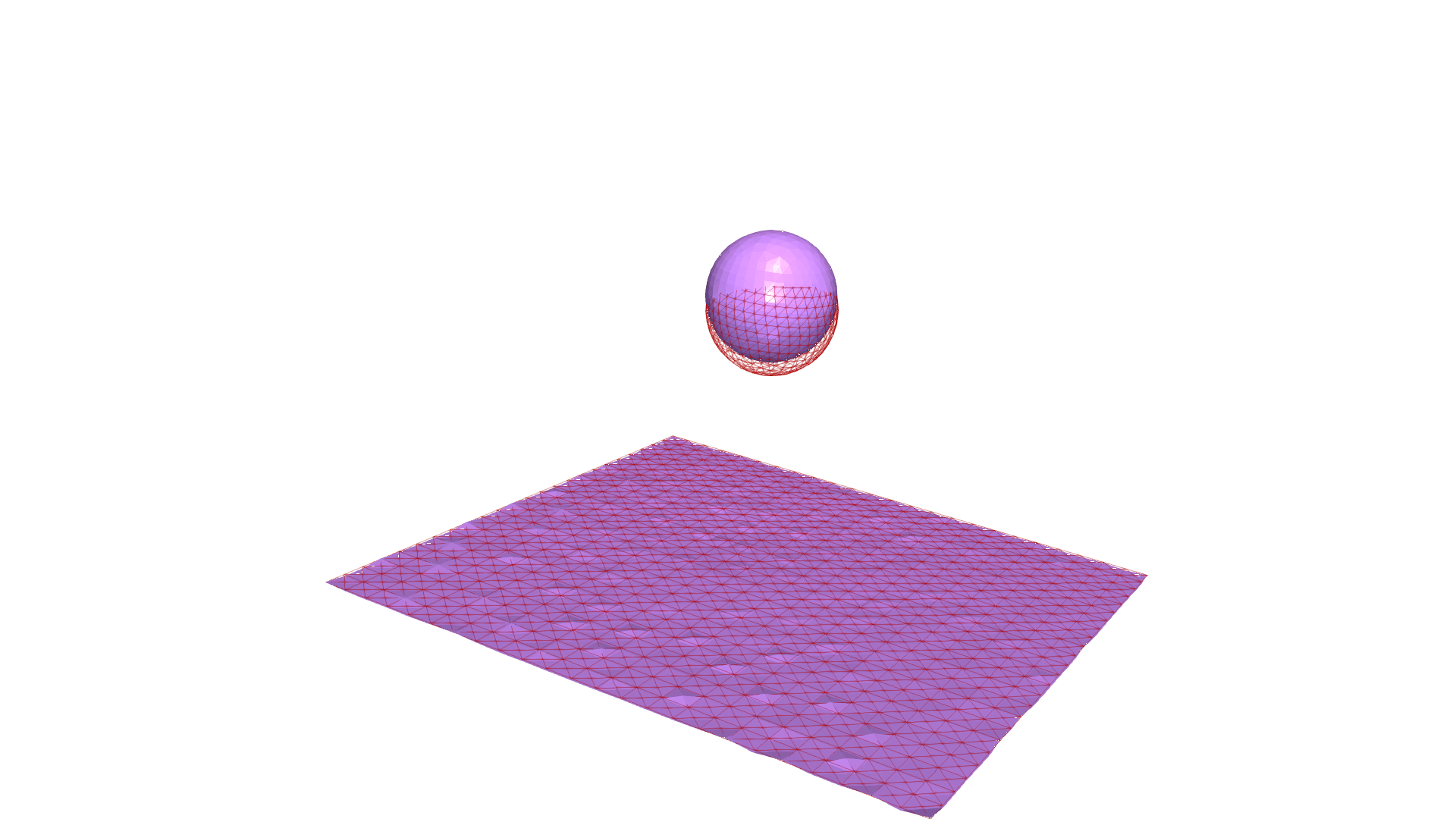}%
    \img{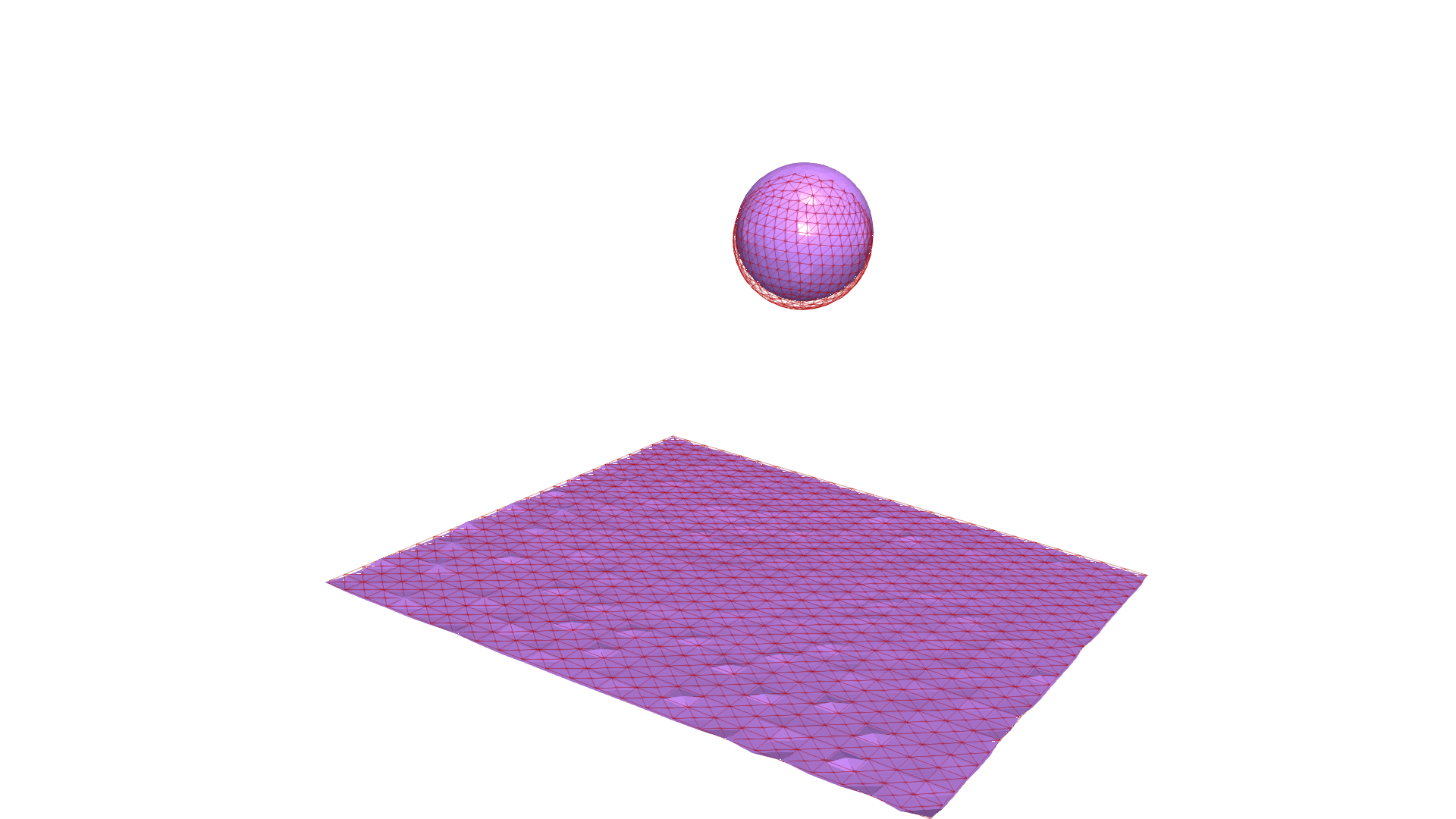}%
    \img{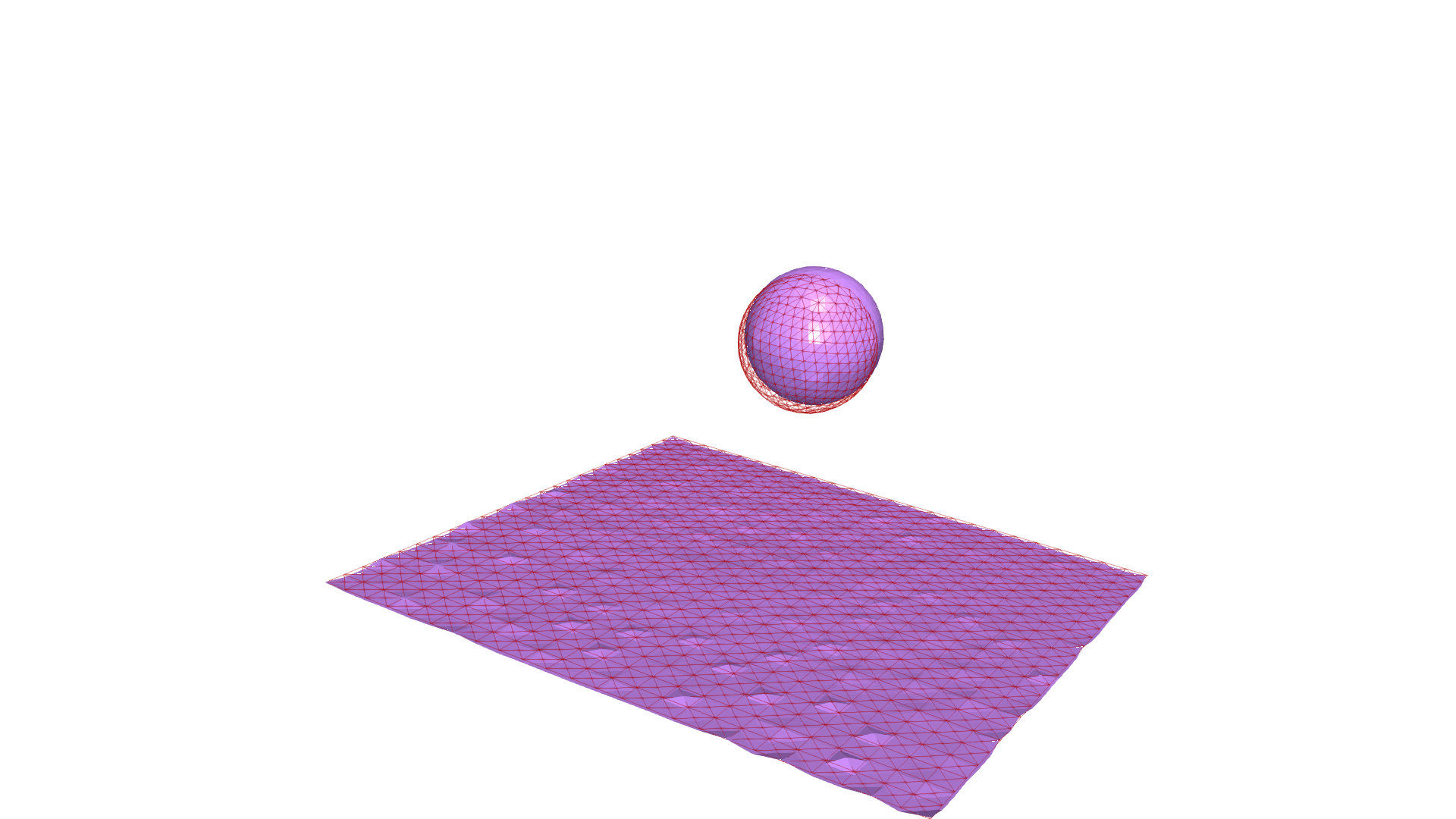}\\[0.15em]

    % Row 7: GNN Encoder
    \rowlabel{GNN \\ Encoder}%
    \img{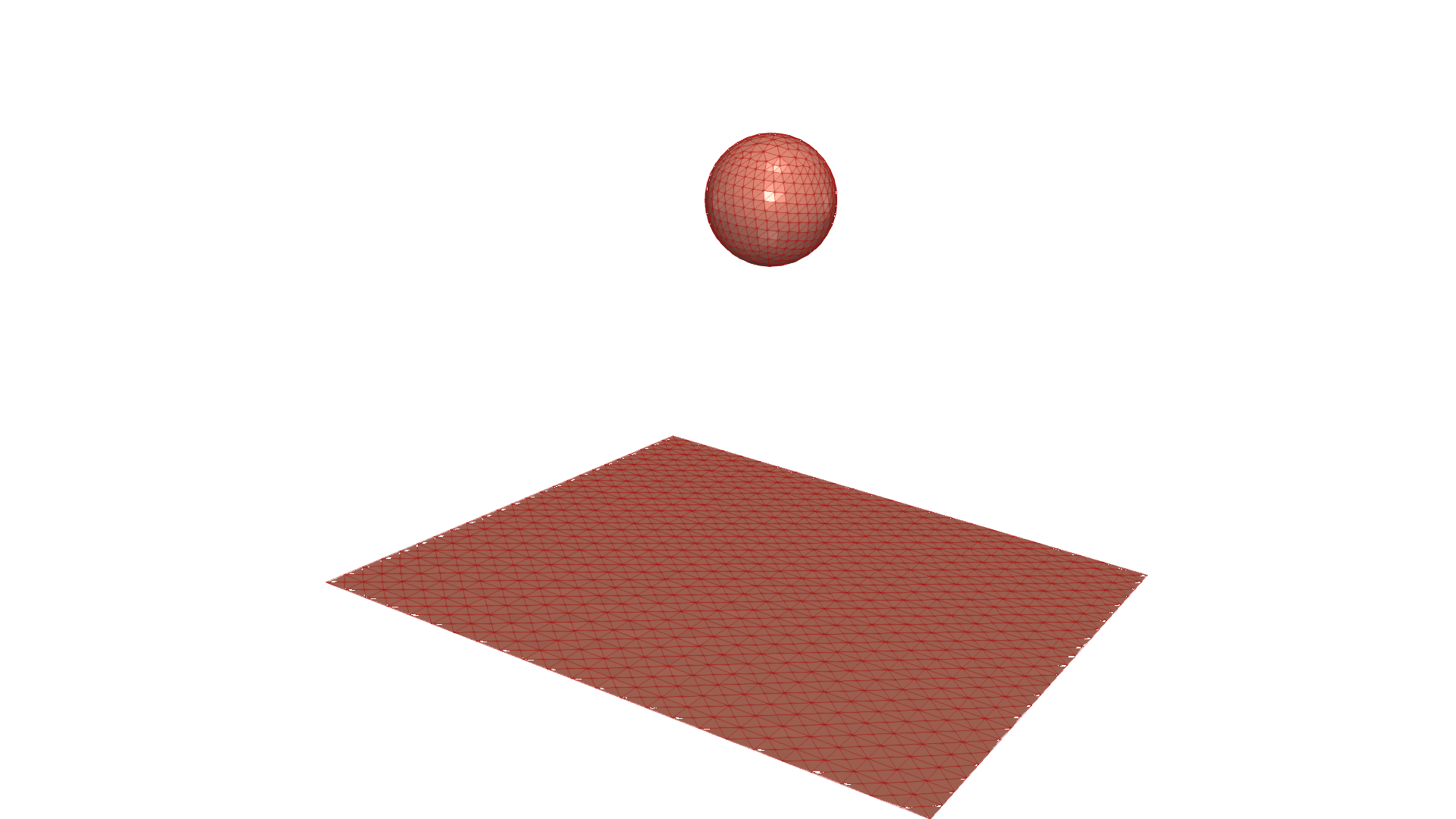}%
    \img{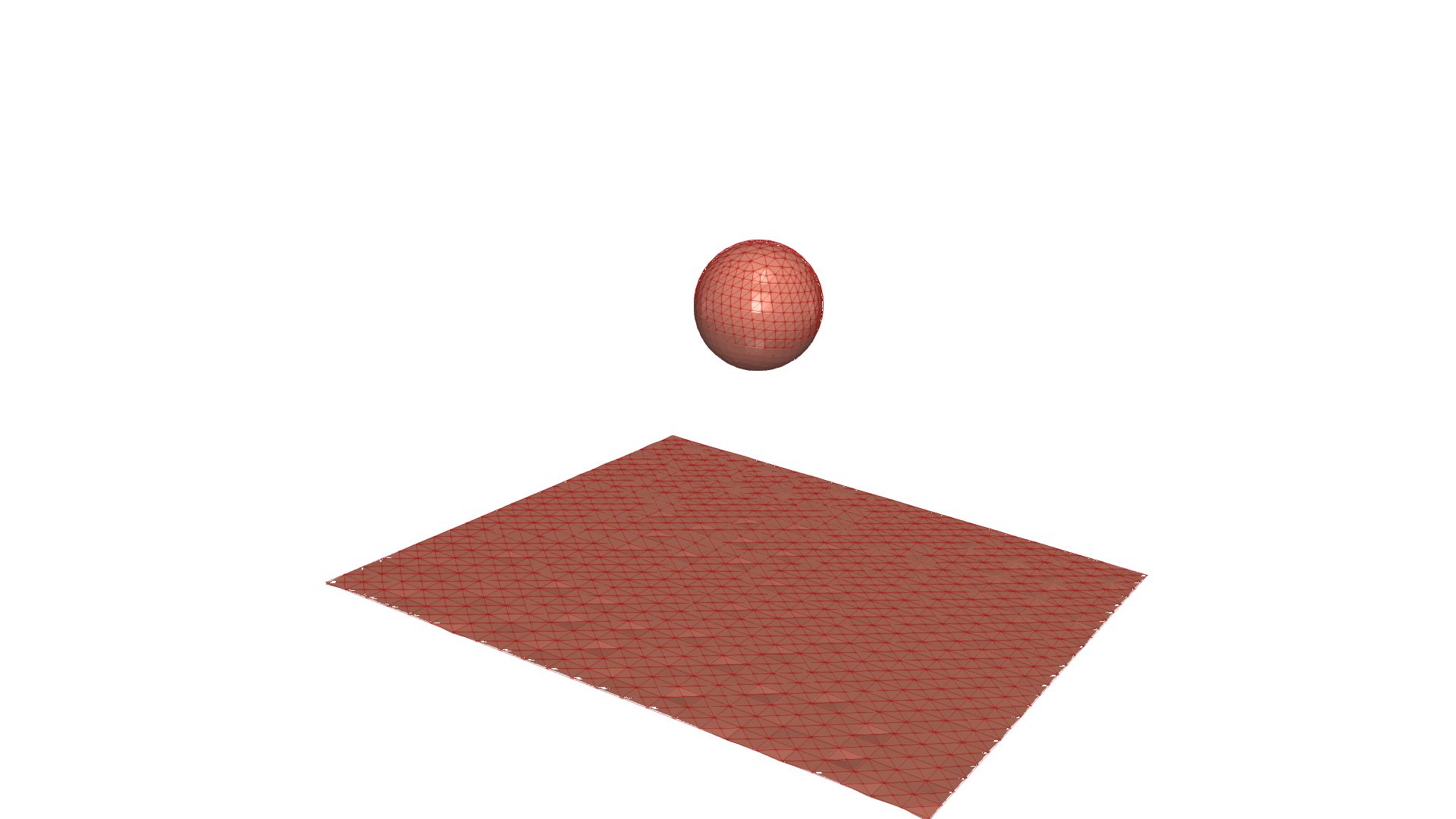}%
    \img{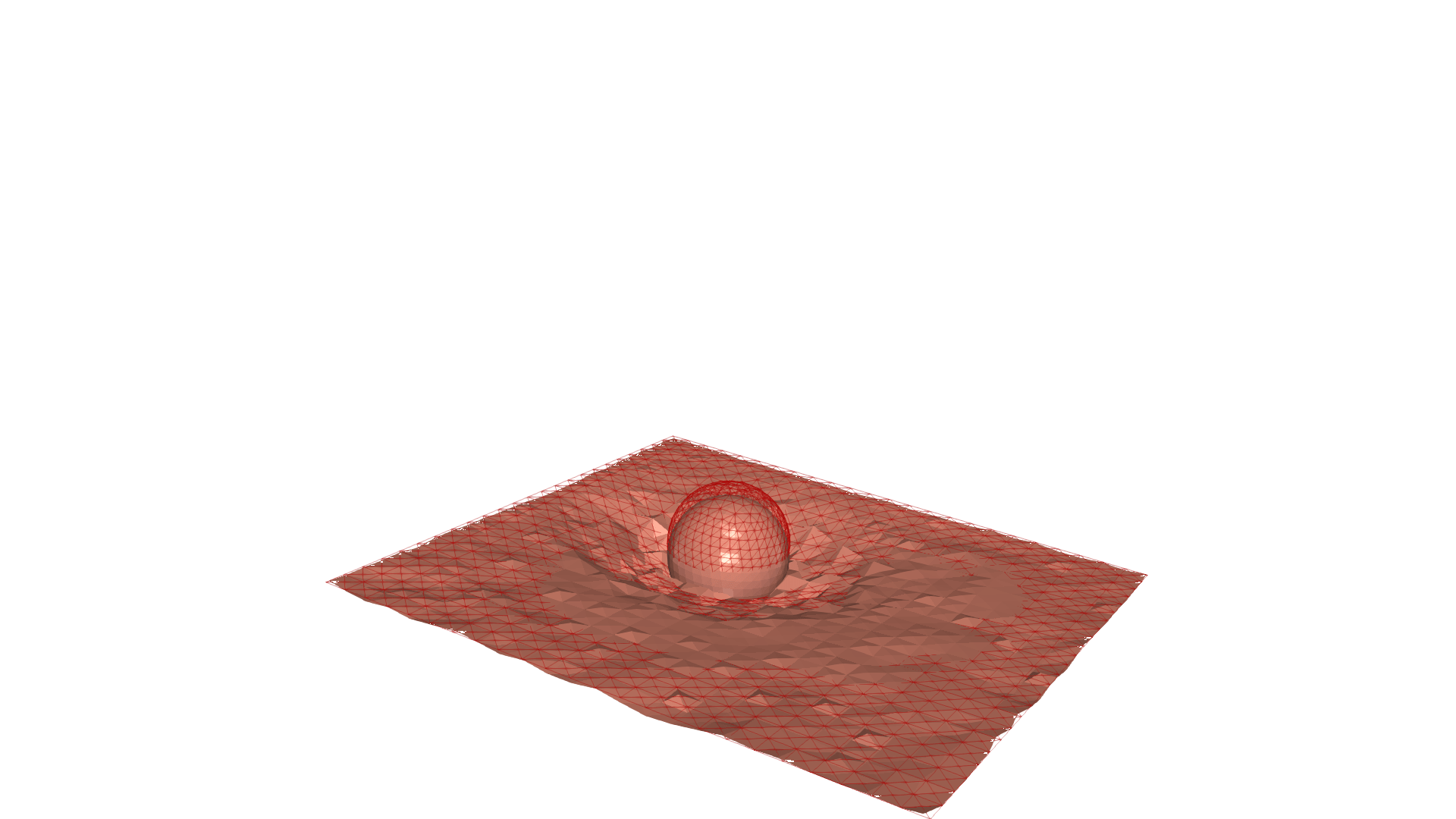}%
    \img{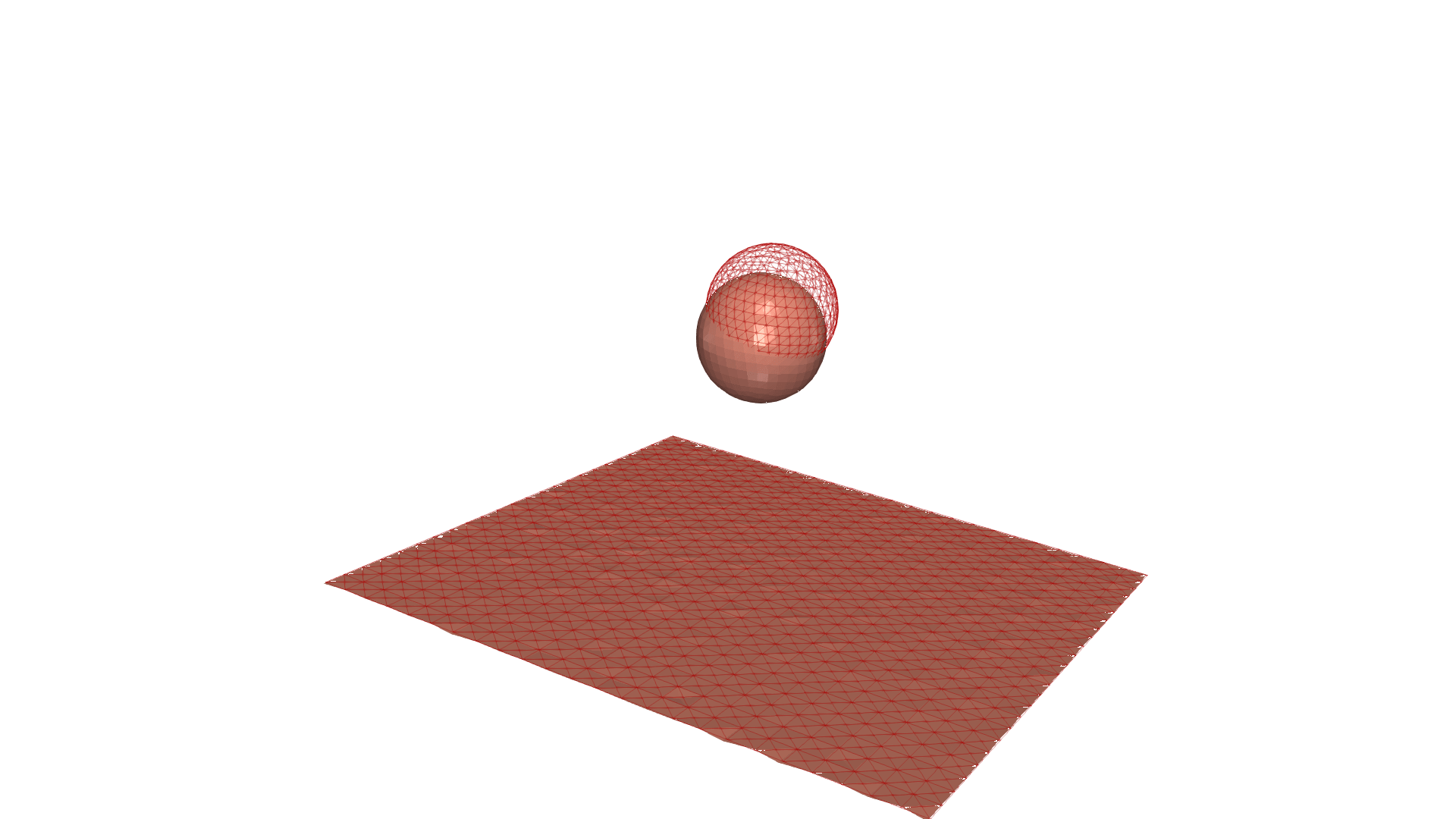}%
    \img{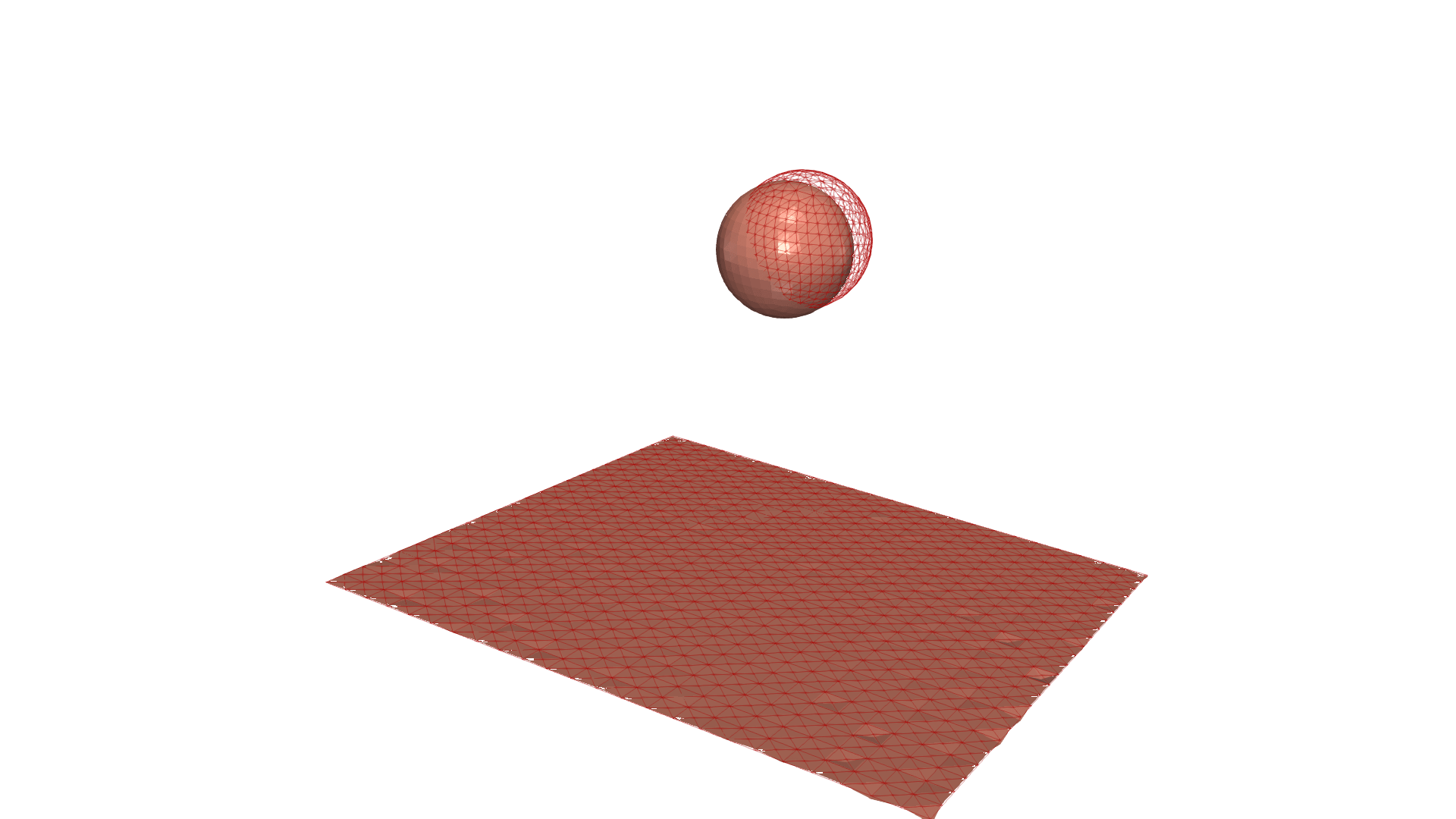}%
    \img{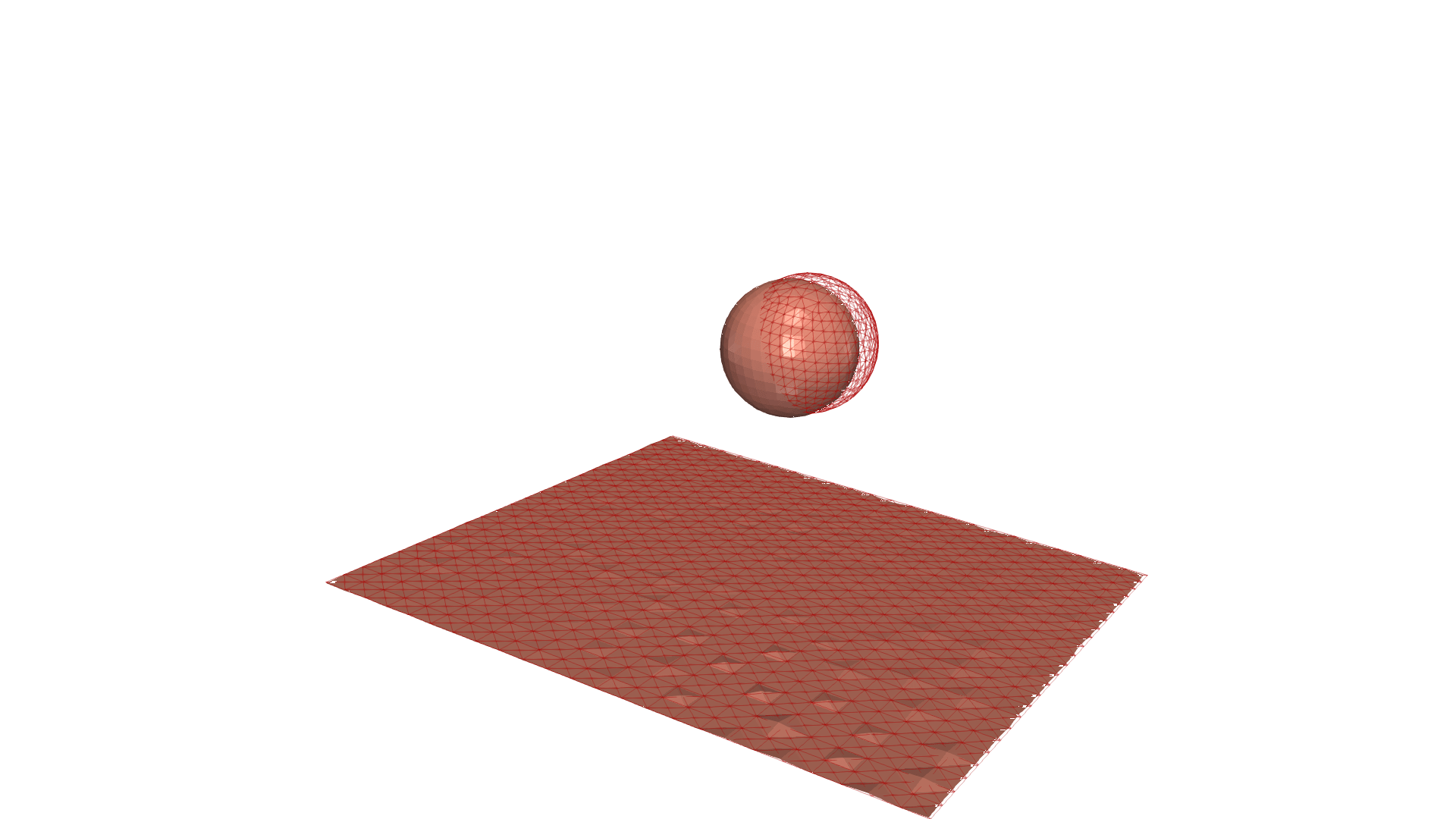}\\[0.15em]

    % Row 8: PSTNet Encoder
    \rowlabel{PSTNet \\ Encoder}%
    \img{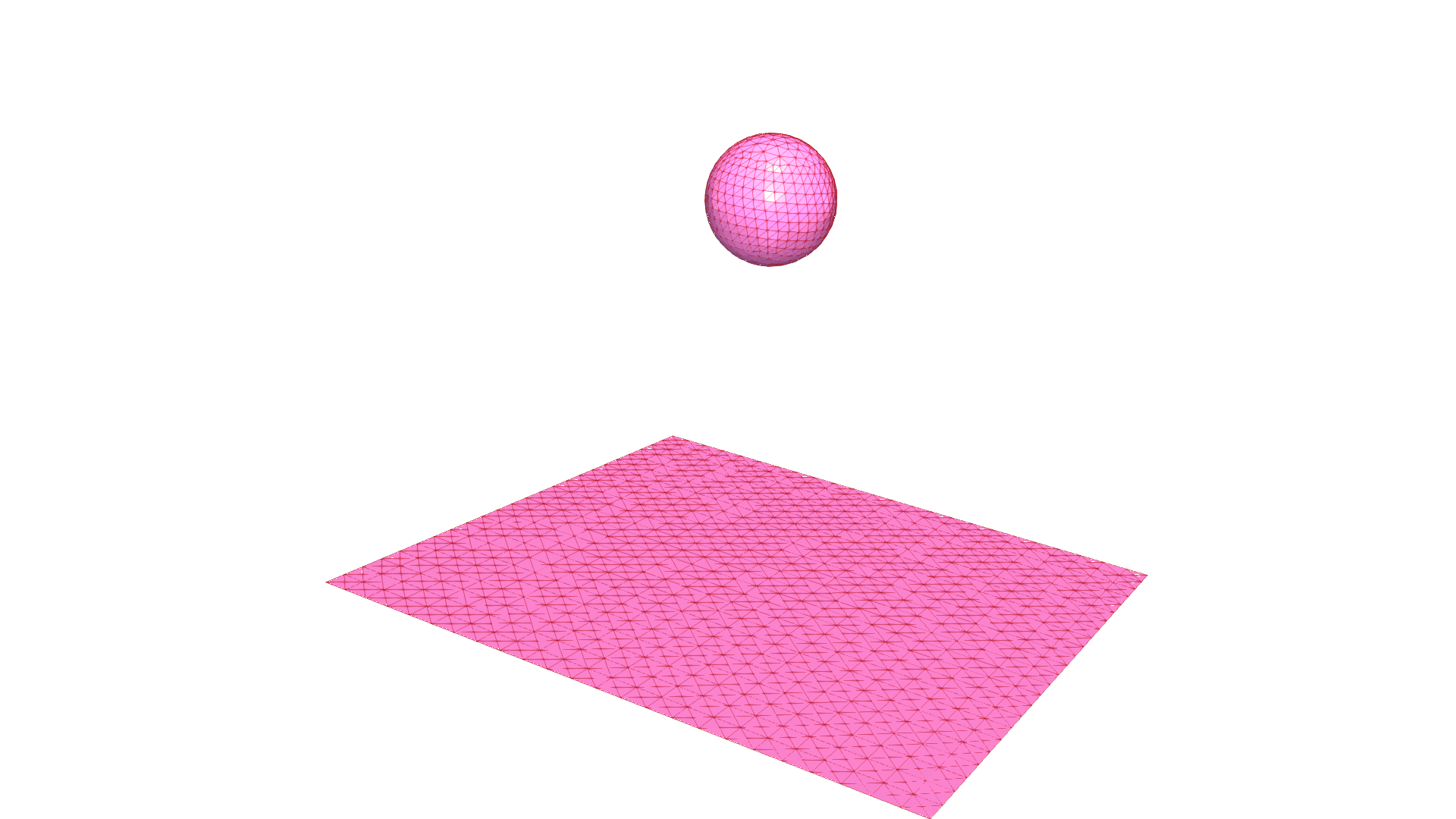}%
    \img{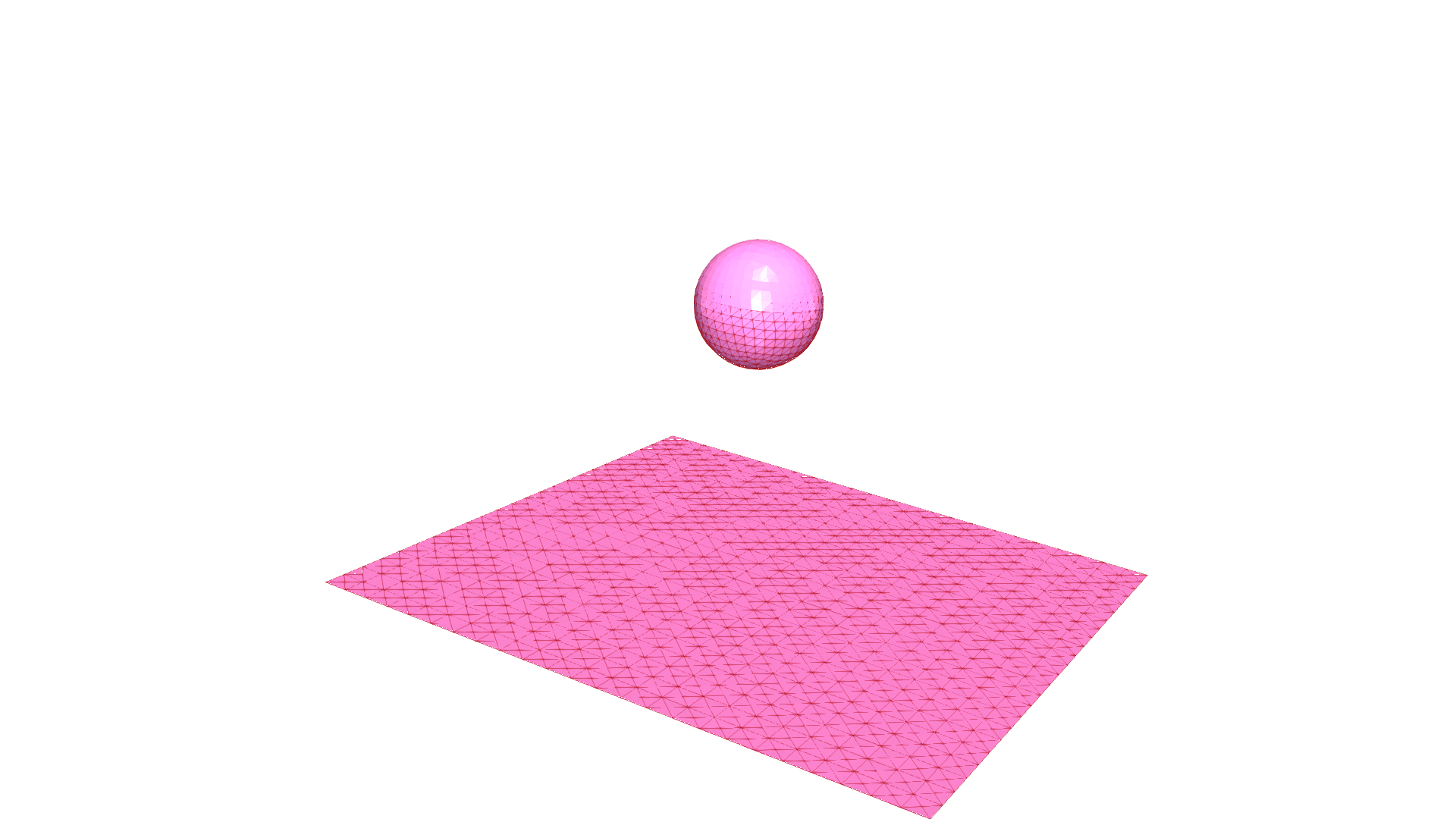}%
    \img{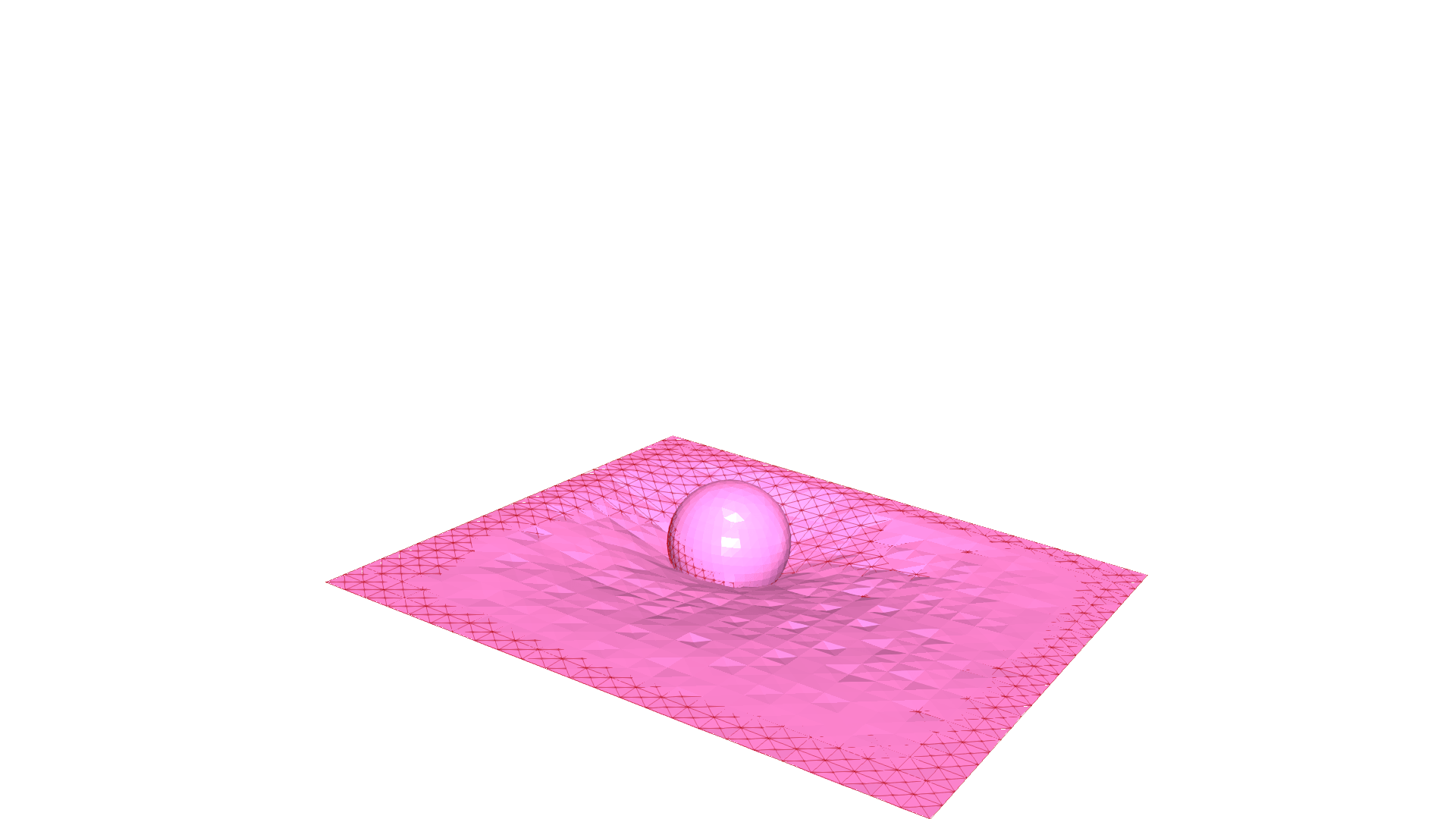}%
    \img{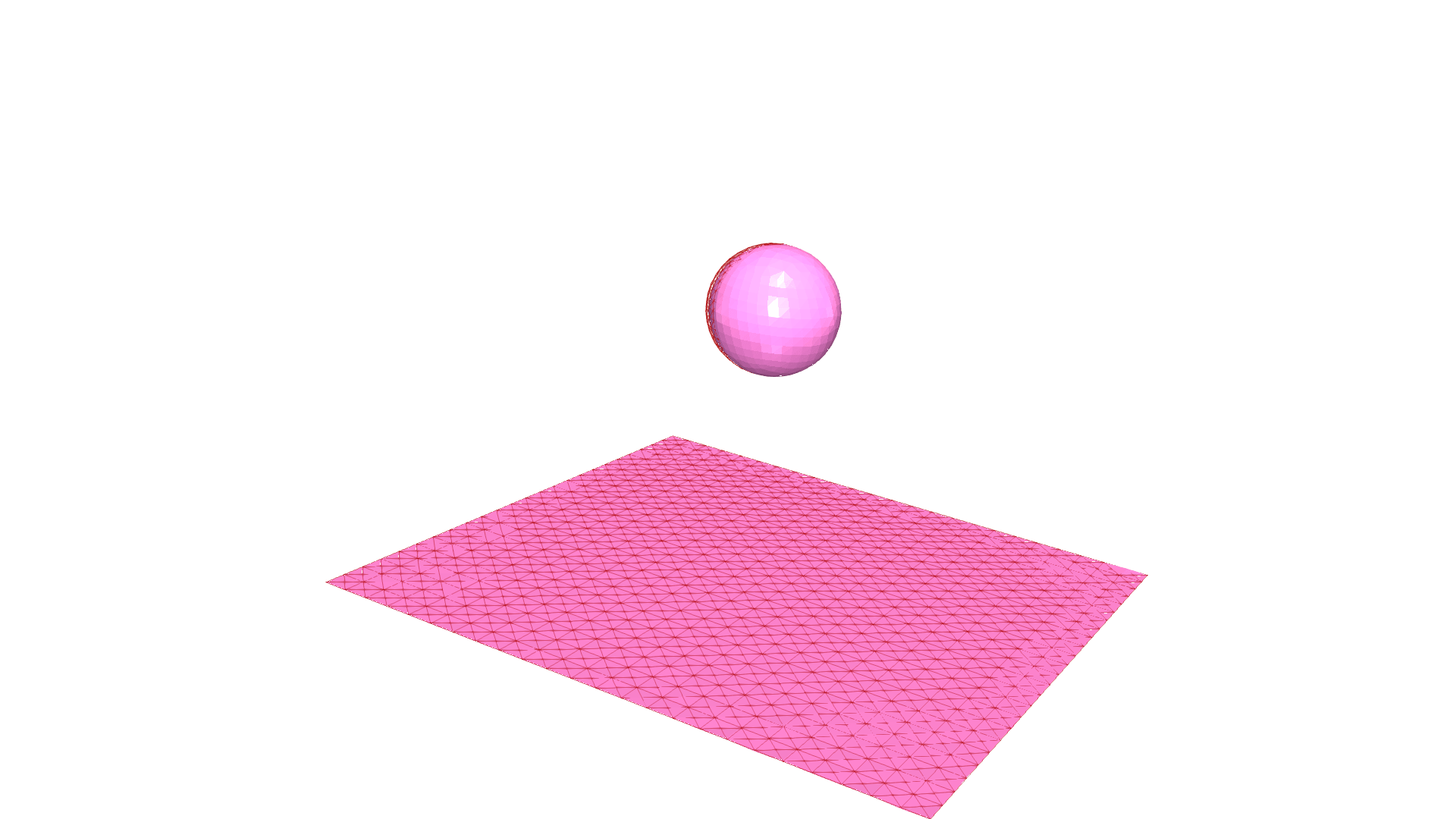}%
    \img{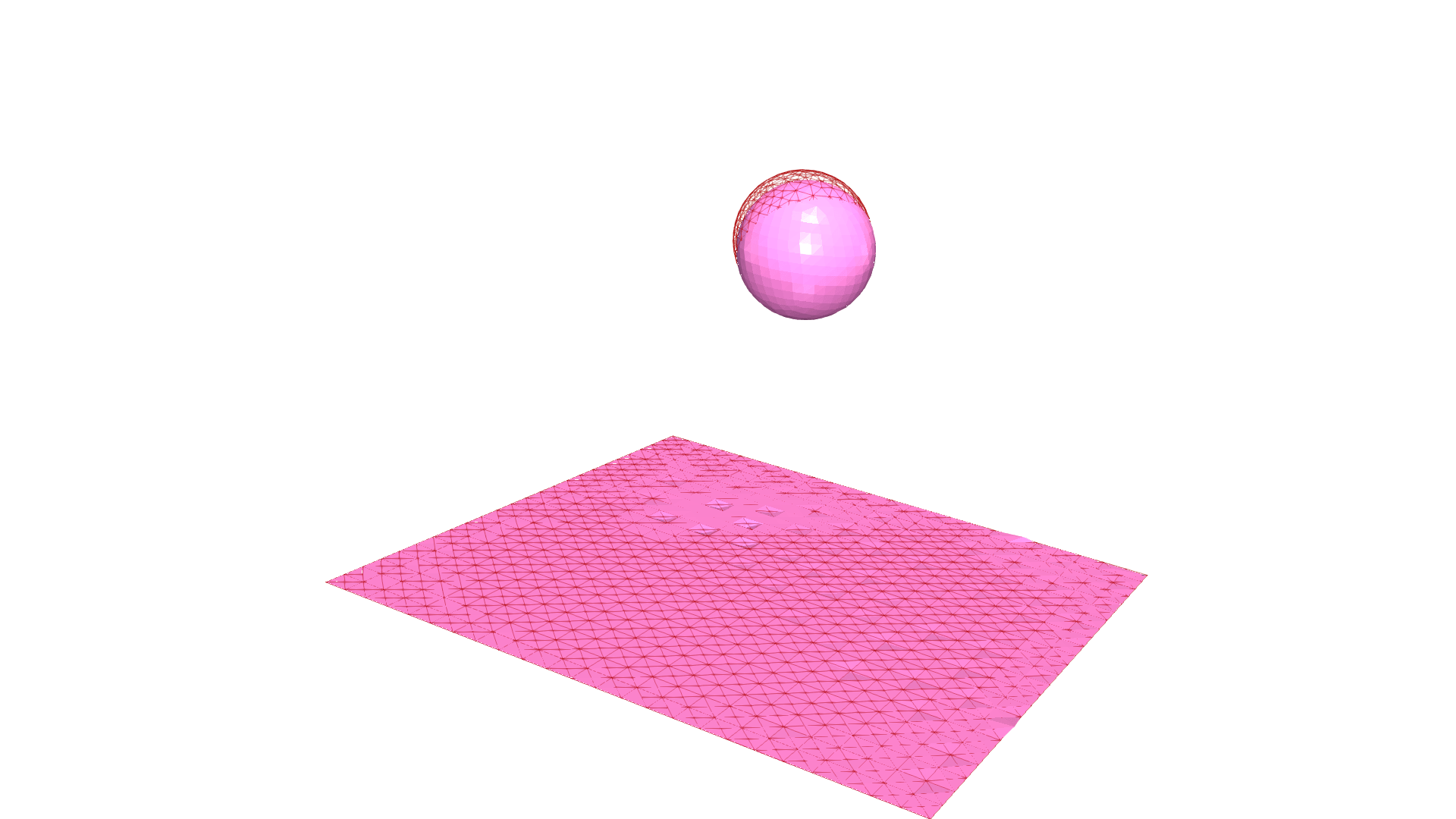}%
    \img{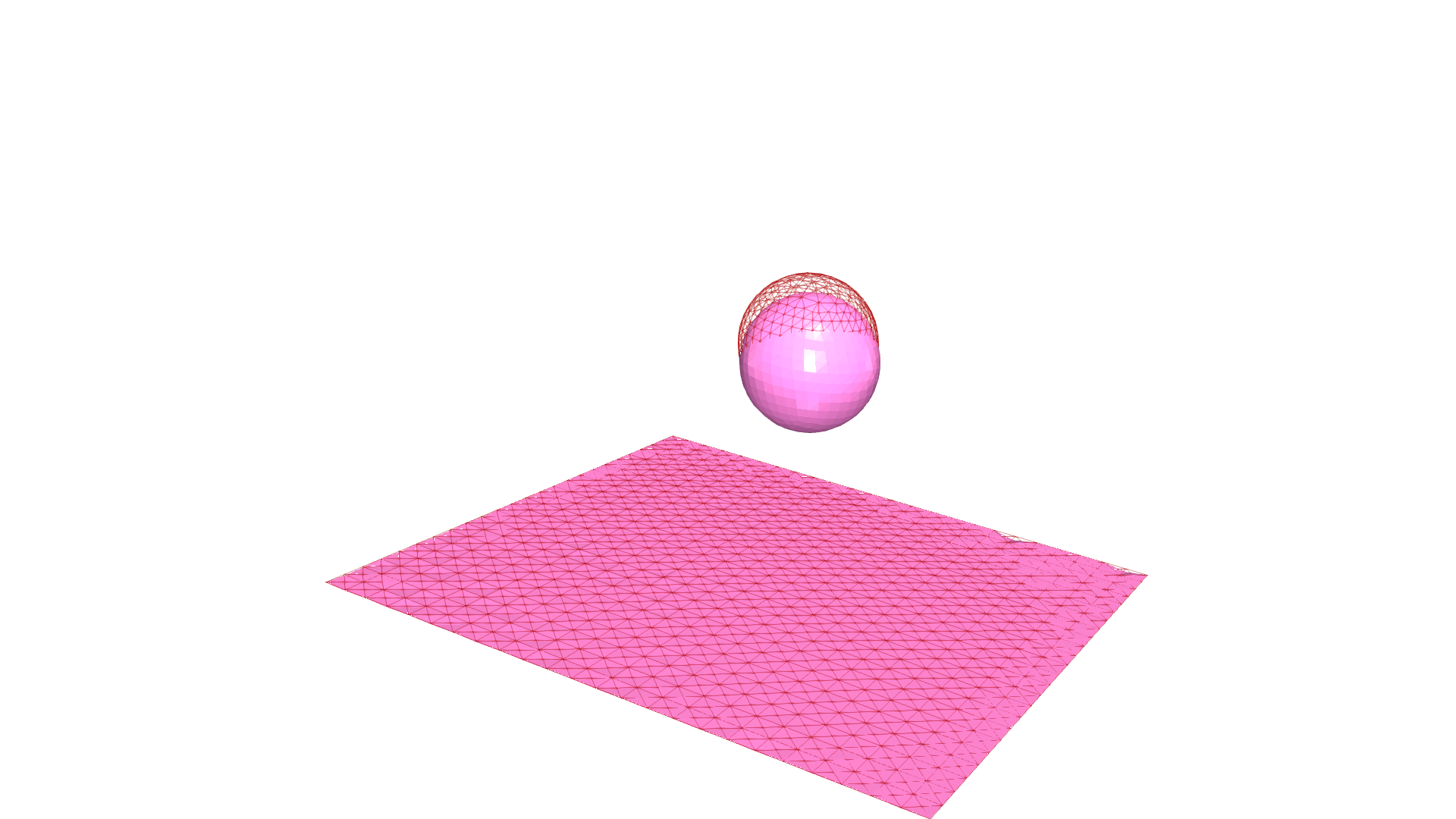}\\[0.15em]

    % Row 9: Pointcloud
    \rowlabel{Pointcloud}%
    \begin{minipage}[c]{0.155\textwidth}\centering
        \pointcloud{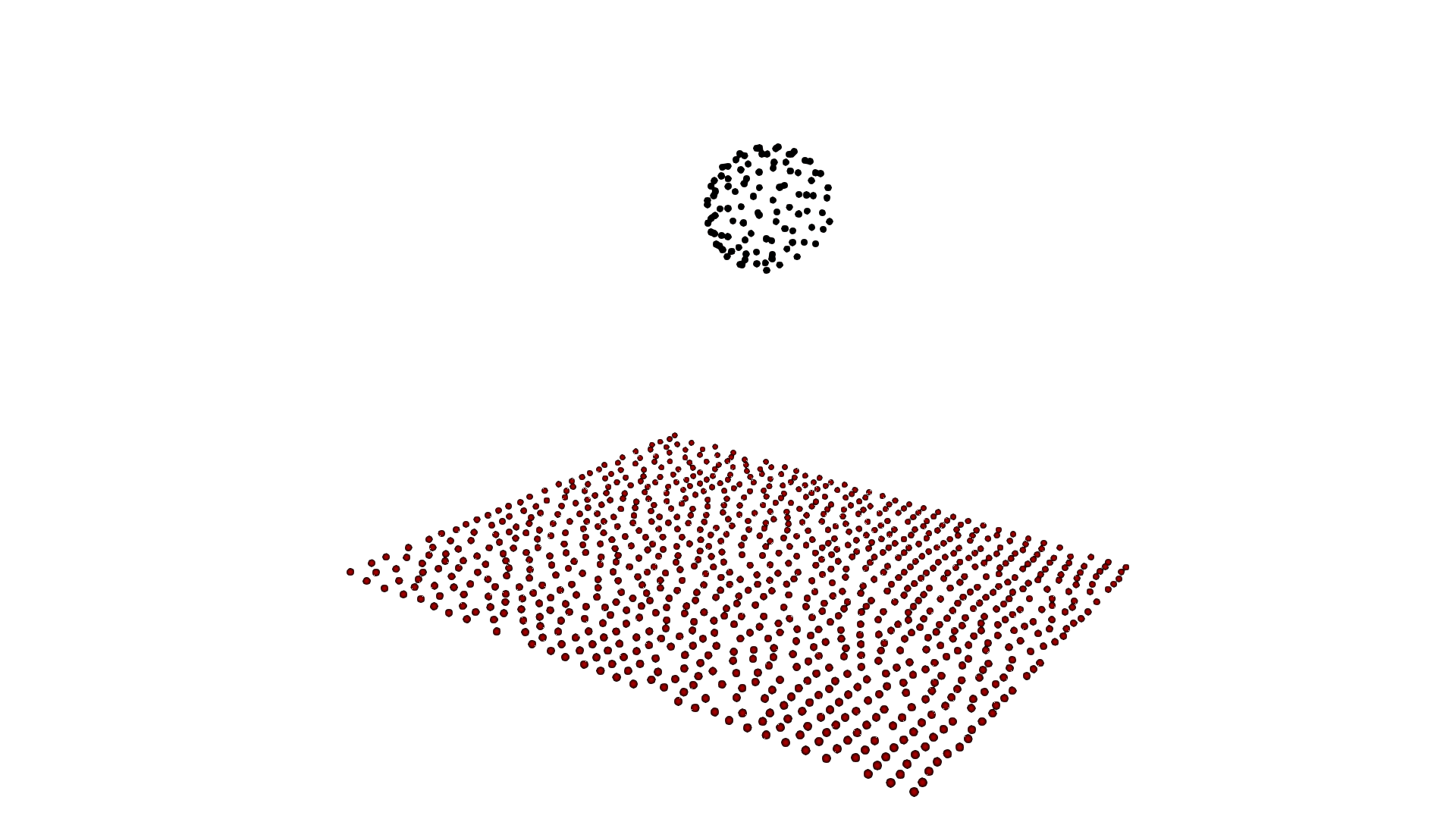}\\[2pt]
        \footnotesize$t{=}0$
    \end{minipage}%
    \begin{minipage}[c]{0.155\textwidth}\centering
        \pointcloud{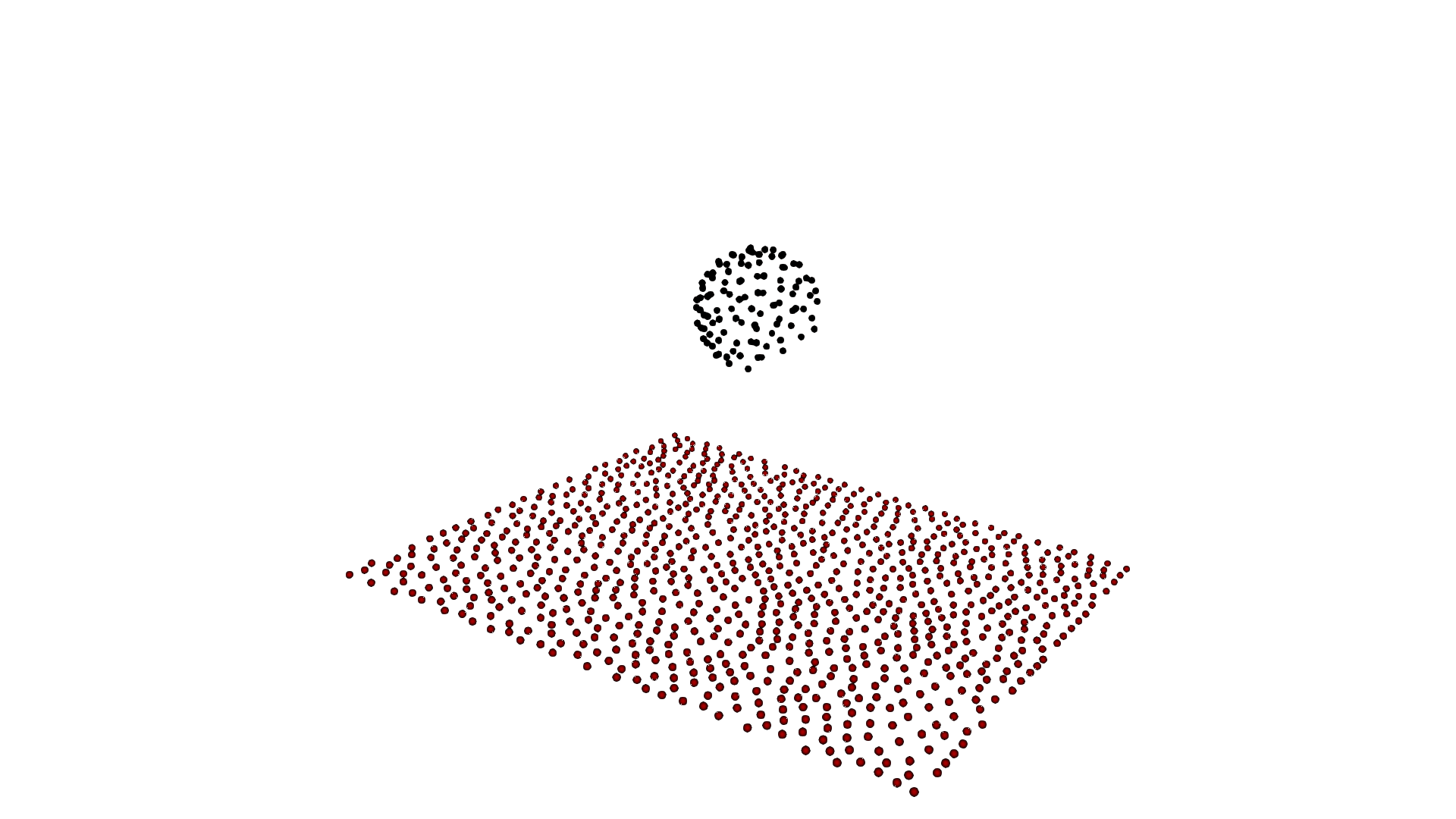}\\[2pt]
        \footnotesize$t{=}5$
    \end{minipage}%
    \begin{minipage}[c]{0.155\textwidth}\centering
        \pointcloud{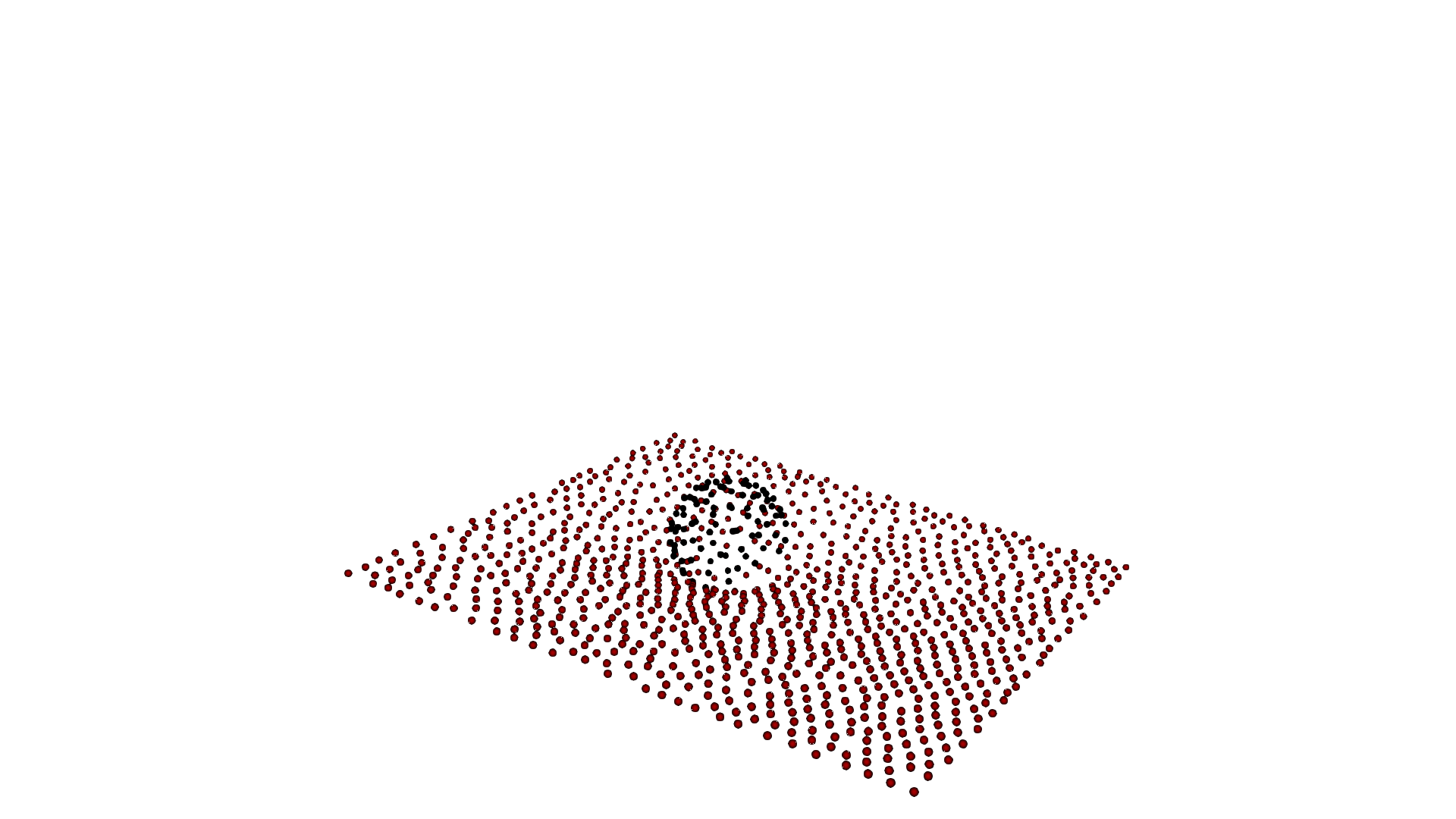}\\[2pt]
        \footnotesize$t{=}10$
    \end{minipage}%
    \begin{minipage}[c]{0.155\textwidth}\centering
        \pointcloud{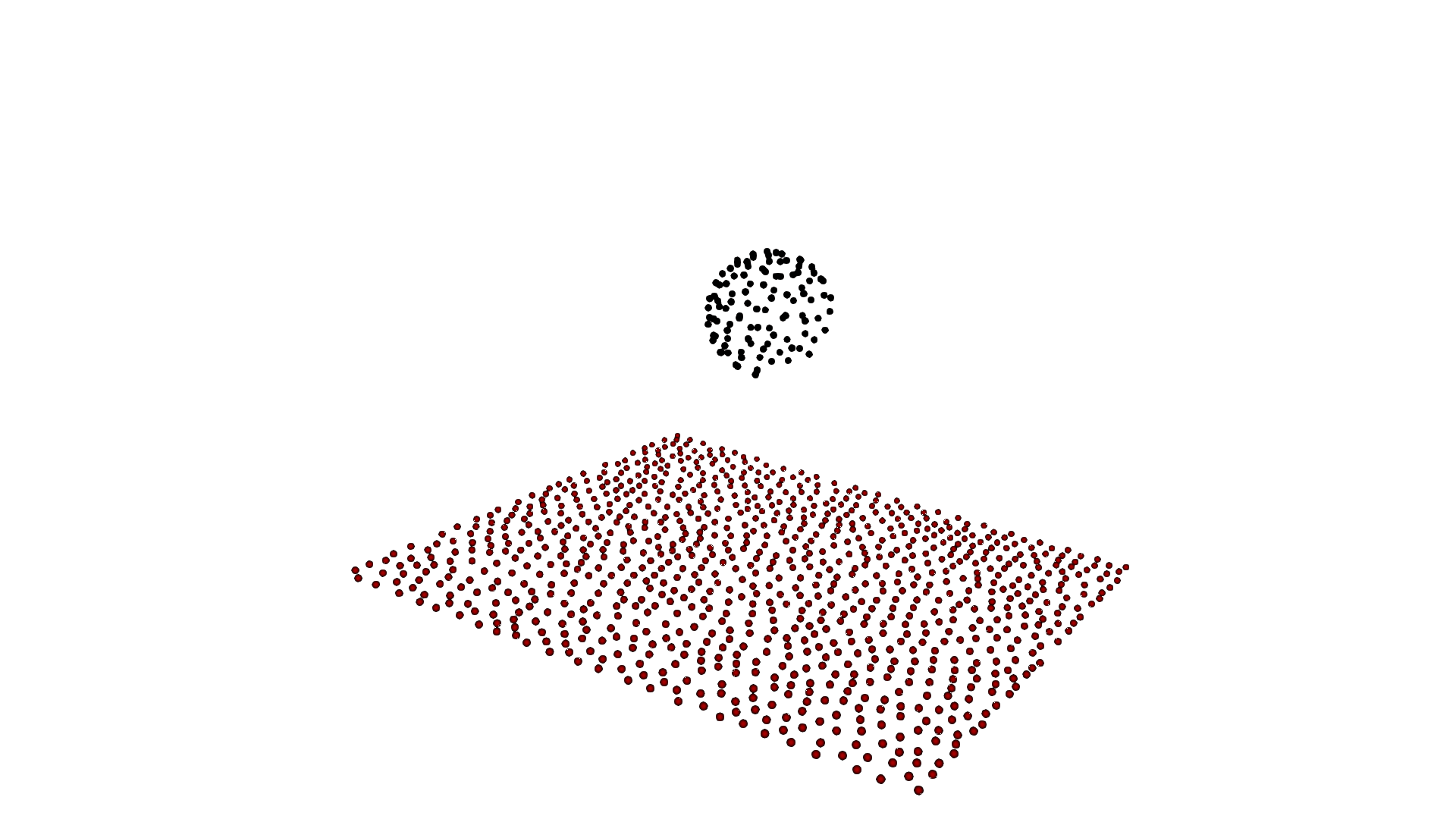}\\[2pt]
        \footnotesize$t{=}15$
    \end{minipage}%
    \begin{minipage}[c]{0.155\textwidth}\centering
        \pointcloud{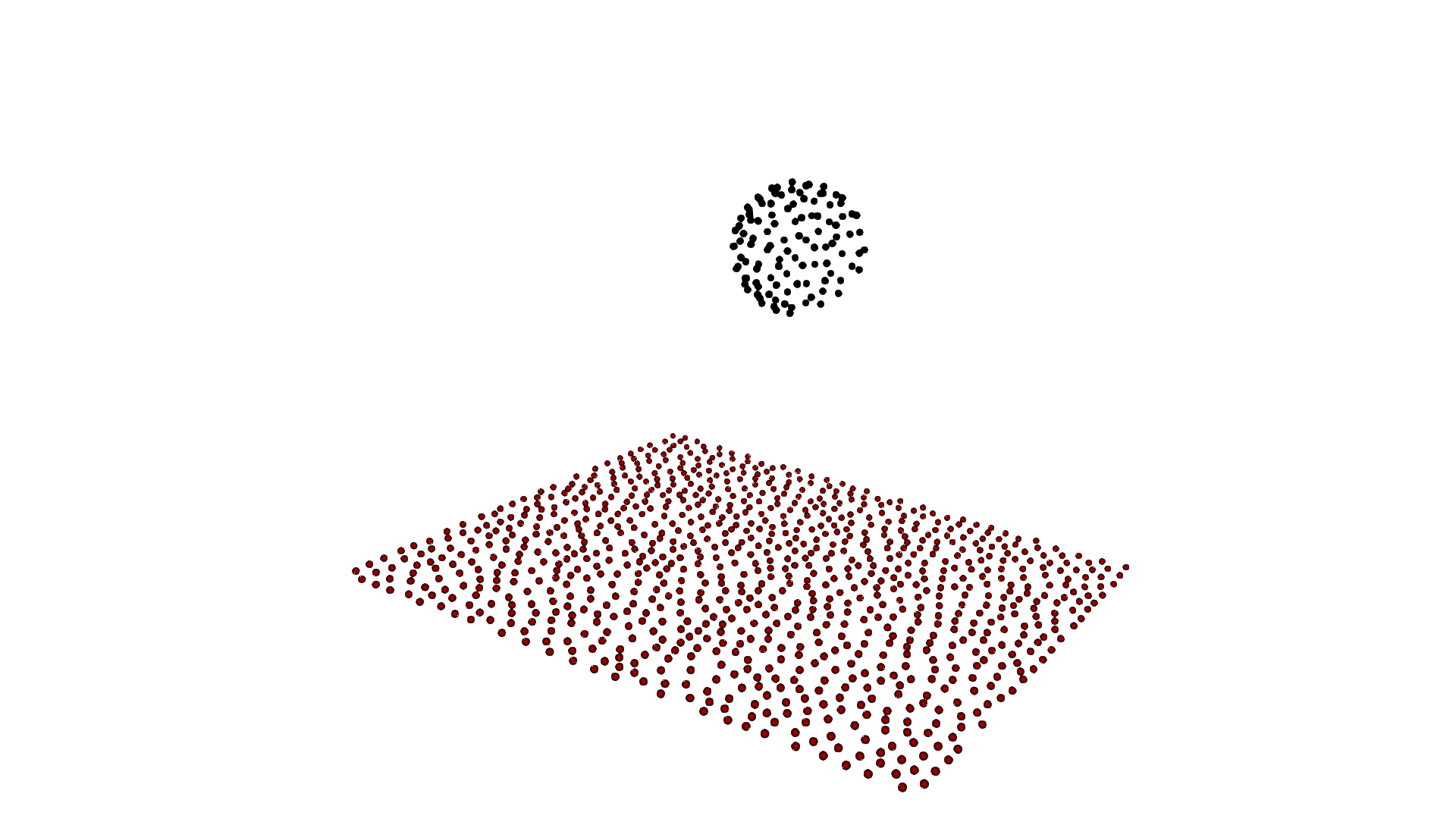}\\[2pt]
        \footnotesize$t{=}20$
    \end{minipage}%
    \begin{minipage}[c]{0.155\textwidth}\centering
        \pointcloud{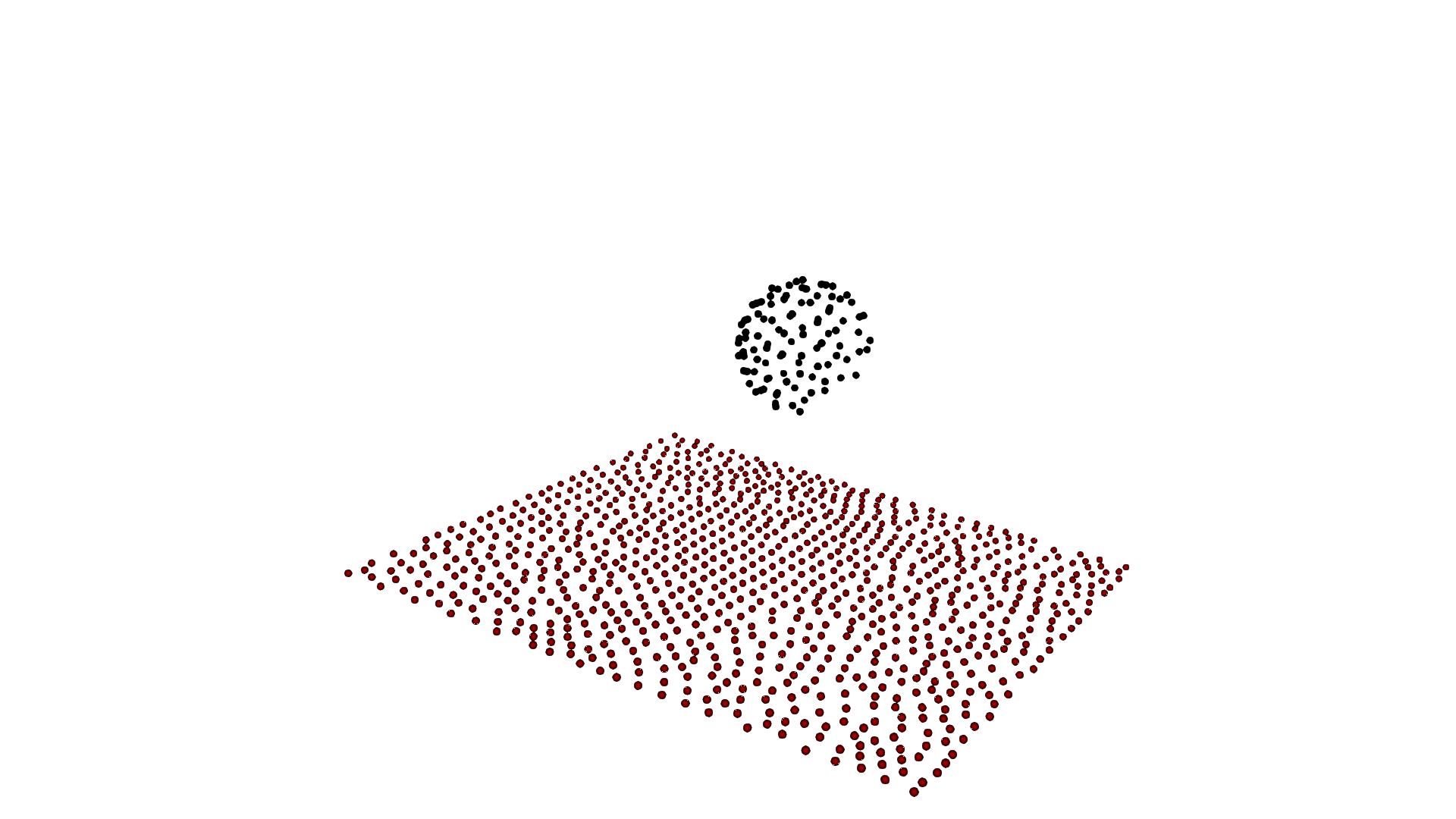}\\[2pt]
        \footnotesize$t{=}24$
    \end{minipage}

    \caption{
Predicted simulation of a \texttt{Trampoline} test task by
\textcolor{TabBlue}{PEACH} (blue),
\textcolor{TabGray}{No Context} (gray),
\textcolor{TabCyan}{MANGO} (cyan),
\textcolor{TabGreen}{Oracle} (green),
\textcolor{TabOrange}{No Context (MGN)} (orange),
\textcolor{TabPurple}{Oracle (MGN)} (purple),
\textcolor{TabBrown}{GNN Encoder} (brown), and
\textcolor{TabPink}{PSTNet Encoder} (pink).
All visualizations show the colored \textbf{predicted mesh} and a \textbf{\textcolor{red}{wireframe}} (red) of the ground-truth simulation. The last row shows an exemplary point cloud sequence from the context set.
    }
    \label{fig:qualitative_trajectories_trampoline}
\end{figure*}

\begin{figure*}[p]
    \centering
    {\Large \textbf{Bending Beam}}\\[1em]
    \vfill

    % helper macros (adjust trim once here)
    \newcommand{\rowlabel}[1]{%
        \begin{minipage}[c]{0.05\textwidth}\centering\rotatebox{90}{\scriptsize\textbf{\shortstack{#1}}}\end{minipage}%
    }
    \newcommand{\img}[1]{%
        \includegraphics[width=0.158\textwidth, valign=m]{#1}%
    }
    \newcommand{\pointcloud}[1]{%
        \includegraphics[width=\textwidth, trim=10cm 6cm 10cm 6cm, clip, valign=m]{#1}%
    }

    % Row 1: PEACH
    \rowlabel{PEACH}%
    \img{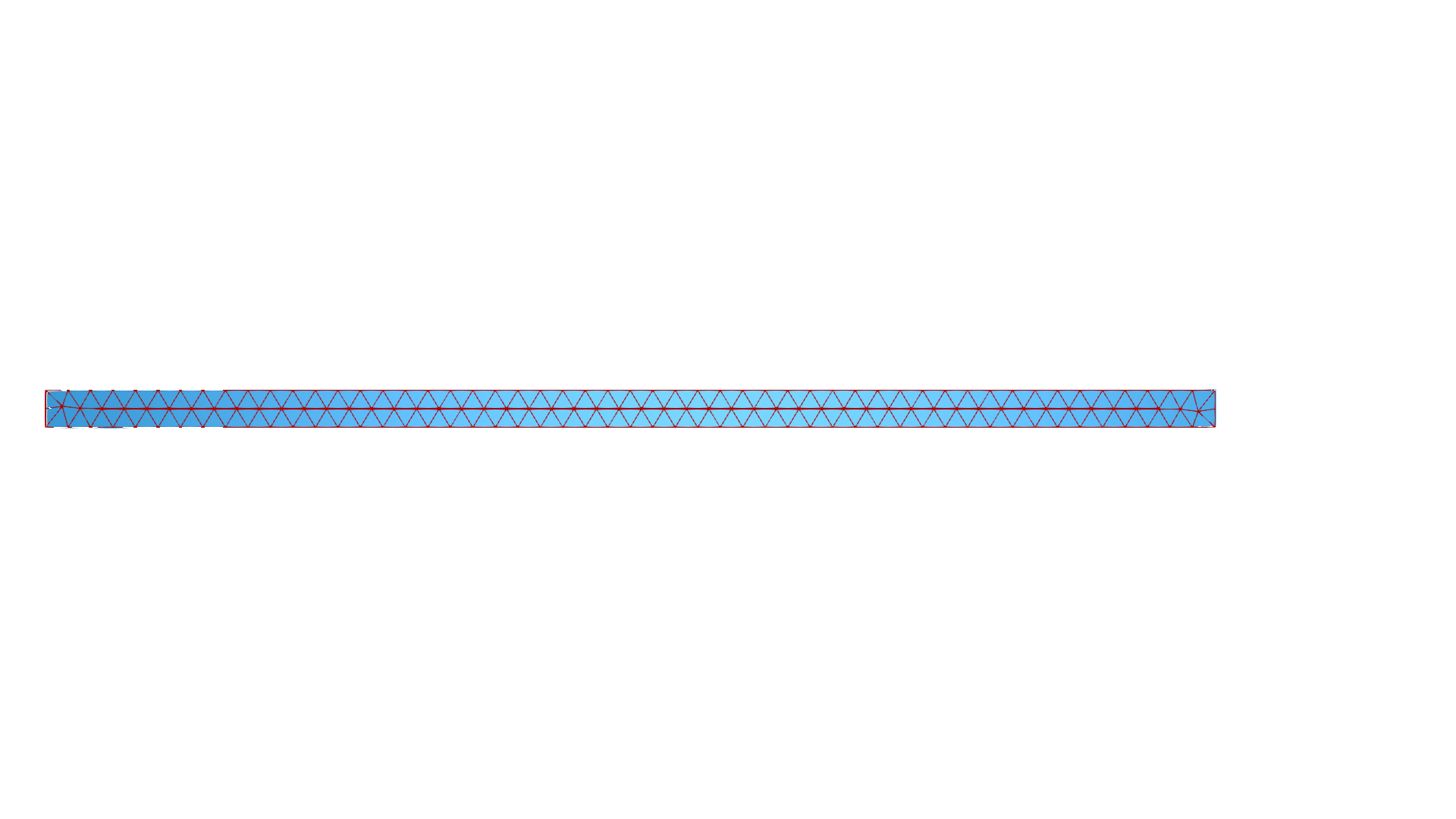}%
    \img{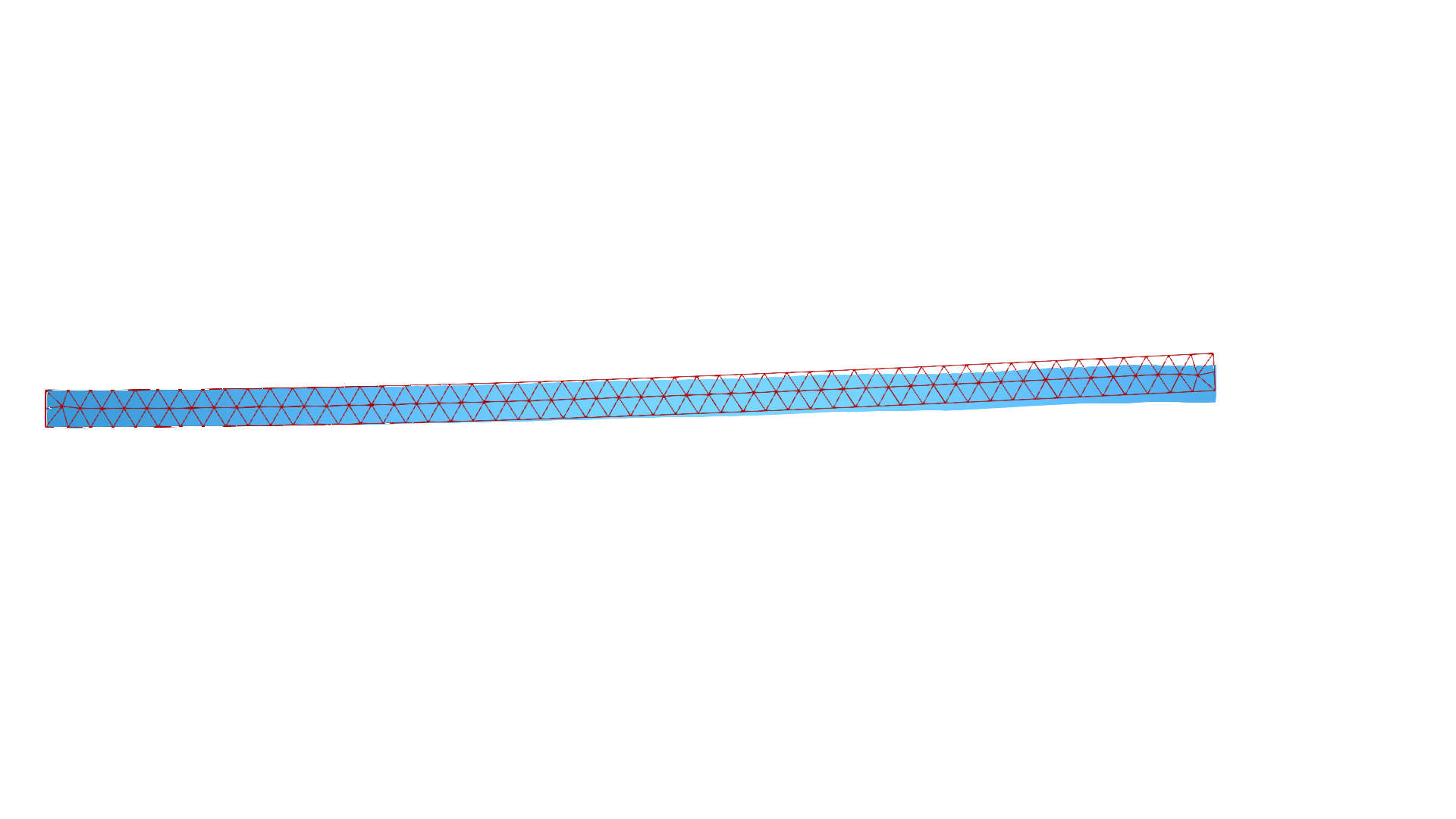}%
    \img{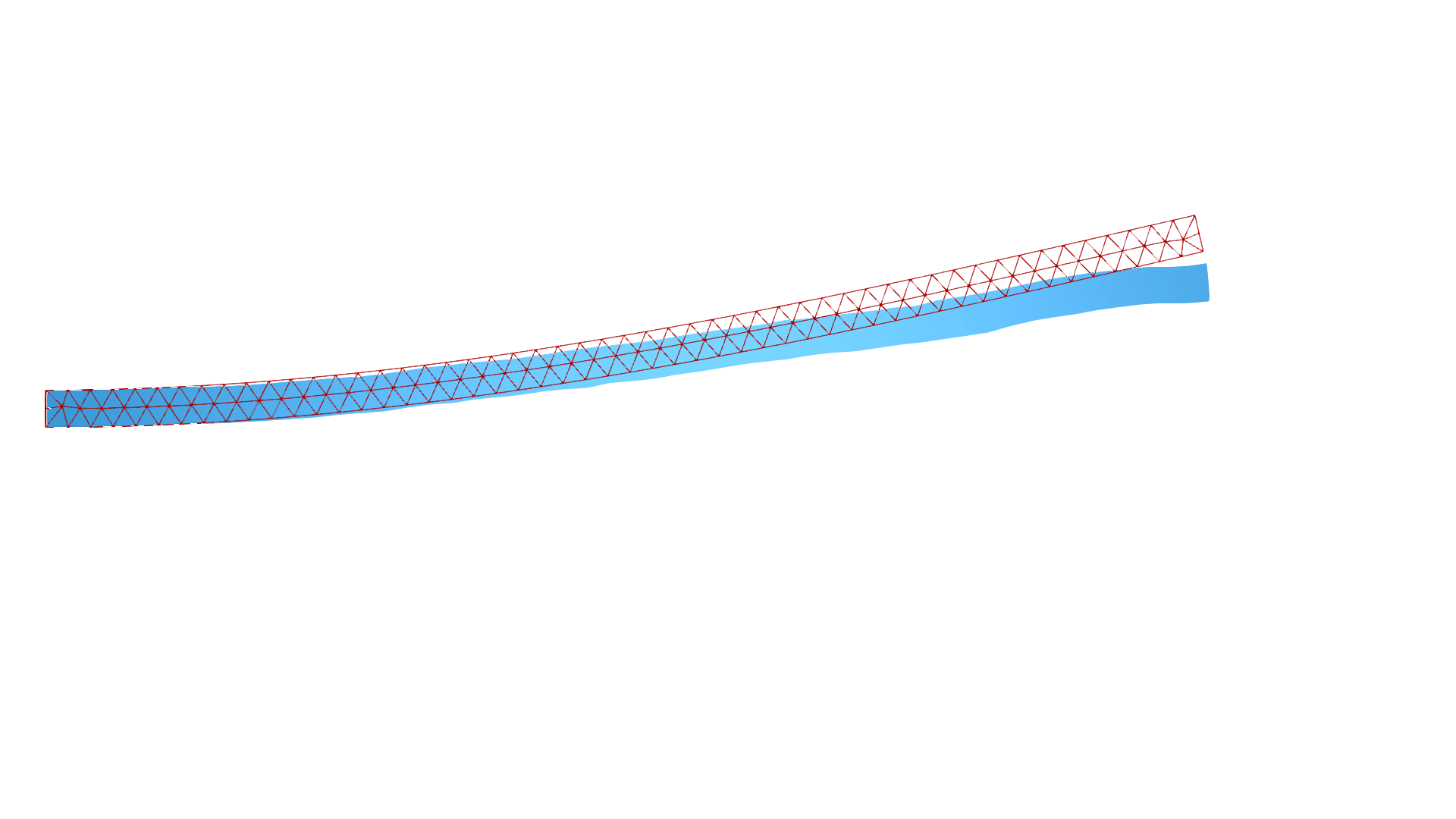}%
    \img{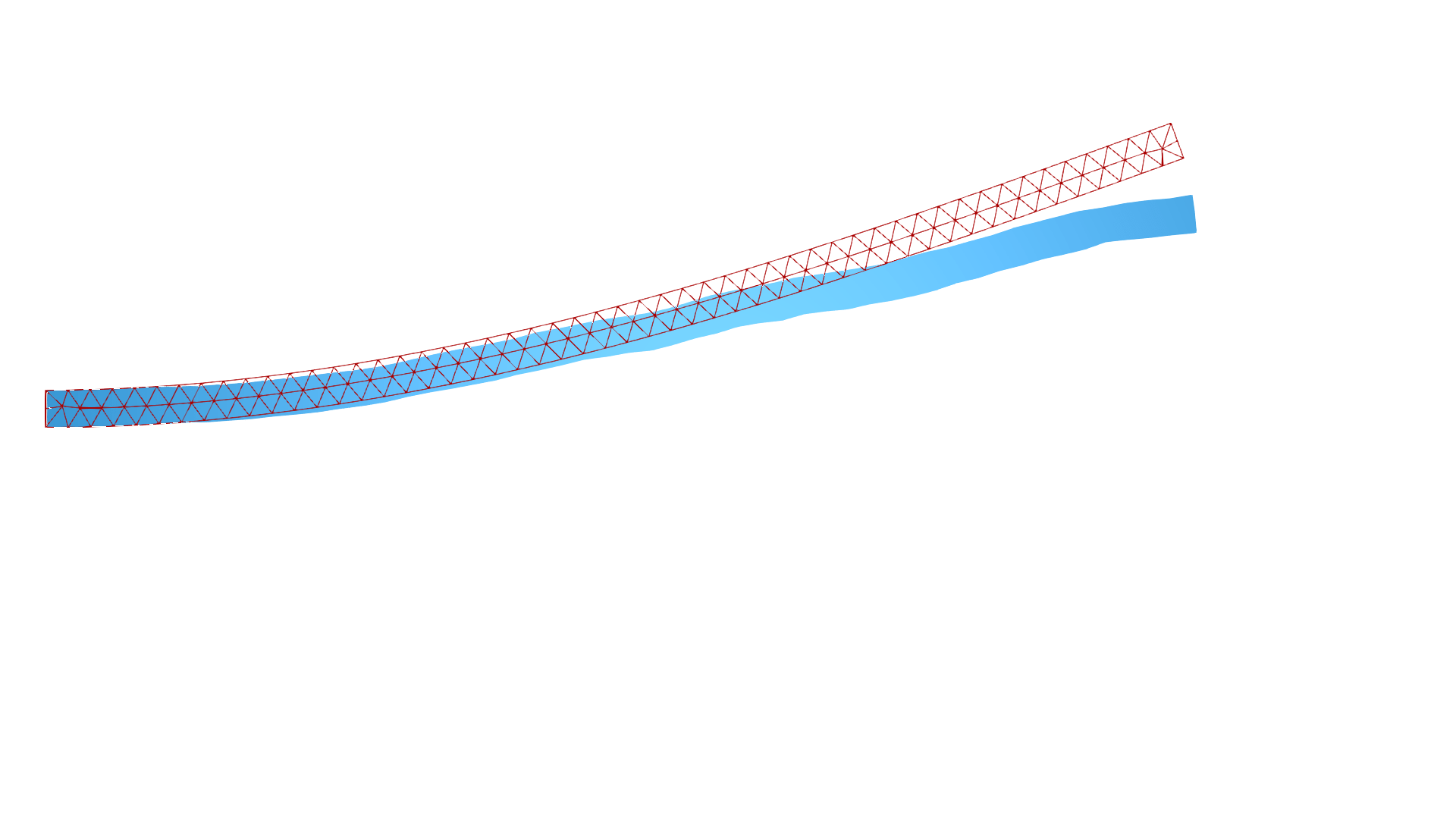}%
    \img{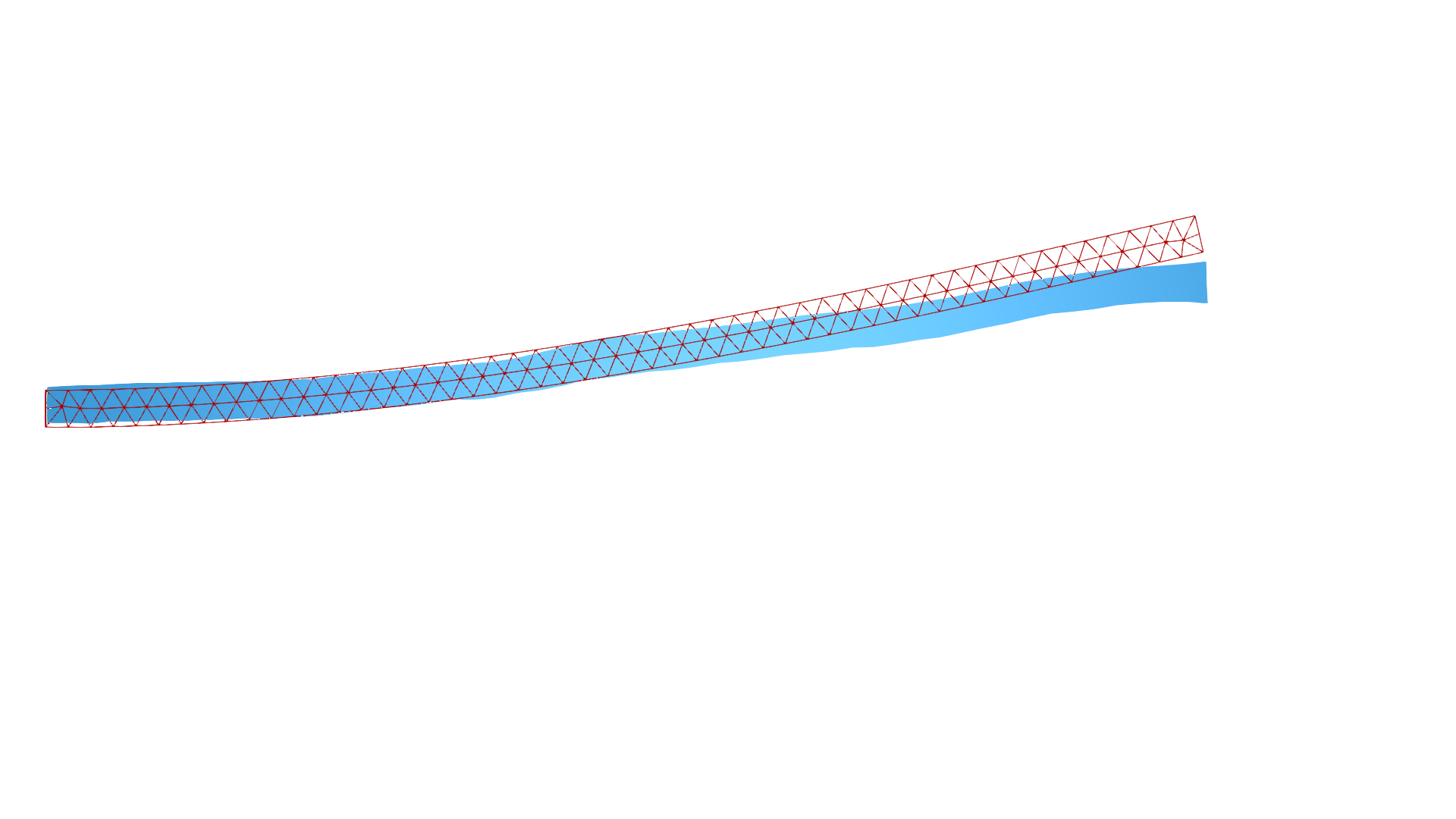}%
    \img{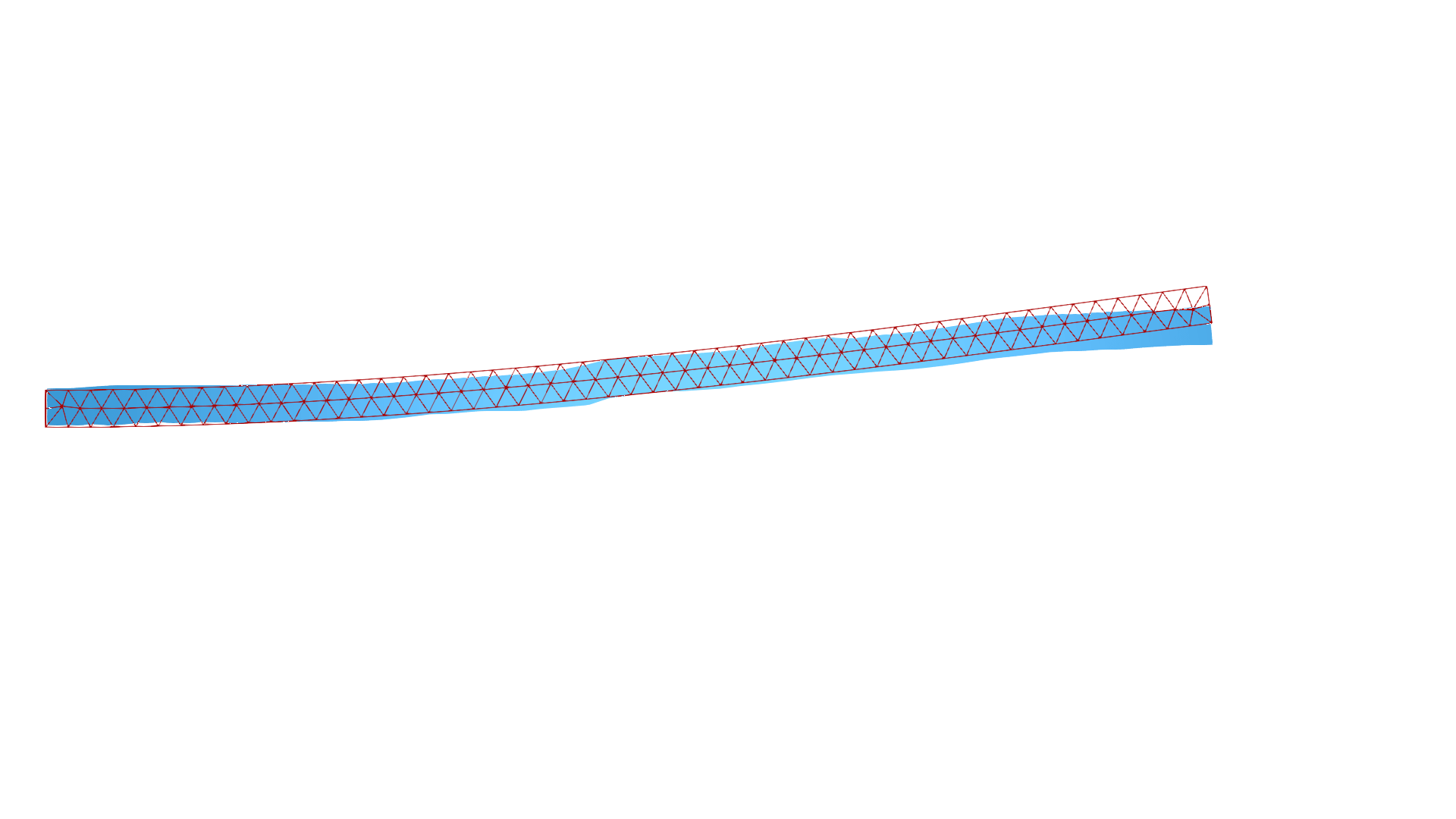}\\[0.6em]

    % Row 2: No Context
    \rowlabel{No Context}%
    \img{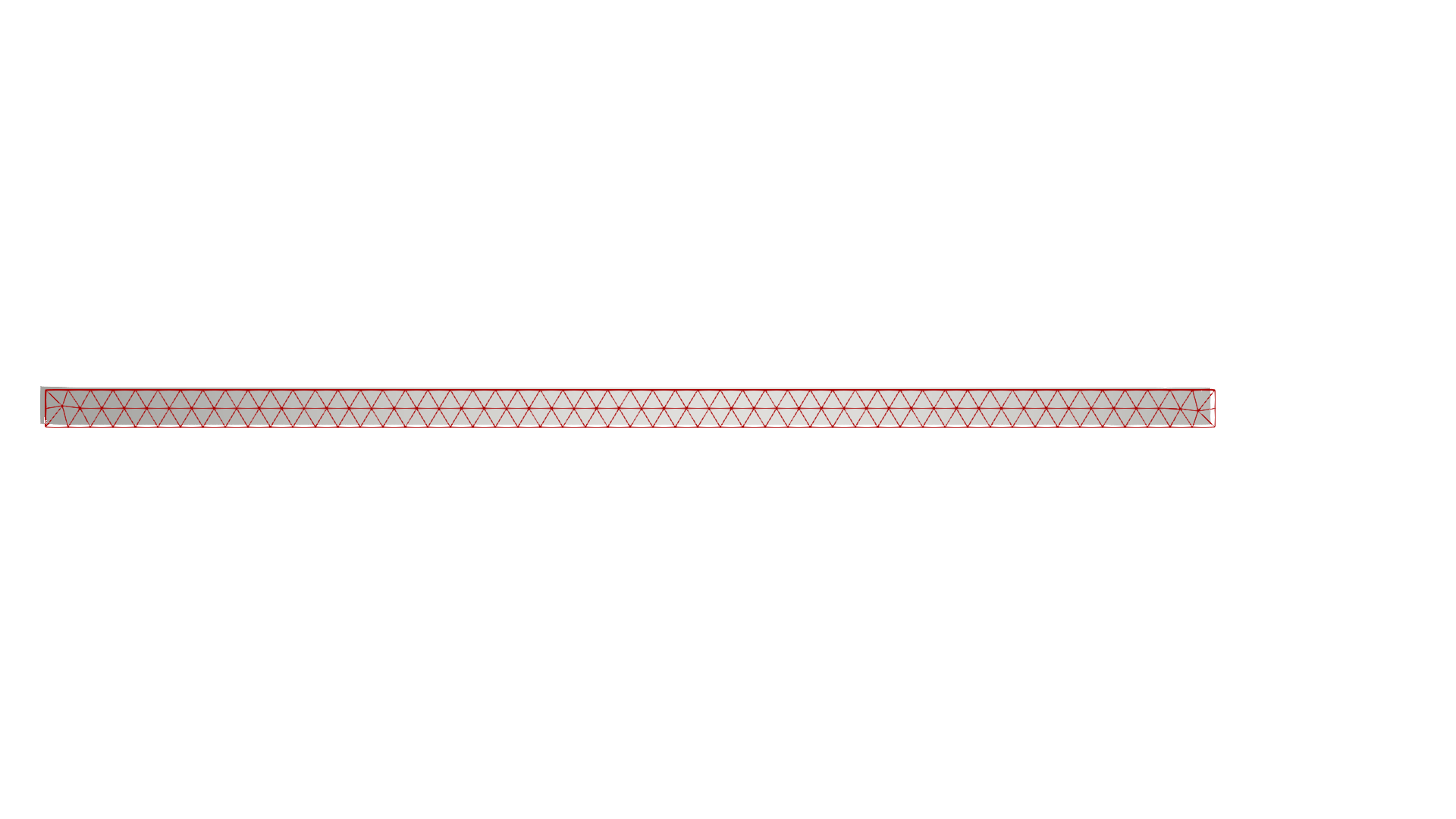}%
    \img{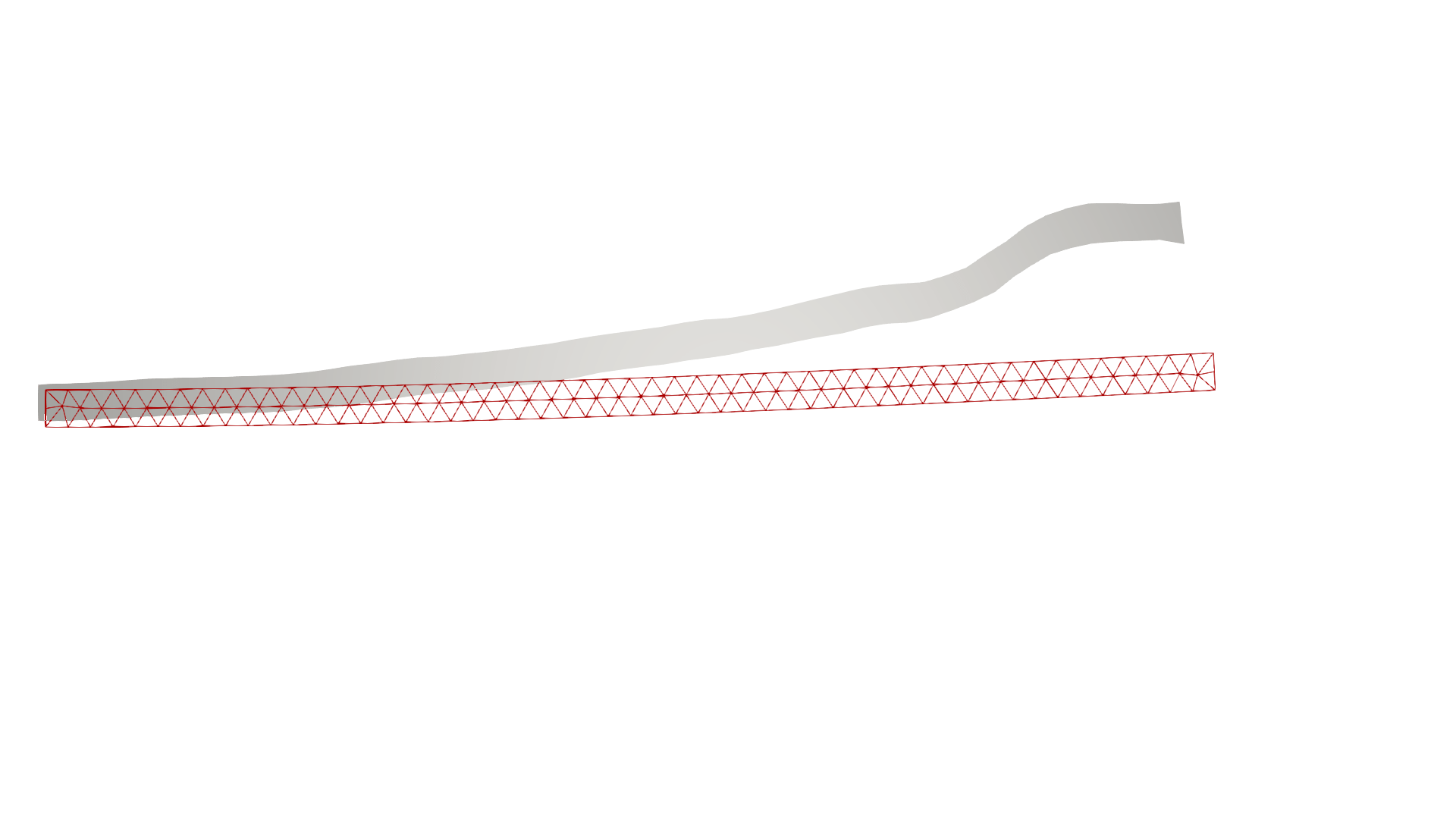}%
    \img{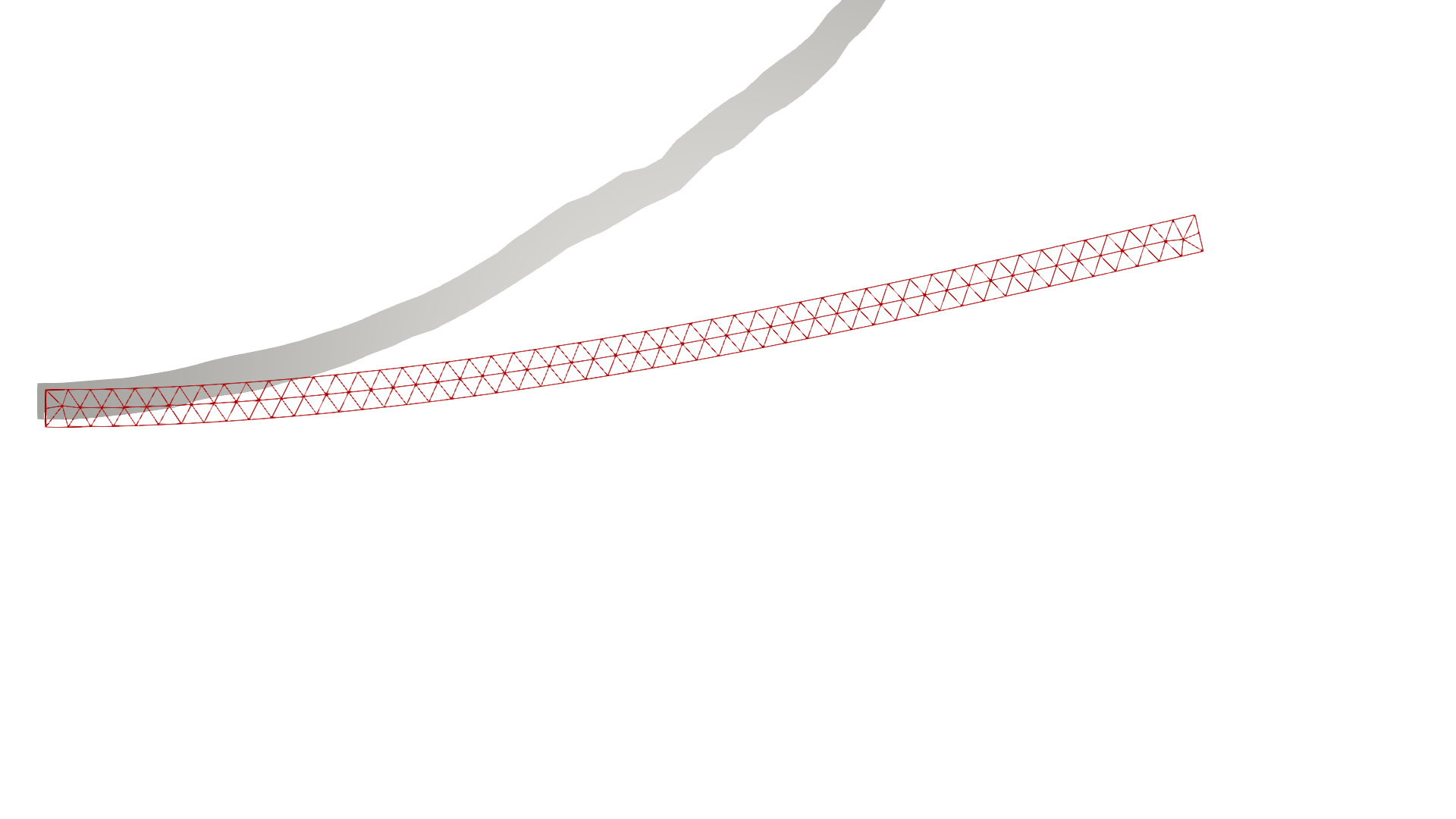}%
    \img{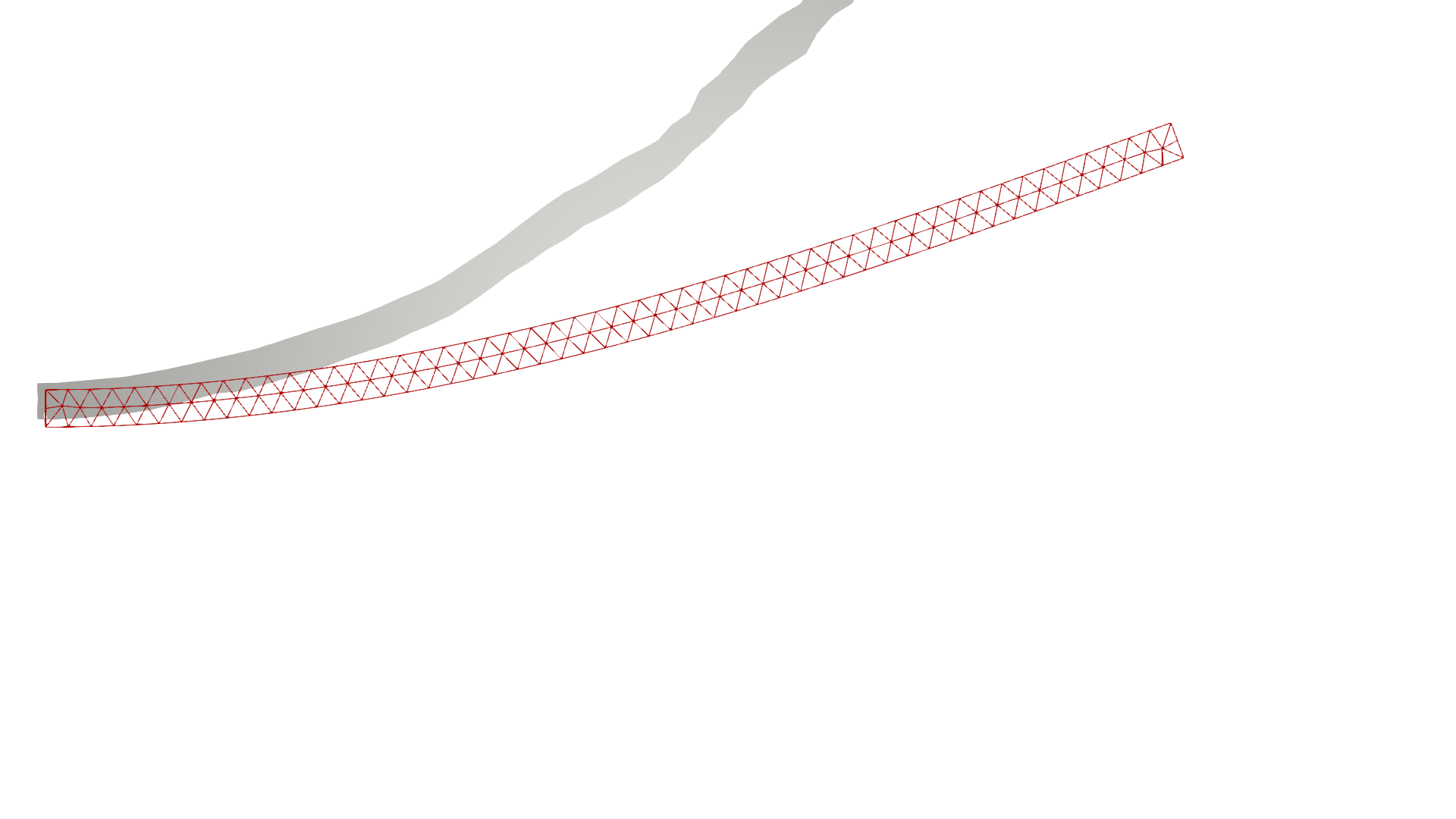}%
    \img{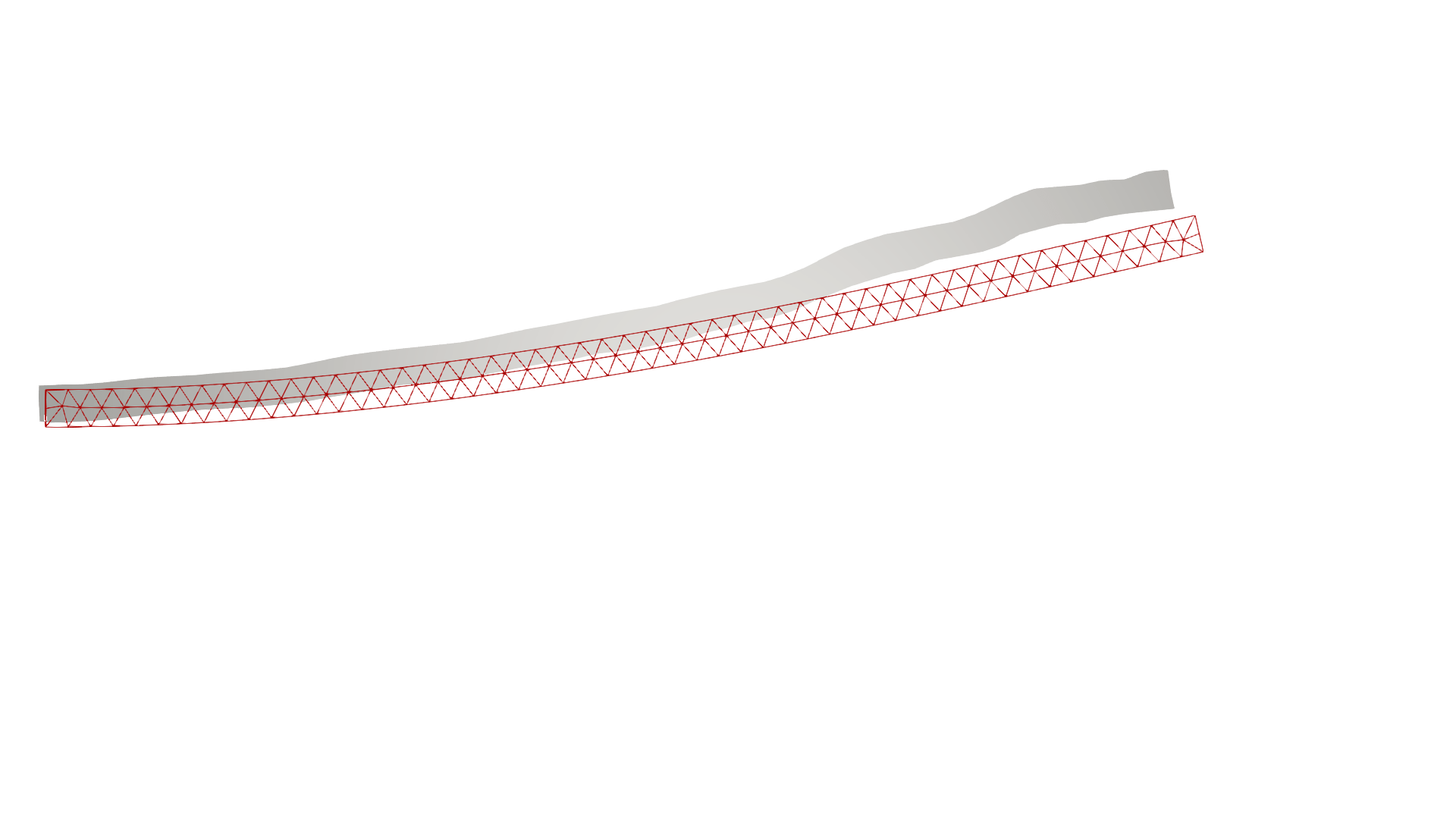}%
    \img{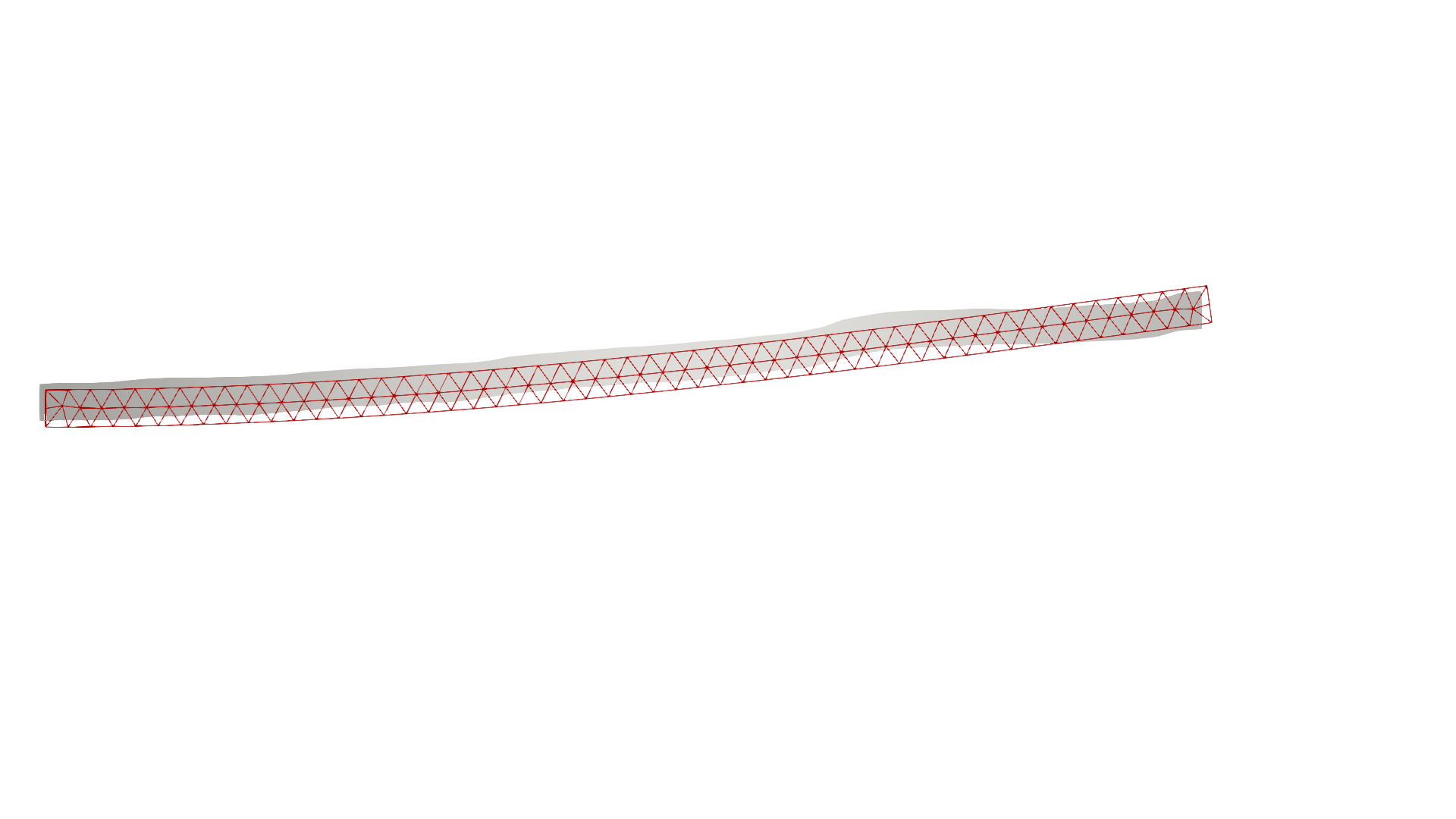}\\[0.6em]

    % Row 3: MANGO Decoder
    \rowlabel{MANGO}%
    \img{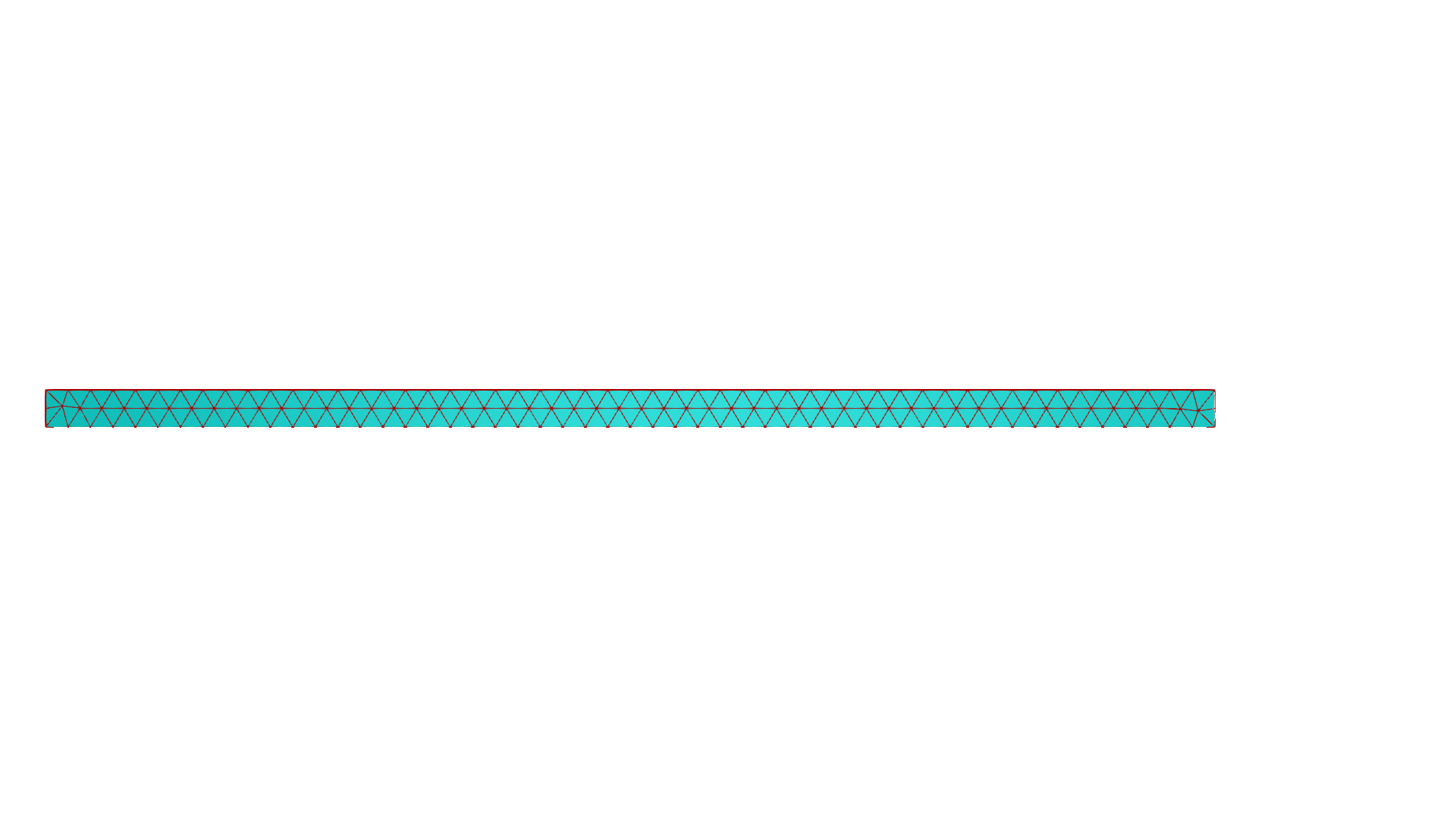}%
    \img{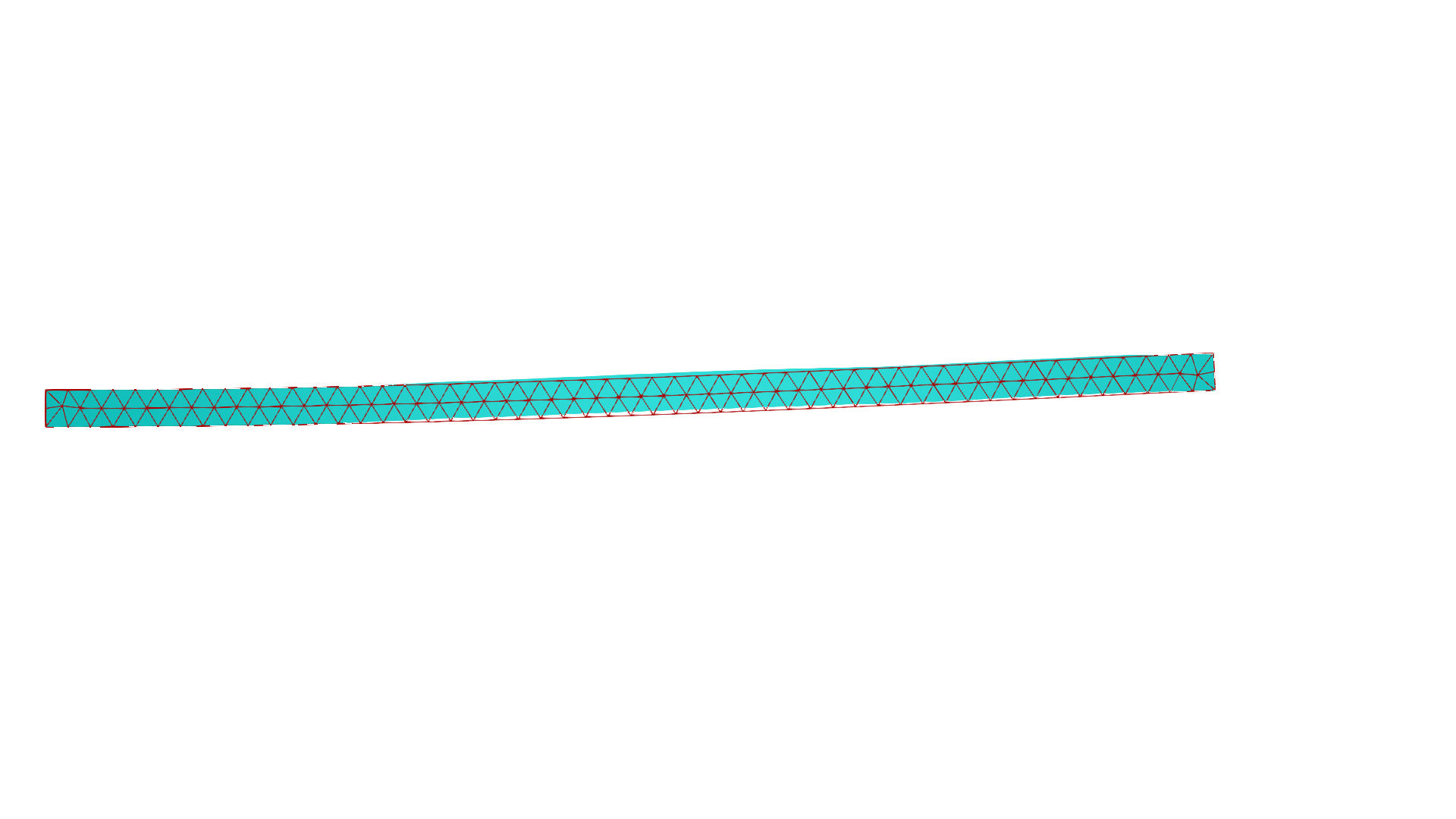}%
    \img{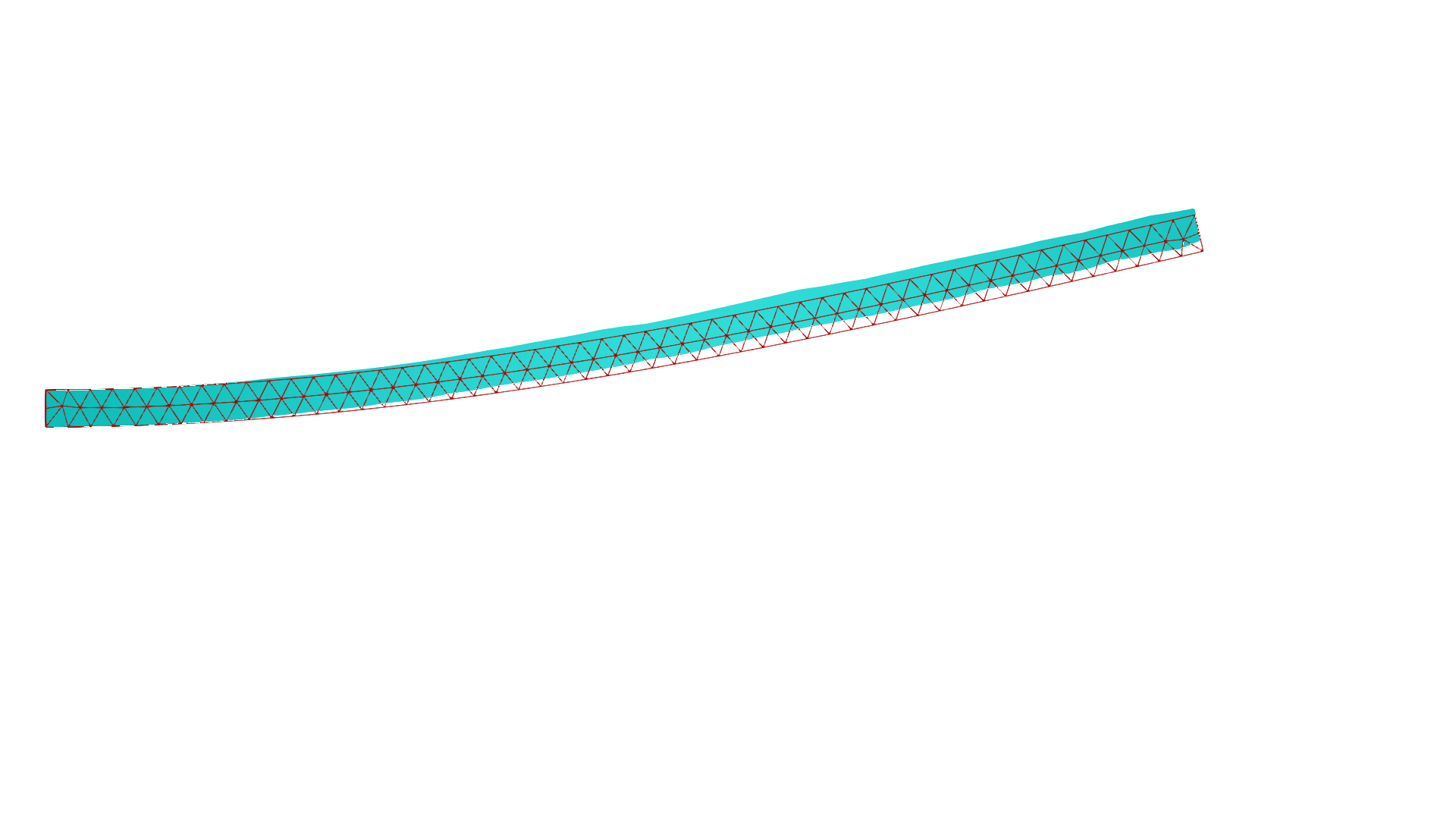}%
    \img{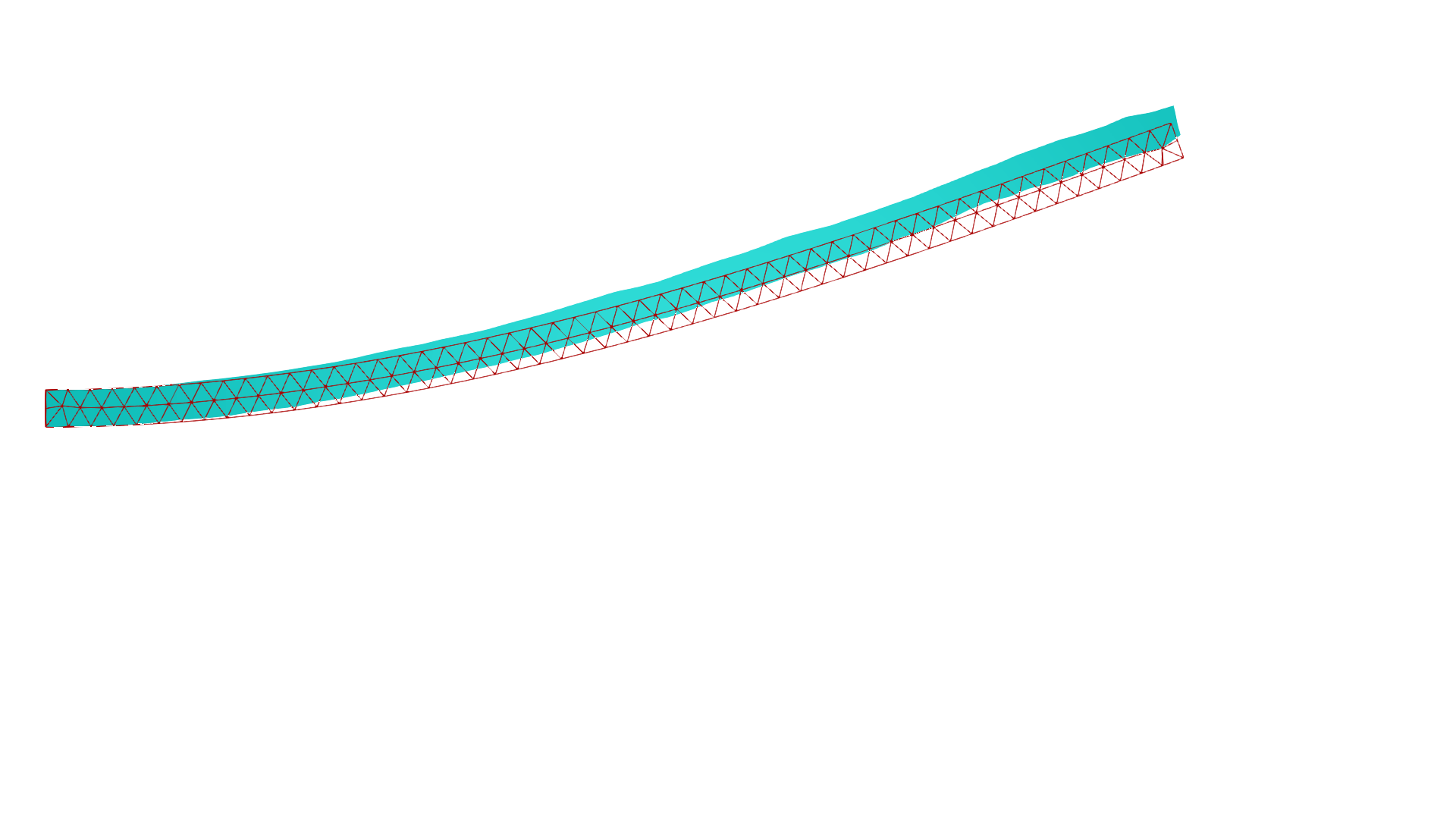}%
    \img{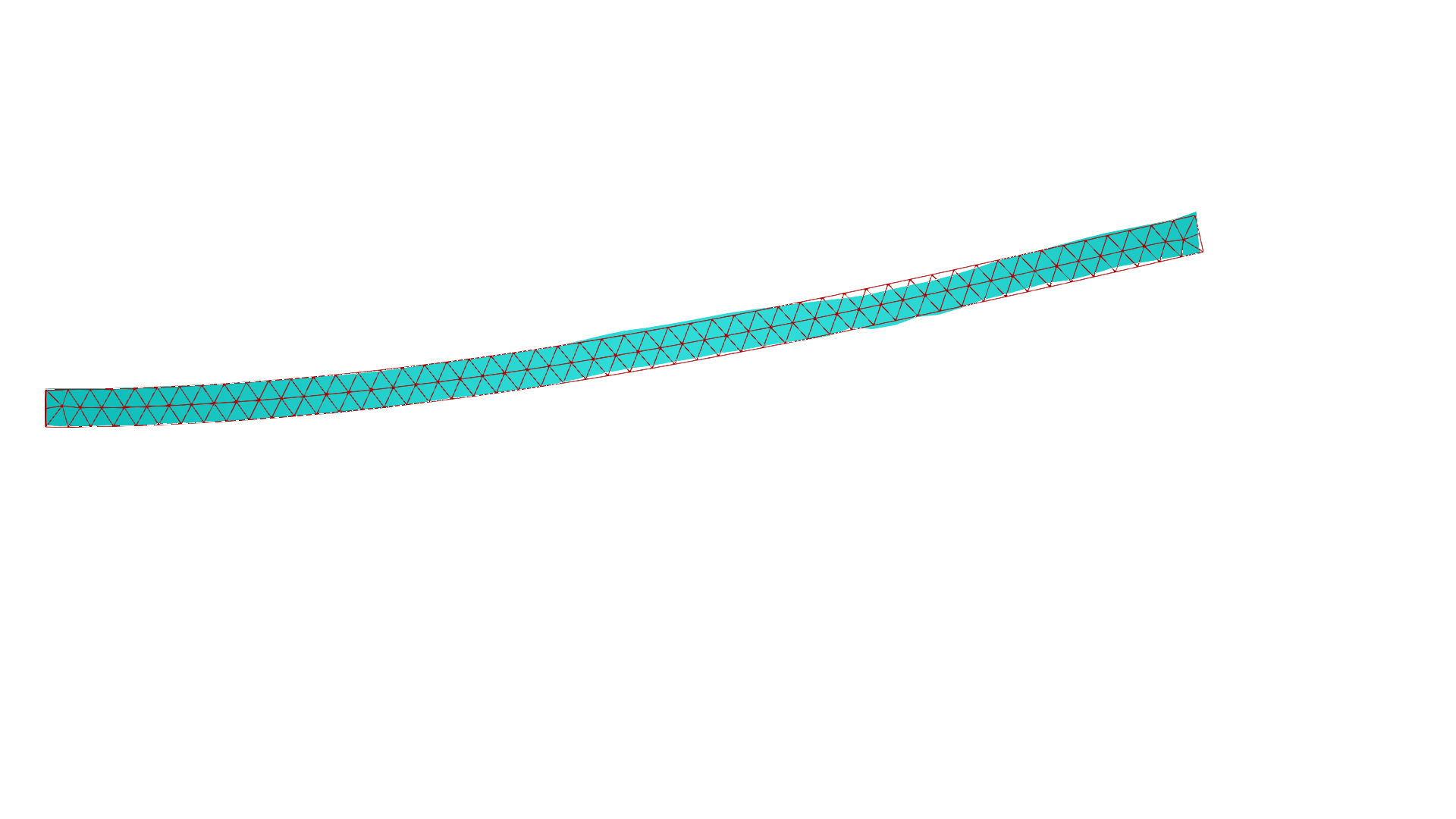}%
    \img{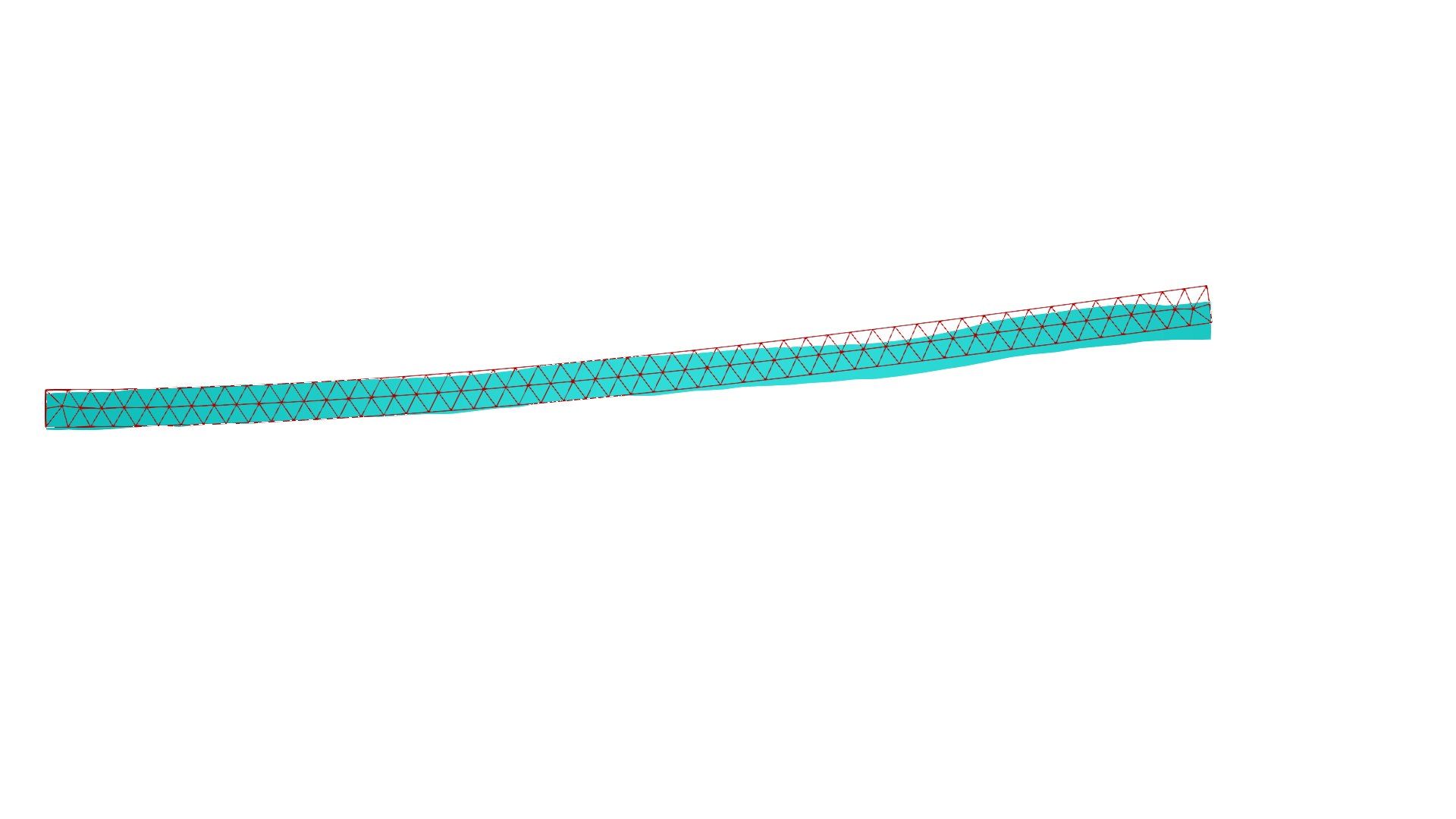}\\[0.6em]

    % Row 4: MANGO Oracle
    \rowlabel{Oracle}%
    \img{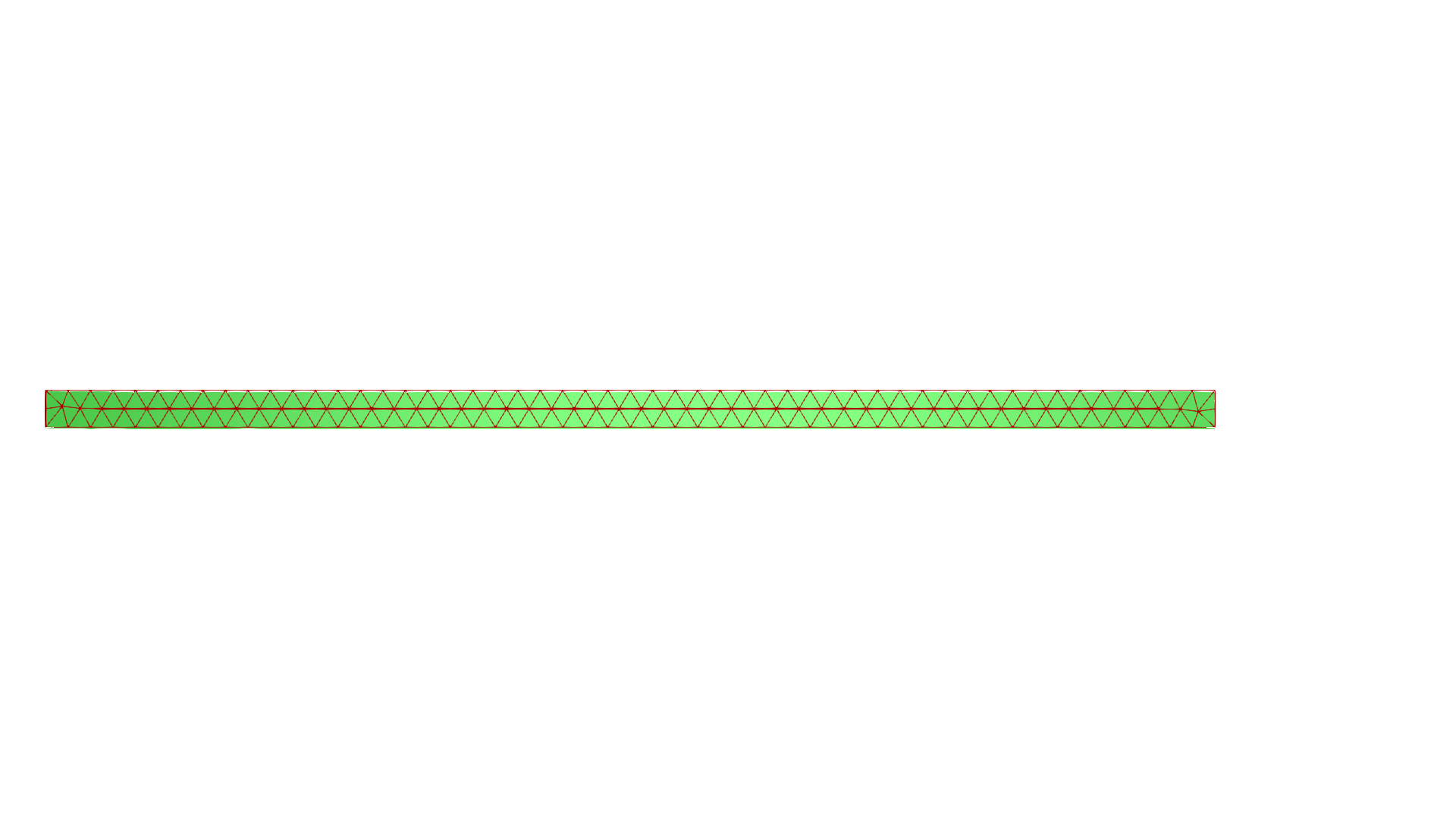}%
    \img{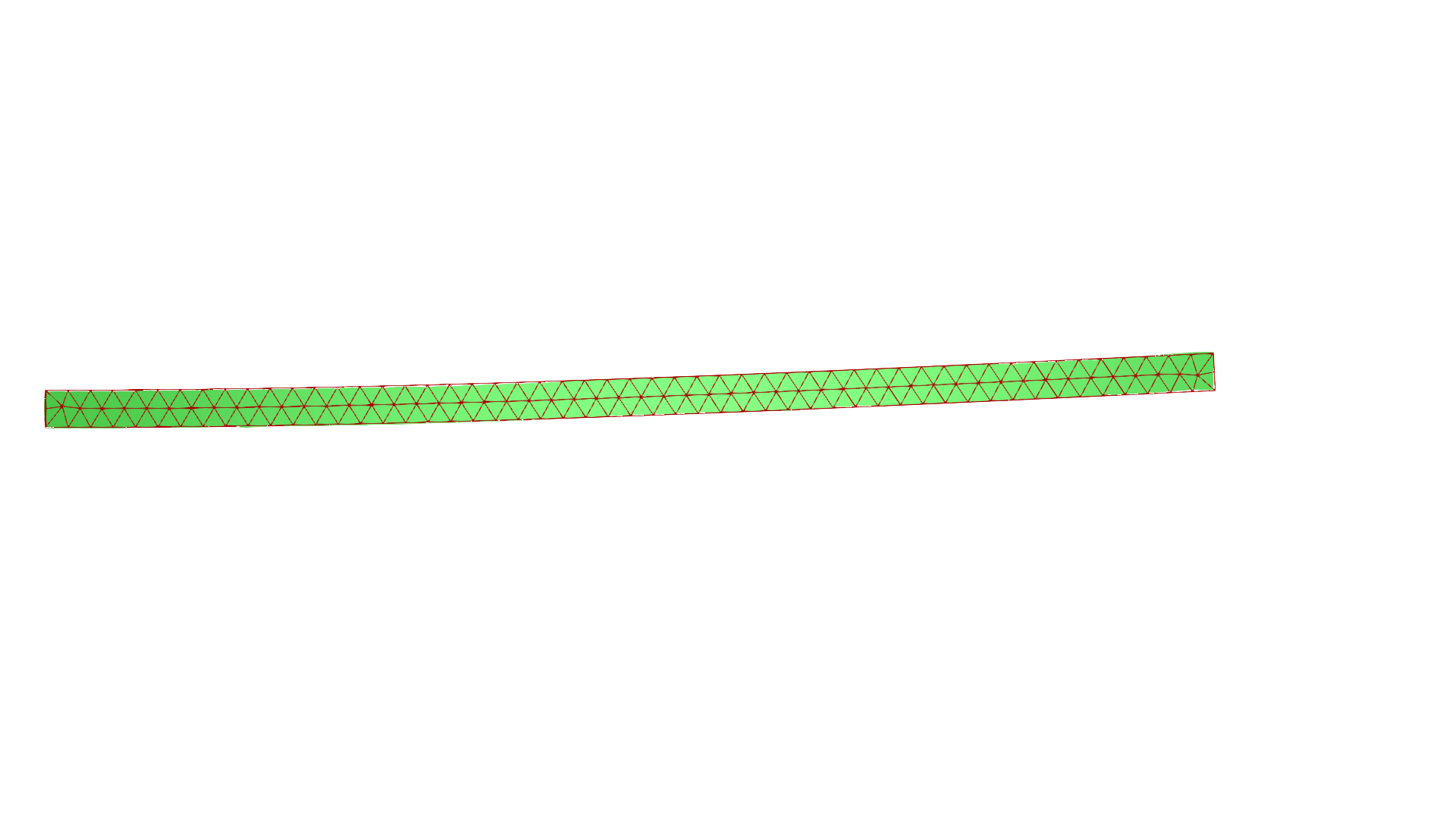}%
    \img{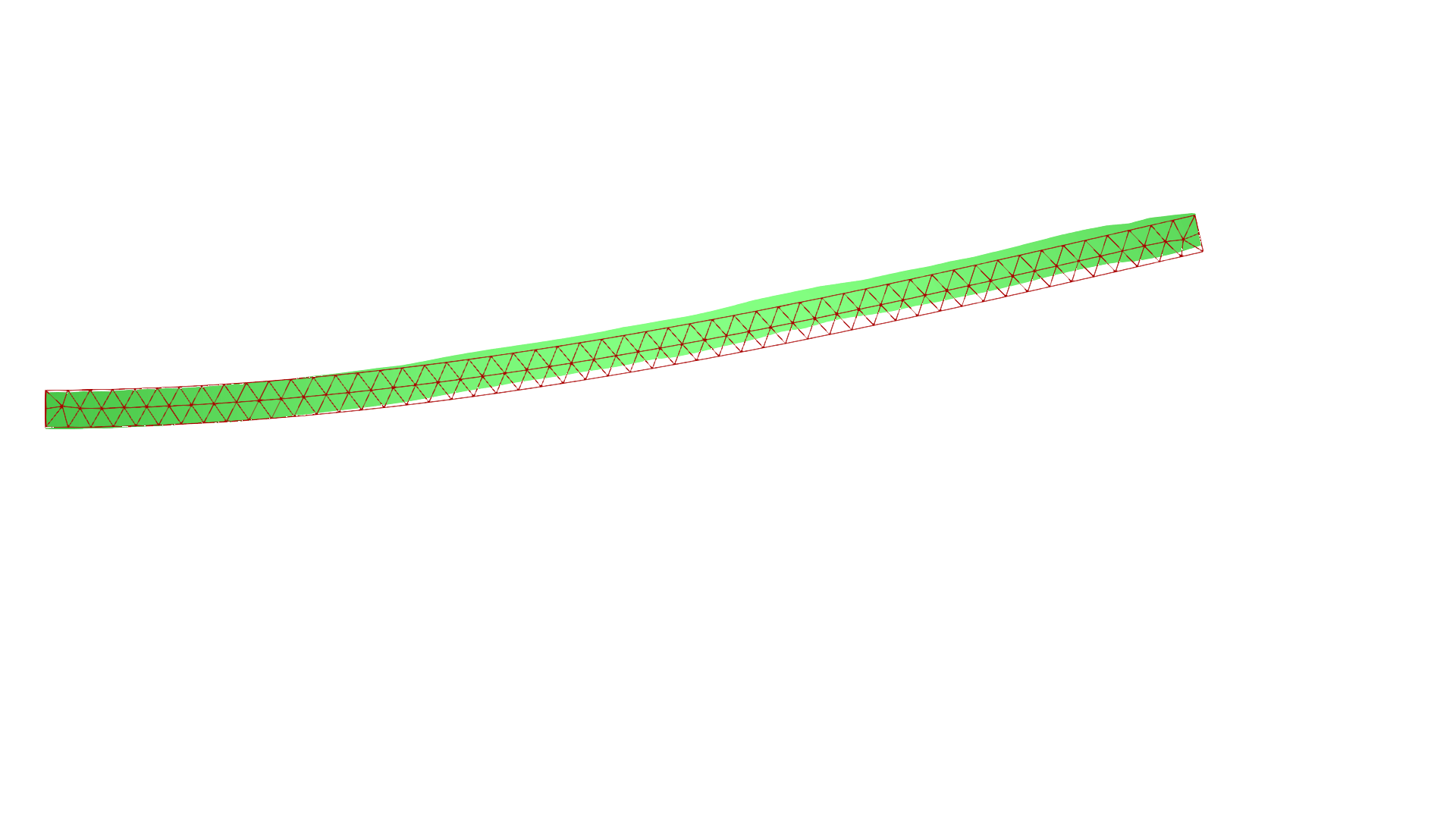}%
    \img{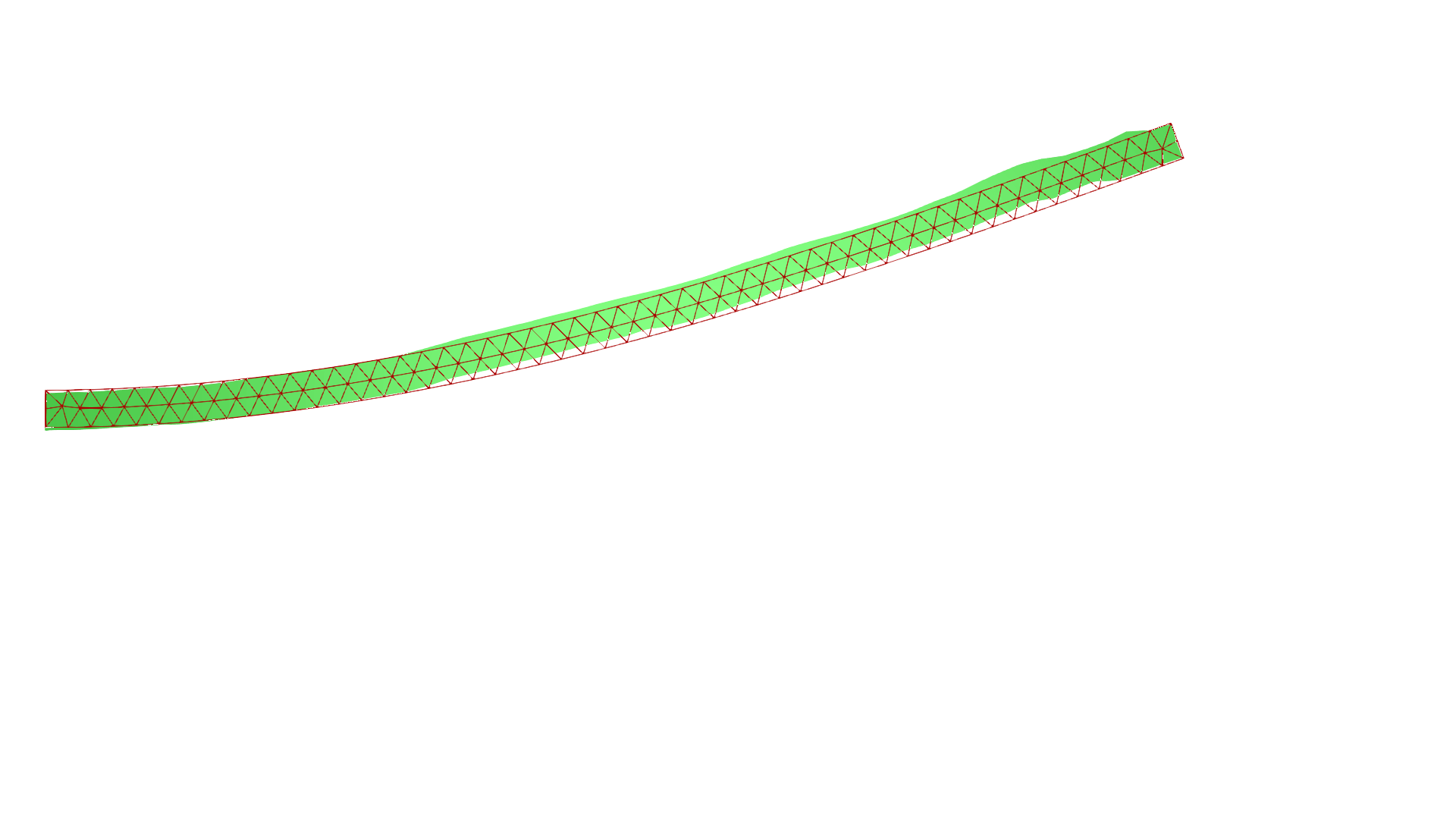}%
    \img{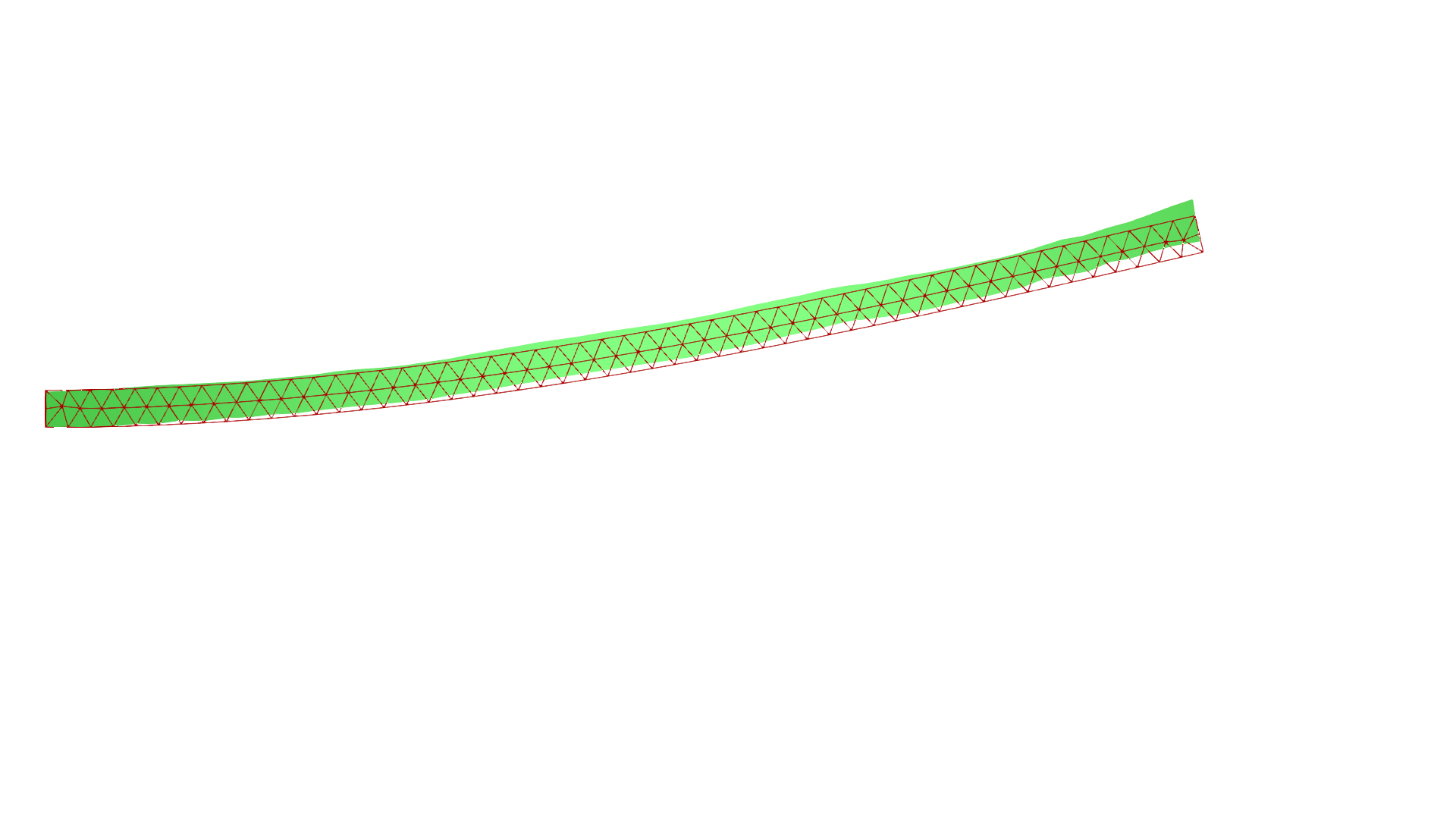}%
    \img{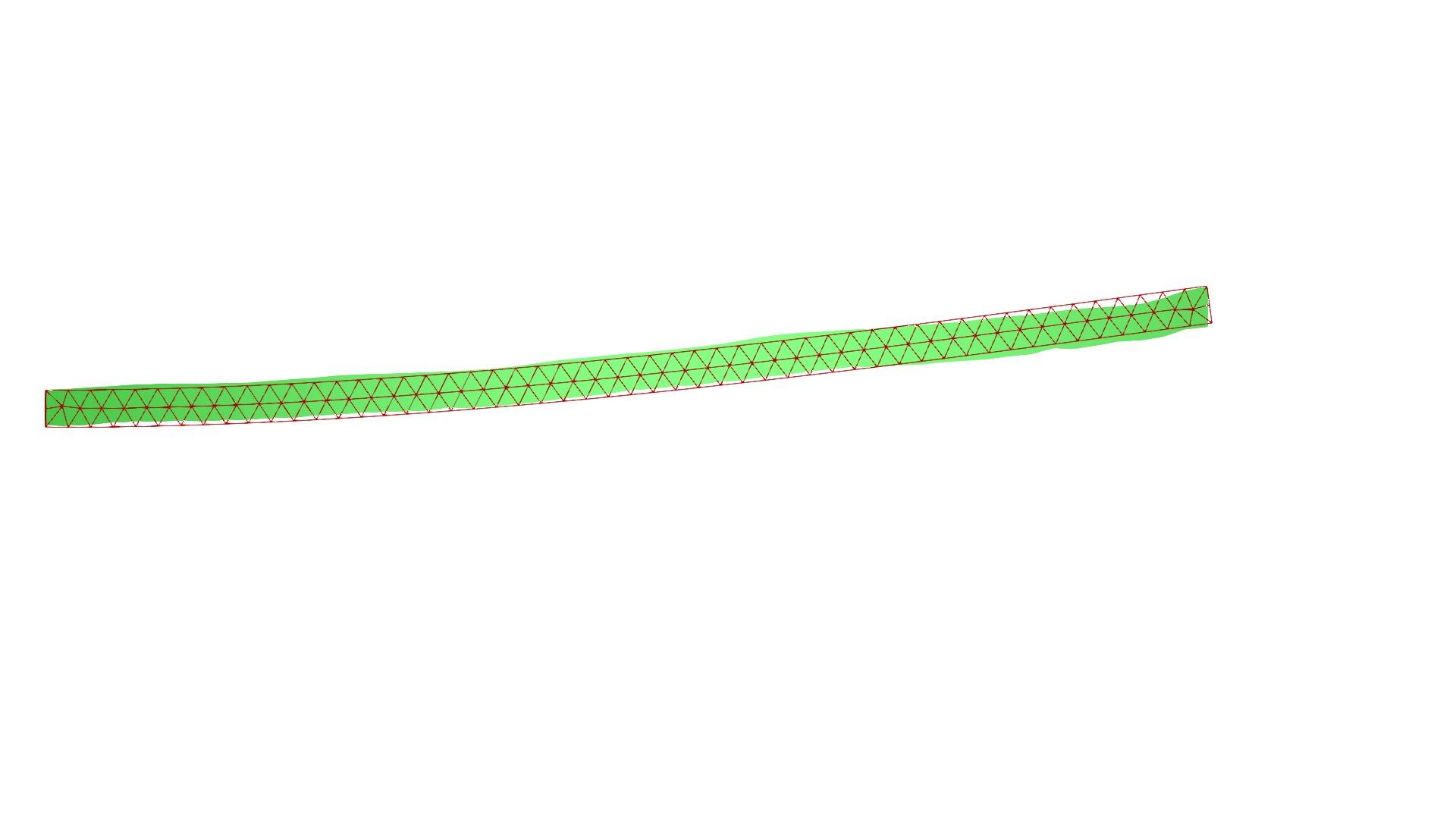}\\[0.6em]

    % Row 5: MGN
    \rowlabel{No Context \\ (MGN)}%
    \img{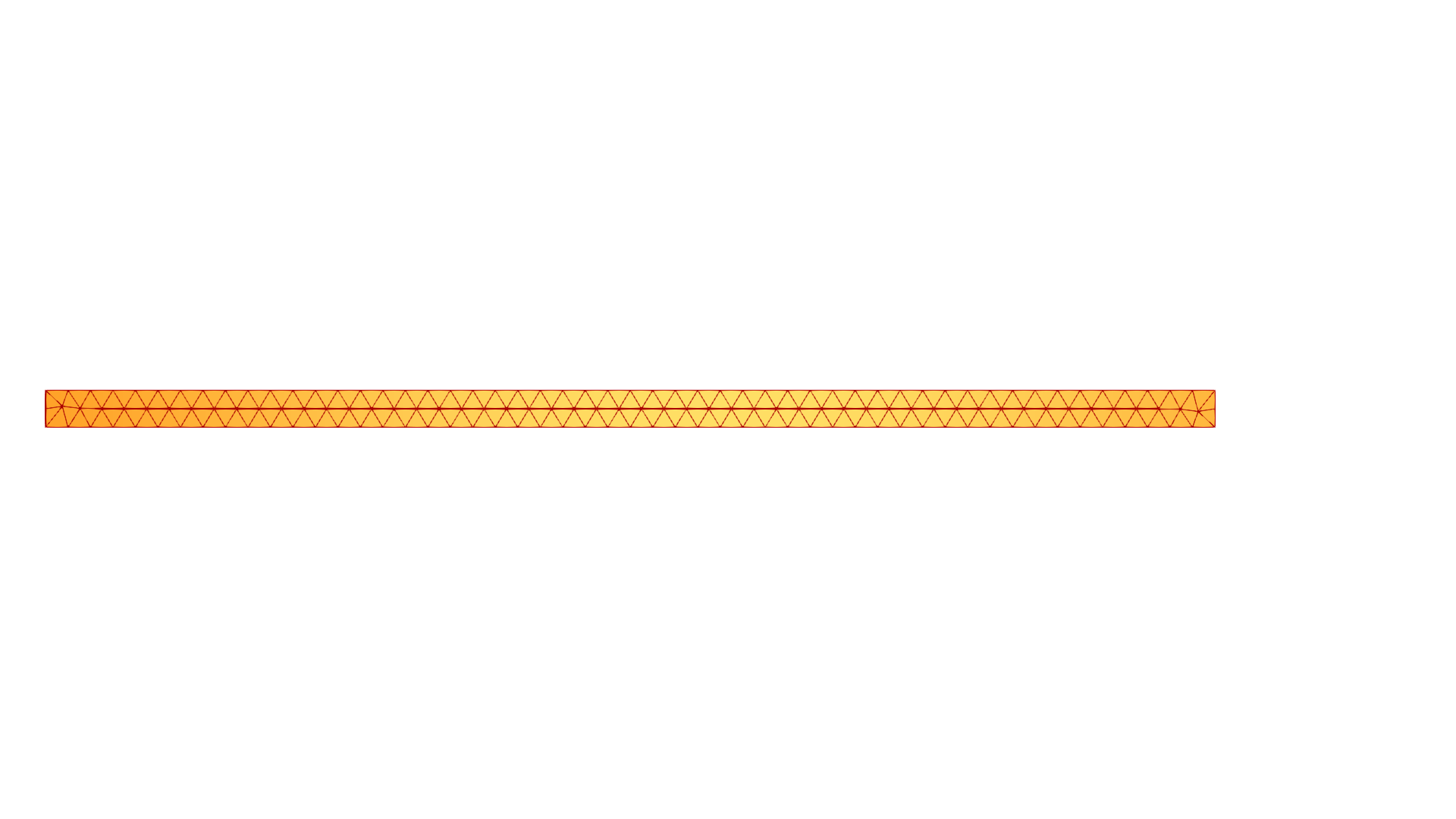}%
    \img{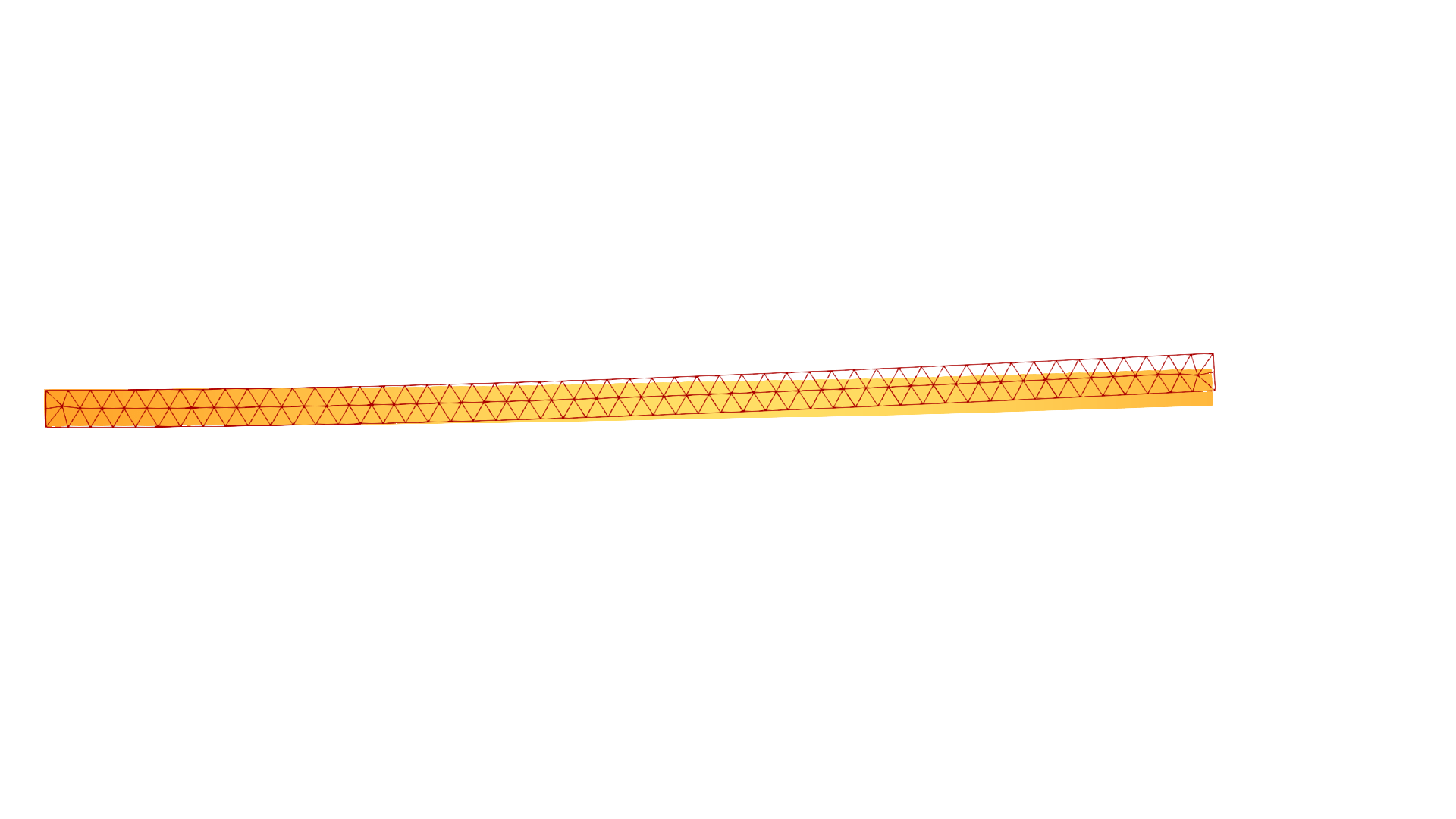}%
    \img{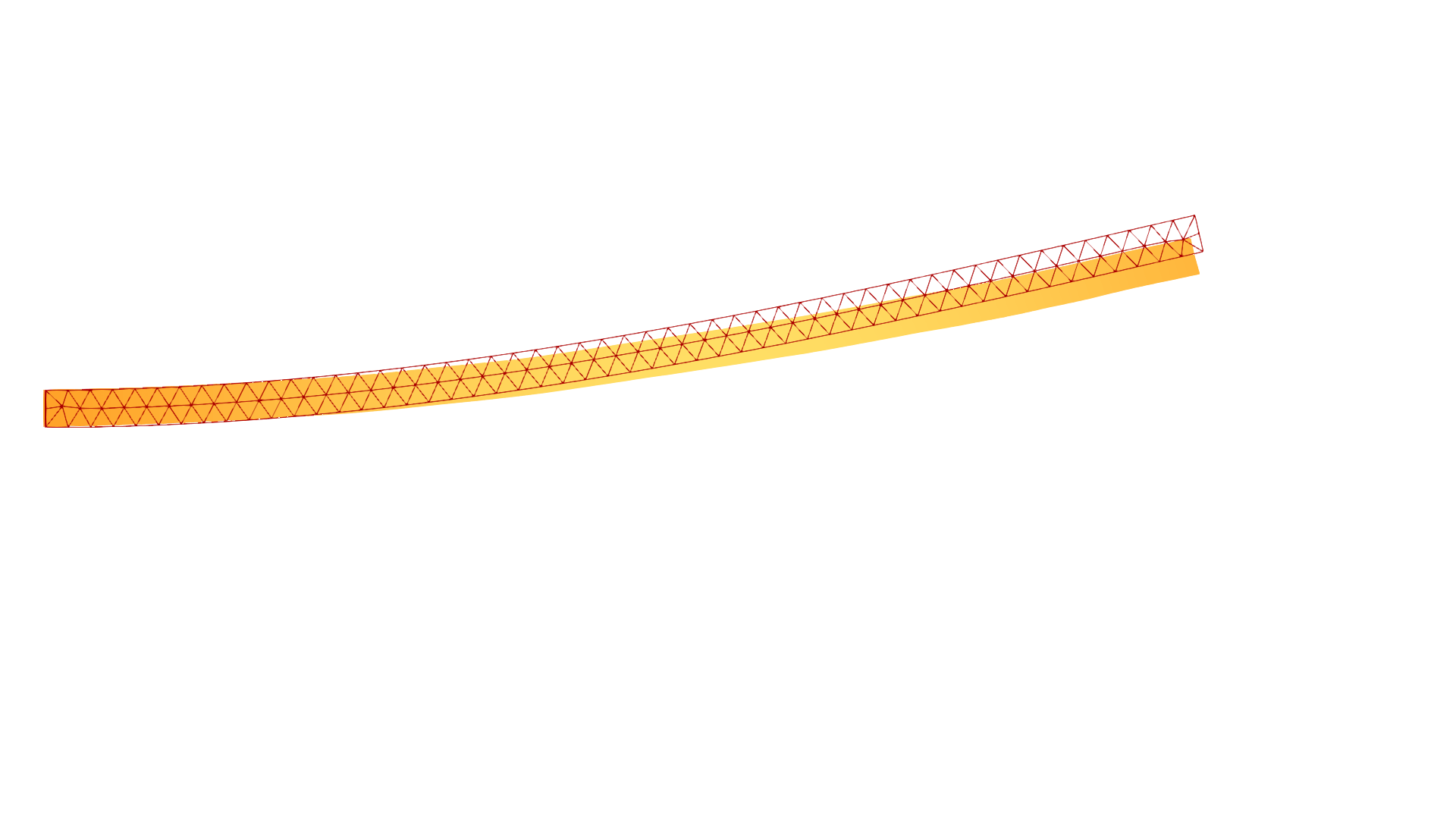}%
    \img{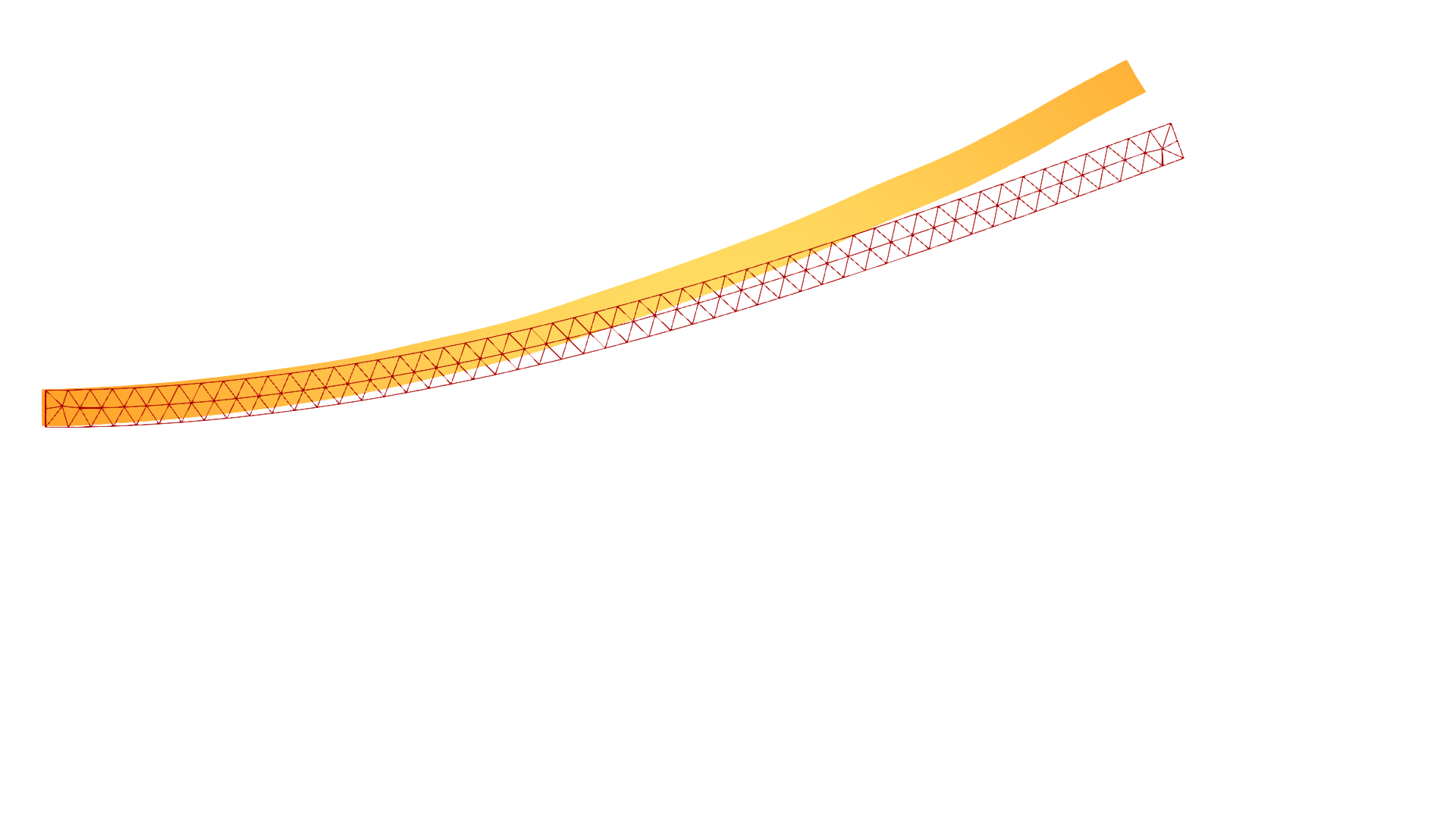}%
    \img{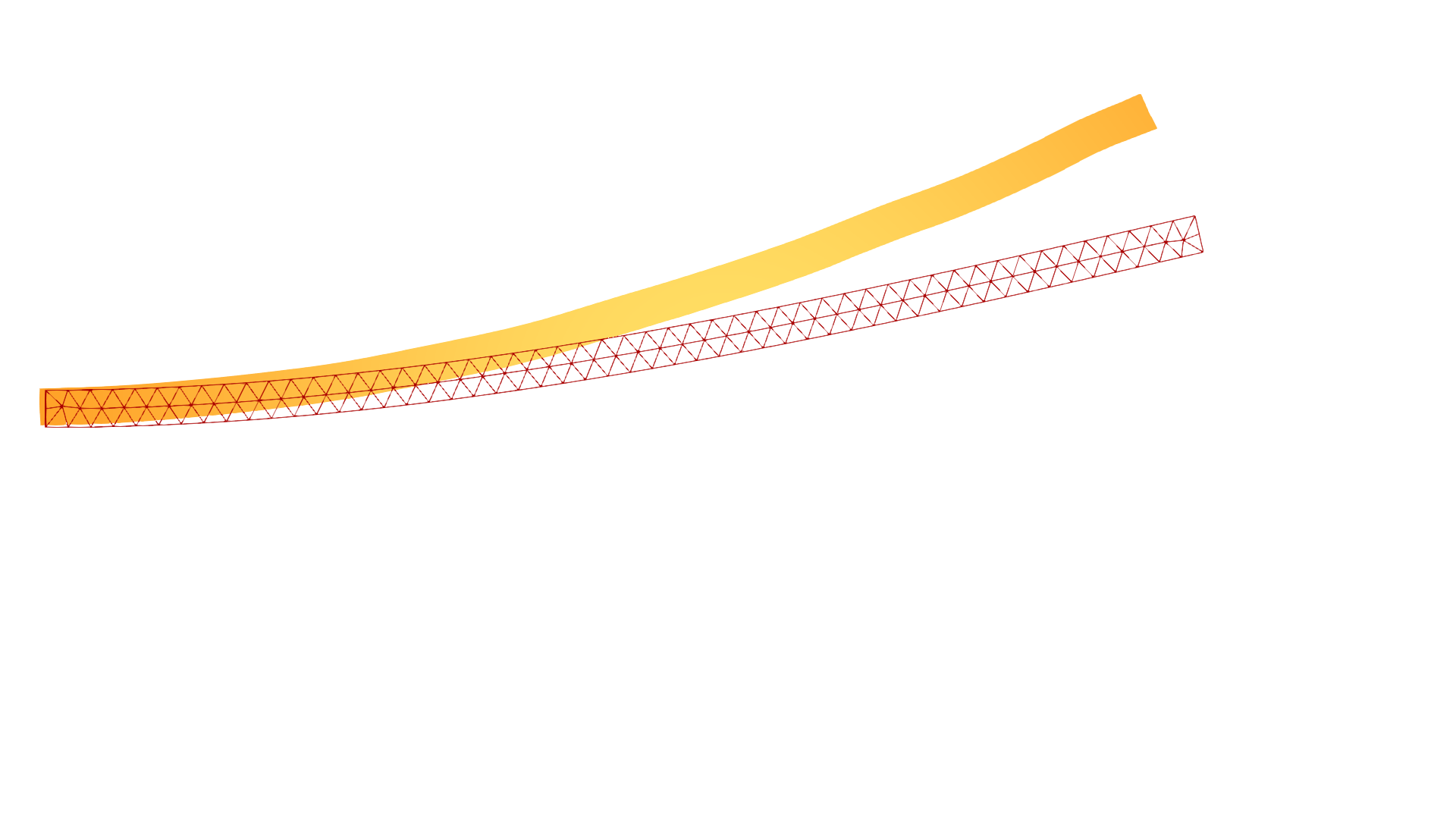}%
    \img{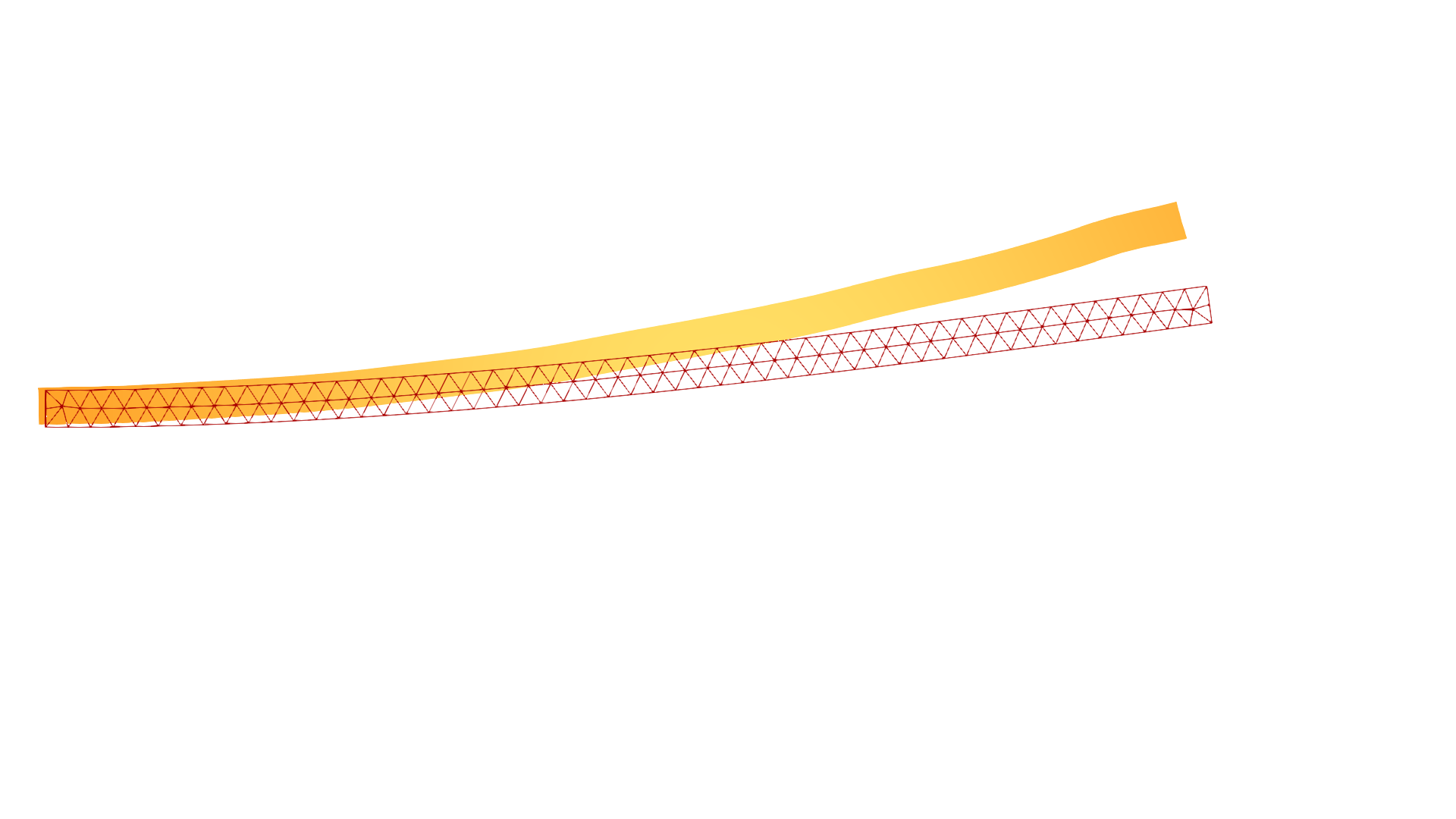}\\[0.6em]

    % Row 6: MGN Oracle
    \rowlabel{Oracle \\ (MGN)}%
    \img{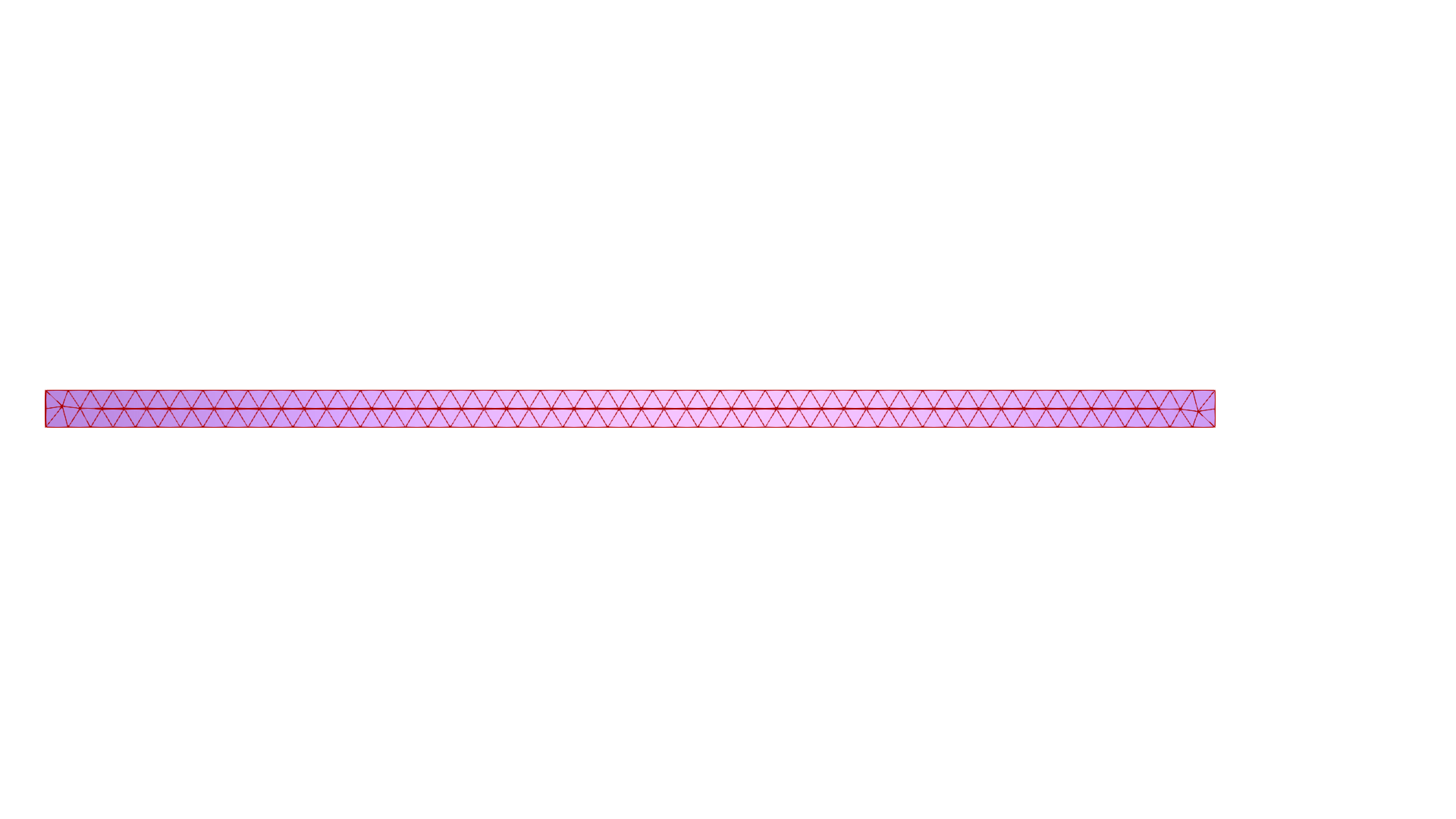}%
    \img{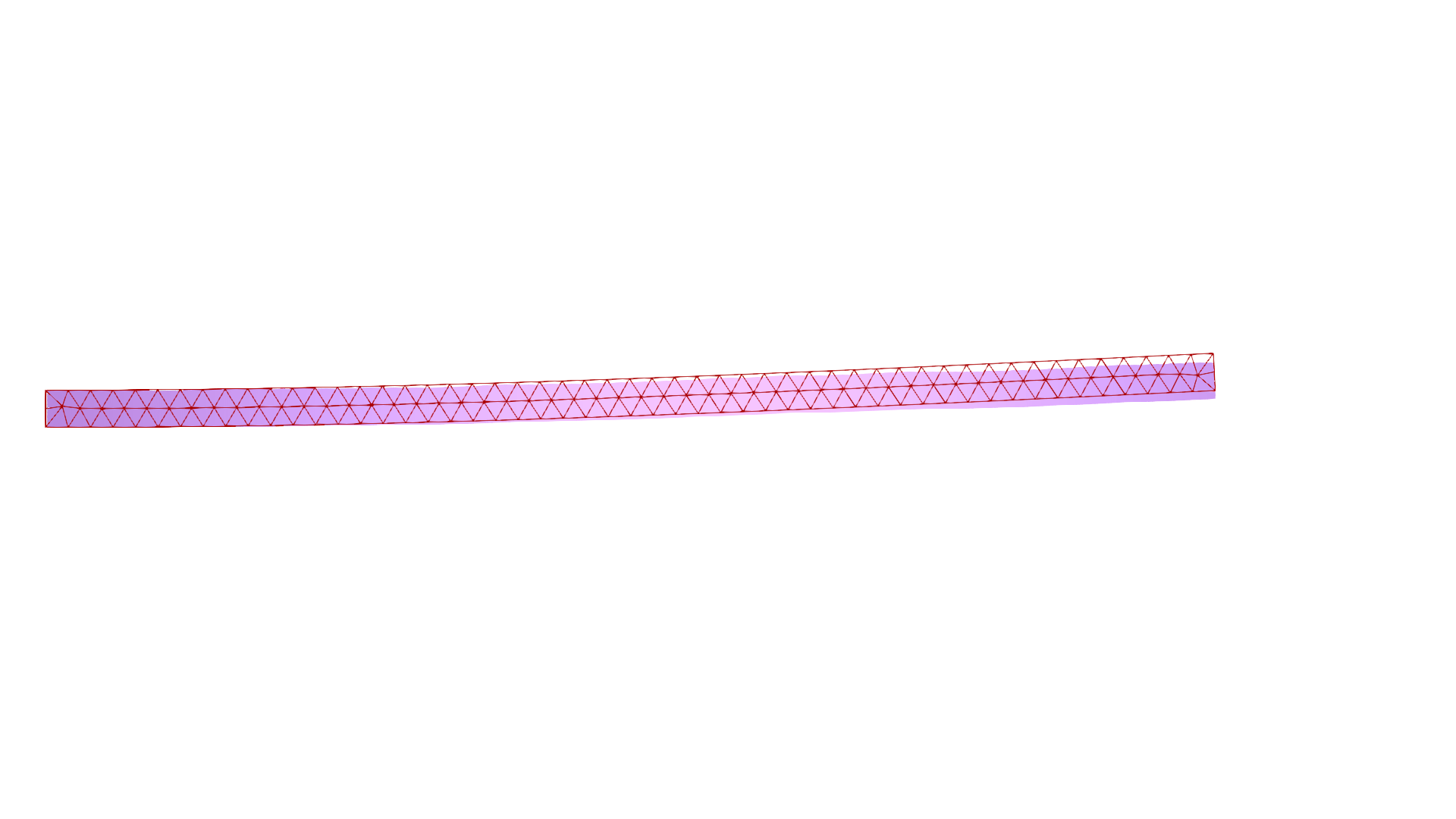}%
    \img{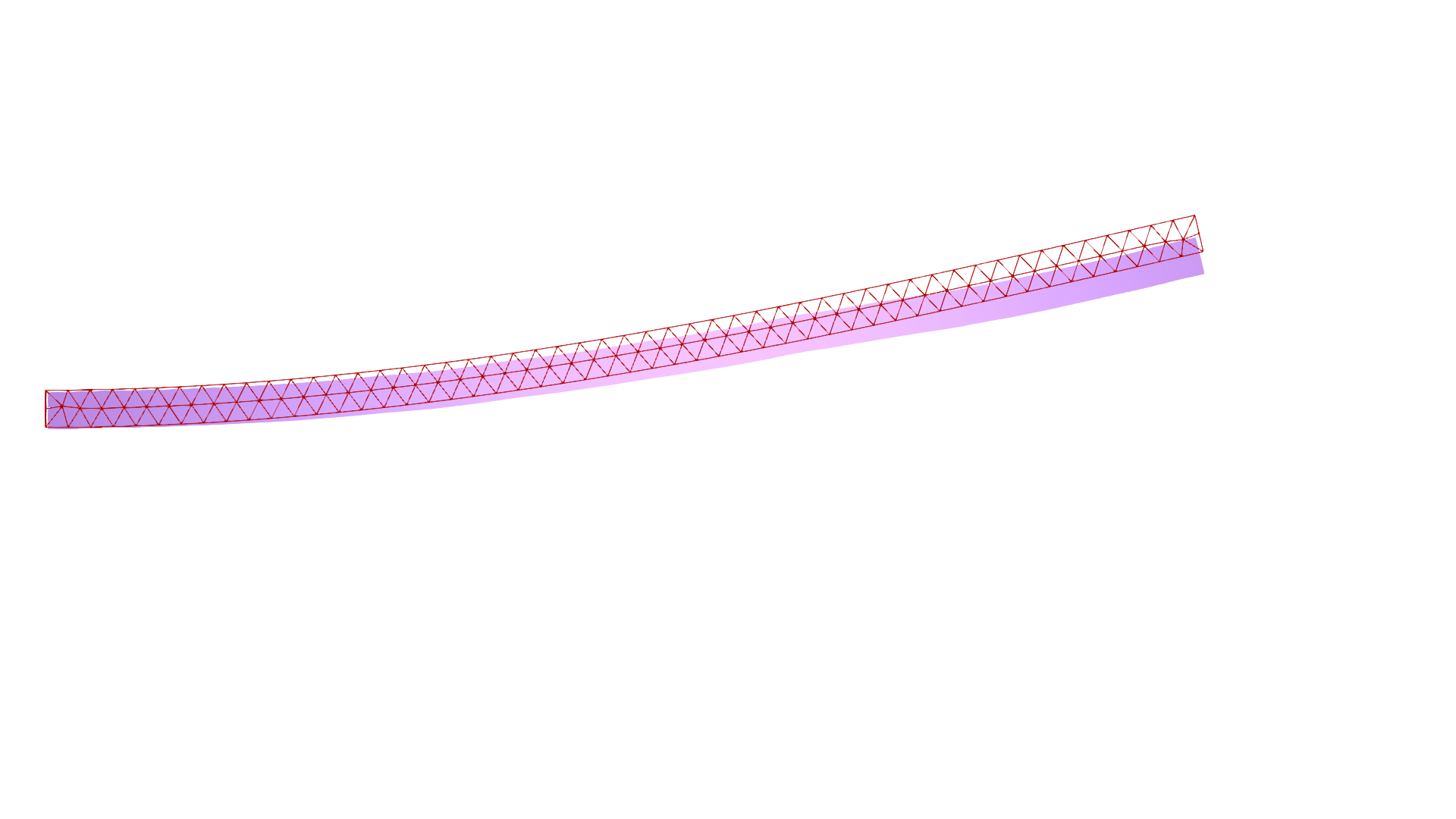}%
    \img{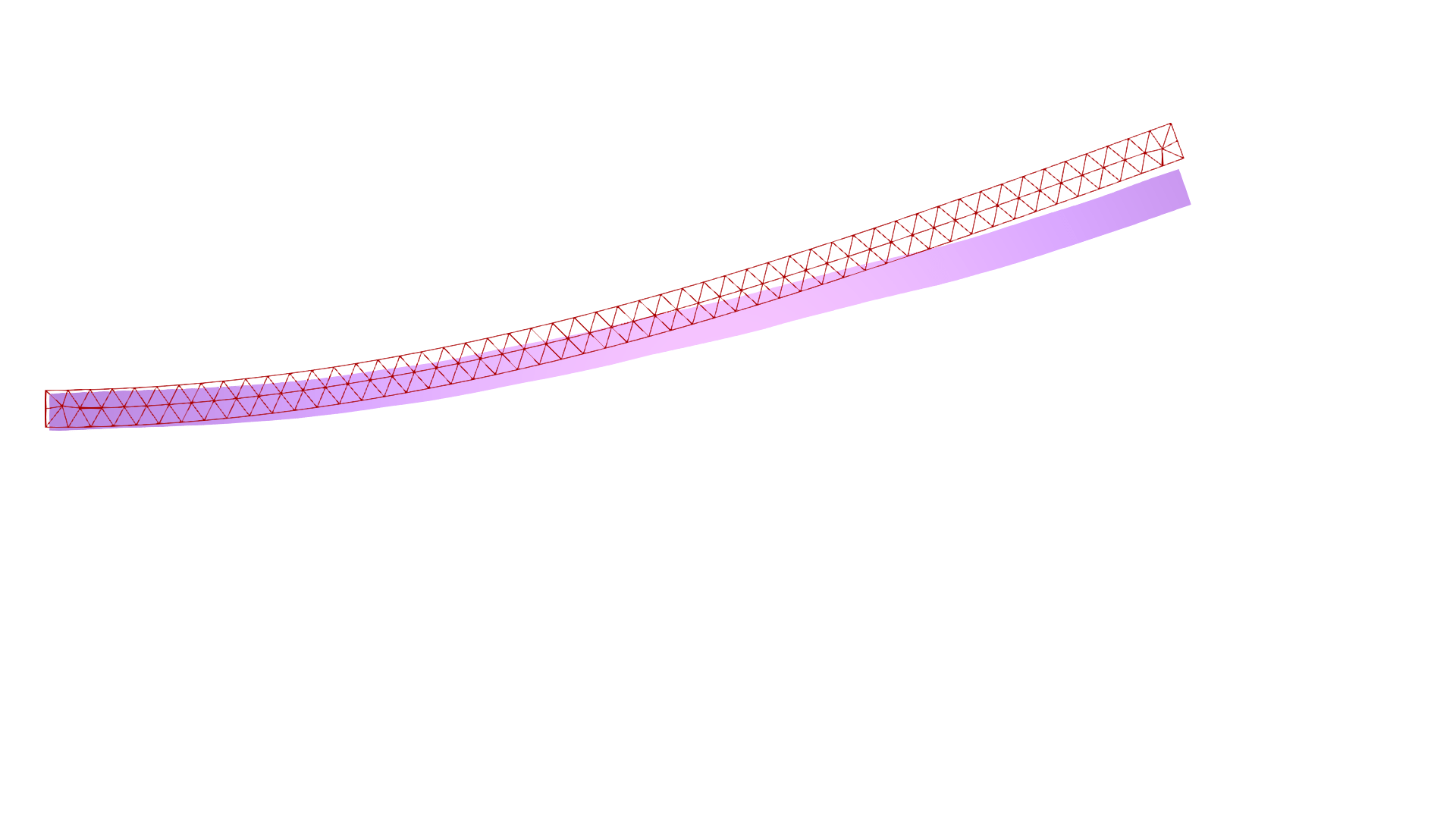}%
    \img{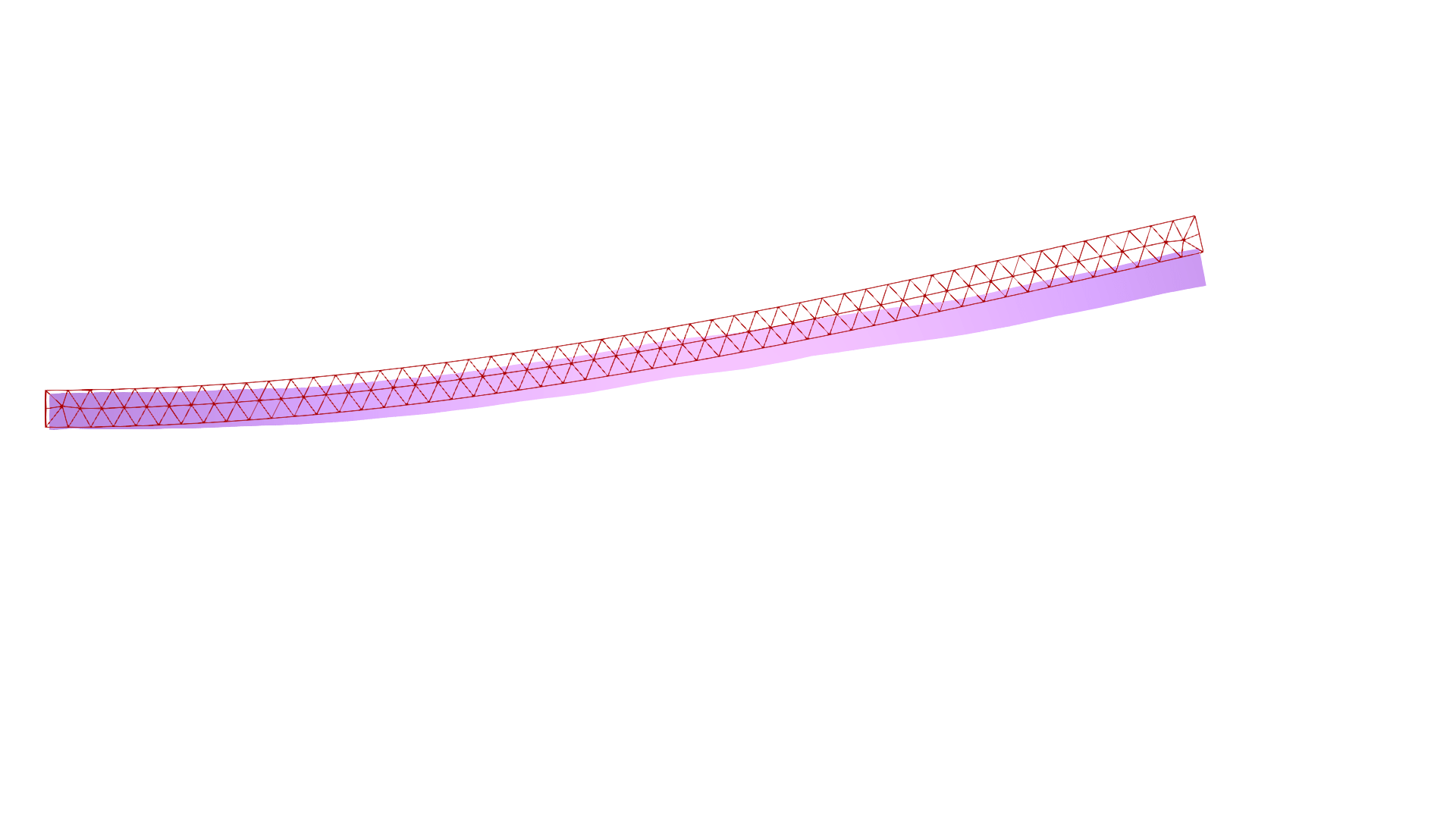}%
    \img{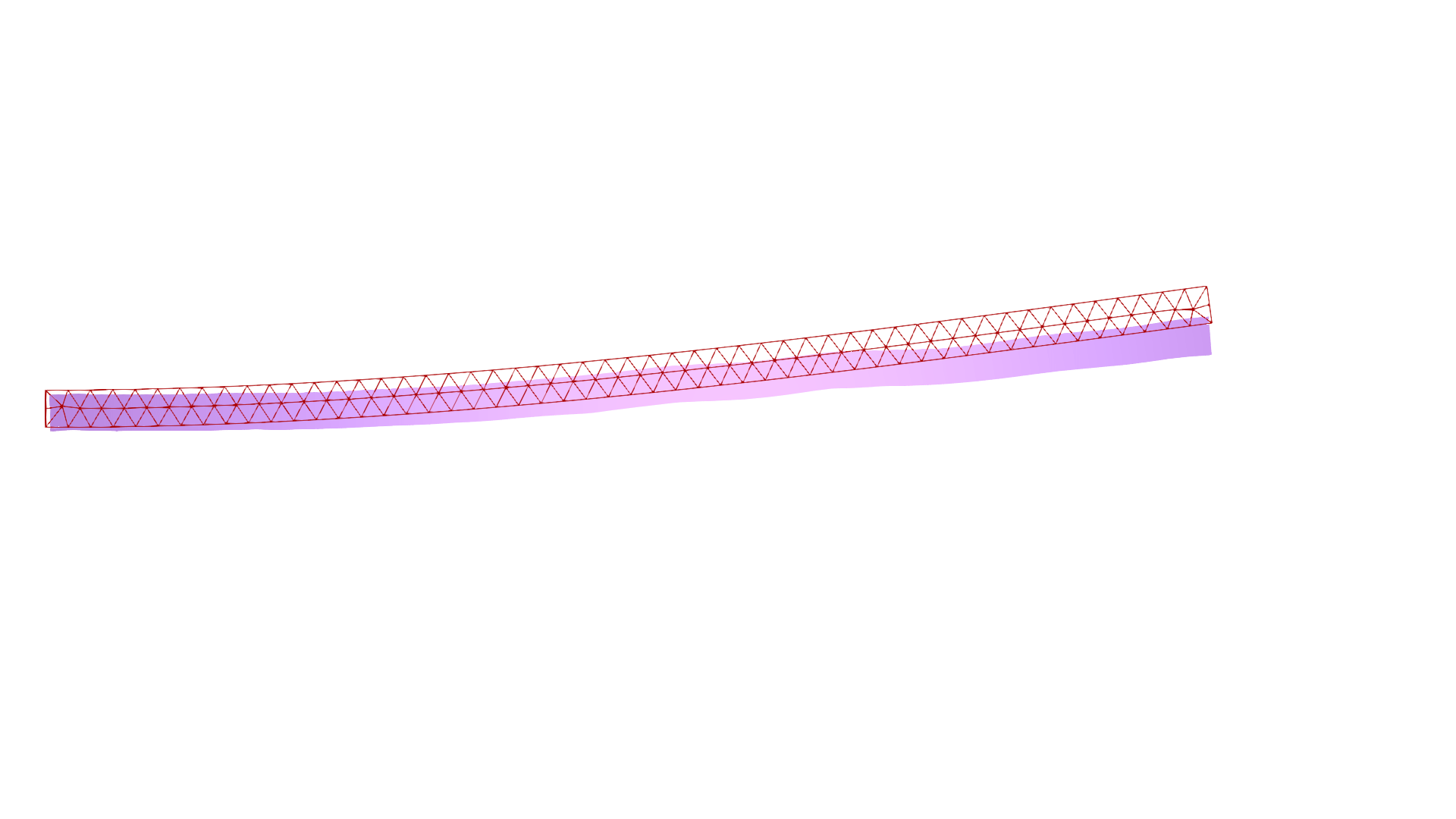}\\[0.6em]

    % Row 7: GNN Encoder
    \rowlabel{GNN \\ Encoder}%
    \img{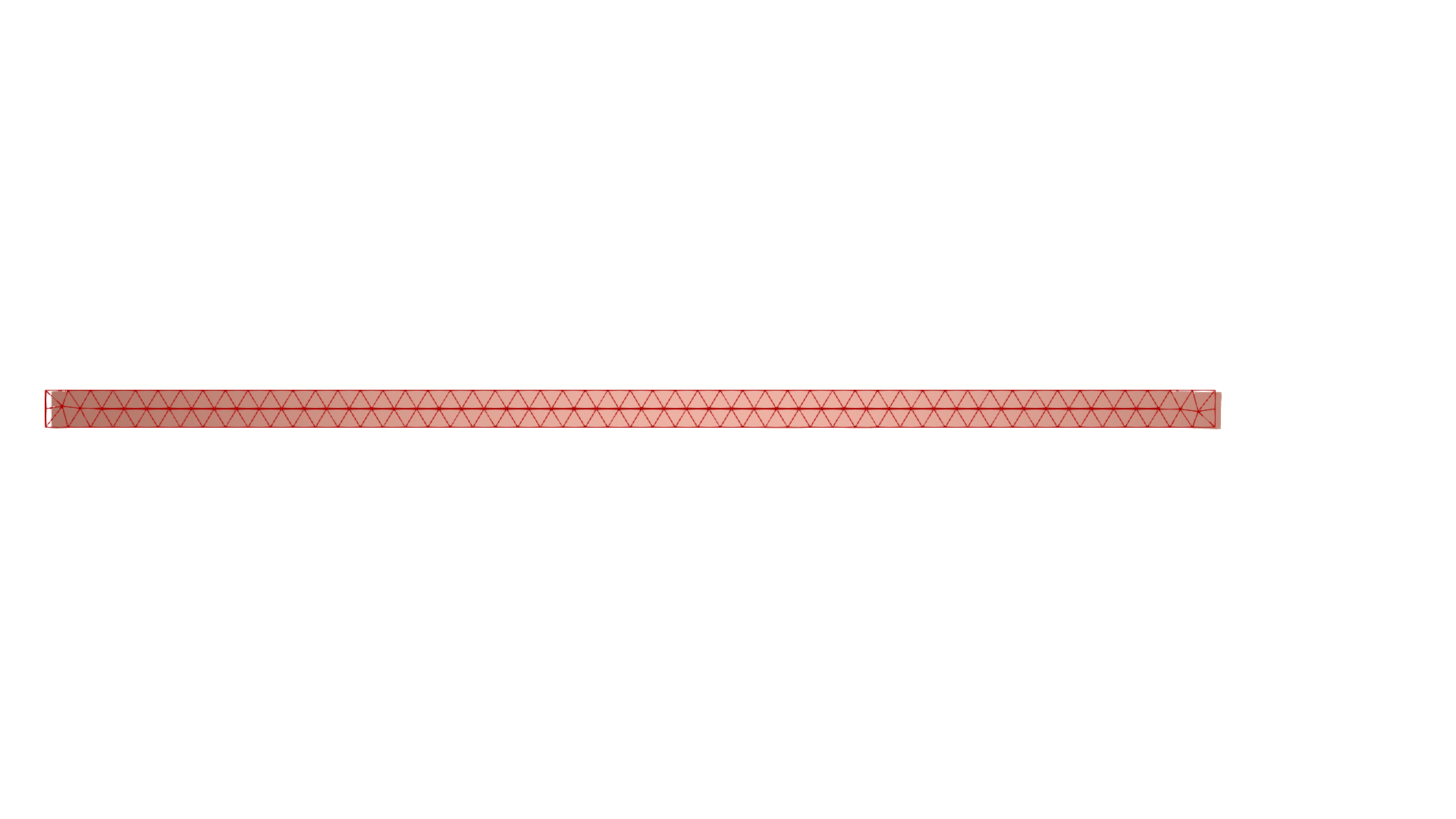}%
    \img{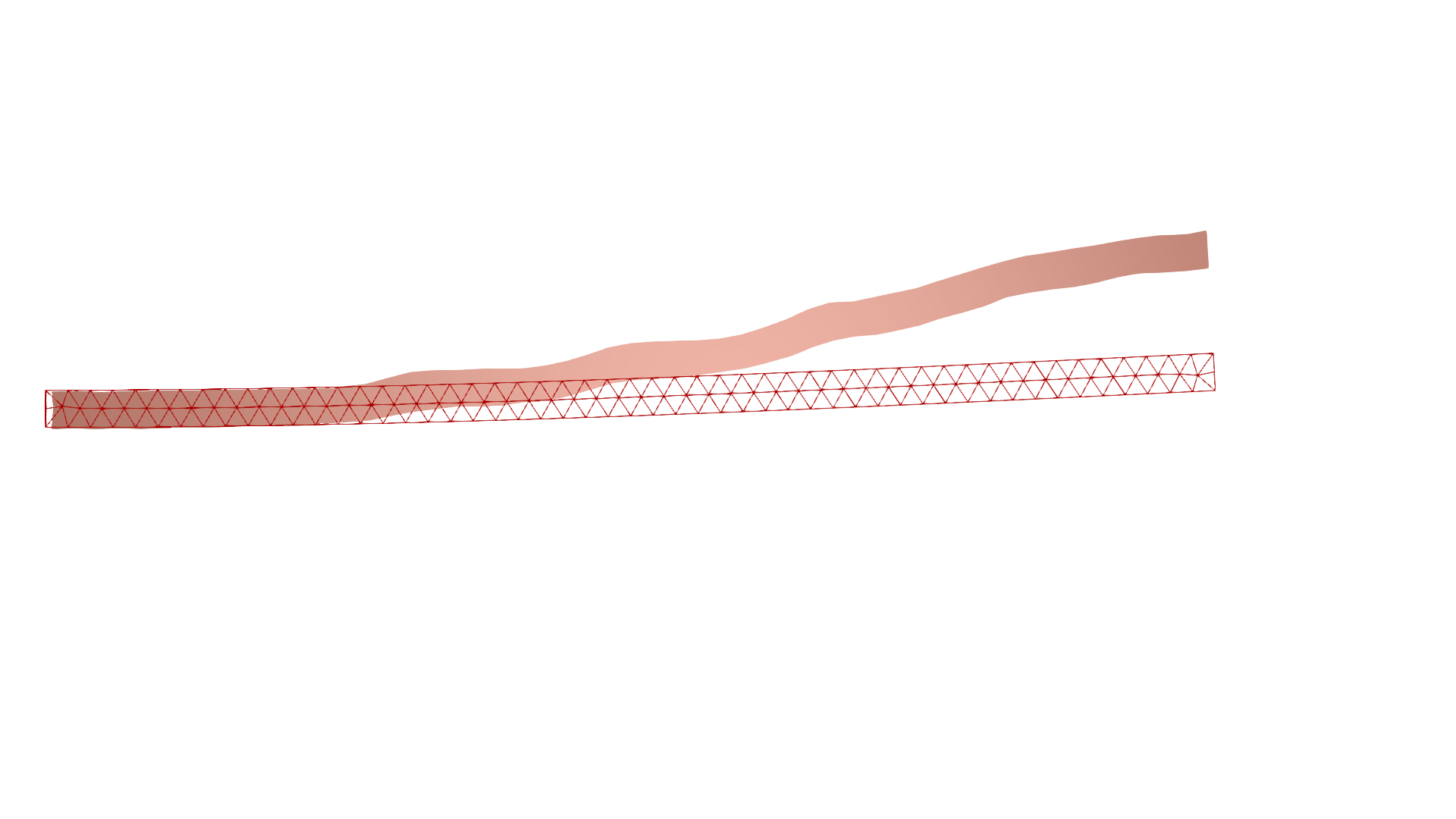}%
    \img{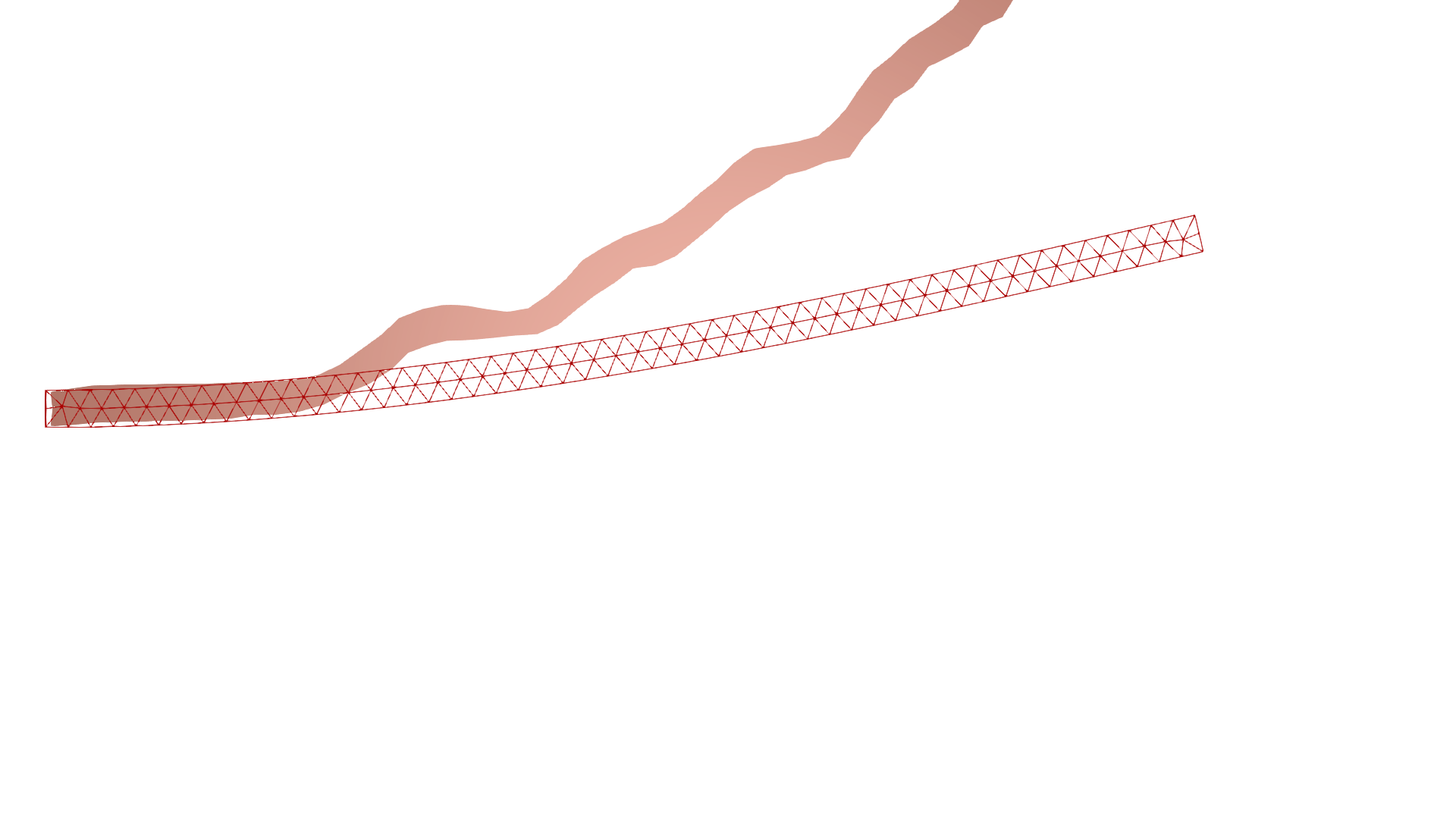}%
    \img{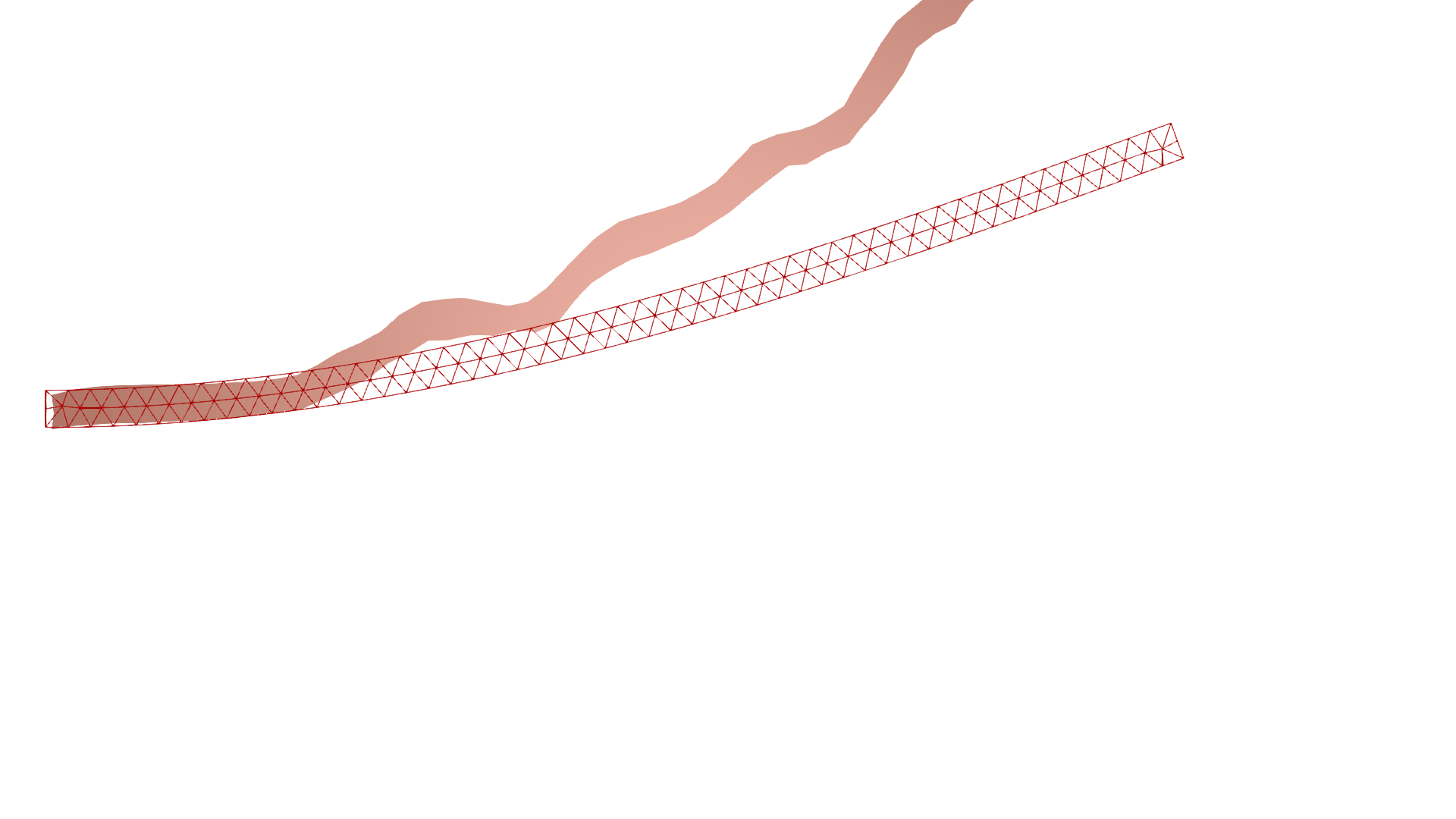}%
    \img{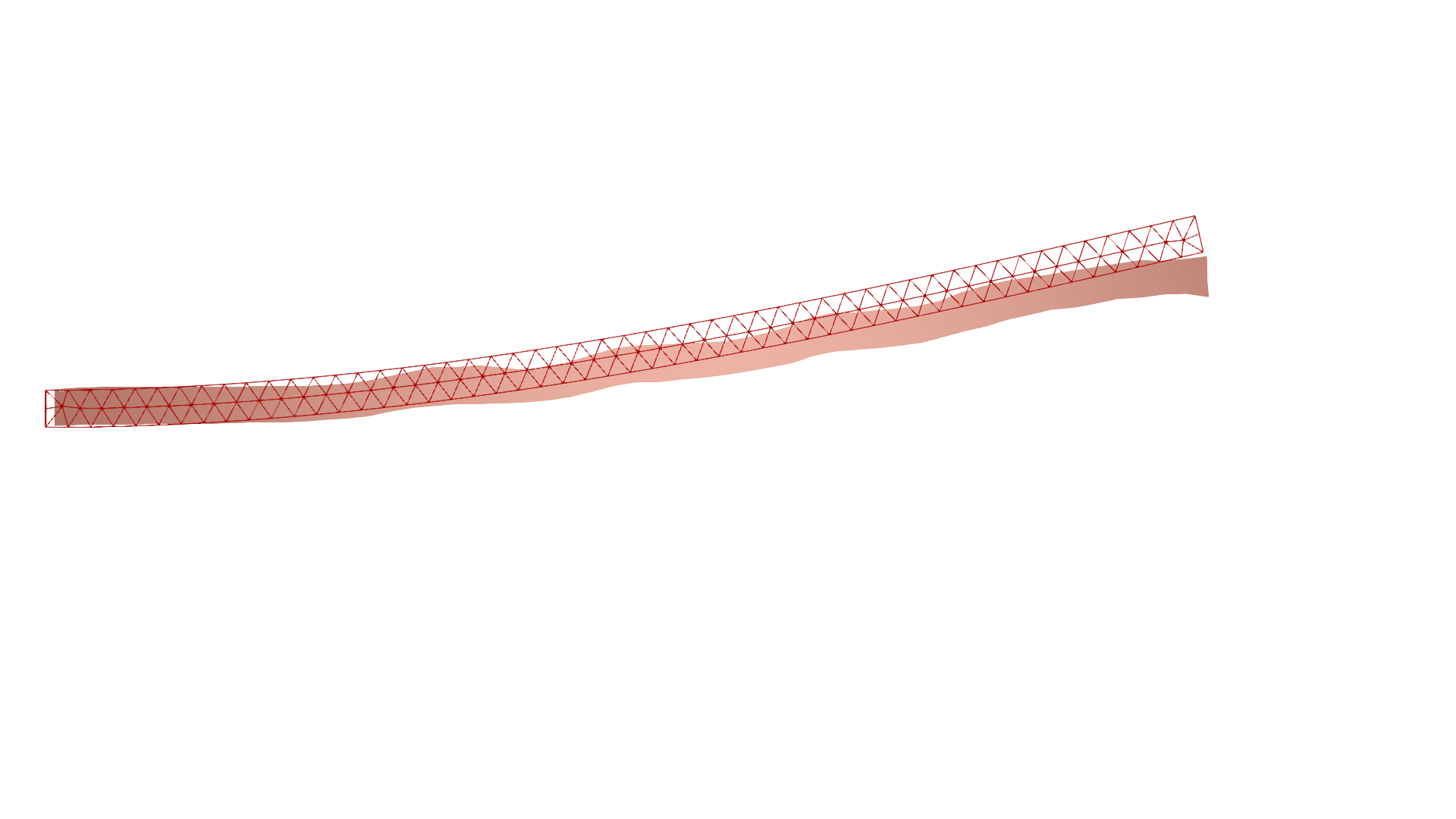}%
    \img{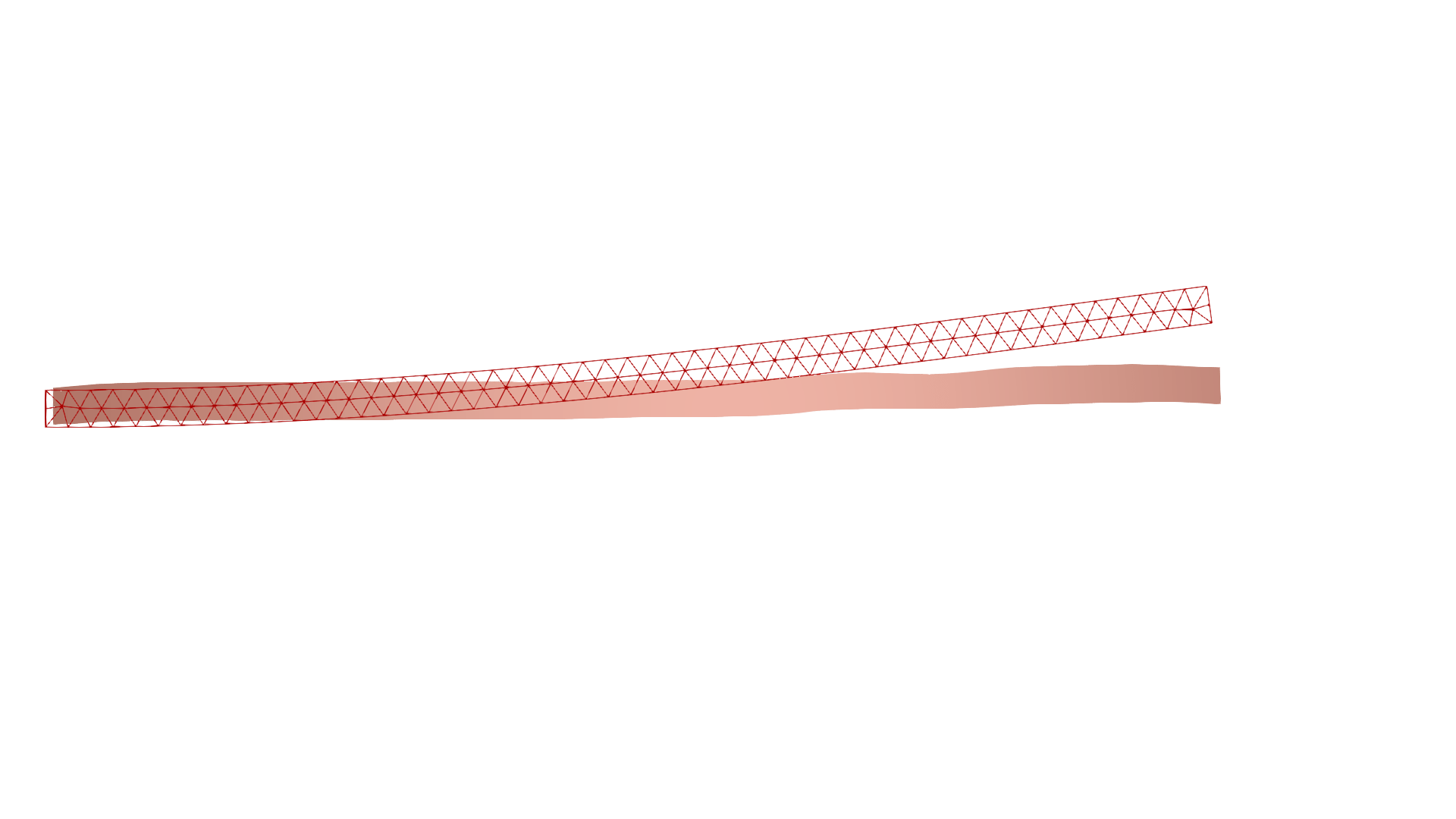}\\[0.6em]

    % Row 8: PSTNET Encoder
    \rowlabel{PSTNet \\ Encoder}%
    \img{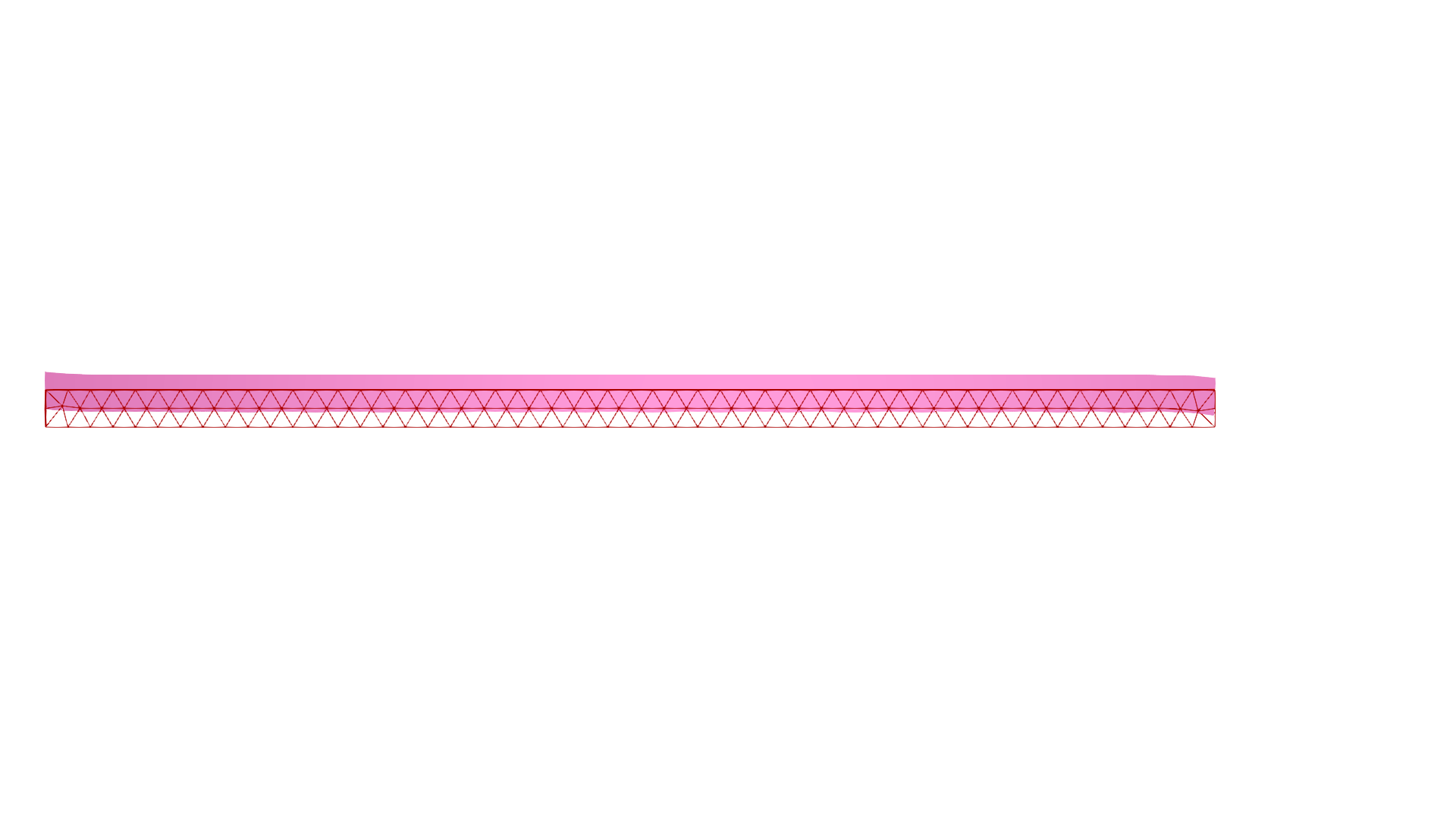}%
    \img{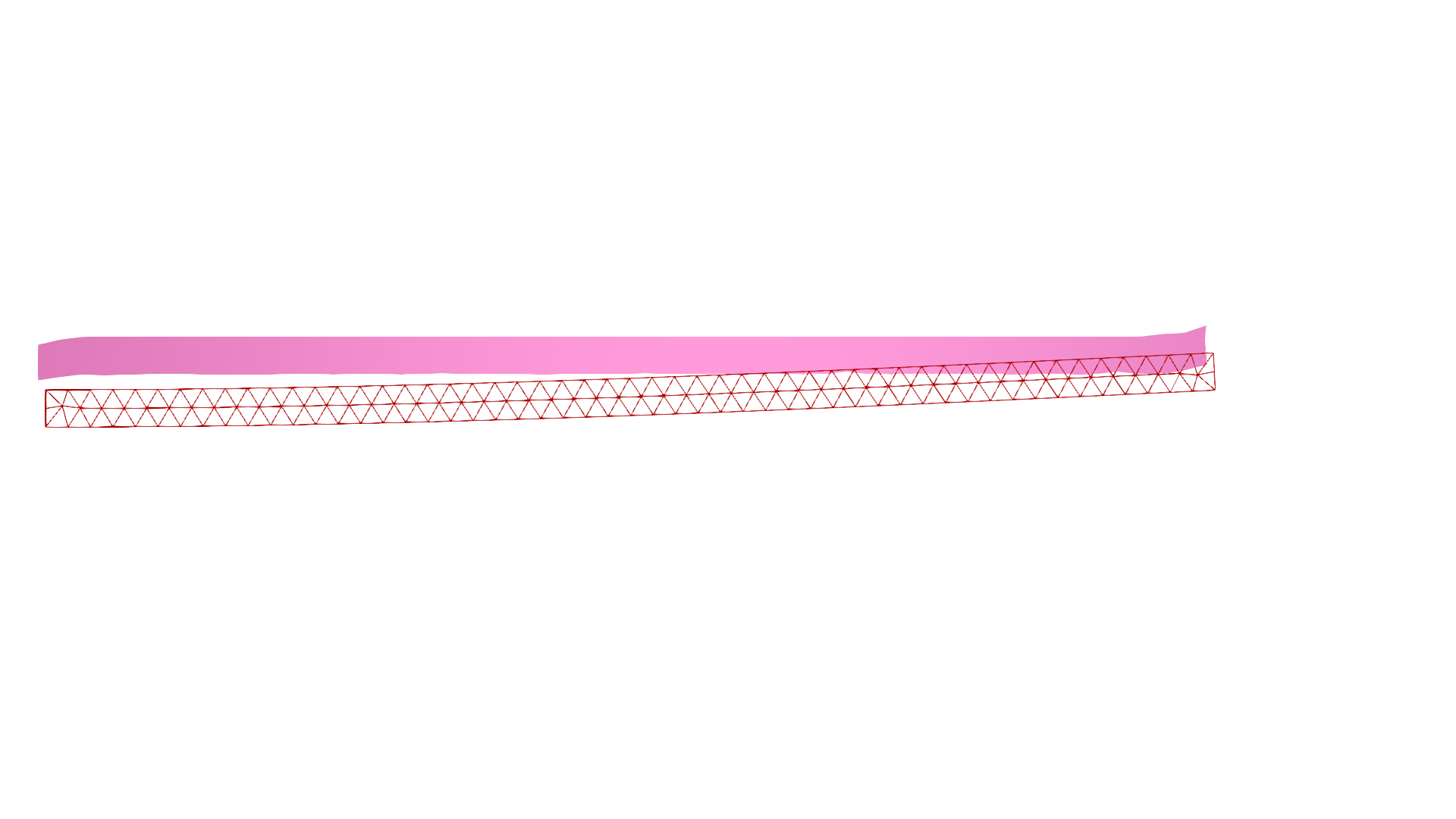}%
    \img{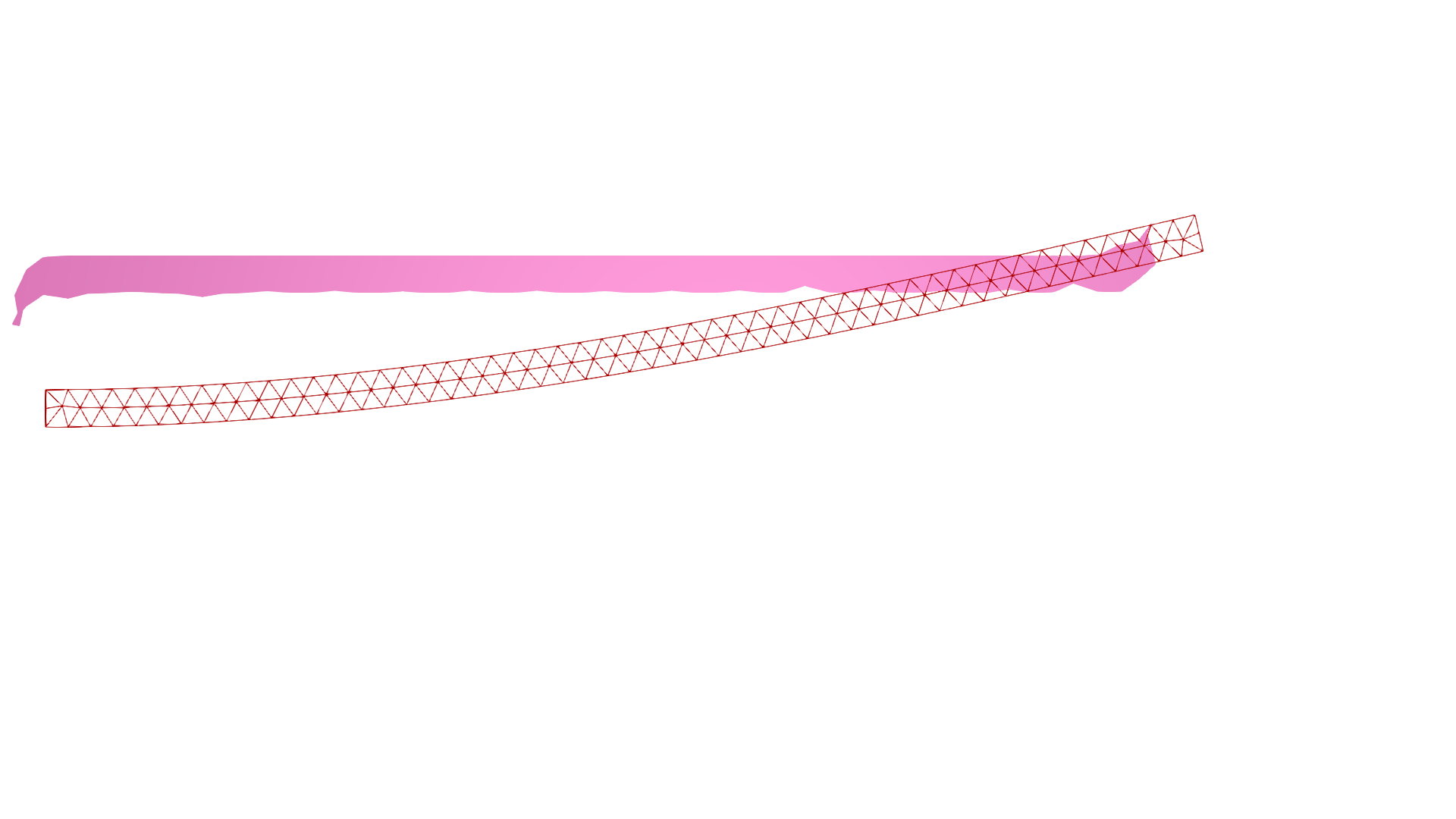}%
    \img{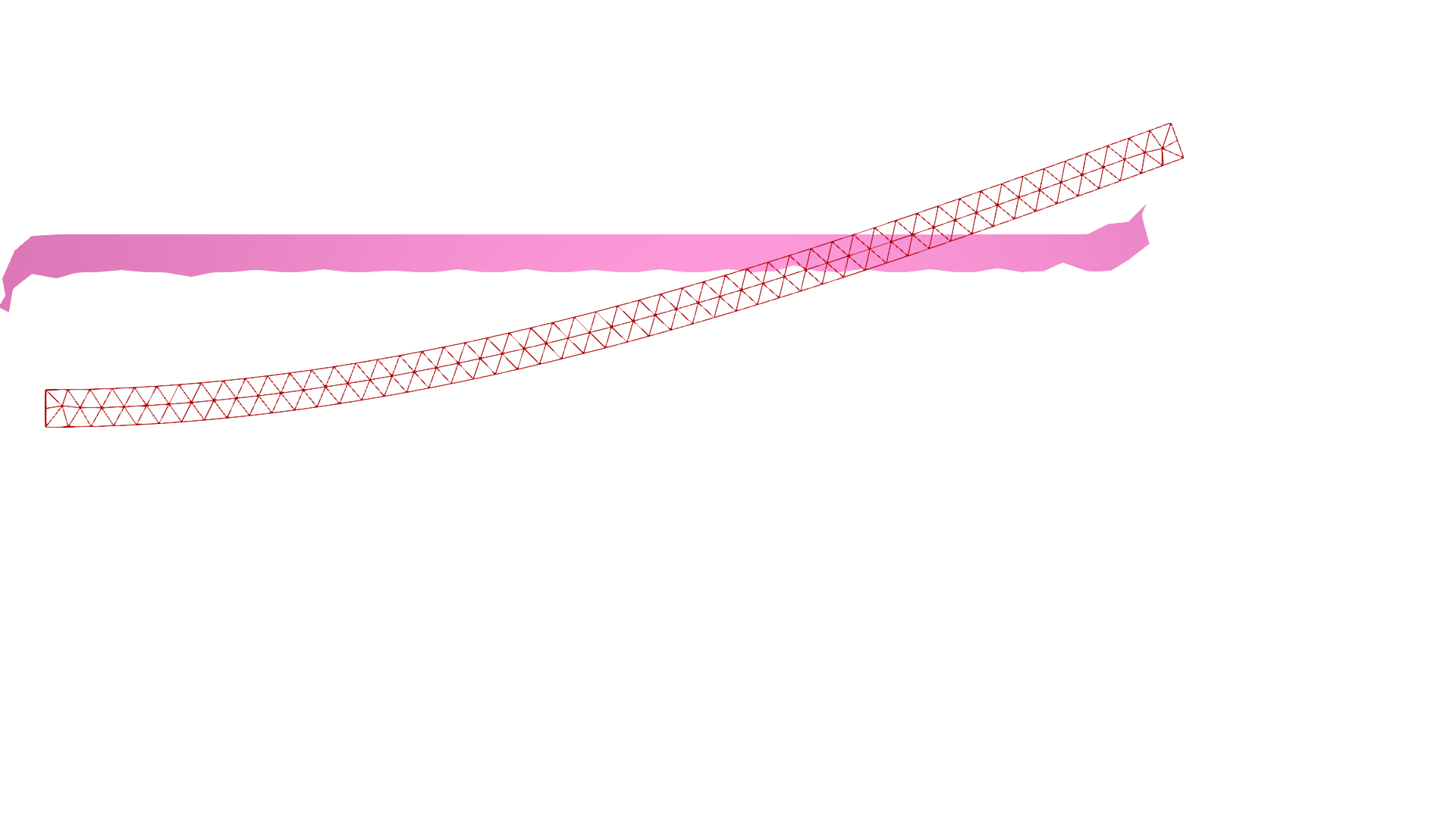}%
    \img{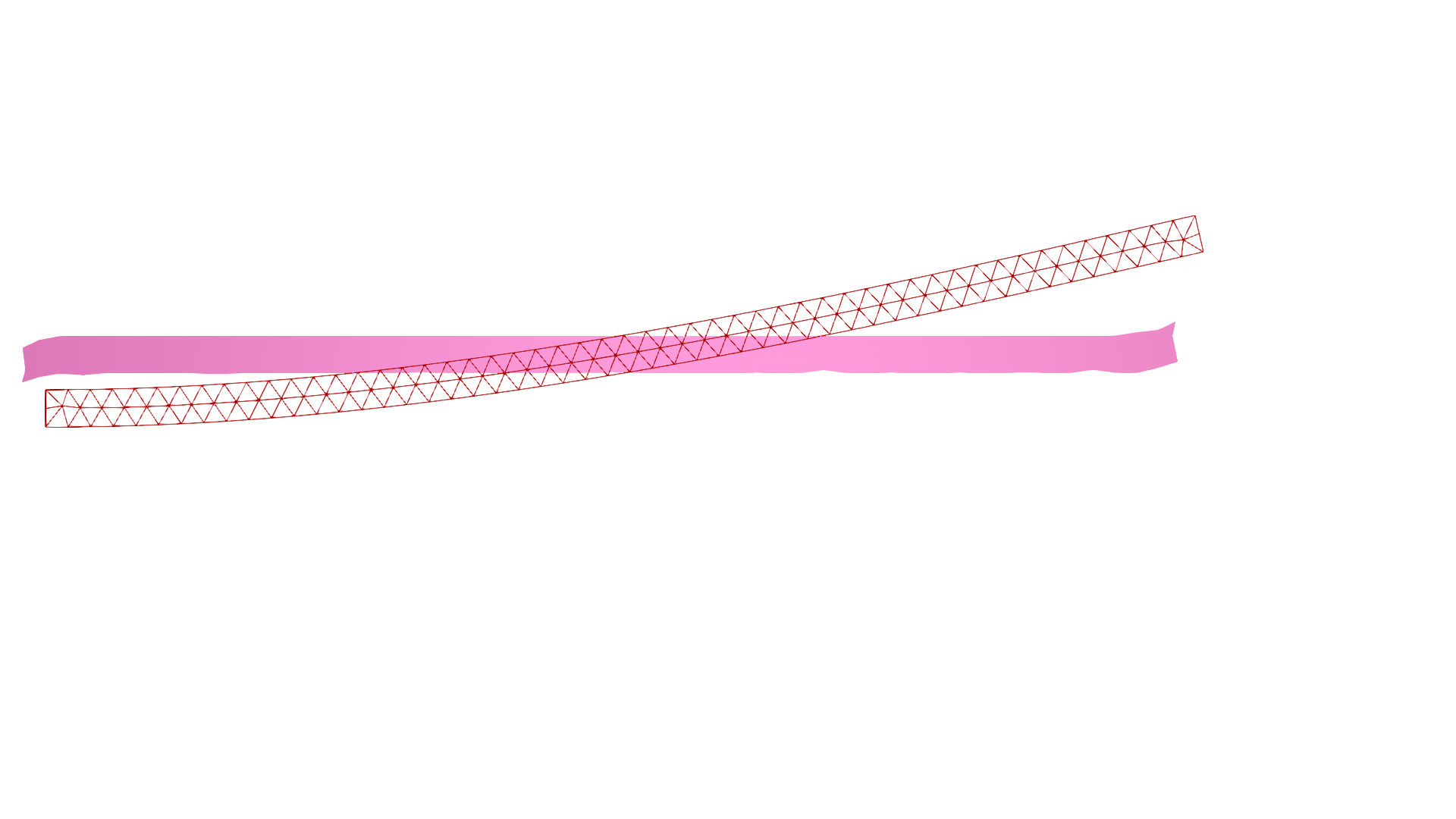}%
    \img{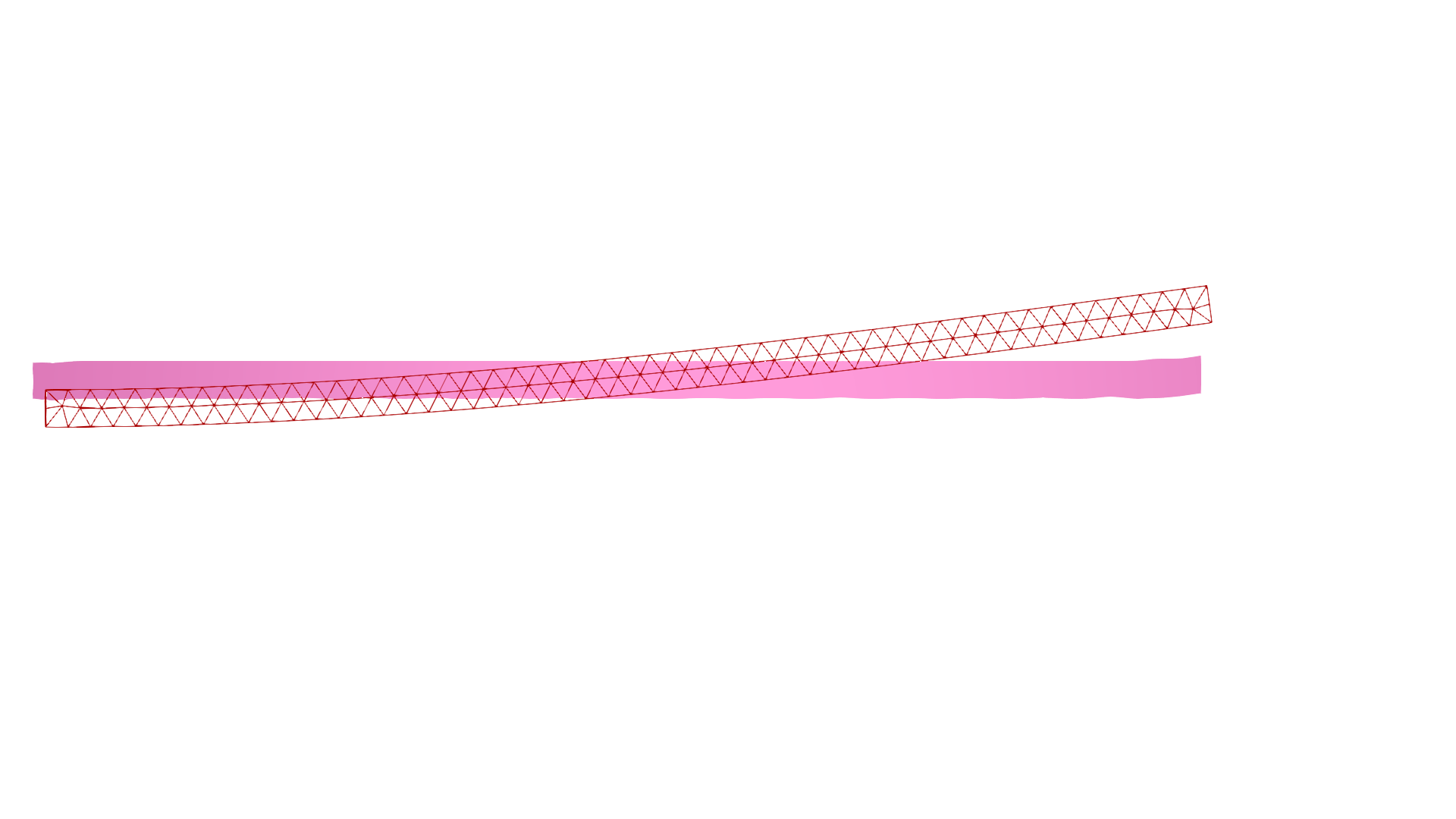}\\[0.6em]

    % Row 9: Pointcloud (special crop only for this row)
    \rowlabel{Pointcloud}%
    \begin{minipage}[c]{0.158\textwidth}\centering
        \pointcloud{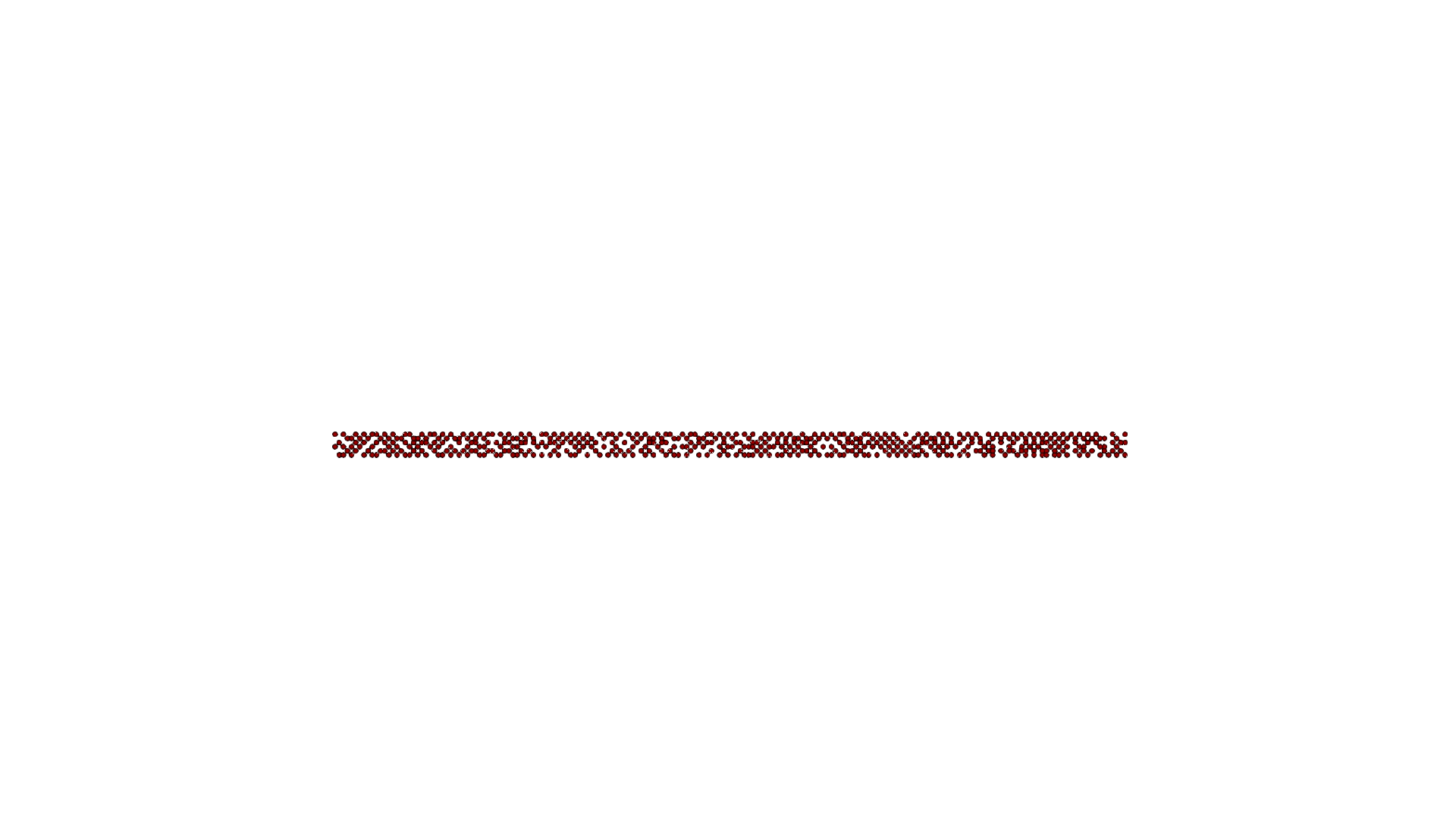}\\[2pt]
        \footnotesize$t{=}0$
    \end{minipage}%
    \begin{minipage}[c]{0.158\textwidth}\centering
        \pointcloud{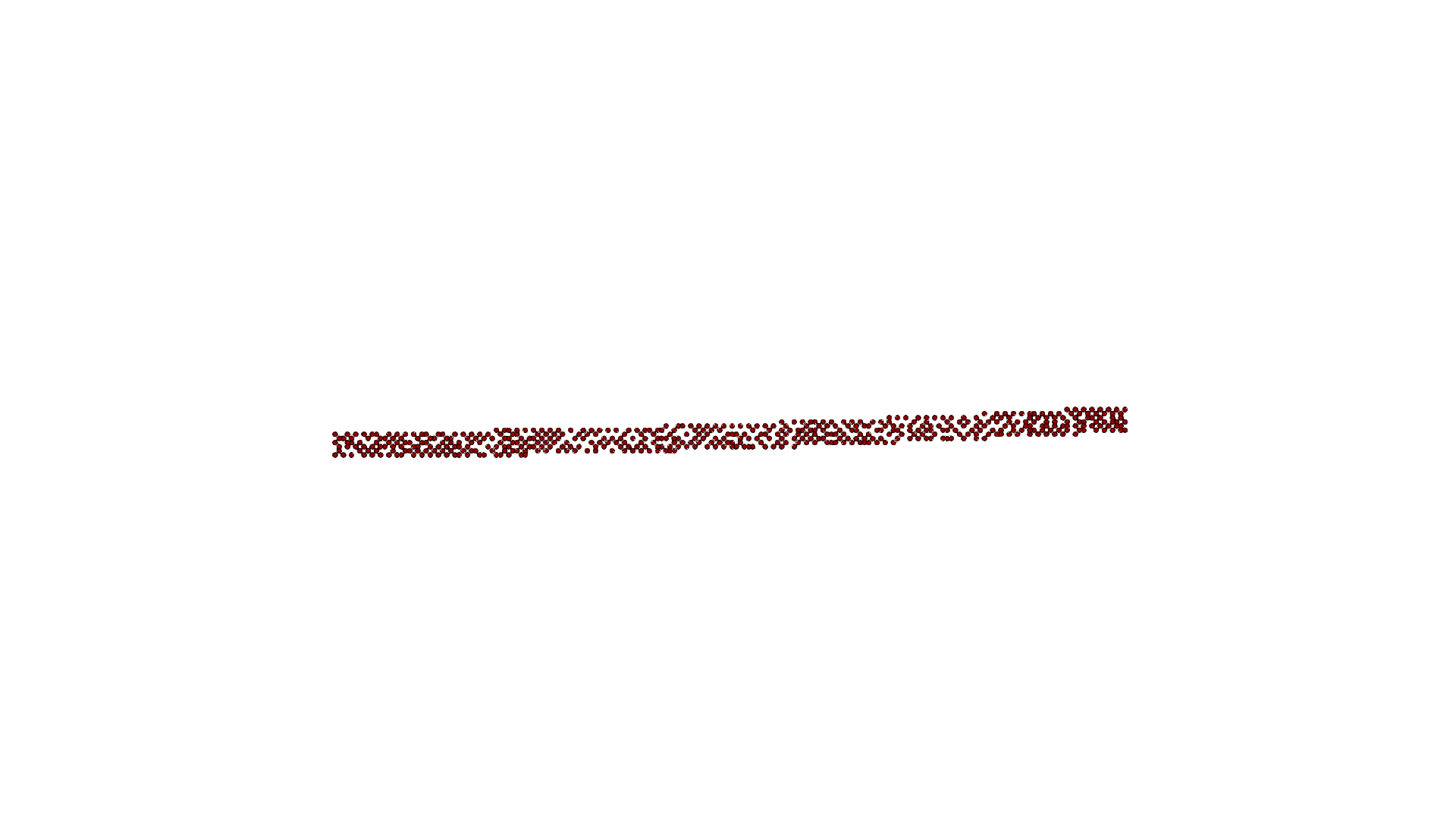}\\[2pt]
        \footnotesize$t{=}20$
    \end{minipage}%
    \begin{minipage}[c]{0.158\textwidth}\centering
        \pointcloud{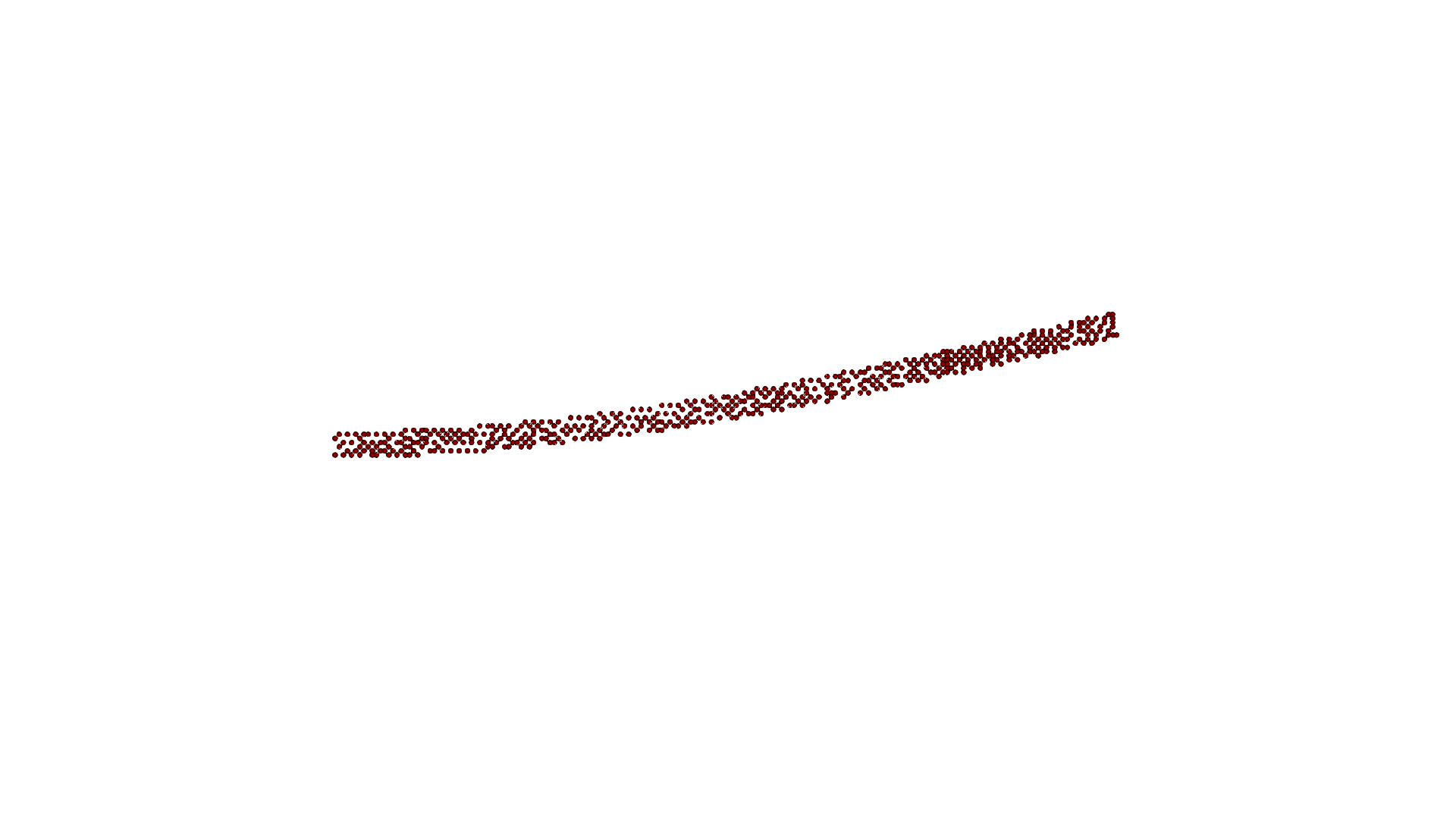}\\[2pt]
        \footnotesize$t{=}40$
    \end{minipage}%
    \begin{minipage}[c]{0.158\textwidth}\centering
        \pointcloud{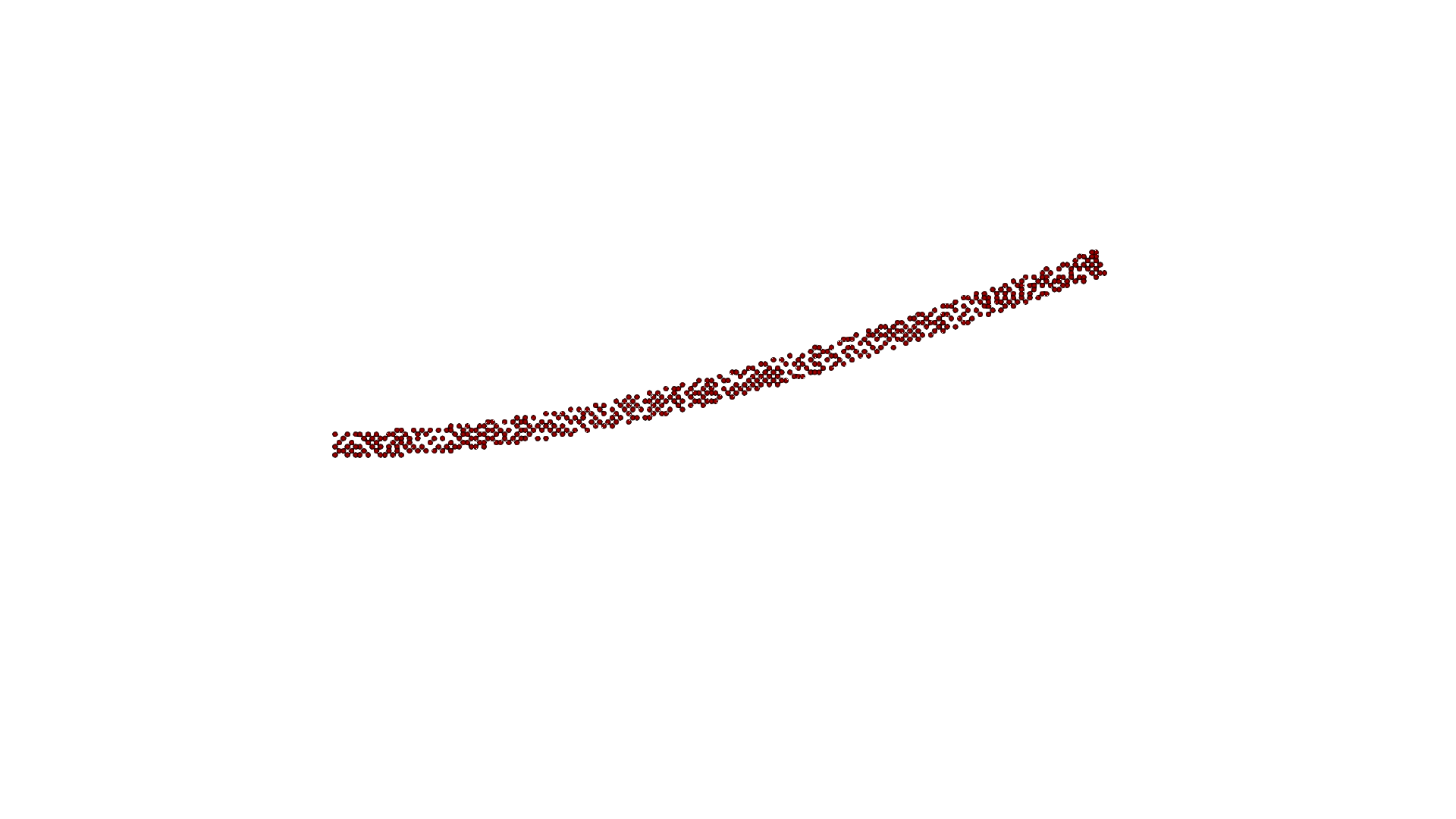}\\[2pt]
        \footnotesize$t{=}60$
    \end{minipage}%
    \begin{minipage}[c]{0.158\textwidth}\centering
        \pointcloud{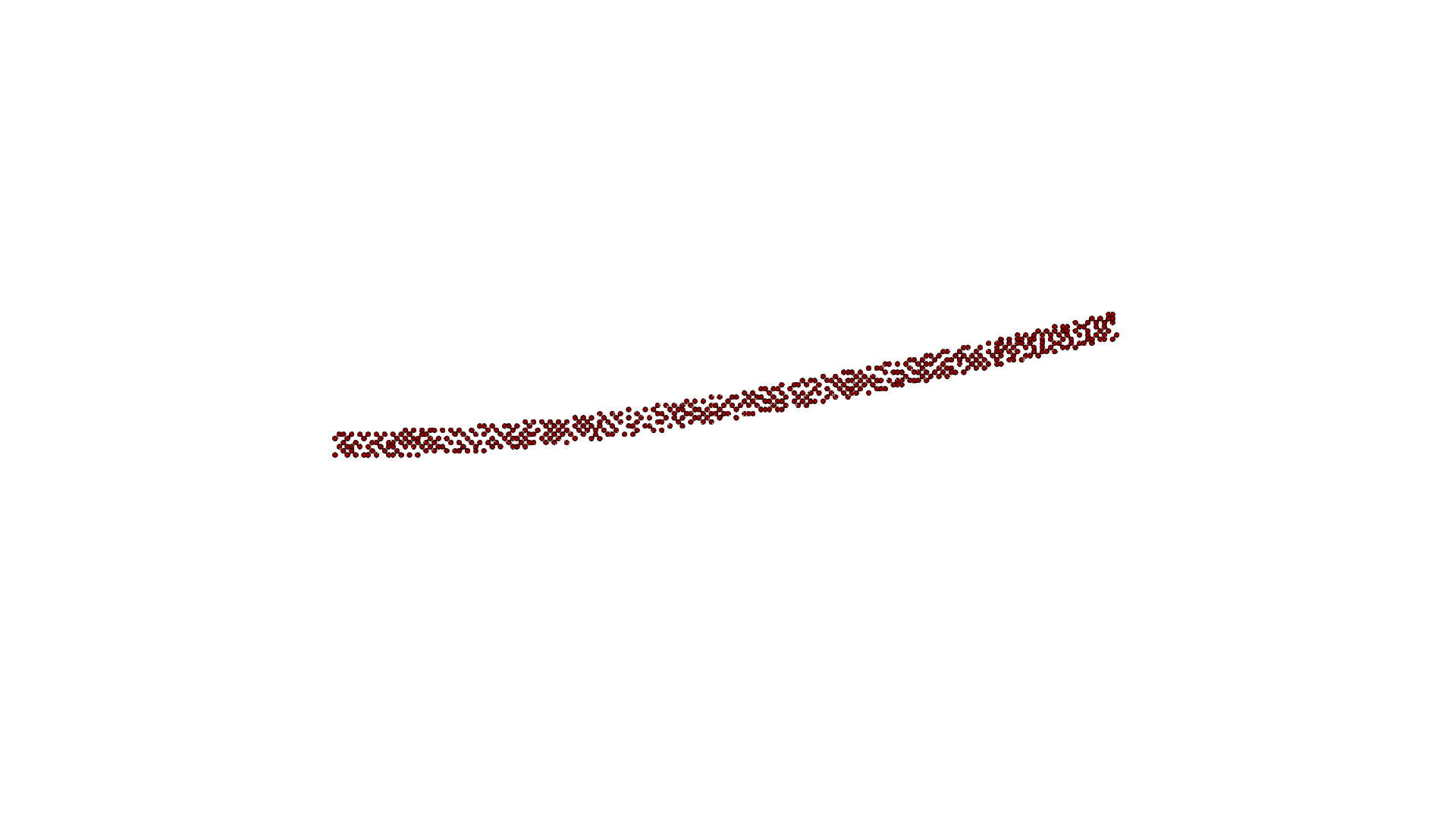}\\[2pt]
        \footnotesize$t{=}80$
    \end{minipage}%
    \begin{minipage}[c]{0.158\textwidth}\centering
        \pointcloud{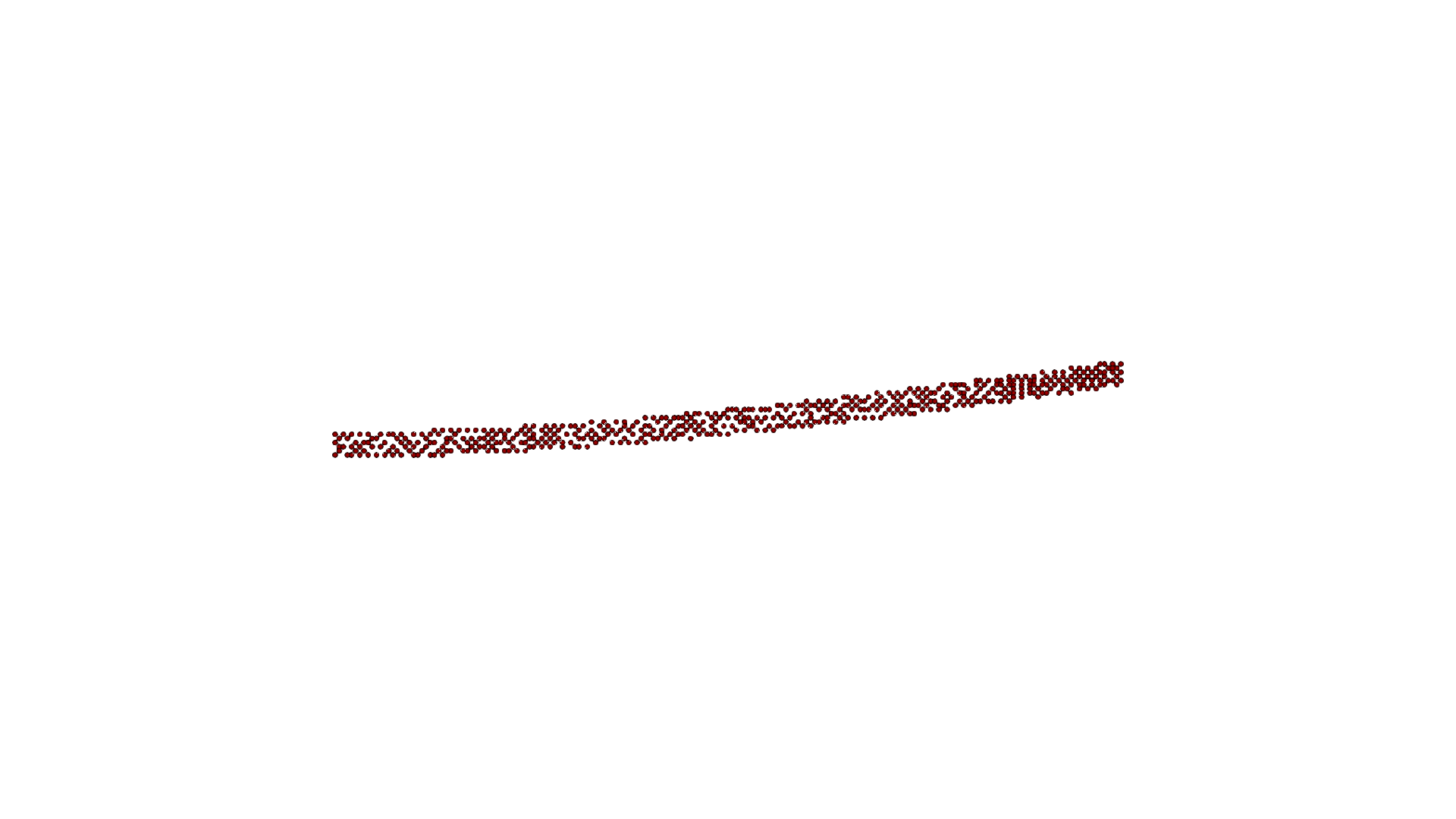}\\[2pt]
        \footnotesize$t{=}100$
    \end{minipage}

    \par\vfill
    \caption{
Predicted simulation of a \texttt{Bending Beam} test task by
\textcolor{TabBlue}{PEACH} (blue),
\textcolor{TabGray}{No Context} (gray),
\textcolor{TabCyan}{MANGO} (cyan),
\textcolor{TabGreen}{Oracle} (green),
\textcolor{TabOrange}{No Context (MGN)} (orange),
\textcolor{TabPurple}{Oracle (MGN)} (purple),
\textcolor{TabBrown}{GNN Encoder} (brown), and
\textcolor{TabPink}{PSTNet Encoder} (pink).
All visualizations show the colored \textbf{predicted mesh} and a \textbf{\textcolor{red}{wireframe}} (red) of the ground-truth simulation. The last row shows an exemplary point cloud sequence from the context set.
    }
    \label{fig:qualitative_trajectories_bbv}
\end{figure*}

We provide qualitative trajectory predictions for all evaluated datasets.
Figure~\ref{fig:qualitative_trajectories_db} shows results for the \texttt{Deforming Block} dataset,
Figure~\ref{fig:qualitative_trajectories_sd} for \texttt{Sheet Deformation},
Figure~\ref{fig:qualitative_trajectories_trampoline} for the \texttt{Trampoline} dataset, and
Figure~\ref{fig:qualitative_trajectories_bbv} for \texttt{Bending Beam}.

% =============================================================
%  DEFORMING BLOCK  –  Single figure (all methods)
% =============================================================

% =============================================================
%  SHEET DEFORMATION  –  Single figure (all methods)
% =============================================================

% =============================================================
%  TRAMPOLINE  –  Single figure (all methods)
% =============================================================

% =============================================================
%  BENDING BEAM  –  Single figure (all methods)
% =============================================================

\clearpage

\end{document}